\documentclass[twoside,11pt,dvipsnames, openleft]{book}

\usepackage{varioref}
\usepackage{jmlr2e}
\usepackage{bm}
\usepackage{dsfont}

\renewcommand{\arraystretch}{1.15}

\usepackage[english]{babel}

\usepackage{imakeidx}
\makeindex[columns=2, title=Alphabetical Index]

\usepackage{stackengine}

\usepackage{cancel}
\usepackage{dsfont}

\usepackage{extarrows}

\usepackage[framemethod=tikz]{mdframed}

\usepackage{amsmath}
\usepackage{arydshln}
\numberwithin{equation}{chapter}

\usepackage{amsmath}
\usepackage{blkarray}

\usepackage[absolute,overlay]{textpos}

\usepackage[final]{pdfpages}

\usepackage[noframe]{showframe}
\usepackage{framed}
\usepackage{lipsum}

\usepackage{xcolor}

\usepackage{tcolorbox}
\usepackage{graphicx}    
\usepackage[hang]{subfigure}

\usepackage{algorithm}
\usepackage{algpseudocode}
\usepackage{amsmath}
\usepackage{graphics}
\usepackage{epsfig}
\usepackage{blkarray}
\usepackage{listings}
\usepackage{booktabs}  

\usepackage{tikz}
\usetikzlibrary{mindmap,trees,backgrounds}
\usetikzlibrary{arrows.meta}
\usetikzlibrary{patterns}

\usetikzlibrary{decorations.text}

\usetikzlibrary{calc,backgrounds}

\newcommand*\circled[1]{\tikz[baseline=(char.base)]{
		\node[shape=circle,draw,inner sep=2pt] (char) {#1};}}

\usepackage{changepage}                 

\usepackage{hhline}

\usetikzlibrary{positioning,shapes,shadows,arrows}
\tikzstyle{condition}=[rectangle, draw=black, rounded corners, fill=colorqr, drop shadow,
text centered, anchor=north, text=black, text width=3cm]
\tikzstyle{abstract}=[rectangle, draw=black, rounded corners, fill=blue!30, drop shadow,
text centered, anchor=north, text=black, text width=3cm]
\tikzstyle{comment}=[rectangle, draw=black, rounded corners, fill=color1, drop shadow,
text centered, anchor=north, text=black, text width=3cm]
\tikzstyle{myarrow}=[->, >=open triangle 90, thick]
\tikzstyle{line}=[-, thick]

\usepackage{tikz-3dplot}
\tikzset{>=latex}
\usetikzlibrary{matrix}

\usepackage{sidecap}
\sidecaptionvpos{figure}{t}

\usepackage{verbatimbox}

\usepackage[mathscr]{euscript}

\usepackage{stmaryrd}

\newlength{\offsetpage}
	{\end{adjustwidth}}
	{\end{adjustwidth}}

\usepackage{minitoc}   
\let\cleardoublepage\clearpage
\usepackage[titletoc]{appendix}

\definecolor{titlepagecolor}{cmyk}{75,68,67,90}
\definecolor{titlepagecolor2}{rgb}{1.0, 0.08, 0.58}
\definecolor{emerald}{rgb}{0.31, 0.78, 0.47}
\definecolor{deeppink}{HTML}{D14064}
\definecolor{lowpink}{HTML}{ffe6ec}
\newcommand{\partcolor}{gray!65} 
\definecolor{lowblue}{HTML}{E1EBFE}

\usepackage{afterpage}

\makeatletter \renewcommand*\cleardoublepage{
	\clearpage
	\if@twoside   
	\ifodd\c@page 
	\hbox{}\newpage
	\if@twocolumn\hbox{}   
	\newpage
	\fi
	\fi
	\fi
} \makeatother
\let\originalpart=\part
\def\part#1{\cleardoublepage\clearpage \pagecolor{\partcolor} \originalpart{#1}\nopagecolor }

\usepackage{blkarray}

\renewcommand\thesection{\thechapter.\arabic{section}}

\newcommand{\longeq}{\mathrel{\rule[0.5ex]{2em}{0.5pt}\hspace{-2em}\rule[-0.5ex]{2em}{0.5pt}}}

\newtcolorbox{eqgraybox}{
	colback=gray!8,
	colframe=gray!8,
	boxrule=0pt,
	arc=0pt,
	top=0pt,
	bottom=0pt,
	left=4pt,
	right=4pt,
	boxsep=0pt,
	nobeforeafter,
	parbox=false,
	before=\par\noindent,
	after=\par
}

\newtcolorbox{mybox}{
	colback=\mdframecolor,
	colframe=\mdframecolor,
	boxrule=0pt,
	arc=0pt,
	top=8pt,        
	bottom=8pt,     
	left=1pt,       
	right=1pt,
boxsep=0pt,
before=\vspace{0pt},
after=\vspace{0pt},
parbox=false,
before=\par\noindent,
}
\newcommand{\bthetastar}{\bm{\theta}^*}

\newcommand{\diff}{\mathop{}\!\mathrm{d}}

\newcommand{\indicatorS}{\delta_{\mathbb{S}}}

\newcommand{\hadaprod}{\circ}

{\endMakeFramed}

\makeatletter
\renewcommand{\l@section}{\@dottedtocline{1}{1.5em}{2.2em}}
\renewcommand{\l@subsection}{\@dottedtocline{2}{4.0em}{3.2em}}
\renewcommand{\l@subsubsection}{\@dottedtocline{3}{7.1em}{4.3em}}
\makeatother

\usepackage[labelfont=bf]{caption}
\newcommand\myhrulefill[1]{\leavevmode\leaders\hrule height#1\hfill\kern0pt}
\DeclareCaptionFormat{myformat}{
	{\color[RGB]{0,0,0}\myhrulefill{0.12em}}
	\\#1#2#3
}
\usepackage{yfonts,color,lettrine}
\definecolor{caligraphcolor}{HTML}{74AECB}

\usepackage{setspace}

\AtBeginDocument{\setlength{\DefaultFindent}{0.5em}}
\usepackage[shortlabels]{enumitem}  
\usepackage{enumitem}

\newcommand*{\eitemi}{\tikz \draw [baseline, ball color=structurecolor,draw=none] circle (2pt);}
\newcommand*{\eitemii}{\tikz \draw [baseline, fill=structurecolor,draw=none,circular drop shadow] circle (2pt);}
\newcommand*{\eitemiii}{\tikz \draw [baseline, fill=structurecolor,draw=none] circle (2pt);}
\setlist[enumerate,1]{label=\color{black}\arabic*.,itemsep=0pt,partopsep=0pt,parsep=\parskip,topsep=5pt}
\setlist[enumerate,2]{label=\color{black}(\alph*).,itemsep=0pt,partopsep=0pt,parsep=\parskip,topsep=5pt}
\setlist[enumerate,3]{label=\color{black}\Roman*.,itemsep=0pt,partopsep=0pt,parsep=\parskip,topsep=5pt}
\setlist[enumerate,4]{label=\color{black}\Alph*.,itemsep=0pt,partopsep=0pt,parsep=\parskip,topsep=5pt}
\setlist[itemize,1]{label={\eitemi},itemsep=0pt,partopsep=0pt,parsep=\parskip,topsep=5pt}
\setlist[itemize,2]{label={\eitemii},itemsep=0pt,partopsep=0pt,parsep=\parskip,topsep=5pt}
\setlist[itemize,3]{label={\eitemiii},itemsep=0pt,partopsep=0pt,parsep=\parskip,topsep=5pt}

\usepackage{xcolor}
\usepackage[Conny]{fncychap}

\makeatletter  

\def\mghrulefill#1{\color{black}\leavevmode\leaders\hrule\@height #1\hfill\kern\z@}
\makeatother

\definecolor{chaptertitle}{RGB}{0,0,128}
\definecolor{chapternum}{RGB}{255,0,0}

\usepackage[colorlinks]{hyperref}
\usepackage{hyperref}
\hypersetup{
	colorlinks=true, 
	linktoc=all,     
	linkcolor=mydarkblue,  
	anchorcolor=blue,
	citecolor=mydarkgreen,
}

\usepackage{fancyhdr, blindtext}

\newcommand{\changefonts}{%
	\fontsize{9}{11}\selectfont
}

\fancypagestyle{plain}{%

	\fancyhf{}  
}

\usepackage{amsthm}
\newtheoremstyle{normalfontstyle} 
{3pt}                           
{3pt}                           
{\normalfont}                   
{}                              
{\bfseries}                     
{}                             
{ }                             
{}                              

\usepackage{thmtools}
\declaretheoremstyle[
spaceabove=3pt,
spacebelow=3pt,
headfont=\bfseries,
notefont=\bfseries, 
notebraces={(}{)}, 
bodyfont=\normalfont,
postheadspace=1em, 
]{normalfontboldhead}

\theoremstyle{normalfontstyle}
\newcommand{\BlackBox}{\rule{1.5ex}{1.5ex}}  
\renewenvironment{proof}{\par\noindent{\bf Proof\ }}{\hfill\BlackBox\\[2mm]}

\declaretheorem[style=normalfontboldhead, name=Definition, numberlike=theo]{definitionT}
\newmdenv[skipabove=7pt,
skipbelow=7pt,
rightline=false,
leftline=true,
topline=false,
bottomline=false,
linecolor=mydarkblue,
innerleftmargin=5pt,
innerrightmargin=5pt,
innertopmargin=0pt,
leftmargin=2cm,
rightmargin=0cm,
linewidth=4pt,
innerbottommargin=0pt]{dBox}
\newenvironment{definition}{\begin{dBox}\begin{definitionT}}{\end{definitionT}\end{dBox}}

\declaretheorem[style=normalfontboldhead, name=Exercise, numberlike=theo]{exerciseC}
\newmdenv[skipabove=7pt,
skipbelow=7pt,
rightline=false,
leftline=true,
topline=false,
bottomline=false,
linecolor=mydarkgreen,
innerleftmargin=5pt,
innerrightmargin=5pt,
innertopmargin=0pt,
leftmargin=2cm,
rightmargin=0cm,
linewidth=4pt,
innerbottommargin=0pt]{eBox}
\newenvironment{exercise}{\begin{eBox}\begin{exerciseC}}{\end{exerciseC}\end{eBox}}

\declaretheorem[style=normalfontboldhead, name=Note, numberlike=theo]{noteC}
\newmdenv[skipabove=7pt,
skipbelow=7pt,
rightline=false,
leftline=true,
topline=false,
bottomline=false,
linecolor=mydarkgreen,
innerleftmargin=5pt,
innerrightmargin=5pt,
innertopmargin=0pt,
leftmargin=2cm,
rightmargin=0cm,
linewidth=4pt,
innerbottommargin=0pt]{nBox}
\newenvironment{noteb}{\begin{nBox}\begin{noteC}}{\end{noteC}\end{nBox}}

\declaretheorem[style=normalfontboldhead, name=Remark, numberlike=theo]{remarekC}
\newmdenv[skipabove=7pt,
skipbelow=7pt,
rightline=false,
leftline=true,
topline=false,
bottomline=false,
linecolor=mydarkpurple,
innerleftmargin=5pt,
innerrightmargin=5pt,
innertopmargin=0pt,
leftmargin=2cm,
rightmargin=0cm,
linewidth=4pt,
innerbottommargin=0pt]{rBox}
\newenvironment{remark}{\begin{rBox}\begin{remarekC}}{\end{remarekC}\end{rBox}}

\declaretheorem[style=normalfontboldhead, name=Assumption, numberlike=theo]{assumptionC}
\newmdenv[skipabove=7pt,
skipbelow=7pt,
rightline=false,
leftline=true,
topline=false,
bottomline=false,
linecolor=mydarkpurple,
innerleftmargin=5pt,
innerrightmargin=5pt,
innertopmargin=0pt,
leftmargin=2cm,
rightmargin=0cm,
linewidth=4pt,
innerbottommargin=0pt]{asBox}

\declaretheorem[style=normalfontboldhead, name=Condition, numberlike=theo]{conditionC}
\newmdenv[skipabove=7pt,
skipbelow=7pt,
rightline=false,
leftline=true,
topline=false,
bottomline=false,
linecolor=mydarkpurple,
innerleftmargin=5pt,
innerrightmargin=5pt,
innertopmargin=0pt,
leftmargin=2cm,
rightmargin=0cm,
linewidth=4pt,
innerbottommargin=0pt]{cdBox}

\declaretheorem[style=normalfontboldhead, name=Example, numberlike=theo]{exampleC}
\newmdenv[skipabove=7pt,
skipbelow=7pt,
rightline=false,
leftline=false,
topline=false,
bottomline=false,
linecolor=mydarkgreen,
innerleftmargin=1pt,
innerrightmargin=5pt,
innertopmargin=0pt,
leftmargin=2cm,
rightmargin=0cm,
linewidth=4pt,
innerbottommargin=0pt]{xBox}
\newenvironment{example}{\begin{xBox}\begin{exampleC}}{\exampbar\end{exampleC}\end{xBox}}

\usepackage{adforn}  
\newcommand{\xchaptertitle}{Chapter~\thechapter~}
\newcommand{\problemname}{Problems}
\newenvironment{problemset}[1][\xchaptertitle~\problemname]{
	\vspace*{10pt}
	\begin{center}
		\phantomsection\addcontentsline{toc}{section}{\texorpdfstring{\xchaptertitle~\problemname}{\problemname}}
		\markright{#1}
		\textcolor{structurecolor}{\Large\bfseries\adftripleflourishleft~#1~\adftripleflourishright}
	\end{center}
	\begin{enumerate}[ref=\thechapter.\theenumi]}{
\end{enumerate}}

\usepackage{color}   

\definecolor{winestain}{rgb}{0.5,0,0}
\definecolor{colorGreenOcre}{RGB}{51,102,0} 
\definecolor{colorBlue2}{RGB}{200,207,248}
\definecolor{mydarkblue}{rgb}{0,0.08,0.45}

\newcommand{\mdframecolor}{gray!10}

\definecolor{mylightbluetitle}{RGB}{60,113,183}
\definecolor{mylightbluetext}{rgb}{0,0.08,0.45}
\definecolor{structurecolorblue}{RGB}{60,113,183}
\definecolor{structurecolorgreen}{RGB}{63,145,182}

\colorlet{structurecolor}{structurecolorblue}
\definecolor{structurecolorelegant}{RGB}{60,113,183}
\definecolor{structurecolorlt}{RGB}{31,119,185}

\definecolor{structurecolorHighTheoremBlue}{RGB}{220,227,248}
\definecolor{structurecolorHighTheoremGreen}{RGB}{188,222,231}
\colorlet{structurecolorHighTheorem}{structurecolorHighTheoremBlue}
\definecolor{mdframecolorRemark}{RGB}{186,94,103}

\definecolor{mydarkblue}{rgb}{0,0.08,0.45}
\definecolor{mydarkred}{rgb}{0.70,0.00,0.00}
\definecolor{mydarkgreen}{rgb}{0.00,0.30,0.00}
\definecolor{mydarkyellow}{RGB}{197,151,13}
\definecolor{mydarkpurple}{RGB}{90,35,140}
\definecolor{mydarkgray}{RGB}{64,64,64}

\definecolor{color0}  {RGB}{174,225,254} 
\definecolor{color1}  {RGB}{220,227,248} 
\definecolor{color2}  {RGB}{28,130,185} 
\definecolor{color3}  {RGB}{255,253,250} 
\definecolor{colormiddleright}  {RGB}{245,253,250} 
\definecolor{colorbottomleft}  {RGB}{255,243,250} 
\definecolor{coloruppermiddle}  {RGB}{255,253,230} 
\definecolor{colormiddleleft}  {RGB}{255,244,237}
\definecolor{colorcr}  {RGB}{249,253,232} 
\definecolor{colorreduction}  {RGB}{255,235,254} 
\definecolor{colorqr}  {RGB}{254,221,199} 
\definecolor{colorbiconjugate}  {RGB}{251,149,161} 
\definecolor{colorsvd}  {RGB}{215,247,235} 
\definecolor{colorupperright}  {RGB}{239,246,251} 
\definecolor{colorspectral}  {RGB}{206,226,243} 
\definecolor{colorbottomright}  {RGB}{220,224,236} 
\definecolor{coloreigenvalue}  {RGB}{197,203,224} 
\definecolor{colorcp} {RGB}{217, 234, 186} 
\definecolor{colorcpborder} {RGB}{233, 243, 216} 
\definecolor{colorupperleft}  {RGB}{235,243,240} 
\definecolor{colorsemidefinite}  {RGB}{217,232,226} 
\definecolor{colormiddle} {RGB}{235, 240,255}
\definecolor{colorlu}  {RGB}{220,227,255} 
\definecolor{colorals}  {RGB}{240,230,255} 
\definecolor{coloralsbkg}  {RGB}{248,243,255} 
\definecolor{canaryyellow}{rgb}{1.0, 0.75, 0.0}
\definecolor{bluepigment}{rgb}{0.0, 0.0, 1.0}
\definecolor{canarypurple}{RGB}{208, 13, 241}
\definecolor{colorGreenOcre}{RGB}{51,102,0} 
\definecolor{colorBlue1}  {RGB}{220,227,248}
\definecolor{colorBlue2}{RGB}{200,207,248}
\definecolor{shadecolor}{gray}{0.75}

\definecolor{color0}  {RGB}{174,225,254} 
\definecolor{color1}  {RGB}{220,227,248} 
\definecolor{color2}  {RGB}{28,130,185} 
\definecolor{color3}  {RGB}{255,253,250} 
\definecolor{color0}  {RGB}{174,225,254} 
\definecolor{color1}  {RGB}{220,227,248} 
\definecolor{color2}  {RGB}{28,130,185} 
\definecolor{color3}  {RGB}{255,253,250} 

\definecolor{colormiddleright}  {RGB}{245,253,250} 
\definecolor{colorbottomleft}  {RGB}{255,243,250} 
\definecolor{coloruppermiddle}  {RGB}{255,253,230} 
\definecolor{colormiddleleft}  {RGB}{255,244,237}

\definecolor{colorcr}  {RGB}{249,253,232} 
\definecolor{colorreduction}  {RGB}{255,235,254} 
\definecolor{colorqr}  {RGB}{254,221,199} 
\definecolor{colorbiconjugate}  {RGB}{251,149,161} 

\definecolor{colorsvd}  {RGB}{215,247,235} 

\definecolor{colorupperright}  {RGB}{239,246,251} 
\definecolor{colorspectral}  {RGB}{206,226,243} 

\definecolor{colorbottomright}  {RGB}{220,224,236} 
\definecolor{coloreigenvalue}  {RGB}{197,203,224} 

\definecolor{colorcp} {RGB}{217, 234, 186} 
\definecolor{colorcpborder} {RGB}{233, 243, 216} 

\definecolor{colorupperleft}  {RGB}{235,243,240} 
\definecolor{colorsemidefinite}  {RGB}{217,232,226} 

\definecolor{colormiddle} {RGB}{235, 240,255}
\definecolor{colorlu}  {RGB}{220,227,255} 

\definecolor{colorals}  {RGB}{240,230,255} 
\definecolor{coloralsbkg}  {RGB}{248,243,255} 

\definecolor{canaryyellow}{rgb}{1.0, 0.75, 0.0}
\definecolor{bluepigment}{rgb}{0.0, 0.0, 1.0}

\definecolor{canarypurple}{RGB}{208, 13, 241}
\definecolor{brightlavender}{rgb}{0.44, 0.16, 0.39}
\newcommand{\bpi}{\bm{\pi}}
\newcommand{\uniformdist}{\text{Unif}}

\newcommand{\normal}{\mathcal{N}}

\newcommand{\real}{\mathbb{R}}
\newcommand{\prob}{\Pr}

\mathchardef\mhyphen="2D

\newcommand{\gap}{\,\,\,\,\,\,\,\,}

\newcommand{\diag}{\mathrm{diag}}

\newcommand{\indicator}{\mathds{1}}

\newcommand{\bernoulli}{\mathrm{Bernoulli}}      
\newcommand{\bernoullidist}{\mathrm{Bernoulli}}

\newcommand{\smu}{\mu}
\newcommand{\ssigma}{\sigma}

\newcommand{\tr}{\mathrm{tr}}

\usepackage{amsmath}

\newcommand{\exampbar}{\hfill $\square$\par}

\newcommand{\cspace}{\mathcal{C}}
\newcommand{\nspace}{\mathcal{N}}

\newcommand{\Corr}{\mathbb{C}\mathrm{orr}}
\newcommand{\Cov}{\mathbb{C}\mathrm{ov}}

\newcommand{\Exp}{\mathbb{E}}

\newcommand{\Var}{\mathbb{V}\mathrm{ar}}

\newcommand{\argmax}{\operatorname*{\text{arg max}}}
\newcommand{\argmin}{\operatorname*{\text{arg min}}} 

\newcommand\abs[1]{\left\lvert#1\right\rvert}

\newcommand\norm[1]{\left\lVert#1\right\rVert}

\newcommand\normone[1]{\left\lVert#1\right\rVert_1}

\newcommand\normtwo[1]{\left\lVert#1\right\rVert_2}
\newcommand\normtwobig[1]{\big\lVert#1\big\rVert_2}
\newcommand\normp[1]{\left\lVert#1\right\rVert_p}

\newcommand\normf[1]{\left\lVert#1\right\rVert_F}

\newcommand\norminf[1]{\left\lVert#1\right\rVert_{\infty}}

\newcommand\innerproduct[1]{\left\langle#1\right\rangle}

\newcommand{\sgn}{\mathrm{sign}}

\newcommand{\rank}{\mathrm{rank}}

\newcommand{\trace}{\mathrm{tr}}

\mathchardef\mhyphen="2D
\newcommand{\complex}{\mathbb{C}}
\newcommand{\naturalset}{\mathbb{N}}

\def\1{\bm{1}}

\newcommand{\R}{\mathbb{R}}

\newcommand{\softmax}{\mathrm{softmax}}

\newcommand{\entropy}{\mathrm{H}}
\newcommand{\KL}{D_{\mathrm{KL}}}
\newcommand{\JS}{D_{\mathrm{JS}}}

\newcommand{\vf}{\bv}

\newcommand{\topone}{{(1)}}

\newcommand{\topzero}{{(0)}}
\newcommand{\toptminus}{{(t-1)}}

\newcommand{\toptzero}{{(t)}}

\newcommand{\toptone}{{(t+1)}}
\newcommand{\topT}{{(T)}}

\newcommand{\bzero}{\boldsymbol{0}}

\newcommand{\balpha}{{\boldsymbol\alpha}}
\newcommand{\bbeta}{{\boldsymbol\beta}}

\newcommand{\bdelta}{{\boldsymbol\delta}}

\newcommand{\bepsilon}{{\boldsymbol\epsilon}}

\newcommand{\bgamma}{{\boldsymbol\gamma}}

\newcommand{\blambda}{{\boldsymbol\lambda}}
\newcommand{\bmu}{{\boldsymbol\mu}}

\newcommand{\bomega}{{\boldsymbol\omega}}
\newcommand{\bphi}{{\boldsymbol\phi}}
\newcommand{\bpsi}{{\boldsymbol\psi}}

\newcommand{\bsigma}{{\boldsymbol\sigma}}

\newcommand{\btheta}{{\boldsymbol\theta}}

\newcommand{\bxi}{{\boldsymbol\xi}}
\newcommand{\bzeta}{{\boldsymbol\zeta}}

\newcommand{\bLambda}{{\boldsymbol\Lambda}}
\newcommand{\bOmega}{{\boldsymbol\Omega}}
\newcommand{\bPhi}{{\boldsymbol\Phi}}

\newcommand{\bPsi}{{\boldsymbol\Psi}}
\newcommand{\bSigma}{{\boldsymbol\Sigma}}
\newcommand{\bTheta}{{\boldsymbol\Theta}}

\newcommand{\widehatbepsilon}{{\widehat\bepsilon}}

\newcommand{\widehatbtheta}{{\widehat\btheta}}

\newcommand{\widebarblambda}{{\overline\blambda}}

\newcommand{\widebarbx}{\overline{\bm{x}}}

\newcommand{\widehatq}{\widehat{q}}

\newcommand{\widehatx}{\widehat{x}}
\newcommand{\widehaty}{\widehat{y}}

\newcommand{\widehatbx}{\widehat{\bm{x}}}
\newcommand{\widehatby}{\widehat{\bm{y}}}
\newcommand{\widehatbz}{\widehat{\bm{z}}}

\newcommand{\bmathcalD}{\bm{\mathcal{D}}}
\newcommand{\bmathcalE}{\bm{\mathcal{E}}}

\newcommand{\mathcalC}{\mathcal{C}}
\newcommand{\mathcalD}{\mathcal{D}}
\newcommand{\mathcalE}{\mathcal{E}}
\newcommand{\mathcalF}{\mathcal{F}}
\newcommand{\mathcalG}{\mathcal{G}}

\newcommand{\mathcalJ}{\mathcal{J}}

\newcommand{\mathcalL}{\mathcal{L}}
\newcommand{\mathcalM}{\mathcal{M}}

\newcommand{\mathcalO}{\mathcal{O}}

\newcommand{\mathcalQ}{\mathcal{Q}}

\newcommand{\mathcalV}{\mathcal{V}}

\newcommand{\mathcalX}{\mathcal{X}}

\newcommand{\mathcalZ}{\mathcal{Z}}

\newcommand{\widetildebA}{\widetilde{\bm{A}}}
\newcommand{\widetildebB}{\widetilde{\bm{B}}}
\newcommand{\widetildebC}{\widetilde{\bm{C}}}
\newcommand{\widetildebD}{\widetilde{\bm{D}}}

\newcommand{\widetildebW}{\widetilde{\bm{W}}}
\newcommand{\widetildebX}{\widetilde{\bm{X}}}

\newcommand{\widetildebc}{\widetilde{\bm{c}}}

\newcommand{\widetildebm}{\widetilde{\bm{m}}}

\newcommand{\widetildebt}{\widetilde{\bm{t}}}

\newcommand{\widetildebv}{\widetilde{\bm{v}}}

\newcommand{\widetildebx}{\widetilde{\bm{x}}}
\newcommand{\widetildeby}{\widetilde{\bm{y}}}
\newcommand{\widetildebz}{\widetilde{\bm{z}}}

\newcommand{\bone}{{\bm{1}}}
\newcommand{\ba}{{\bm{a}}}
\newcommand{\bA}{{\bm{A}}}
\newcommand{\bb}{{\bm{b}}}
\newcommand{\bB}{{\bm{B}}}
\newcommand{\bc}{{\bm{c}}}
\newcommand{\bC}{{\bm{C}}}
\newcommand{\bd}{{\bm{d}}}
\newcommand{\bD}{{\bm{D}}}
\newcommand{\be}{{\bm{e}}}

\newcommand{\bff}{{\bm{f}}}
\newcommand{\bF}{{\bm{F}}}
\newcommand{\bg}{{\bm{g}}}
\newcommand{\bG}{{\bm{G}}}
\newcommand{\bh}{{\bm{h}}}

\newcommand{\bi}{{\bm{i}}}
\newcommand{\bI}{{\bm{I}}}

\newcommand{\bJ}{{\bm{J}}}

\newcommand{\bK}{{\bm{K}}}

\newcommand{\bL}{{\bm{L}}}
\newcommand{\bmm}{{\bm{m}}}
\newcommand{\bM}{{\bm{M}}}

\newcommand{\bp}{{\bm{p}}}

\newcommand{\bq}{{\bm{q}}}
\newcommand{\bQ}{{\bm{Q}}}

\newcommand{\bs}{{\bm{s}}}
\newcommand{\bS}{{\bm{S}}}
\newcommand{\bt}{{\bm{t}}}

\newcommand{\bu}{{\bm{u}}}
\newcommand{\bU}{{\bm{U}}}
\newcommand{\bv}{{\bm{v}}}
\newcommand{\bV}{{\bm{V}}}
\newcommand{\bw}{{\bm{w}}}
\newcommand{\bW}{{\bm{W}}}
\newcommand{\bx}{{\bm{x}}}
\newcommand{\bX}{{\bm{X}}}
\newcommand{\by}{{\bm{y}}}
\newcommand{\bY}{{\bm{Y}}}
\newcommand{\bz}{{\bm{z}}}
\newcommand{\bZ}{{\bm{Z}}}

\def\vmu{{\bm{\mu}}}
\def\vtheta{{\bm{\theta}}}
\def\va{{\bm{a}}}

\def\ve{{\bm{e}}}

\def\vx{{\bm{x}}}

\def\mH{{\bm{H}}}

\def\mJ{{\bm{J}}}

\def\mX{{\bm{X}}}

\def\mSigma{{\bm{\Sigma}}}

\DeclareMathAlphabet{\mathsfit}{\encodingdefault}{\sfdefault}{m}{sl}
\SetMathAlphabet{\mathsfit}{bold}{\encodingdefault}{\sfdefault}{bx}{n}

\def\sA{{\mathbb{A}}}
\def\sB{{\mathbb{B}}}

\def\sD{{\mathbb{D}}}
\def\sE{{\mathbb{E}}}
\def\sF{{\mathbb{F}}}

\def\sS{{\mathbb{S}}}
\def\sT{{\mathbb{T}}}

\def\sX{{\mathbb{X}}}
\def\sY{{\mathbb{Y}}}
\def\sZ{{\mathbb{Z}}}

\def\ra{{\textnormal{a}}}
\def\rb{{\textnormal{b}}}
\def\rc{{\textnormal{c}}}

\def\rx{{\textnormal{x}}}
\def\ry{{\textnormal{y}}}

\def\rva{{\mathbf{a}}}
\def\rvb{{\mathbf{b}}}

\def\rve{{\mathbf{e}}}

\def\rvv{{\mathbf{v}}}

\def\rvx{{\mathbf{x}}}
\def\rvy{{\mathbf{y}}}
\def\rvz{{\mathbf{z}}}

\def\rmA{{\mathbf{A}}}
\def\rmB{{\mathbf{B}}}

\def\rmH{{\mathbf{H}}}

\def\rmY{{\mathbf{Y}}}

\firstpageno{1}

\newcommand{\mytitle}{Generative Models: Principles, Architectures, and Applications}
\begin{document}

\newpage
\thispagestyle{empty}  
\title{\mytitle}

\author{
\begin{center}
\name Jun Lu \\ 
\email jun.lu.locky@gmail.com
\end{center}
}

\frontmatter

\newpage 
\maketitle

\chapter*{\centering \begin{normalsize}Preface\end{normalsize}}

Generative AI has emerged as one of the most transformative forces in modern artificial intelligence, reshaping how we create, imagine, and interact with digital content. From photorealistic images to coherent text, from immersive videos to novel molecular structures, generative models now power applications that were once confined to science fiction. This book is designed to guide readers through the foundational principles, mathematical underpinnings, and practical architectures that underpin this revolution.

Our journey begins with  the two cornerstone families of latent variable generative models: Variational Autoencoders (VAEs) and Generative Adversarial Networks (GANs). Here, we unpack the Evidence Lower-Bound (ELBO), Expectation-Maximization (EM) algorithms, adversarial training, and their many variants, providing both theoretical rigor and practical insights into their strengths and limitations.

The core of the book is dedicated to Diffusion Models, the driving force behind today's state-of-the-art generative systems. We systematically cover both Diffusion Probabilistic Models (DPMs) and Denoising Diffusion Probabilistic Models (DDPMs), walking readers through the forward and reverse processes, the ELBO objective function, noise scheduling, latent diffusion, and score-based interpretations. We also extend this discussion to guidance mechanisms---including classifier and classifier-free guidance---and advanced sampling techniques that enable high-fidelity, controllable generation.

From there, we turn to Flow-Based Models, another powerful class of generative frameworks that offer exact likelihood computation and invertible transformations. We explore  neural differential equations, and score matching, drawing connections to diffusion models to highlight complementary strengths. Finally, we examine the practical architectures that power modern image and video synthesis: U-Nets, ControlNet, Diffusion Transformers (DiTs), and multimodal variants, alongside case studies of real-world systems like Stable Diffusion 3.

No book of this nature is a solitary endeavor. We are deeply grateful to the countless researchers who have paved the way for generative AI, whose pioneering work forms the backbone of this text.  It is our sincere hope that this book will serve as both an educational resource and a source of inspiration as you push the boundaries of what generative models can achieve.


\newpage
\begingroup
\hypersetup{
linkcolor=structurecolor,
linktoc=page,  
}
\dominitoc
\pdfbookmark{\contentsname}{toc} 
\tableofcontents 

\endgroup

%

\chapter*{Notation}\label{notation}


This section provides a concise reference describing notation used throughout this
book.
If you are unfamiliar with any of the corresponding mathematical concepts,
the book describes most of these ideas in Chapter~\ref{chapter_genintroduction} (p.~\pageref{chapter_genintroduction}).

\vspace{0.4in}
\begin{minipage}{\textwidth}
\centerline{\bf Numbers and Arrays}
\bgroup
\def\arraystretch{1.5}
\begin{tabular}{cp{4.25in}}
$\displaystyle a$   & A scalar (integer or real, italics font)\\
$\displaystyle \ba$ & A vector (italics font)\\
$\displaystyle \bA$ & A matrix (italics font)\\
$\displaystyle \bI_D$ & Identity matrix with $D$ rows and $D$ columns\\
$\displaystyle \bI$   & Identity matrix with dimensionality implied by context\\
$\displaystyle \ve_i$ & Standard basis vector $[0,\dots,0,1,0,\dots,0]$ with a 1 at position $i$\\
$\displaystyle \text{diag}(\va)$ & A square, diagonal matrix with diagonal entries given by $\va$\\
$\displaystyle \ra$   & A scalar random variable  (normal fonts)\\
$\displaystyle \rva$  & A vector-valued random variable  (normal fonts)\\
$\displaystyle \rmA$  & A matrix-valued random variable (normal fonts)\\
\end{tabular}
\egroup
\index{Scalar}
\index{Vector}
\index{Matrix}
\end{minipage}

\index{Sets}
\vspace{0.2in}
\begin{minipage}{\textwidth}
\centerline{\bf Sets}
\bgroup
\def\arraystretch{1.5}
\begin{tabular}{cp{4.25in}}
$\displaystyle \sA$ & A set\\
$\displaystyle \varnothing$ & The null set \\
$\displaystyle \real, \complex, \sF\equiv \{\real \text{ or }\complex\}$ & The set of real, complex, either real or complex numbers\\
$\displaystyle \naturalset, \naturalset_0$ & The set of natural numbers $\naturalset=\{1,2,\ldots\}$, $\naturalset_0=\{0\}\cup\naturalset$  \\
$\displaystyle \{0, 1\}$ & The set containing 0 and 1 \\
$\displaystyle \{0, 1, \dots, n \}$ & The set of all integers between $0$ and $n$\\
$\displaystyle [a, b]$ & The real interval including $a$ and $b$\\
$\displaystyle (a, b]$ & The real interval excluding $a$ but including $b$\\
$\displaystyle \sA \backslash \sB$ & Set subtraction, i.e., the set containing the elements of $\sA$ that are not in $\sB$\\
\end{tabular}
\egroup
\index{Scalar}
\index{Vector}
\index{Matrix}
\index{Set}
\end{minipage}

\vspace{0.2in}
\begin{minipage}{\textwidth}
\centerline{\bf Indexing}
\bgroup
\def\arraystretch{1.5}
\begin{tabular}{cp{4.25in}}
$\displaystyle x_i$ & Element $i$ of vector $\bx$, with indexing starting at 1 \\
$\displaystyle \bx_{-i}$ & All elements of vector $\bx$ except for element $i$ \\
$\displaystyle  x_{ij}$ & Element $i, j$ of matrix $\bX$ \\
$\displaystyle \bX_{i, :}=\bX[i,:],\, \bx^{(i)}$ & Row $i$ of matrix $\bX$ \\
$\displaystyle \bX_{:, i}=\bX[:, i],\, \bx_i$ & Column $i$ of matrix $\bX$ \\
$\displaystyle \bX_\sS=\bX[:, \sS], \bX(\sS)$ & $\bX_\sS \in\real^{n\times \abs{\sS}}$ and $\bX(\sS)\in\real^{n\times p}$ if $\bX\in\real^{n\times p}$ \\
\end{tabular}
\egroup
\end{minipage}

\vspace{0.2in}
\begin{minipage}{\textwidth}
\centerline{\bf Linear Algebra Operations}
\bgroup
\def\arraystretch{1.5}
\begin{tabular}{cp{4.25in}}
$\displaystyle \bX^\top$ & Transpose of matrix $\bX$ \\
$\displaystyle \bX \hadaprod \bY $ & Element-wise (Hadamard) product of $\bX$ and $\bY$ \\
$\displaystyle \mathrm{det}(\bX)$ & Determinant of $\bX$ \\
$\displaystyle \cspace(\bX), \nspace(\bX), \mathcalV$ & Column, null  space of $\bX$, and a general space \\
$\displaystyle \rank(\bX)$ & Rank of $\bX$ \\
$\displaystyle \trace(\bX)$ & Trace of $\bX$ \\
$\displaystyle \lambda_{\max}(\bX), \lambda_{\min}(\bX)$ & Largest and smallest eigenvalue of $\bX$ \\
$\displaystyle \sigma_{\max}(\bX), \sigma_{\min}(\bX)$ & Largest and smallest singular value of $\bX$ \\
$\displaystyle  \diag(\bbeta)$ & Diagonal matrix with entries of $\bbeta$
\end{tabular}
\egroup
\index{Transpose}
\index{Element-wise product, Hadamard product}
\index{Hadamard product}
\end{minipage}

\vspace{0.4in}
\begin{minipage}{\textwidth}
\centerline{\bf Calculus}
\bgroup
\def\arraystretch{1.5}
\begin{tabular}{cp{4.25in}}
$\displaystyle\frac{\diff y} {\diff x}$ & Derivative of $y$ with respect to $x$\\ [2ex]
$\displaystyle \frac{\partial y} {\partial x} $ & Partial derivative of $y$ with respect to $x$ \\
$\displaystyle \nabla_{\bx} y $ & Gradient of $y$ with respect to $\bx$ \\
$\displaystyle \nabla_{\bX} y $ & Matrix derivatives of $y$ with respect to $\bX$ \\
$\displaystyle \frac{\partial f}{\partial \vx} $ & Jacobian matrix $\mJ \in \R^{m\times n}$ of $f: \R^n \rightarrow \R^m$\\
$\displaystyle \nabla_\vx^2 f(\vx)\text{ or }\mH( f)(\vx)$ & The Hessian matrix of $f$ at input point $\vx$\\
$\displaystyle \int f(\vx) d\vx $ & Definite integral over the entire domain of $\vx$ \\
$\displaystyle \int_\sS f(\vx) d\vx$ & Definite integral with respect to $\vx$ over the set $\sS$ \\
\end{tabular}
\egroup
\index{Integral}
\index{Hessian matrix}
\end{minipage}

\vspace{0.4in}
\begin{minipage}{\textwidth}
\centerline{\bf Probability and Information Theory}
\bgroup
\def\arraystretch{1.5}
\begin{tabular}{cp{4.25in}}
$\displaystyle \ra \bot \rb$ & The random variables $\ra$ and $\rb$ are independent\\
$\displaystyle \ra \bot \rb \mid \rc $ & They are conditionally independent given $\rc$\\
$\displaystyle \Pr(\bx)$ & A probability distribution over a discrete variable\\
$\displaystyle p(\bx), p_{\rvx}(\bx), f(\bx), f_{\rvx}(\bx)$ & A probability distribution over a continuous variable, or over
a variable whose type has not been specified\\
$\displaystyle \ra \sim P$ & Random variable $\ra$ has distribution $P$\\
$\displaystyle  \Exp_{\rx\sim P} [ g(x) ]\text{ or } \Exp [g(x)]$ & Expectation of $g(x)$ with respect to $P(\rx)$ \\
$\displaystyle \Var[g(x)] $ &  Variance of $g(x)$ under $P(\rx)$ \\
$\displaystyle \Cov[g(x),h(x)] $ & Covariance of $g(x)$ and $h(x)$ under $P(\rx)$\\
$\displaystyle \Corr[g(x),h(x)] $ & Correlation of $g(x)$ and $h(x)$ under $P(\rx)$\\
$\displaystyle \normal ( \mu , \sigma^2)$ &  Gaussian distribution %
 with mean $\mu$ and variance $\sigma^2$ \\
$\displaystyle \normal ( \vmu , \mSigma)$ & Multivariate Gaussian distribution %
 with mean $\vmu$ and covariance $\mSigma$ \\
$\displaystyle \bernoullidist(\pi) $ & Bernoulli distribution with mean $\pi$ \\
$\delta_{\by}\triangleq \delta(\bx-\by)$ & Dirac delta distribution\\
$p_{\text{data}}(\bx)$ & Data distribution\\
\end{tabular}
\egroup
\index{Independence}
\index{Conditional independence}
\index{Variance}
\index{Covariance}
\end{minipage}

\vspace{0.4in}
\begin{minipage}{\textwidth}
\centerline{\bf Functions}
\bgroup
\def\arraystretch{1.5}
\begin{tabular}{cp{4.25in}}
$\displaystyle g: \sA \rightarrow \sB$ & The function $g$ with domain $\sA$ and range $\sB$\\
$\displaystyle g \circ h $ & Composition of the functions $g$ and $h$ \\
  $\displaystyle g(\vx ; \vtheta) $ & A function of $\vx$ parametrized by $\vtheta$.
  (Sometimes we write $g(\vx)$ and omit the argument $\vtheta$ to lighten notation) \\
$\displaystyle \ln(x),\log(x)$ & Natural logarithm of $x$ \\
$\displaystyle \sgn(x)$ & Sign $x$, taking a value among $+1$, $-1$, and 0 \\
$\displaystyle \normp{\bx} $ & $\ell_p$-norm of $\vx$ \\
$\displaystyle \norm{\bx}=\normtwo{\bx} $ & $\ell_2$-norm of $\vx$ \\
$\displaystyle \norm{\bx}=\normone{\bx} $ & $\ell_1$-norm of $\vx$ \\
$\displaystyle \norm{\bx}=\norminf{\bx} $ & $\ell_\infty$ norm of $\vx$ \\
$\displaystyle \indicator\{\mathrm{condition}\}$ & is 1 if the condition is true, 0 otherwise\\
$\displaystyle \indicatorS(\balpha)$ & is 0 if $\balpha\in\sS$, $+\infty$ otherwise\\
$\displaystyle \softmax(\bbeta)$ & Softmax function\\
$\displaystyle \mathcalJ(\cdot)$ & Loss functions\\
$\displaystyle \mathcalF(\cdot)$ & Evidence lower-bound functions\\
$\displaystyle \mathcalL(\cdot)$ & Log-likelihood functions\\
$\displaystyle \ell(\cdot)$ & Likelihood functions\\
$\displaystyle L(\cdot)$ & Lagrangian function\\
\end{tabular}
\egroup
\index{Norm}
\end{minipage}
Sometimes we use a function $g$ whose argument is a scalar but apply
it to a vector, matrix: $g(\vx)$, $g(\mX)$.
This denotes the application of $g$ to the
array elementwise. For example, if $\bC = \sigma(\bX)$, then $c_{ij} = \sigma(x_{ij})$
for all valid values of $i$ and  $j$.

\vspace{0.4in}
\begin{minipage}{\textwidth}
\centerline{\bf Other General Notastions}
\bgroup
\def\arraystretch{1.5}
\begin{tabular}{cp{4.25in}}
$\displaystyle \triangleq$ & Equals by definition\\
$\displaystyle :=, \leftarrow $ & Equals by assignment \\
$\displaystyle \equiv $    & Equals by equivalence \\
$\displaystyle \pi $       & A probability value or 3.141592....\\
$\displaystyle e, \exp $   & 2.71828...
\end{tabular}
\egroup
\end{minipage}

\vspace{0.4in}
\begin{minipage}{\textwidth}
\centerline{\bf Abbreviations}
\bgroup
\def\arraystretch{1.5}
\begin{tabular}{cp{4.25in}}
AR & Autoregressive\\
CDF & Cumulative distribution function\\
PMF & Probability mass function\\
PDF & Probability distribution function\\
i.i.d. & Independently and identically distributed \\
PDF & Probability density function \\
PMF & Probability mass function \\
MVT & Mean value theorem\\
GD & Gradient descent\\
SGD & Stochastic gradient descent \\
MSE & Mean squared error\\
SVD & Singular value decomposition\\
ELBO & Evidence lower-bound \\
KL divergence & Kullback--Leibler divergence \\
AE & Autoencoder\\
VAE & Variational autoencoder model\\
GAN & Generative adversarial network\\
DPM & Diffusion probabilistic model\\
DDPM & Denoising diffusion probabilistic model\\
DDIM & Denoising diffusion implicit model\\
LDM & Latent diffusion model\\
MAF & Masked autoregressive flow\\
IAF & Inverse autoregressive flow\\
ODE & Ordinary differential equation\\
SDE & Stochastic differential equation\\

\end{tabular}
\egroup
\end{minipage}

\vspace{0.4in}
\begin{minipage}{\textwidth}
\centerline{\bf Abbreviations}
\bgroup
\def\arraystretch{1.5}
\begin{tabular}{cp{4.25in}}
FM & Flow matching\\
RK & Runge--Kutta \\
RK2 & The second-order Runge--Kutta method\\
RK4 & The fourth-order Runge--Kutta method\\
FD & Fisher divergence \\
ESM & Energy score matching\\
DSM & Denoising score matching\\
CTMP & Continuous-time Markov process \\
CFM &  Conditional flow matching\\
VP & Variance preserving\\
VE & Variance exploding\\
SNR & signal-to-noise\\
EI & Exponential integrator\\
DEIS & Diffusion exponential integrator sampler\\
NFE & Number of function evaluations\\
LMS & Linear multistep approximation \\
DiT & Diffusion transformer\\

\end{tabular}
\egroup
\end{minipage}

\clearpage

\mainmatter

\newpage 
\chapter{Introduction}\label{chapter_genintroduction}
\begingroup
\hypersetup{
linkcolor=structurecolor,
linktoc=page,  
}
\minitoc \newpage
\endgroup
\section{Introduction and Background}
\lettrine{\color{caligraphcolor}I}
In just a decade, generative models have evolved from niche machine learning curiosities into a cornerstone of modern artificial intelligence, reshaping how we create, interact with, and interpret digital content. From photorealistic images and immersive videos to coherent text and synthetic data, these models have demonstrated an unprecedented ability to learn the underlying patterns of complex data distributions and generate new, plausible samples that reflect the richness of the real world. This book guides readers through the theoretical foundations, architectural innovations, and practical advances that define this transformative field, providing a structured progression from core principles to state-of-the-art systems.

\paragrapharrow{Vector representations of images, videos, audio waveforms, molecular structures.}
In this book, we focus primarily on the task of generating objects represented as vectors $\bx\in\real^D$, including images, videos, and molecular structures.
In computational modeling, complex data types such as images, videos, audio waveforms, and molecular structures are commonly encoded as high-dimensional vectors to enable mathematical analysis and machine learning processing:
\begin{itemize}
\item \textbf{Images:} An image is represented as a tensor $\bx \in \real^{H \times W \times 3}$~\footnote{Although this is a tensor, we still treat it as a vector $\bx$ within the generative modeling framework.}, where $H$ denotes  height, $W$ denotes width, and the third dimension corresponds to the three RGB color channels (Red, Green, Blue). This formulation captures the spatial and chromatic information of the image in a structured numerical format.

\item \textbf{Videos:}  A video is modeled as a temporal sequence of image frames, represented by the tensor $\bx \in \real^{T \times H \times W \times 3}$, where $T$ is the number of frames. Each frame is treated as an individual image, and concatenation along the time axis preserves the dynamic evolution of the visual scene.

\item \textbf{Audio waveforms:} Raw audio signals are continuous one-dimensional time-series data discretized for computational use. An audio waveform is represented as a matrix $\bx \in \real^{S \times C}$, where $S$ denotes the total number of discrete sampling points and $C$ denotes the number of audio channels (e.g., $C=1$ for mono audio, $C=2$ for stereo audio). This structured numerical representation encodes temporal acoustic variations, including amplitude, frequency, and sound duration, fully characterizing the auditory features of the original audio signal.

\item \textbf{Molecular structures:} A molecular structure, such as a protein, is represented by the coordinates of its constituent atoms. This is formulated as a matrix $\bx \in \real^{N \times 3}$, where $N$ is the number of atoms, and each row specifies the 3D spatial coordinates $(x, y, z)$ of a given atom.
\end{itemize}
In all the above examples, the object we aim to generate can be mathematically represented as a vector, typically after flattening. Accordingly, throughout this text, we treat the objects being generated as vectors $\bx\in\real^D$.

\index{Autoregressive (AR) models}
\paragrapharrow{Different paradigms.}
Broadly speaking, generative modeling research falls into two fundamental paradigms. These two frameworks differ drastically in their approaches to factorizing data distributions and in the iterative mechanisms that drive the generative process:
\begin{itemize}
\item \textit{Token-based / sequential refinement (autoregressive, MaskGIT).} 
These models operate within a discrete token space, typically produced by a learned tokenizer such as VQVAE or VQGAN. They model the data distribution as a product of conditional probability terms \citep{van2017neural, esser2021taming}.

\item \textit{Latent-to-data mapping (GANs, VAEs, diffusion models, score-based models, and flow-based models).}
These models learn a transformation that maps a simple prior distribution (most commonly a Gaussian distribution) to the target data distribution, operating primarily in a continuous space.
\end{itemize}

\begin{figure}[h]
\centering
\includegraphics[width=0.95\textwidth]{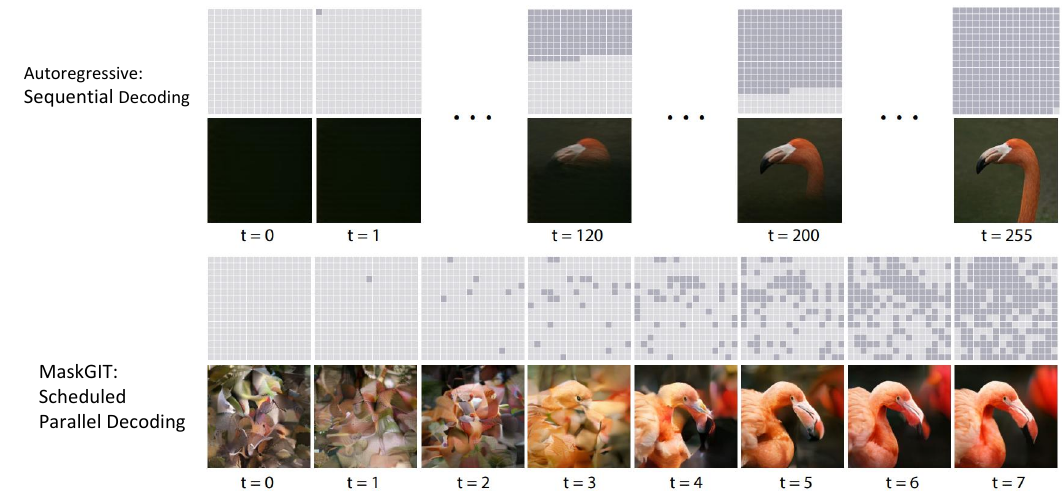}
\caption{
\textbf{The first paradigm of generative models}: Sequential decoding and scheduled parallel decoding. 
In the corresponding visualization, rows 1 and 3 illustrate latent mask inputs at each decoding iteration, while rows 2 and 4 present the generated samples produced at each step. Traditional autoregressive models initialize generation with entirely unknown latent codes (light gray) and fill in latent representations sequentially following a fixed raster scan order (dark gray). MaskGIT decoding also starts with fully unknown latent codes but adopts a parallel generation strategy: it populates latent representations via progressively scattered token predictions (dark gray), with the number of predicted tokens growing rapidly across iterations. Image source: \citet{chang2022maskgit}.
}
\label{fig:autoregres_comp}
\end{figure}

The first paradigm answers the core question: ``Given partial data, what content comes next?". It factorizes the joint data distribution across individual data coordinates and performs iterative generation within the discrete data or token space.
\textit{Autoregressive (AR) models} factorize the probability of data or token sequences $\bx$ as $p(\bx) = \prod p(x_i \mid x_{i-1}, \ldots,x_1)$. These models generate tokens sequentially, with each new token conditioned on all previously generated tokens. Classic examples include PixelRNN and PixelCNN \citep{van2016pixel, van2016conditional}, which explicitly model conditional probability distributions.
\textit{MaskGIT} further extends this framework by relaxing the rigid left-to-right raster ordering enforced by traditional autoregressive models \citep{chang2022maskgit}. Trained with BERT-style masked token prediction, MaskGIT generates content by iteratively unmasking and predicting the most confident tokens in parallel across a small number of decoding steps.
Across all token-based methods, iteration operates over discrete data dimensions or tokens to determine which elements to generate next, and all outputs reside in a discrete space. An intuitive illustration of this paradigm is provided in Figure~\ref{fig:autoregres_comp}.

\begin{figure}[h]
\centering  
\subfigtopskip=2pt 
\subfigbottomskip=9pt 
\subfigcapskip=-5pt 
\includegraphics[width=0.99\textwidth]{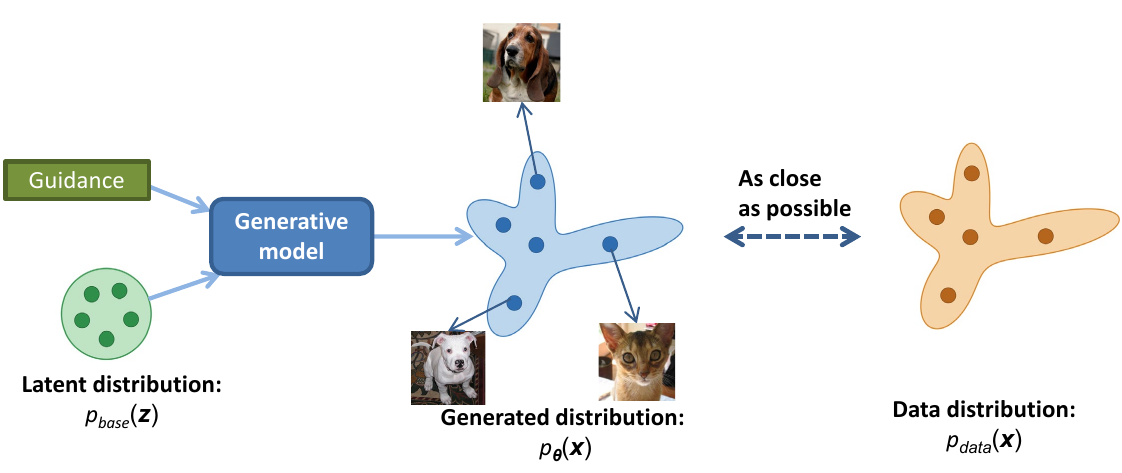}
\caption{
\textbf{The second paradigm of generative models}.
Generative models in this category convert samples drawn from a predefined latent prior distribution (e.g., Gaussian distribution) into samples that conform to the real data distribution.
}
\label{fig:DDPM_gen_model_idea}
\end{figure}

The second paradigm addresses a distinct core question: ``How can we transform random noise into realistic data?". It learns a deterministic or stochastic mapping from a predefined prior distribution to the target data distribution, with iterative processes (where applicable) operating in noise or continuous time space, rather than discrete token space.
Specifically, normalizing flows construct invertible transformations and compute exact data likelihoods via the change-of-variables formula. GANs learn data mappings through adversarial training without explicitly modeling likelihoods. Diffusion and score-based models define a forward progressive noising process and learn the reverse denoising process, with generation iterating over discrete noise levels (timesteps). Flow matching and rectified flow models parameterize a continuous-time velocity field that transports prior distribution samples to realistic data samples via ordinary differential equations (ODEs).
For all models in this paradigm, iterative generation (when present) follows a continuous noise or transport schedule, rather than iterating over individual data dimensions. The latent input is a full-dimensional continuous noise vector, not a partial set of discrete tokens. See Figure~\ref{fig:DDPM_gen_model_idea} for a conceptual demonstration.

While this book focuses primarily on the second paradigm---centered on transforming a simple source distribution into the target distribution of real-world images---the boundary between the two paradigms has become increasingly blurred in recent research. Modern work frequently integrates core ideas from both frameworks.
Hybrid approaches include \textit{continuous autoregressive models}, \textit{token-level diffusion models} (e.g., discrete diffusion and discrete flow models), and \textit{masked autoregressive (MAR) models}. 
These methods unify autoregressive ordering rules with diffusion or flow-matching prediction heads for individual tokens \citep{li2024autoregressive, team2025nextstep}.

\index{Explicit models}
\index{Implicit models}
\paragrapharrow{An alternative classification of generative models.}
As discussed earlier, generative models encompass a diverse set of modeling strategies. A complementary classification framework categorizes generative models based on their parameterization of the underlying data distribution---specifically, whether they define the distribution $p_{\btheta}(\bx)$ explicitly or implicitly, independent of their training objectives:
\begin{itemize}
\item \textit{Explicit models.} These models directly parameterize a probability distribution $p_{\btheta}(\bx)$ via a tractable or approximately tractable density or mass function. Representative examples include AR models, normalization flow models, VAEs,  diffusion models, and flow-matching models.
All of these approaches define $p_{\btheta}(\bx)$ either exactly or through a rigorously tractable lower/upper bound.

\item \textit{Implicit models.} These models characterize the target distribution solely through a stochastic sampling procedure, generally formulated as $\rvx = \mathcalG_{\btheta}(\rvz)$, where the latent noise variable $\rvz $ is sampled from a predefined prior distribution $p_{\text{prior}}$. 
Under this formulation, $p_{\btheta}(\bx)$ does not have a closed-form expression and may not even be fully analytically  defined. 
GANs serve as the canonical example of implicit generative models.
\end{itemize}

\paragrapharrow{Generation as sampling.}
In mainstream generative modeling research, we typically assume access to a finite training dataset consisting of independent and identically distributed (i.i.d.) samples drawn from an intractable, complex
\textit{true data distribution} $p_{\text{data}}(\bx)$. 
These observed data points constitute a finite dataset $\mathcalX=\{\bx_1, \bx_2, \ldots,\bx_N\}\sim p_{\text{data}}$, which acts as a practical surrogate for the inaccessible true distribution. The fundamental objective of generative modeling is to learn a tractable, approximate probability distribution from this finite dataset, as the explicit analytical form of $p_{\text{data}}(\bx)$ is unknown and direct sampling from the true distribution is computationally infeasible.

To further elucidate the essence of the generative task, we formalize the definition of ``data generation" with intuitive examples. Take horse image generation as a typical case: no single ``optimal" horse image exists. Instead, there exists a continuous spectrum of plausible horse images with varying degrees of visual realism and fidelity. In machine learning, this diversity of valid outputs is rigorously modeled as a probability distribution over the entire data space. Mathematically, $p_{\text{data}}: \real^D\rightarrow \real_{+}$ denotes a nonnegative probability density function that assigns a nonnegative likelihood score $p_{\text{data}}(\bx) \geq 0$ to every feasible data sample $\bx\in\real^D$. Samples that better conform to target semantic patterns (e.g., more realistic horse images) yield higher likelihood values under $p_{\text{data}}$. This formulation transforms subjective qualitative evaluations of sample quality into objective quantitative mathematical metrics, and strictly defines the generative task as sampling from the underlying unknown true data distribution $p_{\text{data}}(\bx)$.

To approximate the intractable true data distribution, generative models leverage deep neural networks to parameterize a learnable model distribution $p_{\btheta}(\bx)$, where $\btheta$ denotes the full set of trainable network parameters. The core training objective is to optimize these parameters to obtain an optimal parameter set $\widehatbtheta$, which minimizes the statistical divergence between the learned model distribution $p_{\btheta}(\bx)$ and the true data distribution $p_{\text{data}}(\bx)$. Ideally, the fully optimized model satisfies $p_{\widehatbtheta}(\bx) \approx p_{\text{data}}(\bx)$. After sufficient training, the learned distribution $p_{\widehatbtheta}(\bx)$ acts as a reliable surrogate for the true data distribution, enabling two core capabilities. First, sampling methods such as Monte Carlo sampling can be deployed to generate unlimited novel, realistic data samples. Second, the likelihood $p_{\widehatbtheta}(\bx')$ of any arbitrary sample $\bx'$ can be evaluated to quantify its plausibility under the learned data distribution.

Beyond unconditional generation, numerous real-world applications demand conditional generation guided by auxiliary contextual information, such as \textit{text prompts, categorical labels}, and other conditional variables $\by$. For instance, the model can generate a customized image conditioned on the text prompt: ``A wild brown horse galloping down a snow-covered grassy hillside, with rugged snow-capped mountain peaks towering across the horizon". 
This conditional generation task is mathematically formalized as sampling from the conditional data distribution $\rvx\sim p_{\text{data}}(\cdot\mid \by)$, where $p_{\text{data}}(\cdot\mid \by)$ represents the data distribution constrained by the conditioning variable $\by$. This work primarily focuses on the theoretical framework of unconditional generative models, with  extensions to conditional generative models elaborated in Sections \ref{section:guid_ddpm}, \ref{section:guidance_flow}, and Chapter \ref{chapter:diffarchitect}.

Focusing on the second research paradigm visualized in Figure \ref{fig:DDPM_gen_model_idea}, all mainstream generative models can be uniformly framed as a \textit{distribution transformation problem}. 
This unified framework first defines a simple, tractable base prior distribution $p_{\text{base}}$, conventionally adopted as the standard multivariate Gaussian distribution $p_{\text{base}} = \normal(\bzero, \bI_K)$ with $K\leq D$. A generative model is essentially a learned nonlinear transformation that maps simple prior noise samples to complex real-world data samples. The standard generative process is formalized as follows:
$$
\rvz \sim p_{\text{base}} \;\xrightarrow{\text{Generative Model}}\; \rvx \sim p_{\text{data}}.
$$
This high-level paradigm unifies all dominant generative modeling architectures---including \textit{generative adversarial networks (GANs)}, \textit{variational autoencoders (VAEs)}, \textit{normalizing flows}, \textit{diffusion models}, and \textit{flow-matching models}---under a universal perspective: all these models learn a deterministic or stochastic transformation that converts a simple fixed prior distribution into a complex target data distribution.

\paragrapharrow{Similarity between the two paradigms.}
Despite adopting distinct approaches to content generation, diffusion/flow-matching models and autoregressive (AR) models share several fundamental similarities in their data generation mechanisms, even though their specific implementations and inference pipelines differ substantially. The most essential commonality is that neither paradigm produces complete data outputs in a single forward pass. Instead of one-shot generation, both rely on a step-by-step incremental construction strategy:
\begin{itemize}
\item \textit{AR models:} Generate data sequentially by predicting one token or pixel at a time, constructing the full output in a fixed order (e.g., top-to-bottom or left-to-right for visual data).
\item \textit{Diffusion/flow-matching models:} Produce high-fidelity data through iterative denoising. They gradually refine randomly initialized noisy inputs across numerous steps, transforming pure Gaussian noise into coherent, realistic target data such as clear images.
\end{itemize}
Furthermore, autoregressive models, VAEs, and diffusion models are all fundamentally likelihood-based generative models. Their training objectives uniformly aim to maximize the log-likelihood of training data under their respective generative frameworks, with paradigm-specific optimization strategies:
\begin{itemize}
\item \textit{AR models:} Directly maximize the conditional probability of each individual token given its preceding contextual sequence.
\item \textit{VAE and diffusion models:} Optimize data likelihood indirectly by minimizing the evidence lower-bound (ELBO). This optimization objective can be further decomposed into a sequence of simple, tractable denoising sub-objectives for iterative training.
\end{itemize}

\paragrapharrow{Book structure.}
To conclude, generative AI marks a pivotal paradigm shift in machine learning, shifting the field’s focus from conventional predictive tasks such as classification and regression to creative generative tasks centered on novel data synthesis. This transformative advancement is enabled by major breakthroughs in latent variable modeling, diffusion processes, flow-based optimization methods, and large-scale neural network architectures. These advances have culminated in state-of-the-art systems including \textit{Stable Diffusion}, \textit{DALL-E}, and \textit{Meta Movie Gen}, which significantly bridge the gap between human creativity and machine-generated content. Behind these widely recognized practical applications lies a rigorous mathematical framework built on probability theory, variational inference, and deep learning---the core theoretical foundations that this book systematically elaborates and explains in depth.

The book is structured to help readers build professional expertise in generative modeling progressively, starting from basic foundational concepts and gradually advancing to sophisticated modern architectures, with the organizational breakdown as follows:
\begin{itemize}
\item Part 1: Foundations (Chapters~\ref{chapter_genintroduction}--\ref{chapter:vae_gan}) establishes the fundamental mathematical and theoretical framework of generative modeling. This part covers essential mathematical notations, core probability theories, variational inference principles, and classic latent variable models including VAEs and GANs. It demystifies key concepts such as the evidence lower-bound (ELBO), adversarial training mechanisms, and amortized inference, equipping readers with solid theoretical knowledge and practical tools to analyze and implement basic generative models.

\item Part 2: Diffusion models (Chapter~\ref{chapter:diff}) focuses on the dominant technical paradigm in contemporary generative AI. It comprehensively introduces \textit{diffusion probabilistic models (DPMs)}, \textit{denoising diffusion probabilistic models (DDPMs)}, \textit{latent diffusion models (LDMs)}, and conditional guidance mechanisms. This chapter elaborates on how diffusion models iteratively purify random noise into structured, coherent data, empowering high-quality generation of images, videos, and other complex data modalities.

\item Part 3: Flow-based and score-based models (Chapters~\ref{chapter:flow}--\ref{chapter:scorematch}) explores alternative mainstream generative modeling frameworks, encompassing standard normalizing flows, continuous flow models, and score matching methods. Distinct from diffusion models, these architectures support exact likelihood calculation and deliver stable training performance, serving as effective complementary approaches to diffusion-based generative systems.

\item Part 4: Advanced sampling methods (Chapter~\ref{chapter:dpmsampler}) delve into the practical acceleration and refinement of diffusion model sampling, addressing the computational bottleneck of traditional stepwise generation. Building on the theoretical SDE/ODE foundations of diffusion processes, it introduces state-of-the-art sampling algorithms---including DEIS, DPM-Solver, and DPM-Solver++---that drastically reduce the number of required sampling steps while preserving output fidelity. By bridging continuous-time stochastic dynamics with efficient numerical solvers, this part equips readers with the tools to deploy high-speed, high-quality generative systems for real-world applications, from image synthesis to video generation.

\item Part 5: Modern generative architectures (Chapter~\ref{chapter:diffarchitect}) discusses the key architectural and engineering innovations that underpin modern large-scale generative AI systems. It covers core modules and technologies including \textit{CLIP}, \textit{attention mechanisms}, \textit{U-Nets}, \textit{ControlNet}, \textit{diffusion transformers (DiTs)}, and multimodal generative frameworks. This part also explores latent space manipulation strategies and presents in-depth case studies of cutting-edge models such as \textit{Stable Diffusion 3}.
\end{itemize}

\section{Notations and Mathematical Tools}
In the remainder of this chapter, we introduce and review some fundamental concepts from linear algebra and probability theory that are relevant to generative models. Additional important concepts will be defined and discussed as needed throughout the text for clarity.

In all cases, scalars will be denoted in non-bold font, possibly with subscripts (e.g., $a$, $\alpha$, $\alpha_i$).
Vectors will be represented by \textbf{boldface} lowercase letters, again possibly with subscripts (e.g., $\bmu$, $\bx$, $\bx_n$, $\bz$), and matrices by \textbf{boldface} uppercase letters, also possibly with subscripts (e.g., $\bX$, $\bL_j$).
However, we will use $\bx$ to denote the data object in generative modeling notation, even when it represents a matrix; see Chapter~\ref{chapter:diffarchitect}.
The $i$-th element of a vector $\bz$ will be written as $z_i$ in non-bold font.
To distinguish deterministic quantities from random ones, we adopt the following convention:
\begin{itemize}
\item Random variables (scalars) are denoted using \textit{upright (non-italic)} font (e.g., $\ra$ and $\rb_1$ are random variables, whereas italicized $a$ and $b_1$ denote deterministic scalars).
\item Random vectors are indicated by \textit{upright bold lowercase letters}, possibly with subscripts (e.g., $\rva$ and $\rvb_1$ are random vectors, while italicized bold $\ba$ and $\bb_1$ represent deterministic vectors).
\item Random matrices are denoted by \textit{upright bold uppercase letters}, possibly with subscripts (e.g., $\rmA$ and $\rmB_1$ are random matrices, whereas italicized bold $\bA$ and $\bB_1$ denote deterministic matrices).
\end{itemize}
Subarrays are formed by fixing a subset of indices of a matrix.
The element located in the $i$-th row and $j$-th column of a matrix $\bX$ (i.e., the $(i,j)$ entry) is denoted by $x_{ij}$. 
Consequently, a matrix $\bX\in\real^{N\times D}$ may be expressed as $\bX=\{x_{ij}\}_{i,j=1}^{N, D}=[x_{ij}]$.
For notational convenience, we adopt \textbf{Matlab-style indexing}: 
\begin{itemize}
\item The submatrix consisting of rows $i$ through $j$ and columns $k$ through $m$ of $\bX$ is denoted by  $\bX_{i:j,k:m} \equiv \bX[i:j,k:m]$. 
\item A colon is used to indicate all elements along a dimension. For example, $\bX_{:,k:m} \equiv\bX[:,k:m]$ denotes the submatrix formed by columns $k$ through $m$; and $\bX_{:,k}\equiv\bX[:,k]$ denotes the $k$-th column of $\bX$. 
\item   Alternatively, the $k$-th column of $\bX$ may be denoted more compactly by $\bx_k$; and the $k$-th row of $\bX$ can be denoted as $\bx^{(k)}$.
\end{itemize}

Throughout this book, all vectors are assumed to be column vectors unless explicitly transposed. A row vector is denoted as the transpose of a column vector, e.g., $\bx^\top$. 
Concrete vectors are written using MATLAB-style syntax.
A column vector is written with semicolons $``;"$ separating entries,  e.g., 
$$\bx=[1;2;3] \qquad \text{(column vector)}
$$ 
is a column vector in $\real^3$. Similarly, A row vector uses commas $``,"$ to separate entries, e.g.,
$$\by=[1,2,3]\qquad \text{(row vector)}
$$ 
is a row vector with three entries. 
Equivalently, a column vector may be expressed as the transpose of a row vector: $\by=[1,2,3]^\top$ is a column vector.

The transpose of a matrix $\bX$ is denoted by $\bX^\top$,
and its inverse (when it exists)  by $\bX^{-1}$. 
The $D \times D$ identity matrix is denoted by $\bI_D$ or simply by $\bI$. 
A vector or matrix of all zeros is denoted by the \textbf{boldface} symbol $\bzero$; its dimensions are inferred from context, or explicitly indicated as $\bzero_D$ for a $D$-dimensional zero vector.
Similarly, a vector or matrix of all ones is  denoted by a \textbf{boldface} symbol $\bone$, whose dimensions are clear from  context; or we write $\bone_D$ to denote the vector of all ones with $D$ entries.
Subscripts are often omitted when the dimensions are evident from  context.

\index{Vector norm}
\index{Norm}
\index{Matrix norm}
\subsection{Norms}\label{section:vec_norms}
The concept of a \textit{norm} is fundamental to quantifying the magnitude of vectors or  and enables the formal definition of metrics on normed linear spaces. Norms provide a rigorous numerical measure of the ``size" of vectors and matrices, serving as a foundational tool across numerous applications. Typical use cases include calculating the Euclidean length of a vector and quantifying the scale of matrices in high-dimensional multivariate analysis.

Furthermore, norms facilitate the definition of distances between vectors and matrices. Specifically, the distance between two vectors $\bu$ and $\bv$ is defined as the norm of their difference, namely $\norm{\bu-\bv}$. This distance formulation underpins proximity-based algorithms, including common clustering methods in machine learning.
\footnote{Our discussion focuses exclusively on norms and inner products defined over real vector and matrix spaces. All corresponding results can be directly extended to the complex domain.}

Any valid vector or matrix norm must satisfy three core axioms, as formally stated below (see, for example,  \citet{lu2021numerical} for comprehensive background).
\begin{definition}[Vector norm and matrix nrom\index{Matrix norm}\index{Vector norm}]\label{definition:matrix-norm}
Let $\norm{\cdot}$ denote a generic norm function for vectors and matrices. For any matrix $\bX \in \real^{N\times D}$ and any vector $\bbeta \in \real^D$, the following three properties hold:
\begin{itemize}
\item \textit{Nonnegativity}. $\norm{\bX} \geq 0$ and $\norm{\bbeta}\geq 0$. 
Equality holds if and only if $\bX=\bzero $ and $\bbeta=\bzero$, respectively. 
\item \textit{Positive homogeneity}. $\norm{\lambda \bX} = \abs{\lambda} \cdot \norm{\bX}$ and  $\norm{\lambda \bbeta} = \abs{\lambda} \cdot \norm{\bbeta}$ for any scale $\lambda \in \real$.
\item \textit{Triangle inequality}. For all matrices $\bX, \bY\in \real^{N\times D}$, $\norm{\bX+\bY} \leq \norm{\bX}+\norm{\bY}$.  
For all vectors $\balpha,\bbeta\in \real^D$, $\norm{\balpha+\bbeta} \leq \norm{\balpha}+\norm{\bbeta}$.
\end{itemize}
\end{definition}

Based on the general norm axioms above, we introduce several commonly used vector norms, including the $\ell_1$, $\ell_2$, and $\ell_\infty$-norms, as well as the generalized $\ell_p$-norm.
\begin{definition}[Vector $\ell_1, \ell_2, \ell_\infty$, $\ell_p$-norms]\label{definition:vec_l2_norm}
Given a vector $\bbeta\in\real^D$, the \textit{$\ell_2$ vector norm (Euclidean norm)} is defined as $\normtwo{\bbeta} \triangleq  \sqrt{\beta_1^2+\beta_2^2+\ldots+\beta_D^2}$.
The \textit{$\ell_1$-norm} is defined as
$
\norm{\bbeta}_1 \triangleq \sum_{i=1}^{D} \abs{\beta_i} .
$
And the \textit{$\ell_\infty$-norm} is defined as 
$
\norm{\bbeta}_\infty \triangleq \mathop{\max}_{i=1,2,\ldots,D} \abs{\beta_i} .
$
More generally, the $\ell_p$-norm is defined as $\normp{\bbeta}\triangleq \sqrt[p]{ \sum_{i=1}^{D}\abs{\beta_i}^p  }$ for $p\geq 1$.
\end{definition}

\index{Matrix norm}
While vector norms quantify the ``size" or ``length" of vectors, matrix norms extend this magnitude measurement to matrices with a key distinction: they additionally characterize the linear operator behavior of matrices, specifically their scaling and transformation effects on input vectors.
For a matrix $\bX\in\real^{N\times D}$, we define the matrix Frobenius norm as follows.
\begin{definition}[Matrix Frobenius norm\index{Frobenius norm}]\label{definition:frobernius-in-svd}
The \textit{Frobenius norm} of a matrix $\bX\in \real^{N\times D}$ is defined as 
$$
\normf{\bX} \triangleq  \sqrt{\sum_{i=1,j=1}^{N,D} x_{ij}^2}=\sqrt{\trace(\bX\bX^\top)}=\sqrt{\trace(\bX^\top\bX)} = \sqrt{\sigma_1^2+\sigma_2^2+\ldots+\sigma_r^2}, 
$$
where $\sigma_1, \sigma_2, \ldots, \sigma_r$ denote nonzero singular values of $\bX$, and  $\trace(\bX^\top\bX)$ denotes the trace of $\bX^\top\bX$, i.e., the sum of diagonal entries of $\bX^\top\bX$.
\end{definition}

Another widely adopted matrix norm is the spectral norm, formally defined below.
\begin{definition}[Matrix spectral norm]\label{definition:spectral_norm}
The \textit{spectral norm} of a matrix $\bX\in \real^{N\times D}$ is defined as 
\begin{equation}\label{equation:spectral_norm_eq1}
\normtwo{\bX} = \mathop{\max}_{\bbeta\neq\bzero} \frac{\normtwo{\bX\bbeta}}{\normtwo{\bbeta}}  =\mathop{\max}_{\bu\in \real^D: \norm{\bu}_2=1}  \normtwo{\bX\bu} .
\end{equation}
~\footnote{Throughout this book, we exclusively use $\max$ and $\min$ notation instead of supremum/infimum notation, even when the corresponding extremum is not strictly attained.}
This value is equivalent to the largest singular value of  $\bX$, i.e., $\normtwo{\bX} = \sigma_{\max}(\bX)$; see Section~\ref{section:SVD}.
\end{definition}
For a symmetric matrix $\bX$, its spectral norm is also equivalent to the maximum absolute value of its quadratic form over unit vectors (i.e., maximum absolute eigenvalue):
\begin{equation}\label{equation:spectral_norm_eq2}
\normtwo{\bX}  = \max_{\bu\in \real^D: \norm{\bu}_2=1}\abs{\bu^\top\bX\bu}, \quad \text{if $\bX=\bX^\top$}.
\end{equation}

The spectral norm is upper-bounded by the Frobenius norm:
\begin{equation}
\normtwo{\bX} \leq \normf{\bX}.
\end{equation}
This inequality follows directly from the definitions of the two norms: $\normtwo{\bX}$ corresponds to the maximum singular value $\sigma_{\max}(\bX)$, while $\normf{\bX}$ equals the square root of the sum of squared singular values of $\bX$.
Alternatively, this bound can be proven via the Cauchy--Schwarz inequality:
$$
\normtwo{\bX\bbeta}^2 
= \sum_{i=1}^{N} \left(\sum_{j=1}^{D}  x_{ij}\beta_j\right)^2
\leq \sum_{i=1}^{N} \left(\sum_{j=1}^{D} x_{ij}^2\right) \left(\sum_{j=1}^{D} \beta_j^2\right)
=\normf{\bX}\normtwo{\bbeta},
$$
where the key inequality step relies on the Cauchy--Schwarz inequality.


Notably, the Frobenius norm can be regarded as the matrix analogue of the vector $\ell_2$-norm. For notational simplicity, we omit the full subscripts of the vector $\ell_2$-norm and matrix Frobenius norm when the intended norm is unambiguous from context, following the convention:
$$
\norm{\bX}=\normf{\bX}
\qquad \text{and}\qquad
\norm{\bx}=\normtwo{\bx}.
$$
In contrast, the subscript of the spectral norm $\normtwo{\bX}$ must always be retained and never omitted.

Beyond the operator-based matrix norms introduced above, an alternative approach to defining matrix norms is to vectorize a matrix $\bX\in\real^{N\times D}$ and treat it as a standard vector in $\real^{ND}$. What fundamentally distinguishes rigorous matrix norms from naive vector-induced matrix magnitudes is a critical property known as \textit{submultiplicativity}: for any submultiplicative matrix norm $\norm{\cdot}$, the inequality $\norm{\bX\bY}\leq \norm{\bX}\norm{\bY}$ holds for all compatible matrices $\bX$ and $\bY$. 
See, for example, \citet{lu2021numerical} for more details.

\index{Singular value decomposition}
\subsection{Singular Value Decomposition (SVD)}\label{section:SVD}

This subsection introduces the matrix \textit{singular value decomposition (SVD)}. Before presenting the general SVD formulation, we first review the spectral decomposition of symmetric matrices, which serves as a foundational special case of SVD.
The spectral theorem (or spectral decomposition) for symmetric matrices states that every real symmetric matrix possesses real eigenvalues and is fully diagonalizable over an orthonormal real basis.
\footnote{The spectral decomposition for Hermitian matrices yields an analogous result: Hermitian matrices have real eigenvalues and admit diagonalization with respect to a complex orthonormal basis.}

\begin{theoremHigh}[Spectral decomposition\index{Spectral decomposition}\index{Spectral theorem}]\label{theorem:spectral_theorem}
A real square matrix $\bX \in \real^{D\times D}$ is symmetric if and only if there exist an orthogonal matrix $\bQ$ and a diagonal matrix $\bLambda$ such that
\begin{equation*}
\bX = \bQ \bLambda \bQ^\top,
\end{equation*}
where the columns of $\bQ = [\bq_1, \bq_2, \ldots, \bq_D]$ form a set of mutually orthonormal eigenvectors of $\bX$, and the diagonal entries of $\bLambda=\diag(\lambda_1, \lambda_2, \ldots, \lambda_D)$ are the corresponding real eigenvalues.
In particular, symmetric matrices satisfy the following key properties:
\begin{enumerate}
\item All eigenvalues of a symmetric matrix are \textbf{real-valued}.
\item Its eigenvectors can be selected to form an \textbf{orthonormal} set.
\item The rank of $\bX$ equals the number of its nonzero eigenvalues.
\item If all eigenvalues are distinct, then the corresponding eigenvectors are linearly independent.
\end{enumerate}
\end{theoremHigh}
\begin{proof}
See \citet{lu2021numerical}.
\end{proof}

Spectral decomposition provides a clean diagonal factorization for symmetric square matrices. However, this framework is limited: diagonalization via eigen decomposition is generally unavailable for non-symmetric or non-square matrices. Singular value decomposition (SVD) resolves this limitation by generalizing the decomposition to any real matrix. Instead of a single orthogonal eigenvector matrix, SVD factorizes an arbitrary matrix into the product of two orthogonal matrices and a diagonal matrix of singular values. We now formally state the SVD theorem.

\begin{theoremHigh}[Full singular value decomposition\index{Singular value decomposition}]\label{theorem:full_svd_rectangular}\label{theorem:reduced_svd_rectangular}
Every real $N\times D$ matrix $\bX$ with rank $r$  admits a decomposition of the form
$$
\bX = \bU \bSigma \bV^\top,
$$ 
where $\bSigma\in \real^{N\times D}$ follows the block structure $\bSigma=\footnotesize\begin{bmatrix}
\bSigma_r & \bzero \\
\bzero & \bzero
\end{bmatrix}$ with $\bSigma_r=\diag(\sigma_1, \sigma_2 \ldots, \sigma_r)\in \real^{r\times r}$. 
The singular values are sorted in non-increasing order: $\sigma_1 \geq \sigma_2 \geq \ldots \geq \sigma_r$.
\begin{itemize}
\item The scalars  $\sigma_i$  denote the nonzero \textit{singular values} of  $\bX$. Each singular value $\sigma_i$ equals the positive square root of a nonzero eigenvalue shared by both $\bX^\top \bX$ and $\bX \bX^\top$.

\item $\bU\in \textcolor{black}{\real^{N\times N}}$ is an orthogonal matrix. 
Its first $r$ columns correspond to the eigenvectors of $\bX \bX^\top$ associated with the $r$ nonzero eigenvalues, while the remaining $N-r$ columns form an orthonormal basis for the null space of $\bX^\top$: $\nspace(\bX^\top)\triangleq \{\balpha \mid  \bX^\top\balpha=\bzero\}$.

\item $\bV\in \textcolor{black}{\real^{D\times D}}$ is an orthogonal matrix. 
Its first $r$ columns correspond to the eigenvectors of $\bX^\top \bX$ associated with the $r$ nonzero eigenvalues, while the remaining $D-r$ columns form an orthonormal basis for the null space of $\bX$: $\nspace(\bX)\triangleq \{\bbeta\mid \bX\bbeta=\bzero\}$.

\item The columns of $\bU$ and $\bV$ are referred to as the \textit{left and right singular vectors} of $\bX$, respectively. 
\end{itemize}
Furthermore, the full SVD can be rewritten as a summation of rank-one outer-product matrices: $ \bX = \bU \bSigma \bV^\top = \sum_{i=1}^r \sigma_i \bu_i \bv_i^\top$.
\end{theoremHigh}

For a rank-$r$ matrix $\bX$, the first $r$ singular values satisfy $\sigma_1,\sigma_2,\ldots,\sigma_r>0$, while all subsequent singular values are identically zero: $\sigma_{r+1}=\sigma_{r+2}=\ldots=0$. In most practical applications, the \textit{truncated reduced singular value decomposition (reduced SVD)} is preferred for its compact form.

\begin{figure}[h!]
\centering  
\vspace{-0.35cm}  
\subfigtopskip=2pt  
\subfigbottomskip=2pt  
\subfigcapskip=-5pt  
\subfigure[Reduced SVD decomposition.]{\label{fig:svdhalf}
\includegraphics[width=0.47\linewidth]{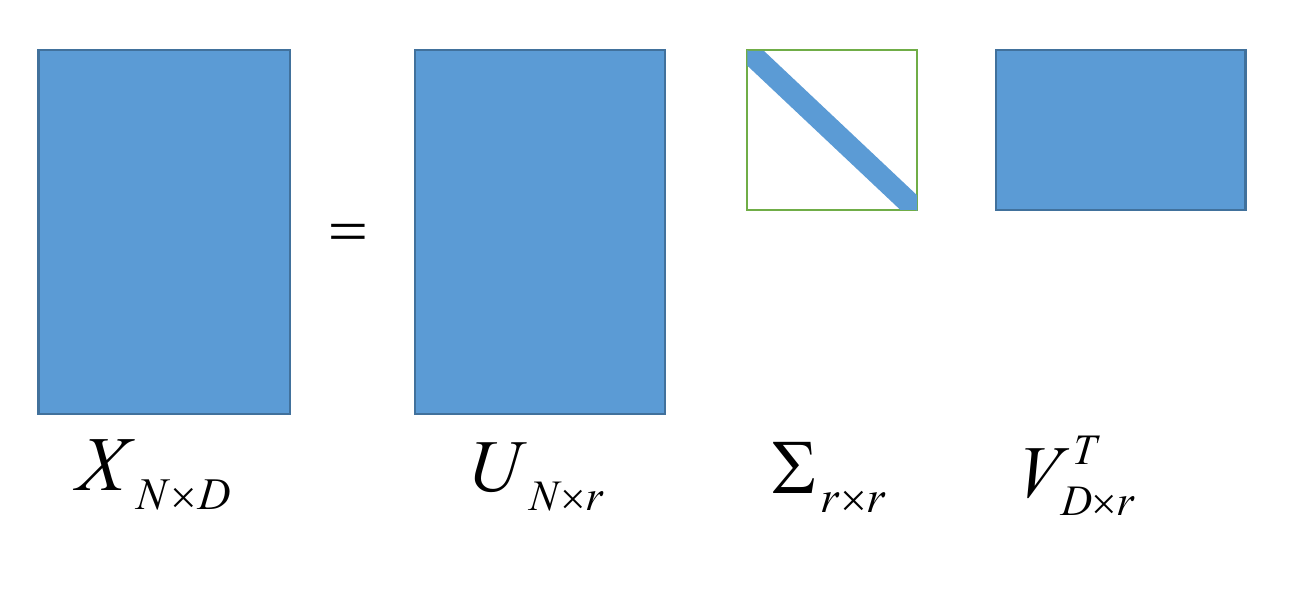}}
\quad 
\subfigure[Full SVD decomposition.]{\label{fig:svdall}
\includegraphics[width=0.47\linewidth]{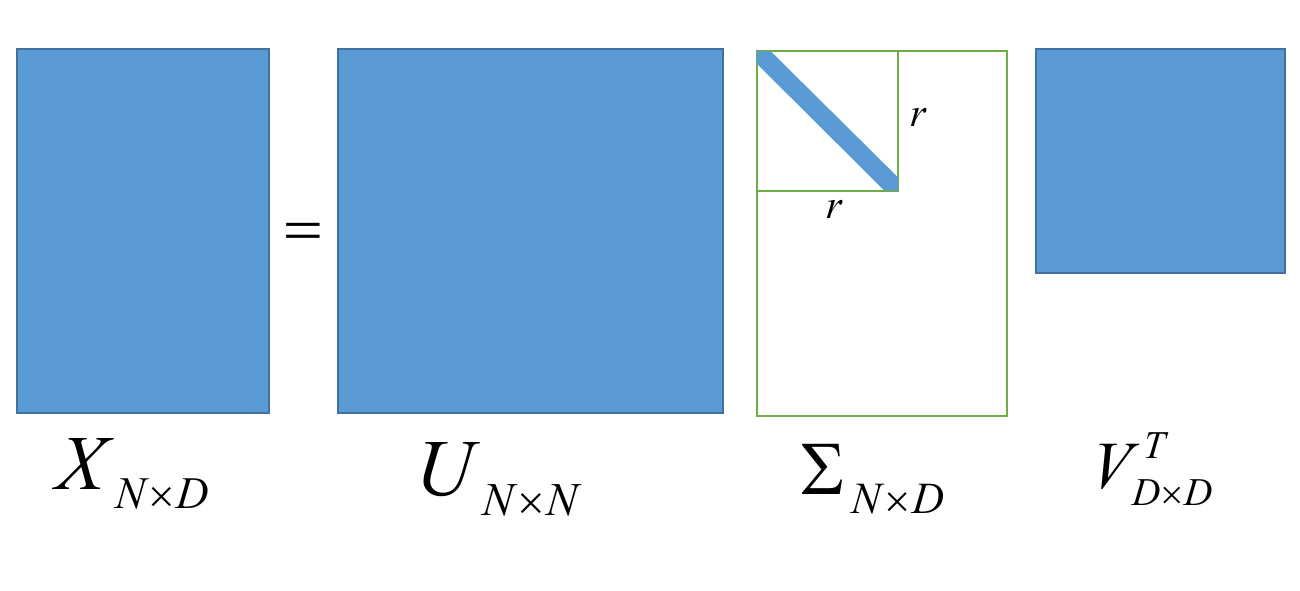}}
\caption{Comparison of the full and reduced SVD. White entries correspond to zero values, while blue entries denote values that are not necessarily zero.}
\label{fig:svd-comparison}
\end{figure}

Given the full SVD $\bX = \bU\bSigma \bV^\top$ of a rank-$r$ matrix $\bX$, we define the truncated submatrices $\bU_r \in \real^{N\times r}$ and $\bV_r \in \real^{D\times r}$ by retaining only the first $r$ columns of $\bU$ and $\bV$, respectively: $\bU = [\bU_r, \bu_{r+1}, \ldots, \bu_N]$ and $\bV = [\bV_r, \bv_{r+1}, \ldots, \bv_D]$. We also retain the positive rank-$r$ singular value matrix $\bSigma_r = \diag(\sigma_1,\ldots,\sigma_r) \in \real^{r\times r}$. This yields the reduced SVD factorization:
\begin{equation}
\bX=\bU_r\bSigma_r\bV_r^\top.
\end{equation}
A visual comparison between full and reduced SVD structures is provided in Figure~\ref{fig:svd-comparison}, where white entries represent zero values and blue entries denote arbitrary nonzero values.

For any matrix $\bX\in\real^{N\times D}$ with reduced SVD $\bX = \bU_r\bSigma_r\bV_r^\top$, straightforward algebraic derivation yields the following decompositions:
\begin{align*}
\bX^\top\bX &= \bV_r\bSigma_r^\top\bU_r^\top\bU_r\bSigma_r\bV_r^\top = \bV_r\bSigma_r^2\bV_r^\top;\\
\bX\bX^\top &= \bU_r\bSigma_r\bV_r^\top\bV_r\bSigma_r^\top\bU_r^\top = \bU_r\bSigma_r^2\bU_r^\top.
\end{align*}
These expressions recover the reduced spectral decompositions of the symmetric positive semidefinite matrices $\bX^\top\bX$ and $\bX\bX^\top$, respectively.
In particular, the singular values of $\bX$ satisfy the fundamental relation:
\begin{equation}\label{equation:sigbd_nearortho_eqp}
\sigma_i = \sigma_i(\bX) = \sqrt{\lambda_i(\bX^\top\bX)} = \sqrt{\lambda_i(\bX\bX^\top)}, \quad i = 1,\ldots,\min\{N, D\}, 
\end{equation}
where $\lambda_1(\bX^\top\bX) \geq \lambda_2(\bX^\top\bX) \geq \ldots$ denotes the sorted eigenvalues of $\bX^\top\bX$ in non-increasing order.
Similarly, the left and right singular vectors of $\bX$ are exactly the eigenvectors obtained from the spectral decomposition of $\bX\bX^\top$ and $\bX^\top\bX$, respectively.
As a result, the SVD of any real matrix can be explicitly constructed from the spectral decompositions of these two symmetric positive semidefinite Gram matrices. This connection also provides a constructive proof for the existence of the matrix SVD.

\begin{exercise}[Proof of SVD]
Based on the above discussion and spectral decomposition theory, prove the existence of the singular value decomposition for an arbitrary real matrix $\bX\in\real^{N\times D}$.
\end{exercise}

\index{Random variable}
\index{Cumulative distribution function}
\subsection{Probability Theory}\label{section:stat_prob}

A \textit{random variable} is a quantity that takes different values governed by inherent randomness, and it is widely used to model uncertain outcomes and stochastic events. In this book, we adopt a consistent typographic convention to distinguish random quantities from their observed realizations: random variables are denoted by lowercase letters in \textbf{upright (non-italic) font}, while their realized or observed values are denoted by \textbf{lowercase italic letters}. For example, $y_1$ and $y_2$ are valid realizations of the random variable $\ry$.

We extend this convention to vector-valued and matrix-valued random quantities. 
\textit{Random vectors (RVs)} are written in {upright bold font} (e.g., $\rvy$), and their specific realizations are written in italic bold font (e.g., $\by$). 
Similarly, \textit{random matrices (RMs)} are denoted by upright bold capital letters (e.g., $\rmY$), whereas their corresponding deterministic realizations adopt italic bold capital letters (e.g., $\bY$). This strict typographic distinction eliminates ambiguity between stochastic quantities and their observed values throughout the text.

A standalone random variable only characterizes the set of all possible outcomes of a stochastic process. To fully quantify stochastic behavior, a random variable must be paired with a probability distribution that quantifies the likelihood of each possible outcome.
In probability theory, the outcome of a random process is modeled as a random variable, whose overall statistical behavior is fully described by its \textit{cumulative distribution function (CDF)}, defined as
\begin{equation}
F(x) \equiv \prob[\rx \leq x].
\end{equation}
In practice, the probabilistic behavior of a random variable $\rx$ is often controlled by an underlying characteristic $\btheta$, referred to as a model parameter. Since the probability of the event $\{\rx \leq x\}$ depends explicitly on $\btheta$, the corresponding CDF is parameterized accordingly:
\begin{equation}
F(x; \btheta) \equiv {\prob}_{\btheta}[\rx \leq x].
\end{equation}
When both the functional form $F(x; \btheta)$ and the true parameter $\btheta$ are known, the probability of any outcome $\rx \leq x$ can be directly evaluated via the parameterized CDF.


\index{Marginal distribution}
\index{Joint probability mass function}
\paragrapharrow{Discrete  random variables.}
Random variables fall into two primary categories: \textit{discrete} and \textit{continuous}. 
A discrete random variable takes values over a finite or countably infinite set of distinct outcomes. These outcomes are not required to be numerical and can instead represent categorical or symbolic states, such as ``red," ``blue," ``success," or ``failure". In contrast, a continuous random variable takes values over an uncountable set, typically a continuous subset of the real line.

Probability distributions assign probabilistic weights to the outcomes of random variables, and their mathematical representation differs substantially for discrete and continuous variables.

For discrete random variables, we adopt the \textit{probability mass function (PMF)}, commonly denoted as $\prob$ or $f$. The PMF maps each feasible outcome of a discrete random variable to its corresponding probability. We use several equivalent notations for the probability that the random variable $\ry$ takes the specific value $y$:
\begin{equation}
\prob(\ry = y)
\quad (\text{or equivalently } \prob(y), \;f_{\ry}(y), \; 
\text{Pr}_{\ry}(y)) .
\end{equation}
By definition, all probabilities lie within the interval $[0,1]$, where $1$ denotes a certain event and $0$ denotes an impossible event. A standard shorthand for specifying the distribution of a random variable uses the distribution symbol ``$\sim$", written as
\begin{equation}
\ry \sim \prob(\ry).
\end{equation}

PMFs can be generalized to characterize {joint probabilistic} behavior across multiple random variables, known as the \textit{joint probability mass function (joint PMF)}. For two discrete random variables $\rx$ and $\ry$, the joint PMF $\prob(\rx = x, \ry = y)$ quantifies the probability that the outcomes $x$ and $y$ occur simultaneously. We frequently use the simplified notations $\prob(x, y)$ or $\prob_{\rx, \ry}(x,y)$. When the joint distribution depends on a parameter set $\btheta$, we write $\prob(x, y \mid \btheta)$  (or $\prob_{\btheta}(x, y)$, $f(x, y; \btheta)$) for compactness.

In practical probabilistic modeling, we often evaluate the likelihood of one event conditioned on the occurrence of another, which is formalized as \textit{conditional probability}. 
The conditional probability of observing $\rx = x$ given that $\ry = y$ is defined as
\begin{equation}\label{equation:bayes_base}
\prob(\rx = x \mid  \ry = y) = \frac{\prob(\rx = x, \ry = y)}{\prob(\ry = y)},
\end{equation}
which is valid whenever $\prob(\ry = y)>0$. 
This fundamental identity serves as the foundation for \textit{Bayes' theorem} \citep{bayes1958essay}.

Given a joint distribution over a collection of random variables, we often derive the \textit{marginal distribution} that describes the statistical behavior of a single subset of variables by marginalizing out the remaining variables. For discrete settings, the marginal distribution of $\rx$ is obtained by summing the joint PMF over all feasible outcomes  of $\ry$:
\begin{equation}
\prob(\rx=x) = \sum_{y} \prob(\rx=x, \ry=y).
\end{equation}

\index{Probability density function}
\paragrapharrow{Continuous random variables.}
For continuous random variables, probability distributions are characterized by a \textit{probability density function (PDF)}, which replaces the PMF used for discrete variables. A valid PDF $p$ must satisfy the following three axiomatic properties:
\begin{itemize}
\item The domain of $p$ covers all possible values of the random variable $\ry$.
\item Unlike a PMF, a PDF value is not bounded above by one, i.e., $p(y)$ may exceed $1$. However, the PDF must be everywhere nonnegative: $p(y)\geq 0$ for all valid outcomes $y$ of $\ry$.
\item The PDF integrates to unity over its entire domain: $\int p(y)\diff y = 1$. Throughout this book, we omit integration limits when the integral is taken over the full real space, such that $\int \equiv \int_{\real}$.
\end{itemize}
Crucially, the PDF value $p(y)$ (also written $f_{\ry}(y)$ or $p_{\ry}(y)$
\footnote{For notational simplicity, we occasionally abbreviate the PDF $p_{\rx_t}$ of a scalar random variable $\rx_t$ or $p_{\rvx_t}$ of a random vector $\rvx_t$ as $p_t$.}) does not represent a probability itself; it represents a probability density. The probability of the random variable falling within an infinitesimal interval of width $\delta y$ centered at $y$ is approximately $p(y)\delta y$.
More formally, for any measurable set $\sA$, the probability that the realized value $y$ belongs to $\sA$ is computed by integrating the PDF over that set:
\begin{equation}
\prob(y\in\sA) = \int_{\sA}p(y)\diff y, 
\quad \text{where } \int p(y)\diff y = 1.
\end{equation}
When the PDF is parameterized by a parameter vector $\btheta$, we adopt several standard and interchangeable notations:
$$
p_\rx(x\mid \btheta),
\quad 
p(x\mid \btheta),
\quad 
p_{\btheta}(x),
\quad 
f_{\rx}(x; \btheta),
\quad  
\text{or simply }\; f(x; \btheta).
$$
The vertical bar or semicolon denotes dependence on the fixed model parameters $\btheta$.

\index{Variance}
\paragrapharrow{Distribution function.}
Every probability distribution can be uniquely characterized by its cumulative distribution function (CDF). For a random variable $\rx$, the CDF is defined as $F(x)=\prob[\rx\leq x]$. The CDF is non-decreasing and right-continuous, and satisfies the following boundary conditions:
$$
0 \leq F(x) \leq 1, \qquad \lim_{x\rightarrow -\infty}F(x) = 0, 
\qquad\text{and}\qquad 
\lim_{x\rightarrow \infty}F(x) = 1.
$$
\begin{subequations}
The \textit{expected value (mean)} $\mu$ and \textit{variance} $\omega^2$ of $\rx$ in the continuous case are defined as
\begin{equation}
\mu = \Exp[\rx] \triangleq \int_{-\infty}^{\infty} x \diff F(x), 
\qquad \omega^2=\Var[\rx] \triangleq \Exp[(\rx - \mu)^2] = \int_{-\infty}^{\infty} (x - \mu)^2 \diff F(x).
\end{equation}
Similarly, in the discrete case, where $\prob(\rx=x)$ denotes the probability mass function, they are defined as 
\begin{equation}
\mu = \Exp[\rx] \triangleq \sum_{x} x \prob(x), \qquad \omega^2 =\Var[\rx]\triangleq \Exp[(\rx - \mu)^2] = \sum_{x} (x - \mu)^2 \prob(x).
\end{equation}
It is straightforward to verify the useful identity: 
\begin{equation}
\Var[\rx] = \Exp[\rx^2] - (\Exp[\rx])^2.
\end{equation}
The \textit{median} of a random variable $\rx$ is any  number $M$ satisfying
\begin{equation}
\prob(\rx \geq M) \geq \frac{1}{2} 
\qquad \text{and} \qquad 
\prob(\rx \leq M) \geq \frac{1}{2}.
\end{equation}
\end{subequations}
\begin{exercise}[Uniform distribution\index{Uniform distribution}]\label{exercise:uniform_dist}
Let $\rx\sim \uniformdist(x\mid a,b)$ follow a uniform distribution over $[a,b]$, whose PDF is given by $p_{\rx}(x) = \frac{1}{b-a}$ for $a\leq x\leq b$ and $p_{\rx}(x)=0$ otherwise. Prove that
$$
\Exp[\rx] = \frac{a+b}{2}
\qquad \text{and}\qquad 
\Var[\rx] =\frac{(b-a)^2}{12}.
$$
\end{exercise}

We now extend the above definitions to multivariate settings. Consider a random vector $\rvx = [\rx_1, \rx_2, \dots, \rx_n]^\top$.
Its \textit{joint cumulative distribution function (joint CDF)}, denoted as $F_{\rvx}(\bx)$, $F(x_1,x_2,\dots,x_n)$, or simply $F(\bx)$, is defined as 
$$
F_{\rvx}(x_1, x_2, \ldots, x_n) = \prob[\rx_1\leq x_1, \rx_2, \leq x_2, \ldots, \rx_n\leq x_n].
$$
The representation of the joint distribution depends on whether the constituent random variables are discrete or continuous:
\begin{itemize}
\item When all the variables are discrete, as discussed above, the joint probability mass function or the joint frequency function can be characterized as 
$$
f_{\rvx}(x_1, x_2, \ldots, x_n) = \prob[\rx_1=x_1, \rx_2=x_2, \ldots, \rx_n=x_n].
$$
\item On the contrary, if the variables are continuous, the \textit{joint probability density function} is a function $f_{\rvx} :\real^n\rightarrow [0, \infty)$ (or simply $f(\bx)$, $p_{\rvx}(\bx)$, $p(\bx)$) such that 
$$
F_{\rvx}(x_1, x_2, \ldots, x_n) = \int_{-\infty}^{x_1}\ldots \int_{-\infty}^{x_n} f_{\rvx}(y_1, y_2, \ldots, y_n)\diff y_1 \ldots \diff y_n.
$$
In this case, when $f_{\rvx}$ is continuous at $\bx=[x_1,x_2, \ldots,x_n]^\top$, we have 
$$
f_{\rvx}(x_1,x_2, \ldots,x_n) =\frac{\partial^n}{\partial x_1 \partial x_2\ldots \partial x_n} F_{\rvx}(x_1,x_2, \ldots,x_n).
$$
The random variables are independent if and only if 
$$
\begin{aligned}
F_{\rvx}(x_1,x_2, \ldots,x_n) &= F_{\rx_1}(x_1) \cdot F_{\rx_2}(x_2)\cdot\ldots \cdot F_{\rx_n}(x_n);\\
\text{or} \quad f_{\rvx}(x_1,x_2, \ldots,x_n) &= f_{\rx_1}(x_1) \cdot f_{\rx_2}(x_2)\cdot\ldots \cdot f_{\rx_n}(x_n).
\end{aligned}
$$
The \textit{conditional or marginal probability density functions} for the continuous cases are analogous to those in the discrete case, except that integration is employed instead of summation. For example, $f_{\rx}(x) = \int f_{\rx,\ry}(x, y)\diff y$. 
See the following sections for more details.
\end{itemize}

\index{Covariance}
\index{Correlation}
Specifically, the joint distribution of two random variables $\rx_i$ and $\rx_j$ is given by the joint cumulative distribution function $F(x_i, x_j) \equiv \prob[\rx_i\leq x_i, \rx_j\leq x_j]$. Then the covariance  between $\rx_i$ and $\rx_j$, denoted $\omega_{ij}$,  is defined as
$$
\omega_{ij} \triangleq \Cov[\rx_i, \rx_j] \triangleq \Exp[(\rx_i - \mu_i)(\rx_j - \mu_j)] 
= \int_{x_i, x_j = -\infty}^{\infty} (x_i - \mu_i)(x_j - \mu_j) \diff F(x_i, x_j),
$$
where $\mu_i = \Exp[\rx_i]$.
Equivalently, $\omega_{ij} = \Exp[\rx_i \rx_j] - \mu_i \mu_j$.

The variance of two random variables quantifies the degree of linear dependence between the two.
A closely related measure is the \textit{correlation}, defined as
\begin{equation}\label{equation:correlation_def}
\Corr[\rx_1, \rx_2] \triangleq \frac{\Cov[\rx_1, \rx_2]}{\sqrt{\Var[\rx_1]\Var[\rx_2]}}.
\end{equation}
While covariance and correlation both capture linear dependence, correlation is scale-invariant and always lies in the interval $[-1,1]$.
It also holds that 
\begin{equation}
\abs{\Corr[\rx_1, \rx_2]} \leq \sqrt{\Var[\rx_1]\Var[\rx_2]}.
\end{equation}

\subsection*{Random Vectors}
For a (continuous) RV $\rvx\in\real^D$, 
the expectation corresponds to the constant vector that minimizes the mean squared error with respect to $\rvx$. This least-squares characterization is formally stated as
\begin{equation}\label{equation:exp_mse}
\Exp[\rvx] 
= \argmin_{\by \in \real^D} \int \normtwo{\bx - \by}^2 p_\rvx(\bx) \diff \bx 
= \int \bx p_\rvx(\bx) \diff \bx. 
\end{equation}
A fundamental tool for computing expectations of transformed random variables is the \textit{law of the unconscious statistician (LOTUS)} (see Problem~\ref{prob:lotus}), which states
\begin{equation}\label{equation:lotus_exp_F}
\Exp[G(\rvx)] = \int G(\bx) p_\rvx(\bx) \diff \bx. 
\end{equation}
When clarification is needed, we explicitly specify the underlying distribution by writing $\Exp_\rvx [G(\rvx)]$ or $\Exp_{p(\bx)} [G(\rvx)]$.
For a random vector $\rvx\in\real^D$ with mean $\bmu = \Exp[\rvx]$, the covariance matrix $\bOmega \in \real^{D \times D}$ is defined as
\begin{equation}
\Cov[\rvx] \triangleq \bOmega = \Exp[(\rvx - \bmu)(\rvx - \bmu)^\top] = \Exp(\rvx\rvx^\top) - \bmu \bmu^\top.
\end{equation}

\subsection*{Distribution Transformation}
We now present key results for the expectation and covariance of linearly and quadratically transformed random vectors.
\begin{lemma}[Linear transformation]\label{lemma:lin_tran_prob}
Let $\rvx \in \real^D$ be a random vector with mean $\bmu\in\real^D$ and covariance matrix $\bOmega\in\real^{D\times D}$. Define the affine transformation $\rvy = \bA\rvx +\bb$, where $\bA \in \real^{M \times D}$ and $\bb\in\real^M$ are deterministic constant matrix and vector, respectively. The transformed mean and covariance satisfy
$$
\Exp[\rvy] = \bA\bmu+\bb 
\qquad\text{and}\qquad  
\Cov[\rvy] = \bA\bOmega\bA^\top.
$$
\end{lemma}
\begin{proof}[of Lemma~\ref{lemma:lin_tran_prob}]
The formula for the mean follows directly from the linearity of expectation. For the covariance, we substitute the definition of the covariance matrix:
$$
\Cov[\rvy] = \Cov[\bA\rvx] = \Exp[\bA(\rvx - \bmu)(\rvx - \bmu)^\top \bA^\top ]= \bA \Exp[(\rvx - \bmu)(\rvx - \bmu)^\top] \bA^\top = \bA\bOmega\bA^\top.  
$$
This completes the proof.
\end{proof}
As a special case, let $\bA=\ba^\top$ be a row vector, so that $\rvy = \ba^\top\rvx$ defines a linear functional of $\rvx$. If $\rvx$ has uncorrelated entries with identical variance, i.e., $\Cov[\rvx] = \sigma^2 \bI$, then the variance of the transformed scalar random variable simplifies to $\Cov[\rvy] = \sigma^2 \ba^\top \ba$.

\begin{lemma}[Quadratic transformation]\label{lemma:quad_tra_prob}
Let $\bA \in \real^{D \times D}$ be a symmetric matrix, and let $\rvx\in\real^D$ be a random vector with mean $\bmu=[\mu_i]\in\real^D$ and covariance  $\bOmega=[\omega_{ij}]\in\real^{D\times D}$. 
Then
$$
\Exp[\rvx^\top\bA\rvx] = \bmu^\top \bA \bmu + \trace(\bA\bOmega),
$$
where $\trace(\bA\bOmega)$ denotes the trace of $\bA\bOmega$, i.e., the sum of its diagonal entries.
\end{lemma}
\index{Trace trick}
\index{Quadratic expectation}
\begin{proof}[of Lemma~\ref{lemma:quad_tra_prob}]
Since $\rvx^\top \bA \rvx = \sum_{i=1}^D \sum_{j=1}^D a_{ij} \rx_i \rx_j$, 
it follows that  $\Exp[\rvx^\top \bA \rvx] = \sum_{i=1}^D \sum_{j=1}^D a_{ij} \Exp[\rx_i \rx_j]$.
Substitute the expectations $ \Exp[\rx_i \rx_j] = \mu_i \mu_j + \omega_{ij} $:
$$
\Exp[\rvx^\top \bA \rvx] 
= {\sum_{i=1}^D \sum_{j=1}^D a_{ij} \mu_i \mu_j} + {\sum_{i=1}^D \sum_{j=1}^D a_{ij} \omega_{ij}} = 
\bmu^\top \bA \bmu + \trace(\bA\bOmega).
$$
This completes the proof.
\end{proof}
In many practical settings, random vectors are centered such that $\bmu=\bzero$. Under zero mean, the quadratic expectation simplifies to $\Exp[\rvx^\top\bA\rvx ] = \trace(\bA\bOmega)$.
If, further, $\bOmega=\sigma^2\bSigma$, then $\Exp[\rvx^\top\bA\rvx ] = \sigma^2\trace(\bA\bSigma)$.
To obtain an unbiased estimator of $\sigma^2$, we can choose a matrix $\bA$ such that $\trace(\bA\bSigma) = 1$ \citep{lu2021rigorous}.

\begin{SCfigure}
\centering
\includegraphics[width=0.5\textwidth]{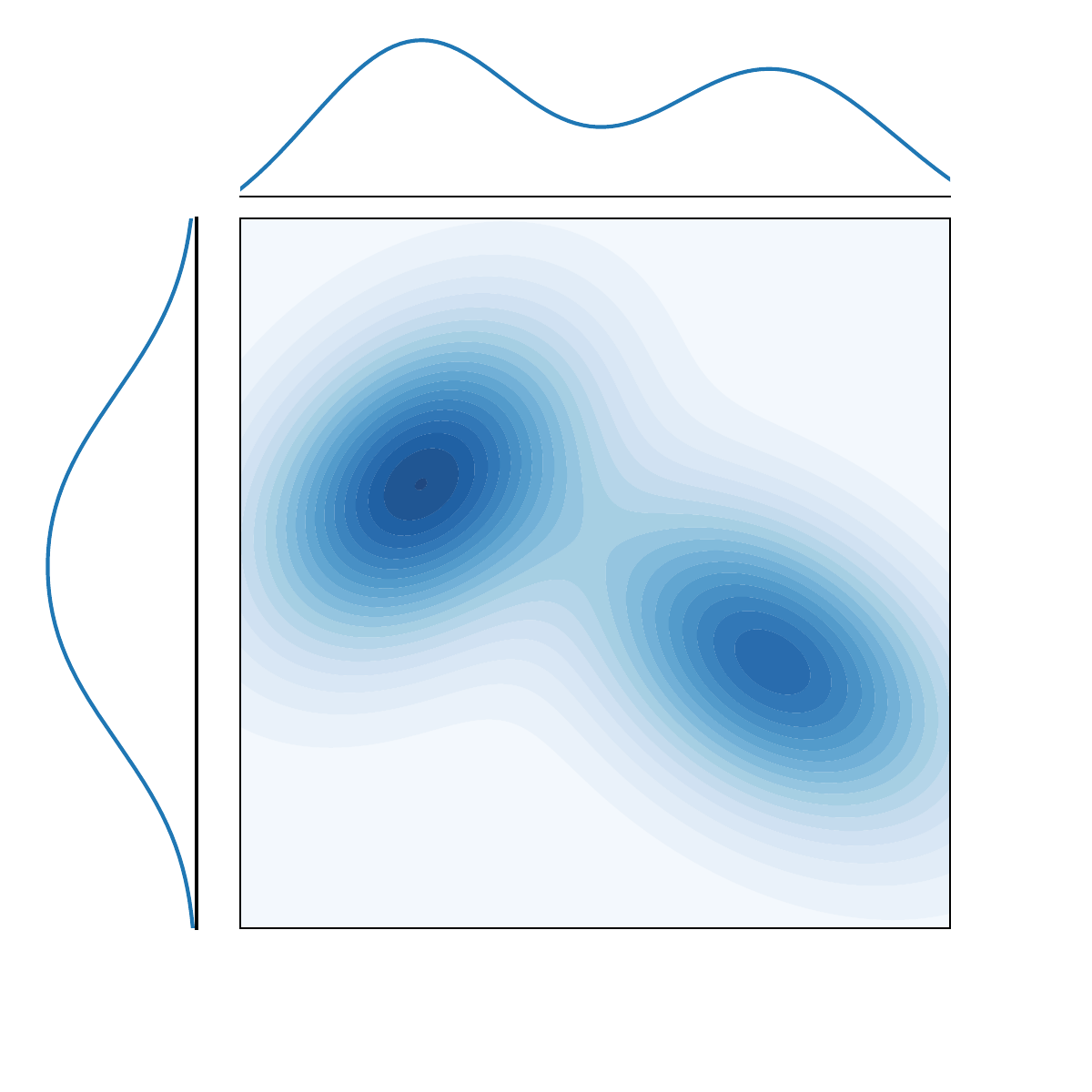}
\caption{
Joint PDF for a 2-D distribution $p_{\rx,\ry}$ (shown in shades of blue) and its marginal distributions $p_\rx$ and $p_\ry$ (shown as blue lines).
}
\label{fig:marginal_prob}
\end{SCfigure}

\index{Unbiasedness property}
\index{Conditional expectation formula}
\subsection*{Conditional Densities and Expectations}
Extending the discrete conditional probability definition in \eqref{equation:bayes_base} to continuous settings, consider two continuous RVs $\rvx, \rvy \in \real^D$. Their joint probability density function $p_{\rvx,\rvy}(\bx,\by)$ yields the corresponding marginal PDFs via integration:
\begin{equation}\label{equation:cond_rvx}
\int p_{\rvx,\rvy}(\bx,\by) \diff\by = p_\rvx(\bx) 
\quad \text{ and }\quad
\int p_{\rvx,\rvy}(\bx,\by) \diff \bx = p_\rvy(\by). 
\end{equation}
Figure~\ref{fig:marginal_prob} illustrates the joint PDF for two scalar random variables ($D=1$). For any fixed outcome $\by$ satisfying $p_\rvy(\by) > 0$, the conditional probability density function $p_{\rvx\mid\rvy}$ characterizes the distribution of $\rvx$ conditioned on the event $\rvy=\by$, and is formally defined as:
\begin{equation}\label{equation:cond_rvx_rvy}
p_{\rvx\mid\rvy}(\bx\mid \by) 
\triangleq \frac{p_{\rvx,\rvy}(\bx,\by)}{p_\rvy(\by)}.
\end{equation}
The conditional PDF $p_{\rvy\mid\rvx}$ is defined symmetrically. From this foundation, Bayes' rule \eqref{equation:bayes_base} generalizes to continuous densities, reversing the conditional relationship via
\begin{equation}
p_{\rvy\mid\rvx}(\by\mid \bx) = \frac{p_{\rvx\mid\rvy}(\bx\mid \by)p_\rvy(\by)}{p_\rvx(\bx)},
\end{equation}
which holds for all $\bx$ such that $p_\rvx(\bx) > 0$.

The conditional expectation $\Exp[\rvx\mid\rvy]$ admits a fundamental least-squares optimality property: it corresponds to the deterministic function $G_*(\rvy)$ that best approximates $\rvx$ while only using information from $\rvy$. Formally, this minimizer is defined as:
\begin{align}
G_* &
\triangleq \argmin_{G:\real^D \to \real^D} \Exp\left[\normtwo{\rvx - G(\rvy)}^2\right] 
= \argmin_{G:\real^D \to \real^D} \int \normtwo{\bx - G(\by)}^2 p_{\rvx,\rvy}(\bx,\by) \diff \bx \diff \by \nonumber \\
&= \argmin_{G:\real^D \to \real^D} \int \left[ \int \normtwo{\bx - G(\by)}^2 p_{\rvx\mid\rvy}(\bx\mid \by) \diff \bx \right] p_\rvy(\by) \diff \by.\label{equation:cond_exp}
\end{align}
For every $\by \in \real^D$ with $p_\rvy(\by) > 0$, the minimizer of the inner integral yields the pointwise conditional expectation function:
\begin{equation}\label{equation:cond_exp_rv}
\Exp[\rvx\mid\rvy=\by] \triangleq G_*(\by) 
= \int \bx p_{\rvx\mid\rvy}(\bx\mid \by) \diff \bx,
\end{equation}
where the optimization step mirrors the least-squares characterization of standard expectation in \eqref{equation:exp_mse}. Substituting the random vector $\rvy$ in place of its deterministic outcome $\by$ gives the random conditional expectation:
\begin{equation}
\Exp[\rvx\mid\rvy] \triangleq G_*(\rvy),
\end{equation}
which is itself a $\real^D$-valued random variable. 
A critical notational distinction must be emphasized: although both quantities are referred to as conditional expectation in standard literature, they are mathematically distinct objects. Specifically, $\Exp[\rvx\mid\rvy=\by]$ is a deterministic function mapping $\real^D$ to $\real^D$, whereas $\Exp[\rvx\mid\rvy]$ is a random variable. This text strictly adheres to the above notation to eliminate ambiguity.

A core and widely used property of conditional expectation is the \textit{unbiasedness property}, which simplifies hierarchical expectation calculations for two RVs $\rvx$ and $\rvy$:
\begin{equation}\label{equation:tower_property}
\Exp\left[\Exp[\rvx\mid \rvy]\right] = \Exp[\rvx].
\end{equation}
Since $\Exp[\rvx\mid\rvy]$ is a random variable dependent solely on $\rvy$, the outer expectation averages out the randomness of $\rvy$ to recover the marginal expectation of $\rvx$. This identity can be rigorously verified via the preceding definitions:
\begin{align*}
\Exp\left[\Exp[\rvx\mid\rvy]\right] 
&= \int \left( \int \bx p_{\rvx\mid\rvy}(\bx\mid \by) \diff \bx \right) p_\rvy(\by) \diff \by 
\stackrel{\eqref{equation:cond_rvx_rvy}}{=} \int \int \bx p_{\rvx,\rvy}(\bx,\by) \diff \bx \diff \by \\
&\stackrel{\eqref{equation:cond_rvx}}{=} \int \bx p_\rvx(\bx) \diff \bx 
= \Exp[\rvx].
\end{align*}
A generalized version of this result is presented in Problem~\ref{prob:prop_expcond}.
Finally, we state a general conditional expectation formula for functions of multiple random variables. For any arbitrary function $G(\rvx, \rvy)$ of the random vectors $\rvx$ and $\rvy$, combining the LOTUS \eqref{equation:lotus_exp_F} with the conditional expectation definition \eqref{equation:cond_exp_rv} yields the \textit{conditional expectation formula}:
\begin{equation}\label{equation:law_uncons_stat}
\Exp[G(\rvx,\rvy)\mid \rvy=\by] = \int G(\bx,\by) p_{\rvx\mid\rvy}(\bx\mid \by) \diff \bx.
\end{equation}

\subsection{Gaussian Distribution and Properties}
The \textit{Gaussian}  (or \textit{normal}) distribution is one of the most widely adopted statistical models in data modeling, machine learning, and signal processing. Its popularity stems from a fundamental probabilistic result: the sum of many independent random variables asymptotically follows a Gaussian distribution, a property rigorously established by the central limit theorem.
\footnote{See, for example, \citet{lu2026first}.} 
Consequently, any measurement or macroscopic quantity composed of numerous independent additive microscopic components tends to be approximately Gaussian. For this reason, observational and sensor noise in engineering and signal processing is routinely modeled as Gaussian noise, owing to its favorable analytical properties and mathematical tractability.

The \textit{multivariate Gaussian distribution} (also referred to as the \textit{multivariate normal (MVN) distribution}) generalizes the univariate Gaussian to vector-valued random variables and characterizes the joint distribution of a set of normally distributed random components. A multivariate Gaussian distribution is fully specified by two finite-dimensional parameters: a mean vector and a positive-definite covariance matrix. The covariance matrix encodes the statistical behavior of the random vector, including individual component variances along its diagonal entries and pairwise linear correlation among components via its off-diagonal entries. Due to its analytical simplicity, closed-form algebraic properties, and wide empirical applicability, the multivariate Gaussian serves as a foundational building block throughout statistics, machine learning, and signal processing. We present its formal definition below.

\begin{definition}[Multivariate Gaussian distribution\index{Gaussian distribution}\index{Normal distribution}]\label{definition:multivariate_gaussian}
A random vector $\rvx \in \real^D$ is said to follow the \textit{multivariate Gaussian distribution (multivariate normal, MVN)} with parameters $\bmu\in\real^D$ and $\bSigma\in\real^{D\times D}$, denoted  $\rvx\sim \normal(\bmu, \bSigma)$, if
$$
\begin{aligned}
p(\bx; \bmu, \bSigma)&= (2\pi)^{-D/2} \abs{\bSigma}^{-1/2}\exp\left\{-\frac{1}{2}(\bx - \bmu)^\top \bSigma^{-1}(\bx - \bmu)\right\},~\footnote{The form of which can be proved using the moment generating function of $D$ i.i.d. univariate standard Gaussian variables.}
\end{aligned}
$$
where $\bmu \in \real^D$ is called the \textit{mean vector}, and $\bSigma\in \real^{D\times D}$ is positive definite and is called the \textit{covariance matrix}. $\abs{\bSigma} = \det(\bSigma)$ denotes the determinant of  $\bSigma$.
The mean, mode, and covariance of the multivariate Gaussian distribution are given by 
\begin{equation*}
\begin{aligned}
\Exp [\rvx] &= \bmu, \qquad 
\mathrm{Mode}[\rvx] = \bmu, \qquad\text{and}\qquad 
\Cov [\rvx] = \bSigma. 
\end{aligned}
\end{equation*}
The covariance matrix can also be expressed as:
$$
\Cov[\rvx]=\Exp[(\rvx-\bmu)(\rvx-\bmu)^\top]=\Exp[\rvx\rvx^\top]-\bmu\bmu^\top.
$$
\end{definition}

\begin{figure}[h]
\subfigure[Gaussian, $\bSigma =\begin{bmatrix}
1&0\\
0&1
\end{bmatrix}. $ ]{\includegraphics[width=0.31
\textwidth]{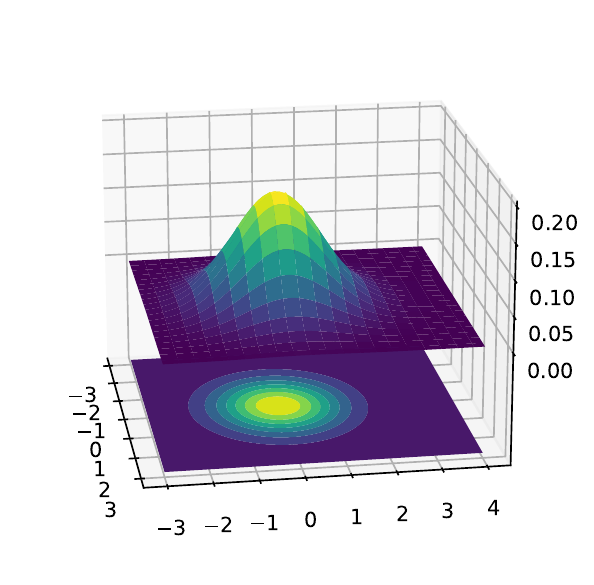} \label{fig:dists_multiGauss_sigma1}}
\subfigure[Gaussian, $\bSigma =\begin{bmatrix}
1&0\\
0&3
\end{bmatrix}.$]{\includegraphics[width=0.31
\textwidth]{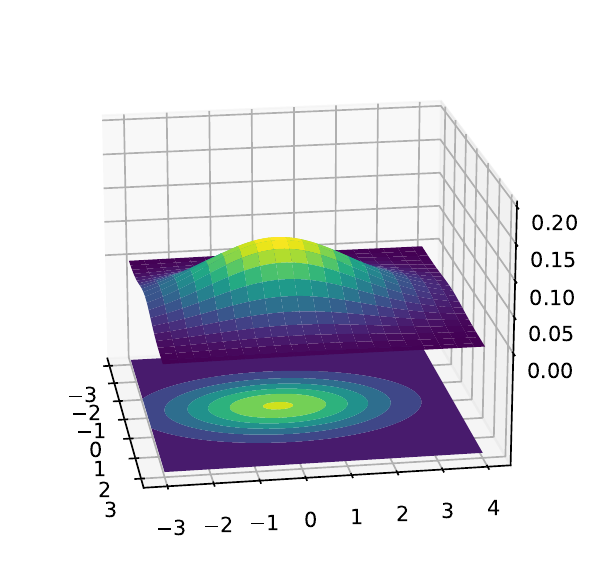} \label{fig:dists_multiGauss_sigma2}}
\subfigure[Gaussian, $\bSigma =\begin{bmatrix}
1&\textendash0.5\\
\textendash0.5&1.5
\end{bmatrix}.$]{\includegraphics[width=0.31 
\textwidth]{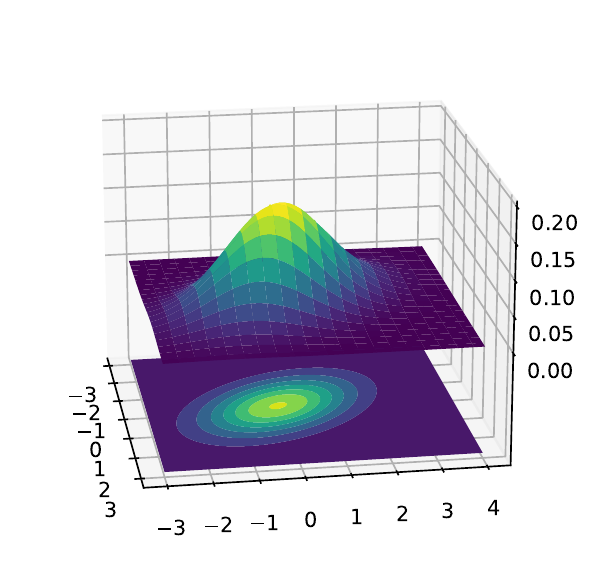} \label{fig:dists_multiGauss_sigma3}}
\subfigure[Gaussian, $\bSigma =\begin{bmatrix}
2&0\\
0&2
\end{bmatrix}. $ ]{\includegraphics[width=0.31
\textwidth]{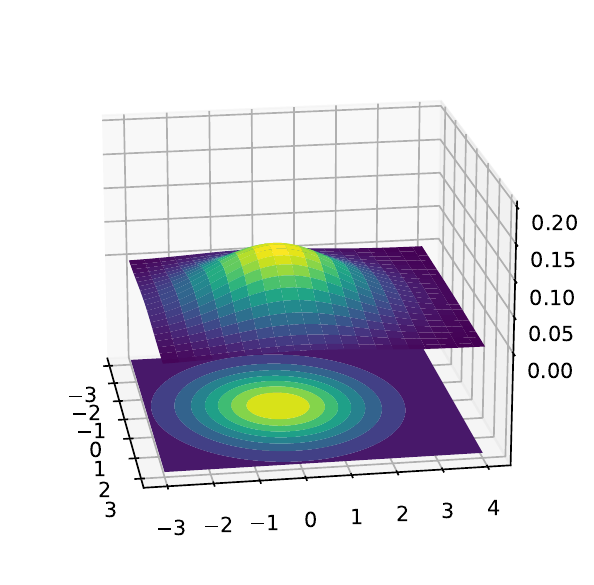} \label{fig:dists_multiGauss_sigma4}}
\subfigure[Gaussian, $\bSigma =\begin{bmatrix}
3&0\\
0&1
\end{bmatrix}.$]{\includegraphics[width=0.31
\textwidth]{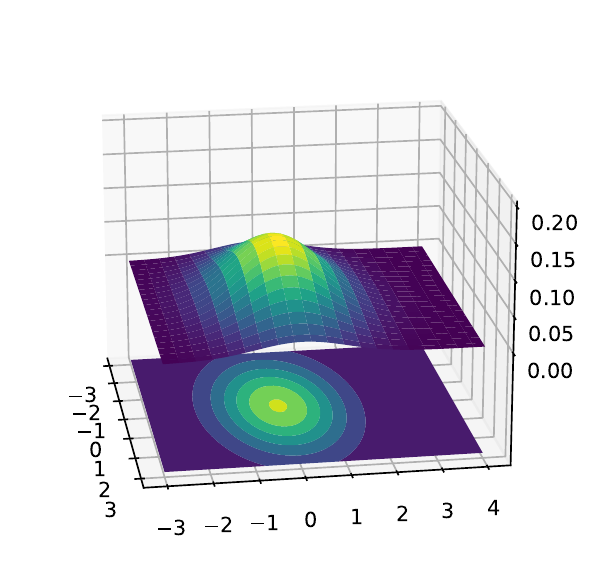} \label{fig:dists_multiGauss_sigma5}}
\subfigure[Gaussian, $\bSigma =\begin{bmatrix}
3&\textendash0.5\\
\textendash0.5&1.5
\end{bmatrix}.$]{\includegraphics[width=0.31
\textwidth]{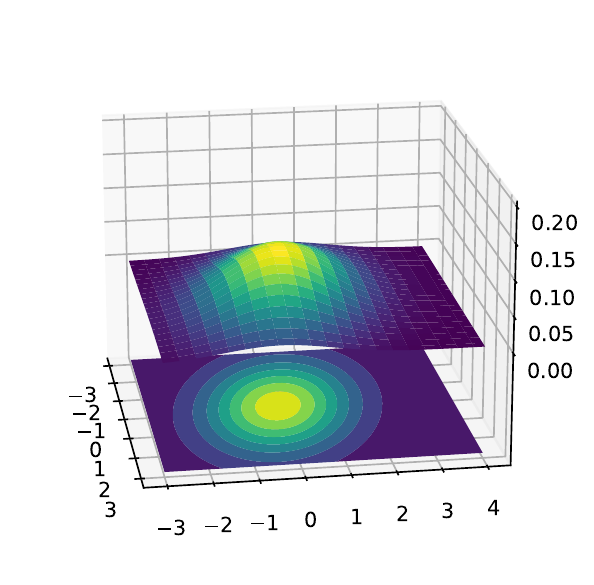} \label{fig:dists_multiGauss_sigma6}}
\centering
\caption{Density and contour plots (\textcolor{mydarkblue}{blue}=low, \textcolor{mydarkyellow}{yellow}=high) of the multivariate Gaussian distribution over the $\real^2$ space for various values of the covariance/scale matrix with a zero-mean vector.  Fig~\ref{fig:dists_multiGauss_sigma1} and \ref{fig:dists_multiGauss_sigma4}: A spherical covariance matrix has a circular shape; 
Fig~\ref{fig:dists_multiGauss_sigma2} and \ref{fig:dists_multiGauss_sigma5}:  A diagonal covariance matrix yields axis-aligned elliptical contours; 
Fig~\ref{fig:dists_multiGauss_sigma3} and \ref{fig:dists_multiGauss_sigma6}: A full covariance matrix results in rotated elliptical contours.}
\centering
\label{fig:multi_gaussian_density}
\end{figure}

Figure~\ref{fig:multi_gaussian_density} visualizes multivariate Gaussian density contours under different covariance structures. A zero-mean Gaussian random vector with identity covariance matrix consists of independent and identically distributed (i.i.d.) standard Gaussian components with zero mean and unit variance. More generally, arbitrary multivariate Gaussian vectors can be constructed via linear transformation of univariate Gaussian samples; see Problem~\ref{problem:multiGauss} for detailed construction.

\subsection*{Properties of Multivariate Gaussian Distribution}\label{section:multi_gauss}

A defining property of Gaussian random vectors is that \textit{affine transformations preserve Gaussianity}. This invariance property can be derived either via the linear transformation rule for probability distributions (Lemma~\ref{lemma:lin_tran_prob}) or using the moment generating function of the multivariate Gaussian distribution \citep{lu2026first}.

\begin{lemma}[Affine transformation of multivariate Gaussian distribution]\label{lemma:affine_mult_gauss}
Let $\rvx\sim \normal(\bmu_x, \bSigma_x)$ and $\rvy\sim \normal(\bmu_y, \bSigma_y)$ be independent $D$-dimensional Gaussian random vectors. For deterministic matrices $\bA,\bB\in\real^{N\times D}$ and deterministic vector $\bc\in\real^N$, the affine combination follows a multivariate Gaussian distribution:
$$
\rvz=\bA\rvx+\bB\rvy +\bc \sim \normal(\bA\bmu_x+\bB\bmu_y+\bc, \bA\bSigma_x\bA^\top +\bB\bSigma_y\bB^\top).
$$
Furthermore, for any fixed vector $\bd\in\real^D$, the linear combination $\bd^\top\rvx$ yields a univariate Gaussian random variable:
$$
\bd^\top\rvx \sim \normal(\bd^\top\bmu_x, \bd^\top\bSigma_x\bd).
$$
\end{lemma}

This result also generalizes to the \textit{sum of independent Gaussian vectors}. 
For a collection of independent Gaussian vectors satisfying $\rvx_i\sim\normal(\bmu_i,\bSigma_i)$ for all $i\in\{1,2,\dots,N\}$, their component-wise sum remains Gaussian:
$$
\sum_{i=1}^{N} \rvx_i \sim \normal\big(\sum_{i=1}^{N}\bmu_i, \sum_{i=1}^{N}\bSigma_i\big)
\gap 
\text{if } \rvx_i\sim\normal(\bmu_i,\bSigma_i), \ \forall\, i\in\{1,2,\ldots,N\}.
$$
A notable special case arises when $\bA=\be_i^\top$, the $i$-th standard basis vector of $\real^D$. The resulting projection extracts the $i$-th entry of $\rvx$:
$$
\rx_i = \be_i^\top\rvx \sim \normal\left(\mu_{x,i},\, \sigma_{x,ii}^2\right),
$$
where $\mu_{x,i}$ denotes the $i$-th entry of the mean vector $\bmu_x$, and $\sigma_{x,ii}^2$ denotes the $i$-th diagonal entry of the covariance matrix $\bSigma_x$. This confirms that every marginal component of a multivariate Gaussian vector follows a univariate Gaussian distribution.

Two critical implications follow directly from Lemma~\ref{lemma:affine_mult_gauss}. 
First, any subvector of a multivariate Gaussian random vector is itself Gaussian. 
Second, an i.i.d. standard Gaussian vector is \textit{isotropic}, meaning its probability distribution is invariant under arbitrary orthogonal transformations.
Formally, consider an i.i.d. standard Gaussian vector $\rvx$. For any orthogonal matrix $\bQ$ (satisfying $\bQ^\top\bQ=\bQ\bQ^\top=\bI$), the transformed vector $\bQ\rvx$ preserves the original distribution. This holds because the transformed mean is $\bQ\bzero = \bzero$ and the transformed covariance matrix is $\bQ\bI\bQ^\top = \bI$. Importantly, isotropy is a stronger property than identical marginal variances: it ensures the entire distribution (not merely second-order moments) remains unchanged under rotation.

\begin{lemma}[Rotations on  multivariate Gaussian distribution]\label{lemma:rotat_multi_gauss}
Multivariate Gaussian distributions with spherical covariance matrices are rotationally invariant. Specifically, for any orthogonal matrix $\bQ$, if $\rvx\sim \normal(\bzero, \sigma^2\bI)$, then the rotated vector satisfies $\bQ\rvx\sim \normal(\bzero, \sigma^2\bI)$.
\end{lemma}

\paragrapharrow{``Standardization and decorrelation."}
A random vector following the zero-mean identity-covariance Gaussian distribution $\normal(\bzero, \bI)$ is termed an \textit{i.i.d. standard Gaussian vector}. Arbitrary multivariate Gaussian vectors can be \textit{standardized} and \textit{decorrelated} via \textit{covariance whitening}. 
For any $\rvx\sim\normal(\bmu, \bSigma)$, the standardized vector follows the standard Gaussian distribution:
\begin{equation}\label{equation:std_mugau_recov}
\rvx\sim\normal(\bmu, \bSigma)
\quad\implies \quad
\rvz =\bSigma^{-1/2}(\rvx-\bmu) \sim \normal(\bzero,\bI).
\end{equation}
Conversely, any general multivariate Gaussian vector can be reconstructed from a standard Gaussian vector via scaling and shifting: if $\rvx\sim\normal(\bmu,\bSigma)$, then 
\begin{equation}\label{equation:gauss_stand}
\rvx=\bmu+\bSigma^{1/2}\bepsilon,
\gap \text{where }\bepsilon\sim\normal(\bzero,\bI).
\end{equation}

Consider $N$ independent and identically distributed samples $\{\bx_1,\bx_2,\dots,\bx_N\}$ drawn from $\normal(\bmu,\bSigma)$, and define the sample mean $\overline{\bx} = \frac{1}{N} \sum_{i=1}^{N} \bx_i$. The scaled sample mean satisfies the Gaussian convergence property:
\begin{equation}\label{equation:mean_mutigau}
\sqrt{N} (\overline{\bx}-\bmu) \sim \normal(\bzero, \bSigma).
\end{equation}

\begin{lemma}[Uncorrelation implies mutual independence for Gaussian random vectors]\label{lemma:uncor_inde_mvn}
For a Gaussian random vector, pairwise uncorrelation among all components guarantees their mutual independence.
\end{lemma}
This property is unique to Gaussian distributions and follows directly from the structure of the multivariate Gaussian PDF. When the covariance matrix is diagonal, all cross-covariance terms vanish, and the joint probability density function factorizes exactly into the product of univariate marginal densities, which is the defining condition for mutual independence.

We now formalize this result for partitioned Gaussian vectors. Let $\rvx \sim \normal(\bmu, \bSigma)$ with $\bmu\in\real^D$, and partition the random vector and its parameters into two subvectors and corresponding submatrices:
$$
\begin{bmatrix}
\rvx_1 \\
\rvx_2
\end{bmatrix}
\sim \normal(\bmu, \bSigma) = \normal
\left(
\begin{bmatrix}
\bmu_1 \\
\bmu_2
\end{bmatrix},
\begin{bmatrix}
\bSigma_{11} & \bSigma_{12} \\
\bSigma_{21} & \bSigma_{22}
\end{bmatrix}
\right).
$$
The subvectors $\rvx_1$ and $\rvx_2$ are mutually independent if and only if the cross-covariance block satisfies $\bSigma_{12} = \bzero$.
More generally, consider a full $D$-dimensional Gaussian vector $\rvx = [\rx_1, \rx_2, \dots, \rx_D]^\top \sim \normal(\bmu, \bSigma)$. Its scalar components are mutually independent if and only if the covariance matrix is purely diagonal:
\begin{equation}\label{equation:iid_mulg_iff7}
\text{the $\rx_i$'s are mutually independent if and only if $\bSigma$ is diagonal.}
\end{equation}
This can be rigorously established as follows:
\begin{proof}[of Lemma~\ref{lemma:uncor_inde_mvn} and Equation~\eqref{equation:iid_mulg_iff7}]
First, suppose all components $\rx_1,\rx_2,\dots,\rx_D$ are mutually independent. Each component follows a univariate Gaussian distribution $\rx_i \sim \normal(\mu_i, \sigma_i^2)$. By independence, the joint PDF factorizes as the product of marginal PDFs:
\begin{align*}
p_{\rvx}(\bx)\ &= \prod_{i=1}^D p_{\rx_i}(x_i) = \prod_{i=1}^D \frac{1}{\sigma_i \sqrt{2\pi}} \exp \left\{ -\frac{1}{2} \frac{(x_i - \mu_i)^2}{\sigma_i^2} \right\} \\
&= \frac{1}{(2\pi)^{D/2} \abs{\diag(\sigma_1^2, \ldots, \sigma_D^2)}^{1/2}} \exp \left\{ -\frac{1}{2} (\bx - \bmu)^\top \diag(\sigma_1^{-2}, \ldots, \sigma_D^{-2}) (\bx - \bmu) \right\}.
\end{align*}
This form corresponds exactly to a multivariate Gaussian distribution with diagonal covariance $\bSigma = \diag(\sigma_1^2,\dots,\sigma_D^2)$.

Conversely, assume $\bSigma$ is diagonal. Reversing the factorization above shows that the joint PDF decomposes into a product of individual marginal PDFs, which implies mutual independence of all scalar components.
\end{proof}

\index{Independence}
This fundamental property yields the following useful corollary concerning linear transformations of Gaussian vectors.
\begin{corollary}[Independence of linear combinations in Gaussian Distributions]
Let $ \rvx \sim \normal(\bmu, \bSigma)$ be a random vector in $\real^{D} $ , and let $ \bA\in\real^{M \times D} $, $ \bB\in\real^{N \times D} $ be fixed matrices. 
Then,
\begin{equation}
\text{$\bA\rvx$ \text{ is independent of } $\bB\rvx$ $\quad\iff\quad$ $\bA\bSigma \bB^\top = \bzero$.}
\end{equation}
\end{corollary}
This result can also be proven via moment generating function properties of multivariate Gaussian distributions; we omit the detailed derivation here.

In addition, standard probabilistic invariance guarantees that measurable functions of independent random vectors remain independent. Specifically, if $\rvx_1$ and $\rvx_2$ are independent, then $G_1(\rvx_1)$ and $G_2(\rvx_2)$ are independent for any pair of measurable functions $G_1, G_2$.

\index{Continuously differentiability}
\index{Second-order partial derivative}
\subsection{Differentiable Functions and Differential Calculus}\label{section:differ_calc}
Differentiability and differential calculus form the foundational framework of mathematical analysis for multi-dimensional functions. This subsection introduces core differentiation concepts that characterize how multivariate functions change with their inputs, providing rigorous tools for function analysis, optimization, and system modeling.

Central to multivariate calculus is the notion of the directional derivative, which quantifies the instantaneous rate of change of a function $G$ at a point $\bx$ along a specified direction vector $\bd$.

\begin{definition}[Directional derivative, partial derivative\index{G\^ateaux derivative}\index{Directional derivative}\index{Partial derivative}]\label{definition:partial_deri}
Consider a function $G$ defined on a domain $\sS\subseteq \real^D$, and let $\bd\in\real^D$ be a nonzero direction vector. The directional derivative of $G$ at $\bx$ along direction $\bd$ is defined as
$$
\mathop{\lim}_{\mu\rightarrow 0}
\frac{G(\bx+\mu\bd) - G(\bx)}{\mu},
$$
provided the limit exists. 
This quantity is also denoted $G^\prime(\bx; \bd)$ or $D_{\bd}G(\bx)$. 
The directional derivative is sometimes called the  \textit{G\^ateaux derivative}.

For each index $i\in\{1,2,\dots,D\}$, the directional derivative evaluated along the $i$-th standard basis vector $\be_i$ (if the limit exists) yields the $i$-th \textit{partial derivative}. Common notations include $\frac{\partial G}{\partial x_i}(\bx)$, $D_{\be_i}G(\bx)$, and $\partial_i G(\bx)$.
\end{definition}

\begin{figure*}[h]
\centering  
\subfigtopskip=2pt 
\subfigbottomskip=9pt 
\subfigcapskip=-5pt 
\includegraphics[width=0.75\textwidth]{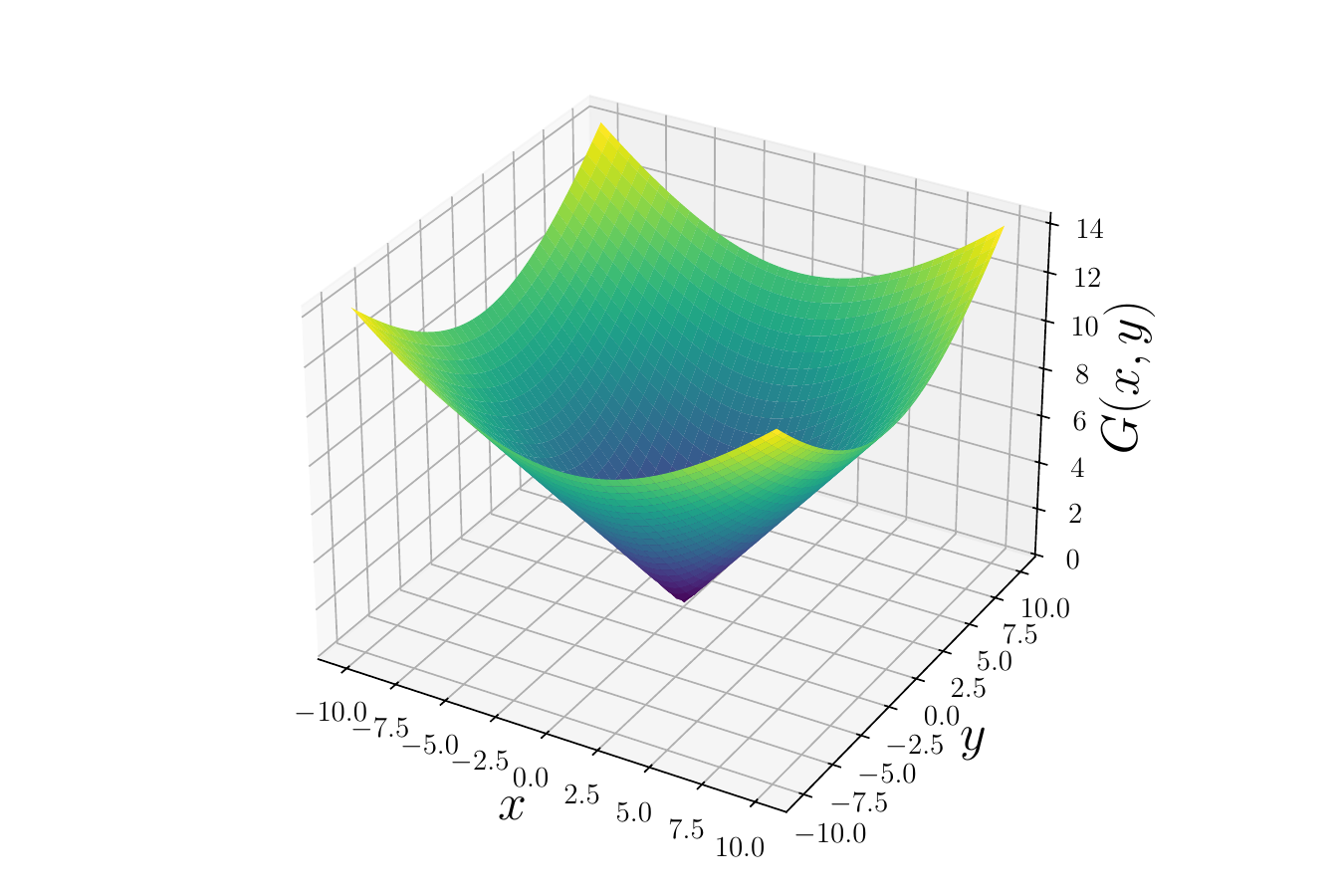}
\caption{Plot for the function $G(x, y) = \sqrt{x^2+y^2}$, in which case any directional derivative for the direction $\bd=[a,b]^\top$ with $a\neq 0$ and $b\neq 0$ at point $[0,0]^\top$ exists. However, the partial derivatives at this point do not exist.}
\label{fig:direc_deri_no_partial}
\end{figure*}

An important subtlety in multivariate differentiation is that a function may admit well-defined directional derivatives in every direction at a given point, yet still lack valid partial derivatives at that point. We illustrate this phenomenon using the function $G(x,y)=\sqrt{x^2+y^2}$ at the origin (Figure~\ref{fig:direc_deri_no_partial}).
For any direction $\bd=[a,b]^\top$ with both entries nonzero, the directional derivative at $(0,0)$ is computed as
$$
G^\prime(0,0; \bd) = \mathop{\lim}_{\mu\rightarrow 0}
\frac{\sqrt{(0+\mu a)^2+(0+\mu b)^2} -\sqrt{0^2+0^2} }{\mu} = \sqrt{a^2+b^2}.
$$
In particular, if $\bd$ is a unit vector, the directional derivative equals $1$.
In contrast, the coordinate partial derivatives at the origin do not exist. The partial derivatives with respect to $x$ and $y$ are
$$
\begin{aligned}
\frac{\partial G}{\partial x}(0,0) &= \mathop{\lim}_{h\rightarrow 0} \frac{\sqrt{(0+h)^2+0^2}}{h} = \mathop{\lim}_{h\rightarrow 0} \frac{\abs{h}}{h},\\
\frac{\partial G}{\partial y}(0,0) &= \mathop{\lim}_{h\rightarrow 0} \frac{\sqrt{0^2+(0+h)^2}}{h} = \mathop{\lim}_{h\rightarrow 0} \frac{\abs{h}}{h}.
\end{aligned}
$$
The one-sided limits differ: the limit equals $1$ for $h\to 0^+$ and $-1$ for $h\to 0^-$. Since the two-sided limit does not converge, the partial derivatives are undefined at the origin.

When all partial derivatives of $G$ exist at a point $\bx\in\real^D$, the \textit{gradient} of $G$ at $\bx$ is defined as the column vector assembling all partial derivatives:
\begin{equation}
\nabla G(\bx)=
\begin{bmatrix}
\frac{\partial G}{\partial x_1} (\bx),
\frac{\partial G}{\partial x_2} (\bx),
\ldots,
\frac{\partial G}{\partial x_D} (\bx)	
\end{bmatrix}^\top
\in \real^D.
\end{equation}

\begin{exercise}
Prove the following two counterintuitive properties of multivariate functions:
\begin{enumerate}
\item A function may possess well-defined partial derivatives at a point yet fail to be continuous at that point.
\footnote{A function $G(\bx)$ is continuous at $\ba$ if $\lim_{\bx\to\ba} G(\bx) = G(\ba)$; a function is continuous overall if it is continuous at every point in its domain.}
\item A function may be continuous at a point yet lack well-defined partial derivatives at that point.
\end{enumerate}
\end{exercise}

A function $G$ defined on an open set $\sS\subseteq \real^D$  is called \textit{differentiable}
if all its partial derivatives exist (i.e., the derivatives exist in univariate cases, and the gradients exist in multivariate cases). This is actually the definition of \textit{Fr\'echet differentiability}.

\begin{definition}[(Fr\'echet) differentiability\index{(Fr\'echet) differentiability}]
Let $f:\real^D\rightarrow \real$. 
The function $G$ is \textit{(Fr\'echet) differentiable} at $\bx\in\real^D$ if there exists a vector $\bg\in\real^D$ such that
$$
\mathop{\lim}_{\bd\rightarrow\bzero} \frac{G(\bx+\bd)-G(\bx)-\bg^\top\bd}{\normtwo{\bd}}=0.
$$
The vector $\bg$ is unique and coincides with the gradient $\nabla G(\bx)$.
\end{definition}

\index{Continuously differentiable}
A function $G$ defined on an open set $\sS\subseteq \real^D$ is \textit{continuously differentiable} on $\sS$ if all of its first-order partial derivatives exist and are continuous throughout $\sS$. 

Importantly, differentiability does not imply continuous differentiability: a function can be differentiable everywhere while possessing discontinuous partial derivatives.
A classic one-dimensional counterexample demonstrates this distinction. Define
$$
G(x) = 
\begin{cases}
x^2 \sin(\frac{1}{x}), & \text{if } x \neq 0, \\
0, & \text{if } x = 0.
\end{cases}
$$
This function is differentiable over the entire real line. For all $x\neq 0$, $G(x)$ is a product of two differentiable functions and is therefore differentiable. At $x=0$, the derivative is verified via the limit definition:
$$
G^\prime(0) = \mathop{\lim}_{\mu\rightarrow 0}
\frac{G(\mu) - G(0) }{\mu}
=\mathop{\lim}_{\mu\rightarrow 0} \frac{\mu^2 \sin(\frac{1}{\mu}) - 0 }{\mu}
=\mu \sin(\frac{1}{\mu}) 
$$
The limit evaluates to zero because $\abs{\sin(1/\mu)}\leq 1$ for all $\mu\neq 0$, yielding a bounded oscillation that vanishes as $\mu\to 0$. Thus, $G'(0)=0$.
Nevertheless, $G(x)$ is not continuously differentiable at $x=0$. For $x\neq 0$, direct differentiation gives
$$
G^\prime(x) = 2x \sin(\frac{1}{x}) - \cos(\frac{1}{x}).
$$
As $x\to 0$, the term $\cos(1/x)$ oscillates indefinitely without convergence, so $\lim_{x\to 0}G'(x)$ does not exist. The first derivative is therefore discontinuous at the origin.

For any differentiable function, the directional derivative admits an exact closed-form expression in terms of the gradient:
\begin{equation}\label{equation:direc_contdiff}
G^\prime(\bx; \bd) = \nabla G(\bx)^\top \bd, \gap \text{for all }\bx\in\sS \text{ and }\bd\in\real^D.
\end{equation} 
\begin{proof}[of \eqref{equation:direc_contdiff}]
The identity holds trivially for $\bd=\bzero$. For nonzero $\bd$, differentiability implies
$$
0 = \lim_{\mu \to 0^+} \frac{G(\bx + \mu \bd) - G(\bx) - \langle \nabla G(\bx), \mu \bd \rangle}{\normtwo{\mu \bd}} 
= \lim_{\mu \to 0^+} \left[ \frac{G(\bx + \mu \bd) - G(\bx)}{\mu \normtwo{\bd}} - \frac{\langle \nabla G(\bx), \bd \rangle}{\normtwo{\bd}} \right].
$$
Factoring constants yields
$$
\begin{aligned}
&f'(\bx; \bd) = \lim_{\mu \to 0^+} \frac{G(\bx + \mu \bd) - G(\bx)}{\mu}\\
&= \lim_{\mu \to 0^+} \left\{ \normtwo{\bd} \left[ \frac{G(\bx + \mu \bd) - G(\bx)}{\mu \normtwo{\bd}} - \frac{\langle \nabla G(\bx), \bd \rangle}{\normtwo{\bd}} \right] + \langle \nabla G(\bx), \bd \rangle \right\}
&= \langle \nabla G(\bx), \bd \rangle.
\end{aligned}
$$
This proves \eqref{equation:direc_contdiff}.
\end{proof}

The definition of differentiability further implies the linear approximation property
\begin{equation}
\mathop{\lim}_{\bd\rightarrow \bzero}
\frac{G(\bx+\bd) - G(\bx) - \nabla G(\bx)^\top \bd}{\normtwo{\bd}} = 0,\gap 
\text{for all }\bx\in\sS,
\end{equation}
which is equivalently written in asymptotic form as
\begin{equation}
G(\by) = G(\bx)+\nabla G(\bx)^\top (\by-\bx) + o(\normtwo{\by-\bx}),
\end{equation}
where the \textit{small-oh} function $o(\cdot): \real_+\rightarrow \real$ is a one-dimensional function satisfying $\frac{o(\mu)}{\mu}\rightarrow 0$ as $\mu\rightarrow 0^+$.~\footnote{Note that we also use the standard \textit{big-Oh} notation to describe the asymptotic behavior of functions.
Specifically, the notation $g(\bd) = \mathcalO(\normtwo{\bd}^p)$ means that there are positive numbers
$C_1$ and $\delta$ such that $\abs{g(\bd)} \leq  C_1 \normtwo{\bd}^p$ for all $\normtwo{\bd}\leq\delta$. In practice it is
often equivalent to $\abs{g(\bd)} \approx C_2\normtwo{\bd}^p$ for  sufficiently small $\bd$, where $C_2$ is  another positive constant.
The \textit{soft-Oh} notation is employed to hide poly-logarithmic factors, i.e., $f = \widetilde{\mathcalO}(g)$ will
imply $f = \mathcalO(g \log^c(g))$ for some absolute constant $c$.}
This first-order approximation directly implies continuity: differentiable functions are always continuous $\mathop{\lim}_{\bx\rightarrow \ba} G(\bx) = G(\ba)$.

The partial derivative $\frac{\partial G}{\partial x_i} (\bx)$ is also a real-valued function of $\bx\in\sS$ that can be partially differentiated. The $j$-th partial derivative of $\frac{\partial G}{\partial x_i} (\bx)$ is defined as 
$$
\frac{\partial^2 G}{\partial x_j\partial x_i} (\bx)=
\frac{\partial \left(\frac{\partial G}{\partial x_i} (\bx)\right)}{\partial x_j} (\bx).
$$
This is called the ($j,i$)-th \textit{second-order partial derivative} of function $G$.
A function $G$ defined over an open set $\sS\subseteq$ is called \textit{twice continuously differentiable} over $\sS$ if all the second-order partial derivatives exist and are continuous over $\sS$. In the setting of twice continuously differentiability, the second-order partial derivative are symmetric:
$$
\frac{\partial^2 G}{\partial x_j\partial x_i} (\bx)=
\frac{\partial^2 G}{\partial x_i\partial x_j} (\bx).
$$
The \textit{Hessian} of the function $G$ at a point $\bx\in\sS$ is defined as the symmetric $D\times D$ matrix 
$$
\nabla^2G(\bx)=
\begin{bmatrix}
\frac{\partial^2 G}{\partial x_1^2} (\bx) & 
\frac{\partial^2 G}{\partial x_1\partial x_2} (\bx) & \ldots &
\frac{\partial^2 G}{\partial x_1\partial x_D} (\bx)\\
\frac{\partial^2 G}{\partial x_2\partial x_1} (\bx) & 
\frac{\partial^2 G}{\partial x_2\partial x_2} (\bx) & \ldots &
\frac{\partial^2 G}{\partial x_2\partial x_D} (\bx)\\
\vdots & 
\vdots & \ddots &
\vdots\\
\frac{\partial^2 G}{\partial x_D\partial x_1} (\bx) & 
\frac{\partial^2 G}{\partial x_D\partial x_2} (\bx) & \ldots &
\frac{\partial^2 G}{\partial x_D^2} (\bx)
\end{bmatrix}.
$$

For high-dimensional functions, we have the following approximation results.
\begin{theoremHigh}[Mean value theorem]\label{theorem:mean_approx}
Let $f:\sS\rightarrow \real$ be a  continuously differentiable function over an open set $\sS\subseteq\real^D$, and given two points $\bx, \by\in\sS$. Then, there exists a point $\bxi\in[\bx,\by]$ such that 
$$
G(\by) = G(\bx)+ \nabla G(\bxi)^\top (\by-\bx).
$$

\end{theoremHigh}
\begin{proof}[of Theorem~\ref{theorem:mean_approx}]
Without loss of generality, we assume $\bxi(t) = \bx + t(\by - \bx)$ with $t\in[0,1]$. 
Then we define $g(t) \triangleq G(\bxi(t))$. We can apply the ordinary mean value theorem to $g(t)$, to get
$$
g(1) = g(0) + g'(\alpha),
$$
for some $\alpha$ in the interval $[0, 1]$. The derivative of $g(t)$ is
$$
g'(t) = \langle \nabla G(\bxi(t)), \by - \bx \rangle,
$$
where
$$
\nabla G(\bxi(t)) = \left[\frac{\partial G}{\partial \xi_1}(\bxi(t)), 
\frac{\partial G}{\partial \xi_2}(\bxi(t)),
\ldots, \frac{\partial G}{\partial \xi_D}(\bxi(t))\right]^\top.
$$
Therefore,
$
g'(\alpha) = \innerproduct{\nabla G(\bxi(\alpha)), \by - \bx}.
$
Since $\bxi(\alpha) = (1 - \alpha)\bx + \alpha \by$, this completes the proof.
\end{proof}

We provide linear approximation and quadratic approximation theorems without a proof.
\begin{theoremHigh}[Linear approximation theorem]\label{theorem:linear_approx}
Let $f:\sS\rightarrow \real$ be a twice continuously differentiable function on an open set $\sS\subseteq\real^D$, and let $\bx, \by\in\sS$. Then, there exists a point $\bxi\in[\bx,\by]$ such that 
$$
G(\by) = G(\bx)+ \nabla G(\bx)^\top (\by-\bx) + \frac{1}{2} (\by-\bx)^\top \nabla^2 G(\bxi) (\by-\bx),
$$ 
or 
$$
G(\by) = G(\bx)+\nabla G(\bx)^\top (\by-\bx) + o(\normtwo{\by-\bx}),
$$
or
$$
G(\by) = G(\bx)+\nabla G(\bx)^\top (\by-\bx) + \mathcalO(\normtwo{\by-\bx}^2).
$$
\end{theoremHigh}
This theorem, also known as the \textit{first-order Taylor series expansion}, suggests that the error
in the linear approximation is of the order of the square of the distance between $\bx$ and $\by$.
And the first-order Taylor series expansion $G(\by) \approx G(\bx) + \nabla G(\bx)^\top(\by -\bx)$ is exact only for
the neighborhood of $\by$ at the point $\bx$.

\begin{theoremHigh}[Quadratic approximation theorem]\label{theorem:quad_app_theo}
Let $f:\sS\rightarrow \real$ be a twice continuously differentiable function on an open set $\sS\subseteq\real^D$, and let $\bx, \by\in\sS$. Then it follows that 
$$
G(\by) = G(\bx)+ \nabla G(\bx)^\top (\by-\bx) + \frac{1}{2} (\by-\bx)^\top \nabla^2 G(\bx) (\by-\bx)
+
o(\normtwo{\by-\bx}^2),
$$
or 
$$
G(\by) = G(\bx)+ \nabla G(\bx)^\top (\by-\bx) + \frac{1}{2} (\by-\bx)^\top \nabla^2 G(\bx) (\by-\bx)
+
\mathcalO(\normtwo{\by-\bx}^3).
$$
\end{theoremHigh}
This theorem indicates that the error in the quadratic approximation is of the order of the cube of the distance between $\bx$ and $\by$, making it a more accurate approximation when $\by$ is close to $\bx$.

\begin{problemset}

\item 
Verify that the vector $\ell_2$-norm, matrix Frobenius norm, and matrix spectral norm satisfy all three defining conditions for valid matrix norms specified in Definition~\ref{definition:matrix-norm}.

\item \label{problem:multiGauss} 
Given the ability to generate univariate standard Gaussian samples $\normal(0, 1)$, construct a sampling scheme to generate multivariate Gaussian random vectors following $\normal(\bmu, \bSigma)$, where $\bmu\in\real^D$ and the positive-definite covariance matrix admits the factorization $\bSigma=\bC\bC^\top$ \citep{lu2021numerical, lu2022matrix}.
\textit{Hint: Let $\rx_1, \rx_2, \dots, \rx_D$ be i.i.d. standard Gaussian variables and define the vector $\rvx=[\rx_1, \rx_2, \dots, \rx_n]^\top$. The affine transformation $\bC\rvx + \bmu$ yields the desired multivariate Gaussian vector: $\bC\rvx +\bmu\sim \normal(\bmu, \bSigma)$.}

\item Provide rigorous proofs for Lemma~\ref{lemma:affine_mult_gauss} (affine invariance of multivariate Gaussian distributions) and Lemma~\ref{lemma:rotat_multi_gauss} (rotational invariance of spherical Gaussian distributions).

\item \textbf{Joint Gaussian density \citep{lu2023bayesian}.} Let $\mathcalX=\{x_1, x_2, \ldots, x_N\}$ be drawn i.i.d. from a Gaussian distribution of $\normal(x\mid \mu, \sigma^2)$.
Show that the joint probability distribution of $\mathcalX$ can be written as
\begin{equation}\label{equation:uni_gaussian_likelihood}
\begin{aligned}
p(\mathcal{X} \mid \smu, \ssigma^2) 
&= (2\pi)^{-N/2}  (\ssigma^2)^{-N/2} \exp\left\{-\frac{1}{2 \ssigma^2}  \left[  N(\overline{x} - \smu)^2 +  N S_{\overline{x}} \right] \right\},
\end{aligned}
\end{equation}
where $S_{\overline{x}}\triangleq \sum_{n=1}^N(x_n - \overline{x})^2$ and $\overline{x} \triangleq \frac{1}{N} \sum_{i=1}^{N}x_i$.
Extend this result to the multivariate setting. 
Show that the likelihood of $N$ random observations $\mathcal{X} = \{\bx_1, \bx_2, \ldots , \bx_N \}$,  generated by a multivariate Gaussian with mean vector $\bmu$ and covariance matrix $\bSigma$, is given by 
$$
\begin{aligned}
&\gap p(\mathcal{X} \mid \bmu, \bSigma) =\prod^N_{n=1} \normal (\bx_n\mid \bmu, \bSigma) 
= (2\pi)^{-ND/2} \abs{\bSigma}^{-N/2}\exp\left\{-\frac{1}{2} \tr( \bSigma^{-1}\bS_{\bmu} )  \right\}\\
&= (2\pi)^{-ND/2} \abs{\bSigma}^{-N/2}\exp\left\{-\frac{N}{2}(\bmu - \overline{\bx})^\top \bSigma^{-1}(\bmu - \overline{\bx})\right\}  \exp\left\{-\frac{1}{2}\tr( \bSigma^{-1}\bS_{\overline{x}} )\right\},
\end{aligned}
$$
where 
$$
\bS_{\bmu} \triangleq \sum^N_{n=1}(\bx_n - \bmu)(\bx_n - \bmu)^\top,\quad
\bS_{\overline{x}} \triangleq \sum^N_{n=1}(\bx_n - \overline{\bx})(\bx_n - \overline{\bx})^\top, \quad
\overline{\bx} \triangleq\frac{1}{N}\sum^N_{n=1}\bx_n.
$$
The matrix $\bS_{\overline{x}}$ is the \textit{matrix of sum of squares} and is also known as the \textit{scatter matrix}.

\item \label{prob:marg_mvn} \textbf{Marginal and conditional of MVN \citep{lu2023bayesian}.}
Let $\rvx$ and $\rvy$ be jointly Gaussian random vectors with 
$$
\rvz=
\begin{bmatrix}
\rvx\\
\rvy 
\end{bmatrix}
\sim 
\normal\left(
\begin{bmatrix}
\bmu_x\\
\bmu_y 
\end{bmatrix}
,
\begin{bmatrix}
\bA & \bC\\
\bC^\top & \bB
\end{bmatrix}
\right)=
\normal\left(
\begin{bmatrix}
\bmu_x\\
\bmu_y 
\end{bmatrix}
,
\begin{bmatrix}
\widetildebA & \widetildebC\\
\widetildebC^\top & \widetildebB
\end{bmatrix}^{-1}
\right).~\footnote{
Given nonsingular $\bM$ and its inverse $\bM^{-1}$; and suppose appropriate sizes for the following partitions \citep{williams2006gaussian}:
$$
\bM=
\begin{bmatrix}
\bA & \bB\\
\bC & \bD
\end{bmatrix},
\gap 
\bM^{-1}=
\begin{bmatrix}
\widetildebA & \widetildebB\\
\widetildebC & \widetildebD
\end{bmatrix}.
$$
We have 
\begin{equation}\label{equation:mt_inv}
\begin{aligned}
&\widetildebA=\bA^{-1}+\bA^{-1}\bB\widetildebD\bC\bA^{-1}&=& (\bA-\bB\bD^{-1}\bC)^{-1} , \\
&\widetildebB=-\bA^{-1}\bB\widetildebD&=& -\widetildebA\bB\bD^{-1},\\
&\widetildebC=-\widetildebD\bC\bA^{-1}&=& -\bD^{-1}\bC\widetildebA, \\
& \widetildebD=(\bD-\bC\bA^{-1}\bB)^{-1}&=& \bD^{-1}+\bD^{-1}\bC\widetildebA\bB\bD^{-1},
\end{aligned}
\end{equation}
}
$$
where $\rvx$ and $\rvy$ are \textit{independent} if and only if $\Cov[\rvx,\rvy]=\bC=\bzero$.
Show that every marginal distribution of a multivariate Gaussian distribution is itself a multivariate Gaussian distribution, and the conditional distribution $\rvx\mid \rvy$ also follows a multivariate Gaussian distribution:
\begin{equation}
\begin{aligned}
\rvx
\sim\normal(\bmu_x,\bA),
\gap 
\rvx\mid \rvy=\by 
&\sim \normal(\bmu_x+\bC\bB^{-1}(\by-\bmu_y), \bA-\bC\bB^{-1}\bC^\top)\\
&=\normal(\bmu_x-\widetildebA^{-1}\widetildebC(\by-\bmu_y), \widetildebA^{-1});\\
\rvy
\sim\normal(\bmu_y,\bB),
\gap 
\rvy\mid \rvx=\bx
&\sim \normal(\bmu_y+\bC^\top\bA^{-1}(\bx-\bmu_x), \bB-\bC^\top\bA^{-1}\bC)\\
&=\normal(\bmu_y-\widetildebB^{-1}\widetildebC^\top(\bx-\bmu_x), \widetildebB^{-1}).\\
\end{aligned}
\end{equation}

\item  \label{prob:linear_gaus_model} \textbf{Linear Gaussian models: Affine dependence of Gaussian variables.}
Suppose random vectors $\rvx\sim\normal(\bmu, \bSigma)$ and $\rvy\mid \rvx=\bx\sim\normal(\bA\bx+\bb, \bM)$.
Note $\rvy$ is not simply the affine transformation $\bA\rvx+\bb$, but it follows that $\rvy=\bA\bx+\bb+\bepsilon$ where $\bepsilon\sim \normal(\bzero, \bM)$.
Show that 
$$
\rvy\sim \normal(\bA\bmu+\bb, \bM+\bA\bSigma\bA^\top),
\gap 
\rvx \mid \rvy \sim \normal\big(\bL\big\{\bA^\top\bM^{-1}(\rvy-\bb)+\bSigma^{-1}\bmu  \big\}, \bL\big), 
$$
where $\bL=(\bSigma^{-1}+\bA^\top\bM^{-1}\bA)^{-1}$.
\textit{Hint: Use Problem~\ref{prob:marg_mvn}, compute the cross-covariance of $\rvx$ and $\rvy$ by $\Cov[\rvx,\rvy]=\Exp[(\rvx-\bmu_x)(\rvy-\bmu_y)^\top]=\bSigma\bA^\top$ where $\bmu_x=\bmu$ and $\bmu_y=\Exp[\rvy]=\bA\bmu+\bb$, and use Woodbury matrix identity: $(\bA+\bB\bD\bC)^{-1} = \bA^{-1} - \bA^{-1} \bB(\bD^{-1} + \bC\bA^{-1}\bB)^{-1}\bC\bA^{-1}$ for appropriate matrices $\bA,\bB,\bC$, and $\bD$; see, for example,  \citet{lu2021numerical}}.

\item \label{prob:lotus} \textbf{Properties of expectation: LOTUS.}
Let $\rx$ and $\ry$ be two random variables, and let $a, b$ be scalars. Show that $\Exp[a\rx+b\ry] = a\cdot\Exp[\rx]+b\cdot \Exp[\ry]$.
Given further a  function $h$, show that 
$$
\Exp[h(\rx)] = \sum_{x} h(x) \prob(x)
\qquad \text{and}\qquad 
\Exp[h(\rx)] = \int_{-\infty}^{\infty} h(x) d F(x),
$$
in the discrete and continuous cases, respectively.

\item \label{prob:prop_expcond}\textbf{Properties of expectation of conditionals.}
Let $\rx$ and $\ry$ be two random variables, and let $h$ be a  function.
Show that
\begin{enumerate}
\item Note that $\Exp[\rx\mid\ry]$ is a function of $\ry$. However, if $\rx$ is independent of $\ry$, then $\Exp[\rx\mid \ry] = \rx$.
\item $\Exp[c\mid \rx] =\rx$, where $c$ is a constant.
\item Linearity: $\Exp[a\rx_1 + b\rx_2\mid \ry] = a\cdot\Exp[\rx_1\mid \ry] + b\cdot\Exp[\rx_2\mid \ry]$.
\item Conditional constant: $\Exp[h(\ry)\rx\mid \ry] = h(\ry)\Exp[\rx\mid \ry]$, where $h(\ry)$ is called a \textit{conditional constant} w.r.t. $\ry$.
\item Monotonicity: if $\rx_1 \leq \rx_2$, then $\Exp[\rx_1\mid \ry] \leq \Exp[\rx_2\mid \ry]$.
\item Tower property: $\Exp\left[\Exp[\rx\mid \ry]\mid h(\ry)\right] = \Exp[\rx\mid h(\ry)]$; that is, $h(\ry)$ conveys information at most as $\ry$.
\item Unbiasedness: $\Exp\big\{\Exp[h(\rx,\ry)\mid \ry]\big\} = \Exp[h(\rx,\ry)]$; specially, $\Exp\big[\Exp[\rx\mid \ry]\big] = \Exp[\rx]$.
\item Least squares: $\Exp\big[\big(\ry - \Exp[\ry\mid \rx]\big)^2\big] \leq \Exp\big[\big(\ry - h(\rx)\big)^2\big]$ for any function $h$. This also means  $g(\rx)\triangleq  \Exp[\ry\mid \rx]$ is the best estimate in the least squares sense.
\end{enumerate}

\item \textbf{Sum of random variables by convolution.}
Let $ \rx $ and $ \ry $ be continuous random variables with probability density functions $ f_{\rx} $ and $ f_{\ry} $. Show that the density function of $ \rx + \ry $ is the convolution of $ f_{\rx} $ with $ f_{\ry} $:
$$
f_{\rx+\ry}(u) = \int_{-\infty}^{+\infty} f_{\rx}(u - v) f_\ry(v) \, dv.
$$

\item \textbf{Properties of variance and correlation.}
Let $\rx, \rx_1, \rx_2, \ry, \rvx$ be random variables or vectors, and  let $a,b$ be constants. 
Show that 
\begin{itemize}
\item Let $\bOmega$ be a real symmetric matrix. Then $\bOmega$ is positive semidefinite  if and only if $\bOmega$ is the covariance matrix of some random vector $\rvx$.
\item $\Var[\rx] = \Exp[\rx^2] - (\Exp[\rx])^2 = \Cov[\rx,\rx]$.
\item $\Var[a\rx + b] = a^2 \Var[\rx]$.
\item $\Var\big[\sum_i \rx_i\big] = \sum_i \Var[\rx_i] + \sum_{i \neq j} \Cov[\rx_i, \rx_j]$.
\item $\Cov[\rx_1, \rx_2] = \Exp[\rx_1 \rx_2] - \Exp[\rx_1]\Exp[\rx_2]$.
\item $\Cov[a\rx_1 + b\rx_2, \ry] = a\cdot\Cov[\rx_1, \ry] + b\cdot \Cov[\rx_2, \ry]$; that is, covariance is linear in one variable.
\item If $\Exp[\rx_1^2] + \Exp[\rx_2^2] < \infty$, then the following are equivalent:
\begin{enumerate}
\item[(i)] $\Exp[\rx_1 \rx_2] = \Exp[\rx_1]\Exp[\rx_2]$;
\item[(ii)] $\Cov[\rx_1, \rx_2] = 0$;
\item[(iii)] $\Var[\rx_1 \pm \rx_2] = \Var[\rx_1] + \Var[\rx_2]$.
\end{enumerate}
Note that independence will imply these three last properties, but none of these properties imply independence.
\item Let $h$ be a nondecreasing function such that $\Exp[\rx^2]<\infty$ and $\Exp[h(\rx)^2]<\infty$. Then $\Cov[\rx, h(\rx)]>0$.
\end{itemize}

\item  Let $\rx\sim \normal(\mu_x, \sigma_x^2)$ and $\ry\sim \normal(\mu_y, \sigma_y^2)$ be two Gaussian random variables. If $\rx$ and $\ry$ are uncorrelated, then $\rx+\ry \sim \normal(\mu_x+\mu_y, \sigma_x^2+\sigma_y^2).$
Show that if $\rx$ and $\ry$ have correlation coefficient $\rho$ (see \eqref{equation:correlation_def}), then
$$
\rx+\ry \sim \normal(\mu_x+\mu_y, \sigma_x^2+\sigma_y^2+2\rho\sigma_x\sigma_y).
$$

\item \textbf{Conditional variance and law of total variance.}
Let $\rx$ and $\ry$ be two random variables. The \textit{condition variance} of $\rx$ given $\ry$ is defined as 
\begin{equation}
\Var[\rx \mid \ry] \triangleq \Exp\big[ \big(\rx - \Exp[\rx \mid \ry]\big)^2 \mid \ry \big] =\Exp[\rx^2\mid \ry ] - (\Exp[\rx \mid \ry ])^2.
\end{equation}
The conditional variance tells us how much variance is left if we use $\Exp[\rx\mid \ry]$ to ``predict" $\rx$.
Prove the \textit{law of total variance}:
\begin{equation}
\Var[\rx] = \Exp\big[\Var[\rx\mid \ry]\big] + \Var\big[\Exp[\rx\mid \ry]\big].
\end{equation}
\textit{Hint: $\Var[\rx] = \Exp[\rx^2] -(\Exp[\rx])^2 = \Exp[\Exp[\rx^2\mid \ry ]] -(\Exp[\Exp[\rx\mid \ry]])^2$ by the unbiasedness property in Problem~\ref{prob:prop_expcond}.}
\end{problemset}


\chapter{Latent Variable Generative Models: VAEs and GANs}\label{chapter:vae_gan}
\begingroup
\hypersetup{
	linkcolor=structurecolor,
	linktoc=page,  
}
\minitoc \newpage
\endgroup

\lettrine{\color{caligraphcolor}G}
Generative models employ machine learning algorithms to learn a probability distribution $p_{\text{data}}(\bx)$ from training data, enabling the synthesis of new, plausible examples from that distribution. 
For instance, a generative model trained on human images learns to generate novel faces by capturing the underlying statistical structure of the training set. 
Formally, we can frame such a model as $p_{\btheta}(\bx)$, where $\bx$ denotes a data vector in the observation space and $\btheta$ represents the model's learnable parameters. For many practical applications, we extend this to conditional generative models of the form $p_{\btheta}(\bx \mid \by)$, where $\by$ encodes conditioning variables---such as specifying that a generated face image should depict a man,  a woman, or a child.

Across diverse data modalities, observed data $\bx$ is often understood as arising from an associated, unobserved latent variable $\bz$, such that the data is generated via $p(\bx \mid \bz)$. A vivid analogy for this latent variable perspective is \textit{Plato's Allegory of the Cave} \citep{luo2022understanding}: prisoners in a cave perceive only two-dimensional shadows cast by unseen three-dimensional objects, never witnessing the objects themselves. These shadows---their observable reality---are determined by higher-dimensional abstract entities they cannot directly behold.

Similarly, real-world data can be interpreted as a ``projection" of higher-level latent representations that encode abstract properties (e.g., color, shape, or semantic meaning in images). 
Just as cave dwellers infer the existence of hidden objects from shadows, we aim to approximate these latent representations that generate our observations. A critical caveat to this analogy, however, is that in generative modeling---unlike Plato's allegory---we typically seek \textbf{lower-dimensional latent representations}, as higher-dimensional latents are impractical to learn without strong priors. This dimensionality reduction acts as a form of compression, often revealing semantically meaningful structure underlying the observed data.

With this \textit{latent variable framework} established, we now turn to the foundational inference and optimization techniques that underpin modern \textit{latent variable generative models}. 
Directly maximizing the likelihood of observed data $p_{\btheta}(\bx)$ is often intractable, as it requires integrating over the unobserved latent variable $\bz$:
$$
p_{\btheta}(\bx) = \int p_{\btheta}(\bx \mid \bz) \, p_{\btheta}(\bz) \diff \bz.
$$
This integral is typically high-dimensional and cannot be computed in closed form for most practical models. To address this challenge, we first introduce the \textit{evidence lower-bound (ELBO)}, a tractable surrogate objective that forms the basis for \textit{variational inference (VI)}, and we connect it to the \textit{expectation-maximization (EM)} algorithm---a classic approach for maximum likelihood estimation with latent variables. Together, these tools provide the mathematical foundation for learning both the latent distribution $p_{\btheta}(\bz)$ and the generative mapping $p_{\btheta}(\bx \mid \bz)$, which we will later build upon in other deep generative models.

Building on this latent variable intuition, we next consider autoencoders---a class of neural networks that learn to compress data into a low-dimensional latent space and then reconstruct the original input from this compressed representation. While standard autoencoders learn a deterministic mapping from data to latents, they lack a probabilistic interpretation of the latent space, limiting their utility as generative models. To address this, we introduce the \textit{variational autoencoder (VAE)}, which frames the autoencoder within the probabilistic latent variable framework we established: it treats the latent space as a probability distribution $p(\bz)$, uses variational inference to approximate the intractable posterior $p(\bz \mid \bx)$, and maximizes the ELBO to jointly learn the encoder (inference model) and decoder (generative model). This marriage of neural networks and probabilistic inference gives VAEs their powerful generative capabilities, setting the stage for our later discussion of \textit{adversarial generative networks (GANs)} as an alternative approach to latent variable modeling.

\begin{figure}[htbp]
\centering
\includegraphics[width=0.9\textwidth]{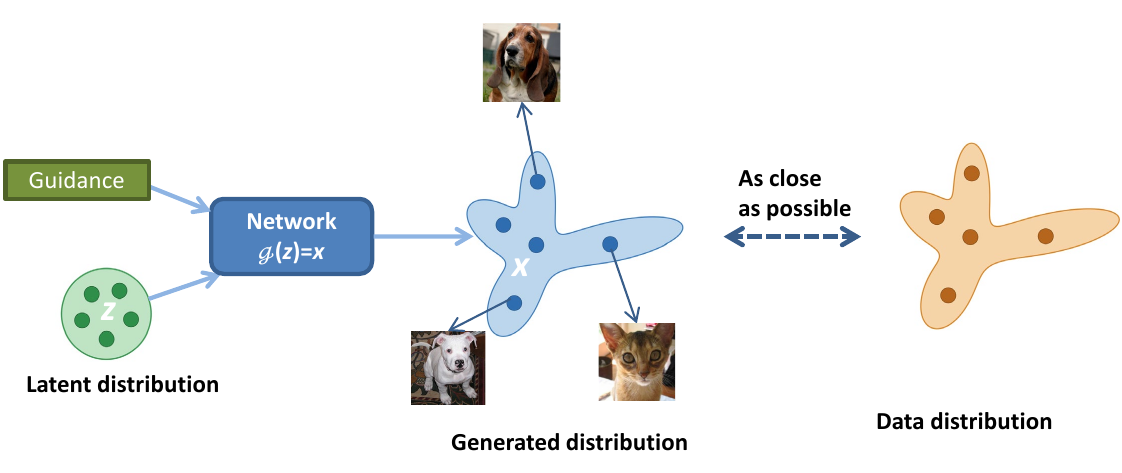} 
\caption{Conceptual illustration of generative models as distribution transformation problems. }
\label{fig:gen_as_prob_trans}
\end{figure}

As illustrated in Figure~\ref{fig:gen_as_prob_trans}, this core intuition frames generative modeling fundamentally as a \textit{distribution transformation problem}: the model learns to map a simple, predefined latent distribution $p(\bz)$ (represented by the compact latent space $\bz$) to the complex, high-dimensional data distribution $p_{\text{data}}(\bx)$ (the manifold of real-world observations, such as images) via a learned transformation $\mathcalG(\bz) = \bx$. This transformation aligns samples from the latent space with the structure of real data, enabling the synthesis of new, plausible examples that match the target distribution.

\section{Background: ELBO, EM, and VI}\label{section:vbi_root}
\index{Variational inference}
\index{Expectation-maximization (EM)}
\index{Kullback--Leibler (KL) divergence}

This section presents the foundational principles of  \textit{variational inference} (\textit{VI} or \textit{variational Bayesian inference}), the evidence lower-bound (ELBO), and expectation-maximization (EM). 
Together, these concepts establish the mathematical framework underlying modern latent-variable generative models, including variational autoencoders (VAEs; see Section~\ref{section:ae_vae}), and diffusion models (see Chapter~\ref{chapter:diff}).
Variational inference is an approximate inference method that seeks to closely match the intractable true posterior distribution by minimizing a statistical divergence between the learned approximation and the target posterior. In VI, this divergence is quantified by the \textit{Kullback--Leibler (KL) divergence}, which measures how much one probability distribution deviates from a reference distribution. By reformulating Bayesian inference as an optimization problem---namely, minimizing KL divergence---VI optimizes the parameters of a simple, tractable surrogate distribution, termed the variational distribution, to best approximate the analytically intractable posterior. This formulation achieves a principled balance between approximation quality and computational efficiency, yielding scalable inference solutions for complex probabilistic models \citep{jordan1999introduction, wainwright2008graphical}.

\subsection{Motivating Model: Latent Variables and Alternating Methods}\label{section:lvm}
\index{Variational inference}
\index{Hidden variables}
\index{Latent variables}
\index{Latent variable models}
\index{Global latent variables}

\begin{SCfigure}
\centering
\includegraphics[width=0.35\textwidth]{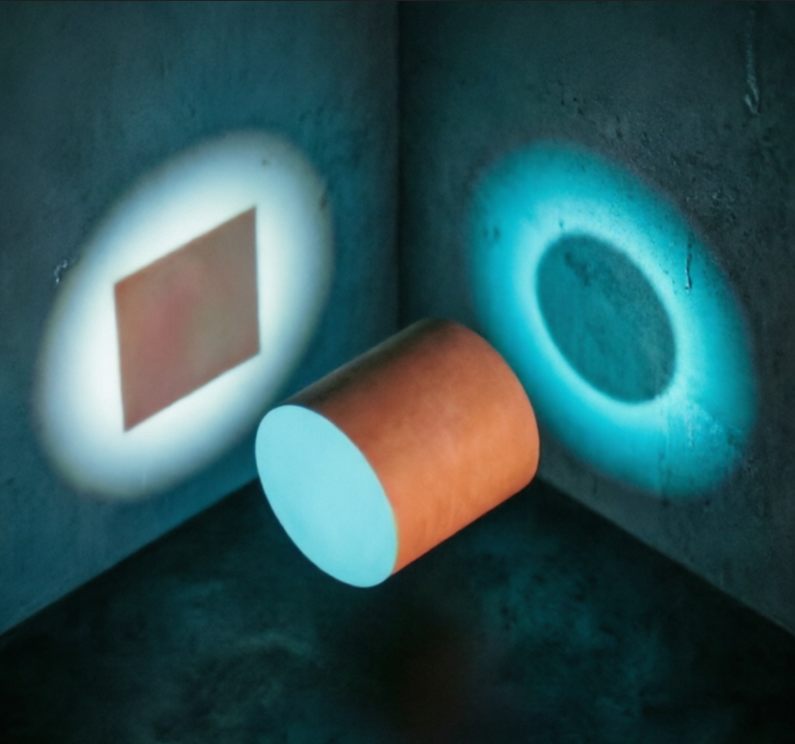}
\caption{Orthographic projection of a cylinder onto two perpendicular planes (the walls of a corner), clearly showing how a single 3D object can appear as two distinct 2D shapes when viewed from orthogonal directions.}
\label{fig:cylinder}
\end{SCfigure}

Consider a statistical model that jointly generates two random vectors, $\rvx\in \sX$ and $\rvz\in \sZ$, instead of one, following a distribution from the parametric family $\mathcalF = \{p_{\btheta} = p_{\btheta}(\cdot , \cdot) \mid \btheta\in\bTheta\} $. 
Formally, $(\rvx, \rvz) \sim p_{\bthetastar}$ for some true parameter $\bthetastar\in \bTheta$. 
In practice, however, only realizations of  $\rvx$ are observable, where  $\rvx\sim p_{\text{data}}(\rvx)$,  while the components of  $\rvz$ remain unobserved.
Specifically, although the unknown true parameter $\bthetastar$ generates  $N$ pairs i.i.d. pairs  $(\bx_1, \bz_1), (\bx_2, \bz_2), \ldots , (\bx_N, \bz_N)$, we only have access to the observed dataset $\mathcalX=\{\bx_1, \bx_2, \ldots, \bx_N\}$. 
The unobserved components $\rvz$ are therefore termed  \textit{latent variables} or \textit{hidden variables}. 
For instance, latent variables may correspond to projections of a cylinder from two viewpoints, which can be used to reconstruct the original cylinder; see Figure~\ref{fig:cylinder} (similar to the ``Plato's Allegory of the Cave").
Other common examples include the component indicators in Gaussian and Bernoulli mixture models  
(see Problems~\ref{prob:mix_of_gauss}--\ref{prob:mix_of_bern})
\footnote{\textcolor{black}{While more complex generative structures are possible, we restrict our attention here to the standard setting in which each observed data point is associated with its own latent variable}.}.
Latent variables are integral to model formulation but absent from observational data. Though unmeasurable, they are critical for explaining the intrinsic structural patterns and statistical variability of observed data.

\begin{SCfigure}
\centering
\includegraphics[width=0.45\textwidth]{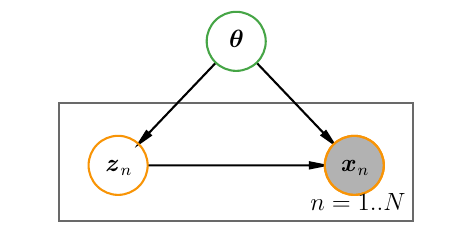}
\caption{Graphical model representation of latent variable models. Green circles denote global latent variables, orange circles represent observed (shaded) and latent (unshaded) variables, and plates indicate repeated structures across data points.}
\label{fig:lvm}
\end{SCfigure}

Within this framework, the model parameter $\btheta$ is typically interpreted as a \textit{global latent variable}, as it governs the overall data-generating process and influences all observations and their corresponding latent variables.
In contrast, the sample-specific hidden variables $\{\bz_n\}_{n=1}^N$ are referred to as \textit{local latent variables} (or simply \textit{latent variables}). 
Each $\bz_n$ is uniquely associated with its corresponding observation
$\bx_n$ and captures the local variation specific to that individual data point.  
This hierarchical relationship between global parameters, local latent variables, and observed data is illustrated in the graphical model in Figure~\ref{fig:lvm}.

Latent variables naturally emerge in a wide range of real-world applications. In \textit{clustering tasks}, for example, they correspond to the unknown cluster assignments of individual data points \citep{beal2003variational, jain2017non, lu2021survey}. In \textit{topic modeling}, latent variables encode the implicit semantic topics underlying document collections, while in \textit{image analysis}, they capture high-level visual structures and abstract features that cannot be directly observed from raw pixel values \citep{blei2003latent}. In this way, \textit{latent variable models (LVMs)} enable the discovery of hidden data patterns and support rigorous, interpretable inference about the underlying data-generating mechanism.

\paragrapharrow{Maximum likelihood estimation and KL divergence.}
The most direct approach to parameter estimation in such models is to compute the \textit{maximum likelihood (ML) estimator} of the parameter $\btheta$. 
This is achieved  by maximizing the \textit{marginal likelihood} (also known as \textit{model evidence}), which is obtained by marginalizing out the latent variables $\{\bz_n\}$:
\begin{equation}\label{equation:lvm_maxmarg}
\widehatbtheta 
= \argmax_{\btheta\in\bTheta}\,\ell(\btheta)
\simeq  \argmax_{\btheta\in\bTheta}\, 
\left\{\ell(\btheta; \bx_1, \bx_2, \ldots, \bx_N) 
\triangleq \prod_{n=1}^N \int p_{\btheta}(\bz_n, \bx_n) \diff\bz_n
\right\},
\end{equation}
where we have assumed the support of the random vector $\rvz$ is continuous in \eqref{equation:lvm_maxmarg}. When it is discrete, the integral is replaced by a sum: $\ell(\btheta; \bx_1, \bx_2, \ldots, \bx_N)=\prod_{n}\sum_{\bz_n\in\sZ} p_{\btheta}(\bz_n, \bx_n) \diff\bz_n$. 

Note that $\ell(\btheta) \triangleq  \Exp_{p_{\text{data}}(\bx)}[p_{\btheta}(\bx)]$ (where $p_{\text{data}}(\bx)$ denotes the true data-generating distribution) defines the \textit{likelihood function},
\footnote{For convenience, we use the notation $\Exp_{\bx\sim  p_{\text{data}}(\bx)} [\ln p_{\btheta}(\bx)]$ or simply $\Exp_{ p_{\text{data}}(\bx)} [\ln p_{\btheta}(\bx)]$  instead of $\Exp_{\rvx\sim p_{\text{data}}} [\ln p(\rvx\mid \btheta)]$ for expectations. 
Here, $\rvx$ denotes a random vector, and $\bx$ denotes a concrete realization. The former notation is widely used in the machine learning community.
By the same convention, we write $\bx \sim p(\bx)$ instead of $\rvx \sim p(\bx)$ to indicate that $\bx$ is distributed according to $p(\bx)$.}
and $\ell(\btheta; \bx_1, \bx_2, \ldots, \bx_N) $ denotes the \textit{empirical likelihood function} given observations $\mathcalX=\{\bx_1,\bx_2, \ldots,\bx_N\}$.
Without loss of generality, we  use the two terms \textbf{interchangeably} and employ the symbol `$\simeq$' to represent  this relationship.

Alternatively, the marginal likelihood function $\ell(\btheta; \bx_1, \bx_2, \ldots, \bx_N)$ can be derived using the \textit{chain rule of probability}:
\begin{equation}\label{equation:lvm_maxmarg_v2}
\widehatbtheta = \mathop{\argmax}_{\btheta\in\bTheta}\, 
\left\{\ell(\btheta; \bx_1, \bx_2, \ldots, \bx_N) 
=  \prod_{n=1}^N  \frac{p_{\btheta}(\bx_n, \bz_n)}{p_{\btheta}(\bz_n\mid \bx_n)}
\right\},
\end{equation}
However, in most practical scenarios, maximizing the marginal likelihood is computationally intractable.
This is   either because it  requires integrating out all latent variables $\{\bz_n\}$ in \eqref{equation:lvm_maxmarg}, which is infeasible  for complex models, or because it relies on access to the true posterior encoder $p_{\btheta}(\bz_n\mid \bx_n)$ in \eqref{equation:lvm_maxmarg_v2}.
For instance, when the support $\sZ$ is discrete, summing over $\sZ$ involves $\abs{\sZ}^N$ terms~\footnote{$\abs{\sZ}$ denotes  the cardinality of the set $\sZ$.}, making direct evaluation and optimization of the likelihood impractical.
\footnote{This intractability is characteristic of many latent variable models and motivates the use of approximate inference techniques---such as Monte Carlo sampling, variational inference, or the EM algorithm---which avoid direct computation of the marginal likelihood by approximating either the posterior distribution or the likelihood itself. These methods yield tractable solutions even in high-dimensional or complex latent spaces.
}

Direct maximization of the product of likelihoods is typically intractable, so we instead optimize the \textit{log-likelihood}, which transforms the product into a sum and simplifies optimization.
Let the \textit{Kullback--Leibler (KL) divergence}  between distributions  $P$ and $Q$ be defined as
\begin{equation}\label{equation:kl_def_vae}
\KL[P \parallel Q] \triangleq \int P(x) \ln \left( \frac{P(x)}{Q(x)} \right) \diff x 
=\Exp_{P} \left[\ln \left( \frac{P(x)}{Q(x)}\right) \right]
\geq 0
\end{equation}
with equality if and only if  $P=Q$. That is, KL divergence is always \textbf{nonnegative} and equals zero precisely when the two distributions are identical.
The KL divergence provides a fundamental measure of discrepancy between two probability distributions.
We now establish the equivalence between \textit{maximum likelihood estimation (MLE)} and \textit{KL divergence minimization} for a model without latent variables,  $p_{\btheta}(\bx)$ with parameter $\btheta$:
\begin{proposition}[MLE and KL divergence minimization]
\begin{equation}\label{equation:equiv_mle_kl}
\begin{aligned}
\argmax_{\btheta} \Exp_{ p_{\text{data}}(\bx)} [\ln p_{\btheta}(\bx)]
&=\argmax_{\btheta} \int p_{\text{data}}(\bx) \ln \frac{p_{\btheta}(\bx)}{p_{\text{data}}(\bx)}\diff \bx\\
&=\argmin_{\btheta} \KL[p_{\text{data}}(\bx)\parallel p_{\btheta}(\bx)],
\end{aligned}
\end{equation}
where $p_{\text{data}}(\bx)$ denotes the true data-generating distribution.
\end{proposition}
The derivation in \eqref{equation:equiv_mle_kl} thus confirms that maximizing the likelihood of  observed data is equivalent 
to minimizing the KL divergence between the true data distribution $p_{\text{data}}(\bx)$ and the model's predictive distribution $p_{\btheta}(\bx)$.

\paragrapharrow{Alternating maximization (AM) algorithm.}
To overcome this intractability, a widely used strategy is the \textit{alternating maximization (AM)} approach. This iterative method alternates between two steps: (1) inferring  latent variables given the current parameter estimate, and (2) updating  model parameters using the inferred latent variables.
More concretely, suppose the true parameter $\bthetastar$ were known. 
We could then estimate each latent variable via \textit{maximum a posteriori (MAP)} assignment:
$$
\text{(AM-Step 1): }\quad 
\widehatbz_n = \mathop{\argmax}_{\bz\in\sZ} p_{\bthetastar}(\bz \mid \bx_n),\quad \forall\, n\in\{1,2,\ldots,N\}.
$$
Conversely, if the latent variables $\{\bz_n\}$ were known, the ML estimate of $\btheta$ would be:
$$
\text{(AM-Step 2): }
\qquad 
\widehatbtheta_{\text{MLE}} =\mathop{\argmax}_{\btheta\in\bTheta} \, \ell(\btheta; \{(\bx_n,\bz_n)\}_{n=1}^N ).
$$
This step refines parameter estimates using the complete dataset, including the inferred latent variables. 
This alternating procedure yields a practical algorithm for fitting latent variable models; see Algorithm~\ref{alg:am_lvm}.

\paragrapharrow{Limitation and EM algorithm.}
Although intuitive, this alternating maximization framework has significant drawbacks---especially when the latent space $\sZ$ is large or highly structured. 
At each iteration $t$, the algorithm performs a ``hard assignment,"  mapping each data point $\bx_n$ to a single latent value $\widehatbz_n^\toptzero \in \sZ$ (we use the superscript $(t)$ to denote the iteration index). 
This discards other plausible latent configurations $\bz^\prime$  for which the posterior probability $p_{\btheta^\toptzero}(\bz^\prime \mid  \bx_n )$ may remain substantial, even if lower than the maximum value $ p_{\btheta^\toptzero}(\bz_n^\toptzero \mid \bx_n)$. 
As a result, valuable uncertainty information is lost, which can lead to suboptimal or unstable parameter estimates.

The \textit{expectation-maximization (EM)} algorithm is designed to address this issue \citep{baum1970maximization, dempster1977maximum}.
Instead of making hard assignments, EM preserves uncertainty by computing the expected \textit{complete-data log-likelihood} under the current posterior distribution over  latent variables. 
This ``soft assignment"  weights all possible latent values by their posterior probabilities, enabling the algorithm to account for the full range of plausible interpretations  for each observation. Consequently, EM generally produces more robust and accurate parameter estimates than basic alternating maximization.
To achieve this, the EM algorithm alternately maximizes the \textit{evidence lower-bound (ELBO)} (via coordinate ascent), which---as its name suggests---is a lower bound of the evidence.

\index{Alternating maximization}
\index{Latent variable models}
\index{EM algorithm}
\begin{algorithm}[h!] 
\caption{Alternating Maximization (AM) for Latent Variable Models}
\label{alg:am_lvm}
\begin{algorithmic}[1] 
\Require Observed data points $\{\bx_1, \bx_2, \ldots, \bx_N\}$;
\State \textbf{initialize:} $\btheta^\topone$; 
\State Set maximum number of iterations $C$;
\State $t=0$; \Comment{Iteration counter}
\While{$t<C$} 
\State $t=t+1$;
\For{$n=1,2,\ldots, N$}
\State Step-1: $\widehatbz_n^\toptzero \leftarrow \mathop{\argmax}_{\bz\in\sZ} p_{\btheta^\toptzero}(\bz \mid \bx_n)$;
\Comment{(AM$_1$)}
\EndFor
\State Step-2: $\btheta^\toptone \leftarrow\mathop{\argmax}_{\btheta\in\bTheta}\, \ell(\btheta; \{(\bx_n,\widehatbz_n^\toptzero)\}_{n=1}^N )$;
\Comment{(AM$_2$)}
\EndWhile
\State Output $\btheta^\toptzero$;
\end{algorithmic} 
\end{algorithm}

\index{Evidence lower-bound}
\index{Variational free-energy}
\index{ELBO}
\index{VFE}
\index{KL divergence}
\index{EM algorithm}
\index{Model evidence}
\subsection{ELBO and VFE}\label{section:elbo_cfe}
As noted earlier, similar to alternating maximization for latent variable models, the EM algorithm also aims to increase the marginal likelihood over successive iterations. 
However, directly maximizing the marginal likelihood (also called \textit{model evidence}; see Equation~\eqref{equation:lvm_maxmarg}) is often intractable or computationally prohibitive. 
To overcome this issue, the EM algorithm introduces a \textbf{proxy function} that acts  as a tractable lower bound on the marginal likelihood. 
The core idea of EM is to construct such a proxy, referred to as the \textit{Q-function}, which lower-bounds the log-marginal likelihood and reformulates  the original parameter estimation problem as a bivariate optimization over both the model parameter $\btheta$ and an auxiliary distribution over the latent variables.

To derive the EM algorithm, we assume---without loss of generality---that the latent variables are supported on a continuous domain.
For a latent variable model,
the log-marginal likelihood function $\ell(\btheta)\simeq \ell(\btheta; \{\bx_n\})$ from \eqref{equation:lvm_maxmarg} satisfies:
\begin{align}
\mathcalL\left(\btheta\right) 
&\triangleq\ln \ell(\btheta)
\simeq \sum_{n=1}^N \ln \int p_{\btheta}(\bz_n, \bx_n) \diff\bz_n
=\sum_{n=1}^N \ln \int q_{\bz_n}(\bz_n) \frac{p_{\btheta}(\bz_n, \bx_n)}{ q_{\bz_n}(\bz_n)} \diff\bz_n\nonumber\\
&\geq \sum_{n=1}^N  \int q_{\bz_n}(\bz_n) \ln\frac{p_{\btheta}(\bz_n, \bx_n)}{ q_{\bz_n}(\bz_n)} \diff\bz_n\nonumber\\
&= \underbrace{\sum_{n=1}^N  \int q_{\bz_n}(\bz_n) \ln p_{\btheta}(\bz_n, \bx_n) \diff\bz_n}_{\text{Explain data/expected energy}} \underbrace{- \sum_{n=1}^N  \int q_{\bz_n}(\bz_n) \ln q_{\bz_n}(\bz_n) \diff\bz_n}_{\text{Entropy}} \nonumber\\
&\triangleq 
\mathcalF\left(q_{\bz_1}(\bz_1), q_{\bz_2}(\bz_2), \ldots, q_{\bz_N}(\bz_N), \btheta\right)
\equiv \mathcalF\left(\{q_{\bz_n}(\bz_n)\}_{n=1}^N, \btheta\right), \label{equation:elbo_ineq}
\end{align}
where the inequality follows from  Jensen's inequality, exploiting  the concavity of the logarithm function.
This quantity $\mathcalF\left(\{q_{\bz_n}(\bz_n)\}_{n=1}^N, \btheta\right)$ is known as the \textit{evidence lower-bound} (\textit{ELBO}, also referred to as the \textit{marginal log-likelihood lower-bound} or  \textit{variational lower-bound}).
The term $q_{\bz_n}(\bz_n)$ denotes an \textbf{arbitrary} probability distribution over the  latent variables, called the \textit{variational distribution}.
If each data point $n$ is governed by distinct parameters, we may write the variational distribution as $q_{\bz_n}(\bz_n) \triangleq q(\bz_n\mid \bx_n, \blambda_n)$.
\footnote{
For example, in VAEs, $q_{\bz_n}(\bz_n) = q_{\blambda} (\bz\mid\bx_n)$ corresponds to an encoder network with parameters $\blambda$ \textbf{shared} across all data points; see \eqref{equation:vae_elbo_obj}.
In diffusion models, $q_{\bz_n}(\bz_n) = q (\bz_1:\bz_T\mid\bx_0)$ represents the \textbf{deterministic} forward process; see \eqref{equation:elbo_expanded_ddpm}.}
As a hindsight (in the context of constrained EM optimization or variational inference), our goal is to maximize the ELBO while keeping $\btheta$ fixed. 
The first term in the ELBO  encourages the variational distributions $q_{\bz_n}(\bz_n)$ to place  high probability on latent configurations that accurately explain the observed data (the \textit{expected energy}); while the second term---the \textit{entropy}---promotes uncertainty by spreading probability mass over multiple configurations, thus avoiding overconfident approximations.

\begin{SCfigure}
\centering
\includegraphics[width=0.6\textwidth]{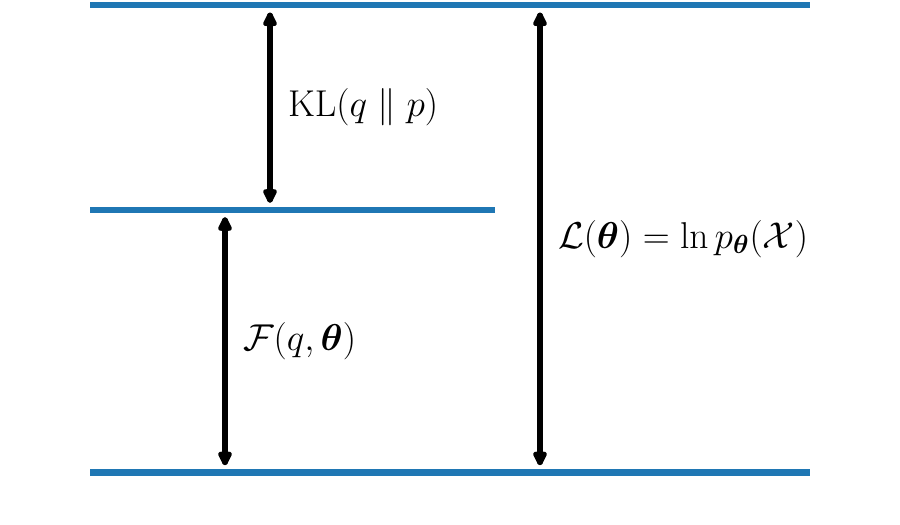}
\caption{Illustration of the ELBO decomposition
given by \eqref{equation:elbo_vfe_neg} or \eqref{equation:elbo_vfe_negv2}, which holds for
any choice of distribution $q(\bz)$.
Because  $\KL[q\parallel p] \geq 0$,
the quantity $\mathcalF(q, \btheta)$ is a lower bound on the log-marginal likelihood
function $\mathcalL(\btheta)=\ln p_{\btheta}(\mathcalX)$.}
\label{fig:ELBO_decom}
\end{SCfigure}

\paragrapharrow{ELBO decomposition.}
The ELBO can be further interpreted via its inherent connection to KL divergence. Specifically,
\begin{subequations}
\begin{equation}\label{equation:elbo_vfe_neg}
\begin{aligned}
\mathcalF\big(\{q_{\bz_n}&(\bz_n)\}_{n=1}^N, \btheta\big) 
=\sum_{n=1}^N \Exp_{q_{\bz_n}} \left[\ln\frac{p_{\btheta}(\bz_n, \bx_n)}{ q_{\bz_n}(\bz_n)}\right]
= \sum_{n=1}^N  \int q_{\bz_n}(\bz_n) \ln\frac{p_{\btheta}(\bz_n, \bx_n)}{ q_{\bz_n}(\bz_n)} \diff\bz_n\\
&=\sum_{n=1}^N  \int q_{\bz_n}(\bz_n) \ln p_{\btheta}(\bx_n) \diff\bz_n  + \sum_{n=1}^N\int q_{\bz_n}(\bz_n) \ln\frac{p_{\btheta}(\bz_n \mid \bx_n)}{ q_{\bz_n}(\bz_n)} \diff\bz_n \\
&=\sum_{n=1}^N\ln p_{\btheta}(\bx_n) - \sum_{n=1}^N\KL\left[ q_{\bz_n}(\bz_n) \parallel p_{\btheta}(\bz_n \mid \bx_n)\right]\\
&\triangleq
-\mathcalL_{\text{VFE}}\left(\{q_{\bz_n}(\bz_n)\}_{n=1}^N, \btheta\right),
\end{aligned}
\end{equation}
where $\mathcalL_{\text{VFE}}\big(\left\{q_{\bz_n}(\bz_n)\right\}_{n=1}^N, \btheta\big)$ denotes the \textit{variational free-energy (VFE)}, 
a concept originating from statistical physics \citep{neal1998view}. 
The VFE can be decomposed as the negative entropy (see Problem~\ref{problem:entropy_mgau}) of $q_{\bz}(\bz)$ minus the expected energy under $q_{\bz}(\bz)$:
$$
\mathcalL_{\text{VFE}}\left(\left\{q_{\bz_n}(\bz_n)\right\}_{n=1}^N, \btheta\right) = 
\underbrace{\sum_{n=1}^N  \int q_{\bz_n}(\bz_n)\ln{ q_{\bz_n}(\bz_n)} \diff\bz_n}_{\text{negative entropy}} 
- \underbrace{\sum_{n=1}^N  \int q_{\bz_n}(\bz_n) \ln{p_{\btheta}(\bz_n, \bx_n)}\diff\bz_n}_{\text{expected energy}}.
$$
The above derivation implies that, with model parameter $\btheta$ fixed, maximizing the ELBO is equivalent to minimizing the KL divergence between the variational distribution $q_{\bz_n}(\bz_n)$ and the true latent posterior $p_{\btheta}(\bz_n \mid \bx_n)$.
\footnote{In the deep learning community, the quantity $p_{\btheta}(\bz_n \mid \bx_n)$ is commonly referred to as a \textit{recognition network} or \textit{inference network}, which can be parameterized via neural networks \citep{hinton1995wake}. Accordingly, the variational distribution $q_{\bz_n}(\bz_n)$ is optimized to approximate this inference network.}
Furthermore, rearranging Equation~\eqref{equation:elbo_vfe_neg} yields a direct decomposition of the log-marginal likelihood $\mathcalL(\btheta)=\ln p_{\btheta}(\mathcalX)$ as the sum of the ELBO and the accumulated KL divergence:
\begin{equation}\label{equation:elbo_vfe_negv2}
\underbrace{\ln p_{\btheta}(\mathcalX) }_{\text{Evidence}}
= 
\underbrace{\mathcalF\big(\{q_{\bz_n}(\bz_n)\}_{n=1}^N, \btheta\big)}_{\text{ELBO}} 
+
\underbrace{\sum_{n=1}^N\KL\left[ q_{\bz_n}(\bz_n) \parallel p_{\btheta}(\bz_n \mid \bx_n)\right]}_{\text{KL divergence}}
,
\end{equation}
where $\mathcalX=\{\bx_1,\bx_2,\ldots,\bx_N\}$ denotes the full set of observed data samples.
\end{subequations}
This, again, confirms that the ELBO is a lower bound of $\mathcalL(\btheta)$ since the KL divergence is nonnegative. 
The KL term quantifies the gap between the ELBO and the exact log-marginal likelihood, thereby measuring the tightness of the variational lower bound. A visual illustration of this decomposition is provided in Figure~\ref{fig:ELBO_decom}.

\paragrapharrow{ELBO as a proxy.}
Maximizing the ELBO $\mathcalF(q, \btheta)$ with respect to both the model parameter $\btheta$  and the approximate posterior distributions $q_{\bz_n}(\bz_n)$ serves two complementary objectives:
\begin{enumerate}[(i)]
\item Maximizing the ELBO minimizes the KL divergence between $q_{\bz_n}(\bz_n)$ and $p_{\btheta}(\bz_n\mid \bx_n)$.
For fixed $\btheta$, the log-likelihood $\ln p_{\btheta}(\mathcalX)$ is a constant. 
Maximizing the ELBO $\mathcalF$ (with respect to $q$) is therefore exactly equivalent to minimizing the KL divergence:
\begin{align*}
\max_q \mathcalF(q, \btheta) 
\quad\iff \quad 
\min_q\KL\left[ q_{\bz_n}(\bz_n) \parallel p_{\btheta}(\bz_n \mid \bx_n)\right].
\end{align*}
This implies that the approximate posterior $q$ becomes as close as possible to the true posterior $p_{\btheta}(\bz \mid \bx)$ or  $p_{\btheta}(\bz_n \mid \bx_n)$ for each observed data point.
\item Maximizing the ELBO tightens the lower bound on the log-likelihood.
Since the ELBO is a lower bound for $\ln p_{\btheta}(\mathcalX)$, increasing the ELBO raises this lower bound, which indirectly promotes an increase in the true log-likelihood.
\end{enumerate}
In practice, we optimize both $\btheta$ and $q$ in tandem:
\begin{itemize}
\item For fixed $q$: Optimize $\btheta$ to maximize the ELBO, which pushes up the log-likelihood of the data under the model.
\item For fixed $\btheta$: Optimize $q$ to maximize the ELBO, which is equivalent to minimizing the KL divergence between $q$ and the true posterior $p_{\btheta}(\bz_n\mid \bx_n)$.
\end{itemize}
When the ELBO is maximized, the KL divergence term becomes zero, and the ELBO equals the true log-likelihood. 
At this optimum, the approximate posterior $q$ matches the true posterior $p_{\btheta}(\bz_n\mid \bx_n)$ exactly.
See Section~\ref{section:em_uncons} for the \textit{unconstrained EM case} (in which case the variational distribution $q_{\bz_n}(\bz_n)$ is equivalent to  the true posterior distribution $p_{\btheta}(\bz_n\mid \bx_n)$ at each iteration).

The ELBO decomposition in \eqref{equation:elbo_vfe_negv2}  is also key to extending the {unconstrained EM algorithm}  to the \textit{constrained EM algorithm} (in which case the  true posterior distribution $p_{\btheta}(\bz_n\mid \bx_n)$ is \textbf{intractable}, and the variational distribution \colorbox{\mdframecolor}{$q_{\bz_n}(\bz_n) \triangleq q(\bz_n\mid \bx_n, \blambda_n)$} is restricted to a tractable family of distributions parameterized by $\blambda_n$).
We briefly explain the motivation for maximizing the ELBO in constrained EM algorithms; see Section~\ref{section:em_constrained} for further details.
Having introduced latent variables $\bz_n$ for each sample $n$, our goal is to infer the underlying latent structure that explains the observed data. 
In other words, we aim to optimize the parameters of the variational posterior $q_{\bz_n}(\bz_n) \triangleq q(\bz_n\mid \bx_n, \blambda_n)$---where each data-point-specific variational distribution depends on the variational parameter $\blambda_n$---to match the true posterior  $p_{\btheta}(\bz_n\mid \bx_n)$ exactly. This is achieved by minimizing their KL divergence (ideally to zero).
Unfortunately, direct minimization of this KL divergence is intractable, as the true posterior $p_{\btheta}(\bz_n\mid \bx_n)$ is unavailable.
However, on the left-hand side of \eqref{equation:elbo_vfe_negv2}, the data likelihood $\bx_n$ (and thus the evidence term  $\ln p_{\btheta}(\bx_n)$) is constant with respect to $\blambda_n$. 
This is because it is computed by marginalizing the latent variables $\bz_n$ from the joint distribution $p(\bx_n, \bz_n)$ and is entirely independent of $\blambda_n$. 
Since the ELBO and the KL divergence sum to a constant, maximizing the ELBO with respect to $\blambda_n$ necessarily minimizes the KL divergence by the same amount. 
The ELBO can therefore be maximized as a proxy for learning the true latent posterior: the more we optimize the ELBO, the closer the approximate posterior becomes to the true posterior. Furthermore, after training, the ELBO can be used to estimate the likelihood of observed or generated data, as it is trained to approximate the model evidence $\ln p(\bx_n)$.

\paragrapharrow{KL divergence behavior.} 
The direction of the KL divergence strongly influences the behavior of the variational approximation:
\begin{itemize}
\item Minimizing  $\KL[q\parallel p]$ (the \textit{reverse} or \textit{exclusive KL}) produces  \textit{mode-seeking} or \textit{zero-forcing} behavior: $q$ is forced to zero wherever $p$ is zero, often concentrating its mass around a single mode of $p$ or one of the modes of $p$.
\item In contrast, minimizing $\KL[p\parallel q]$ (the \textit{forward} or \textit{inclusive KL}) yields  \textit{mass-covering} or \textit{mean-seeking} behavior: $q$ must assign non-negligible probability over the entire support of $p$, resulting in wider, more inclusive approximations   (see Problem~\ref{problem:forward_rever_KL}).
\end{itemize}

\index{EM algorithm}
\subsection{EM for Unconstrained Optimization}\label{section:em_uncons}
We now introduce the EM algorithm for latent variable models.
Although its general formulation is due to \citet{dempster1977maximum}, the core idea had already appeared in earlier work addressing specific problems  \citep{baum1970maximization}.
Similar to alternating maximization, the EM algorithm iterates between two steps:
the \textit{E-step}, which computes the posterior distribution of the latent variables given the current parameter estimate, and the \textit{M-step},  which updates the model parameters (or global latent variables) by maximizing a surrogate objective derived from the E-step.
Using the ELBO derived in \eqref{equation:elbo_ineq}, and denoting the estimates $q_{\bz}(\bz)$ and $\btheta$  at iteration $t$  as $q^\toptzero_{\bz}(\bz)$ and $\btheta^\toptzero$, respectively, the E- and M-step updates are given by
\begin{mybox}
\begin{subequations}
\begin{align}
&\textbf{E-Step: } \quad q^\toptone_{\bz_n}(\bz_n) &\leftarrow& \mathop{\argmax}_{q_{\bz_n}} \mathcalF\left(\left\{q_{\bz_n}(\bz_n)\right\}_{n=1}^N, \btheta^{\textcolor{mylightbluetext}{(t)}}\right),\quad \forall\, n;\\
&\textbf{M-Step: } \quad \btheta^\toptone &\leftarrow& \mathop{\argmax}_{\btheta\in\bTheta} \mathcalF\left(\left\{q^{\textcolor{mylightbluetext}{(t+1)}}_{\bz_n}(\bz_n)\right\}_{n=1}^N, \btheta\right).
\end{align}
\end{subequations}
\end{mybox}
The EM algorithm can thus be interpreted as follows: the E-step approximates the conditional distributions of the local latent variables  $\{\bz_n\}$, while the M-step updates the global parameter $\btheta$.
These two steps are repeated until the sequence $\{\btheta^\toptzero\}$ converges.
Under mild regularity conditions, convergence to a local maximum of the marginal likelihood is guaranteed \citep{gupta2011theory, jain2017non}. 
Since the log-likelihood may possess multiple local maxima, it is common practice to run the EM algorithm multiple times from different initializations $\btheta^\topone$.
The final estimate is then selected as the solution that achieves the highest likelihood across all runs.

\index{Lagrangian function}
\index{Functional derivatives}
\index{Variational derivatives}
\paragrapharrow{E-Step.} 
To derive the E-step, we maximize the ELBO $\mathcalF\big(\left\{q_{\bz_n}(\bz_n)\right\}_{n=1}^N, \btheta^\toptzero\big)$ with respect to each $q_{\bz_n}(\bz_n)$, subject to the normalization constraint $\int q_{\bz_n}(\bz_n) \diff\bz_n =1$ for all $n\in\{1,2,\ldots,N\}$.
Introducing \textit{Lagrange multipliers} $\{\gamma_n\}_{n=1}^N$,  the associated Lagrangian is (see, for example, \citet{boyd2004convex}):
$$
L\left(q_{\bz_n}(\bz_n), \bgamma\right)=
\mathcalF\left(\left\{q_{\bz_n}(\bz_n)\right\}_{n=1}^N, \btheta^\toptzero\right) + \sum_{n} \gamma_n \left(\int q_{\bz_n}(\bz_n) \diff\bz_n -1\right).
$$
Taking the functional (or variational) derivative of $L$ with respect to  $q_{\bz_n}(\bz_n)$ and setting the result to zero yields:
\begin{equation}\label{equation:qzi_fllai}
\int \ln \frac{p_{\btheta^\toptzero}(\bz_n, \bx_n )}{q_{\bz_n}(\bz_n)} \diff\bz_n - 1+\gamma_n = 0
\implies 
q^\toptone_{\bz_n}(\bz_n) = \exp(\gamma_n-1) p_{\btheta^\toptzero}(\bz_n, \bx_n),\, \forall\, n.
\end{equation}
Plugging in the expressions into the Lagrangian function $L\left(q_{\bz_n}(\bz_n), \blambda\right)$ and solving for the maximum gives
$$
\int \exp(\gamma_n-1)p_{\btheta^\toptzero}(\bz_n, \bx_n) \diff\bz_n=0 
\quad\implies \quad
\gamma_n = 1-\ln \int p_{\btheta^\toptzero}(\bz_n, \bx_n) \diff\bz_n.
$$
Substituting $\gamma_n$  backinto \eqref{equation:qzi_fllai}, we obtain 
\begin{equation}\label{equation:cons_EM_estep}
q^\toptone_{\bz_n}(\bz_n) = p_{\btheta^\toptzero}(\bz_n \mid \bx_n), \quad  \forall\, n.
\end{equation}
Hence, the optimal variational distribution in the E-step is exactly the true posterior of the latent variables given the current parameter estimate $\btheta=\btheta^\toptzero$.
Substituting this posterior  $q^\toptone_{\bz_n}(\bz_n)$ back into the ELBO $\mathcalF\big(\left\{q_{\bz_n}(\bz_n)\right\}_{n=1}^N, \btheta^\toptzero\big)$ shows that the ELBO forms a \textbf{tight} bound of the logarithm of the marginal likelihood function $\mathcalL(\btheta) $, meaning the bound becomes an equality (see Problem~\ref{problem:elbo_equa_em}): 
\begin{equation}\label{equation:cons_EM_estep2}
\mathcalF\left(\left\{q_{\bz_n}(\bz_n)\right\}_{n=1}^N, \btheta^\toptzero\right)=
\mathcalL(\btheta^\toptzero).
\end{equation}
In other words, the KL divergence term in the ELBO decomposition shown in Figure~\ref{fig:ELBO_decom} or \eqref{equation:elbo_vfe_negv2} vanishes.

\index{Q-function}
\paragrapharrow{M-Step.}
With the posterior $q^\toptone_{\bz_n}(\bz_n)= p_{\btheta^\toptzero}(\bz_n \mid \bx_n)$ for all $n\in\{1,2,\ldots,N\}$ fixed from the E-step, 
we now optimize the global parameter  $\btheta$. 
In the literature, the E-step of the EM algorithm is understood as constructing the so-called \textit{Q-function} \citep{gupta2011theory, jain2017non}:
$$
\begin{aligned}
&\quad \mathcalF\left(\left\{q_{\bz_n}(\bz_n) = \textcolor{mylightbluetext}{p_{\btheta^\toptzero}(\bz_n \mid \bx_n)}\right\}_{n=1}^N, \textcolor{mylightbluetext}{\btheta}\right)
=\sum_{n}  \int p_{\btheta^\toptzero}(\bz_n \mid \bx_n) \ln\frac{p_{\btheta}(\bz_n, \bx_n)}{ p_{\btheta^\toptzero}(\bz_n \mid \bx_n)} \diff\bz_n\\
&=\underbrace{\sum_{n}  \int p_{\btheta^\toptzero}(\bz_n \mid \bx_n) \ln p_{\btheta}(\bz_n, \bx_n) \diff\bz_n}_{\triangleq \text{$Q(\btheta \mid \btheta^\toptzero)$}} 
-\underbrace{\sum_{n}  \int p_{\btheta^\toptzero}(\bz_n \mid \bx_n)\ln p_{\btheta^\toptzero}(\bz_n \mid \bx_n) \diff\bz_n}_{\text{entropy of $p(\bz_n \mid \bx_n, \btheta^\toptzero$})}.
\end{aligned}
$$
Because the second term (the entropy) does not depend on $\btheta$, maximizing the ELBO is equivalent to maximizing only the Q-function.
\begin{align}
\textbf{M-Step: } \quad \btheta^\toptone &\leftarrow \mathop{\argmax}_{\btheta\in\bTheta} 
\sum_{n}  \int p_{\btheta^\toptzero}(\bz_n \mid \bx_n) \ln p_{\btheta}(\bz_n, \bx_n) \diff\bz_n \nonumber\\
&=\mathop{\argmax}_{\btheta\in\bTheta} Q(\btheta \mid \btheta^\toptzero).
\label{equation:cons_em_mstep}
\end{align}

\paragrapharrow{Role of ELBO, and Q-function.}
The EM algorithm is a fundamental optimization framework for latent variable models, with the ELBO serving as its core driving mechanism. Specifically, the E-step tightens the ELBO bound to match the current model likelihood, while the M-step maximizes the ELBO to update model parameters. This two-stage iterative process ensures monotonic improvement and steady convergence toward a local maximum of the marginal likelihood function. The complete EM procedure is summarized in Algorithm~\ref{alg:em_alg}:
\begin{enumerate}
\item \textit{E-step.} Compute the posterior distributions of the latent variables  $  \{\bz_n\}  $  conditioned on the  observations  $ \{\bx_n\} $  and current parameter estimate $\btheta=\btheta^\toptzero$. 
Based on the decomposition $\ln p_{\btheta}(\mathcalX) = \text{ELBO}+\text{KL}$ (see Figure~\ref{fig:ELBO_decom}), since $\ln p_{ \btheta^\toptzero}(\mathcalX)$ is constant with respect to $q$, maximizing ELBO is equivalent to minimizing the total KL divergence $\sum_{n=1}^N\KL\left[ q_{\bz_n}(\bz_n) \parallel p_{\btheta}(\bz_n \mid \bx_n)\right]$.
\item \textit{M-step.} Update the parameter $\btheta^\toptone$ by maximizing the expected (empirical) complete-data log-likelihood or the Q-function $Q(\btheta \mid \btheta^\toptzero)$.
\end{enumerate}
Because the E-step often admits a closed-form solution (as shown above), the EM algorithm effectively reduces to iteratively constructing and optimizing the Q-function. 
The Q-function can be further compactly written as 
\begin{equation}
Q(\btheta \mid \btheta^\toptzero) 
= \sum_{n=1}^{N} \Exp_{{\bz\sim q}} \big[\ln p_{\btheta}(\bz_n, \bx_n)\big]
= \sum_{n=1}^{N} \Exp_{\underbrace{\bz\sim p_{\btheta^\toptzero}(\cdot \mid \bx_n )}_{\text{conditional prob.}}}  \underbrace{\big[\ln p_{\btheta}(\bz_n, \bx_n )\big]}_{\text{log-joint prob.}},
\end{equation}
where the expectation is evaluated over the latent posterior distribution obtained from the previous iteration, weighting the log-joint probability of observed and latent variables.

\begin{algorithm}[h] 
\caption{Expectation-Maximization (EM) Algorithm}
\label{alg:em_alg}
\begin{algorithmic}[1] 
\Require Observed data points $\{\bx_1, \bx_2, \ldots, \bx_N\}$;
\State \textbf{initialize:} $\btheta^\topone$; 
\State Set maximum number of iterations $C$;
\State $t=0$; \Comment{Iteration counter}
\While{$t<C$} 
\State $t=t+1$;
\State E-step: $q^\toptone_{\bz_n}(\bz_n) \leftarrow p_{\btheta^\toptzero}(\bz_n \mid \bx_n), \, \forall\, n\in\{1,2,\ldots,N\}$;
\State M-step: $\btheta^\toptone \leftarrow \mathop{\argmax}_{\btheta\in\bTheta} Q(\btheta \mid \btheta^\toptzero)$;
\EndWhile
\State Output $\btheta^\toptzero$;
\end{algorithmic} 
\end{algorithm}

The Q-function exhibits favorable theoretical properties as a surrogate objective.
Critically, any parameter update that increases $Q(\btheta \mid \btheta^\toptzero)$  is guaranteed to increase the log-marginal likelihood $\mathcalL\left(\btheta\right)$. 
Furthermore, for widely used models including \textit{Gaussian mixture models} and \textit{mixture regression} (see Problems~\ref{prob:mix_of_gauss}--\ref{prob:mix_of_bern}), the Q-function can be constructed and optimized efficiently in closed form  \citep{jain2017non, lu2023bayesian}.

\paragrapharrow{Soft assignment.}
We further decompose the global Q-function into point-wise contributions from individual data samples $\bx_n$ (\textit{point-wise Q-function}):
$$
\begin{aligned}
Q(\btheta \mid \btheta^\toptzero) 
&=
\sum_{n}  \underbrace{\int p_{\btheta^\toptzero}(\bz_n \mid \bx_n) \ln p_{\btheta}(\bz_n, \bx_n) \diff\bz_n}_{\triangleq \text{$\bQ_{\bx_n}(\btheta \mid \btheta^\toptzero)$}}
&=\sum_{n} \bQ_{\bx_n}(\btheta \mid \btheta^\toptzero),
\end{aligned}
$$
where we define the posterior weight $\bw_{\bz_n}\triangleq p_{\btheta^\toptzero}(\bz_n \mid \bx_n)$, yielding the per-sample Q-function:
\begin{equation}
\bQ_{\bx_n}(\btheta \mid \btheta^\toptzero) 
=
\int \bw_{\bz_n} \ln p_{\btheta}(\bz_n, \bx_n) \diff\bz_n.
\end{equation}
This decomposition highlights the core advantage of the EM framework.
Rather than performing \textit{hard assignment}---which maps each observation $\bx_n$ to a single most-likely latent configuration $\widehatbz_n^\toptzero = \mathop{\argmax}_{\bz\in\sZ} p_{\btheta^\toptzero}(\bz \mid \bx_n)$---EM adopts \textit{soft assignment}. 
It incorporates the full posterior distribution $p_{\btheta^\toptzero}(\bz_n \mid \bx_n)$ to weight all plausible latent configurations.

In comparison, the alternating maximization algorithm (Algorithm~\ref{alg:am_lvm}) relies on hard assignment: it selects only the single most probable latent state $\widehatbz_n$ under the current posterior for parameter updates. By contrast, the EM algorithm leverages the full posterior uncertainty, yielding more robust, stable parameter updates---particularly for multimodal or highly uncertain latent posterior distributions.

\index{Stochastic EM}
\index{Ascent direction}
\index{Gradient descent}
\paragrapharrow{Stochastic EM algorithm.} 
In large-scale data scenarios, evaluating  the full Q-function across all $N$ data points in every EM iteration incurs prohibitive computational costs. 
To resolve this issue, the \textit{stochastic EM algorithm} is adopted, which approximates the maximization (M) step using a single randomly sampled data point or a small mini-batch. At iteration $t$, after randomly sampling a data index $n$, the parameter update proceeds as follows:
\begin{equation}
\btheta^\toptone \leftarrow \btheta^\toptzero + \eta_t \nabla Q_{\bx_n}(\btheta^\toptzero \mid \btheta^\toptzero),
\end{equation}
where $\eta_t>0$ denotes the step size at iteration $t$, and $\nabla Q_{\bx_n}(\btheta^\toptzero \mid \btheta^\toptzero)$  represents a gradient \textit{ascent direction}. 
On expectation, taking a positive step along this gradient increases the expected log-likelihood of the model \citep{lu2022gradient}.

\index{MAP EM}
\paragrapharrow{Maximum a posteriori (MAP EM).} 
To incorporate prior domain knowledge into parameter estimation, we impose a prior distribution $p(\btheta)$ 
over the model parameter and maximize the \textit{log-posterior objective} instead of the standard log-marginal likelihood. 
The \textit{MAP parameter estimate} is defined as:
\begin{subequations}\label{equation:mapem_all}
\begin{equation}\label{equation:mapem1}
\widehatbtheta_{\mathrm{MAP}} =
\mathop{\argmax}_{\btheta\in\bTheta}
\sum_{n}\ln \int p_{\btheta}(\bz_n, \bx_n) \diff\bz_n +\ln p(\btheta).
\end{equation}
The E-step remains identical to that of the standard EM algorithm, as the parameter prior is independent of the latent variables. 
In contrast, the M-step is modified to incorporate the log-prior regularization term, yielding the \textit{MAP M-step objective}:
\begin{equation}\label{equation:mapem2}
\textbf{MAP M-Step:}\quad  \mathop{\argmax}_{\btheta\in\bTheta} Q(\btheta \mid \btheta^\toptzero)+\ln p(\btheta).
\end{equation}
\end{subequations}

We demonstrate the practical implementation of the EM algorithm using the \textit{Gaussian mixture model (GMM)}, a canonical and widely adopted statistical model.~\footnote{See, for example, \citet{jain2017non, lu2021survey} for further details.} 
GMMs are extensively utilized in clustering, density estimation, and topic modeling (where they extract latent semantic topics from document corpora).
\begin{example}[Gaussian mixture model (GMM)\index{Gaussian mixture model}\index{Mixture of Gaussians}]\label{example:gmm_twoclus}
Consider a Gaussian mixture model (mixture of Gaussians) with two modes or clusters. 
The model parameters are $\btheta=\{\pi_k, \bmu_k, \bSigma_k\}_{k\in\{0,1\}}$, and the joint density is
$$
f_{\btheta} \left( \cdot,\cdot \mid \btheta \right) = \pi_0\cdot \normal_0 + \pi_1 \cdot \normal_1,
$$
where $\normal_k=\normal(\cdot \mid \bmu_k, \bSigma_k), k=\{0,1\}$ denotes a multivariate Gaussian density (Definition~\ref{definition:multivariate_gaussian}), and $\pi_0+\pi_1=1$ are the mixture coefficients.
For simplicity, assume equal mixing proportions ($\pi_0=\pi_1=1/2$)  and shared isotropic covariance matricex ($\bSigma_0=\bSigma_1=\bI$). 
We then aim to estimate only the means: $\btheta=\{\bmu_0, \bmu_1\}$. 
Each observed data point  $\bx_n$ ($n=1,2,\ldots,N$) is associated with a discrete latent variable $z_n\in\{0,1\}$ indicating its component membership: $\normal_0$ or $\normal_1$. 
The joint and posterior distributions are:
$$
\begin{aligned}
p_{\btheta}(\bx_n, z_n  ) &= \normal_{z_n}(\bx_n \mid \bmu_{z_n}, \bSigma_{z_n}),\\
p_{\btheta}(z_n \mid \bx_n) &= \frac{p_{\btheta}(\bx_n, z_n  )}{p_{\btheta}(\bx_n)}
=
\frac{p_{\btheta}(\bx_n, z_n  )}{p_{\btheta}(\bx_n,z_n  )+p_{\btheta}(\bx_n,1-z_n  )}, \quad n\in\{1,2,\ldots,N\}.
\end{aligned}
$$
Accordingly, the Q-function for the EM algorithm is constructed as the expectation of the complete-data log-likelihood under the posterior distribution of latent variables:
$$
\begin{aligned}
Q(\btheta \mid \btheta^\toptzero) 
&=\sum_{n} \bQ_{\bx_n}(\btheta \mid \btheta^\toptzero)
=
\sum_{n}  \underbrace{\sum_{z_n\in\{0,1\}} p_{\btheta^\toptzero}(z_n \mid \bx_n) \ln p_{\btheta}(z_n, \bx_n )}_{\text{$\bQ_{\bx_n}(\btheta \mid \btheta^\toptzero)$}}\\
&=\sum_{n}  \left\{p_{\btheta^\toptzero}(0 \mid \bx_n) \ln p_{\btheta}(0, \bx_n ) + p_{\btheta^\toptzero}(1 \mid \bx_n) \ln p_{\btheta}(1, \bx_n )\right\}.
\end{aligned}
$$
This Q-function admits an analytical closed-form maximization solution in the M-step, enabling explicit update rules for the mean parameters $\bmu_0$ and $\bmu_1$.
The resulting EM algorithm iteratively alternates between two core steps: the E-step computes soft probabilistic cluster responsibilities for each data point, while the M-step updates cluster means via weighted averaging of observed data points.
Generalized extensions of this basic two-component GMM framework, including multi-component mixtures, full-rank covariance matrices, and non-Gaussian mixture distributions, are explored in Problems~\ref{prob:mix_of_gauss}--\ref{prob:mix_of_bern}.
\end{example}

\index{Variational EM algorithm}
\index{Constrained EM algorithm}
\subsection{EM for Constrained Optimization}\label{section:em_constrained}
The standard unconstrained EM algorithm yields the exact latent posterior during the E-step, such that $q^\toptone_{\bz_n}(\bz_n) = p_{\btheta^\toptzero}(\bz_n \mid \bx_n), \, \forall\, n$ (see \eqref{equation:cons_EM_estep}).
In practical scenarios, however, this approach faces critical limitations: real-world data is often governed by multiple interacting latent variables, rendering the true latent posterior  $p_{\btheta^\toptzero}(\bz_n \mid \bx_n),  \, \forall\, n$ computationally intractable  \citep{williams1991mean, ghahramani1995factorial, beal2003variational, turnertwo}. 
To resolve this issue, variational Bayesian methods constrain the posterior to a \textbf{tractable} distribution family, giving rise to the \textit{constrained EM algorithm}, also widely known as the \textit{variational EM (VEM) algorithm}.

Suppose we constrain the approximate posterior to a parametric family $\mathcalQ=\{q_{\bz}(\bz \mid \blambda)= q(\bz \mid \blambda)\mid  \blambda\in{\Lambda}\}$, where $\blambda$ is called the \textit{variational parameter}. 
In practice, the variational distribution for each latent variable $\bz_n$ is conditioned on its corresponding observation $\bx_n$. 
We thus write $q(\bz_n \mid \blambda_n)$, where $\blambda_n=\{\bx_n, \widebarblambda_n\}$, and only the learnable component $\widebarblambda_n$ is updated during model inference.
For simplicity, we denote and optimize over the unified variational parameter $\blambda \in \Lambda$ throughout our analysis.
At iteration $t$, the E-step becomes (using Equation~\eqref{equation:elbo_vfe_neg}, which shows that maximizing the ELBO is equivalent to minimizing the KL divergence when fixing the model parameter $\btheta=\btheta^\toptzero$):
\begin{align}
\textbf{E-Step: } \quad q^\toptone(\bz_n \mid \blambda_n) 
\leftarrow 
&\mathop{\argmax}_{\blambda \in\Lambda} \mathcalF\left(\left\{q(\bz_n \mid \blambda_n)\right\}_{n=1}^N, \btheta^\toptzero\right),\quad \forall\, n\in\{1,2,\ldots,N\} \nonumber\\
=&\mathop{\argmin }_{\blambda \in\Lambda} \sum_{n}\KL\left[ q(\bz_n \mid \blambda_n) \parallel p_{\btheta^\toptzero}(\bz_n \mid \bx_n)\right].
\end{align}
In other words, we seek the  \textit{variational posterior} $q^\toptone(\cdot \mid \blambda_n) $ that is closest---in KL divergence---to the true (but intractable) posterior $p_{\btheta^\toptzero}(\bz_n \mid \bx_n)$.
Since the predefined variational family  $\mathcalQ$ does not generally include the exact true posterior, the ELBO is no longer a tight bound. 
A nonnegative gap persists between the ELBO value and the log-marginal likelihood of the data.

The M-step retains the standard update rule of unconstrained EM (see \eqref{equation:cons_em_mstep}) but adopts the optimized variational posterior in place of the intractable exact posterior:
\begin{align}
\textbf{M-Step: } \quad \btheta^\toptone &\leftarrow \mathop{\argmax}_{\btheta\in\bTheta} 
\sum_{n}  \int q^\toptone(\bz_n \mid \blambda_n)  \ln p_{\btheta}(\bz_n, \bx_n) \diff\bz_n
\end{align}

A natural question arises: how can we minimize the KL divergence when the true posterior $p_{\btheta^\toptzero}(\bz_n \mid \bx_n)$ is analytically intractable? 
The core solution lies in the structural properties of well-designed variational families. If the variational distribution $q(\bz \mid \blambda)$ supports tractable moment computation (e.g., Gaussian distributions, fully specified by first- and second-order moments), the optimization process only requires expectations evaluated under the true posterior, such as $\Exp[\bz_n \mid \bx_n, \btheta^\toptzero]$ and $\Exp[\bz_n\bz_n^\top \mid \bx_n, \btheta^\toptzero]$. We elaborate on this mechanism in greater detail in the following sections.

\index{Mean-field approximation}
\index{Lagrangian function}
\index{Functional derivatives}
\index{Variational derivatives}
\index{Structured variational inference}
\subsubsection*{Mean-Field Approximation of Hidden Variables}
Let $\mathcalX=\mathcalX(\bx_{1:N})=\{\bx_1,\bx_2,\ldots,\bx_N\}$ denote the set of observed data, where each observation satisfies $\bx_n\in\real^D$. 
Similarly, let  $\mathcalZ=\mathcalZ(\bz_{1:N})=\{\bz_1,\bz_2,\ldots,\bz_N\}$ denote the corresponding collection of latent variables, with  $\bz_n\in\real^Q$.
The \textit{mean-field approximation (MFA)} assumes full factorization of the variational posterior across all latent dimensions, formulated as:
$$
q_{\bz_n}(\bz_n) =\prod_{q=1}^{Q} q_{z_{nq}}(z_{nq}), 
\quad \,\forall\, n\in\{1,2,\ldots,N\}.
$$
Under this factorization assumption, the ELBO can be expanded as follows:
\begin{align}
\mathcalF(\{q_{\bz_n}(\bz_n)\}, \btheta) 
&= \sum_{n=1}^N  \int q_{\bz_n}(\bz_n) \ln\frac{p_{\btheta}(\bz_n, \bx_n)}{ q_{\bz_n}(\bz_n)} \diff\bz_n
= \sum_{n=1}^N  \int \prod_{q=1}^{Q} q_{z_{nq}}(z_{nq}) \ln\frac{p_{\btheta}(\bz_n, \bx_n)}{ \prod_{q=1}^{Q} q_{z_{nq}}(z_{nq})} \diff\bz_n \nonumber\\
&= \sum_{n=1}^N  \int \prod_{q=1}^{Q} q_{z_{nq}}(z_{nq}) \ln p_{\btheta}(\bz_n, \bx_n) -  \sum_{q=1}^{Q} q_{z_{nq}}(z_{nq})\ln q_{z_{nq}}(z_{nq})  \diff\bz_n.
\end{align}

To derive the MFA-based E-step at iteration $t$, we maximize the ELBO $\mathcalF\left(\left\{q_{\bz_n}(\bz_n)\right\}, \btheta^\toptzero\right)$ with respect to each individual latent factor   $q_{z_{nq}}(z_{nq})$, subject to the normalization constraints: $\int q_{z_{nq}}(z_{nq}) \diff z_{nq} =1$ for all $n\in\{1,2,\ldots,N\}, q\in\{1,2,\ldots, Q\}$.
We introduce a set of Lagrange multipliers $\{\gamma_{nq}\}$ to construct the constrained Lagrangian function:
$$
L\left(q_{z_{nq}}(z_{nq}), \bgamma\right)=
\mathcalF\left(\left\{q_{\bz_n}(\bz_n)\right\}_{n=1}^N, \btheta^\toptzero\right) + \sum_{n,q} \gamma_{nq} \left(\int q_{z_{nq}}(z_{nq}) \diff z_{nq} -1\right).
$$
Setting the functional  derivative  of the Lagrangian function with respect to $q_{z_{nq}}(z_{nq})$ to zero yields the optimal update rule for each latent factor:
\begin{align}
&\ln q^\toptone_{z_{nq}}(z_{nq}) = \int \left[ \prod_{k\neq q}^{Q}q_{z_{nk}}(z_{nk}) \ln p(\bz_n, \bx_n \mid\btheta^\toptzero)\right]\diff \bz_{n/q} +\gamma_{nq}-1 
\nonumber\\
&\implies q^\toptone_{z_{nq}}(z_{nq}) = \frac{1}{\mathcalC_{nq}}\exp\left\{\int \left[ \prod_{k\neq q}^{Q}q_{z_{nk}}(z_{nk}) \ln p(\bz_n, \bx_n \mid\btheta^\toptzero)\right]\diff \bz_{n/q}\right\},\label{equation:em_uncon_mf}
\end{align}
where $\mathcalC_{nq}$ denotes the normalization constant, $\diff {\bz_{n/q}}$ denotes integration over all dimensions of $\bz_n$ excluding $z_{nq}$, 
and $\prod_{k\neq q}^{Q}$ denotes the product of all elements except the $q$-th item.
More compactly, this update can be written as:
\begin{equation}\label{equation:em_uncon_mf_comp}
q^\toptone_{z_{nq}}(z_{nq}) 
\leftarrow
\frac{1}{\mathcalC_{nq}}\exp\left\{\Exp_{q(-z_{nq})} \left[ \ln p(\bz_n, \bx_n \mid\btheta^\toptzero)\right]\right\},
\end{equation}
where the expectation can be taken over all $n^\prime \in\{1,2,\ldots,N\}, q^\prime\in\{1,2,\ldots, Q\}$ except $\{n^\prime =n, q^\prime =q\}$.
Thus, the ELBO is maximized iteratively: at each step, we update one factor $q_{z_{nq}}(z_{nq})$ while holding all others fixed (i.e., keeping the remaining factors $q_{z_{n^\prime q^\prime}}(z_{n^\prime q^\prime})$ constant, where $n^\prime \in\{1,2,\ldots,N\}$, $q^\prime\in\{1,2,\ldots, Q\}$, and $\{n^\prime \neq n, q^\prime \neq q\}$). This coordinate ascent procedure is repeated until convergence.

A natural extension of mean-field variational inference is \textit{structured variational inference} (\textit{structured VI}) \citep{saul1996mean}, which retains dependencies among subsets of latent variables to relax the strict full-factorial constraint. While structured VI yields more accurate posterior approximations than standard MFA, it increases optimization complexity and often leads to intractable analytical or computational solutions.

\subsection{Final Remarks}
We extend the unconstrained EM algorithm---where the variational distribution $q_{\bz_n}(\bz_n)$ equals the true posterior distribution $p_{\btheta}(\bz_n\mid \bx_n)$ at each iteration---to the constrained EM algorithm. In the constrained setting, the true posterior $p_{\btheta}(\bz_n\mid \bx_n)$ is intractable, and the variational distribution {$q_{\bz_n}(\bz_n) \triangleq q(\bz_n\mid \blambda_n)$} is restricted to a tractable family parameterized by $\blambda_n$.
Full \textit{variational Bayesian inference} generalizes the basic constrained EM algorithm to fully Bayesian settings by treating both local latent variables $\{\bz\}$ and global model parameters $\btheta$ as random variational quantities rather than fixed constants. It factorizes the joint variational posterior over latent variables and model parameters, derives coordinate-ascent variational Bayesian expectation (VBE) and variational Bayesian maximization (VBM) update rules under mean-field assumptions, and unifies EM, MAP estimation, and classical variational inference within a single ELBO optimization framework. This lays complete Bayesian mathematical foundations for subsequent approximate inference algorithms.
For further details, see, for example, \citet{lu2023bayesian}.

Building directly on this Bayesian framework, \textit{Monte Carlo / stochastic variational inference} addresses a key practical limitation of traditional coordinate ascent: many complex, non-conjugate generative models do not yield closed-form ELBO gradients.
Monte Carlo variational inference introduces sample-based stochastic gradient estimators, including the score function and reparameterization trick, paired with variance-reduction techniques such as \textit{Rao--Blackwellization} and \textit{control variates}. This enables tractable numerical optimization for intractable likelihoods even in the absence of analytical closed-form solutions.
For further details, see, for example, \citet{wingate2013automated, ranganath2014black}.

In contrast, \textit{amortized variational inference} resolves the scalability bottleneck caused by per-data-point variational parameters in traditional variational inference. Rather than optimizing independent latent posterior parameters $\blambda_n$ for each individual observation, it trains a shared encoder network that maps any input $\bx$ directly to the variational posterior $q_{\blambda}(\bz\mid\bx)$. This one-time training procedure amortizes inference computation across all samples, enables fast prediction on unseen data, and constitutes the core architectural innovation of variational autoencoders, bridging classical variational theory with deep generative neural networks.
See Section~\ref{section:vae_pca} for more details.

\index{Autoencoder}
\section{Autoencoder and Variational Autoencoder}\label{section:ae_vae}

Building on the variational framework introduced in the previous section, this section extends deterministic \textit{autoencoders (AEs)} to probabilistic \textit{variational autoencoders (VAEs)}.
AEs and their probabilistic variant, VAEs, represent some of the most influential architectures in unsupervised and self-supervised learning. They provide a deep learning framework for learning compact, structured data representations, supporting applications including denoising, anomaly detection, and generative modeling.
Standard AEs compress data into fixed latent points but lack a structured latent space suitable for generative tasks. VAEs address this limitation by encoding inputs into Gaussian latent distributions using amortized neural encoders, optimizing the evidence lower-bound (ELBO) objective, which balances reconstruction error and KL-divergence regularization on the latent space.

A particularly appealing property of these models is that they generalize and extend well-established concepts from classical data analysis, especially matrix decomposition methods such as principal component analysis (PCA) and nonnegative matrix factorization (NMF) \citep{lu2021numerical, lu2022matrix, lu2023bayesian}.
At its core, an AE comprises two transformations, implemented as neural networks: an encoder that maps input data $\bx$ to a lower-dimensional \textit{latent code} (also called a \textit{latent representation} or \textit{embedding}) $\bz$, useful for downstream tasks including information retrieval and matching, dimensionality reduction and visualization, and representation learning for classification and clustering \citep{aggarwal2020linear}; and a decoder that reconstructs the original input from this latent code.

When both the encoder and decoder are constrained to be linear and the reconstruction loss is defined as mean squared error, the optimal solution for an AE with a low-dimensional bottleneck is mathematically equivalent to PCA---a matrix decomposition method that factorizes the data matrix $\bX\in\real^{N\times D}$ into orthogonal components that capture the maximum variance \citep{lu2021numerical, lu2022matrix, lu2023bayesian}.
In this respect, the AE can be seen as a nonlinear, learnable generalization of matrix factorization, where the learned ``factors" are no longer limited to linear subspaces or nonnegativity constraints, but instead capture complex, hierarchical patterns through deep nonlinear mappings.

The VAE extends this foundation by adopting a probabilistic perspective: rather than outputting a single point estimate $\bz$, the encoder produces the parameters of a distribution (typically Gaussian) over the latent space. The VAE then optimizes a lower bound on the log-likelihood of the data---the evidence lower-bound (ELBO)---which trades off reconstruction accuracy against regularization imposed by a prior distribution (usually a standard normal). This regularization promotes a smooth, well-structured latent space, enabling meaningful interpolation and novel sample generation.

Key innovations discussed also include the reparameterization trick for differentiable latent sampling, the $\beta$-VAE for adjustable constraint tradeoffs, discrete vector-quantized AE (VQ-AE) variants, and hierarchical VAEs. This section characterizes VAEs as tractable, interpretable density estimators with smooth latent manifolds that support effective interpolation and generative modeling.

\begin{figure}[h]
\centering
\includegraphics[width=0.99\textwidth]{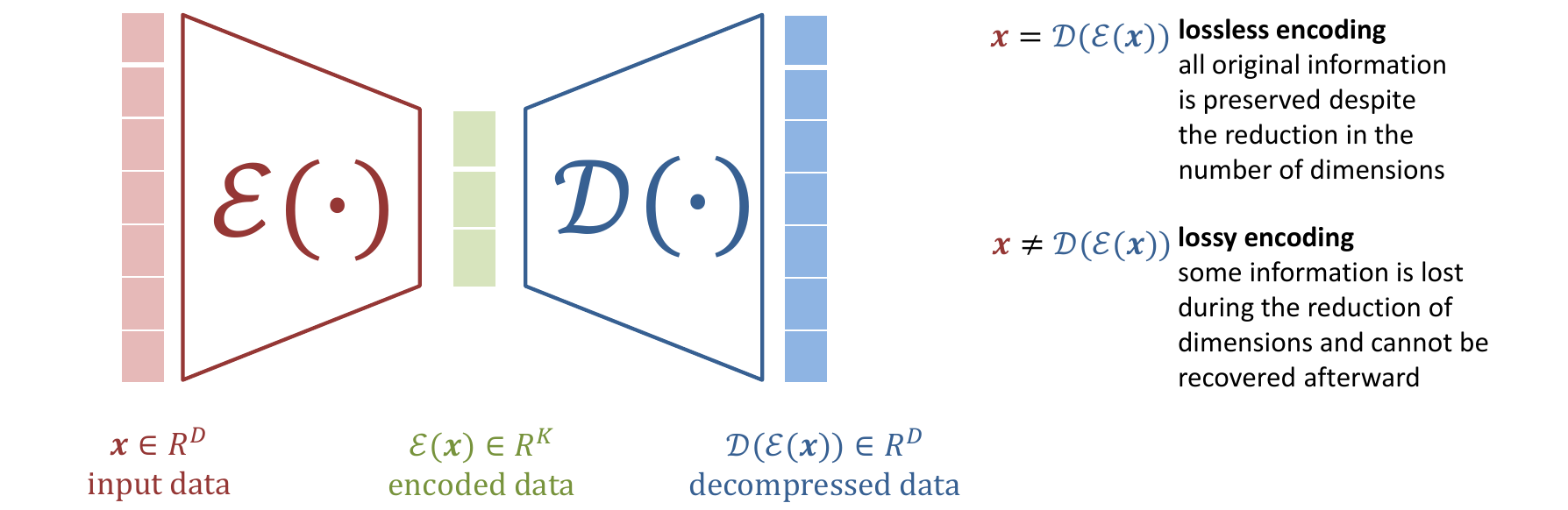}
\caption{Description of an autoencoder.}
\label{fig:autoencoder}
\end{figure}
\subsection{Autoencoder}
In machine learning, an \textit{autoencoder (AE)} performs dimensionality reduction by reducing  the number of features that characterize a dataset (denoted as $\bx$ in Figure~\ref{fig:autoencoder}). 
This process relies on an \textit{encoding stage}, denoted as $\mathcalE(\bx): \real^D \to \real^K$, which either selects a subset of the original input features or constructs a compact set of new, derived features.
The complementary  \textit{decoding stage}, formulated as   $\mathcalD(\mathcalE(\bx)): \real^K \to \real^D$, reconstructs the original input from the resulting \textit{compressed  representation}. 
In practice, the latent dimension  $K$ is consistently set much smaller than the original input dimension $D$ to achieve effective compression. 
For image data, for instance, an input may have a dimensionality of $D = 3 \times 1024 \times 1024$ (corresponding to  channels, height, and width dimensions). 
It is common to downsample such inputs to a latent dimension of $K = 3 \times \frac{1024}{16} \times \frac{1024}{16}$. 
Collectively, the encoding function $\mathcalE(\cdot)$ and decoding function $\mathcalD(\cdot)$ constitute a complete autoencoder model.

Ideally, $\mathcalE(\cdot)$ and $\mathcalD(\cdot)$ are optimized to yield high-fidelity reconstruction, such that the reconstructed output $\mathcalD(\mathcalE(\bx))$ closely matches the original input $\bx$ on average. However, compression via AEs is often \textbf{lossy}, depending on the underlying data distribution, latent space dimensionality, and encoder architectural design. This means certain input information is irreversibly discarded during encoding and cannot be fully recovered in the decoding phase.

Accordingly, the core objective of AE learning is to identify the optimal pair of encoder and decoder functions from a predefined function family. Specifically, given a set of feasible encoder functions $\sE$ and decoder functions $\sD$, the goal is to learn the encoder-decoder pair that maximizes information retention during encoding and consequently minimizes reconstruction error during decoding. This learning paradigm yields the following optimization formulation:
$$
(\mathcalE,\mathcalD) = \mathop{\argmin}_{(\mathcalE,\mathcalD)\in(\sE,\sD)} G\left(\bx, \mathcalD(\mathcalE(\bx))\right),
$$
where $G(\cdot, \cdot)$  denotes a loss function that quantifies the discrepancy between the original input and its reconstructed counterpart. The function families $\sE$ and $\sD$ can be instantiated with various flexible function approximators, most commonly multilayer perceptrons and deep neural networks \citep{lecun2015deep, goodfellow2016deep}. 

Notably, when both the encoder and decoder are constrained to linear transformations---forming a \textit{linear autoencoder}---the optimal solution is mathematically equivalent to \textit{principal component analysis (PCA)} or \textit{singular value decomposition (SVD; \textcolor{black}{Theorem~\ref{theorem:reduced_svd_rectangular}})}, provided the loss function $G$ is defined based on the Frobenius norm or spectral norm (see Problem~\ref{theorem:young-theorem_frob} or \citet{lu2021numerical, lu2022matrix}).
\footnote{Strictly speaking, PCA (or SVD) corresponds to a special case of linear AEs where the learned basis vectors are orthonormal. In contrast, generic linear AEs impose no orthogonality or normalization constraints on their weight vectors. Further discussions can be found in \citet{lu2021numerical, lu2023bayesian}.}
As a result, the weight matrices that define the linear transformation in Figure~\ref{fig:autoencoder} span the \textit{principal subspace} of the input data, though their corresponding weight vectors are not necessarily orthogonal or unit-normalized. This mathematical equivalence is intuitive: both PCA and linear AEs implement linear dimensionality reduction by minimizing the same sum-of-squares reconstruction loss.

A natural approach to overcome the representational limitations of linear manifolds is to integrate nonlinear activation functions, as widely adopted in deep neural networks. With nonlinear components, more sophisticated AE architectures are capable of achieving aggressive dimensionality reduction while maintaining low reconstruction error. Intuitively, encoders and decoders with sufficient expressive capacity can compress arbitrarily high-dimensional input data into a one-dimensional latent space. Theoretically, an encoder with unlimited representational power could map $N$ distinct input data points to unique scalar values ranging from $1$ to $N$ (or arbitrary integers on the real line), and the paired decoder could perfectly invert this mapping to recover the original data with zero information loss.

Nevertheless, two critical caveats must be considered in this context. First, lossless dimensionality reduction achieved via over-parameterized AEs typically comes at the cost of latent space regularity, resulting in latent representations that lack \textbf{interpretable} and structurally exploitable properties. Second, the ultimate goal of dimensionality reduction is not merely to reduce input dimensionality, but to preserve the essential structural information of the original data within the low-dimensional latent representations. For these reasons, the latent dimension size and the architectural complexity (i.e., depth) of AEs---which jointly determine the degree and quality of data compression---must be carefully tuned according to the downstream tasks of dimensionality reduction.

\paragrapharrow{Solution via truncated SVD or PCA.}
As noted above, when the encoder and decoder are restricted to linear transformations under mean squared error loss, the optimal solution is mathematically equivalent to principal component analysis (PCA) or singular value decomposition (SVD).
Assume the data are centered such that the sample mean $\widebarbx=\bzero$. 
Let the data matrix $\bX \in \real^{N\times D}$ hold the $N$ observations as rows.
As shown in Problem~\ref{theorem:young-theorem_frob}, the \textit{truncated SVD (TSVD; \textcolor{black}{Theorem~\ref{theorem:reduced_svd_rectangular}})}---which zeros out all but the top $K$ singular values---yields  the best rank-$K$ approximation of $\bX$ in the Frobenius norm. Denote this approximation by $\widetildebX=\bU_K\bSigma_K\bV_K^\top$, where $\bU_K\in\real^{N\times K}$, $\bV_K\in\real^{D\times K}$, and $\bSigma_K\in\real^{K\times K}$. 
From the perspective of PCA, the encoder maps each data point $\bx_n\in\real^D$ (the $n$-th row of $\bX$) to its coordinates in the principal subspace:
\begin{equation}
\text{encoder: }\quad \mathcalE(\bx_n) = \bV_K^\top \bx_n, \quad \bx_n\in\real^{D},\quad \forall\, n\in\{1,2,\ldots, N\}.
\end{equation}
Here, the columns of $\bV_K$ are the orthonormal eigenvectors associated with the $K$ largest eigenvalues of the data covariance matrix \citep{lu2021numerical, lu2023bayesian}. 
Since $\bV_K^\top\bV_K=\bI_K$, it follows that $\bV_K^\top\widetildebX^\top=\bSigma_K\bU_K^\top$.
The corresponding decoder reconstructs the input by projecting back into the original space:
\begin{equation}
\text{decoder: }\quad \mathcalD(\mathcalE(\bx_n)) = \bV_K \mathcalE(\bx_n), \quad  \forall\, n\in\{1,2,\ldots, N\}.
\end{equation}
Thus, the full reconstruction is $\mathcalD(\mathcalE({\bX}^\top))=\bV_K\bSigma_K\bU_K^\top$, which is exactly  the truncated SVD of $\bX^\top$. 
This confirms that when $\sE$ and $\sD$ are linear, the AE is equivalent to PCA/SVD.

\paragrapharrow{Multi-aspect information.}
Autoencoders are trained to reconstruct their input from a compressed latent code. This objective encourages the encoder to learn a representation that preserves all information necessary for accurate reconstruction, which inherently captures distinct, meaningful aspects of the data.
For example, in images, this includes not only the overall shape of objects but also textures, colors, and spatial relationships; in audio, it encompasses linguistic content, speaker identity, pitch, and prosody; in text, it covers syntactic structure, semantic meaning, and even stylistic nuances.
Overall, this multi-aspect information arises for the following reasons:
\begin{itemize}
\item \textit{The reconstruction objective as a ``forcing function".}
The core training signal of an AE is the reconstruction loss: the decoder must produce an output that matches the original input as closely as possible. To achieve this, the encoder cannot discard critical information. Instead, it must encode every feature the decoder requires to faithfully recover the input.

\item \textit{The latent space as a structured information bottleneck.}
The latent vector serves as a compressed information bottleneck. Since it has significantly fewer dimensions than the raw input, the encoder is forced to organize information efficiently. This promotes the emergence of disentangled or structured representations, where different dimensions or directions in the latent space correspond to distinct attributes of the data.
For instance, one direction in the latent space might control ``speaker identity" in audio, while another governs ``emotional tone."
This structured layout enables the latent vector to simultaneously encode multiple independent factors of variation present in the data.

\item  \textit{Unsupervised learning of inherent data structure.}
Importantly, these distinct aspects are learned in an unsupervised manner, without explicit labels for properties such as texture or semantics. The AE identifies these aspects because they represent the underlying factors that explain variation in the training data.
If changing the speaker's voice alters the audio signal, the model learns to encode this variation in the latent space.
Similarly, if modifying the syntax of a sentence changes its representation, the model learns to capture this distinction.
\end{itemize}
In summary, the combination of the reconstruction objective and the compressed bottleneck drives the AE to learn a rich, structured latent representation that encodes the full range of distinct aspects present in the input data.

\paragrapharrow{Anomaly detection.}
Once an AE has been trained on a collection of normal data samples, it can be used to detect anomalous samples.
When a previously unseen sample is passed into the trained AE:
\begin{itemize}
\item Normal samples: The AE has thoroughly learned their underlying patterns, so the reconstruction loss remains small and the output closely matches the input.
\item  Anomalous samples: The model has not encountered such patterns during training and therefore cannot reconstruct them accurately, leading to a large reconstruction loss. The large reconstruction loss is used as the signal to flag the sample as an anomaly.
\end{itemize}

\paragrapharrow{Other formulations.}
AE  can also be trained by directly minimizing a reconstruction loss:
\begin{equation}\label{equation:ae_mse_first}
\mathcalJ(\blambda,\btheta) 
=\Exp_{\bx \sim p_{\text{data}}} \left[ \normtwo{\mathcalD_\btheta(\mathcalE_\blambda(\bx)) - \bx}^2 \right]
\simeq  \frac{1}{2}\sum_{n=1}^{N}\normtwo{\mathcalD_\btheta (\mathcalE_\blambda(\bx_n )) - \bx_n}^2,  
\end{equation}
where $\blambda, \btheta$ represent the model parameters (e.g., the weights of a neural network), and we assume the dataset $\mathcalX=\{\bx_1,\bx_2,\ldots,\bx_N\}$ is sampled such that $\rvx\sim p_{\text{data}}$.

In general, the encoder  $\mathcalE_\blambda(\bx_n)$ can be implemented using various model architectures---including linear models, nonlinear models, or deep neural networks \citep{bishop2006pattern}. 
However, to prevent the model from simply learning the identity mapping (which would achieve perfect reconstruction but provide no useful compression), it is essential to constrain the capacity of the latent representation. One common approach is to limit the dimensionality of the hidden (bottleneck) layer, as noted previously.
An alternative strategy is to encourage sparsity in the internal representation using regularization. A popular choice is the \textit{$\ell_1$ penalty}, which promotes sparse activations \citep{lu2026first}. 
This yields the following regularized objective function:
\begin{equation}
\mathcalJ(\blambda,\btheta) = \frac{1}{2}\sum_{n=1}^{N}\normtwo{\mathcalD_\btheta (\mathcalE_\blambda(\bx_n )) - \bx_n}^2
+
\lambda\normone{\btheta},
\end{equation}
where $\lambda\in\real_+$ controls the strength of regularization.

Another effective approach is the \textit{denoising autoencoder} \citep{vincent2008extracting}, which encourages the model to learn robust data representations by training it to reconstruct clean inputs from corrupted versions. 
Specifically, each input $\bx_n$ is artificially corrupted (e.g., by adding noise) to form $\widetildebx_n$, which is then fed into the AE. The model is trained to minimize the loss:
\begin{equation}
\mathcalJ(\blambda,\btheta) = \frac{1}{2}\sum_{n=1}^{N}\normtwo{ \mathcalD_\btheta (\mathcalE_\blambda(\widetildebx_n ))  - \bx_n}^2.
\end{equation}
By learning to ``undo" the corruption, the model captures meaningful structural properties of the data. 
For example, in image data, it may learn that neighboring pixels are strongly correlated, allowing it to correct noisy or missing pixel values.

The most common form of corruption employs additive Gaussian noise. Alternatively, when certain input dimensions are randomly masked out (set to zero), the model is known as a \textit{masked autoencoder}. For instance, \citet{he2022masked} used deep networks to reconstruct complete images from partially observed (masked) inputs.

\index{Text latent representation}
\paragrapharrow{Text representation.}
In the \textit{seq2seq2seq autoencoder} (also called an \textit{unsupervised summarization autoencoder}), the latent space consists not of real-valued or binary vectors, but of natural language text sequences \citep{wang2018learning}.
The framework operates as follows:
\begin{itemize}
\item \textit{Encoder (first seq2seq stage).} 
The encoder takes an input document and compresses it into a shorter text sequence (the latent representation), for instance using Transformer architectures \citep{vaswani2017attention}. This sequence is not a numerical vector, but a natural-language summary of the original document.

\item \textit{Decoder (second seq2seq stage).} The decoder then takes this summary and reconstructs the full original document from it.
\end{itemize}
The AE's training objective---reconstructing the input from the latent representation---ensures that the intermediate text sequence acts as a semantically faithful summary. It must capture all key information, core meaning, and essential details needed to regenerate the original document.
However, this latent text representation may not be human-readable. To produce human-readable outputs from the encoder, a \textit{discriminator} can be used to constrain the generator's output to match the style and structure of human-written sentences.

\index{Binary latent representation}
\paragrapharrow{Binary latent representation.}
In AEs, a binary representation is a discrete latent encoding scheme in which the encoder maps input data into a vector of binary values (typically 0 or 1). Unlike continuous latent representations, which use real-valued vectors, binary representations rely on a set of independent bits, where each bit encodes a discrete feature or attribute of the input.

Each binary dimension in the latent vector corresponds to a distinct, interpretable property of the input---for example, one bit may encode the presence or absence of glasses, while another indicates the gender of a subject. During training, the AE learns to set these bits so that the decoder can still accurately reconstruct the original input. This creates a structured bottleneck, forcing the model to compress the input into a set of discrete, independent features rather than arbitrary real-valued vectors.

Binary representations offer several advantages: they reduce latent-space dimensionality, improve interpretability by assigning clear semantic meaning to each bit, and enable efficient manipulation of specific attributes (e.g., flipping a single bit to toggle a feature such as glasses on or off). They also bridge continuous latent spaces and fully discrete tokenization schemes (such as one-hot vectors), providing a structured intermediate framework for representation learning.

\index{Codebook}
\index{Vector quantized-AE}
\index{VQAE}
\index{VQVAE}
\subsection{Vector Quantized-AE (VQAE)}
The  \textit{vector quantized-autoencoder (VQAE)} is another simple variant of the AE that learns  discrete representations \citep{van2017neural}.
In addition to an encoder and decoder, the VQAE also learns a \textit{codebook}. 
For instance, in speech processing, the codebook encodes phonetic information \citep{chorowski2019unsupervised}.

The VQAE operates as follows.
First, input data (e.g., an image) is mapped by the encoder $ \mathcalE(\bx) $ to a continuous latent  representation $ \bz_e(\bx) $:
\begin{equation}
\bz_e(\bx) = \mathcalE(\bx).
\end{equation}
This representation $ \bz_e(\bx) $ is a real-valued vector that captures the features of the input data, yet remains continuous---much like the latent representations in standard AEs.

Next comes the vector quantization module, the core innovation of the VQAE.
In this stage, the continuous representation $ \bz_e(\bx) $ is projected into a discrete embedding vector space (a codebook, typically denoted $ \be_d $). This space contains $ Q $ embedding vectors $ \be_1, \be_2, \ldots, \be_Q $, each of fixed dimensionality and representing a distinct point in the embedding space.
To map the continuous representation $ \bz_e(\bx) $ to a discrete embedding vector, the VQ module selects the embedding vector $ \be_d $ that is nearest to $ \bz_e(\bx) $:
\begin{equation}
\bz_q(\bx) = \be_d, \quad \text{where} \quad d = \argmin_j \normtwo{\bz_e(\bx) - \be_j }.
\end{equation}
This operation converts the continuous latent vector $ \bz_e(\bx) $ into a discrete representation $ \bz_q(\bx) $, which corresponds to one of the embedding vectors $ \be_d $ in the codebook.

\begin{figure}[h]
\centering  
\vspace{-0.2cm} 
\subfigtopskip=2pt 
\subfigbottomskip=2pt 
\subfigcapskip=-5pt 
\includegraphics[width=0.95\textwidth]{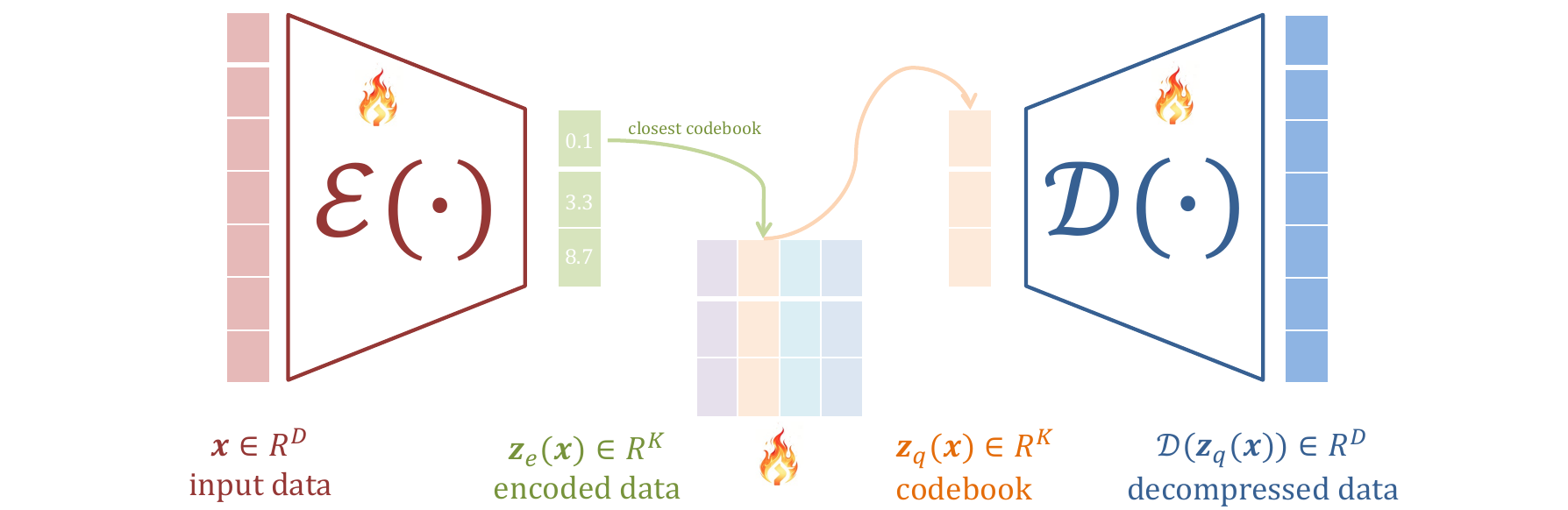}
\caption{Illustration of a VQAE. The ``flame" symbol indicates that the corresponding component is learned from data.}
\label{fig:illus_vqae}
\end{figure}

The quantized discrete embedding vector $ \bz_q(\bx) $ is then fed into  the decoder $ \mathcalD(\bz_q(\bx)) $ to reconstruct the input:
\begin{equation}
\widehatbx = \mathcalD(\bz_q(\bx)).
\end{equation}
The decoder reconstructs the quantized discrete embedding back into the original data space, producing an output $ \widehatbx $ that approximates the input $\bx$ as closely as possible.
An illustration of this process is shown in Figure~\ref{fig:illus_vqae}.

\index{Straight-through estimator}
\index{Stop gradient operator}
\paragrapharrow{Optimization of VQAE.}
To train the encoder and decoder, we first define the overall optimization objective for the VQAE. Since the VQAE is fundamentally an AE, its loss should include the reconstruction error between the original input and the reconstructed output:
\begin{equation}
\mathcalJ_{\text{VQAE-Recon}} = \normtwo{\bx - \mathcalD(\bz_q(\bx))}^2.
\end{equation}
However, direct optimization using this loss is infeasible. 
In the loss function, $ \bz_q(\bx) $ serves as the decoder input. 
The mapping from the encoder output $ \bz_e(\bx) $ to $ \bz_q(\bx) $ is non-differentiable,
so gradients cannot be backpropagated from the decoder to the encoder via standard gradient descent.
A natural solution is to directly copy gradients from $ \bz_q(\bx) $ to $ \bz_e(\bx) $.

The VQAE employs the \textit{straight-through estimator} to achieve this gradient copying \citep{bengio2013estimating}. 
This technique decouples forward and backward computations, enabling custom gradient definitions for non-differentiable operations. Central to this approach is the \texttt{sg} (stop-gradient) operator:
\begin{equation}
\text{sg}(\bx) =
\begin{cases}
\bx, & \text{(in forward propagation)}; \\
\bzero, & \text{(in backward propagation)}.
\end{cases}
\end{equation}
In short, the \texttt{sg} operator preserves its input during the forward pass but suppresses all gradients during backpropagation (i.e., it acts as a constant for differentiation).

Using this operator, we reformulate the reconstruction loss to copy gradients from $ \bz_q(\bx) $ to $ \bz_e(\bx) $:
\begin{equation}
\mathcalJ_{\text{VQAE-Recon}} = \normtwo{\bx - \mathcalD\big( \bz_e(\bx) + \text{sg}\left[ \bz_q(\bx) - \bz_e(\bx) \right] \big)}^2.
\end{equation}
During forward propagation, the decoder uses the quantized vector $ \bz_q(\bx) $ to compute the loss;
during backpropagation, gradients flow as if the decoder input were $ \bz_e(\bx) $:
$$
\mathcalJ_{\text{VQAE-Recon}} =
\begin{cases}
\normtwo{\bx - \mathcalD(\bz_q(\bx))}^2, & \text{(in forward propagation)}; \\
\normtwo{\bx - \mathcalD(\bz_e(\bx)) }^2, & \text{(in backward propagation)}.
\end{cases}
$$
This technique enables gradient propagation and allows stable training of the encoder and decoder.

\paragrapharrow{Optimization of  the embedding space.}
The preceding discussion assumes a pre-trained embedding space. We now describe how the codebook is optimized.
The goal of embedding space learning is for each codebook vector to summarize a cluster of encoder outputs.
For example, a vector representing blurry nighttime street scenes should aggregate encoder outputs from all images of dimly lit urban roads with low visibility and nighttime lighting characteristics.
Thus, each embedding vector should be as close as possible to its matched encoder output. Letting $ \bz_e(\bx) $ denote the encoder output and $ \bz_q(\bx) $ its nearest codebook neighbor, we define the loss to be:
$$
\mathcalJ_{\text{vq}} = \normtwo{\bz_e(\bx) - \bz_q(\bx)}^2.
$$

However, \citet{van2017neural} argued that the encoder and embedding vectors should be updated at different rates. They again used the \texttt{sg} operator to split this loss into two terms, where $ \beta $ controls the relative learning rate of the encoder. The authors found the algorithm is robust to $ \beta $, with values between $0.1$ and $2.0$ performing similarly well:
\begin{equation}
\mathcalJ_{\text{vq}} = \underbrace{\normtwo{\text{sg}[\bz_e(\bx)] - \bz_q(\bx)}^2 }_{\text{Codebook loss}}
+ \beta \cdot \underbrace{\normtwo{\bz_e(\bx) - \text{sg}[\bz_q(\bx)]}^2}_{\text{Commitment loss}}.
\end{equation}
The first term originates from the classic vector quantization  algorithm in dictionary learning and serves to optimize the codebook embedding space, hence known as the \textit{codebook loss}. The \texttt{sg} operator ensures this term provides no gradients to the encoder parameters and only updates the codebook embedding vectors.
The second term is known as the \textit{commitment loss}, which regularizes the encoder's outputs to prevent them from drifting too far from the codebook vectors. Here, the \texttt{sg} blocks gradients from updating the embedding vectors, so this term only affects the encoder parameters.

Rather than learning codebook vectors via gradient descent on the codebook loss term ($\normtwo{\text{sg}[\bz_e(\bx)] - \bz_q(\bx)}^2$), we can update them directly using running  averages of the encoder outputs assigned to each code. 
This approach typically results in more stable and faster training.
The core idea is that each codebook vector $\be_q$ should lie at the centroid (mean) of all encoder outputs mapped to it. In practice, an \textit{exponential moving average (EMA)} maintains this centroid as a running average across training batches.

\paragrapharrow{The final loss.}
Combining all components, the full VQAE loss function is defined as follows. We introduce a hyperparameter $\alpha$ to balance the contributions of the different loss terms alongside the reconstruction error:
\begin{equation}
\mathcalJ_{\text{VQAE}} = \mathcalJ_{\text{VQAE-Recon}}
+ \alpha\cdot \normtwo{\text{sg}[\bz_e(\bx)] - \bz_q(\bx)}^2 
+ \beta \cdot\normtwo{\bz_e(\bx) - \text{sg}[\bz_q(\bx)]}^2.
\end{equation}

The VQAE is also widely referred to as  \textit{VQVAE} in the literature, as it retains the VAE framework (see Section~\ref{section:vae}) while adopting a specific structural choice:
the posterior $q(\bz\mid\bx)$ is deterministic and categorical, assigning probability 1 to the single nearest codebook index and 0 to all others (a one-hot distribution); and the prior $p(\bz)$ over the discrete codes is assumed uniform during training.
The KL divergence between a one-hot categorical $q$ and a uniform prior over $Q$ categories is:
$\KL[q\parallel p] = \ln  Q$, which is constant and independent of encoder parameters.
Only the reconstruction, codebook, and commitment losses remain.
Structurally, the model is therefore a VAE: it includes an encoder that produces a posterior, a prior distribution, and an ELBO objective. The variational framework simply becomes degenerate, as the KL term reduces to a constant due to the deterministic one-hot posterior and uniform prior.

\index{Variational autoencoder}
\subsection{Variational Autoencoder (VAE)}\label{section:vae}

Thus far, we have introduced the dimensionality reduction task and established autoencoders (AEs) as fundamental encoder-decoder architectures trainable via gradient descent. Building on this foundation, we now extend the discussion to generative modeling, elaborate on the critical limitations of standard AEs for content generation, and formally introduce \textit{variational autoencoders (VAEs)} as a principled solution to these drawbacks.

\begin{figure}[h]
\centering
\includegraphics[width=0.99\textwidth]{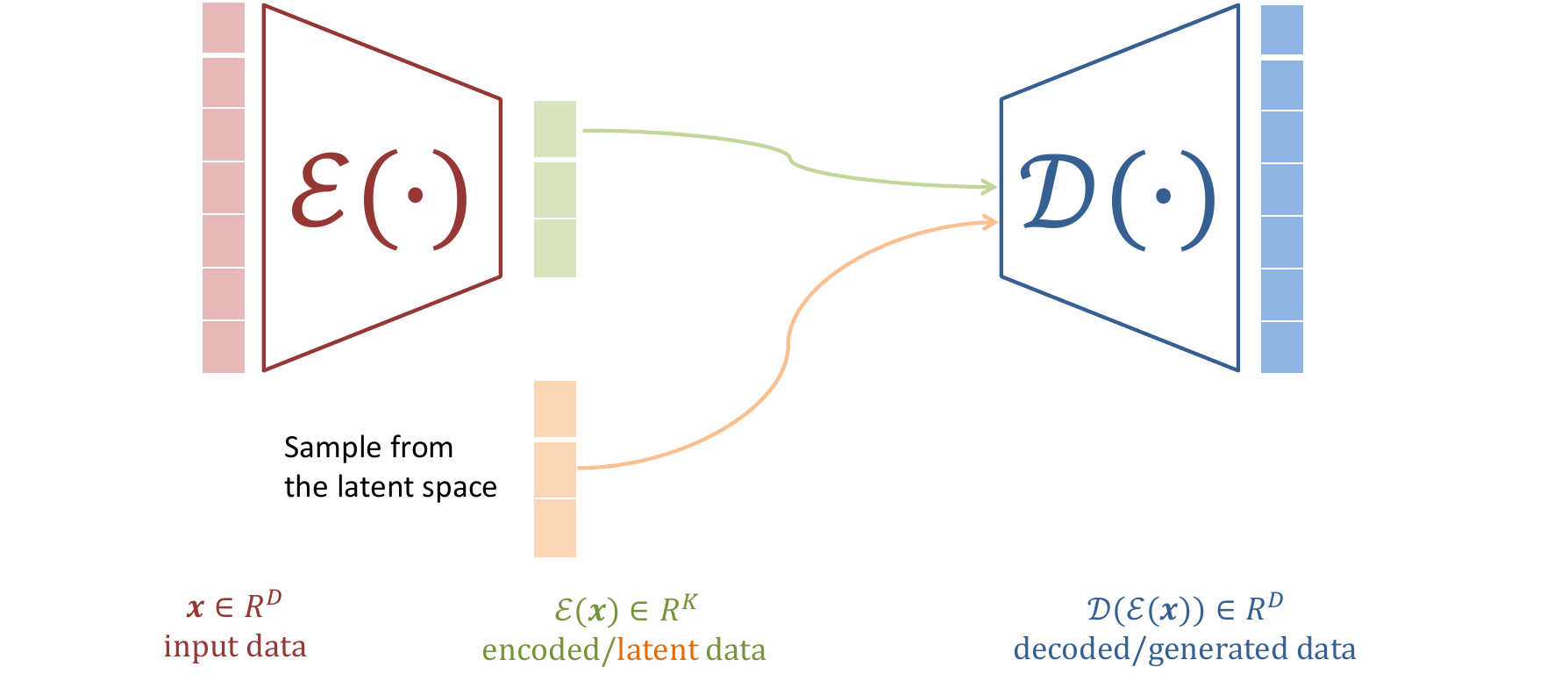}
\caption{
Once an AE has been trained, new data can be generated by decoding points randomly sampled from the latent space.
The quality and relevance of the resulting generated data rely on the regularity of the latent space.
}
\label{fig:ae_gen}
\end{figure}

\paragrapharrow{AE as a generative model.}
Once trained, a standard AE yields a paired encoder and decoder, yet it cannot inherently generate novel data samples. Intuitively, if the encoder produces a well-structured, regular latent space during training, we can sample arbitrary latent vectors from this space and decode them to synthesize new data. The perceptual quality and semantic coherence of such generated samples are entirely dependent on the \textbf{structural regularity} of the learned latent space \citep{anwar2021vae}, as visualized in Figure~\ref{fig:ae_gen}.

However, standard AEs lack inherent constraints to guarantee a regular, well-organized latent space. The structure of the AE latent space is implicitly determined by the input data distribution, latent dimensionality, and network architecture, with no explicit regularization during training. This makes it impossible to rigorously ensure that the learned latent space is suitable for reliable generative sampling.
The core limitation stems from the AE's training objective. The standard reconstruction loss defined in \eqref{equation:ae_mse_first} only optimizes accurate input reconstruction and imposes no constraints on the latent distribution $p_{\text{latent}}(\bz)$, where latent vectors are computed as $ \bz = \mathcalE_\blambda(\bx) $ for training samples $ \bx \sim p_{\text{data}} $. For generative modeling, we aim to learn a valid latent distribution that can be sampled stochastically, whose samples are then decoded via $ \mathcalD_\btheta $ to approximate the true data distribution $p_{\text{data}}(\bx)$. Unfortunately, standard AEs offer no control over the shape and properties of $p_{\text{latent}}(\bz)$.

Without explicit regularization, the learned latent distribution often turns out highly irregular, fragmented, and intractable for stochastic sampling---even when the AE achieves nearly perfect reconstruction accuracy. The model only prioritizes fitting training data and may compress the original data distribution into a disordered latent space that cannot support valid generative sampling. To resolve this critical limitation and enforce a well-behaved, tractable latent distribution for generation, we reformulate the deterministic AE framework into a principled probabilistic generative model, which yields the VAE paradigm.

This latent space irregularity issue is further illustrated by an extreme one-dimensional toy example. Consider a sufficiently powerful AE that can map all training samples to distinct points on a one-dimensional latent axis and perfectly reconstruct the original inputs with zero reconstruction error. While this achieves optimal reconstruction performance, the unconstrained latent space leads to severe overfitting. Large regions of the latent space will not correspond to valid data patterns, and decoding latent points from these unoccupied regions produces meaningless, incoherent outputs. Although this example is intentionally extreme, the underlying problem is universal for standard AEs: unregularized latent spaces consistently degrade generation performance.

\begin{figure}[h]
\centering
\includegraphics[width=0.99\textwidth]{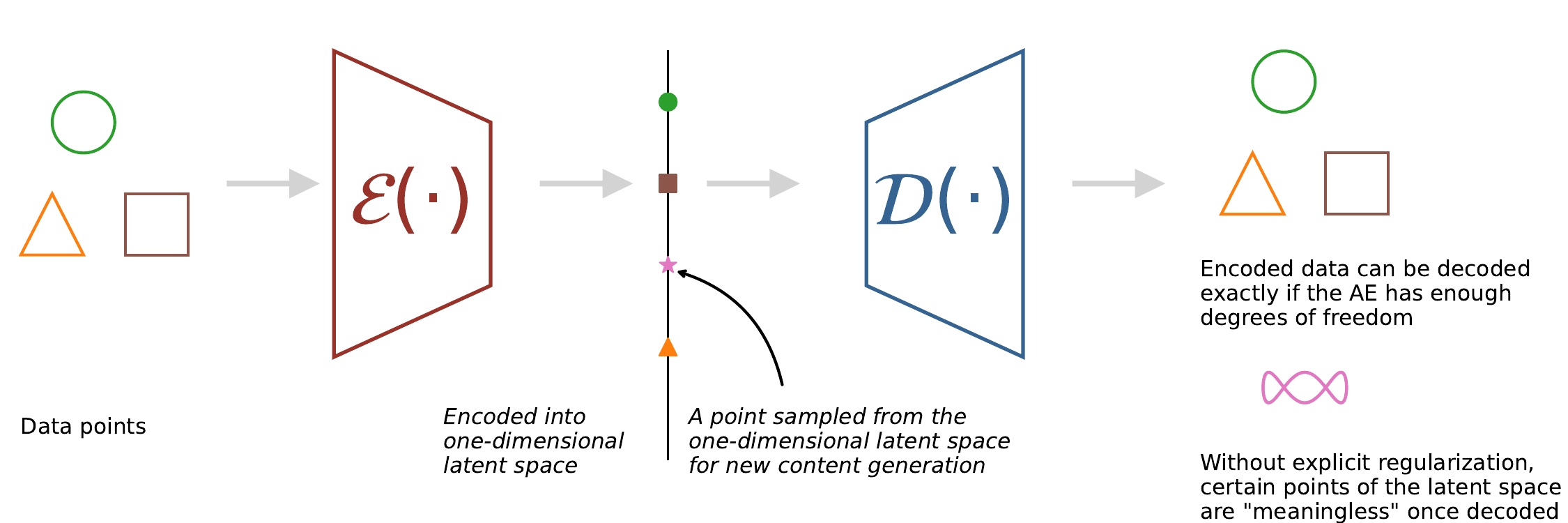}
\caption{Conceptual illustration of an irregular latent space that inhibits the use of AEs for generating new content.
Adapted from \citet{anwar2021vae}.}
\label{fig:vae_latent_irregular}
\end{figure}

This inherent flaw of unregularized AE latent spaces is universal across model architectures. Since the standard AE training objective only optimizes reconstruction accuracy with no constraints on latent space structure, the model inevitably overfits the training data and produces disorganized latent distributions. As visualized in Figure~\ref{fig:vae_latent_irregular}, randomly sampling from such unstructured latent regions generates noisy, semantically invalid, and nonsmooth data samples. By contrast, VAEs address this core issue by integrating probabilistic modeling and explicit latent regularization, constraining the latent space to follow a regular, sampler-friendly distribution and enabling stable, high-quality generative modeling.

\paragrapharrow{From AE to VAE.}
As discussed above, to use the decoder of an AE for generative purposes, we must ensure that the latent space is sufficiently \textbf{regular}.
One way to achieve such regularity is to introduce explicit regularization during training.
Thus, a VAE can be viewed as an AE whose training is regularized to prevent overfitting and ensure the latent space exhibits favorable properties that support generative modeling.

\begin{figure}[htp]
\centering
\includegraphics[width=0.99\textwidth]{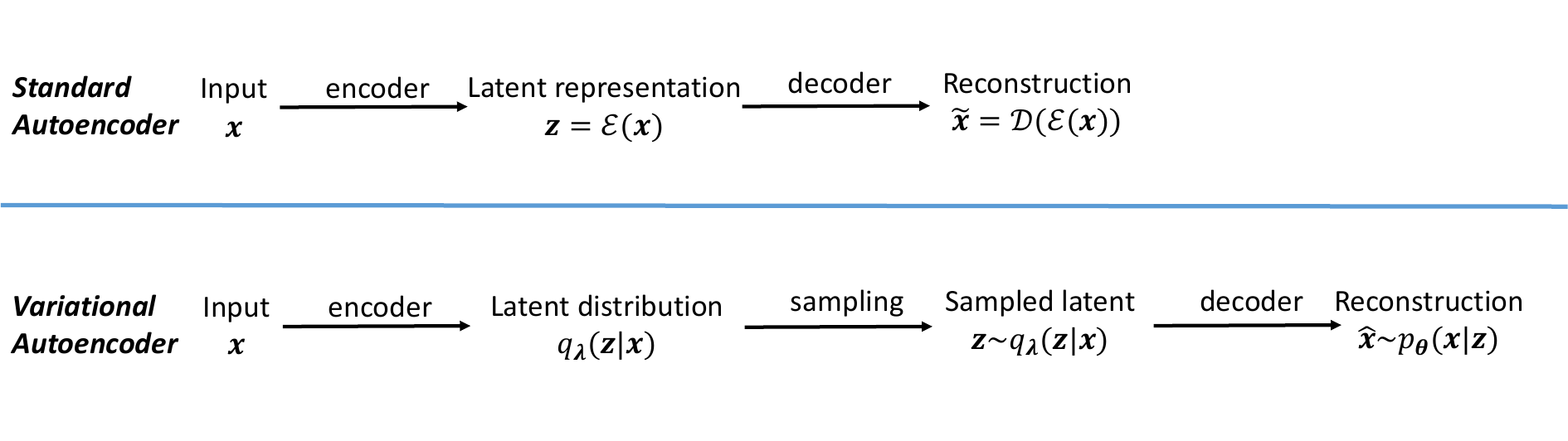}
\caption{Comparison between a standard AE and a VAE.}
\label{fig:ae_vae}
\end{figure}

Like a standard AE, a VAE is an architecture consisting of an encoder and a decoder, trained to minimize the reconstruction error between the reconstructed data and the original input.
However, to impose regularization on the latent space, we slightly modify the encoding-decoding process: instead of mapping an input to a single point in latent space, we encode it as a distribution over the latent space. The model is then trained as follows:
\begin{itemize}
\item The input is encoded as distribution over the latent space;
\item A point is sampled from this distribution in the latent space;
\item The sampled point is decoded, and the reconstruction error is computed;
\item The reconstruction error is backpropagated through the network.
\end{itemize}
See Figure~\ref{fig:ae_vae} for an illustration for the comparison between a standard AE and a VAE.

In practice, the encoded distributions are modeled as normal distributions, allowing the encoder to be trained to output the mean and covariance matrix that define these Gaussians.
One reason to encode an input as a distribution with non-zero variance, rather than a single point, is that it naturally enables \textbf{latent space regularization}: the distributions produced by the encoder are encouraged to remain close to a standard normal distribution. 
A second motivation for distributional encoding relates to \textbf{guidance for the content generation} (see Section~\ref{section:other_issue_vae}). 
Without a probabilistic encoding, when training on text-image pairs that share the same text prompt but correspond to different images, the encoder would map them to the same deterministic latent point.
This creates a critical training inconsistency: diverse visual content associated with identical textual conditions is compressed into a single fixed latent vector, preventing the model from capturing visual variation within the same prompt.
Consequently, the generator fails to learn meaningful generative patterns and tends to produce blurry, repetitive, low-diversity images for identical text prompts, severely limiting generalization and generation flexibility.

Thus, the loss function minimized during VAE training consists of two components:
a ``reconstruction term" (at the output layer), which optimizes the fidelity of the encoding-decoding process,
and a ``regularization term" (at the latent layer), which structures the latent space by encouraging the encoder's output distributions to approximate a standard normal distribution.
This regularization term is formulated as the KL divergence (see \eqref{equation:kl_def_vae}) between the learned distribution and a standard Gaussian, whose motivation is further explained in the next section.

\paragrapharrow{Continuity and completeness from the regularization.}
The regularity required of the latent space to enable generative modeling can be characterized by two key properties: 
\textbf{continuity} (two nearby points in the latent space should not produce completely different content when decoded) and \textbf{completeness}  (for a given target distribution, any point sampled from the latent space should yield ``meaningful" content after decoding).
For example, when sampling a latent vector between the square
``\begin{tikzpicture}[line width=1.5pt, color=brown]
\draw (0,0) -- (0.25,0) -- (1/4,1/4) -- (0,1/4) -- cycle;
\end{tikzpicture}"
and the triangle
``\begin{tikzpicture}[line width=1.5pt, color=orange]
\draw (0,0) -- (0.25,0) -- (0.25/2,1/4) -- cycle;
\end{tikzpicture}"
in Figure~\ref{fig:vae_latent_irregular}, we expect the generated output to be an isosceles trapezoid
``\begin{tikzpicture}[line width=1.5pt, color=pink]
\draw (0,0) -- (0.5,0) -- (1.5/4,1/4) -- (0.5/4,1/4) -- cycle;
\end{tikzpicture}"
rather than a meaningless pattern.

\begin{figure}[htp]
\centering
\includegraphics[width=0.99\textwidth]{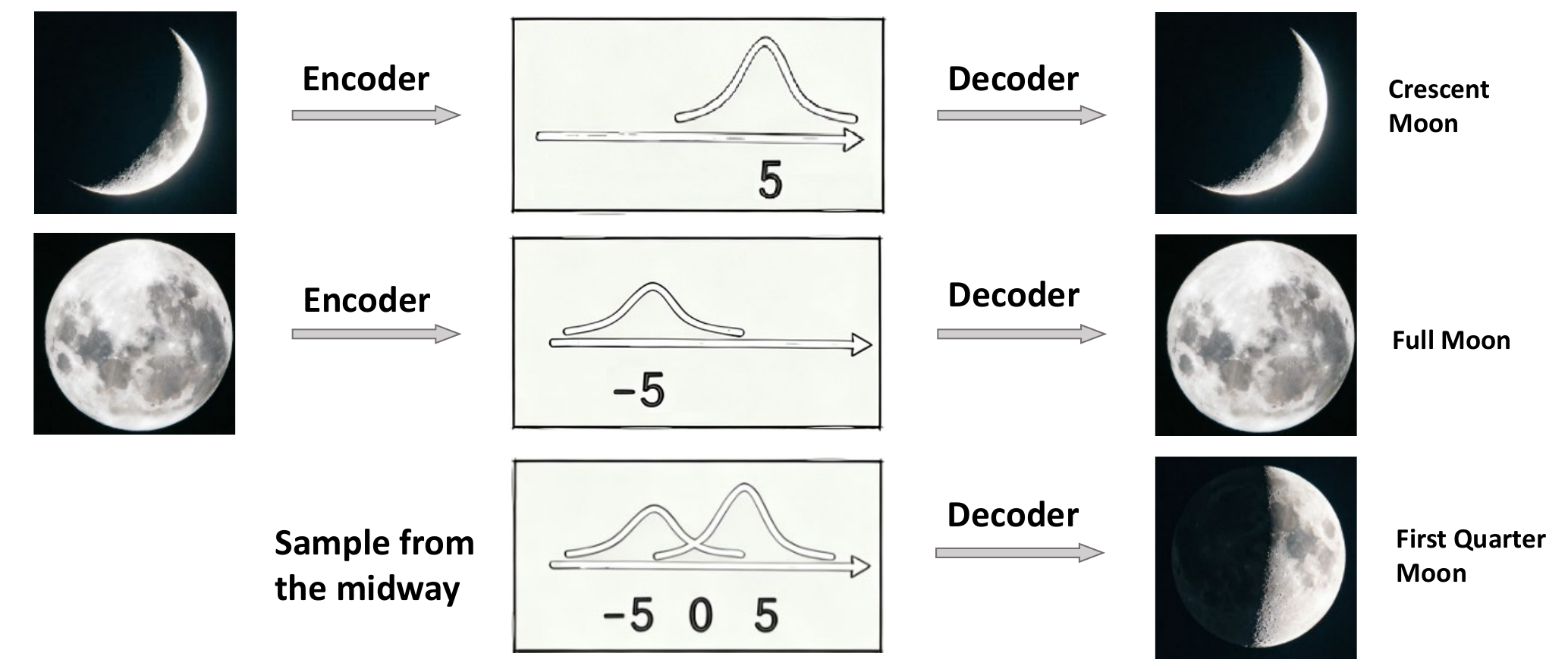}
\caption{Generation using a continuous and smooth latent space.}
\label{fig:vae_latent_regular}
\end{figure}

To provide another illustration, a VAE maps an input $\bx$ to a probability distribution over the latent space, typically a Gaussian distribution parameterized  by a mean ($\bmu$) and variance ($\bsigma^2$).
Figure~\ref{fig:vae_latent_regular} shows that the gray boxes do not represent single points or coordinates. Instead, they represent bell curves (Gaussian distributions).
The ``Crescent Moon" is encoded as a distribution centered around the latent value $5$; 
the ``Full Moon" is encoded as a distribution centered around the latent value $-5$.
The most significant capability demonstrated here is \textit{latent interpolation}. 
Because the latent space is continuous and probabilistic, we can sample a point $\bz$ from a region between two learned concepts to generate  novel, coherent images.
Consider the third row: no input image is provided on the left. Instead, the model samples a latent vector from the region labeled $0$, midway between $5$ and $-5$. The corresponding output is a ``First Quarter Moon".

We observe that the continuity and completeness achieved through regularization tend to create a smooth ``gradient" over the information encoded in the latent space. For instance, a latent point lying midway between the means of two distributions learned from distinct training samples should decode to content that is semantically intermediate between the two original inputs, as it could plausibly be sampled by the decoder in either context.

The core objective of a VAE is therefore to learn a continuous latent manifold. By constraining the encoder to output distributions, the model ensures that similar inputs map to overlapping or nearby distributions in latent space. This avoids ``holes" in the latent space where the decoder would produce undefined or nonsensical outputs.

\begin{figure}[h!]
\centering  
\vspace{-0.35cm} 
\subfigtopskip=2pt 
\subfigbottomskip=2pt 
\subfigcapskip=-5pt 
\subfigure[VAE.]{\label{fig:lvm_VAE}
\includegraphics[width=0.431\linewidth]{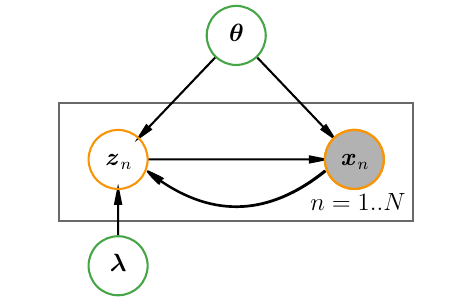}}
\subfigure[Parameter flow in VAE.]{\label{fig:lvm_VAE_flow}
\includegraphics[width=0.431\linewidth]{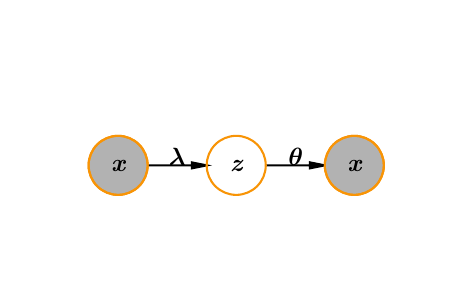}}
\caption{Graphical representation for VAE. Use the variational distribution $q_{\blambda}(\bz\mid \bx)$ to approximate the intractable posterior $p_{\btheta}(\bz\mid \bx)$.}
\label{fig:lvm_VAE_and_flow}
\end{figure}

\subsection{VAE via Generative Intuition}
To introduce the VAE framework \citep{kingma2013auto, rezende2014stochastic, kingma2019introduction}, we first adopt a perspective grounded in generative intuition, as outlined earlier in this section, rather than relying on background from the EM algorithm or variational inference presented in Section~\ref{section:vbi_root}.
Recall that the reconstruction loss in \eqref{equation:ae_mse_first} is insufficient for training an AE to serve as a generative model in latent space.
This is because the latent distribution $p_{\text{latent}}(\bz)$, defined by $\bz = \mathcalE_\blambda(\bx)$ with $\bx \sim p_{\text{data}}$ ($\bx\in\real^D$) and $ \mathcalE_\blambda(\cdot)$ denoting the encoder parameterized by $\blambda$,  is not explicitly constrained. 
A generative model for $p_{\text{data}}(\bx)$ can then be constructed by passing samples from a latent generative model through the decoder $\mathcalD_\btheta$, parameterized by $\btheta$.
This setup is illustrated in Figure~\ref{fig:lvm_VAE_and_flow}.

A critical limitation of standard AEs, as currently formulated, is that we exert little control over $p_{\text{latent}}(\bz)$. Consequently, there is no guarantee that $p_{\text{latent}}(\bz)$ will be well-behaved---for instance, smooth, simple, or Gaussian-like---making it unsuitable for generative modeling. Although encoding compresses the input data, it may also warp the data distribution $p_{\text{data}}$ into a highly complex, difficult-to-model latent distribution $p_{\text{latent}}$.
Thus, by ensuring the latent distribution $p_{\text{latent}}$ remains well-behaved and easy to model, we can use $\bz\sim p_{\text{latent}}$ and $\mathcalD_\btheta(\bz)$ to define a valid generative model.

\index{Maximum likelihood estimation}
\index{KL divergence}
\paragrapharrow{Probabilistic formulation and reconstruction error.}
A VAE extends the deterministic standard AE by relaxing the assumption that the encoder and decoder are deterministic functions. 
Specifically, we define a probabilistic encoder $q_\blambda(\bz\mid\bx)$ with parameters $\blambda$ and a probabilistic decoder $p_\btheta(\bx\mid\bz)$ with parameters $\btheta$. A common choice is to use Gaussian distributions:
\begin{equation}\label{equation:vae_intui_gaussform}
q_\blambda(\bz\mid\bx) = \normal(\bz\mid \bmu_\blambda(\bx), \diag(\bsigma_\blambda^2(\bx))), \quad p_\btheta(\bx\mid\bz) = \normal(\bx\mid \bmu_\btheta(\bz), \sigma_\btheta^2(\bz) \bI_D),
\end{equation}
where $\bmu_\blambda(\bx) \in \real^K$, $\bsigma_\blambda^2(\bx) \in \real^K_{+}$, $\bmu_\btheta(\bz) \in \real^D$, and $\sigma_\btheta^2(\bz) \in \real_{+}$ are neural network outputs, and ``$\diag(\cdot)$" denotes  a diagonal matrix. 
Encoding and decoding now involve sampling:
\begin{align*}
\bz &\sim q_\blambda(\cdot\mid\bx); & \text{(encode)} \\
\bx &\sim p_\btheta(\cdot\mid\bz). & \text{(decode)}
\end{align*}
Note that setting $\bsigma_\blambda(\cdot) = \bzero$ and $\sigma_\btheta(\cdot) = \bzero$ recovers a deterministic AE.

A natural objective function arises from maximum likelihood estimation, as shown in \eqref{equation:equiv_mle_kl}:
\begin{equation}
\mathcalJ_{\text{VAE-Recon}}(\blambda, \btheta) = -\Exp_{\bx \sim p_{\text{data}}(\bx), \bz \sim q_\blambda(\bz\mid\bx)} [\ln p_\btheta(\bx\mid\bz)].
~\footnote{Note again that, for notational convenience, we adopt the notation 
$\Exp_{\bx \sim p_{\text{data}}(\bx), \bz \sim q_\blambda(\bz\mid\bx)} [\ln p_\btheta(\bx\mid\bz)]$ 
or simply $\Exp_{p_{\text{data}}(\bx), q_\blambda(\bz\mid\bx)} [\ln p_\btheta(\bx\mid\bz)]$ 
instead of $\Exp_{\rvx \sim p_{\text{data}}(\bx), \rvz \sim q_\blambda(\bz\mid\bx)} [\ln p_\btheta(\rvx\mid\rvz)]$ for expectations. Here, $\rvx$ denotes a random vector, and $\bx$ denotes a concrete vector value. 
This convention is standard in machine learning.
Similarly, we write $\bx \sim p(\bx)$ instead of $\rvx \sim p(\bx)$ to indicate that $\bx$ is distributed according to $p(\bx)$.}
\end{equation}
Two key differences distinguish this from the standard AE loss:
the encoder now produces a distribution from which we sample $\bz \sim q_\blambda(\bz\mid\bx)$, and the loss evaluates the negative log-likelihood of the original input under the probabilistic decoder. Intuitively, the loss measures how probable the original data point $\bx$ is under the full stochastic encoding-decoding process, averaging over all possible encodings and decodings.
For the Gaussian formulation above, this reconstruction loss simplifies to:
\begin{equation}
\mathcalJ_{\text{VAE-Recon}}(\blambda, \btheta) = \Exp_{\bx \sim p_{\text{data}}(\bx), \bz \sim q_{\blambda}(\bz\mid\bx)} 
\left[ \frac{\normtwo{\bx - \bmu_{\btheta}(\bz)}^2}{2\sigma_{\btheta}^2(\bz)}  + \frac{D}{2} \ln \sigma_{\btheta}^2(\bz) \right] 
+ \text{const},
\end{equation}
where we use the probability density of the multivariate Gaussian distribution (see Definition~\ref{definition:multivariate_gaussian}). 
The VAE reconstruction loss is therefore closely related to the standard AE loss in \eqref{equation:ae_mse_first}; the primary distinction is that we now average over all possible encodings $\bz \sim q_{\blambda}(\cdot\mid\bx)$. 
The second term, which depends on the decoder variance, balances reconstruction accuracy and predictive uncertainty.
Many practical implementations fix $\sigma_{\blambda}(\bx)$ and $\sigma_{\btheta}(\bz)$ to learned scalar constants (independent of $\bx$ and $\bz$, respectively). 
This stabilizes training and avoids numerical instability when learning variance parameters. Under this setting, the VAE reconstruction loss reduces to a stochastic variant of the standard AE loss, up to constant terms:
\begin{equation}
\mathcalJ_{\text{VAE-Recon}}(\blambda, \btheta) = \Exp_{\bx \sim p_{\text{data}}(\bx), \bz \sim q_{\blambda}(\bz\mid\bx)} 
\left[ \frac{1}{2\sigma_{\btheta}^2} \normtwo{\bx - \bmu_{\btheta}(\bz)}^2 \right] + \text{const}.
\end{equation}

\index{Jensen--Shannon (JS) divergence}
\paragrapharrow{Regularization.}
Let us now revisit our goal: we aim to encode the data distribution $p_{\text{data}}(\bx)$ such that its image in the latent space forms a well-behaved or easy-to-learn distribution $p_{\text{latent}}(\cdot)$. 
The fact that VAEs encode inputs as distributions rather than deterministic points is not sufficient to guarantee the continuity and completeness properties described at the start of the section. Without a properly defined regularization term, the model may learn to minimize reconstruction error by effectively ``ignoring" the probabilistic nature of the encoder and behaving nearly like a classical AE, leading to overfitting (see Figure~\ref{fig:vae_latent_irregular}). To achieve this, the encoder could either produce distributions with extremely small variances (approaching point masses) or distributions with widely separated means that lie far apart in the latent space. In both scenarios, the probabilistic structure is misused, negating the intended benefits, and continuity and/or completeness are not achieved.

To avoid these behaviors, we must regularize both the mean and the covariance matrix of the distributions output by the encoder. In practice, this regularization is implemented by encouraging the encoded distributions to remain close to a standard Gaussian distribution (zero-mean and unit-variance). This constraint encourages covariance matrices near the identity, preventing degenerate point-like distributions, and means near $\bzero$, preventing encoded representations from becoming overly scattered in the latent space.

Toward this end, we introduce a prior distribution $p_{\text{prior}}(\bz)$ over latent variables  $\bz$. 
As noted  above, we set $p_{\text{prior}} = \normal(\bzero, \bI_K)$,  an isotropic Gaussian with $K<D$. 
This choice of prior represents the ``ideal" form for the latent distribution: a normal distribution is simple to model and supports our goal of obtaining a trainable latent space. The core idea is therefore to regularize the encoder so that the induced latent distribution remains as close as possible to $p_{\text{prior}}$, which we achieve using the auxiliary loss
\begin{equation}
\mathcalJ_{\text{VAE-Prior}}(\blambda) = \Exp_{\bx \sim p_{\text{data}}(\bx)} \left[ \KL(q_{\blambda}(\cdot\mid\bx) \parallel p_{\text{prior}}) \right],
\end{equation}
where $\KL[\cdot\parallel \cdot]$ denotes the KL divergence (see \eqref{equation:kl_def_vae}). 
The loss $\mathcalJ_{\text{VAE-Prior}}$ has an intuitive interpretation: we want the encoding distribution for any data point $\bx$ to resemble a Gaussian. Applying this constraint across all $\bx$ naturally encourages the aggregate latent distribution to also resemble a Gaussian.
With this regularization term, we prevent the model from mapping data to widely separated regions in latent space and instead encourage the encoded distributions to overlap, thereby satisfying the continuity and completeness conditions outlined earlier in this section. As with any regularization, this improvement in latent space structure comes at the cost of slightly increased reconstruction error on the training data.

As an alternative, the \textit{Jensen--Shannon (JS) divergence} may be used in place of the KL divergence within this framework:
\begin{equation}
\JS[P\parallel Q] = 
\frac{1}{2} \KL[P \parallel \tfrac{1}{2}(P+Q)]
+\frac{1}{2} \KL[Q \parallel \tfrac{1}{2}(P+Q)].
\end{equation}

\paragrapharrow{Final loss.}
To construct the full loss function for a VAE, we combine the reconstruction loss and the prior regularization loss using a weighting parameter $\beta \geq 0$. This yields the VAE training objective:
\begin{mybox}
\begin{subequations}\label{equation:vae_loss_generaintuition}
\begin{align}
\mathcalJ_{\text{VAE}}&(\blambda, \btheta) = \mathcalJ_{\text{VAE-Recon}}(\blambda, \btheta) + \beta\cdot \mathcalJ_{\text{VAE-Prior}}(\blambda) \\
&= -\Exp_{p_{\text{data}}(\bx),  q_{\blambda}(\bz\mid\bx)} \left[ \ln p_{\btheta}(\bx \mid \bz) \right] 
+ \beta\cdot \Exp_{p_{\text{data}}(\bx)} \left[ \KL(q_{\blambda}(\cdot\mid\bx) \parallel p_{\text{prior}}) \right]
\label{equation:vae_int_EXP}\\
&\simeq 
\underbrace{-\sum_{n=1}^{N}\Exp_{q_{\blambda}(\bz_n\mid\bx_n)} \left[ \ln p_{\btheta}(\bx_n \mid \bz_n) \right] }_{\text{Data fit}}
+ \beta \cdot
\underbrace{\sum_{ n=1}^{N}\left[ \KL(q_{\blambda}(\bz_n\mid\bx_n) \parallel p_{\text{prior}}(\bz_n)) \right]}_{\text{Regularization/Prior enforcement loss}},
\label{equation:vae_int_X}
\end{align}
\end{subequations}
\end{mybox}
where the first term encourages accurate reconstruction of data from latent samples (data fitting), and the second term encourages the latent distribution to remain close to a Gaussian (regularization).
The approximation in \eqref{equation:vae_int_X} corresponds to the loss evaluated over a dataset $\mathcalX=\{\bx_1, \bx_2, \ldots,\bx_N\}$, i.e., the \textit{complete-data loss}.
The parameter $\beta$ controls the relative strength between the two components.

\subsection{VAE via Joint-Space KL Divergence}
In this subsection, we derive the full VAE loss function via a joint-space perspective. First, note that both the encoder and decoder define valid joint distributions over the input data $\bx$ and  latent  variables  $\bz$:
\begin{align}
q_\blambda(\bx,\bz) &= p_{\text{data}}(\bx)\cdot q_\blambda(\bz\mid\bx); &&\text{(encoder joint distribution)} \label{equation:vae_joint_prob1}\\
p_\btheta(\bx,\bz) &= p_{\text{prior}}(\bz)\cdot p_\btheta(\bx\mid\bz). &&\text{(decoder joint distribution)} \label{equation:vae_joint_prob2}
\end{align}
VAE training can therefore be interpreted as optimizing the parameters $\blambda$ and $\btheta$ to align the encoder and decoder joint distributions. This alignment can be formally quantified using the KL divergence between the two joint data-latent distributions:
\begin{subequations}
\begin{align}
\KL\big[q_\blambda(\bx,\bz) \parallel &p_\btheta(\bx,\bz)\big]
= \KL\big[p_{\text{data}}(\bx)q_\blambda(\bz\mid \bx) \parallel p_{\text{prior}}(\bz)p_\btheta(\bx\mid \bz)\big] \\
&= \Exp_{\diamondsuit}\left[\ln\left(\frac{p_{\text{data}}(\bx)q_\blambda(\bz\mid \bx)}{p_{\text{prior}}(\bz)p_\btheta(\bx\mid \bz)}\right)\right] \\
&= \Exp_{\diamondsuit}\big[\ln p_{\text{data}}(\bx)\big]
+ \Exp_{\diamondsuit}\left[\ln\left(\frac{q_\blambda(\bz\mid \bx)}{p_{\text{prior}}(\bz)}\right)\right]
- \Exp_{\diamondsuit}\big[\ln p_\btheta(\bx\mid \bz)\big],
 \label{equation:vae_kl_jointpdf}
\end{align}
\end{subequations}
where the expectation is defined over$\diamondsuit = \bx\sim p_{\text{data}}(\bx), \bz\sim q_\blambda(\bz\mid\bx)$.

We now analyze each term in the decomposed KL expression individually. First,
\begin{equation}
\Exp_{\diamondsuit}\big[\ln p_{\text{data}}(\bx)\big]
= \Exp_{p_{\text{data}}(\bx)}\big[\ln p_{\text{data}}(\bx)\big] = C, 
\end{equation}
which yields a constant value $C$ independent of the encoder and decoder parameters $\blambda$ and $\btheta$. 
Second,
\begin{equation}
\Exp_{\diamondsuit}\left[\ln\left(\frac{q_\blambda(\bz\mid \bx)}{p_{\text{prior}}(\bz)}\right)\right]
= \Exp_{p_{\text{data}}(\bx)}\big[\KL\big(q_\blambda(\bz\mid \bx)\parallel p_{\text{prior}}(\bz)\big)\big].
\end{equation}
This term penalizes deviations of the encoder posterior $q_\blambda(\bz\mid \bx)$  from the predefined latent prior $p_{\text{prior}}(\bz)$, thereby enforcing latent-space regularization.
Third,
\begin{equation}
-\Exp_{p_{\text{data}}(\bx), q_\blambda(\bz\mid\bx)}\big[\ln p_\btheta(\bx\mid \bz)\big] 
\end{equation}
represents the average negative log-likelihood over the data and latent samples, which constitutes the standard reconstruction loss component. 
By discarding the constant term and combining the regularization and reconstruction objectives, we arrive at a key interpretation: the full VAE loss is equivalent to the KL divergence measured over the joint data-latent space:
\begin{mybox}
\begin{subequations}\label{equation:vae_jpdf_loss_ALL}
\begin{align}
\mathcalJ_{\text{VAE}}&(\blambda,\btheta)+ \text{const}
= \KL\big[q_\blambda(\bx,\bz)\parallel p_\btheta(\bx,\bz)\big]   \label{equation:vae_jpdf_loss1}\\
&= - \Exp_{p_{\text{data}}(\bx), q_\blambda(\bz\mid\bx)}\big[\ln p_\btheta(\bx\mid\bz)\big]
+\Exp_{p_{\text{data}}(\bx)}\big[\KL\big(q_\blambda(\bz\mid\bx)\parallel p_{\text{prior}}(\bz)\big)\big]
\label{equation:vae_jpdf_loss2} \\
&\simeq  
\underbrace{- \sum_{n=1}^{N} \Exp_{ q_\blambda(\bz_n\mid\bx_n)}\big[\ln p_\btheta(\bx_n\mid\bz_n)\big]}_{\text{Data fit}}
+\underbrace{\sum_{n=1}^{N}\KL\big[q_\blambda(\bz_n\mid\bx_n)\parallel p_{\text{prior}}(\bz_n)\big]}_{\text{Regularization/Prior enforcement loss}} .
\label{equation:vae_jpdf_loss3} 
\end{align}
\end{subequations}
\end{mybox}
This derivation reveals that the VAE training objective can be fundamentally interpreted as a KL divergence minimization problem defined over the joint space of input data and latent variables.

\paragrapharrow{Interpretation of the VAE loss in \eqref{equation:vae_jpdf_loss_ALL}.}
We now interpret the VAE as a generative model. 
To generate new samples, we first sample latent variables from the prior distribution, $\bz \sim p_{\text{prior}} $ (e.g., the standard Gaussian  $p_{\text{prior}} = \normal(\bzero,\bI_K)$), 
and then sample data points from the decoder distribution conditioned on these latent variables: $\bx \sim p_\btheta(\cdot\mid\bz)$. 
This sampling procedure yields the marginal generative distribution:
$$
p_\btheta(\bx) = \int p_\btheta(\bx\mid \bz)\,p_{\text{prior}}(\bz)\diff \bz.
$$
We next prove that training a VAE effectively optimizes this learned generative distribution $p_\btheta(\bx)$ to approximate the true data distribution $p_{\text{data}}(\bx)$.
To establish this result, we first introduce a useful KL divergence decomposition lemma.
\begin{lemma}[KL divergence chain decomposition]\label{lemma:kl_chain}
Let $p(\bx,\bz),q(\bx,\bz)$ denote joint distributions over variables $\bx\in\real^{D_1}, \bz\in\real^{D_2}$. 
The KL divergence between the two joint distributions decomposes as follows:
$$
\KL\big[p(\bx,\bz)\parallel q(\bx,\bz)\big]
= \KL\big[p(\bx)\parallel q(\bx)\big]
+ \Exp_{p(\bx)}\big[\KL\big(p(\bz\mid \bx)\parallel q(\bz\mid \bx)\big)\big].
$$
Since the second term is strictly nonnegative, this decomposition directly implies the data-processing inequality:
\begin{equation}\label{equation:kl_chain_ineq}
\KL\big[p(\bx)\parallel q(\bx)\big]
\leq \KL\big[p(\bx,\bz)\parallel q(\bx,\bz)\big]. 
\end{equation}
\end{lemma}
\begin{proof}[of Lemma~\ref{lemma:kl_chain}]
Write out the KL divergence, we have 
\begin{align*}
\KL\big[p(\bx,\bz)\parallel q(\bx,\bz)\big]
&= \Exp_{p(\bx,\bz)}\left[\ln\frac{p(\bx,\bz)}{q(\bx,\bz)}\right] 
= \Exp_{p(\bx,\bz)}\left[\ln\left(\frac{p(\bz\mid \bx)}{q(\bz\mid \bx)}\frac{p(\bx)}{q(\bx)}\right)\right] \\
&= \Exp_{p(\bx)}\left[\ln\frac{p(\bx)}{q(\bx)}\right] 
+\Exp_{p(\bx,\bz)}\left[\ln\frac{p(\bz\mid \bx)}{q(\bz\mid \bx)}\right] \\
&= \KL\big[p(\bx)\parallel q(\bx)\big]
+ \Exp_{p(\bx)}\big[\KL\big(p(\bz\mid \bx)\parallel q(\bz\mid \bx)\big)\big].
\end{align*}
This completes the proof.
\end{proof}

Applying Lemma~\ref{lemma:kl_chain} to the VAE joint distributions yields the following key inequality:
\begin{equation}\label{equation:vaejoint_dl_ineq}
\mathcalJ_{\text{VAE}}(\blambda,\btheta)
= \KL\big[q_\blambda(\bx,\bz)\parallel p_\btheta(\bx,\bz)\big] + \text{const}
\geq \KL\big[p_{\text{data}}(\bx)\parallel p_\btheta(\bx)\big] + \text{const} ,
\end{equation}
where we use the property that the data marginal distribution of the encoder joint distribution $q_\blambda(\bx,\bz)$ is exactly the true data distribution $p_{\text{data}}(\bx)$; see \eqref{equation:vae_joint_prob1}.
This result reveals that the VAE loss function minimizes an upper bound on the KL divergence between the true data distribution $p_{\text{data}}(\bx)$ and the VAE's learned generative distribution $p_\btheta(\bx)$. This core property validates the VAE as a valid generative model.

By symmetry, we can derive a second complementary inequality from the same decomposition:
\begin{equation}\label{equation:vaejoint_dl_ineq2}
\mathcalJ_{\text{VAE}}(\blambda,\btheta)
= \KL\big[q_\blambda(\bx,\bz)\parallel p_\btheta(\bx,\bz)\big] + \text{const}
\geq \KL\big[q_\blambda(\bz)\parallel p_{\text{prior}}(\bz)\big] + \text{const} .
\end{equation}
This indicates that the VAE objective in \eqref{equation:vae_jpdf_loss_ALL} also minimizes an upper bound on the KL divergence between the aggregated latent distribution and the predefined latent prior distribution.

\subsection{VAE via Maximizing ELBO}\label{section:vae_pca}
We now  introduce the VAE framework by maximizing the evidence lower-bound (ELBO).
We have already seen that the marginal likelihood function for a latent variable model (see Section~\ref{section:lvm}) is given by
\begin{equation}\label{equation:vae_likelihood}
p_{\btheta}(\bx ) = \int p_{\btheta}(\bx\mid\bz) p(\bz) \diff\bz.
\end{equation}
When the conditional likelihood $p_{\btheta}(\bx\mid\bz)$ is parameterized by a deep neural network with weights $\btheta$,
this high-dimensional integral over latent variables $\bz$ becomes analytically intractable and cannot be computed exactly. 
The VAE resolves this issue by optimizing a tractable lower bound on the log-likelihood throughout training \citep{kingma2013auto, rezende2014stochastic, kingma2019introduction}.
This third perspective on the VAE framework is built upon two core principles:
(i) Optimization of the ELBO to approximate the intractable exact log-likelihood, establishing a fundamental connection to the EM algorithm;
(ii) \textit{Amortized inference:} rather than fitting a separate posterior approximation for each individual data point, we train a shared encoder network that maps input observations $\bx$ to approximate latent posterior distributions over $\bz$: $q_{\blambda}(\bz_n\mid\bx_n)$ for any data point $\bx_n$.

We consider a generative model in which the observation likelihood $p_{\btheta}(\bx\mid\bz)$  for inputs $\bx\in\real^D$ is defined by a deep neural network $\bmu_{\btheta}(\bz)$.
For instance, this network typically outputs the mean parameters of a Gaussian observation distribution.
We further impose a standard isotropic Gaussian prior over the $K$-dimensional latent variables $\bz\in\real^K$:
\begin{equation}\label{equation:vae_latent_prior}
p(\bz) =p_{\text{prior}}(\bz) = \normal(\bz\mid\bzero,\bI_K).
\end{equation}
To derive the VAE training objective, we revisit the standard ELBO decomposition from \eqref{equation:elbo_vfe_neg}. 
For any \textbf{arbitrary}  latent distribution $q(\bz)$, the exact log-likelihood decomposes as
\begin{equation}\label{equation:elbo_kl_vae}
\ln p_\btheta(\bx) = \mathcalF(\btheta) + \KL[q(\bz) \parallel p_\btheta(\bz\mid\bx)],
\end{equation}
where  $\KL[P \parallel Q] \geq 0$ denotes the {KL divergence} between distributions $P$ and $Q$, with equality if and only if $P=Q$ (see \eqref{equation:kl_def_vae}). 
Then term $\mathcalF(\btheta)$ corresponds to the ELBO (see also \eqref{equation:elbo_ineq}), formally defined as:
\begin{equation}\label{equation:vae_elbo}
\mathcalF(\btheta) 
\triangleq  \int q(\bz) \ln \left\{ \frac{p_{\btheta}(\bx\mid\bz) p_{\text{prior}}(\bz)}{q(\bz)} \right\} \diff\bz,
\end{equation}

Since the KL divergence is strictly nonnegative, we obtain the lower-bound relation:
\begin{equation}\label{equation:vae_lower_bound}
	\ln p_\btheta(\bx) \geq \mathcalF.
\end{equation}
This confirms that $\mathcalF$ constitutes a valid lower bound on the intractable log-likelihood $\ln p_\btheta(\bx)$. 
Unlike the exact marginal log-likelihood, the ELBO can be efficiently approximated via Monte Carlo sampling, a paradigm referred to as \textit{stochastic variational inference}.
Consequently, the ELBO serves as a practical, tractable surrogate for maximum-likelihood-based VAE training.

\paragrapharrow{Complete-data log-likelihood.}
Now consider an observed dataset $\mathcalX = \{\bx_1, \bx_2, \ldots,\bx_N\}$, assumed to contain independent and identically distributed samples from the model. The expected log-likelihood over the data distribution decomposes as
\begin{equation}\label{equation:vae_log_likelihood_sum}
\Exp_{p_{\text{data}}(\bx)} [\ln p_\btheta(\bx)]
\simeq 
\ln p_\btheta(\mathcalX) = \sum_{n=1}^N \mathcalF_n + \sum_{n=1}^N \KL[q_{\bz_n}(\bz_n\mid \blambda_n) \parallel p_\btheta(\bz_n\mid\bx_n)],
\end{equation}
where the per-data-point ELBO is defined as 
\begin{equation}\label{equation:vae_elbo_per_data}
\mathcalF_n = \mathcalF_n(\blambda_n, \btheta) 
\triangleq  \int q_{\bz_n}(\bz_n\mid \blambda_n) \ln \left\{ \frac{p_{\btheta}(\bx_n\mid\bz_n) p_{\text{prior}}(\bz_n)}{q_{\bz_n}(\bz_n\mid \blambda_n)} \right\} \diff\bz_n.
\end{equation}
As in models such as {mixture models} (see Problems~\ref{prob:mix_of_gauss}--\ref{prob:mix_of_bern}), we introduce a separate latent variable $\bz_n$ for each observation $\bx_n$.
Consequently, each $\bz_n$ has its own approximate posterior  $q_{\bz_n}(\bz_n\mid \blambda_n)$, which in principle can be optimized independently.

\paragrapharrow{Intractability and approximation.}
Because \eqref{equation:vae_log_likelihood_sum} holds for any choice of $q_{\bz_n}(\bz_n\mid \blambda_n)$, we may choose the family of distributions that maximizes $\mathcalF_n$ (or equivalently, minimizes the KL divergence to the true posterior, $\KL[q_{\bz_n}(\bz_n\mid \blambda_n) \parallel p_{\btheta}(\bz_n\mid\bx_n)]$). 
In simpler models such as Gaussian mixtures or probabilistic/Bayesian PCA, the exact posterior $p_{\btheta}(\bz_n\mid\bx_n)$ can be computed analytically within the E-step of the EM algorithm \citep{lu2023bayesian}. 
Setting $q_{\bz_n}$ equal to this true posterior results in zero KL divergence, so the ELBO matches the exact log-likelihood (see also the unconstrained EM algorithm in Section~\ref{section:em_uncons}).
By Bayes' theorem~\eqref{equation:bayes_base}, the exact posterior is given by
\begin{equation}\label{equation:vae_bayes_posterior}
p_\btheta(\bz_n\mid\bx_n) = \frac{p_{\btheta}(\bx_n\mid\bz_n) p_{\text{prior}}(\bz_n)}{p_\btheta(\bx_n)}.
\end{equation}
While the numerator is easy to evaluate---because $p_{\text{prior}}(\bz_n)$ is a standard Gaussian (Equation~\eqref{equation:vae_latent_prior}) and $p_{\btheta}(\bx_n\mid\bz_n)$ is given by the decoder network---the denominator is exactly the intractable marginal likelihood $p_\btheta(\bx_n)$. 
We are therefore forced to approximate the posterior.

\subsubsection*{VAE with Amortized Variational Inference}
In principle, we could introduce a separate set of parameters $\blambda_n$ for each approximate posterior $q_{\bz_n}(\cdot\mid \blambda_n)$ and optimize these parameters individually. 
However, this approach is computationally prohibitive for large datasets, and the approximate posteriors would need to be re-estimated after every update to $\btheta$. Instead, the VAE adopts a more scalable strategy: it introduces a second neural network---the encoder---that amortizes the cost of inference by sharing parameters across all data points. This encoder maps each input $\bx_n$ to the parameters of its approximate posterior $q_{\bz_n}$, enabling efficient and joint optimization of both generative and inference models.

\begin{figure}[h!]
\centering  
\vspace{-0.35cm} 
\subfigtopskip=2pt 
\subfigbottomskip=2pt 
\subfigcapskip=-5pt 
\subfigure[(Standard) VI.]{\label{fig:lvm_VI}
	\includegraphics[width=0.431\linewidth]{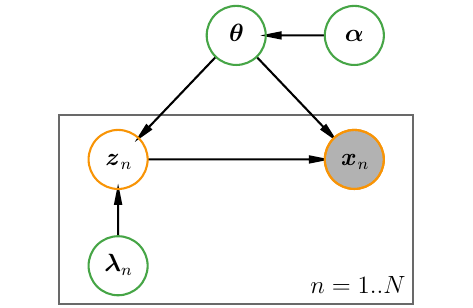}}
\subfigure[Amortized VI.]{\label{fig:lvm_VI_amortize}
	\includegraphics[width=0.431\linewidth]{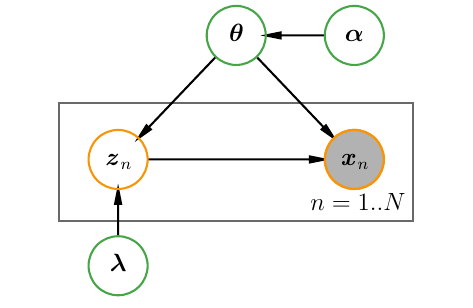}}
\caption{Graphical model representation of latent variable models under variational inference. Green circles denote prior variables, orange circles represent observed and latent variables, and plates  indicate replicated structures. }
\label{fig:lvm_and_hyp_vb}
\end{figure}

In the VAE, rather than computing a separate posterior approximation {$q_{\textcolor{mylightbluetext}{\bz_n}}(\bz_n\mid\blambda_n)$} for each data point $\bx_n$ separately, we train a single neural network---referred to as the \textit{encoder network} or \textit{recognition model}---to approximate all such posteriors simultaneously. 
This framework  is known as \textit{amortized variational inference} (or simply \textit{amortized inference}).~\footnote{see, for example, \citet{lu2023bayesian} and references therein for further details.} 
The encoder defines a conditional distribution $q_{\textcolor{mylightbluetext}{\blambda}}(\bz\mid\bx)$, parameterized by $\blambda$,  that maps any input $\bx$ to a distribution over the latent space; see Figure~\ref{fig:lvm_and_hyp_vb} for a comparison between standard VI and amorized VI.
The objective function---the ELBO---now depends on \textbf{both} the generative parameters $\btheta$ and the inference parameters $\blambda$; see \eqref{equation:vae_elbo_vae}. 
We maximize this bound jointly over both parameter sets using gradient-based optimization methods \citep{goodfellow2016deep, lu2022gradient}.

The VAE uses a shared, data-dependent approximate  posterior: $q_{\blambda}(\bz\mid \bx )$ with variational parameter $\blambda$:
$
q_{\blambda}(\bz_n\mid \bx_n ) = \normal\big(\bz_n \mid \bmu_{\blambda}(\bx_n), \diag(\bsigma_{\blambda}^2(\bx_n))\big),
$
where $\bmu_{\blambda}$ and $\bsigma_{\blambda}^2$ are parameterized by a neural network.
This yields an algorithm that closely resembles training a standard AE, but with a stochastic encoding step that injects learnable noise (see Figure~\ref{fig:lvm_VAE_flow}), hence the name variational autoencoder.
The VAE objective is to maximize the following ELBO jointly over $\btheta$ and $\blambda$:
\begin{mybox}
\begin{subequations}\label{equation:vae_elbo_obj}
\begin{align}
&\argmin_{\blambda,\btheta} \mathcalJ_{\text{VAE}}(\blambda, \btheta)=\argmin_{\btheta, \blambda} \mathcalF(\blambda, \btheta) 
\simeq \argmin_{\btheta, \blambda}\sum_{n=1}^{N} \mathcalF_n\\
&=\argmax_{\blambda,\btheta} 
\sum_{n=1}^{N} \Exp_{q_{\blambda}(\bz_n\mid\bx_n)} 
\left[ \ln \frac{p_{\btheta}(\bz_n, \bx_n  )}{q_{\blambda}(\bz_n\mid\bx_n)} \right]\\
&= \argmax_{\blambda,\btheta} \sum_{n=1}^{N}\int q_{\blambda}(\bz_n\mid\bx_n) \ln \left\{ \frac{p_{\btheta}(\bx_n\mid\bz_n) p_{\text{prior}}(\bz_n)}{q_{\blambda}(\bz_n\mid\bx_n)} \right\} \diff\bz_n  \\
&=
\argmax_{\blambda,\btheta} 
\underbrace{\sum_{n=1}^{N} \Exp_{q_{\blambda}(\bz_n\mid\bx_n)} 
	\left[ \ln {p_{\btheta}(\bx_n \mid\bz_n  )} \right]}_{\text{Data fit}}
\underbrace{-\sum_{n=1}^{N}\KL\left[q_{\blambda}(\bz_n\mid\bx_n) \parallel p(\bz_n  ) \right]}_{\text{Regularization/Prior enforcement loss}}. \label{equation:vae_elbo_vae}
\end{align}
\end{subequations}
\end{mybox}
Once again, the symbol `$\simeq$' denotes the approximation of the population-level expected loss by the empirical dataset loss.
The first term can be
interpreted as the (negative) reconstruction error, corresponding to the fidelity of reconstructing the observed data $\bx_n$ from the latent representations; 
this ensures the model learns meaningful latent variables from which the original data can be reliably regenerated. 
The second KL term acts as a regularizer, measured by the divergence between the approximate posterior and the prior $p_{\text{prior}}(\bz_n)$. Minimizing this term encourages the encoder to learn a proper distribution rather than collapsing to a Dirac delta function. 
This tradeoff is natural in Bayesian inference and reflects the balance that must be struck between confidence in the observed data and confidence in the prior.

\begin{remark}[Connection to EM algorithms]
Consistent with the EM framework (Section~\ref{section:vbi_root}),  the core training goal remains maximum-likelihood estimation of the generative parameters $\btheta$:
$\mathop{\max}_{\btheta} p(\mathcalX\mid\btheta) \simeq \mathop{\max}_{\btheta} \sum_{n=1}^{N} \ln p(\bx_n \mid \btheta ). $
Standard variational inference approaches address this objective via a constrained EM algorithm (Section~\ref{section:em_constrained}), which iteratively maximizes the per-sample ELBO:
$$
\mathop{\max}_{\btheta,\{\blambda_n\}} 
\sum_{n=1}^{N} \Exp_{q_{\textcolor{mylightbluetext}{\bz_n}}(\bz_n\mid\textcolor{mylightbluetext}{\blambda_n})} 
\left[ \ln \frac{p_{\btheta}(\bz_n, \bx_n  )}{q_{\textcolor{mylightbluetext}{\bz_n}}(\bz_n\mid\textcolor{mylightbluetext}{\blambda_n})} \right], 
\quad \forall\, n=1,2,\ldots,N.
$$ 
In contrast, the VAE leverages amortized inference via a single shared, data-dependent posterior approximation  $q_{\blambda}(\bz\mid \bx )$ parameterized by global variational weights $\blambda$:
$ q_{\textcolor{mylightbluetext}{\blambda}}(\bz_n\mid \bx_n ) = \normal(\bz_n \mid \bmu_{\blambda}(\bx_n), \diag(\bsigma_{\blambda}^2(\bx_n))) $
for all data samples  $\bx_n\in\{\bx_1,\bx_2,\ldots,\bx_N\}$. 
The mean and variance outputs $\bmu_{\blambda}$ and $\bsigma_{\blambda}^2$ are implemented as a neural network, as detailed in the subsequent section.
\end{remark}

\begin{remark}[Connection to the joint-space KL divergence perspective]
We previously established the equivalence between the VAE loss derived from joint-space KL divergence \eqref{equation:vae_jpdf_loss3} and the ELBO-based objective presented in \eqref{equation:vae_elbo_vae}.
Rearranging the terms from the joint-space KL divergence in \eqref{equation:vae_kl_jointpdf} yields complementary theoretical interpretations, most notably the formal definition of the evidence lower-bound. For fixed input  $\bx$, we derive:  
\begin{align}
\Exp_{q_\blambda(\bz\mid\bx)}\left[\ln\left(\frac{q_\blambda(\bz\mid \bx)}{p_{\text{prior}}(\bz)p_\btheta(\bx\mid \bz)}\right)\right]
&= \Exp_{q_\blambda(\bz\mid\bx)}\left[\ln\left(\frac{q_\blambda(\bz\mid \bx)}{p_\btheta(\bz\mid \bx)}\right)-\ln p_\btheta(\bx)\right] \nonumber \\
&= \KL\big[q_\blambda(\bz\mid \bx)\parallel p_\btheta(\bz\mid \bx)\big]-\ln p_\btheta(\bx), \label{equation:vaejoint_refom}
\end{align}
where the first equality is obtained from Bayes' rule:
$ p_\btheta(\bz\mid \bx)={p_\btheta(\bx\mid \bz)p_{\text{prior}}(\bz)}/{p_\btheta(\bx)}$.
We may thus rearrange \eqref{equation:vaejoint_refom} to obtain
\begin{equation}
\Exp_{q_\blambda(\bz\mid\bx)}\left[\ln\left(\frac{p_\btheta(\bx\mid \bz)p_{\text{prior}}(\bz)}{q_\blambda(\bz\mid \bx)}\right)\right]
+ \KL\big[q_\blambda(\bz\mid \bx)\parallel p_\btheta(\bz\mid \bx)\big]
=\ln p_\btheta(\bx).
\end{equation}
This identity yields the standard ELBO inequality:
\begin{equation}
\underbrace{\Exp_{q_\blambda(\bz\mid\bx)}\left[\ln\left(\frac{p_\btheta(\bx\mid \bz)p_{\text{prior}}(\bz)}{q_\blambda(\bz\mid \bx)}\right)\right]}_{\equiv \text{ELBO} =\mathcalF(\bx;\blambda,\btheta)}
\leq \underbrace{\ln p_\btheta(\bx)}_{\text{Evidence}}. 
\end{equation} We may now rewrite $\mathcalJ_{\text{VAE}}(\blambda, \btheta)$ from \eqref{equation:vae_jpdf_loss2} in terms of the ELBO via
\begin{align}
\mathcalJ_{\text{VAE}}(\blambda, \btheta)
&= \KL\big[q_\blambda(\bx,\bz)\parallel p_\btheta(\bx,\bz)\big] + \text{const} \nonumber\\
&= \Exp_{p_{\text{data}}(\bx)}\Exp_{q_\blambda(\bz\mid\bx)}\left[\ln\left(\frac{p_{\text{data}}(\bx)q_\blambda(\bz\mid \bx)}{p_\btheta(\bx\mid \bz)p_{\text{prior}}(\bz)}\right)\right] + \text{const} \nonumber\\
&= -\Exp_{p_{\text{data}}(\bx)}\big[\mathcalF(\bx;\blambda,\btheta)\big] + \text{const}.
\end{align}
This final expression confirms that minimizing the standard VAE loss is mathematically equivalent to maximizing the expected ELBO over the empirical data distribution.
\end{remark}

\index{KL annealing}
\index{$\beta$-VAE}
\subsection{Gaussian Parameterization of VAE Encoders and Decoders}
From all perspectives, a VAE consists of two neural networks with independent parameters that are trained jointly:
an \textit{encoder network} that maps an observed data vector $\bx$ to a distribution over the latent variable $\bz$, and a \textit{decoder network}---the original generative model---that maps a latent vector $\bz$ back to the data space. 
This architecture resembles a standard neural network–based AE, but with a critical difference: rather than producing a single point estimate in the latent space, the encoder outputs a probability distribution over $\bz$.
The encoder effectively learns an approximate probabilistic inverse of the decoder, consistent with Bayes' theorem; see Figure~\ref{fig:ae_vae}.

As noted earlier, a common choice for the encoder is a Gaussian distribution with diagonal covariance, where both the mean $\bmu_{\blambda}$ and  variance $\bsigma_{\blambda}^2$ are outputs of a neural network that takes a fixed input $\bx$:
\begin{equation}\label{equation:vae_encoder_gaussian}
\begin{aligned}
q_{\blambda}(\bz\mid \bx ) 
&= \normal(\bz \mid \bmu_{\blambda}(\bx), \diag(\bsigma_{\blambda}^2(\bx)))
= \prod_{k=1}^K \normal\left(z_k \mid 
(\bmu_\blambda(\bx))_k, 
(\bsigma_\blambda^2(\bx))_k
\right),
\end{aligned}
\end{equation}
Note that the mean components can take any real value, so the corresponding output units typically use a linear activation function. In contrast,  variances must be nonnegative; therefore, the network usually outputs log-variances or applies an exponential activation (e.g., $\exp(\cdot)$) to ensure positivity.

Similarly, the decoder is often modeled as a Gaussian distribution with diagonal covariance:
\begin{equation}\label{equation:vae_dec_gauss_diag}
p_{\btheta}(\bx\mid \bz)
=\normal(\bx\mid \bmu_{\btheta}(\bz), \diag(\bsigma^2_{\btheta}(\bz))), 
\end{equation}
where $\bmu_{\btheta}$ and $\bsigma_{\btheta}^2$ are produced by neural network transformations applied to the latent vector $\bz$. 
In practice, the decoder variance $\bsigma^2_{\btheta}$ is often \textbf{fixed} (e.g., set to 1), and only the mean $\bmu_{\btheta}$ is learned.
Alternatively, a scalar value $\sigma^2_{\btheta}$ is learned such that the variance of $p_{\btheta}(\bx\mid \bz)$ is given by $\sigma^2_{\btheta}\bI_D$.
The reconstructed output $\widehatbx_n = \bmu_{\btheta}(\bz)$ is then compared against  the original input $\bx$.
For binary data (e.g., binarized MNIST), a Bernoulli likelihood is commonly used in place of a Gaussian.
To encode or decode a variable, we sample
\begin{align*}
\bz &\sim q_\blambda(\cdot\mid\bx); & \text{(encode)} \\
\bx &\sim p_\btheta(\cdot\mid\bz). & \text{(decode)}
\end{align*}
Note that when $\bsigma_\blambda(\cdot) = \bzero$ and $\bsigma_\btheta(\cdot) = \bzero$ for all inputs, we recover a standard AE.

By convention, the prior is chosen to be a standard multivariate Gaussian:
$$
p_{\text{prior}}(\bz)=\normal(\bzero, \bI_K).
$$
This choice is convenient because (i) it is straightforward to sample from, and (ii) the KL divergence between two Gaussians has a closed-form expression (see Problem~\ref{problem:kl_vae}). 
Furthermore, it encourages the encoder to produce latent representations that are smooth, continuous, and centered around the origin with unit variance, supporting  meaningful interpolation and sampling in the latent space.

In practice, many researchers scale the KL regularization term by a hyperparameter $\beta$ to control the trade-off between reconstruction fidelity and latent structure.
Too illustrate this,
we note that the KL term in the loss \eqref{equation:vae_jpdf_loss2} or ELBO \eqref{equation:vae_elbo_vae} regularizes the encoder to align its output with the prior $p_{\text{prior}}(\bz)$, ensuring that samples from the prior produce  realistic data when passed through the decoder. However, two failure modes may arise:
\begin{enumerate}[label=(\roman*)]
\item \textit{Posterior collapse.} The encoder disregards the input and outputs a distribution close to the prior, i.e., $q_{\blambda}(\bz\mid \bx)\approx p_{\text{prior}}(\bz)$. This renders the latent code uninformative. 
Symptoms include low-quality reconstructions (blurry outputs) and a KL divergence near zero.
\item \textit{Poor generative quality.} Reconstructions are accurate, but samples generated from  $p_{\text{prior}}(\bz)$ are unrealistic. 
In this case, the encoder fits the data too closely, causing $q_{\blambda}(\bz\mid \bx)$ to diverge substantially  from  $p_{\text{prior}}(\bz)$, so prior samples lie in low-density regions of the latent space used during training.
\end{enumerate}
Both issues can be alleviated by introducing a weighting coefficient $\beta>0$ on the KL term:
\begin{mybox}
\begin{equation}\label{equation:vae_beta_loss_decomposed}
\begin{aligned}
\mathcalJ_{\text{VAE}}^\beta&(\blambda, \btheta)
= \mathcalJ_{\text{VAE-Recon}}(\blambda, \btheta) + \beta \cdot\mathcalJ_{\text{VAE-Prior}}(\blambda)  \\
&=-\Exp_{p_{\text{data}}(\bx),  q_{\blambda}(\bz\mid\bx)} \left[ \ln p_{\btheta}(\bx \mid \bz) \right] 
+ \beta\cdot \Exp_{p_{\text{data}}(\bx)} \left[ \KL(q_{\blambda}(\cdot\mid\bx) \parallel p_{\text{prior}}) \right].
\end{aligned}
\end{equation}
\end{mybox}
This formulation yields the \textit{$\beta$-VAE} \citep{hoffman2016elbo, higgins2017beta} and is consistent with the loss in \eqref{equation:vae_loss_generaintuition} from a generative intuition perspective.
If reconstructions are poor, decrease $\beta$; if generated samples are poor, increase  $\beta$. 
In many implementations, $\beta$ is scheduled to start small and increase gradually during training, a procedure known as \textit{KL annealing}.

To make this loss more concrete, we derive the KL divergence for the Gaussian case (see Problems~\ref{problem:kl_gauss1}--\ref{problem:kl_gauss2}). 
Assume the encoder follows a Gaussian distribution as defined in \eqref{equation:vae_encoder_gaussian}. 
We then obtain:
$$
\begin{aligned}
\mathcalJ_{\text{VAE-Prior}}(\blambda) 
&= \Exp_{p_{\text{data}}(\bx)} \left[ \KL(q_{\blambda}(\cdot\mid\bx) \parallel \normal(\cdot\mid\bzero, \bI_K)) \right] 
= \Exp \left[ \frac{1}{2} \mathcalD(\bsigma_{\blambda}^2(\bx)) + \frac{1}{2} \normtwo{\bmu_{\blambda}(\bx)}^2 \right],
\end{aligned}
$$
where $\mathcalD(\bv) \triangleq \sum_{i=1}^K v_i - \ln v_i - 1$. The function $\mathcalD(v)$ has a unique minimum at $v = 1$.
This loss is intuitive: The mean $\bmu_{\blambda}(\bx)$ is penalized for deviating from zero, and the variance $\bsigma_{\blambda}^2(\bx)$ is penalized for deviating from one. 
For the full VAE loss, we assume a scalar variance for the encoder---that is, $\diag(\bsigma^2_{\btheta}(\bz))=\sigma^2_{\btheta}(\bz)\bI_D$--yielding:
\begin{mybox}
\begin{align}
& \mathcalJ_{\text{VAE}}^\beta(\blambda, \btheta) 
= \mathcalJ_{\text{VAE-Recon}}(\blambda, \btheta) + \beta \cdot\mathcalJ_{\text{VAE-Prior}}(\blambda) \nonumber \\
\small& = \Exp_{p_{\text{data}}(\cdot), q_{\blambda}(\cdot\mid\bx)} 
\left[ 
\underbrace{\frac{\normtwo{\bx - \bmu_{\btheta}(\bz)}^2}{2\sigma_{\btheta}^2(\bz)} }_{\text{recon. error}} 
+ \underbrace{\frac{D}{2} \ln \sigma_{\btheta}^2(\bz)}_{\text{confidence}} 
+ \underbrace{\frac{\beta}{2} \mathcalD(\bsigma_{\blambda}^2(\bx))}_{\text{latent var}\rightarrow \bone} 
+ \underbrace{\frac{\beta}{2} \normtwo{\bmu_{\blambda}(\bx)}^2}_{\text{latent mean}\rightarrow\bzero} 
\right] .
\label{equation:vae_intui_loss_isoto}
\end{align}
\end{mybox}
The four terms in this loss function are highly intuitive. 
The first term is the standard reconstruction error. The second term characterizes the decoder's uncertainty: a smaller variance indicates a more ``confident" decoder but also imposes a heavier penalty on reconstruction errors. Furthermore, we encourage the latent variance to approach 1 and the latent mean to approach 0, so that the latent distribution remains close to a standard Gaussian.

The final loss function is optimized using gradient-based methods, most commonly stochastic gradient descent (SGD) or its variants, operating on mini-batches of data \citep{goodfellow2016deep, lu2022gradient}. Although $\btheta$ and 
$\blambda$ are updated jointly, the procedure can be conceptually interpreted as alternating between refining the encoder (inference network) and the decoder (generative model), analogous to the E- and M-steps in the EM algorithm.

Another key distinction from classical EM is that, for a fixed $\btheta$, optimizing $\blambda$ does not generally drive the KL divergence to zero. This is because the encoder network---despite its flexibility---cannot perfectly represent the true posterior 
$p_{\btheta}(\bz\mid \bx)$. Several factors contribute to this residual gap:
\begin{enumerate}[label=(\roman*)]
\item The true posterior may be non-Gaussian or non-factorized, potentially exhibiting complex dependencies across latent dimensions.
\item Even deep neural networks have finite representational capacity and cannot exactly model arbitrary distributions.
\item The optimization is approximate due to stochastic gradients, finite training iterations, and convergence to local optima---limitations also present in constrained EM methods (see Section~\ref{section:em_constrained}).
\end{enumerate}
Consequently, the ELBO remains a strict lower bound on the true log-likelihood, as illustrated in Figure~\ref{fig:ELBO_decom}.

\index{Reparameterization trick}
\subsection{Other Issues in VAEs}\label{section:other_issue_vae}

Unfortunately, even with the decomposition in \eqref{equation:vae_elbo_vae} with Gaussian parameterizations \eqref{equation:vae_encoder_gaussian} and \eqref{equation:vae_dec_gauss_diag}, 
the ELBO remains intractable to compute exactly. 
This is because the first term involves an integral over the latent variable $\bz_n$, and the integrand---due to the nonlinear decoder network---has a complex dependence on $\bz_n$. For a single data point $\bx_n$, the contribution to the ELBO can be written as (see \eqref{equation:vae_elbo_vae}):
\begin{equation}\label{equation:vae_elbo_decomposed}
\mathcalF_n(\btheta,\blambda) 
=
\Exp_{q_{\blambda}(\bz_n\mid\bx_n)} \left[ \ln {p_{\btheta}(\bx_n \mid\bz_n  )} \right] 
- \KL[q_{\blambda}(\bz_n\mid\bx_n) \parallel p_{\text{prior}}(\bz_n)].
\end{equation}
The second term is a KL divergence between two Gaussian distributions and has a closed-form expression (see Problem~\ref{problem:kl_vae}):
\begin{equation}
\KL[q_{\blambda}(\bz_n\mid\bx_n) \parallel p_{\text{prior}}(\bz_n)] 
= \frac{1}{2} \sum_{k=1}^K \left\{  (\bmu_\blambda(\bx_n))_k^2 + (\bsigma_\blambda^2(\bx_n))_k - \ln (\bsigma_\blambda^2(\bx_n))_k  - 1\right\}.
\end{equation}
For the first term in \eqref{equation:vae_elbo_decomposed}, a natural approach is to approximate the expectation using a Monte Carlo estimator:
\begin{equation}\label{equation:vae_monte_carlo_approx}
\int q_{\blambda}(\bz_n\mid\bx_n) \ln p_{\btheta}(\bx_n\mid\bz_n) \diff\bz_n \simeq \frac{1}{S} \sum_{s=1}^{S} \ln p_{\btheta}(\bx_n\mid\bz_n^{(s)}),
\quad \bz_n^{(s)}\sim q_{\blambda}(\bz_n\mid\bx_n).
\end{equation}
While this estimate is straightforward to differentiate with respect to $\btheta$, 
computing gradients with respect to $\blambda$ is problematic: the samples $\bz_n^{(s)}$  depend on $\blambda$ through the encoder distribution, yet once drawn, they are treated as fixed constants. Consequently, standard backpropagation cannot propagate gradients through the sampling operation.
Conceptually, fixing $\bz_n$ to sampled values blocks the flow of gradient information to the encoder parameters $\blambda$; that is, when the ELBO is estimated using fixed samples from $q_{\blambda}(\bz_n\mid \bx_n)$, the error signal cannot be backpropagated through the stochastic sampling step to update the encoder network.

This issue is resolved by the \textit{reparameterization trick}, which rewrites the sampling procedure so that randomness is decoupled from the model parameters. 
Specifically, if $\bepsilon\sim\normal(\bzero, \bI)$, then
\begin{equation}\label{eq:reparam_trick}
\bz = \diag(\bsigma)  \bepsilon + \bmu
\end{equation}
follows a Gaussian distribution $\normal(\bmu, \diag(\bsigma^2))$  (see Lemma~\ref{lemma:affine_mult_gauss}). 
Applying this to our encoder, we replace direct sampling from $q_{\blambda}(\bz_n \mid \bx_n )$ with:
\begin{equation}\label{equation:vae_sample_reparam}
\bz_n^{(s)}\sim q_{\blambda}(\bz_n \mid \bx_n )
\quad\implies \quad
\bepsilon_n^{(s)} \sim \normal(\bzero, \bI), \bz_n^{(s)} 
= \bmu_{\blambda}(\bx_n)+ \bsigma_{\blambda}(\bx_n)\hadaprod \bepsilon_n^{(s)},
\end{equation}
where $\hadaprod$ denotes the \textit{Hadamard product}, and $s = 1,2,\ldots,S$ indexes Monte Carlo samples.
Note that we usually set $\bsigma_{\blambda}(\bx_n) = \exp(\frac{1}{2} \ln \bsigma^2_{\blambda}(\bx_n))$ to ensure nonnegativity of the standard deviation.
This reformulation makes the dependence on $\blambda$ explicit and differentiable, enabling gradient-based optimization via automatic differentiation.

Although the reparameterization trick applies primarily to continuous latent variables, alternative gradient estimators exist for discrete cases (e.g., the REINFORCE estimator; \citep{williams1992simple}). However, these methods typically suffer from high variance. Thus, the reparameterization trick also acts as an effective variance-reduction technique \citep{lu2023bayesian}.

Under our modeling assumptions, the full VAE loss function  (averaged over a mini-batch $\sT\subset\{1,2,\ldots,N\}$) becomes:
\begin{mybox}
\begin{equation}\label{equation:vae_loss}
\mathcalJ(\blambda, \btheta) = -\mathcalF 
\simeq  \frac{N}{\abs{\sT}}\sum_{n\in\sT} \left\{ \frac{1}{2} \sum_{k=1}^K \left(\mu_{nk}^2 + \sigma_{nk}^2  - \ln \sigma_{nk}^2 -1 \right) 
- \frac{1}{S} \sum_{s=1}^S \ln p_{\btheta}(\bx_n\mid\bz_n^{(s)}) \right\}.
\end{equation}
\end{mybox}
where the latent sample $\bz_n^{(s)}$ is constructed as  $z_{nk}^{(s)} = \sigma_{nk} \epsilon_{nk}^{(s)} + \mu_{nk}$, with $\mu_{nk} = (\bmu_{\blambda}(\bx_n))_k$ and $\sigma_{nk} = (\bsigma_{\blambda}(\bx_n))_k$.
To simplify further, we may again set $\sigma_{\btheta}^2(\bz) = \sigma^2$ to a constant value, yielding:
\begin{mybox}
\begin{equation}\label{equation:vaeloss_cst_var}
\mathcalJ(\blambda, \btheta)
\simeq
\frac{N}{\abs{\sT}}\sum_{n\in\sT} \left\{ \frac{1}{2} \sum_{k=1}^K \left(\mu_{nk}^2 + \sigma_{nk}^2  - \ln \sigma_{nk}^2 -1 \right) 
+ \frac{1}{S}
\sum_{s=1}^S  \frac{1}{2\sigma^2} \normtwo{\bx_n - \widehatbx_n^{(s)}}
\right\},
\end{equation}
\end{mybox}
where $\widehatbx_n^{(s)} = \bmu_\btheta(\bz_n^{(s)}) 
=  \bmu_\btheta\big(\bmu_{\blambda}(\bx_n)+ \bsigma_{\blambda}(\bx_n)\hadaprod \bepsilon_n^{(s)}\big)$.
In practice, learning a data-dependent Gaussian decoder variance $\sigma_{\btheta}^2$ can be numerically unstable  and may lead to degenerate solutions without proper regularization. 
For improved stability, many implementations thus fix $p_{\btheta}(\bx\mid\bz) = \normal(\bx\mid \bmu_{\btheta}(\bz), \sigma^2 \bI_D)$ with a constant $\sigma^2$, which makes the reconstruction term proportional to the mean squared error (up to constants), as shown in \eqref{equation:vaeloss_cst_var}.

In practice, the number of Monte Carlo samples per data point is often set to $S=1$. Although this produces  a noisy estimate of the ELBO, the noise is compatible with stochastic gradient optimization and generally leads to faster and more efficient training.


VAE training proceeds as follows: for each data point in a mini-batch, 
(i) perform a forward pass through the encoder to obtain $\bmu_{\blambda}(\bx_n)$ and $\bsigma_{\blambda}(\bx_n)$, (ii) sample  $\bepsilon_n^{(s)}$ and compute $\bz_n^{(s)}$ via reparameterization, (iii) pass $\bz_n^{(s)}$ through the decoder to evaluate the reconstruction log-likelihood or the full ELBO in \eqref{equation:vae_loss}, and (iv) compute gradients of the ELBO with respect to both $\btheta$ and  
$\blambda$ via automatic differentiation. 
The complete procedure is summarized in Algorithm~\ref{alg:amo_vae}.

\begin{algorithm}[h] 
\caption{Variational Autoencoder (VAE) using Reparameterization Trick}
\label{alg:amo_vae}
\begin{algorithmic}[1] 
\Require Observed data points $\mathcalX=\{\bx_1, \bx_2, \ldots, \bx_N\}\in\real^D$;
\State \textbf{initialize:} Model parameters $\btheta,  \blambda$; 
\State \textbf{parameters:} Step size $\eta$, MC sample number $S$, latent dimension $K<D$;
\State Set maximum number of iterations $C$, iteration counter: $t=0$;
\While{$t<C$} 
\State Sample mini-batch $\sT\subset \{1,2,\ldots,N\}$; 
\Comment{(VAE$_1$)}
\State $\bepsilon^{(s)}\sim\normal(\bzero, \bI_K)$ for all $s=1,2,\ldots,S$;  
\Comment{(VAE$_2$)}
\State $\bz_n^{(s)} \leftarrow  \bmu_{\blambda}(\bx_n)+ \bsigma_{\blambda}(\bx_n)\hadaprod \bepsilon_n^{(s)}$  for all $n=1,2,\ldots,N$, $s=1,2,\ldots,S$;    
\Comment{(VAE$_3$)}
\State $\mathcalJ \leftarrow \frac{N}{\abs{\sT}}\sum_{n\in\sT} \left\{ \frac{1}{2} \sum_{k} \left(\mu_{nk}^2 + \sigma_{nk}^2 - \ln \sigma_{nk}^2 -1 \right) 
-\frac{1}{S} \sum_{s} \ln p_{\btheta}(\bx_n\mid\bz_n^{(s)}) \right\}$;
\Comment{(VAE$_4$)}
\State $\btheta \leftarrow \btheta -\eta \nabla_{\btheta}\mathcalJ $; \Comment{(VAE$_5$)}
\State $\blambda \leftarrow \blambda -\eta \nabla_{\blambda}\mathcalJ $; \Comment{(VAE$_6$)}
\State  Increment iteration counter: $t\leftarrow t+1$;
\EndWhile
\State Output $\btheta$, $\blambda$;
\end{algorithmic} 
\end{algorithm}

\paragrapharrow{Evaluation and generative process.}
After training, to evaluate how well the model represents a new test point $\widetildebx$, we use the ELBO 
$\mathcalF$ as a tractable lower bound on the log-likelihood. For a tighter estimate, it is preferable to sample from the approximate posterior {$q_{\blambda}(\bz\mid\widetildebx)$} rather than from the prior $p_{\text{prior}}(\bz)$, since the former concentrates probability mass in regions relevant to $\widetildebx$.

Once the model is trained and evaluated, the encoder network is discarded. 
New data points are generated by sampling from the prior $p_{\text{prior}}(\bz)$ and forward-propagating through the decoder network to obtain samples in the data space: $p_{\btheta}(\bx\mid \bz)$.
This differs from standard AEs, where the latent code is a deterministic function of the input. In a VAE, the encoder outputs a distribution over latent codes, and actual codes are obtained by sampling---making the model inherently probabilistic (see Figure~\ref{fig:ae_vae}).

This probabilistic framework makes VAEs particularly effective for generative tasks, such as image and sequence synthesis \citep{kingma2013auto, rezende2014stochastic}. Because the latent space is regularized to be smooth and continuous, \textbf{interpolating} between two latent vectors typically yields meaningful transitions in the data space.
See Figure~\ref{fig:vae_latent_regular} for an illustrative example.

However, VAEs are known to sometimes produce blurrier images compared to alternatives such as GANs \citep{goodfellow2020generative}. 
This occurs because the reconstruction objective encourages averaging over plausible outputs to minimize expected error.

\paragrapharrow{Conditional VAE.}
Several variants of the VAE have been developed. 
For image data, encoders typically use convolutional layers, while decoders employ transposed convolutions. In a \textit{conditional VAE}, both the encoder and decoder receive an additional guidance variable $\by$ (e.g., a class label). The prior over the latent variable can either remain the standard $p_{\text{prior}}(\bz)$ or be extended to a conditional prior $p(\bz\mid\by)$, which may be parameterized by a separate neural network. Training proceeds identically to the standard VAE.
During generation, the user can specify a particular value of $\by$ to guide the model toward producing more relevant or targeted outputs.

\paragrapharrow{General framework.}
Rather than imposing a standard Gaussian assumption for $p_{\text{prior}}(\bz)$, we may analyze the model under general latent distributions. We still employ the reparameterization:
$$
\bz_n\sim q_{\blambda}(\bz_n \mid \bx_n )
\implies 
\bepsilon_n \sim \normal(\bzero, \bI), \bz_n = \bmu_{\blambda}(\bx_n)+ \bsigma_{\blambda}(\bx_n)\hadaprod \bepsilon_n,
$$
where $\hadaprod$ represents the {Hadamard product}. And the corresponding ELBO objective in \eqref{equation:vae_elbo} becomes
$$
\mathop{\max}_{\btheta,\blambda} 
\sum_{n=1}^{N} \Exp_{p(\bepsilon)} 
\left[ \ln \frac{p_{\btheta}\big(\left\{\bmu_{\blambda}(\bx_n)+ \bsigma_{\blambda}(\bx_n)\hadaprod \bepsilon_n\right\}, \bx_n \big)}{q_{\blambda}\big(\left\{\bmu_{\blambda}(\bx_n)+ \bsigma_{\blambda}(\bx_n)\hadaprod \bepsilon_n\right\}\mid\bx_n\big)} \right].
$$
A Monte Carlo approximation yields:
$$
\sum_{n=1}^{N} 
\frac{1}{S}
\sum_{s=1}^{S}
\left[ \ln \frac{p_{\btheta}\big(\left\{\bmu_{\blambda}(\bx_n)+ \bsigma_{\blambda}(\bx_n)\hadaprod \bepsilon_n^{(s)}\right\}, \bx_n \big)}{q_{\blambda}\big(\left\{\bmu_{\blambda}(\bx_n)+ \bsigma_{\blambda}(\bx_n)\hadaprod \bepsilon_n^{(s)}\right\}\mid\bx_n\big)} \right],
\quad
\bepsilon_n^{(s)}\sim\normal(\bzero, \bI),
$$
which enables end-to-end gradient-based learning via backpropagation.

\paragrapharrow{VAEs as latent models.}
From the above discussion, VAEs can be regarded as full generative models, where the encoder exists only to assist training of the complementary decoder, which maps a Gaussian prior distribution to the target data distribution.
Samples can then be generated by sampling
$$
\text{$\bz \sim p_{\text{prior}}$ and then $\bx \sim p_\btheta(\bx\mid\bz)$}.
$$
In  Section~\ref{section:ldm_intro}, we will introduce \textit{latent diffusion models (LDMs)}, 
in which VAEs are used as latent encoders to map the original high-dimensional data space into a compact latent space, after which a separate generative model is trained within this learned latent space.

Why, then, do we choose to train an additional generative model in the latent space? The reason lies in the so-called \textit{amortization gap} between the left- and right-hand sides of both \eqref{equation:vaejoint_dl_ineq} and \eqref{equation:vaejoint_dl_ineq2}, which arises from the chain decomposition of KL divergence and corresponds exactly to the gap in the information-processing inequality.
This gap vanishes if and only if $q_\blambda(\bz\mid\bx) = p_\btheta(\bz\mid\bx)$, meaning the encoder perfectly represents the true posterior. 
Thus, even though minimizing $\KL\big[q_\blambda(\bx,\bz)\parallel p_\btheta(\bx,\bz)\big]$ implies minimizing  $\KL\big[q_\blambda(\bz)\parallel p_{\text{prior}}(\bz)\big]$ (see \eqref{equation:vaejoint_dl_ineq2}), a reduction in the former does not guarantee a proportional reduction in the latter. 
As a result, after training terminates, neither $\KL\big[q_\blambda(\bx,\bz)\parallel p_\btheta(\bx,\bz)\big]$  nor the amortization gap
\begin{equation}
	\KL\big[q_\blambda(\bx,\bz)\parallel p_\btheta(\bx,\bz)\big] - \KL\big[q_\blambda(\bz)\parallel p_{\text{prior}}(\bz)\big]
\end{equation}
is fully minimized, so $q_\blambda(\bz)\neq p_{\text{prior}}(\bz)$. 

This behavior can also be interpreted from the ELBO decomposition in \eqref{equation:vae_lower_bound}. 
The gap between the model evidence (log-likelihood) $\ln p_\btheta(\bx)$ and the ELBO $\mathcalF$ is zero if and only if $\KL(q(\bz) \parallel p_\btheta(\bz\mid\bx))=0$ (see \eqref{equation:elbo_kl_vae}). 
An improvement in the ELBO does not guarantee a corresponding improvement in the true log-likelihood.
At the end of training, the ELBO is generally not fully maximized, and the gap remains non-negligible.

Furthermore, note  that during training,  the decoder learns to reconstruct inputs from samples drawn from $q_\blambda(\bz)$ rather than the prior $p_{\text{prior}(\bx)}(\bz)$. 
Switching to sampling from $p_{\text{prior}}(\bz)$ 
at generation time therefore introduces a distributional shift relative to the training distribution. In practice, however, this mismatch is beneficial rather than harmful.

Finally, empirical evidence shows that \textit{diffusion} and \textit{flow models} are generally more expressive than the convolutional architectures typically used for VAE decoders. It is therefore sensible to delegate the more challenging parts of generative modeling to a dedicated latent-space model; see Chapters~\ref{chapter:diff}--\ref{chapter:flow}. 
Additionally, variational formulations of diffusion and flow models can themselves be interpreted as instances of VAEs; see, for example, Section~\ref{section:dpm_sec}.

\subsection{Hierarchical Variational Autoencoder}\label{section:hvae}

\begin{figure}[h!]
\centering  
\vspace{-0.35cm} 
\subfigtopskip=2pt 
\subfigbottomskip=2pt 
\subfigcapskip=-5pt 
\subfigure[VAE.]{\label{fig:lvm_HVAE_raw}
\includegraphics[width=0.27\linewidth]{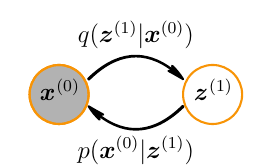}}
\subfigure[Hierarchical VAE.]{\label{fig:lvm_HVAE}
\includegraphics[width=0.65\linewidth]{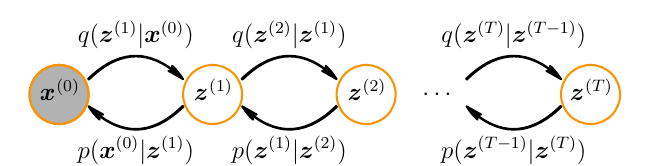}}
\caption{Comparison of standard VAE and hierarchical VAE, illustrating  the hierarchical VAE with $T$ hierarchical latent levels. 
The generative process is formulated as a Markov chain, where each latent $\bz^\toptzero$ is generated only from the preceding latent variable $\bz^\toptone$.
The shaded cycle represents the observed variable, while the unshaded cycles represent unobserved latent variables.}
\label{fig:lvm_HVAE_all}
\end{figure}

A \textit{hierarchical variational autoencoder} (\textit{HVAE}, also referred to as the  \textit{recursive VAE})   is a generalized variant of the standard VAE that extends the vanilla VAE framework by introducing multi-level hierarchical latent variables \citep{sonderby2016ladder, luo2022understanding}. 
In this hierarchical formulation, latent variables are hierarchically structured, such that lower-level latent variables are generated from more abstract, higher-level latent representations.

Conventional HVAE models with $T$ hierarchical levels allow each latent variable to depend on all preceding latent variables in the hierarchy. In contrast, this work focuses on a restrictive and simplified variant termed \textit{Markovian HVAE (MHVAE)}.
In MHVAE, the entire generative process follows a strict Markov chain property: each hierarchical decoding step is Markovian, where the generation of each latent variable $\bz^\toptzero$ depends only on its immediate preceding latent variable $\bz^\toptone$. We denote the original input data as $\bx^\topzero = \bx$. Conceptually and visually, this architecture can be interpreted as a sequential stacking of standard VAE modules, as visualized in Figure~\ref{fig:lvm_HVAE_all}.

Mathematically, the joint generative distribution and approximate posterior distribution of the  MHVAE are formally defined as follows:
\begin{align}
p_{\btheta}(\bx^\topzero, \bz^\topone, \ldots, \bz^\topT) 
&= p(\bz_T)p_\btheta(\bx^\topzero\mid\bz^\topone)\prod_{t=2}^{T}p_\btheta(\bz^\toptminus\mid\bz^\toptzero); 
\label{equation:hvae_joint_prob}\\
q_\blambda(\bz^\topone, \ldots, \bz^\topT\mid\bx^\topzero) 
&= q_\blambda(\bz^\topone \mid\bx^\topzero)\prod_{t=2}^{T}q_\blambda(\bz^\toptzero \mid\bz^\toptminus).
\label{equation:hvae_joint_post}
\end{align}
Based on the above formulations, we derive the ELBO for MHVAE as:
\begin{align}
\ln p_{\btheta}(\bx^\topzero) 
&\geq \Exp_{q_\blambda(\bz^\topone, \ldots, \bz^\topT\mid\bx^\topzero)} \left[ \ln \frac{p_{\btheta}(\bx^\topzero, \bz^\topone, \ldots, \bz^\topT)}{q_\blambda(\bz^\topone, \ldots, \bz^\topT\mid\bx^\topzero)} \right] \label{equation:elbo_hvae}\\
&= \Exp_{q_\blambda(\bz^\topone, \ldots, \bz^\topT\mid\bx^\topzero)} \left[ \ln \frac{p(\bz_T)p_\btheta(\bx^\topzero\mid\bz^\topone)\prod_{t=2}^{T}p_\btheta(\bz^\toptminus\mid\bz^\toptzero)}{q_\blambda(\bz^\topone\mid\bx^\topzero)\prod_{t=2}^{T}q_\blambda(\bz^\toptzero\mid\bz^\toptminus)} \right].
\end{align}
where the inequality follows similarly from the ELBO decomposition \eqref{equation:elbo_vfe_negv2}, and the last equality follows from   plugging in the joint distribution \eqref{equation:hvae_joint_prob} and posterior \eqref{equation:hvae_joint_post} (cf. \eqref{equation:vae_elbo_vae}).
As we will show in Chapter~\ref{chapter:diff},  this hierarchical ELBO objective can be further decomposed into interpretable components when analyzing diffusion models.

\index{Generative adversarial networks (GANs)}
\index{Generator}
\index{Discriminator}
\section{Generative Adversarial Networks (GANs)}\label{section:gans}

A VAE establishes a probabilistic generative framework grounded in latent variable inference. It posits that real-world data originates from a low-dimensional latent space. The encoder maps observed data to the posterior distribution of latent variables, while the decoder reconstructs the original data from sampled latent codes.
The optimization objective of VAE is the ELBO, which balances the regularization of latent distributions and reconstruction fidelity. Essentially, VAE learns an explicit probability density function of data and achieves generative modeling by sampling latent variables and decoding the sampled latent representations.

\textit{Generative adversarial networks (GANs)} \citep{goodfellow2020generative}, by contrast, employ a dual adversarial training paradigm composed of two core components: a \textit{generator} and a \textit{discriminator}. The generator synthesizes fake samples that mimic the distribution of real data, whereas the discriminator differentiates authentic real samples from generator-synthesized counterparts.
The two networks engage in iterative adversarial competition to promote mutual performance improvement. Through confrontational optimization, the generator progressively learns to capture the underlying distribution of real data without explicit density modeling, ultimately producing highly realistic synthetic samples.

Accordingly, GANs differ fundamentally from VAEs in three key aspects:
\begin{itemize}
\item \textit{Modeling form.} VAEs implement explicit probabilistic distribution modeling, while GANs perform implicit data distribution fitting.
\item \textit{Training mechanism.} VAEs are trained by optimizing a combined loss function of reconstruction loss and regularization terms, whereas GANs update network parameters via a minimax adversarial game. 
\item \textit{Latent space property.} VAEs leverage a statistically interpretable and constrained latent variable space, whereas the latent space of GANs is not subject to strict statistical constraints.
\end{itemize}

\begin{figure}[htbp]
\centering
\includegraphics[width=0.99\textwidth]{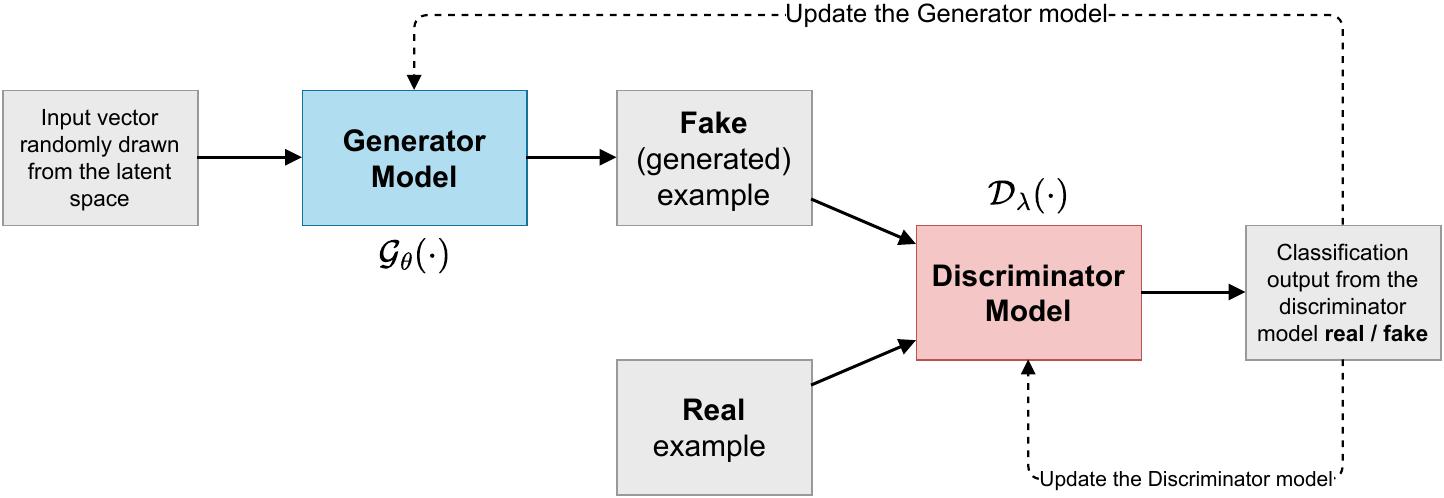}
\caption{
Schematic illustration of the GAN framework.	
In this architecture, the discriminator neural network $\mathcalD_\blambda(\bx)$ is trained to differentiate genuine training samples from synthetic samples generated by the generator network $\mathcalG_\btheta(\bz)$. The generator is optimized to maximize the discriminator's classification error by generating high-fidelity, realistic images. Conversely, the discriminator learns to minimize this error iteratively, improving its capability to accurately distinguish between real and synthetic data samples.
}
\label{fig:gan_schematic}
\end{figure}

\index{Zero-sum game}
\subsection{Adversarial Training}
We consider a generative model that maps a latent space $\bz$ to the data space $\bx$ via a nonlinear transformation. 
We first define a latent distribution $p_{\text{latent}}(\bz)$,  typically chosen as a standard Gaussian distribution:
\begin{equation}
p_{\text{latent}}(\bz) = \normal(\bz\mid \bzero, \bI).
\end{equation}
Coupled with this latent prior is a nonlinear mapping $\bx = \mathcalG_\btheta(\bz)$ implemented by a deep neural network with trainable parameters  $\btheta$, termed the \textit{generator}. 
Together, these two components implicitly induce a data distribution over  $\bx$. 
Our objective is to align this learned distribution with the empirical distribution of a training dataset $\mathcalX=\{\bx_n\}_{n=1}^N$. 
Nevertheless, direct maximum likelihood optimization over the generator parameters $\btheta$ is infeasible, since the exact likelihood of the generated data distribution generally lacks a closed-form expression.
To address this issue, \textit{generative adversarial networks (GANs)} introduce a secondary neural network, the \textit{discriminator}, which is jointly trained with the generator. The discriminator serves to provide informative training signals for updating the generator's weights throughout the optimization process. A schematic overview of this adversarial training framework is presented in Figure~\ref{fig:gan_schematic}.

The discriminator network is designed to differentiate authentic real samples sourced from the training dataset from synthetic ``fake" samples generated by the generator, and is optimized via \textbf{minimizing} standard classification loss. In contrast, the generator aims to \textbf{maximize} this classification error by producing synthetic samples that closely mimic the statistical distribution of real training data. The two networks thus compete against one another during the training process, giving rise to the term ``\textit{adversarial}". 
This training paradigm constitutes a classic \textit{zero-sum game}, where any performance improvement of one network corresponds to a performance degradation of the other.

This adversarial framework enables the discriminator to furnish effective supervisory training signals for generator optimization, converting the intractable unsupervised density estimation task into a feasible supervised learning problem.

To formalize the above training objective, we define a binary classification target variable as follows:
\begin{align}
r &= 1, & & \text{real data},  \\
r &= 0, & & \text{synthetic data}. 
\end{align}
The discriminator adopts a single output unit paired with a \textit{logistic-sigmoid activation function}, whose output quantifies the probability that an input data vector $\bx$ is authentic real data:
\begin{equation}
P(r=1) = \mathcalD_\blambda(\bx). 
\end{equation}
Training of the discriminator is conducted using the standard \textit{cross-entropy loss function}, expressed as:
\begin{equation}\label{equation:gan_ce_func}
\mathcalJ_{\text{CE}} = -\frac{1}{N} \sum_{n=1}^N \left\{ r_n \ln d_n + (1-r_n) \ln(1-d_n) \right\} ,
\end{equation}
where $d_n = \mathcalD_\blambda(\bx_n)\in[0,1]$ denotes the discriminator's output for the $n$-th input data vector, and the loss is normalized by the total number of data samples. 

The training batch consists of two types of samples: real data samples $\bx_n$ and  synthetic samples generated as $\mathcalG_\btheta(\bz_n)$, where each latent variable $\bz_n$  is randomly sampled from the predefined latent distribution $p_{\text{latent}}(\bz)$. 
Given the label assignment rule $r_n=1$ for real samples and $r_n=0$ for synthetic samples, the cross-entropy loss in  \eqref{equation:gan_ce_func} can be rewritten in terms of statistical expectations over data and latent distributions:
\begin{mybox}
\begin{subequations}\label{equation:gan_error_all}
\begin{align}
\mathcalJ_{\text{GAN}}(\blambda, \btheta) 
&= - \Exp_{p_{\text{data}}(\bx)} \big[\ln \mathcalD_\blambda(\bx)\big]
-  \Exp_{p_{\text{latent}}(\bz)} \big[\ln(1 - \mathcalD_\blambda(\mathcalG_\btheta(\bz)))\big] \\
&\simeq  -\frac{1}{N_{\text{real}}} \sum_{n \in \text{real}} \ln \mathcalD_\blambda(\bx_n)
- \frac{1}{N_{\text{synth}}} \sum_{n \in \text{synth}} \ln(1 - \mathcalD_\blambda(\mathcalG_\btheta(\bz_n))) ,
\end{align}
\end{subequations}
\end{mybox}
where $p_{\text{data}}(\bx)$ represents the underlying distribution of real training data, and $p_{\text{latent}}(\bz)$ denotes the prior noise distribution in the latent space.
In practical training with the dataset  $\mathcalX=\{\bx_1, \bx_2, \ldots,\bx_N\}$, the number of real samples $N_{\text{real}}$ is typically set equal to the number of synthetic samples $N_{\text{synth}}$. 
The integrated generator-discriminator architecture supports end-to-end training via stochastic gradient descent (SGD), with gradients computed through backpropagation. 

A distinctive feature of GAN optimization is its adversarial alternating training scheme: the loss function is \textbf{minimized with respect to the discriminator parameters $\boldsymbol{\blambda}$} but \textbf{maximized with respect to the generator parameters $\btheta$}:
\begin{align}
&\min_{\mathcalD_\blambda} 
- \Exp_{p_{\text{data}}(\bx)} \big[\ln \mathcalD_\blambda(\bx)\big]
-  \Exp_{p_{\text{latent}}(\bz)} \big[\ln(1 - \mathcalD_\blambda(\mathcalG_\btheta(\bz)))\big]; \\
&\max_{\mathcalG_\btheta} 
-  \Exp_{p_{\text{latent}}(\bz)} \big[\ln(1 - \mathcalD_\blambda(\mathcalG_\btheta(\bz)))\big].
\end{align}
Specifically, one SGD step is executed for each network using a mini-batch of samples, followed by the generation of new synthetic samples for the next iteration.

This maximization objective for the generator can be implemented via standard gradient-descent optimization by inverting the sign of the gradient, leading to the following parameter update rules:
\begin{align}
\blambda \leftarrow \blambda + \Delta \blambda,
\qquad \Delta \blambda &= -\eta \nabla_{\blambda} \mathcalJ_n(\blambda, \btheta); \label{equation:gan_lambda_gd} \\
\btheta \leftarrow \btheta + \Delta \btheta,
\qquad 
\Delta \btheta &= \eta \nabla_{\btheta} \mathcalJ_n(\blambda, \btheta) , \label{equation:gan_theta_gd}
\end{align}
where $\mathcalJ_n(\blambda, \btheta)$ denotes the per-sample or mini-batch loss function, and $\eta$ is the learning rate.  
The opposite signs in the two update rules reflect the adversarial dynamic: the discriminator is updated to reduce classification loss, while the generator is updated to increase this loss. 
When the generator achieves optimal performance, the generated synthetic samples become statistically indistinguishable from real data, causing the discriminator to output a uniform probability of 0.5 for all inputs. 

Upon training convergence, the discriminator is discarded, and the trained generator is solely retained for inference. New data samples can be synthesized by sampling latent vectors from the latent prior distribution and forwarding them through the generator network. Theoretical analysis further demonstrates that given sufficiently expressive generator and discriminator networks, a fully converged GAN is capable of exactly matching the learned generative distribution $\mathcalG_{\btheta}(\bz), \bz\sim p_{\text{latent}}(\bz)$ to the true underlying data distribution $p_{\text{data}}(\bx)$.

\index{Conditional GANs}
\index{Autoencoding conditional GANs}
\subsection{Conditional GANs and  Variants}

The standard GAN model discussed above generates samples from the unconditional distribution $p(\bx)$. 
For instance, it can synthesize horse images when trained on a horse image dataset. 
Extending this framework, \textit{conditional GANs (CGANs)} \citep{mirza2014conditional} are designed to sample from the conditional distribution $p(\bx\mid \by)$, where the guidance vector $\by$ can encode categorical attributes such as different horse breeds. 
To enable conditional generation, both the generator and discriminator networks take $\by$ as an additional input variable, and model training relies on labeled paired training samples $\{\bx_n, \by_n\}$. After training converges, targeted class-specific images can be generated by feeding the corresponding class vector into the conditioning variable $\by$. Compared with training independent GAN models for each individual class, the CGAN framework enables joint learning of shared internal representations across all classes, thus achieving more efficient data utilization.

In practical implementation, the conditional information $\by$ corresponding to each data sample $\bx$ is first encoded into a low-dimensional latent variable via a deep parameterized network $\mathcalE_\bphi$ (consistent with the encoding mechanism of autoencoders): $\bz_y = \mathcalE_\bphi(\by)$. The resulting conditional latent vector $\bz_y$ is then concatenated with the original latent input and fed into the generator network. The full CGAN objective function is formulated as follows:
\begin{mybox}
\begin{subequations}\label{equation:cgan_error_all}
	\small
\begin{align}
\mathcalJ_{\text{CGAN}}&(\blambda, \btheta, \bphi) 
= - \Exp_{p_{\text{data}}(\bx, \by)} \big[\ln \mathcalD_\blambda(\bx\mid \by)\big]
-  \Exp_{p_{\text{latent}}(\bz), p(\by)} \big[\ln(1 - \mathcalD_\blambda\big(\mathcalG_\btheta(\bz, \mathcalE_\bphi(\by))\mid\by\big))\big] \\
&\simeq  -\frac{1}{N_{\text{real}}} \sum_{n \in \text{real}} \ln \mathcalD_\blambda(\bx_n\mid\by_n)
- \frac{1}{N_{\text{synth}}} \sum_{n \in \text{synth}} \ln\big(1 - \mathcalD_\blambda\big(\mathcalG_\btheta(\bz_n, \mathcalE_\bphi(\by_n))\mid \by_n\big)\big) .
\end{align}
\end{subequations}
\end{mybox}
The above equation again presents the exact theoretical loss and its discrete empirical approximation based on real and synthetic sample batches.

\begin{figure}[h!]
\centering                      
\vspace{-0.35cm}                 
\subfigtopskip=2pt               
\subfigbottomskip=2pt            
\subfigcapskip=-5pt              
\subfigure[Generator in CGAN.]{\label{fig:generator_cgan}
	\includegraphics[width=0.55\linewidth]{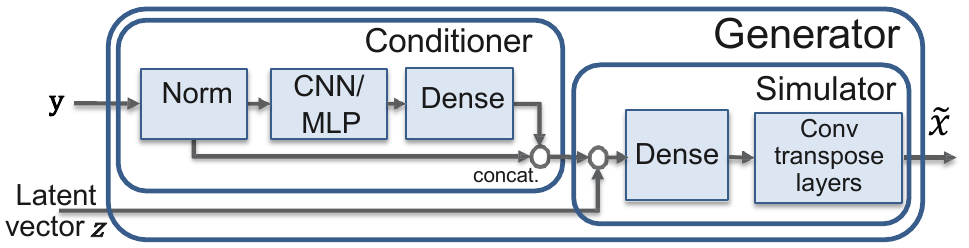}}
\subfigure[Discriminator in CGAN.]{\label{fig:discriminator_cgan}
	\includegraphics[width=0.38\linewidth]{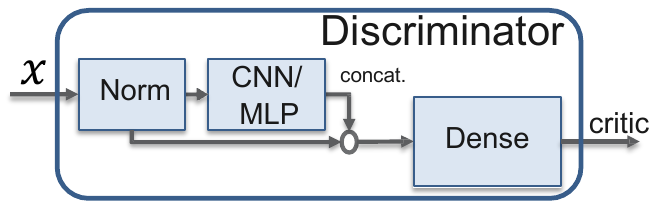}}
\caption{
Schematic illustration of a CGAN, in which a discriminator neural network  is trained to distinguish between real samples from the training set, and synthetic samples produced by the generator network. 
The generator takes latent $\bz$ and guided information $\by$ as inputs, which aims to maximize the error of the discriminator network by producing realistic images, whereas the discriminator network tries to minimize the same error by becoming better at distinguishing real from synthetic examples. 
Adapted from \citet{lu2022autoencoding, lu2022hybrid}.
}
\label{fig:cgan_all}
\end{figure}

\begin{SCfigure}
\centering
\includegraphics[width=0.6\textwidth]{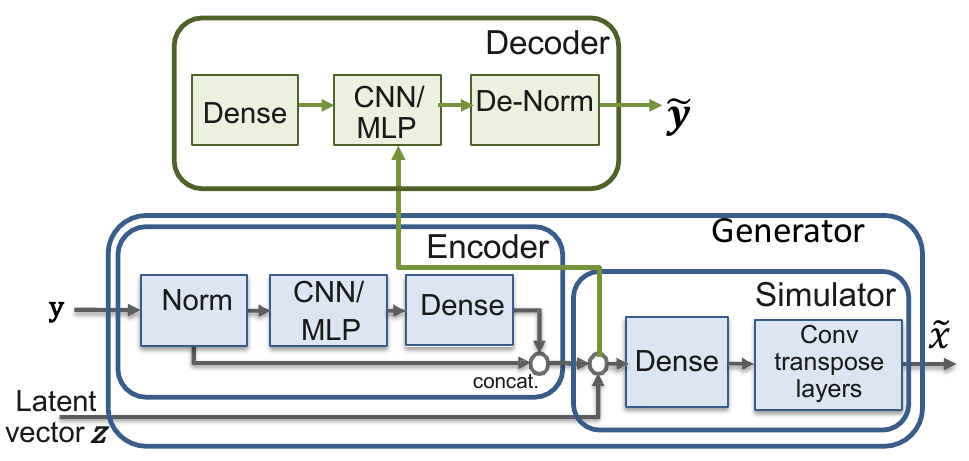}
\caption{Schematic illustration of the generator in an ACGAN, in which the decoder is used to enforce that the encoder keeps raw information as much as possible.
Adapted from \citet{lu2022autoencoding, lu2022hybrid}.
}
\label{fig:generator_acgan}
\end{SCfigure}

A critical limitation of the standard CGAN framework is its over-reliance on the \textit{conditioner} to extract discriminative features for misleading the discriminator (see Figure~\ref{fig:generator_cgan}). This limitation is negligible when the discriminator is fully and optimally trained. However, in most practical scenarios---particularly for tasks with limited real data (e.g., financial time series modeling)---the discriminator cannot be perfectly optimized. This imperfect discriminator training causes the CGAN conditioner to discard essential intrinsic information from the original guidance data (or historical data in financial time series modeling). 
To address this issue, the \textit{autoencoding conditional GAN (ACGAN)} model achieves a balanced optimization between conditional information extraction and adversarial generation. It integrates an autoencoder module to preserve the intrinsic attributes of conditional input data (e.g., inherent patterns in financial time series) during adversarial training \citep{lu2022autoencoding, lu2022hybrid}. The objective functions for generators in CGAN and ACGAN are compared below:
\begin{mybox}
\small
\begin{align}
\text{(CGAN)} \,\, &\max_{\mathcalG_\btheta, \mathcalE_\bphi}  
-  \Exp_{p_{\text{latent}}(\bz), p(\by)} 
\big[\ln(1 - \mathcalD_\blambda\big(\mathcalG_\btheta(\bz, \mathcalE_\bphi(\by))\mid\by\big))\big];\\
\text{(ACGAN)} \,\, &\max_{\mathcalG_\btheta, \mathcalE_\bphi, \mathcalD_\bpsi}  
-  \Exp_{p_{\text{latent}}(\bz), p(\by)} 
\big[\ln(1 - \mathcalD_\blambda\big(\mathcalG_\btheta(\bz, \mathcalE_\bphi(\by))\mid\by\big))\big]
-\beta_e f(\mathcalD_\bpsi (\mathcalE_\bphi(\by)), \by),
\end{align}
\end{mybox}
where $f(\cdot,\cdot)$ denotes a general loss function (e.g.,  mean square loss function); $\mathcalD_\bpsi$ denotes the decoder that reconstruct the original guidance information $\by$ from the encoded information $\mathcalE_\bphi(\by)$.
The hyperparameter  $\beta_e$ balances the weight of the autoencoder regularization term, which is defined as the \textit{autoencoding penalty (AP)}. 
This penalty term constrains the encoder to retain essential guidance information, avoiding information loss caused by adversarial overfitting. The structural illustration of the ACGAN generator is presented in Figure~\ref{fig:generator_acgan}.

\index{Smoothed GANs}
\index{Least-squares GANs}
\subsection{Other Issues  in GANs}

While GANs can generate high-quality outputs, they are notoriously difficult to train effectively because of their adversarial learning framework. Furthermore, unlike conventional error minimization, there is no reliable measure of training progress, as the objective function can both increase and decrease during optimization.
A key training challenge is known as \textit{mode collapse}, where the generator's weights adapt during training such that all latent variable samples $\bz$ are mapped to only a limited subset of valid outputs. In extreme cases, the generator may produce just one or a very small number of distinct output values $\bx$. The discriminator then assigns a value of 0.5 to these samples, causing training to stall. For instance, a GAN trained on handwritten digit images might only generate the digit``8". Although the discriminator cannot distinguish these synthetic ``8"s from real ones, it fails to detect that the generator is not producing the full set of digit classes.

\begin{figure}[htbp]
\centering
\includegraphics[width=0.8\textwidth]{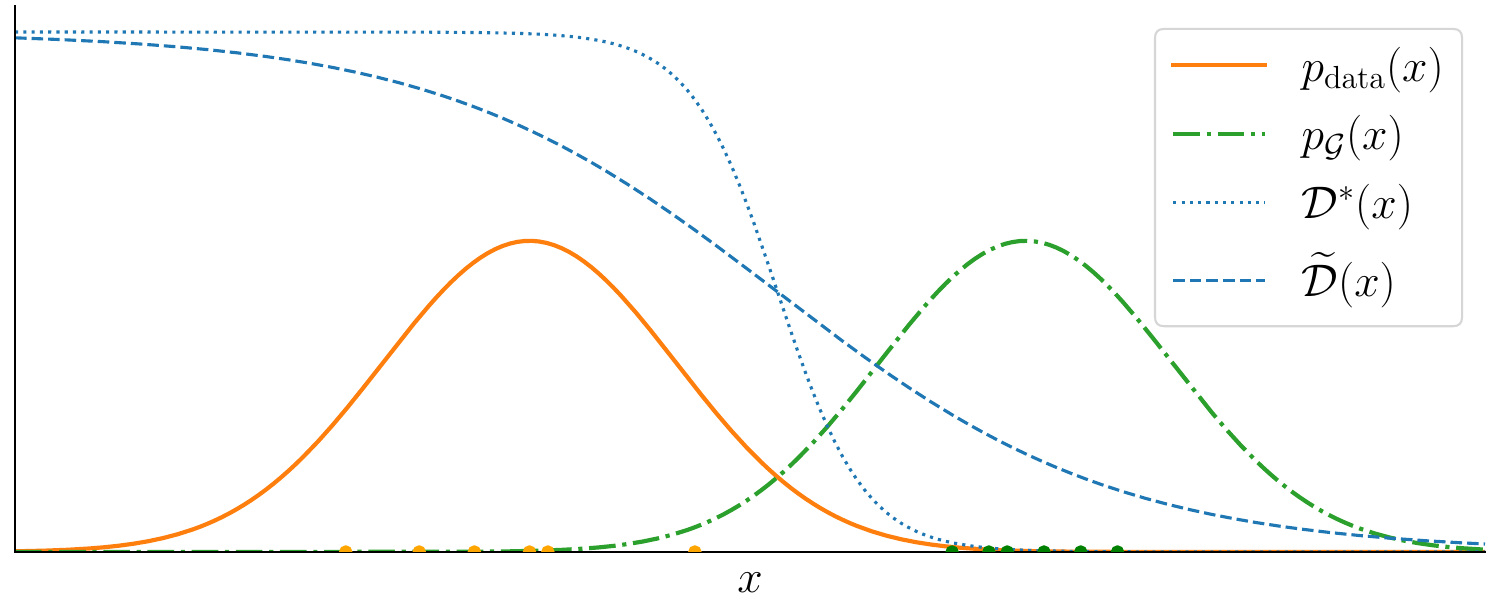}  
\caption{Conceptual diagram illustrating the challenges of training GANs.
It depicts a one-dimensional data space $x$ containing a fixed but unknown data distribution $p_{\text{data}}(x)$ and an initial generator distribution $p_\mathcalG(x)$. The optimal discriminator $\mathcalD^*(x)$ exhibits near-zero gradients near real and synthetic data points, severely slowing learning. A smoothed discriminator $\widetilde{\mathcalD}(x)$ can accelerate the training process.
}
\label{fig:gan_collapse}
\end{figure}

\paragrapharrow{Smoothed GANs.}
Further insight into the training difficulties of GANs can be gained from Figure~\ref{fig:gan_collapse}, which illustrates a simple 1D data space $x$ with samples $\{x_n\}$ drawn from a fixed but unknown data distribution $p_{\text{data}}(x)$. 
Figure~\ref{fig:gan_collapse} also depicts the initial generative distribution $ p_\mathcalG(x) $ together with samples drawn from this distribution. 
Since the real and generated distributions differ significantly, the optimal discriminator $ \mathcalD^*(x) = \frac{ p_{\text{data}}(x)}{ p_{\text{data}}(x)+ p_\mathcalG(x)} $
\footnote{This follows because $x\mapsto a\ln(x)+b\ln(1-x)$ is maximized over [0,1] at $\frac{a}{a+b}$.}
is straightforward to learn and decays very steeply, with near-zero gradients in regions around real and generated samples. 
Consider the second term in the GAN loss function \eqref{equation:gan_error_all}. 
For a fixed generator, $\mathcalD_\blambda(\mathcalG_\btheta(\bz) )$ is zero across the region covered by synthetic samples. 
As a result, small updates to the generator parameters $\btheta$ cause minimal changes in the discriminator's output, leading to small gradients and extremely slow learning.

This issue can be resolved by employing a \textbf{smoothed} variant $\widetilde{\mathcalD}(x)$ of the discriminator function,  as illustrated in Figure~\ref{fig:gan_collapse}, which provides stronger gradients to guide the training of the generator network. 
On the other hand, the \textit{least-squares GAN (LSGAN)} replaces the cross-entropy error function~\eqref {equation:gan_ce_func} with a \textit{mean squared  error loss}, thereby achieving smoothing by using a real-valued discriminator output 
(where $\mathcalD_\blambda(\bx)\in\real$, rather than a probability constrained to $[0,1]$) to mitigate gradient saturation \citep{mao2017least}:
\begin{itemize}
\item The discriminator is trained to minimize the squared error between its predictions and the corresponding target labels for both real and synthetic data:
\begin{mybox}
\begin{equation}
\min_{\blambda}
\left\{
\mathcalJ_{\text{LSGAN}}^D(\blambda) = \Exp_{p_{\text{data}}(\bx)} \left[ \left( \mathcalD_\blambda(\bx) - b \right)^2 \right] 
+ \Exp_{p_{\text{latent}}(\bz)} \left[ \left( \mathcalD_\blambda(\mathcalG_\btheta(\bz)) - a \right)^2 \right]
\right\}.
\end{equation}
\end{mybox}
The first term pushes $\mathcalD_\blambda(\bx)$ toward $b$ (typically 1) for real samples, and  the second term pushes $\mathcalD_\blambda(\mathcalG_\btheta(\bz))$ toward $a$ (typically 0) for fake/generated data.
\item The generator is trained to minimize the squared error between the discriminator's outputs on fake samples and the target label $c$ (typically set to $b$):
\begin{mybox}
\begin{equation}
\min_{\btheta}
\left\{
\mathcalJ_{\text{LSGAN}}^G(\btheta) 
= \Exp_{p_{\text{latent}}(\bz)} \left[ \left( \mathcalD_\blambda(\mathcalG_\btheta(\bz)) - c \right)^2 \right]
\right\},
\end{equation}
\end{mybox}
where  $c=1$ in most implementations.
This encourages the generator to produce samples that move toward the discriminator's decision boundary, rather than being ignored entirely (as occurs in standard GANs when generated samples lie far from this boundary).
\end{itemize}

\begin{SCfigure}
\centering
\includegraphics[width=0.6\textwidth]{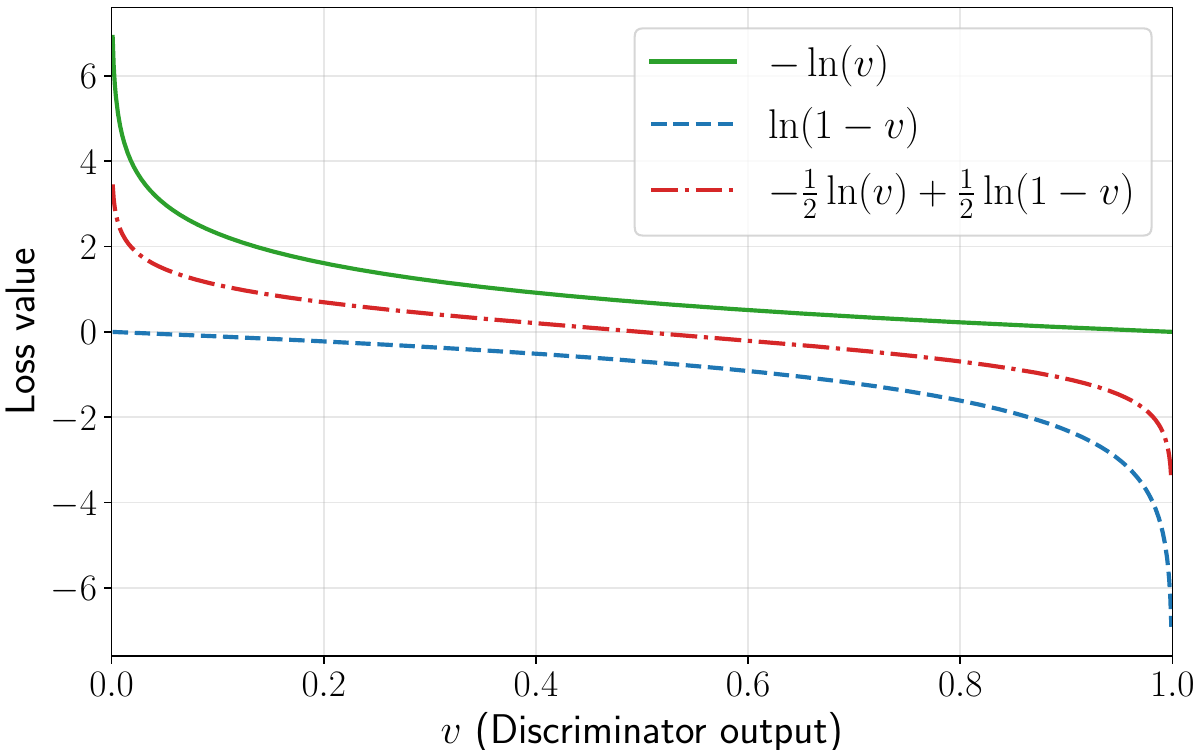}
\caption{Plots of $-\ln(v)$ and $\ln(1-v)$, illustrating the sharply contrasting gradient behavior near $v = 0$ and $v = 1$.}
\label{fig:GAN_lninv}
\end{SCfigure}

\paragrapharrow{Modified generator loss.}
Many other modifications to the GAN loss function and training algorithm have been introduced to stabilize and improve training \citep{mescheder2018training}. 
A widely adopted adjustment involves replacing the generator term in the original loss function
\begin{equation}
\max_{\mathcalG_\btheta} - \Exp_{p_{\text{latent}}(\bz)} [\ln(1 - \mathcalD_\blambda(\mathcalG_\btheta(\bz_n) ))]
\end{equation}
with the following alternative formulation:
\begin{mybox}
\begin{equation}
\max_{\mathcalG_\btheta}\Exp_{p_{\text{latent}}(\bz)} [\ln \mathcalD_\blambda(\mathcalG_\btheta(\bz_n) )]. 
\end{equation}
\end{mybox}
While the first expression minimizes the likelihood that a generated sample is classified as fake, the second maximizes the probability that it is classified as real. The distinct characteristics of these two formulations can be explained using Figure~\ref{fig:GAN_lninv}. When the generator distribution $p_\mathcalG(\bx)$ differs substantially from the true data distribution $p_{\text{data}}(\bx)$, the value $\mathcalD_\blambda(\mathcalG_\btheta(\bz))$ becomes near zero. In this regime, the original form yields an extremely small gradient, while the modified form produces a large gradient, resulting in more rapid learning.

In rare instances, the two loss functions can be combined to maintain a large gradient at both extremes, as illustrated by the red curve in Figure~\ref{fig:GAN_lninv}:
\begin{mybox}
\begin{equation}
\max_{\mathcalG_\btheta}\Exp_{p_{\text{latent}}(\bz)} [
\tfrac{1}{2} \ln \mathcalD_\blambda(\mathcalG_\btheta(\bz_n) )
- \tfrac{1}{2} \ln(1 - \mathcalD_\blambda(\mathcalG_\btheta(\bz_n) ))
].
\end{equation}
\end{mybox}

\paragrapharrow{Wasserstein GANs.}
A more direct approach to encourage the generator distribution $p_\mathcalG(\bx)$  to converge toward the data distribution $p_{\text{data}}(\bx)$ is to adjust the loss function to directly quantify the distance between these two distributions in the data space. 
This separation can be measured using the \textit{Wasserstein distance}, also referred to as the earth mover's distance. 
Intuitively, we can visualize the distribution $p_\mathcalG(\bx)$ as a pile of soil that is gradually relocated in small amounts to form the target distribution $p_{\text{data}}(\bx)$. The Wasserstein metric corresponds to the total mass of soil moved multiplied by the average distance it is transported. Among all possible ways to reshape the soil pile to match $p_{\text{data}}(\bx)$, the Wasserstein distance is defined by the plan that minimizes this average moving distance.

In practice, this distance cannot be computed directly, so it is approximated using a discriminator network with real-valued outputs. Early approximations constrained the gradient $\nabla_{\bx} \mathcalD_\blambda(\bx)$ of the discriminator with respect to the input $\bx$ using weight clipping, leading to the original \textit{Wasserstein GAN} \citep{arjovsky2017wasserstein}. A more robust refinement introduces an explicit gradient penalty, resulting in the \textit{gradient-penalized Wasserstein GAN} \citep{gulrajani2017improved}, whose loss function is defined as follows:
\begin{mybox}
\begin{subequations}
\begin{align}
\mathcalJ_{\text{WGAN}}(\blambda, \btheta) 
&=
- \Exp_{p_{\text{data}}(\bx)} \big[\ln \mathcalD_\blambda(\bx) - \beta_w \cdot\left( \normtwo{\nabla_{\bx} \mathcalD_\blambda(\bx)}^2 - 1 \right)^2\big] \nonumber\\ 
&\quad\quad\quad -  \Exp_{p_{\text{latent}}(\bz)} \big[\ln(1 - \mathcalD_\blambda(\mathcalG_\btheta(\bz)))\big]\\
&\simeq  
-\frac{1}{N_{\text{real}}} \sum_{n \in \text{real}} \left[ \ln \mathcalD_\blambda(\bx_n) - \beta_w \cdot\left( \normtwo{\nabla_{\bx_n} \mathcalD_\blambda(\bx_n)}^2 - 1 \right)^2 \right]  \nonumber\\
&\quad\quad\quad- \frac{1}{N_{\text{synth}}} \sum_{n \in \text{synth}} \ln (1-\mathcalD_\blambda(\mathcalG_\btheta(\bz_n)) ),
\end{align}
\end{subequations}
\end{mybox}
where $\beta_w$ governs the relative weighting of the gradient penalty term.

\begin{problemset}
\item \textbf{ELBO derivation: alternative perspective.} In \eqref{equation:elbo_vfe_negv2}, we show that the evidence (log-marginal likelihood) can be written as the sum of the ELBO and a KL divergence. 
Since the KL divergence is always nonnegative, this implies that the ELBO forms a lower bound on the evidence.
For a single data point $\bx$ and its corresponding latent variable $\bz$, this lower bound can be expressed as
$$
\ln p_{\btheta}(\bx) 
\geq \Exp_{q(\bz\mid \bx)} \left[\ln \frac{p_{\btheta}(\bx, \bz)}{q(\bz\mid \bx)}\right].
$$
Instead of deriving this bound using the standard ELBO decomposition, derive it alternatively using Jensen's inequality.
Similarly, derive the lower bound~\eqref{equation:elbo_hvae} for the HVAE:
$$
\ln p_{\btheta}(\bx^\topzero) 
\geq 
\Exp_{q_\blambda(\bz^\topone, \ldots, \bz^\topT\mid\bx^\topzero)} \left[ \ln \frac{p_{\btheta}(\bx^\topzero, \bz^\topone, \ldots, \bz^\topT)}{q_\blambda(\bz^\topone, \ldots, \bz^\topT\mid\bx^\topzero)} \right].
$$



\item \label{problem:elbo_equa_em} 
In the EM algorithm (Section~\ref{section:em_uncons}), consider  iteration $t$. 
Suppose that we set $q^\toptone_{\bz_n}(\bz_n) = p_{\btheta^\toptzero}(\bz_n \mid \bx_n)$ for all $n=1,2,\ldots,N$ in the E-step. Show that the evidence lower-bound (ELBO) becomes tight, i.e., $\mathcalF(\{q_{\bz_n}(\bz_n)\}_{n=1}^N, \btheta^\toptzero)=
\mathcalL(\btheta^\toptzero) $.

\item \label{problem:forward_rever_KL} \textbf{Exclusive/inclusive KL.}
When fitting a parametric distribution $q_{\blambda}$ to a target distribution $p$ by minimizing the KL divergence $\KL[q_{\blambda} \parallel p]$ with respect to $\blambda$, this is known as \textit{reverse/exclusive KL} minimization. 
Show that this approach exhibits \textit{mode-seeking} (or \textit{zero-forcing}) behavior: the minimization forces $q_{\blambda}(x)=0$ wherever  $p(x)=0$, often causing $q_{\blambda}$ to concentrate around a single mode of $p$. 
Conversely, when  fitting $q_{\blambda} $ to $p$ by minimizing  $\KL[p\parallel q_{\blambda} ]$ with respect to $\blambda$, this is called \textit{forward/inclusive KL}. 
Show that this leads to \textit{mass-covering} (or \textit{mean-seeking}) behavior: $q_{\blambda}$ must assign non-negligible probability mass wherever $p$ does.

\item \label{problem:kl_vae}
\textbf{KL of Gaussians.}  Let  $q(\bx)=\normal( \bmu, \diag(\bsigma^2))$ and $p(\bx)=\normal( \bzero_D, \bI_D)$, where $\bmu, \bsigma^2\in\real^D$ (see Definition~\ref{definition:multivariate_gaussian}). Show that $\KL[q\parallel p]=\frac{1}{2}\sum_{n=1}^{D}(\mu_n^2+\sigma^2_n-\ln \sigma^2_n-1)$. 
This expression is commonly used as the KL regularization term in variational autoencoders (VAEs); see Equation~\eqref{equation:vae_elbo_vae}.

\item \label{problem:kl_gauss1} \textbf{KL of Gaussians.} Let  $p(x)=\normal(\mu_1, \sigma_1^2)$ and $q(x)=\normal(\mu_2,\sigma_2^2)$ with $\sigma_1, \sigma_2 \in \real_{+}$. 
Show that 
\begin{equation}
\KL[p \parallel q ] = \ln\frac{\sigma_2}{\sigma_1} +\frac{\sigma_1^2+(\mu_1-\mu_2)^2}{2\sigma_2^2}-\frac{1}{2}.
\end{equation}
Let $p(\bx) = \normal(\bx\mid \bmu_1, \sigma_1^2\bI_D)$ and $q(\bx) = \normal(\bx\mid \bmu_2, \sigma_2^2\bI_D)$ be isotropic Gaussians with $\sigma_1, \sigma_2 \in \real_{+}$ and $\bx \in \real^D$. Show that 
\begin{equation}\label{equation:kl_iso}
\KL[p \parallel q] 
= \frac{1}{2} \left[ \mathcalD\left(\frac{\sigma_1^2\bI_D}{\sigma_2^2\bI_D}\right) 
+ \frac{\normtwo{\bmu_1 - \bmu_2}^2}{\sigma_2^2} \right],  
\,\text{ with } \mathcalD(\bv) \triangleq \sum_{i=1}^D v_i - \ln v_i - 1.
\end{equation}
The expression above is intuitive: If the mean and variances coincide,  then $\KL[p \parallel q] = 0$. 
Further, the divergence increases with the squared  distance $\normtwo{\bmu_1 - \bmu_2}^2$ between the mean vectors. Finally, the function $\mathcalD(\alpha)$ has a unique minimum at $\alpha = 1$ so that $\KL[p \parallel q]$ is minimized when $\sigma_1 = \sigma_2$.

More generally, with diagonal covariance Gaussians $p(\bx) = \normal(\bx\mid \bmu_1, \diag(\bsigma_1^2))$ and $q(\bx) = \normal(\bx\mid \bmu_2, \diag(\bsigma_2^2))$ with $\bsigma_1, \bsigma_2 \in \real_{+}^D$ and $\bx \in \real^D$,  \eqref{equation:kl_iso} becomes
\begin{equation}\label{equation:kl_diag}
\KL[p \parallel q] 
= \frac{1}{2} \left[  \mathcalD\left(\frac{\bsigma_1^2}{\bsigma_2^2}\right) 
+ (\bmu_1 - \bmu_2)^\top \diag(\bsigma_2^2)^{-1} (\bmu_1 - \bmu_2) \right], 
\end{equation}
where $\frac{\bsigma_1^2}{\bsigma_2^2}$ denotes element-wise division.

\item \label{problem:kl_gauss2} \textbf{KL of Gaussians.}
Now consider the multivariate case: let $\normal_1(\bx)=\normal(\bmu_1, \bSigma_1)$ and $\normal_2(\bx)=\normal(\bmu_2,\bSigma_2)$ (Definition~\ref{definition:multivariate_gaussian}) with   $\bx\in\real^D$. Show that 
\begin{align}
\KL[\normal_1\parallel &\normal_2] 
= 
\frac{1}{2}\ln\abs{\bSigma_2\bSigma_1^{-1}}+\frac{1}{2}\trace\bSigma_2^{-1}
\big( (\bmu_1-\bmu_2)(\bmu_1-\bmu_2)^\top +\bSigma_1-\bSigma_2 \big)\\
&= \frac{1}{2} \left[ \ln \frac{\abs{\bSigma_2}}{\abs{\bSigma_1}} - D + \trace(\bSigma_2^{-1} \bSigma_1) + (\bmu_2 - \bmu_1)^\top \bSigma_2^{-1} (\bmu_2 - \bmu_1) \right].
\end{align}
More generally, for an arbitrary distribution $p(\bx)$ and a multivariate Gaussian $\normal(\bx)=\normal(\bmu, \bSigma)$ with $\bx\in\real^D$, show that 
$$
\KL[p\parallel \normal ]=
\int \frac{1}{2}(\bx-\bmu)^\top\bSigma^{-1}(\bx-\bmu) \diff\bx 
+
\frac{1}{2}\ln \abs{\bSigma}+\frac{D}{2}\ln2\pi + \int p(\bx)\ln p(\bx) \diff\bx.
$$

\item \label{problem:entropy_mgau} \textbf{Entropy of Gaussians.} 
As introduced in Section~\ref{section:elbo_cfe}, \textit{entropy} is closely related to the KL divergence. The entropy of a distribution $p(\bx)$ is defined as
\begin{equation}
\entropy[p(\bx)] = -\int p(\bx)\ln p(\bx) \diff\bx.
\end{equation}
For a multivariate Gaussian random vector $\rvx\sim \normal(\bmu,\bSigma)$  with $\rvx\in\real^D$ (Definition~\ref{definition:multivariate_gaussian}), show that 
$$
\entropy[\normal(\bmu,\bSigma)] = \frac{1}{2}\ln \abs{\bSigma}+ \frac{D}{2} \ln(2\pi e).
$$

\item \label{prob:mix_of_gauss} \textbf{Mixture of Gaussians.}
Consider a dataset $\mathcalX=\{\bx_1,\bx_2, \ldots,\bx_N\}\in\real^D$ generated from a Gaussian mixture model with $K$ components:
\begin{equation}
p(\bX\mid \bpi, \{\bmu_k\}, \{\bSigma_k\}) = 
\prod_{n=1}^{N}\sum_{k=1}^{K} \pi_k \normal(\bx_n \mid \bmu_k, \bSigma_k),
\end{equation}
where $\bX\in\real^{N\times D}$ contains the data points as rows, $\bmu_k\in\real^D$,  $\bSigma\in\real^{D\times D}$, and $\bpi=[\pi_1,\pi_2,\ldots,\pi_K]^\top$ are the \textit{mixing coefficients} satisfying $0\leq \pi_k\leq 1$  and $\sum_{k=1}^{K}\pi_k=1$.
Introduce a $K$-dimensional binary latent variable $\{\bz_n\}\in\{0,1\}^K$ for each sample $n=1,2,\ldots,N$, 
such that $z_{nk}\in\{0,1\}$ and $\sum_{k=1}^{K}z_{nk}=1$.
Using the EM algorithm (Algorithm~\ref{alg:em_alg}), derive the update equations for the parameters  $\bpi, \{\bmu_k\}, \{\bSigma_k\}$.
Show that the E-step computes the posterior responsibilities as
$$
\zeta_{nk} \leftarrow
\frac{\pi_k^{\text{old}}\normal({\bx_n}\mid \bmu_k^{\text{old}}, \bSigma_k^{\text{old}})}{\sum_{\ell=1}^{K}\pi_\ell^{\text{old}}\normal({\bx_n}\mid \bmu_\ell^{\text{old}}, \bSigma_\ell^{\text{old}})}, 
\quad \forall\, n=1,2,\ldots,N, k=1,2,\ldots,K,
$$
so that  $p(z_{nk}=1\mid \bx_n) = \zeta_{nk}$.
Then show that the M-step updates the parameters as follows:
\begin{align*}
N_k &\leftarrow \sum_{n=1}^{N}\zeta_{nk}; 
&&\bmu_k^{\text{new}} \leftarrow \frac{1}{N_k}\sum_{n=1}^{N} \zeta_{nk} \bx_n;\\
\bSigma_k^{\text{new}} &\leftarrow \frac{1}{N_k}\sum_{n=1}^{N} \zeta_{nk} (\bx_n-\bmu_k)(\bx_n-\bmu_k)^\top;
&&\pi_k^{\text{new}} \leftarrow \frac{N_k}{N}, 
\qquad \forall\,n, k.
\end{align*}
\textit{Hint: See Example~\ref{example:gmm_twoclus} for the special case of two clusters.}

\item \label{prob:mix_of_bern} \textbf{Mixture of Bernoullis.}
Following the same setup as in Problem~\ref{prob:mix_of_gauss}, but now assume each observation $\bx_n\in\{0,1\}^{D}$ is binary. Consider the Bernoulli mixture model:
\begin{equation}
p(\bX\mid \bpi, \{\btheta_k\}) = 
\prod_{n=1}^{N}\sum_{k=1}^{K} \pi_k \bernoulli(\bx_n \mid \btheta_k)
=\prod_{n=1}^{N}\sum_{k=1}^{K} \pi_k \prod_{d=1}^D\left(\theta_{kd}^{x_{nd}}  (1-\theta_{kd})^{(1-x_{nd})} \right),
\end{equation}
where $\btheta_k\in\real^D$ and each component satisfies $0\leq \theta_{kd}\leq 1$ for $d=1,2,\ldots,D$.
Again, introduce a $K$-dimensional binary latent variable $\{\bz_n\}\in\{0,1\}^K$ for each sample $n=1,2,\ldots,N$, 
with $z_{nk}\in\{0,1\}$ and $\sum_{k=1}^{K}z_{nk}=1$.
Show that the E-step computes
$$
\zeta_{nk} \leftarrow
\frac{\pi_k^{\text{old}} \bernoulli({\bx_n}\mid \btheta_k^{\text{old}})}{\sum_{\ell=1}^{K}\pi_\ell^{\text{old}} \bernoulli({\bx_n}\mid \btheta_\ell^{\text{old}})}, 
\quad \forall\, n=1,2,\ldots,N, k=1,2,\ldots,K,
$$
so that  $p(z_{nk}=1\mid \bx_n) = \zeta_{nk}$.
Then show that the M-step updates the parameters as
\begin{align*}
N_k &\leftarrow \sum_{n=1}^{N}\zeta_{nk}; 
\qquad \btheta_k^{\text{new}} \leftarrow \frac{1}{N_k}\sum_{n=1}^{N} \zeta_{nk} \bx_n;
\qquad \pi_k^{\text{new}} \leftarrow \frac{N_k}{N}, 
\quad \forall\,n, k.
\end{align*}

\index{Truncated SVD}
\index{Eckart-Young-Mirsky theorem}
\index{Low-rank approximation}
\item \label{theorem:young-theorem_frob}\textbf{Eckart--Young--Mirsky theorem and truncated SVD (TSVD) \citep{stewart1993early, lu2021numerical}.}
Suppose we wish to approximate a rank-$R$ matrix $\bA\in \real^{M\times N}$ by a lower-rank matrix  $\bB$ of rank $K$ ($K<R$), measured in the Frobenius norm (Definition~\ref{definition:frobernius-in-svd}):
$$
\bB = \mathop{\arg\min}_{\rank(\bB)\leq K} \, \normf{\bA - \bB}.
$$
Let $\bA_K$ be the \textit{truncated SVD} (TSVD) of $\bA$ with the top $K$ terms, i.e., $\bA_K = \sum_{i=1}^{K} \sigma_i\bu_i\bv_i^\top$ from the SVD of $\bA=\sum_{i=1}^{R} \sigma_i\bu_i\bv_i^\top$ by zeroing out the  smallest $R-K$ singular values of $\bA$. 
Show that $\bA_K$ is the optimal rank-$K$ approximation to $\bA$ in terms of the Frobenius norm, satisfying $\normf{\bA-\bA_K}^2 = \sum_{i\geq K+1}\sigma_i^2$.
What is the corresponding optimal approximation under the spectral norm (Definition~\ref{definition:spectral_norm})?

\item \textbf{One-hot autoencoder.} 
Utilize the MNIST dataset \citep{lecun1998mnist} to train a one-hot autoencoder, with the latent representation consisting of ten-dimensional one-hot vectors. 
Use these learned latent representations to train a classification network, and compare the resulting performance with that of other baseline models.

\end{problemset}

\newpage 
\chapter{Diffusion Models}\label{chapter:diff}
\begingroup
\hypersetup{
linkcolor=structurecolor,
linktoc=page,  
}
\minitoc \newpage
\endgroup
\lettrine{\color{caligraphcolor}W}
We observe that a highly effective strategy for building expressive generative models is to define a distribution $p(\bz)$ over a latent variable $\bz$, which is then mapped to the data space  $\bx$ via a deep neural network (see Figure~\ref{fig:gen_as_prob_trans} for an illustration).
This framework allows  $p(\bz)$ to adopt a simple, fixed form---such as the standard Gaussian distribution $\normal(\bzero, \bI)$---since the transformative flexibility of deep neural networks enables general-purpose modeling of complex data distributions over $\bx$. In Chapter~\ref{chapter:vae_gan}, we covered two prominent instantiations of this modeling framework: variational autoencoders (VAEs) and generative adversarial networks (GANs),  each adopting unique strategies for neural network design and training.

This chapter elaborates on another class of models within the same general framework: \textit{diffusion models}, formally termed \textit{denoising diffusion probabilistic models (DDPMs)}  \citep{sohl2015deep, ho2020denoising}. 
These models have emerged as the state-of-the-art approach across numerous generative tasks. Widely deployed in practical systems including Stable Diffusion and DALL-E 2 for image generation \citep{rombach2022high, ramesh2021zero, ramesh2022hierarchical} and WaveGrad for speech synthesis \citep{chen2020wavegrad}, DDPMs are generative models that learn to systematically remove noise from data through a controlled, probabilistic iterative process.
DDPM operation consists of two core phases:
\begin{itemize}
\item \textit{Forward diffusion (noising).}
Beginning with clean, real-world data (e.g., images, audio), small increments of Gaussian noise are repeatedly added across numerous timesteps until the original data signal is completely obscured, leaving only pure random Gaussian noise. This forward noising process is fixed, deterministic in procedure, and fully probabilistic in its output distribution.
\item \textit{Reverse diffusion (denoising).}
A neural network is trained to predict and subtract the injected noise step by step. Starting from pure random noise, the model performs iterative denoising to recover realistic data samples that align with the distribution of the training dataset.
\end{itemize}

Unlike VAEs, which generate data via single-step latent decoding, diffusion models can be interpreted as a hierarchical variant of VAEs. In this formulation, the encoder distribution is fixed by the predefined forward noising process, while only the generative reverse distribution is learned end-to-end. Diffusion models therefore synthesize new data samples through continuous, multi-step stochastic evolution.

Diffusion models offer distinct practical advantages: they feature simplified training pipelines, scale efficiently on parallel computing hardware, and eliminate the training instabilities and optimization challenges inherent to adversarial training. Empirically, they produce generative outputs matching or exceeding the quality of GAN-generated samples. Their primary limitation lies in computational efficiency: the multi-step iterative denoising process requires numerous forward passes through the neural network, increasing the computational cost of sample generation.

\section{Diffusion Probabilistic Models (DPMs)}\label{section:dpms}

The release of OpenAI's DALL-E series \citep{ramesh2021zero, nichol2021glide, ramesh2022hierarchical} has greatly popularized high-fidelity image generation, bringing diffusion model algorithms to widespread attention among researchers and industry practitioners. Originally proposed in 2015, diffusion models draw core inspiration from non-equilibrium thermodynamics \citep{sohl2015deep}. This chapter systematically introduces the fundamental principles of diffusion models.

\subsection{Diffusion Probabilistic Model}\label{section:dpm_sec}

\textit{Diffusion probabilistic models (DPMs)}, also simply referred to as \textit{diffusion models (DMs)}, were first proposed by \citet{sohl2015deep} in 2015. The original work motivates the model design through non-equilibrium statistical physics. In this chapter, we introduce and analyze DPMs from a probabilistic perspective.

\subsubsection{VAE and  Hierarchical VAE}
\begin{figure}[h!]
\centering  
\vspace{-0.35cm} 
\subfigtopskip=2pt 
\subfigbottomskip=2pt 
\subfigcapskip=-5pt 
\subfigure[VAE.]{\label{fig:lvm_HVAE_raw_DDPM}
\includegraphics[width=0.27\linewidth]{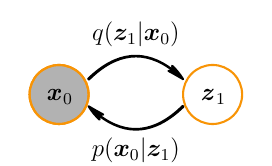}}
\subfigure[Hierarchical VAE.]{\label{fig:lvm_HVAE_DDPM}
\includegraphics[width=0.65\linewidth]{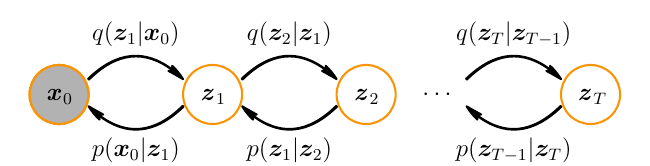}}
\caption{Comparison of standard VAE and hierarchical VAE, depicting the hierarchical VAE framework with $T$ hierarchical latent variables. 
The generative process is modeled as a Markov chain, where each latent variable  $\bz_t$ is generated only from the preceding  latent variable  $\bz_{t+1}$.
Shaded circles represent observed variables, while unshaded circles correspond to unobserved latent variables.}
\label{fig:lvm_HVAE_all_DDPM}
\end{figure}

We previously compare standard VAEs and hierarchical VAEs (HVAEs) in Section~\ref{section:hvae} and illustrate their structural differences in Figure~\ref{fig:lvm_HVAE_all}.
For consistent and simplified notation throughout this chapter, we adopt subscripts instead of superscripts to index sequence dimensions. To clarify the notational convention: prior chapters use subscripts to index individual data samples within a dataset (i.e., $\bx_n\in\mathcalX=\{\bx_1, \bx_2, \ldots, \bx_N\}$) and superscripts to denote sequential indices of latent or data variables. The key architectural distinctions between standard VAEs and HVAEs relevant to DDPM modeling are visualized in Figure~\ref{fig:lvm_HVAE_all_DDPM}.

From a theoretical perspective, diffusion models can be regarded as an extension of VAEs. Accordingly, we initiate our analysis from the fundamental VAE framework. We first review the basic VAE model (see Figure~\ref{fig:lvm_HVAE_raw_DDPM}), which involves two core variables: the observed variable $\rvx$ and the latent hidden variable $\rvz$.
For a training data sample (e.g., an input image) denoted as $\bx=\bx_0$, we treat the sample as a $D$-dimensional vector lying in the space $\real^D$. The autoencoder pipeline first encodes the input $\bx_0$ into a latent representation $\bz_1$, and then reconstructs the original input $\bx_0$ by decoding from $\bz_1$. Specifically, $\bx_0$ serves as the input of the encoder and the output of the decoder, whereas $\bz_1$ acts as the encoder output and the decoder input. As visualized in the figure, the conditional distribution $p_{\btheta}(\bx_0\mid\bz_1)$ characterizes the decoding process, and $q_{\blambda}(\bz_1\mid\bx_0)$ corresponds to the encoding process.

A standard VAE performs only a single encoding step. This naturally raises a question: can the encoding process be extended to multiple iterative steps?
As illustrated in Figure~\ref{fig:lvm_HVAE_DDPM}, extending the single-step VAE encoding to multi-step iterations yields a Markov chain-like chained structure with bidirectional propagation.
The left-to-right propagation corresponds to a progressive encoding process, equivalent to iteratively executing the VAE encoder $T$ times. Conversely, the right-to-left propagation represents a progressive decoding process, which is equivalent to $T$ iterative executions of the VAE decoder. We define the multi-step encoding procedure as the \textit{forward process} (or \textit{forward trajectory}) and the multi-step decoding procedure as the \textit{reverse process} (or \textit{reverse trajectory}).

Both the forward and reverse processes satisfy the Markov property, where the state at each timestep $t$ depends solely on the state at the previous timestep. In this framework, $q(\bz_t \mid \bz_{t-1})$ denotes a single-step encoding operation, and $p(\bz_{t-1} \mid \bz_{t})$ denotes a single-step decoding operation. All latent variables across timesteps $\bz_{1:T}$ (where $\bz_{1:T}$ denotes $\bz_1, \bz_2, \ldots, \bz_T$) collectively form the complete latent variable set of the hierarchical model.
The joint probability distribution of the entire hierarchical VAE model is formulated as:
\begin{equation}\label{equation:hvae_joint_pdf}
p_{\btheta}(\bx_0, \bz_{1:T}) = p(\bz_T)  \left\{\prod_{t=2}^T p_\btheta(\bz_{t-1} \mid \bz_t)\right\}
p_\btheta(\bx_0\mid \bz_1),
\end{equation}
where the conditional distribution $p_\btheta(\bz_{t-1} \mid \bz_t)$ is parameterized by a deep neural network with trainable parameters  $\btheta$, and the prior distribution $p(\bz_T)$ is determined by the noise corruption introduced during the multi-step encoding process (detailed in subsequent discussions).

The posterior distribution of the latent variables  $q_\blambda(\bz_{1:T} \mid \bx_0)$ conditioned on the observed input $\bx_0$ can be decomposed into a product of conditional transition distributions:
\begin{equation}\label{equation:hvae_posterior_decomp_ddpm}
q_\blambda(\bz_{1:T} \mid \bx_0) = q_\blambda(\bz_1 \mid \bx_0) \prod_{t=2}^T q_\blambda(\bz_t \mid \bz_{t-1}),
\end{equation}
where each conditional distribution $q_\blambda(\bz_t \mid \bz_{t-1})$ is also modeled by a deep neural network with trainable parameters $\blambda$.

In this hierarchical framework, only the variable $\bx=\bx_0$ remains observable, and the objective is to maximize the log-likelihood $\ln p_{\btheta}(\bx_0)$ of observed data samples. Following the standard VAE derivation, we derive the evidence lower-bound (ELBO) for optimization (see Section~\ref{section:elbo_cfe} for comprehensive derivation details):
\begin{equation}\label{equation:elbo_deri_hvae_ddpm}
\begin{aligned}
\ln p_{\btheta}(\bx_0) 
&= \ln \int p_{\btheta}(\bx_0, \bz_{1:T})  \diff\bz_{1:T} 
= \ln \int \frac{p_{\btheta}(\bx_0, \bz_{1:T}) q_\blambda(\bz_{1:T} \mid \bx_0)}{q_\blambda(\bz_{1:T} \mid \bx_0)} \diff\bz_{1:T} 
\\
&= \ln \Exp_{q_\blambda(\bz_{1:T} \mid \bx_0)} \left[ \frac{p(\bx_0, \bz_{1:T})}{q_\blambda(\bz_{1:T} \mid \bx_0)} \right]
\geq \Exp_{q_\blambda(\bz_{1:T} \mid \bx_0)} \left[ \ln \frac{p(\bx_0, \bz_{1:T})}{q_\blambda(\bz_{1:T} \mid \bx_0)} \right], 
\end{aligned}
\end{equation}
where the last inequality follows from Jensen's inequality.
Formula (\ref{equation:elbo_deri_hvae_ddpm}) is not much different from VAE, except that $\bz$ in VAE is replaced with the multi-timestep latent sequence $\bz_{1:T}$ (cf. \eqref{equation:vae_elbo_vae}).

\begin{figure}[htbp]
\centering
\includegraphics[width=0.9\textwidth]{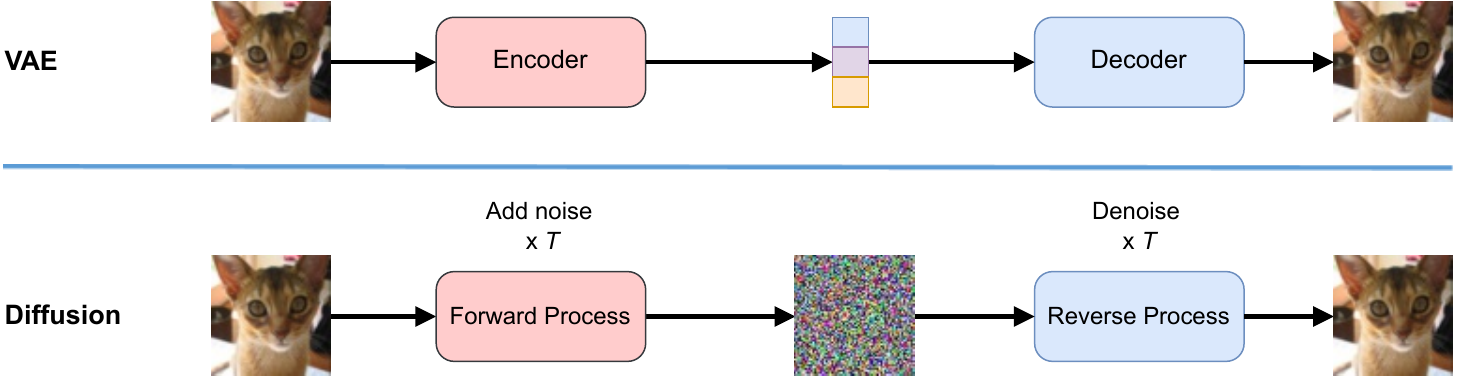} 
\caption{Conceptual illustration of the comparison between VAE and diffusion models. }
\label{fig:ddpm_vae_compare}
\end{figure}
\subsubsection{Diffusion Model}

Extending the standard VAE to a Markovian chain structure yields the Markovian hierarchical VAE (MHVAE), which serves as the fundamental prototype of diffusion models. In fact, diffusion models can be obtained by introducing minor modifications to the MHVAE framework. The core differences between diffusion models and MHVAE lie in the following three aspects:
\begin{enumerate}[label=(A\arabic*)]
\item \textit{Dimension consistency.} 
Each step of the forward and backward processes adopts identical input and output dimensions, such that all latent variables $\bz_t$ share the same dimensionality as the original input $\bx_0$.
\footnote{This is the reason why many existing studies denote $\bz_t$ as $\bx_t$. In this work, we retain the latent variable symbol $\bz_t$ to explicitly distinguish the functional roles of the two variables.} This setup differs from conventional VAEs, where the latent dimension of $\bz$ is typically smaller than the dimension of the observed input $\bx$.

\item \textit{Fixed forward transformation.} 
In the multi-step forward encoding process, the encoder distribution $q(\bz_t \mid \bz_{t-1})=q_{\blambda}(\bz_t \mid \bz_{t-1})$ is no longer a learnable neural network parameterized by $\blambda$. Instead, it is fixed as a predefined linear Gaussian transformation, whose explicit formulation is presented in the subsequent content.

\item \textit{Standard normal prior asymptotics.} 
Benefiting from the linear Gaussian encoder constraint and the inherent Markov chain properties, the latent variable $\bz_T$ asymptotically obeys a Gaussian distribution as the total timestep $T \to \infty$. In other words, $\bz_T$ converges to a Gaussian distribution with increasing $T$. By imposing a decay factor smaller than 1 on the linear Gaussian transformation, $\bz_T$ can be guaranteed to converge to the standard normal distribution, namely $p(\bz_T)\sim\normal(\bzero, \bI_D)$ in ~\eqref{equation:hvae_joint_pdf}, where $\bzero$ denotes the zero mean and $\bI_D$ denotes the isotropic unit covariance matrix.

\end{enumerate}
An intuitive conceptual comparison between standard VAEs and diffusion models is provided in Figure~\ref{fig:ddpm_vae_compare}.

\begin{figure}[htbp]
\centering
\includegraphics[width=0.9\textwidth]{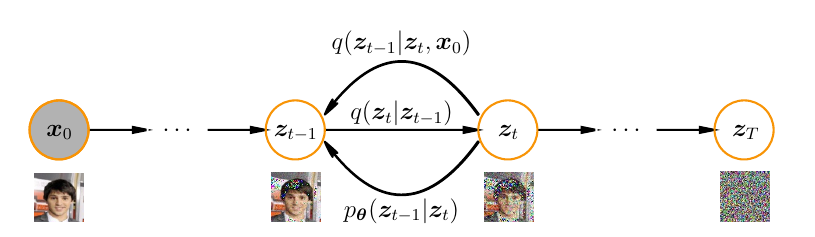} 
\caption{Graphical representation of the diffusion model. 
The original data $\bx=\bx_0$ is shown by the {shaded node}, since it is an observed variable, whereas the noise-corrupted data $\bz_1, \ldots, \bz_T$ are considered to be latent variables. The noise process is defined by the forward distribution $q(\bz_t\mid\bz_{t-1})$ and can be viewed as an encoder. Our goal is to learn a model $p_\btheta(\bz_{t-1}\mid\bz_t)$ that tries to reverse this noise process and which can be viewed as a decoder. As we will see later, the conditional distribution $q(\bz_{t-1}\mid\bz_t, \bx_0)$ plays an important role in defining the training procedure.}
\label{fig:diffusion_graphical_model}
\end{figure}
\subsection{Forward and Reverse Processes}\label{section:forward_rev_ddpm}

The process of a diffusion model is illustrated in Figure~\ref{fig:diffusion_graphical_model}, which corresponds to a Bayesian network (directed graphical model) with a Markov chain structure. 
The variable $\bx=\bx_0$ serves as the model's starting point and represents the distribution of real-world data (e.g., real images). 
Starting from $\bx_0$, the data distribution gradually evolves into a standard Gaussian distribution $\bz_T$ following this Markov chain structure.
The joint probability distribution of the entire graphical network is formulated as $p_{\btheta}(\bx_0, \bz_{1:T})$. 
Given the observed data vector $\bx=\bx_0$ where $\bx\sim p_{\text{data}}(\bx)$, the conditional joint distribution of the latent variables $\bz_{1:T}$ is defined as:
\begin{equation}\label{equation:forward_joint_ddpm}
q(\bz_1, \ldots, \bz_t \mid \bx_0) = q(\bz_1 \mid \bx_0) \prod_{\tau=2}^{t} q(\bz_\tau \mid \bz_{\tau-1}), 
\quad t=2,3,\ldots,T.
\end{equation}
This left-to-right Markovian evolution is termed the \textit{forward process} (also referred to as the \textit{diffusion process}, \textit{forward diffusion}, or \textit{forward noising}).
\footnote{
Analogous to the encoder module in a VAE, this forward process is fixed rather than trainable (refer to Assumption (A2)). 
Notably, this terminology differs from that used for flow-based models, where the latent-to-data mapping is defined as the forward process, as detailed in Chapter~\ref{chapter:flow}.}
The forward process iteratively injects noise into the real data sample $\bx_0$, ultimately converting it into a pure Gaussian random variable $\bz_T$: $\bz_T\sim\normal(\bzero, \bI_D)$,
\footnote{See Problem~\ref{prob:dpm_zT}.}
where $T$ denotes  the total number of diffusion steps, 
which corresponds to the number of noise injection operations in the forward process and matches the number of denoising steps required for the subsequent reverse process, as elaborated below and visualized in Figure~\ref{fig:diffusion_graphical_model}.

In contrast, the right-to-left evolution in Figure~\ref{fig:diffusion_graphical_model} constitutes the inverse of the forward diffusion process and is thus defined as the \textit{reverse process} (or \textit{denoising process}, \textit{reverse diffusion}). 
This process progressively transforms the standard Gaussian noise variable  $\bz_T$ into a realistic data sample $\bx_0$. 
In generative AI, this transformation pipeline corresponds to novel image generation, hence it is also known as the  \textit{image generation process}. 
Statistically, this procedure essentially implements sampling from the target joint probability distribution, and is therefore additionally referred to as the \textit{sampling process}. 
Consistent with this reverse evolution order, the joint distribution $p_{\btheta}(\bx_0, \bz_{1:T})$ can be factorized as follows (same as the HVAE formulation~\eqref{equation:hvae_joint_pdf}):
\begin{equation}\label{equation:reverse_joint_ddpm}
p_{\btheta}(\bx_0, \bz_{1:T}) = p(\bz_T) \left\{\prod_{t=T-1}^1 p_{\btheta}(\bz_t \mid \bz_{t+1})\right\} p_{\btheta}(\bx_0\mid \bz_1).
\end{equation}

\subsection*{Forward Process: Deterministic Gaussian Transformation}

Suppose we sample a data point (e.g., an image) from the training set and denote this observation as $\bx=\bx_0$.
The true probability density function of the real data distribution $p_{\text{data}}(\bx_0)$ remains unknown. 
However, we can access a collection of real data samples, which serve as observed realizations of  $\bx_0$. 
In this formulation, $\bx_0$ is the observed variable, while $\bz_{1:T}$ represent unobserved latent variables. 
Accordingly, the joint distribution of the entire Markovian network simplifies to the conditional distribution $q(\bz_{1:T} \mid \bx_0)$,
\footnote{
Note again that, consistent with standard diffusion model literature, latent variables are occasionally denoted as $\bx_1, \ldots, \bx_T$ in other works, with $\bx_0$ reserved for the observed data. This book adopts $\bz$ for latent variables and $\bx$ for observed variables to maintain notational consistency throughout the text.} 
which can be decomposed via the probability chain rule as \eqref{equation:forward_joint_ddpm}:
\begin{equation}\label{equation:forward_conditional_ddpm}
q(\bz_{1:T} \mid \bx_0) = q(\bz_1 \mid \bx_0) \prod_{t=2}^T q(\bz_t \mid \bz_{t-1}) = \prod_{t=1}^T q(\bz_t \mid \bz_{t-1}).
\end{equation}

Following  Assumption (A2), each latent variable $\bz_t$ follows a Gaussian distribution, and the forward encoder distribution $q(\bz_t \mid \bz_{t-1})$ corresponds to a fixed linear Gaussian transformation.  
Specifically, a linear Gaussian transformation implies that the mean of $\bz_t$ is a linear function of the preceding latent variable $\bz_{t-1}$.
Given a training data sample $\bx=\bx_0$, we corrupt the data by injecting independent Gaussian noise into each dimension of the vector. Treating the input data as a $D$-dimensional vector $\bx_0 \in \real^D$, the resulting noisy variable $\bz_1 \in \real^D$ is defined as:
\begin{subequations}\label{equation:forward_step_z1_ddpm_ALL}
\begin{align}
\bz_1 &= \sqrt{1 - \beta_1} \bx_0 + \sqrt{\beta_1} \bepsilon_1, \quad \text{with $\bepsilon_1 \sim \normal(\bzero, \bI_D)$}
\label{equation:forward_step_z1_ddpm}\\
\iff \quad q(\bz_1\mid\bx_0) &= \normal(\bz_1 \mid \sqrt{1 - \beta_1} \bx_0, \beta_1 \bI_D).
\label{equation:forward_step_z1_ddpm_v2}
\end{align}
\end{subequations}
where   $\beta_1 < 1$ controls the noise variance at the first diffusion step.  
The  weighted coefficients $\sqrt{1 - \beta_1}$ and $\sqrt{\beta_1}$ in~\eqref{equation:forward_step_z1_ddpm} or~\eqref{equation:forward_step_zt_ddpm} are carefully designed. They ensure that the mean of $\bz_t$ gradually approaches zero and its variance gradually approaches the identity matrix as the diffusion step progresses.

This noise injection procedure is repeated iteratively with independent Gaussian noise at each step, generating a sequence of progressively noisier variables $\bz_2, \ldots, \bz_T$.
Given a predefined \textit{noise scheduler} (or \textit{variance scheduler}) $\beta_1, \beta_2, \ldots,\beta_T$,
the update rule for each subsequent noisy latent variable is formulated as:
\begin{subequations}\label{equation:forward_step_zt_ddpm_ALL}
\begin{align}
\bz_t &= \sqrt{1 - \beta_t} \bz_{t-1} + \sqrt{\beta_t} \bepsilon_t, \quad \text{with $\bepsilon_t \sim \normal(\bzero, \bI_D)$}
\label{equation:forward_step_zt_ddpm}\\
\iff\quad  q(\bz_t \mid \bz_{t-1}) &= \normal\left( \bz_t \mid \sqrt{1 - \beta_t} \bz_{t-1},\, \beta_t \bI_D \right), 
\quad t=2,3,\ldots,T,
\label{equation:q_zt_given_zt_minus_1_ddpm}
\end{align}
\end{subequations}
where  the mean $\bmu_{\bz_t}$  of the conditional distribution $q(\bz_t \mid \bz_{t-1})$  is strictly a linear function of  $\bz_{t-1}$.
Equation~\eqref{equation:q_zt_given_zt_minus_1_ddpm} 
indicates that each forward step essentially adds zero-mean Gaussian noise with covariance $\beta_t \bI_D$ to the scaled version of the previous latent variable $\sqrt{1-\beta_t} \bz_{t-1}$.

\begin{figure*}[h]
\centering  
\subfigtopskip=2pt 
\subfigbottomskip=9pt 
\subfigcapskip=-5pt 
\includegraphics[width=0.99\textwidth]{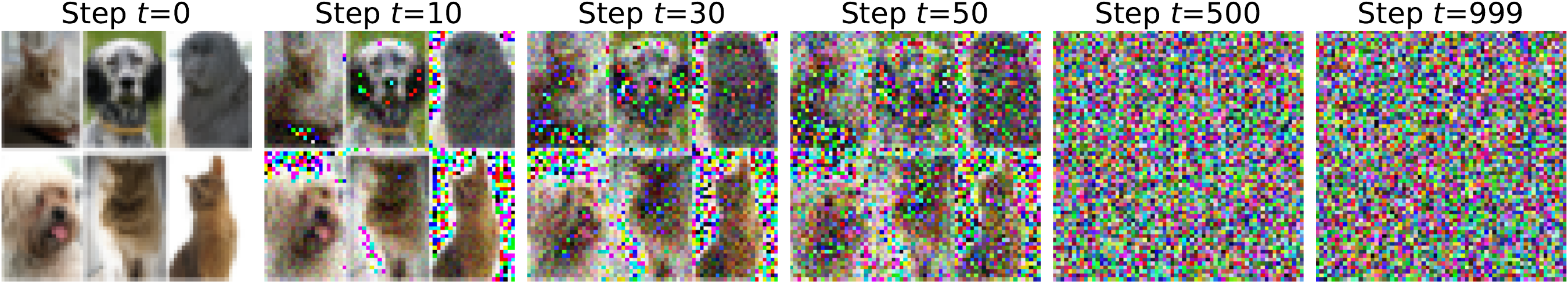}
\caption{Diffusion step for a set of images with $t=0, 10, 30, 50, 500, 999$ steps. In the process, we adopt a \textit{linear variance schedule} for the forward diffusion process \textcolor{black}{(i.e., $\beta_t = \beta_1+ \frac{t-1}{T-1}(\beta_T-\beta_1)$, for $t=1,2,\ldots,T$ and $T=1000$)}, following the protocol established by \citet{ho2020denoising}. 
This schedule ensures that the signal is gradually obscured by Gaussian noise, transitioning from nearly identity transformations at early steps to near-pure noise at the final step $T$.}
\label{fig:diffusion_celeb}
\end{figure*}

The sequence of conditional distributions defined in  \eqref{equation:q_zt_given_zt_minus_1_ddpm} forms a well-defined Markov chain, which can be represented as the probabilistic graphical model illustrated in Figure~\ref{fig:diffusion_graphical_model}. As the step index $t$ increases, the latent variable $\bz_t$ gradually degrades into pure Gaussian noise. As visualized in Figure~\ref{fig:diffusion_graphical_model} and Figure~\ref{fig:diffusion_celeb}, clean input images are progressively corrupted until they become indistinguishable from random Gaussian noise.
This degradation process can be intuitively analogized to \textit{ink diffusing in water}.
Initially concentrated ink gradually spreads uniformly throughout the liquid over time, rendering the original ink droplet unrecognizable and the entire solution uniformly turbid. This diffusion-inspired mechanism gives the model its name: the \textit{diffusion models} or \textit{diffusion probabilistic models (DPMs)}.

To guarantee that the latent mean converges to zero  ($\bmu_{\bz_t} \to \bzero$), all variance parameters are constrained to $\beta_t \in (0, 1)$. 
These hyperparameters can be manually predefined and follow a monotonic increasing schedule:  $\beta_1 < \beta_2 < \ldots < \beta_T<1$.
\footnote{The $\beta_t$ coefficients can be either manually fixed as static hyperparameters \citep{ho2020denoising} or optimized as learnable model parameters \citep{kingma2021variational}.}
Intuitively, small noise variance is applied at early diffusion steps, resulting in mild data corruption and slow diffusion. As the step proceeds, increasing noise variance amplifies corruption and accelerates the diffusion process.
The monotonic increasing design of the $\{\beta_t\}$ schedule offers two core advantages:
\begin{itemize}
\item Large noise magnitudes in early steps would cause drastic, irreversible distortion to the original data distribution, making it infeasible for the reverse denoising process to recover authentic signals.
\item Insufficient noise injection in late-stage steps (when data is already nearly randomized) leads to negligible distribution changes, slowing convergence and requiring excessive steps to achieve full Gaussianization.
\end{itemize}
The increasing variance schedule enables a slow-start and accelerating diffusion trajectory, which is optimal for stable and complete forward corruption. In summary, the forward diffusion process iteratively superimposes calibrated Gaussian noise on the original data until the structured data distribution evolves into pure standard Gaussian noise. Critically, the forward encoder distribution $q(\bz_t \mid \bz_{t-1})$ is non-trainable; it is fixed as a predefined linear Gaussian transformation throughout training and inference.

\subsection*{Forward Process: Diffusion Kernel}

\begin{figure}[H]
\centering  
\subfigtopskip=2pt 
\subfigbottomskip=9pt 
\subfigcapskip=-5pt 
\includegraphics[width=0.8\textwidth]{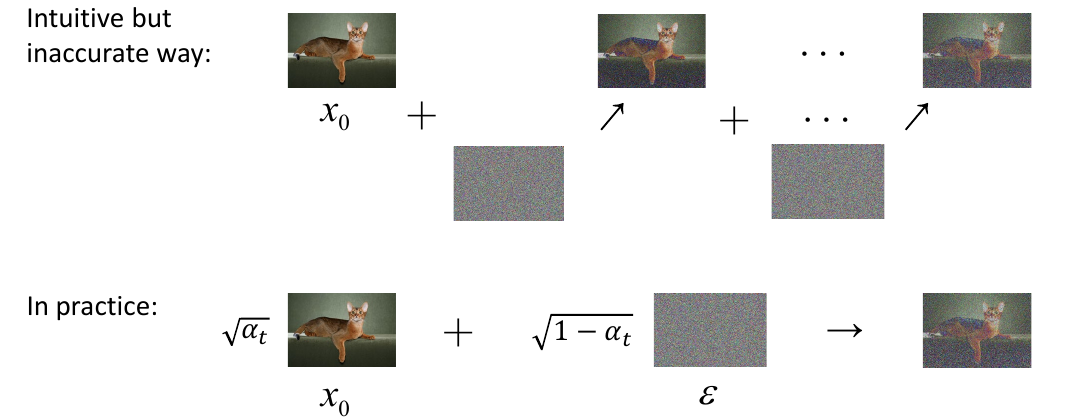}
\caption{Illustration of diffusion kernel. In practice, noise is not added incrementally but applied in a single step corresponding to timestep $t$.}
\label{fig:DDPM_fast_ff}
\end{figure}

To compute $\bz_t$ at each timestep of the forward diffusion process, the most straightforward approach is sequential iteration from $\bx_0$ to $\bz_T$ using the transition formula in \eqref{equation:q_zt_given_zt_minus_1_ddpm}. At each iteration, the noise term $\bepsilon_t$ is sampled from a standard normal distribution, following the noise sampling strategy used in standard VAEs (see Algorithm~\ref{alg:amo_vae}).
While this iterative update scheme is mathematically valid, it suffers from low computational efficiency,  especially when the total number of diffusion steps $T$ is large. 
By exploiting the linear Gaussian property of the transition distribution $q(\bz_t \mid \bz_{t-1})$, we derive a closed-form solution that directly computes $\bz_t$ from the original input data $\bx_0$ for any arbitrary timestep $t$. This closed-form expression supports \textbf{parallel computation} of all intermediate latent states $\{\bz_t\}$ and significantly  accelerates the forward diffusion process.
Specifically, the joint distribution of the latent variables, conditioned on the observed data vector $\bx=\bx_0$ ($\bx\sim p_{\text{data}}(\bx)$), is given by
\begin{equation}\label{equation:ddpm_joint_latent_dist_ddpm}
q(\bz_1, \ldots, \bz_t \mid \bx_0) = q(\bz_1 \mid \bx_0) \prod_{\tau=2}^{t} q(\bz_\tau \mid \bz_{\tau-1}).
\end{equation}
Marginalizing over the intermediate variables $\bz_1, \ldots, \bz_{t-1}$ yields  the \textit{diffusion kernel} (also called \textit{transition kernel} or \textit{noising kernel}):
\begin{subequations}\label{equation:ddpm_diffusion_kernel_all}
\begin{align}
q(\bz_t \mid \bx_0) &= \int q(\bz_{1:t} \mid \bx_0) \diff\bz_{1:t-1}
= \normal\left( \bz_t \mid \sqrt{\alpha_t} \bx_0,\, (1 - \alpha_t) \bI_D \right)
\label{equation:diffusion_kernel_ddpm} \\
\iff \quad  \bz_t &= \sqrt{\alpha_t} \bx_0 + \sqrt{1 - \alpha_t} \bepsilon_t, 
\quad\text{with } \bepsilon_t \sim \normal(\bzero, \bI_D).
\label{equation:diffusion_kernel_zt_form_ddpm}
\end{align}
\end{subequations}
where we define
\begin{equation}\label{equation:ddpm_alpha_t_ddpm}
\alpha_t \triangleq \prod_{\tau=1}^{t} (1 - \beta_\tau),
\end{equation}
and the implication~\eqref{equation:diffusion_kernel_zt_form_ddpm} follows from \eqref{equation:gauss_stand};
see Problem~\ref{prob:ddpm_trans_kernel}.
This also implies 
\begin{equation}\label{equation:diffusion_kernel_zt_form_x0_ddpm}
\bx_0 = \frac{\bz_t - \sqrt{1-\alpha_t}\bepsilon_t}{\sqrt{\alpha_t}}, 
\quad 
\bepsilon_t \sim \normal(\bzero, \bI_D).
\end{equation}
Note that $\bepsilon_t$ now represents the total noise added to the original image, rather than the incremental noise added at the current step of the Markov chain.
An illustration of this diffusion kernel is shown in Figure~\ref{fig:DDPM_fast_ff}.


We observe that each intermediate distribution takes a simple closed-form Gaussian form that can be directly sampled. This property is highly beneficial for training  DPMs or DDPMs, as it enables efficient stochastic gradient descent using randomly selected intermediate terms in the Markov chain without evaluating the entire chain sequentially.

After sufficiently many steps, the raw data becomes indistinguishable from Gaussian noise. 
In the limit $T \to \infty$, we have
\begin{equation}\label{equation:limit_zT_ddpm}
q(\bz_T \mid \bx_0) = \normal(\bz_T \mid \bzero, \bI_D)
\quad\implies\quad 
q(\bz_T) = \normal(\bz_T \mid \bzero, \bI_D).
\end{equation}
at which point all information about the original data is lost. 
The resulting marginal distribution for $\bz_T$ follows directly from the fact that the right-hand side expression is independent of $\bx_0$.
The choice of coefficients $\sqrt{1 - \beta_t}$ and $\sqrt{\beta_t}$ introduced earlier  (e.g., in~\eqref{equation:forward_step_zt_ddpm}) guarantees that once the Markov chain converges to a zero-mean, unit-covariance distribution, further updates maintain this stationary distribution.

\subsection*{Reverse Process: Data-Conditioned Distribution}

Our objective is to learn to reverse the noise-adding process, so it is natural to examine the inverse of the conditional distribution $q(\bz_t \mid \bz_{t-1})$, which can be derived via Bayes' theorem~\eqref{equation:bayes_base} as
\begin{equation}\label{equation:ddpm_true_reverse}
q(\bz_{t-1} \mid \bz_t) = \frac{q(\bz_t \mid \bz_{t-1}) q(\bz_{t-1})}{q(\bz_t)}.
\end{equation}
The marginal distribution $q(\bz_{t-1})$ can be expressed as
\begin{equation}\label{equation:marginal_zt_minus_1}
q(\bz_{t-1}) = \int q(\bz_{t-1} \mid \bx_0) p_{\text{data}}(\bx_0) \diff\bx_0,
\end{equation}
where $q(\bz_{t-1} \mid \bx_0)$ corresponds to the conditional Gaussian defined in~\eqref{equation:diffusion_kernel_ddpm}. 
However, this distribution is \textbf{intractable}, as it requires integration over the unknown data density $p_{\text{data}}(\bx_0)$.
Approximating this integral using training dataset samples results in a complex distribution represented as a Gaussian mixture.

Instead, we focus on the \textbf{data-conditioned reverse distribution} $q(\bz_{t-1} \mid \bz_t, \bx_0)$ (see also Figure~\ref{fig:diffusion_graphical_model}), which we will show shortly follows a simple Gaussian form. 
Intuitively, this makes sense: while it is challenging to infer a less-noisy latent state from a noisy observation alone, the problem becomes far more manageable when the original clean data $\bx_0$ is also known. 
This form will turn out to be useful when we rewrite the ELBO in Section~\ref{section:diff_elbo}.
This conditional distribution can be derived using Bayes' theorem~\eqref{equation:bayes_base}:
\begin{equation}\label{equation:bayes_conditional_xtminus}
q(\bz_{t-1} \mid \bz_t, \bx_0) 
= \frac{q(\bz_t \mid \bz_{t-1}, \bx_0) q(\bz_{t-1} \mid \bx_0)}{q(\bz_t \mid \bx_0)}
= \frac{q(\bz_t \mid \bz_{t-1}) q(\bz_{t-1} \mid \bx_0)}{q(\bz_t \mid \bx_0)},
\end{equation}
where the second equality follows from the Markov property of the forward diffusion process.
The quantity $q(\bz_t \mid \bz_{t-1})$  is given by~\eqref{equation:q_zt_given_zt_minus_1_ddpm};
as a function of $\bz_{t-1}$, this takes the form of an exponential of a quadratic form. 
Similarly, the quantity $q(\bz_{t-1} \mid \bx_0)$ in the numerator of~\eqref{equation:bayes_conditional_xtminus} is the diffusion kernel from~\eqref{equation:diffusion_kernel_ddpm}, which also involves an exponential quadratic in $\bz_{t-1}$. 
The denominator of~\eqref{equation:bayes_conditional_xtminus} can be disregarded since it is constant with respect to $\bz_{t-1}$. 
Consequently, the right-hand side of~\eqref{equation:bayes_conditional_xtminus} is a Gaussian distribution, whose mean and covariance can be identified through the method of \textit{matching Gaussian sufficient statistics}, yielding
\begin{equation}\label{equation:ddpm_reverse_conditional_gaussian}
q(\bz_{t-1} \mid \bz_t, \bx_0) = \normal\left( \bz_{t-1} \mid \bmu^q_t(\bx_0, \bz_t),\, \bSigma^q_t \right),
\end{equation}
with
\begin{align}
\bmu^q_t(\bx_0, \bz_t) 
&= \frac{(1 - \alpha_{t-1}) \sqrt{1 - \beta_t} \bz_t + \sqrt{\alpha_{t-1}} \beta_t \bx_0}{1 - \alpha_t}, \label{eq:mean_reverse} \\
\bSigma^q_t &= (\sigma^q_t)^2\bI_D, 
\qquad \qquad (\sigma^q_t)^2 = \frac{\beta_t (1 - \alpha_{t-1})}{1 - \alpha_t}, \label{eq:variance_reverse}
\end{align}
where we have used the definition from~\eqref{equation:ddpm_alpha_t_ddpm}.

\subsection*{Reverse Process: Neural Network Approximation}\label{section:dpm_approx}

We have established that the forward encoder is defined by a sequence of Gaussian conditional distributions $q(\bz_t \mid \bz_{t-1})$.
As illustrated in Figure~\ref{fig:diffusion_graphical_model}, the reverse process represents a right-to-left decoding operation.
Starting from $\bz_T$ (which corresponds to pure random Gaussian noise), the model progressively reconstructs meaningful structured data such as images by iteratively denoising the latent variables. 
At each timestep $t$, he objective is to estimate $\bz_{t-1}$ from the noisy latent state $\bz_t$. 
The conditional distribution governing this backward transition is denoted  $q(\bz_{t-1} \mid \bz_t)$.
Under this reverse process, we factorize the joint distribution $p(\bx_0, \bz_{1:T})$ as follows:
\begin{equation}\label{equation:rev_joint_ddpm}
p(\bx_0, \bz_{1:T}) = p(\bz_T) \left\{\prod_{t=T}^{2} q(\bz_{t-1} \mid \bz_t)\right\}  q(\bx_0 \mid \bz_1).
\end{equation}
The density of $p(\bz_T)$ is known analytically and follows a standard multivariate Gaussian distribution, i.e., $p(\bz_T) \sim \normal(\bzero, \bI_D)$. 
In contrast, the reverse conditional $q(\bz_{t-1} \mid \bz_t)$ is \textbf{intractable} to compute directly. Even when derived via Bayes' theorem, its normalization constant involves integrals without closed-form expressions, making the exact distribution computationally infeasible to evaluate.
Furthermore, it would require integration over all possible values of the initial vector $\bx_0$, whose distribution corresponds to the unknown data distribution  $p_{\text{data}}(\bx_0)$ that we aim to model. 

\paragrapharrow{Neural network approximation.}
Instead, we learn an approximation to the reverse distribution using a parameterized distribution $p_\btheta(\bz_{t-1} \mid \bz_t)$ implemented by a deep neural network, where $\btheta$ denotes the network's weights and biases: 
\begin{equation}
q(\bz_{t-1} \mid \bz_t)
\approx
p_{\btheta}(\bz_{t-1} \mid \bz_t).
\end{equation}
This reverse step acts analogously to the decoder in a variational autoencoder and is visualized in Figure~\ref{fig:diffusion_graphical_model}.

\begin{figure}[h]
\centering  
\subfigtopskip=2pt 
\subfigbottomskip=9pt 
\subfigcapskip=-5pt 
\includegraphics[width=0.95\textwidth]{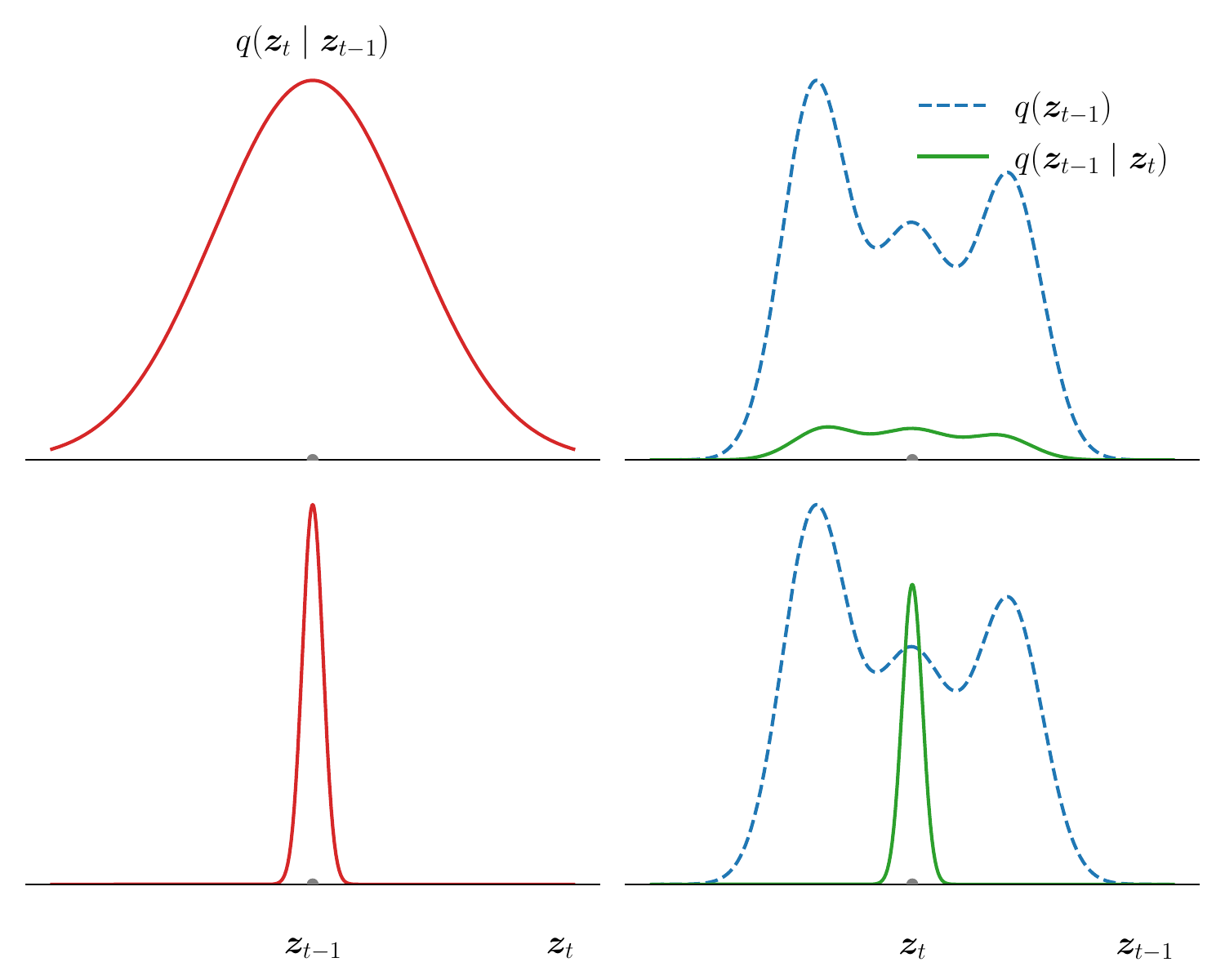}
\caption{
Illustration of the reverse distribution $q(\bz_{t-1}\mid\bz_t)$ derived via Bayes' theorem~\eqref{equation:bayes_conditional_xtminus} for scalar random variables.
\textbf{Upper.} The blue curve on the right shows the marginal distribution $q(\bz_{t-1})$ modeled as a mixture of three Gaussians, while the left panel displays the forward Gaussian noise process $q(\bz_t\mid\bz_{t-1})$ as a distribution over $\bz_t$ centered at $\bz_{t-1}$. 
Their product, after normalization, yields the distribution $q(\bz_{t-1}\mid\bz_t)$, plotted in green for a fixed value of $\bz_t$.
Since the left-hand distribution is relatively broad, corresponding to a large variance $\beta_t$, the resulting reverse distribution exhibits a complex multimodal shape.
\textbf{Bottom.} The forward distribution  $q(\bz_t \mid \bz_{t-1})$  in the left panel has a much smaller variance $\beta_t$. 
The corresponding reverse distribution $q(\bz_{t-1} \mid \bz_t)$, shown in green on the right, is nearly Gaussian with a variance similar to that of $q(\bz_t \mid \bz_{t-1})$.
Adapted from \citet{bishop2023deep}.}
\label{fig:ddpm_gauss_rev}
\end{figure}

\paragrapharrow{Form of the approximation.}
Intuitively, setting small step variances in the forward process~\eqref{equation:forward_step_zt_ddpm_ALL} such that $\beta_t \ll 1$ , ensures that changes in the latent vector between consecutive steps remain modest, which in turn simplifies learning the inverse transformation.  
\textbf{More precisely, if $\beta_t \ll 1$, then the true reverse distribution $q(\bz_{t-1} \mid \bz_t)$ becomes approximately Gaussian in $\bz_{t-1}$.}
This behavior follows from~\eqref{equation:ddpm_true_reverse}, since the right-hand side depends on $\bz_{t-1}$ through $q(\bz_t \mid \bz_{t-1})$ and $q(\bz_{t-1})$. 
If the forward transition distribution $q(\bz_t \mid \bz_{t-1})$ is a sufficiently narrow Gaussian (i.e., $\beta_t \ll 1$), then the marginal distribution $q(\bz_{t-1})$ carries substantial probability mass only over regions where $q(\bz_t \mid \bz_{t-1})$ is also concentrated, 
 implying that $q(\bz_{t-1} \mid \bz_t)$ is likewise approximately Gaussian.
 This heuristic is supported by a simple illustrative example shown in Figures~\ref{fig:ddpm_gauss_rev}.
 
However, because the per-step variances are small, a large number of steps is required to ensure that the latent distribution $\bz_T$ produced by the forward noising process remains close to a standard Gaussian. This increases the computational cost of generating new samples; in practice, $T$ often ranges in the thousands.

The above discussion motivates approximating the true reverse process $q(\bz_{t-1} \mid \bz_t)$ by a Gaussian distribution.
We can formally verify that the reverse distribution $q(\bz_{t-1} \mid \bz_t)$ is approximately Gaussian by performing a Taylor series expansion of $\ln q(\bz_{t-1} \mid \bz_t)$ around  $\bz_t$ with respect to $\bz_{t-1}$. 
This expansion also demonstrates that for small step variances, the covariance of the reverse distribution $q(\bz_{t-1} \mid \bz_t)$ is close to the covariance $\beta_t \bI_D$ of the forward noise process $q(\bz_t \mid \bz_{t-1})$. 

\paragrapharrow{The final Gaussian approximation.}
Accordingly, we model the \textit{reverse process} (or \textit{reverse diffusion}) using a Gaussian distribution of the form
\begin{equation}\label{equation:ddpm_reverse_model_gaussian}
\begin{aligned}
p_\btheta(\bz_{t-1} \mid \bz_t) 
&= \normal\left( \bz_{t-1} \mid \bmu^\btheta_t(\bz_t),\, (\sigma^\btheta_t)^2 \bI_D \right), \quad t= 2,3,\ldots,T;\\
p_\btheta(\bx_0 \mid \bz_1) 
&= \normal\left( \bx_0 \mid \bmu^\btheta_1(\bz_1),\, (\sigma^\btheta_1)^2 \bI_D \right),
\end{aligned}
\end{equation}
where {$\bmu^\btheta_t(\bz_t)$} is a deep neural network parameterized by  $\btheta$, and $(\sigma^\btheta_t)^2$ is either fixed (for instance, $(\sigma^\btheta_t)^2=\beta_t$ or $(\sigma^\btheta_t)^2=(\sigma^q_t)^2$ \citep{ho2020denoising}) or learned end-to-end \citep{nichol2021improved}. 
The network explicitly takes the timestep $t$ as input to adapt to the varying values of $\beta_t$ across different stages of the Markov chain. 
Similar to the amortization inference for VAEs (see Figure~\ref{fig:lvm_and_hyp_vb}), this design allows a single network to invert all steps of the chain, rather than requiring a separate model for each step. 

Considerable flexibility exists in choosing the neural network architecture for $\bmu^\btheta_t(\bz_t)$, as long as its output matches the input dimensionality. 
Under this constraint, \textit{{U-Net}} or a \textit{diffusion transformer (DiT)} architectures are standard choices for image generation tasks \citep{ronneberger2015u, peebles2023scalable}. See Chapter~\ref{chapter:diffarchitect} for more details.

The complete reverse process approximating \eqref{equation:rev_joint_ddpm} then forms a Markov chain expressed as
\begin{equation}\label{equation:ddpm_reverse_chain}
p(\bx_0, \bz_{1:T})
\approx
p_\btheta(\bx_0, \bz_{1:T} ) = p(\bz_T) \left\{ \prod_{t=2}^{T} p_\btheta(\bz_{t-1} \mid \bz_t) \right\} p_\btheta(\bx_0 \mid \bz_1),
\end{equation}
where $p(\bz_T)$  is given by $\normal(\bz_T \mid \bzero, \bI_D)$. 
After training, we can sample from the simple Gaussian prior over $\bz_T$ and map it to a sample from the data distribution $p_{\text{data}}(\bx_0)$ (approximated as the marginal distribution $p_{\btheta}(\bx_0)$) through a sequence of reverse steps obtained by repeatedly applying the trained network.
This again reflects the core paradigm of this book: transforming samples from an initial simple latent distribution (e.g., a Gaussian) into samples from the target data distribution; see Figure~\ref{fig:DDPM_gen_model_idea} for an illustration.

\subsection{The Objective Function (ELBO)}\label{section:diff_elbo}

We now turn to defining an objective function for training the neural network.
A standard approach for learning the unknown parameters of a probabilistic model is maximum likelihood estimation (see, for example, \eqref{equation:lvm_maxmarg}), which involves maximizing the log-probability of the observed data point $\bx_0$. 
In our case, the full joint distribution of the model is given $p_\btheta(\bx_0, \bz_{1:T})$.
However, we only observe realizations of  $\bx_0$, while the latent variables $\bz_{1:T}$ are unobserved. 
We therefore maximize the marginal distribution $p_\btheta(\bx_0)$ rather than the joint distribution $p_\btheta(\bx_0, \bz_{1:T})$.
The marginal likelihood can be obtained by integrating out the latent variables:
\begin{equation}\label{equation:likelihood_integral_ddpm}
p_\btheta(\bx_0) = \int p_\btheta(\bx_0, \bz_{1:T})   \diff\bz_{1:T}.
\end{equation}
where $p_\btheta(\bx_0, \bz_{1:T})$ is defined in~\eqref{equation:ddpm_reverse_chain}. 
This formulation corresponds to the general latent variable model (LVM) \eqref{equation:lvm_maxmarg}, where the latent variables are given by $\mathcalZ = \{\bz_1, \bz_2, \ldots, \bz_T\}$ and the observed variable is $\bx_0$. 
Notably, all latent variables share the same dimensionality as the data space, a property that does not hold for VAEs or GANs, as discussed in Chapter~\ref{chapter:vae_gan}.

\subsection*{Deriving the ELBO for DPMs}
From~\eqref{equation:likelihood_integral_ddpm}, we observe that the likelihood requires integration over all possible trajectories that could map Gaussian noise to the observed data point. These integrals are \textbf{intractable}, as they involve integration across the highly nonlinear and complex mappings induced by the neural network.
We therefore use an approach analogous to that employed for VAEs (and hierarchical VAEs), and instead maximize a lower bound on the log-likelihood known as the \textit{evidence lower-bound (ELBO)}, introduced in Section~\ref{section:elbo_cfe}. We rederive this bound specifically for diffusion models below. For \textbf{any} valid distribution $q(\mathcalZ)$, the following identity---referred to as the ELBO decomposition---always holds (see equations~\eqref{equation:elbo_vfe_negv2}, \eqref{equation:elbo_kl_vae}, or Figure~\ref{fig:ELBO_decom}):
\begin{equation}\label{equation:elbo_decom_ddpm}
\underbrace{\ln p_\btheta(\bx_0)}_{\text{Evidence}} 
=  \underbrace{\int q(\mathcalZ) \ln \left\{ \frac{p_\btheta(\bx_0, \mathcalZ  )}{q(\mathcalZ)} \right\} \diff\mathcalZ}_{\triangleq\mathcalF(\btheta), \text{ Evidence}} 
+ \underbrace{\KL[q(\mathcalZ) \parallel p_\btheta(\mathcalZ \mid \bx_0)]}_{\text{KL divergence}},
\end{equation}
where $\mathcalF$ denotes the ELBO, 
and $\KL[P \parallel Q]$ denotes the {KL divergence}  between distributions $P$ and $Q$; see \eqref{equation:kl_def_vae}.
Since the KL divergence is nonnegative, we have 
\begin{equation}\label{equation:elbo_inequality_ddpm}
\ln p_\btheta(\bx_0) \geq \mathcalF(\btheta).
\end{equation}
Given that the exact log-likelihood is intractable, we train the neural network by maximizing this lower bound $\mathcalF(\btheta)$.

To proceed, we first derive an explicit expression for the diffusion model ELBO. 
Note that the inequality \eqref{equation:elbo_inequality_ddpm} holds for any valid distribution $q(\mathcalZ)$.
In many ELBO applications, such as the VAE, $q(\mathcalZ)$ is equipped with \textbf{learnable} parameters $\blambda$ ($q(\mathcalZ) =q_{\blambda}(\mathcalZ\mid\bx_0)$, the encoder of VAEs), typically implemented via a deep neural network, and the ELBO is maximized with respect to both these parameters and those of the joint distribution $p_\btheta(\bx_0, \mathcalZ  )$.
Optimizing $q(\mathcalZ)$ tightens the bound, bringing the parameter optimization for $p_\btheta(\bx_0, \mathcalZ  )$ closer to exact maximum likelihood estimation; see Section~\ref{section:vae_pca}.

However, for diffusion models, we set  $q(\mathcalZ)$ to the \textbf{fixed} distribution $q(\mathcalZ)=q(\bz_1, \ldots, \bz_T \mid \bx_0)$ given by the forward Markov chain in~\eqref{equation:ddpm_joint_latent_dist_ddpm}. 
Consequently, the only learnable parameters are those defining the reverse Markov chain model $p_\btheta(\bx_0, \bz_{1:T})$. 

We now substitute the forward diffusion process $q(\bz_1,\ldots,\bz_T \mid \bx_0)$ from~\eqref{equation:ddpm_joint_latent_dist_ddpm} and the reverse process $p_\btheta(\bx_0, \bz_{1:T})$ from~\eqref{equation:ddpm_reverse_chain} into~\eqref{equation:elbo_decom_ddpm}. This allows us to expand the ELBO as
\begin{mybox}
\begin{align}
\mathcalF(\btheta) &= \Exp_q \left[ \ln \left\{ \frac{p(\bz_T) \left\{ \prod_{t=2}^{T} p_\btheta(\bz_{t-1} \mid \bz_t) \right\} p_\btheta(\bx_0 \mid \bz_1)}{q(\bz_1 \mid \bx_0) \prod_{t=2}^{T} q(\bz_t \mid \bz_{t-1})} \right\} \right] \nonumber \\
&= \Exp_q \left[ \ln p(\bz_T) + \ln p_\btheta(\bx_0 \mid \bz_1) 
+ \sum_{t=2}^{T} \ln \left( \frac{p_\btheta(\bz_{t-1} \mid \bz_t)}{q(\bz_t \mid \bz_{t-1})} \right) - \ln q(\bz_1 \mid \bx_0)  \right],
\label{equation:elbo_expanded_ddpm}
\end{align}
\end{mybox}
with the expectation operator defined as
\begin{equation}\label{equation:expectation_q_ddpm}
\begin{aligned}
\Exp_q[\,\cdot\,] \triangleq 
&\int  q(\bz_1 \mid \bx_0) \prod_{t=2}^{T} q(\bz_t \mid \bz_{t-1}) \, [\,\cdot\,]\diff\bz_{1:T}
\equiv \int  \prod_{t=1}^{T} q(\bz_t \mid \bx_0) \, [\,\cdot\,] \diff\bz_{1:T}.
\end{aligned}
\end{equation}

The first term $\ln p(\bz_T)$ on the right-hand side of~\eqref{equation:elbo_expanded_ddpm} corresponds to the fixed standard normal distribution  $\normal(\bz_T \mid \bzero, \bI_D)$. Since it contains no trainable parameters, it acts as a constant offset and can be dropped from the ELBO without affecting optimization.
Similarly, the fourth term $-\ln q(\bz_1 \mid \bx_0)$ is independent of $\btheta$ and may also be omitted.
The second term in~\eqref{equation:elbo_expanded_ddpm} plays the same role as the reconstruction loss in VAEs.
Its expectation can be approximated using Monte Carlo sampling from the distribution of $\bz_1$ given by~\eqref{equation:diffusion_kernel_ddpm}, yielding 
\begin{equation}\label{equation:ddpm_fourth_mc_reconstruction}
\Exp_q \left[ \ln p_\btheta(\bx_0 \mid \bz_1) \right] 
\simeq 
\frac{1}{S} \sum_{s=1}^{S} \ln p_\btheta(\bx_0 \mid \bz_1^{(s)}),
\end{equation}
where $\bz_1^{(s)} \sim \normal(\bz_1 \mid \sqrt{1 - \beta_1} \bx_0, \beta_1 \bI_D)$. 
In contrast to VAEs, we do not need to backpropagate gradients through the sampled values, because the forward distribution $q$ is fixed and the reparameterization trick (see Section~\ref{section:other_issue_vae}) is therefore unnecessary.

The third term on the right-hand side of~\eqref{equation:elbo_expanded_ddpm} consists of a sum over terms involving consecutive latent variables $\bz_{t-1}$ and $\bz_t$. 
Recall from our earlier derivation of the diffusion kernel~\eqref{equation:diffusion_kernel_ddpm} that we can directly sample $q(\bz_{t-1} \mid \bx_0)$ as a Gaussian, and then obtain a corresponding sample of $\bz_t$ via~\eqref{equation:forward_step_zt_ddpm} or~\eqref{equation:diffusion_kernel_zt_form_ddpm}, which is also Gaussian. While this approach is asymptotically correct, sampling pairs of latent variables produces \textbf{high-variance} Monte Carlo estimates, requiring an impractically large number of samples to reduce noise.
Equivalently, this third term can be expressed as an expectation of a KL divergence over two successive random variables:
$$
\Exp_{q(\bz_{t+1} \mid \bz_{t}) q(\bz_{t-1} \mid \bz_{t-2})} \left[ \KL \left( q(\bz_t \mid \bz_{t-1}) \parallel p_\btheta(\bz_t \mid \bz_{t+1}) \right) \right].
$$
Sampling both variables simultaneously again introduces high variance, destabilizing optimization and slowing convergence.
For these reasons, directly optimizing~\eqref{equation:elbo_expanded_ddpm} is not ideal.
Instead, we reformulate the ELBO into a form that can be reliably estimated by sampling only one variable per term.

\subsection*{Rewriting the ELBO via Bayes' theorem}

Building on our earlier discussion of the ELBO, our objective in this section is to reformulate the ELBO using KL divergences, which will subsequently allow us to express it in closed form.
To see this, we use Bayes' theorem~\eqref{equation:bayes_base} to invert the conditional distribution $q(\bz_t \mid \bz_{t-1})$, yielding
\begin{equation}
{q(\bz_t \mid \bz_{t-1}) =} 
q(\bz_t \mid \bz_{t-1}, \bx_0) = \frac{q(\bz_{t-1} \mid \bz_t, \bx_0)  q(\bz_t \mid \bx_0)}{q(\bz_{t-1} \mid \bx_0)}.
\end{equation}
Note that although the forward distribution $q(\bz_t \mid \bz_{t-1})$ is independent of $\bx_0$, the data-conditioned reverse distribution $q(\bz_{t-1} \mid \bz_t, \bx_0)$ explicitly depends on $\bx_0$.
This decomposition enables us to rewrite the third term of the ELBO from \eqref{equation:elbo_expanded_ddpm} as
\begin{equation}\label{equation:ddpm_log_ratio_decomposition}
\ln \left( \frac{p_\btheta(\bz_{t-1} \mid \bz_t)}{q(\bz_t \mid \bz_{t-1})} \right) 
= \ln \left( \frac{p_\btheta(\bz_{t-1} \mid \bz_t)}{q(\bz_{t-1} \mid \bz_t, \bx_0)} \right) 
+ \ln \left( \frac{q(\bz_{t-1} \mid \bx_0)}{q(\bz_t \mid \bx_0)} \right).
\end{equation}
The second term on the right-hand side of~\eqref{equation:ddpm_log_ratio_decomposition} does not depend on $\btheta$ and can thus be discarded.
Substituting~\eqref{equation:ddpm_log_ratio_decomposition} into~\eqref{equation:elbo_expanded_ddpm}, we obtain
\begin{mybox}
\begin{align}
\mathcalF(\btheta) 
&= \Exp_q \left[ 
\ln p_\btheta(\bx_0 \mid \bz_1) +
\sum_{t=2}^{T} \ln \left( \frac{p_\btheta(\bz_{t-1} \mid \bz_t)}{q(\bz_{t-1} \mid \bz_t, \bx_0)} \right) + \ln \frac{p(\bz_T)}{q(\bz_t\mid\bx_0)} \right] +\text{const}
\nonumber\\
&=
\underbrace{\Exp_{q(\bz_1 \mid \bx_0)} \left[ \ln p_\btheta(\bx_0 \mid \bz_1) \right]}_{\text{Reconstruction Term}} 
- \underbrace{\KL \left[q(\bz_T \mid \bx_0) \parallel p(\bz_T) \right]}_{\text{Prior Matching Term}} 
\nonumber \\
&\quad -  \underbrace{\sum_{t=2}^{T} \Exp_{q(\bz_t \mid \bx_0)} \big[ \KL\big(q(\bz_{t-1} \mid \bz_t, \bx_0) \parallel p_\btheta(\bz_{t-1} \mid \bz_t)\big)\big]    \diff\bz_t}_{\text{Consistency Terms}}
+ \text{const},  
\label{equation:ddom_elbo_final}
\end{align}
\end{mybox}
where we have simplified the expectation over $q(\bz_1, \ldots, \bz_T \mid \bx_0)$ in the first term of \eqref{equation:ddom_elbo_final}, as   $\bz_1$ is the only latent variable present in the integrand. 
Under the expectation defined in \eqref{equation:expectation_q_ddpm}, all conditional distributions integrate to one, leaving only the integral over  $\bz_1$. 
Similarly, in the third term of \eqref{equation:ddom_elbo_final}, each integral involves only consecutive latent variables $\bz_{t-1}$ and $\bz_t$, and all other variables can be marginalized out.

\paragrapharrow{Prior matching term.} The second term is the \textit{prior matching term}, which takes the form of a KL divergence and corresponds to the regularization term in the VAE objective \eqref{equation:vae_elbo_vae}.  
The key distinction is that $q(\bz_T \mid \bx_0)$ contains no learnable parameters: it represents the forward process, which we assume to be a known linear Gaussian transformation and thus requires no model training. When $T$ is sufficiently large, this term approaches zero. Moreover, as mentioned previously,  since it is independent of $\btheta$, it can be safely omitted from the optimization objective.

\paragrapharrow{Reconstruction term.} The term $\Exp_{q(\bz_1 \mid \bx_0)} \left[ \ln p_\btheta(\bx_0 \mid \bz_1) \right]$ serves as the \textit{reconstruction term}, analogous to its counterpart in the VAE objective \eqref{equation:vae_elbo_vae}. 
It reconstructs the original data $\bx_0$ from the first-step latent variable $\bz_1$. 
This reconstruction term assigns high probability to the observed data sample and can be trained using the same Monte Carlo strategy as the corresponding term in the VAE, via the sampling approximation in~\eqref{equation:ddpm_fourth_mc_reconstruction}.
Expanding $\ln p_\btheta(\bx_0 \mid \bz_1)$ in the form \eqref{equation:ddpm_reverse_model_gaussian} using the Gaussian probability density function (Definition~\ref{definition:multivariate_gaussian}):
\begin{equation}\label{equation:ddpm_reconz1_logprob}
\begin{aligned}
\ln p_\btheta(\bx_0 \mid \bz_1) 
&= \ln \frac{(2\pi)^{-D/2} }{\abs{(\sigma_1^\btheta)^2\bI_D}^{1/2}} \exp\left\{ -\frac{1}{2}   (\bx_0 - \bmu^\btheta_1(\bz_1))^\top (\sigma_1^\btheta\bI_D)^{-2} (\bx_0 - \bmu^\btheta_1(\bz_1)) \right\} \\
&= -\frac{1}{2(\sigma_1^\btheta)^2} \normtwo{\bx_0 - \bmu^\btheta_1(\bz_1)}^2 +\text{const},
\end{aligned}
\end{equation}
where all additive terms independent of the network parameters $\btheta$ have been absorbed into a constant term that has no effect on training.
From this expression, maximizing $\ln p(\bx_0 \mid \bz_1)$ is equivalent to minimizing the mean squared error between the model output $\bmu^\btheta_1(\bz_1)$ and $\bx_0$. 
Since the model output essentially predicts $\bx_0$, we denote $\bmu^\btheta_1(\bz_1)$ as $\widehatbx^\btheta_1(\bz_1)$:
\begin{equation}\label{equation:ddpm_recon_loss1}
\begin{aligned}
\argmax_\btheta \Exp_{q(\bz_1 \mid \bx_0)} \left[ \ln p_\btheta(\bx_0 \mid \bz_1) \right] 
&= \argmin_\btheta \Exp_{q(\bz_1 \mid \bx_0)} 
\left[\frac{1}{2(\sigma^\btheta_t)^2} \normtwo{\bx_0 - \widehatbx^\btheta_1(\bz_1)}^2 \right].
\end{aligned}
\end{equation}
As shown, this term corresponds to training a model to predict the sample $\bx_0$, with an effective loss function equivalent to mean squared error. The expectation $\Exp_{q(\bz_1 \mid \bx_0)}$ is necessary because the model input $\bz_1$ is a random variable rather than a deterministic value, requiring averaging over its distribution. The distribution of $\bz_1$ is defined in \eqref{equation:forward_step_z1_ddpm_v2}. Following the VAE framework, we approximate this expectation using sampling methods similar to \eqref{equation:ddpm_fourth_mc_reconstruction}.
{For now, we may set aside the detailed implementation of this term. As will become clear in subsequent derivations, this term can ultimately be combined with the third term in the final objective expression.}

\paragrapharrow{Consistency terms.}
The difference lies in the \textit{consistency terms} in~\eqref{equation:ddom_elbo_final}, which are defined between pairs of Gaussian distributions and can therefore be expressed in closed form.
The distribution $q(\bz_{t-1} \mid \bz_t, \bx_0)$ is given by~\eqref{equation:ddpm_reverse_conditional_gaussian}, while the distribution $p_\btheta(\bz_{t-1} \mid \bz_t)$ is defined in~\eqref{equation:ddpm_reverse_model_gaussian}. 
The KL divergence can thus be derived using the closed-form expression for the KL divergence between two Gaussian distributions (see Problems~\ref{problem:kl_gauss1}--\ref{problem:kl_gauss2}):
\begin{align}\label{equation:ddpm_kl_gaussian}
&\KL[q(\bz_{t-1} \mid \bz_t, \bx_0) \parallel p_\btheta(\bz_{t-1} \mid \bz_t)] 
= \frac{1}{2(\sigma^\btheta_t)^2} \normtwo{\bmu^q_t(\bx_0, \bz_t) - \bmu^\btheta_t(\bz_t)}^2 + \text{const},
\end{align}
where $\bmu^q_t(\bx_0, \bz_t)$ is defined in~\eqref{eq:mean_reverse}, and all additive terms independent of the network parameters \(\btheta\) have been absorbed into a constant term that has no influence on training.

As noted earlier, $(\sigma^\btheta_t)^2$ is either \textbf{fixed} (e.g., $(\sigma^\btheta_t)^2=\beta_t$ or $(\sigma^\btheta_t)^2=(\sigma^q_t)^2$ \citep{ho2020denoising}) or \textbf{learned} \citep{nichol2021improved}. When  $(\sigma^\btheta_t)^2=(\sigma^q_t)^2$, the constant term reduces to zero.
Each consistency term in~\eqref{equation:ddom_elbo_final} retains an integral over $\bz_t$, weighted by $q(\bz_t \mid \bx_0)$. 
Once again, this integral can be approximated by sampling from $q(\bz_t \mid \bx_0)$, which can be performed efficiently using the diffusion kernel~\eqref{equation:diffusion_kernel_ddpm}.

We observe that the KL divergence in~\eqref{equation:ddpm_kl_gaussian} takes the form of a simple squared loss function. 
Since we optimize the network parameters to maximize the lower bound in~\eqref{equation:ddom_elbo_final}, this is equivalent to minimizing the squared error, owing to the negative sign preceding the KL divergence terms in the ELBO.

The quantity $\bmu^q_t(\bx_0, \bz_t)$ is defined in \eqref{eq:mean_reverse} as a function of $\bx_0$ and $\bz_t$:
\begin{equation}\label{equation:mean_reverse_revis}
\bmu^q_t(\bx_0, \bz_t) = \frac{(1-\alpha_{t-1})\sqrt{1-\beta_t}\bz_t + \sqrt{\alpha_{t-1}}\beta_t\bx_0}{1-\alpha_t}.
\end{equation}
In contrast, $\bmu^\btheta_t(\bz_t)$  is a function solely  of $\bz_t$. In other words, our parameterized model takes $\bz_t$ as input and outputs $\bmu^\btheta_t$. 
Our objective is to make $\bmu^\btheta_t$ as close as possible to $\bmu^q_t$. 
To this end, we can construct $\bmu^\btheta_t$ to share a similar functional form with $\bmu^q_t$:
\begin{equation}\label{equation:mean_reverse_hat}
\bmu^\btheta_t(\bz_t) \triangleq  \frac{(1-\alpha_{t-1})\sqrt{1-\beta_t}\bz_t + \sqrt{\alpha_{t-1}}\beta_t\widehatbx^\btheta_t(\bz_t)}{1-\alpha_t}
\end{equation}
Rather than directly predicting $\bmu^q_t(\bx_0, \bz_t)$, the model predicts $\widehatbx^\btheta_t(\bz_t)\approx \bx_0$ and then computes the corresponding estimate of $\bmu^q_t(\bx_0, \bz_t)$ using \eqref{equation:mean_reverse_hat}. 
With this definition, the mean squared loss $\normtwo{\bmu^q_t - \bmu^\btheta_t}^2$ simplifies to:
\begin{equation}\label{equation:loss_simplification_hatx_ddpm}
\begin{aligned}
\normtwo{\bmu^q_t(\bx_0, \bz_t) - \bmu^\btheta_t(\bz_t)}^2 
= \left( \frac{\sqrt{\alpha_{t-1}}\beta_t}{1-\alpha_t} \right)^2 \normtwo{\bx_0 - \widehatbx^\btheta_t(\bz_t)}^2.
\end{aligned}
\end{equation}
For each timestep $t$ with $2 \leq t \leq T$, our parameterized model takes $\bz_t$ and $t$ as inputs and predicts $\widehatbx^\btheta_t(\bz_t)$. 
Our goal is to minimize the mean squared error between this prediction and $\bx_0$. Maximizing the negative KL divergence (the third term) in the ELBO function is thus equivalent to minimizing the mean squared error between the model output and $\bx_0$.
\begin{equation}\label{equation:ddpm_kl_to_mse}
\begin{aligned}
&\argmax_\btheta \left[ -\sum_{t=2}^T \Exp_{q(\bz_t\mid\bx_0)} \big(  \KL[q(\bz_{t-1}\mid\bz_t,\bx_0) \parallel p_\btheta(\bz_{t-1}\mid\bz_t)] \big) \right] \\
&= \argmin_\btheta \left[ \sum_{t=2}^T \frac{1}{2(\sigma^\btheta_t)^2} \left(\frac{\sqrt{\alpha_{t-1}}\beta_t}{1-\alpha_t}\right)^2 \Exp_{q(\bz_t\mid\bx_0)} \left[ \normtwo{\widehatbx^\btheta_t(\bz_t) - \bx_0}^2 \right] \right].
\end{aligned}
\end{equation}

\paragrapharrow{The final objective function.}
Finally, by \textbf{omitting constant coefficients}, we combine \eqref{equation:ddpm_recon_loss1} and \eqref{equation:ddpm_kl_to_mse} to derive the final objective function. 
The result is remarkably simple: for any $t \in 1,2,\ldots,T$, the parameterized model takes $\bz_t$ and $t$ as inputs and outputs $\widehatbx^\btheta_t(\bz_t)$. 
The model is trained to minimize the mean squared error between its prediction $\widehatbx^\btheta_t(\bz_t)$ and the ground-truth value $\bx_0$. 
Accordingly, maximizing the ELBO in \eqref{equation:ddom_elbo_final} is approximately equivalent to solving
\begin{mybox}
\begin{align}
\small
\argmax_\btheta \mathcalF(\btheta) 
&\simeq \argmin_\btheta \left\{\Exp_{q(\bz_1\mid\bx_0)} \left[ \normtwo{\bx_0 - \widehatbx^\btheta_1(\bz_1)}^2 \right] + \left[ \sum_{t=2}^T \Exp_{q(\bz_t\mid\bx_0)} \left[ \normtwo{\widehatbx^\btheta_t(\bz_t) - \bx_0}^2 \right] \right]\right\} 
\nonumber \\
&= \argmin_\btheta \sum_{t=1}^T \Exp_{q(\bz_t\mid\bx_0)} \left[\normtwo{\widehatbx^\btheta_t(\bz_t) - \bx_0}^2 \right].
\label{equation:dpm_final_elbo}
\end{align}
\end{mybox}
As before, each term above contains an integral over $\bz_t$, weighted by $q(\bz_t \mid \bx_0)$.
This expectation can be efficiently approximated by sampling from $q(\bz_t \mid \bx_0)$ using the diffusion kernel~\eqref{equation:diffusion_kernel_ddpm}.
Compared to the original ELBO error in~\eqref{equation:elbo_expanded_ddpm}, we now only sample one variable per term, leading to Monte Carlo estimates with significantly lower variance.

When optimizing the model via stochastic gradient descent (SGD), we compute the gradient of the loss function with respect to the network parameters for a randomly sampled training data point $\bx_0$. 
Instead of evaluating the loss for every timestep $t$ in the summation from \eqref{equation:dpm_final_elbo}, we randomly sample a single timestep along the Markov chain for each training sample. Gradients are accumulated across mini-batches and subsequently applied to update the network weights.

Notably, this loss formulation inherently induces implicit data augmentation. For each training sample $\bx_0$, a unique noise realization $\bepsilon_t$ is sampled in every training iteration, generating a distinct noisy latent state $\bz_t$. The above optimization pipeline applies to a single training sample $\bx_0$, and the full gradient computation workflow is summarized in Algorithm~\ref{alg:ddpm_training}.

\begin{algorithm}
\caption{DPM Training Procedure \index{Denoising diffusion probabilistic models}\index{DDPMs}}
\label{alg:ddpm_training}
\begin{algorithmic}[1]
\Require Training data $\mathcalX=\{\bx_n\}$, noise schedule $\{\beta_t\}_t$, the total number of steps $T$ (standard choice: $T = 1,000$, $\beta_1 = 10^{-4}$, $\beta_T = 0.02$);
\State \textbf{initialize:} $\btheta$;
\For{$cnt=0,1,2,\ldots$}
\State $\bx_0\sim \mathcalX$; \Comment{(DPM$_1$)} 
\State $t\sim \uniformdist(\{1,2,\ldots,T\})$; \Comment{(DPM$_2$)} 
\State $\bepsilon\sim\normal(\bzero, \bI_D)$;  \Comment{(DPM$_3$)} 
\State $\bz_t\leftarrow \sqrt{\alpha_t}\bx_0+\sqrt{1-\alpha_t}\bepsilon$; \Comment{(DPM$_4$)} 
\State Update network by minimizing loss: $\mathcalJ(\btheta) \leftarrow \normtwo{\widehatbx^\btheta_t(\bz_t) - \bx_0}^2$; \Comment{(DPM$_5$)} 
\EndFor
\State \Return  Optimized network parameter $\btheta$;
\end{algorithmic}
\end{algorithm}

\subsection{Generating New Samples}\label{section:gen_dpm}

The trained neural network is denoted $\widehatbx^\btheta_t(\bz_t)$, where $t$ denotes the timestep and  $\btheta$ represents  the network parameter. 
It takes $\bz_t$ and $t$ as inputs and outputs a prediction of $\bx_0$. 
sing this model, we obtain an approximate form of the true conditional distribution $q(\bz_{t-1} \mid \bz_t)$ in the reverse process:
\begin{equation}\label{eq:reverse_process_approx}
q(\bz_{t-1} \mid \bz_t) 
\approx
p_\btheta(\bz_{t-1} \mid \bz_t) 
\approx
q(\bz_{t-1} \mid \bz_t, \bx_0) \sim \normal(\bz_{t-1}\mid \bmu^q_t(\bx_0, \bz_t), \bSigma^q_t),
\end{equation}
where the mean $\bmu^q_t(\bx_0, \bz_t)$ and variance $\bSigma^q_t$ are  given by \eqref{eq:mean_reverse} and \eqref{eq:variance_reverse}, respectively, with the true data $\bx_0$ replaced by the model prediction $\widehatbx^\btheta_t(\bz_t)$:
\begin{align}
\bmu^q_t(\bx_0, \bz_t) &= \frac{(1 - \alpha_{t-1})\sqrt{1-\beta_t}\bz_t + \sqrt{\alpha_{t-1}}\beta_t\widehatbx^\btheta_t(\bz_t)}{1 - \alpha_t}; \\
\bSigma^q_t &=  (\sigma^q_t)^2 \bI_D =\frac{\beta_t(1 - \alpha_{t-1})}{1 - \alpha_t} \bI_D.
\end{align}
The reverse generation (sampling) procedure can be summarized as follows:
\begin{enumerate}
\item Set $T$, e.g., $T = 1000$;
\item Sample $\bz_T$ from a standard Gaussian distribution.
\item Use the neural network model to compute $\widehatbx^\btheta_t(\bz_t)$.
\item Compute the mean $\bmu^q_t(\widehatbx^\btheta_t(\bz_t), \bz_t)$ and variance $\bSigma^q_t$ of $p_{\btheta}(\bz_{t-1} \mid \bz_t)$.
\item Sample $\bz_{t-1}$ randomly from $p_{\btheta}(\bz_{t-1} \mid \bz_t)$.
\item Repeat steps 3--5 until $t = 1$.
\end{enumerate}
Complete pseudocode is provided in Algorithm~\ref{alg:dpm_sampling}.
\begin{algorithm}[H]
\caption{Sampling from a DPM\index{Diffusion probabilistic models}}
\label{alg:dpm_sampling}
\begin{algorithmic}[1]
\Require Trained prediction network $\widehatbx^\btheta_t(\bz)$, noise schedule $\{\beta_t\}_t$;
\State Set the total number of steps $T$, e.g., $T = 1000$;
\State Sample  $\bz_T\sim \normal(\bzero, \bI_D)$;
\For{{$t=T,T-1,\ldots, 2$}}
\State Use the neural network model to compute $\widehatbx^\btheta_t(\bz_t)$; \Comment{(SDPM$_1$)} 
\State Compute the mean $\bmu^q_t(\widehatbx^\btheta_t(\bz_t), \bz_t)$ and variance $\bSigma^q_t$ of $p_{\btheta}(\bz_{t-1} \mid \bz_t)$; \Comment{(SDPM$_2$)} 
\State Sample $\bz_{t-1}$ randomly from $p_{\btheta}(\bz_{t-1} \mid \bz_t)$; \Comment{(SDPM$_3$)} 
\EndFor
\State Set the new sample $\bx_0 \leftarrow \widehatbx^\btheta_1(\bz_1)$;
\State \Return New sample $\bx_0$;
\end{algorithmic}
\end{algorithm}

At each timestep $t$, the model directly predicts the original data $\bx_0$. Intuitively, this design appears somewhat unnatural: when $t$ is large, $\bz_t$ is heavily corrupted and far from $\bx_0$. Even though $t$ is provided as an input to the model, it remains challenging for a single parameterization to robustly handle the full range of timesteps. Empirical results confirm this limitation: early diffusion probabilistic models exhibited relatively weak performance and low-quality image generation, as documented in comparative experiments \citep{ho2020denoising}.
Due to these drawbacks, the original DPM formulation proposed in 2015 gained little mainstream attention. 
This changed in 2020, when \citet{ho2020denoising} introduced denoising diffusion probabilistic models (DDPMs) as a significantly improved variant. 
Following this work, OpenAI and Stability AI released models such as DALL-E 2 and Stable Diffusion built upon DDPMs \citep{ramesh2022hierarchical, rombach2022high}, enabling breakthrough performance and bringing diffusion models widespread recognition. In Section~\ref{section:ddpm_key}, we examine the key innovations introduced by DDPMs.

\index{Binomial diffusion}
\subsection{Extension to Binary Data}

Given binary input data $x_0 \in \{0,1\}^D$ he Gaussian-kernel-based DPM can be readily adapted to binary data via the binomial kernel, a formulation known as \textit{binomial diffusion} \citep{sohl2015deep}. In this framework, the forward diffusion process gradually corrupts the original data to converge to a stationary $\bernoullidist(1/2)$ distribution.

\paragrapharrow{Forward kernel.}
At each diffusion step, every bit is stochastically flipped toward a uniform binary distribution governed by the noise coefficient $\beta_t$, following the transition kernel:
\begin{equation}
q(z_t \mid z_{t-1}) = \bernoullidist\left(z_t\mid (1 - \beta_t)z_{t-1} + \tfrac{1}{2}\beta_t\right).
\end{equation}
This definition implies that the Bernoulli sampling probability at timestep $t$ is a convex combination of the previous latent state and the uniform probability $1/2$.
Note that for a scalar $x \in \{0, 1\}$ and probability  parameter $p \in [0, 1]$, the Bernoulli distribution is formally defined as:
$$
\bernoulli(x\mid  p) 
= p^x (1-p)^{1-x} 
= 
\begin{cases}
p, & x = 1; \\
1-p, & x = 0.
\end{cases}
$$

\paragrapharrow{Marginal kernel.} By composing the sequential forward diffusion steps, we define the cumulative product parameter  $\alpha_t \triangleq \prod_{\tau=1}^{t} (1 - \beta_\tau)$.
The marginal distribution of the latent variable $z_t$ conditioned on the original data $x_0$ then takes the form:
\begin{equation}\label{equation:bernoulli_marg_kernel}
q(z_t \mid x_0) = \bernoullidist\left(z_t\mid \alpha_t x_0 + \tfrac{1}{2}(1-\alpha_t)\right).
\end{equation}
A hint of the derivation is provided in Problem~\ref{prob:bernoulli_marg_kernel}. 
As the diffusion step $t$ approaches the total step $T$, $\alpha_t \to 0$, such that the marginal distribution converges to $q(z_T) \to \bernoullidist(1/2)$, regardless of the original input $x_0$.

\paragrapharrow{Posterior (reverse-time conditional).}
Using Bayes' theorem~\ref{equation:bayes_base} for binary state spaces, the true data-dependent reverse conditional distribution $q(z_{t-1} \mid z_t, x_0)$ follows a Bernoulli distribution with the success probability formulated as:
\begin{equation}
q(z_{t-1} = 1 \mid z_t, x_0) = \frac{q(z_t \mid z_{t-1} = 1) q(z_{t-1} = 1 \mid x_0)}{\sum_{k \in \{0,1\}} q(z_t \mid z_{t-1} = k) q(z_{t-1} = k \mid x_0)}.
\end{equation}
All terms in the formulation are derived from the forward kernels defined above. Specifically, the forward transition probabilities are $q(z_t \mid z_{t-1} = 1) = (1 - \beta_t/2)^{z_t} (\beta_t/2)^{1-z_t}$ and  $q(z_t \mid z_{t-1} = 0) = (\beta_t/2)^{z_t} (1 - \beta_t/2)^{1-z_t}$, while the marginal probability of the previous latent state is  $q(z_{t-1} = 1 \mid x_0) = \alpha_{t-1}x_0 + \tfrac{1}{2}(1-\alpha_{t-1})$.
This analytically tractable posterior distribution serves as the optimization target for the learned reverse diffusion model.

\begin{figure}[h]
\centering  
\subfigtopskip=2pt 
\subfigbottomskip=9pt 
\subfigcapskip=-5pt 
\includegraphics[width=0.95\textwidth]{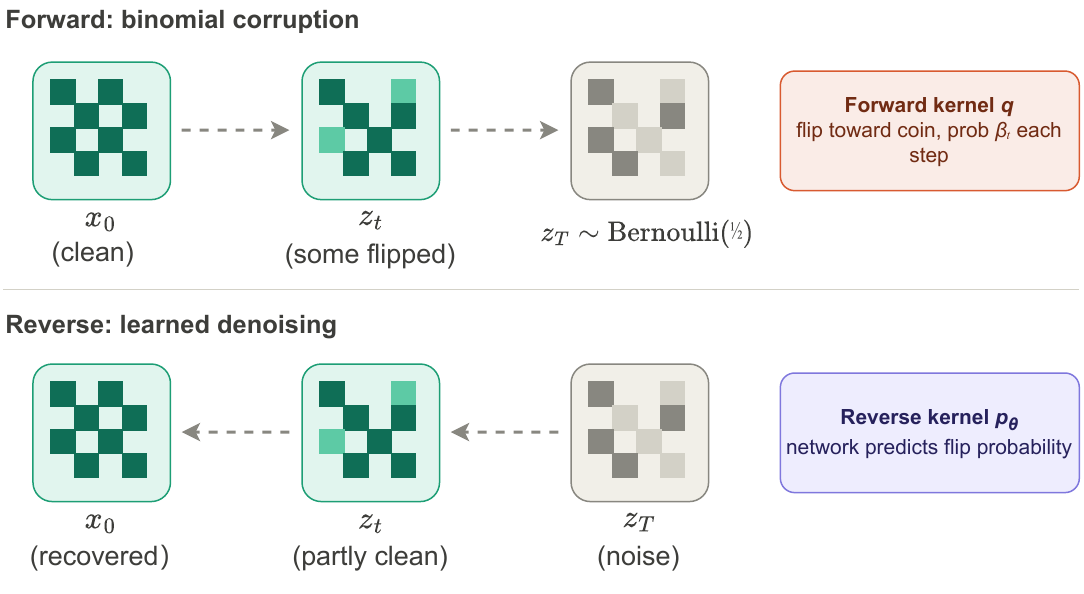}
\caption{
Illustration of the binomial diffusion model.
Green = pixel on (1), light = pixel off (0), gray = corrupted / uncertain.
Each forward step follows a Bernoulli mixture: $z_t = (1-\beta_t)z_{t-1}+\beta_t\cdot\text{coin flip}$.
The reverse chain is trained to reverse one binomial flip step at a time.}
\label{fig:ddpm_binomial}
\end{figure}

\paragrapharrow{Approximate reverse model and variational bound.}
We parameterize the learned reverse diffusion kernel as a Bernoulli distribution:  $p_\btheta(z_{t-1} \mid z_t) = \bernoullidist\left(z_{t-1}\mid p_t^\btheta(z_t)\right)$. Following the paradigm of Gaussian DPMs, the network predicts a reconstructed original data estimate $\widehatx_0 = p_t^\btheta(z_t) \in \textcolor{black}{[0,1]}$, which is substituted into the true posterior to define the approximate reverse kernel:
$$
p_\btheta(z_{t-1} \mid z_t) = q(z_{t-1} \mid z_t, x_0 = p_t^\btheta(z_t)).
$$
Adapting the variational lower bound (ELBO) framework for diffusion probabilistic models from \eqref{equation:ddom_elbo_final}, we derive the corresponding variational objective for the binomial diffusion model:
\begin{align}
\ln p_\btheta(x_0) 
&\geq 
\underbrace{\Exp_q \left[ \ln p_\btheta(x_0 \mid x_1) \right] }_{\mathcalJ_0}
- \underbrace{\Exp_q \left[ \sum_{t>1} \KL[q(z_{t-1} \mid z_t, x_0) \parallel p_\btheta(z_{t-1} \mid z_t)] \right] }_{\mathcalJ_{t-1}}
\nonumber\\
&\quad - \underbrace{\Exp_q \left[ \KL[q(z_T \mid x_0) \parallel p(z_T)]  \right]}_{\mathcalJ_T}.
\label{equation:bernoulli_elbo}
\end{align}
The terminal term $\mathcalJ_T \approx 0$ in practice, since the terminal forward distribution $q(z_T \mid x_0)$ asymptotically matches the uniform prior $p(z_T)=\bernoullidist(1/2)$.
Each intermediate loss term $\mathcalJ_{t-1}$ computes the KL divergence between two Bernoulli distributions $\KL[\bernoullidist(a) \parallel \bernoullidist(b)]$ (see Problem~\ref{prob:kl_bernoulli}),
with
\begin{itemize}
\item $a = q(z_{t-1} = 1 \mid z_t, x_0)$: the success probability of the true posterior, computed using the ground-truth original data $x_0$;
\item $b = q(z_{t-1} = 1 \mid z_t, \widehatx_0)$:  the success probability of the approximate posterior, computed using the model-predicted data $\widehatx_0$.
\end{itemize}
This framework can be seamlessly extended to high-dimensional binary datasets $\bx_0\in\real^D$. A visual illustration of the full pipeline is presented in Figure~\ref{fig:ddpm_binomial}. The final training objective is obtained by summing or averaging the ELBO loss terms across all diffusion steps $t$ and all $D$ binary dimensions. For sampling, we first sample the terminal latent variable $\bz_T \sim \text{Bernoulli}(1/2)$, then perform ancestral sampling over the learned reverse kernel $p_\btheta(\bz_{t-1} \mid \bz_t)$ to recover the original binary data $\bx_0$.

\section{Denoising Diffusion Probabilistic Models (DDPMs)}\label{section:ddpm_key}

As mentioned previously, DDPMs introduced key improvements and optimizations to DPMs, addressing some of their shortcomings and significantly enhancing the quality of data generated by diffusion models. This is what enables diffusion models to achieve outstanding performance in the field of image or video generation. The central improvement introduced by DDPMs is a revision to what the parameterized model predicts:
\begin{enumerate}[label=(A4)]
\item Instead of predicting the original $\bx_0$, the model now predicts the noise added at each timestep, reducing the difficulty of model learning  \citep{ho2020denoising}. 
\end{enumerate}

Why predict noise rather than a less noisy data or the original clean data?
The core motivation stems from the relative learning difficulty of the two prediction tasks.
At intermediate diffusion steps, the latent variable $\bz_t$ is dominated by noise, especially as $t$ approaches $T$. Directly predicting the full clean data $\bx_0$ or a fully denoised latent state $\bz_{t-1}$ from such a heavily corrupted signal requires the model to recover fine-grained, high-frequency structural details from extremely noisy inputs---a highly nonlinear and challenging regression problem.
In contrast, the noise $\bepsilon_t$ that defines the diffusion kernel is zero-mean, independent and identically distributed Gaussian noise across all spatial dimensions and timesteps. Its statistical structure remains simple and consistent throughout the forward process, making the prediction target considerably more predictable and stable.

Empirically, predicting noise also yields lower-variance gradients during training and improved convergence behavior. This is because the model learns a signal with a stationary, standardized distribution rather than one that varies drastically across timesteps and spatial regions (e.g., different image regions).
Furthermore, formulating the learning objective around noise prediction simplifies the loss landscape, mitigates strong covariate shift between early and late diffusion steps, and allows the network to focus on estimating a well-behaved signal rather than hallucinating entire data structures from heavily obscured inputs. This reduction in learning difficulty directly translates to higher-quality generated samples and more stable training dynamics.

\subsection{Predicting the Noise}\label{section:ddpm_rev}

As highlighted earlier, the core modification that yields superior generation quality redefines the neural network's prediction target. Instead of predicting the denoised  data at each Markov chain step, the network learns to estimate the cumulative noise injected into the original clean data to produce the noisy latent sample at the corresponding timestep. To derive this formulation, we first rearrange the diffusion kernel expression in \eqref{equation:diffusion_kernel_zt_form_ddpm} to isolate the clean data term:
\begin{equation}\label{eq:x_from_zt_epsilon}
\bx_0 = \frac{1}{\sqrt{\alpha_t}} \bz_t - \frac{\sqrt{1 - \alpha_t}}{\sqrt{\alpha_t}} \bepsilon_t.
\end{equation}
Substituting this expression into the data-conditioned reverse Gaussian distribution in ~\eqref{equation:ddpm_reverse_conditional_gaussian} allows us to reparameterize the mean $\bmu^q_t(\bx_0, \bz_t)$ of the reverse conditional distribution $q(\bz_{t-1} \mid \bz_t, \bx_0)$ in terms of the original clean data $\bx_0$ and the accumulated noise $\bepsilon_t$, resulting in:
\begin{equation}\label{equation:ddpm_mu_in_terms_of_epsilon}
\bmu^q_t(\bx_0, \bz_t) = \frac{1}{\sqrt{1 - \beta_t}} \left\{ \bz_t - \frac{\beta_t}{\sqrt{1 - \alpha_t}} \bepsilon_t \right\},
\end{equation}
where we have used the definition of $\alpha_t = \prod_{\tau=1}^{t} (1 - \beta_\tau)$ in \eqref{equation:ddpm_alpha_t_ddpm}.
Consistent with this reparameterization, we replace the original network formulation  $\widehatbx^\btheta_t(\bz_t)\equiv \bmu^\btheta_t(\bz_t)$---which directly predicts denoised data for each timestep $t$---with a noise-prediction network $\bepsilon^\btheta_t(\bz_t)$. 
This new network is optimized to estimate the cumulative noise added to $\bx_0$ to generate the noisy latent state $\bz_t$. 
Retracing the derivation steps for \eqref{equation:ddpm_mu_in_terms_of_epsilon} yields the analytical relationship between the data-prediction and noise-prediction network outputs:
\begin{equation}\label{equation:nu_in_terms_of_noisenet}
\bmu^\btheta_t(\bz_t) = \frac{1}{\sqrt{1 - \beta_t}} \left\{ \bz_t - \frac{\beta_t}{\sqrt{1 - \alpha_t}} \bepsilon^\btheta_t(\bz_t) \right\}.
\end{equation}
Note that $\bepsilon_t$  denotes the cumulative noise injected into the original clean data $\bx_0$ (see~\eqref{equation:diffusion_kernel_zt_form_ddpm}). 
Accordingly, the network $\bepsilon^\btheta_t(\cdot)$ predicts the total accumulated noise across all preceding diffusion steps, rather than only the incremental noise added at timestep $t$ alone.

\begin{subequations}\label{equation:ddpm_loss_form_allterms}
\paragrapharrow{Reconstruction term.}
The reconstruction term in the ELBO~\eqref{equation:ddom_elbo_final} can be approximated using~\eqref{equation:ddpm_fourth_mc_reconstruction} with a sampled value of $\bz_1$. Using the form~\eqref{equation:ddpm_reconz1_logprob} for $\ln p_\btheta(\bx_0 \mid \bz_1)$, we obtain
\begin{align}
\ln p_\btheta(\bx_0 \mid \bz_1) 
&= -\frac{1}{2(\sigma^\btheta_t)^2} \normtwo{\bx_0 - \bmu^\btheta_1(\bz_1)}^2 + \text{const} \nonumber \\
&= -\frac{\beta_1}{2(1 - \beta_1)(\sigma^\btheta_t)^2} \normtwo{\bepsilon^\btheta_1(\bz_1) - \bepsilon_1}^2 + \text{const}. 
\label{equation:ddpm_reconstruction_noise_form}
\end{align}
where the last equality follows from substituting $\bmu^\btheta_1(\bz_1)$ using~\eqref{equation:nu_in_terms_of_noisenet}, substituting \(\bx_0\)  $\bx_0$ using~\eqref{equation:diffusion_kernel_zt_form_ddpm} or \eqref{eq:x_from_zt_epsilon}, and applying the identity $\alpha_1 = (1 - \beta_1)$ from~\eqref{equation:ddpm_alpha_t_ddpm}.

\paragrapharrow{Consistent terms.}
Similarly, substituting~\eqref{equation:ddpm_mu_in_terms_of_epsilon} and~\eqref{equation:nu_in_terms_of_noisenet} into the consistent terms \eqref{equation:ddpm_kl_gaussian} yields
\begin{align}
\KL[q(\bz_{t-1} &\mid \bz_t, \bx_0) \parallel p_\btheta(\bz_{t-1} \mid \bz_t)] 
=\frac{1}{2(\sigma^\btheta_t)^2} \normtwo{\bmu^q_t(\bx_0, \bz_t) - \bmu^\btheta_t(\bz_t)}^2 + \text{const} \nonumber \\
&= \frac{\beta_t^2}{2(1 - \alpha_t)(1 - \beta_t)(\sigma^\btheta_t)^2} \normtwo{\bepsilon^\btheta_t(\bz_t) - \bepsilon_t}^2 + \text{const} \nonumber \\
&= \frac{\beta_t^2}{2(1 - \alpha_t)(1 - \beta_t)(\sigma^\btheta_t)^2} \normtwo{\bepsilon^\btheta_t(\sqrt{\alpha_t} \bx_0 + \sqrt{1 - \alpha_t} \bepsilon_t) - \bepsilon_t}^2 + \text{const},
\label{equation:ddpm_kl_noise_prediction}
\end{align}
where the final equality uses the relation  $\bz_t=\sqrt{\alpha_t} \bx_0 + \sqrt{1 - \alpha_t} \bepsilon_t$ from~\eqref{equation:diffusion_kernel_zt_form_ddpm}.
\end{subequations}

\paragrapharrow{The final objective function.}
Observe that~\eqref{equation:ddpm_reconstruction_noise_form} takes exactly the same form as~\eqref{equation:ddpm_kl_noise_prediction} for the special case $t = 1$. 
Thus, both the reconstruction and consistency terms are expressed in terms of the squared difference between the predicted noise and the true noise.
\citet{ho2020denoising} observed empirically that performance is further improved by simply omitting the scalar prefactors in front of the terms in~\eqref{equation:ddpm_loss_form_allterms}, so that all steps in the Markov chain receive equal weighting.
Summing this simplified form of~\eqref{equation:ddpm_loss_form_allterms} over all steps $t=1,2,\ldots,T$ results in a training objective of the form
\begin{equation}\label{equation:ddpm_equaweight_sum_loss}
\mathcalJ(\btheta)
= \sum_{t=1}^T \normtwo{\bepsilon^\btheta_t\bigl(\sqrt{\alpha_t}\bx_0 + \sqrt{1 - \alpha_t}\bepsilon_t\bigr) - \bepsilon_t}^2.
\end{equation}
More compactly, the loss can be written as an expectation over all observed data samples:
\begin{mybox}
\begin{align}
\mathcalJ(\btheta)
&=\Exp_\diamondsuit\left[\normtwo{\bepsilon^\btheta_t\bigl(\sqrt{\alpha_t}\bx_0 + \sqrt{1 - \alpha_t}\bepsilon_t\bigr) 
- \bepsilon_t}^2\right],
\label{equation:ddpm_equaweight_sum_loss2}\\
\text{with }\diamondsuit 
&=  t \sim \uniformdist(\{1,\ldots,T\}),\,  \bx_0 \sim {p_{\text{data}}}(\bx), \,  \bepsilon_t \sim \normal(\bzero, \bI_D). \nonumber
\end{align}
\end{mybox}
Note again that the network $\bepsilon^\btheta_t(\cdot)$ predicts the total noise added to the original data vector $\bx_0$, ot merely the incremental noise introduced at step $t$.
The corresponding gradient computation is presented in Algorithm~\ref{alg:diffusion_training}.

\begin{algorithm}
\caption{DDPM Training Procedure \index{Denoising diffusion probabilistic models}\index{DDPMs}}
\label{alg:diffusion_training}
\begin{algorithmic}[1]
\Require Training data $\mathcalX=\{\bx_n\}$, noise schedule $\{\beta_t\}_t$, the total number of steps $T$ (standard choice: $T = 1,000$, $\beta_1 = 10^{-4}$, $\beta_T = 0.02$);
\State \textbf{initialize:} $\btheta$;
\For{$cnt=0,1,2,\ldots$}
\State $\bx_0\sim \mathcalX$; \Comment{(DDPM$_1$)} 
\State $t\sim \uniformdist(\{1,2,\ldots,T\})$; \Comment{(DDPM$_2$)} 
\State $\bepsilon\sim\normal(\bzero, \bI_D)$;  \Comment{(DDPM$_3$)} 
\State $\bz_t\leftarrow \sqrt{\alpha_t}\bx_0+\sqrt{1-\alpha_t}\bepsilon$; \Comment{(DDPM$_4$)} 
\State Update network by minimizing loss: $\mathcalJ(\btheta) \leftarrow \normtwo{\bepsilon^\btheta_t(\bz_t) - \bepsilon}^2$; \Comment{(DDPM$_5$)} 
\EndFor
\State \Return  Optimized network parameter $\btheta$;
\end{algorithmic}
\end{algorithm}
\subsection{Generating New Samples}\label{section:gen_ddpm}

\begin{figure}[H]
\centering  
\subfigtopskip=2pt 
\subfigbottomskip=9pt 
\subfigcapskip=-5pt 
\includegraphics[width=0.9\textwidth]{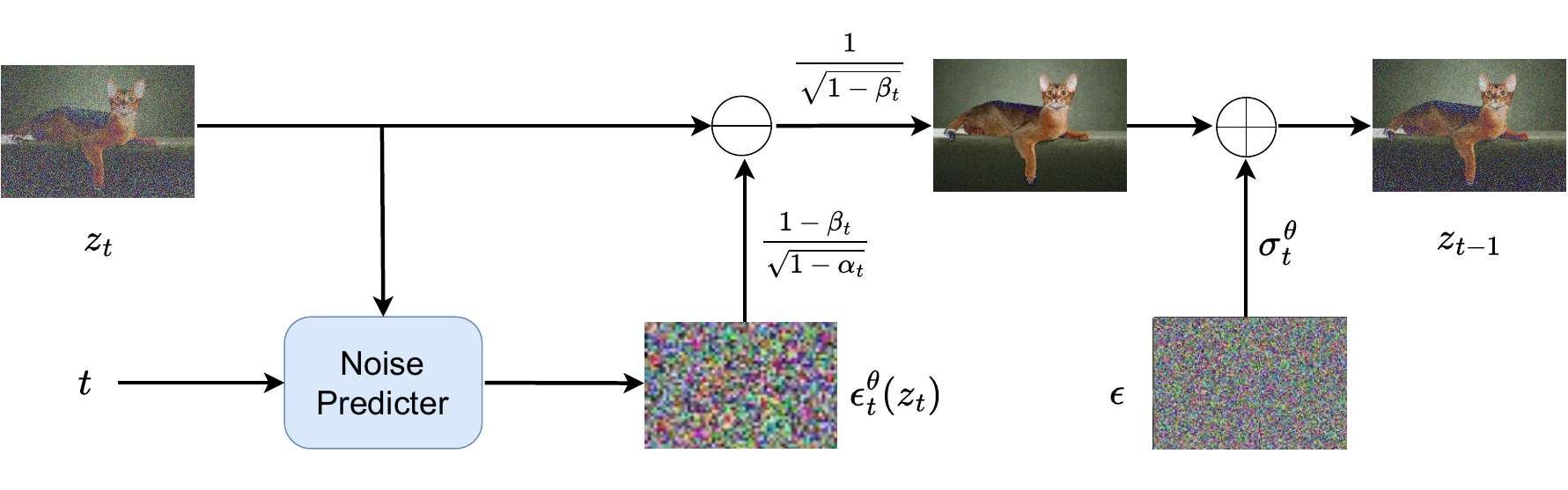}
\caption{
Conceptual illustration of sampling from a DDPM.}
\label{fig:DDPM_sample_add_noise}
\end{figure}

Once the network has been trained, we can generate new samples in the data space by first sampling from the Gaussian distribution $p(\bz_T)\sim \normal(\bzero, \bI_D)$ and then performing successive denoising through each step of the Markov chain. 
Given a denoised latent sample $\bz_t$ at step $t$, we generate a sample $\bz_{t-1}$ in three steps. First we evaluate the output of the neural network, denoted  $\bepsilon^\btheta_t(\bz_t)$. 
From this predicted noise, we evaluate $\bmu^\btheta_t(\bz_t)$ using~\eqref{equation:nu_in_terms_of_noisenet}.
Finally we generate a sample $\bz_{t-1}$ from $p_\btheta(\bz_{t-1} \mid \bz_t) = \normal(\bz_{t-1} \mid \bmu^\btheta_t(\bz_t),\, (\sigma^\btheta_t)^2 \bI_D)$ by adding noise scaled by the variance so that
\begin{equation}\label{equation:ddpm_sampling_step}
\bz_{t-1} = \bmu^\btheta_t(\bz_t) + \sigma^\btheta_t \, \bepsilon,
\quad\text{where $\bepsilon \sim \normal(\bepsilon \mid \bzero, \bI_D)$},
\end{equation}
where setting either $(\sigma^\btheta_t)^2=\beta_t$ or $(\sigma^\btheta_t)^2=(\sigma^q_t)^2$ yields comparable performance \citep{ho2020denoising}. 
The first choice is optimal when  $\bx_0\sim\normal(\bzero,\bI_D)$, 
while the second is optimal when  $\bx_0$ is deterministically fixed to a single point. 
These represent two extreme settings corresponding to upper and lower bounds on the entropy of the reverse process for data with coordinate-wise unit variance  \citep{sohl2015deep}.
Note that the network $\bepsilon^\btheta_t(\cdot)$ predicts the total noise added to the original data vector $\bx_0$ to produce $\bz_t$. However, in the sampling step, we only subtract a fraction $\beta_t / \sqrt{1 - \alpha_t}$ of this noise from $\bz_t$ (see \eqref{equation:nu_in_terms_of_noisenet}), 
then inject additional noise with variance $\sigma^\btheta_t$ to generate $\bz_{t-1}$ (see \eqref{equation:ddpm_reverse_model_gaussian}).
At the final step, when computing the synthetic data sample $\widetildebx_0$, 
we do not add extra noise, as our goal is to produce a noise-free output. 
The full sampling procedure is summarized in Algorithm~\ref{alg:diffusion_sampling} and illustrated in Figure~\ref{fig:DDPM_sample_add_noise}.

One may also wonder why we add $\bepsilon$ to $\bz_{t-1}$ in Step (SDDPM3).
Randomness at each step prevents mode collapse or a reduction in the overall probability mass of the generative process, analogous to sampling in language models. At each step, we do not greedily select the highest-probability token, even though the overall objective is to maximize the probability of the complete sequence. In this context, \textit{beam search} often performs better than naive random sampling.
Nevertheless, at the final step, we use the expectation $\bmu^\btheta_1(\bz_1)$ instead of a sampled value from the corresponding distribution.
This choice is consistent with the first term of our ELBO~\eqref{equation:ddom_elbo_final}. In practice, all regression models ultimately output expected values rather than stochastic samples.

\begin{algorithm}
\caption{Sampling from a DDPM\index{Denoising diffusion probabilistic models}\index{DDPMs}}
\label{alg:diffusion_sampling}
\begin{algorithmic}[1]
\Require Trained denoising network $\bepsilon^\btheta_t(\bz)$, noise schedule $\{\beta_t\}_t$, the total number of steps $T$ (standard choice: $T = 1,000$, $\beta_1 = 10^{-4}$, $\beta_T = 0.02$);
\State $\bz_T\sim \normal(\bzero, \bI_D)$;
\For{\textcolor{black}{$t=T,T-1,\ldots, 2$}}
\State $\bmu^\btheta_t(\bz_t) = \frac{1}{\sqrt{1 - \beta_t}} \left\{ \bz_t - \frac{\beta_t}{\sqrt{1 - \alpha_t}} \bepsilon^\btheta_t(\bz_t) \right\}$; \Comment{(SDDPM$_1$)} 
\State $\bepsilon\sim\normal(\bzero, \bI_D)$;  \Comment{(SDDPM$_2$)} 
\State $\bz_{t-1}\leftarrow \bmu^\btheta_t(\bz_t) +\sigma^\btheta_t\bepsilon \sim p_\btheta(\bz_{t-1} \mid \bz_t)$; \Comment{(SDDPM$_3$)} 
\EndFor
\State \textcolor{black}{$\widetildebx_0\leftarrow \frac{1}{\sqrt{1 - \beta_1}} \left\{ \bz_1 - \frac{\beta_1}{\sqrt{1 - \alpha_1}} \bepsilon^\btheta_1(\bz_1) \right\}$};
\State \Return New sample $\widetildebx_0$;
\end{algorithmic}
\end{algorithm}

\subsection{Latent Diffusion Models (LDMs)}\label{section:ldm_intro}

We have observed that diffusion models can be computationally demanding, as they sequentially reverse a noise process that may involve hundreds or even thousands of steps.
\citet{song2020denoising} introduced a related framework called \textit{denoising diffusion implicit models (DDIMs)}, which relaxes the Markovian assumption of the noise process while preserving the same training objective; see Section~\ref{section:ddims}.
This approach delivers a one or two order-of-magnitude speedup during sampling without compromising the quality of generated samples.

\begin{figure}[h!]
\centering  
\vspace{-0.35cm} 
\subfigtopskip=2pt 
\subfigbottomskip=2pt 
\subfigcapskip=-5pt 
\subfigure[DDPM.]{\label{fig:DDPM_ddpm2ldm1}
\includegraphics[width=0.43\linewidth]{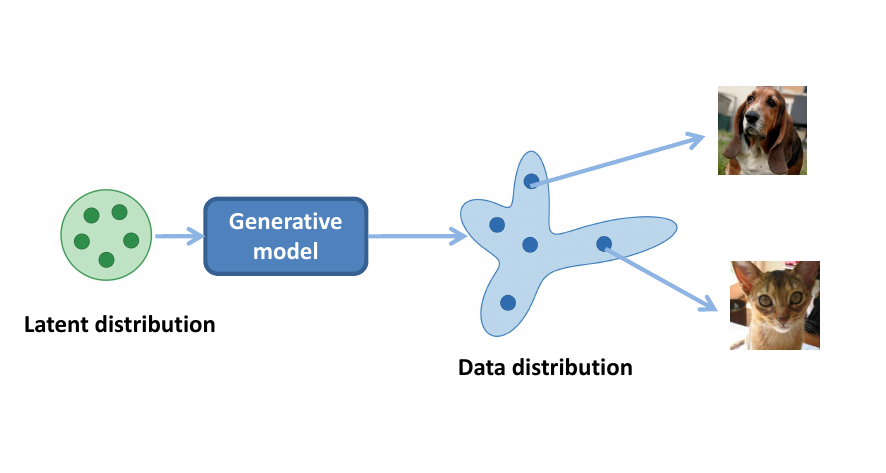}}
\subfigure[LDM.]{\label{fig:DDPM_ddpm2ldm2}
\includegraphics[width=0.54\linewidth]{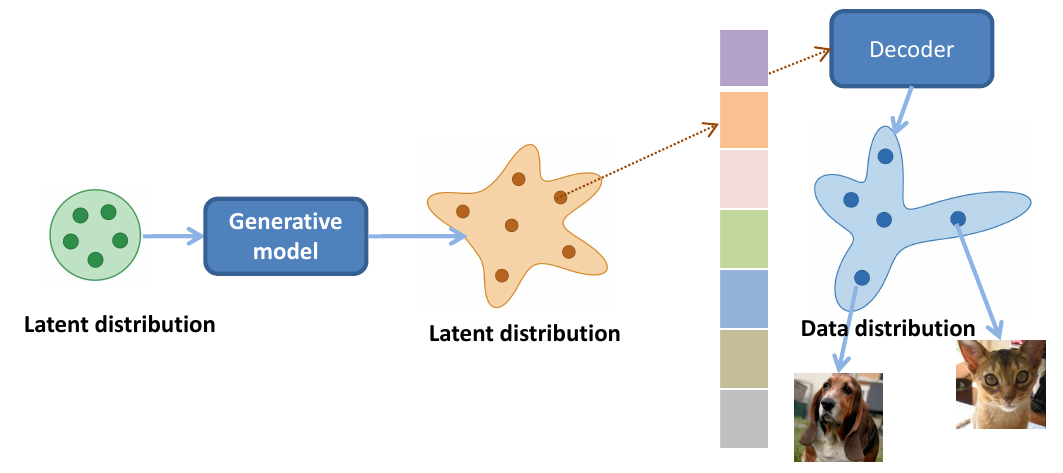}}
\caption{Conceptual illustration of the difference between standard DDPMs and LDMs. 
LDMs represent a computationally efficient variant of  DDPMs. Instead of performing diffusion directly on high-resolution data, LDMs conduct diffusion on compressed latent tokens and use a decoder to reconstruct the final output, enabling scalable generation of high-fidelity samples compared to conventional DDPMs.}
\label{fig:DDPM_ddpm2ldm}
\end{figure}

A different strategy to mitigate the high computational cost of applying diffusion models directly in high-resolution data space is known as \textit{latent diffusion models (LDMs)} \citep{rombach2022high}.
Standard DDPMs carry out the full forward noising and reverse denoising processes directly on the high-dimensional raw data distribution, operating in the pixel space of the target signal. Although this formulation produces high-fidelity samples, it incurs significant computational overhead, as the model must learn denoising transitions across the full dimensionality of the input data.
High-resolution data exhibit extremely large dimensionality, meaning hundreds of iterative diffusion steps require substantial GPU memory and heavy computation. This makes high-quality, high-resolution generation slow and resource-intensive, limiting deployment on standard hardware.

In contrast, LDMs adopt a two-stage paradigm: first, a learned encoder (such as the variational autoencoder, or VAE, described in Section~\ref{section:ae_vae}) maps the data into a low-dimensional latent space that retains only the most semantically relevant information.
Critically, this latent representation---also referred to as a latent code, embedding, token, or latent vector---captures the essential semantic and structural information of the input while discarding redundant noise and fine-grained details.
The diffusion process is then applied exclusively to these compact latent representations. The final signal is reconstructed from the generated latents using a corresponding decoder, or ``detokenizer."
By restricting the computationally expensive diffusion steps to a lower-dimensional manifold, LDMs achieve substantially improved training and sampling efficiency while maintaining the generative quality of the original framework, thus enabling scalable synthesis of high-resolution outputs.
The fundamental difference between DDPMs and LDMs lies in the domain where the diffusion process is defined. A comparison between DDPMs and LDMs is shown in Figure~\ref{fig:DDPM_ddpm2ldm}.
Further details on the architectures used within the LDM framework can be found in Section~\ref{section:large_gen}.


\begin{figure}[h!]
\centering                      
\vspace{-0.35cm}                 
\subfigtopskip=2pt               
\subfigbottomskip=2pt            
\subfigcapskip=-5pt              
\subfigure[Encoding in a LDM.]{\label{fig:ldm_latent}
\includegraphics[width=0.35\linewidth]{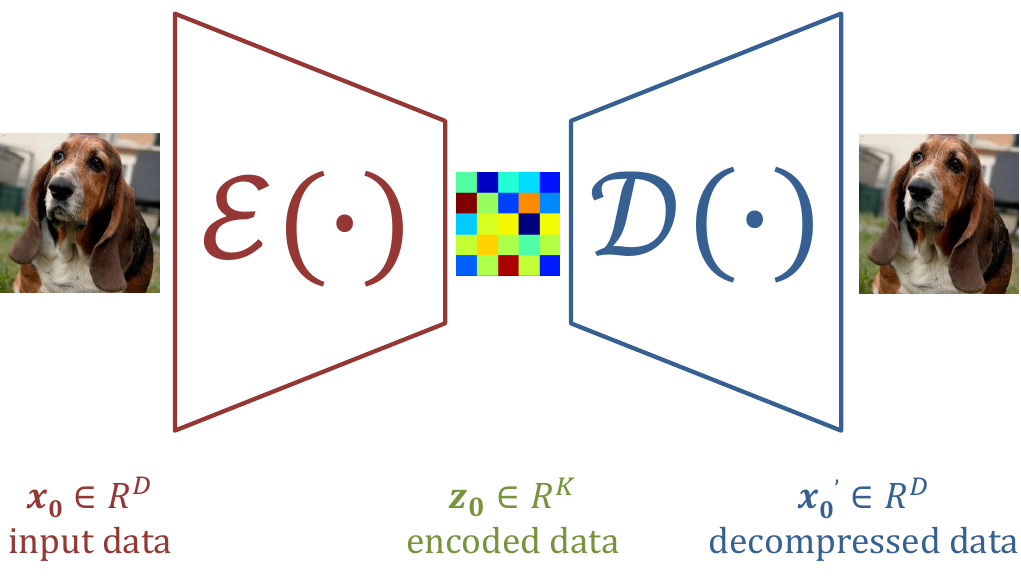}}
\subfigure[DDPM operates in the latent space in a LDM.]{\label{fig:ldm_process}
\includegraphics[width=0.58\linewidth]{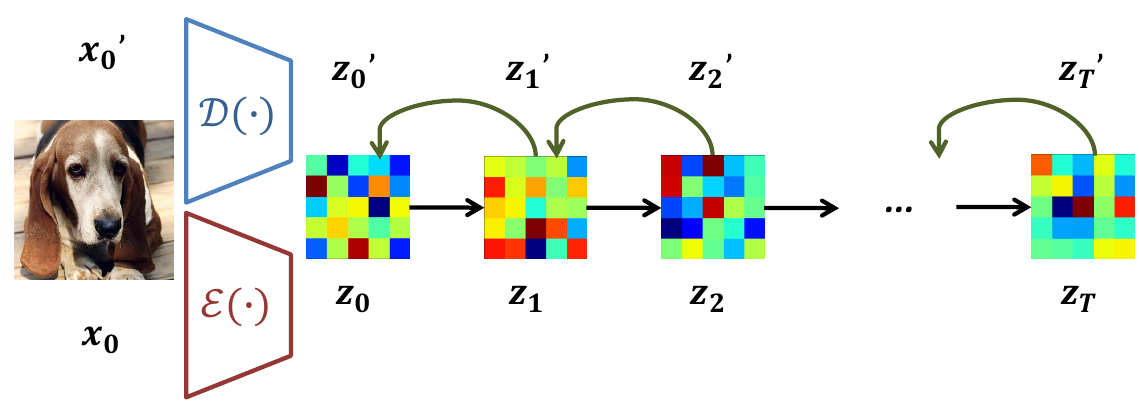}}
\caption{Illustration of latent diffusion models. Encoding the original data $\bx_0$ into a latent vector $\bz_0$; and diffusion is applied to the latent vector in a LDM.}
\label{fig:ldm_process_ALL}
\end{figure}

Specifically, LDMs move the entire diffusion process from high-resolution pixel space into a compressed latent space.
First, a pre-trained autoencoder (such as a variational autoencoder, or VAE) compresses raw data into compact low-dimensional latent representations: $\bz_0=\mathcalE(\bx_0)$.
All forward noising and reverse denoising steps of diffusion are then performed on these small latent vectors rather than full-resolution data.
Intuitively, a well-trained autoencoder can be seen as filtering out high-frequency or semantically irrelevant details, allowing the generative model to focus on important, perceptually meaningful features.
Finally, the decoded latent output $\bz_0'$ is mapped back to a high-quality data: $\bx_0' = \mathcalD(\bz_0')$.
This encoder (paired with its corresponding decoder), or VAE, can be a pre-trained model (see, for example, Algorithm~\ref{alg:amo_vae}) with parameters frozen during the training of the latent diffusion model, as illustrated in Figure~\ref{fig:ldm_latent}.
This compression drastically reduces computational cost and memory consumption while preserving essential visual information. The core diffusion logic remains unchanged; only the operating domain is shifted to latent space for improved efficiency, as shown in Figure~\ref{fig:ldm_process}.

\subsection{Learning the Variance}
\begin{figure}[h!]
\centering                      
\vspace{-0.35cm}                 
\subfigtopskip=2pt               
\subfigbottomskip=2pt            
\subfigcapskip=-5pt              
\subfigure[The ratio $(\sigma^q_t)^2/\beta_t$ across diffusion steps for processes of varying lengths.]{\label{fig:ddpm_noise_div}
\includegraphics[width=0.485\linewidth]{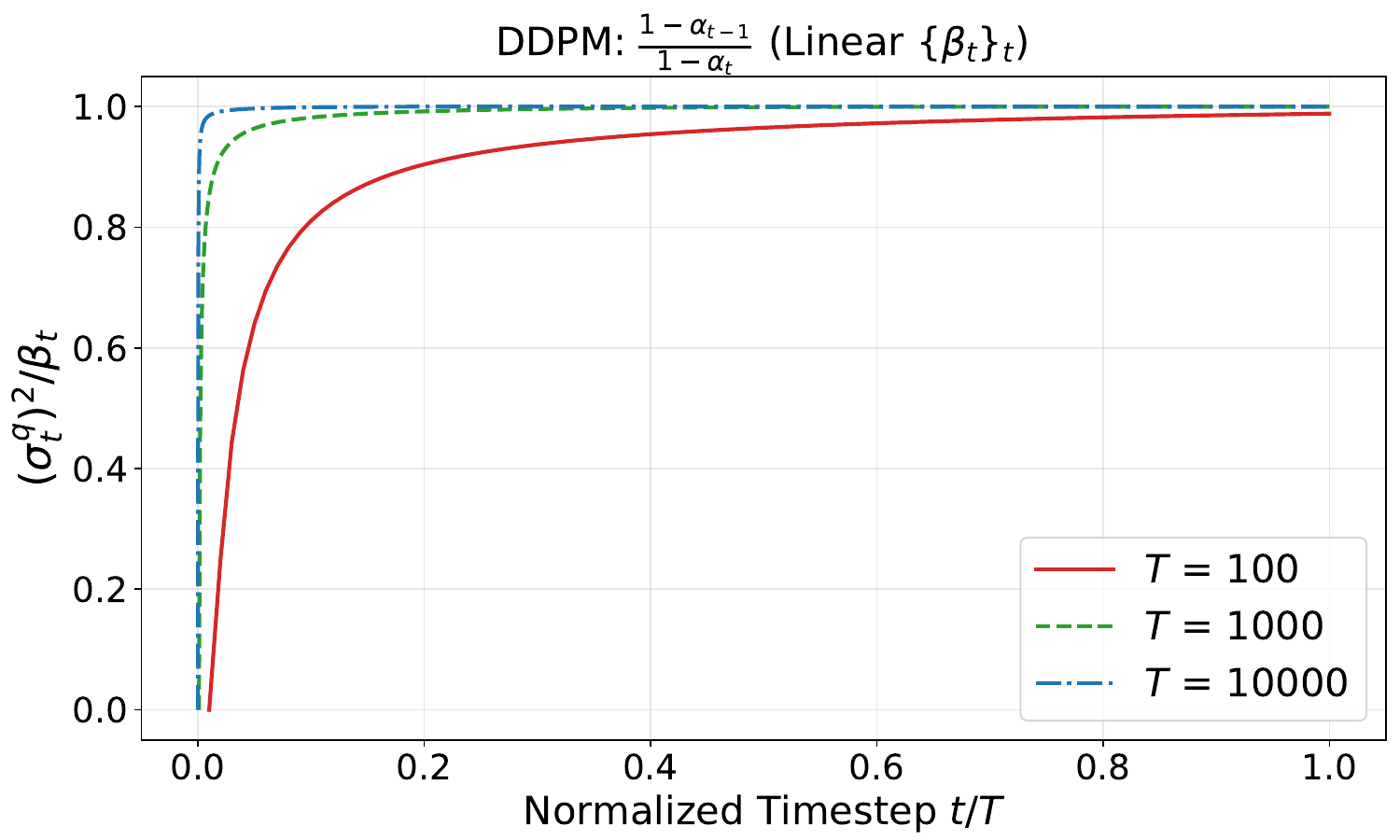}}
\subfigure[Cumulative product $\alpha_t$ under linear and cosine noise schedules.]{\label{fig:ddpm_noise_schedule_imp}
\includegraphics[width=0.485\linewidth]{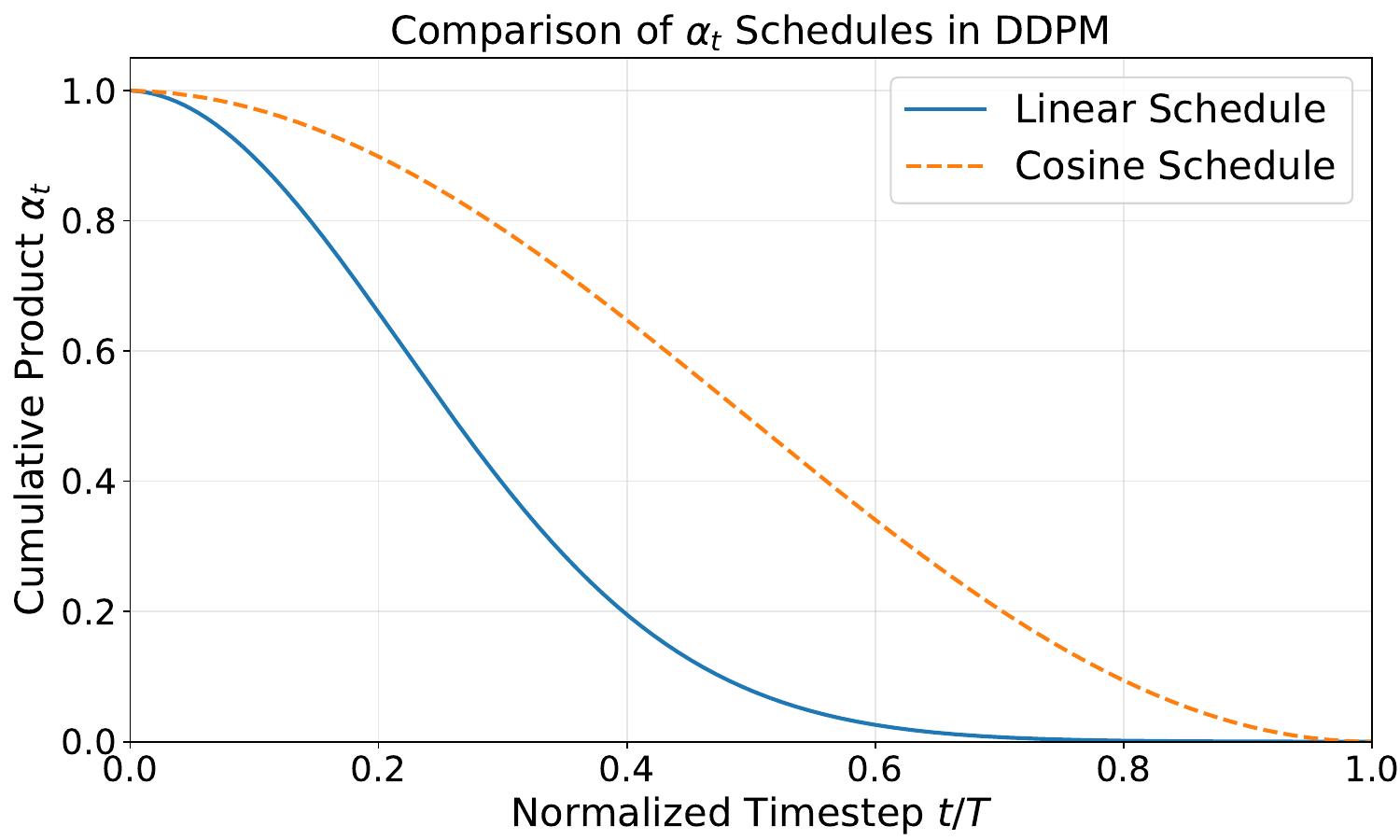}}
\caption{Improving the DDPM with learning the variance and alternative noise schedules.}
\label{fig:ddpm_noise_div_ALL}
\end{figure}

In the sampling step~\eqref{equation:ddpm_sampling_step}, we use a fixed variance $(\sigma^\btheta_t)^2=\beta_t$ (the increment noise at each forward step; see \eqref{equation:forward_step_zt_ddpm_ALL})
or $(\sigma^\btheta_t)^2=(\sigma^q_t)^2= \frac{\beta_t (1 - \alpha_{t-1})}{1 - \alpha_t}$ (the conditional distribution variance of the reverse process; see \eqref{equation:ddpm_reverse_conditional_gaussian}) to model the reverse process via $p_\btheta(\bz_{t-1}\mid \bz_t)=\normal\left( \bz_{t-1} \mid \bmu^\btheta_t(\bz_t),\, (\sigma^\btheta_t)^2 \bI_D \right)$.
The first choice is optimal when  $\bx_0\sim\normal(\bzero,\bI_D)$, 
and the second is optimal when  $\bx_0$ is deterministically fixed to a single point.
These represent two extreme choices corresponding to upper and lower bounds on the entropy of the reverse process for data with coordinatewise unit variance \citep{sohl2015deep}.
In practice, the two choices yield similar results \citep{ho2020denoising}. 
Given that $\beta_t$ and $(\sigma^q_t)^2$ correspond to opposite extremes, it is natural to ask why this choice has little effect on generated samples.
In Figure~\ref{fig:ddpm_noise_div}, we show that $\beta_t$ and $(\sigma^q_t)^2$
are nearly identical except near $t = 0$, where the model handles imperceptible fine details. 
Furthermore, as the number of diffusion steps increases, $\beta_t$ and $(\sigma^q_t)^2$ remain close to each other over a larger portion of the diffusion process.
This suggests that, in the limit of infinitely many diffusion steps, the selection of $(\sigma^\btheta_t)^2$ may have no meaningful impact on sample quality.
In other words, as more diffusion steps are added, the model mean $\bmu^\btheta_t(\bz_t)$ dominates the learned distribution far more than the variance term
$(\sigma^\btheta_t)^2\bI_D$.

Although the above reasoning supports fixing $(\sigma^\btheta_t)^2$ as a reasonable choice for sample quality, it does not address log-likelihood. 
In fact, \citet{nichol2021improved} showed that the first few steps of the diffusion process contribute most heavily to the evidence lower-bound.  
It is therefore plausible that log-likelihood can be improved by selecting a more appropriate form for $(\sigma^\btheta_t)^2$. 
To achieve this, we learn $(\sigma^\btheta_t)^2$ by interpolating between the two extremes $\beta_t$ and $(\sigma^q_t)^2$.
\citet{nichol2021improved} found that interpolation in the log domain performs well:
\begin{equation}
(\sigma^\btheta_t)^2(\bz_t)=
\exp\left\{\zeta\ln\beta_t +(1-\zeta)\ln (\sigma^q_t)^2\right\}.
\end{equation}
Since the original DDPM final objective $\mathcalJ(\btheta)$ in \eqref{equation:ddpm_equaweight_sum_loss2} does not depend on $(\sigma^\btheta_t)^2$, we then combines the final objective function $\mathcalJ(\btheta)$  in \eqref{equation:ddpm_equaweight_sum_loss2} and   the negative ELBO $-\mathcalF(\btheta)$ from \eqref{equation:ddom_elbo_final}:
\begin{equation}
\widehatbtheta=\argmin_\btheta \left\{\mathcalJ'(\btheta) = \mathcalJ(\btheta)-\lambda \mathcalF(\btheta)\right\},
\end{equation}
where a small $\lambda$ (e.g., $\lambda=0.001$) prevents $-\mathcalF(\btheta)$ from overwhelming $\mathcalJ(\btheta)$.

\index{Linear schedule}
\index{Cosine schedule}
\subsection{Choices of Noise Schedule}
In the original DDPM paper \citet{ho2020denoising}, which uses a total of $T = 1000$ steps, the noise schedule is defined to be \textit{linear}:
\begin{equation}
\beta_t = \beta_1 + \frac{t-1}{T-1}\left(\beta_T - \beta_1\right),
\quad 
\alpha_t = \prod_{\tau=1}^{t} (1 - \beta_\tau),
\quad t = 1, 2 \ldots, T.
\end{equation}
In contrast, the \textit{cosine schedule} is defined as follows  \citep{nichol2021improved}:
\begin{equation}
{\alpha}_t = \cos^2\left(\frac{t/T + s}{1 + s} \cdot \frac{\pi}{2}\right),
\qquad
\beta_t = 1 - \frac{{\alpha}_t}{{\alpha}_{t-1}},
\end{equation}
with $s=0.008$ and $\beta_t$ capped at $0.999$.
A comparison of the cumulative product $\alpha_t = \prod_{\tau=1}^{t} (1 - \beta_\tau)$ is shown in Figure~\ref{fig:ddpm_noise_schedule_imp}.
This illustrates that under the cosine schedule, $\alpha_t$ decreases linearly in the middle of the diffusion process, while remaining nearly constant near the boundaries $t = 0$ and $t = T$, thus avoiding abrupt variations in noise magnitude.
By comparison, $\alpha_t$ under the linear schedule decays toward zero much more rapidly, discarding signal information faster than necessary.
This explains why the forward noising process becomes excessively noisy at low resolutions (e.g., for images of size $64 \times 64$ or $32 \times 32$) and therefore contributes little to final sample quality.

\index{Tweedie's formula}
\index{Score matching}
\index{Fisher score}
\index{Stein score}
\subsection{Score-Based Interpretation of DDPMs}\label{section:score_ddpm}

The denoising diffusion models presented in this chapter are closely connected to another family of deep generative models developed largely independently, which are built upon \textit{score matching} \citep{hyvarinen2005estimation, song2019generative}. 
This subsection focuses exclusively on the equivalence between DDPMs and score matching. A more general score-matching framework grounded in stochastic differential equations is introduced in Section~\ref{section:score_mat}.

We first introduce the preliminary background of \textit{Tweedie's formula}. 
In statistics, unknown parameters of probability distributions are commonly estimated from observed samples via approaches such as maximum likelihood estimation and Bayesian estimation. Tweedie's formula serves as a dedicated method for estimating the mean parameter of distributions belonging to the exponential family~\footnote{See, for example, \citet{lu2021rigorous} for more details.}.  As Gaussian distributions fall into the exponential family, Tweedie's formula is applicable to estimating their mean parameters. We omit the mathematical proof of Tweedie's formula and focus solely on its practical application herein.

For a multivariate Gaussian random variable $\by \sim \normal(\by\mid \bmu_\by, \bSigma_\by)$, Tweedie's formula is defined as:
\begin{equation}\label{equation:tweedie_formula}
(\text{Tweedie's formula}) \qquad \Exp[\bmu_\by \mid \by] = \by + \bSigma_\by \underbrace{\nabla_\by \ln p(\by)}_{\text{score}},
\end{equation}
where the left-hand side $\Exp[\bmu_\by \mid \by]$ denotes the conditional expectation of $\bmu_\by$ given the observed sample $\by$, which intuitively corresponds to estimating the mean parameter $\bmu_\by$ from the sample $\by$. 
Here, $\ln p(\by)$ is the log-likelihood of the observed sample, and 
$\bs^p(\by)\triangleq\nabla_\by \ln p(\by)$ denotes the gradient of the sample log-likelihood, termed the  \textit{Stein score}, or simply the \textit{score}. 
\footnote{Note that a distinct concept, the \textit{Fisher score}, refers to the gradient of the log-likelihood with respect to model parameters, i.e., $\nabla_\btheta \ln p(\by\mid \btheta)$. In contrast, the Stein score is defined as the log-likelihood gradient with respect to the observed sample variable $\nabla_\by \ln p(\by)$.}

We now return to the forward transition kernel defined in~\eqref{equation:ddpm_diffusion_kernel_all}. The forward diffusion process yields the marginal distribution $q(\bz_t \mid \bx_0)$:
\begin{equation}
q(\bz_t \mid \bx_0) = \normal(\bz_t\mid \sqrt{\alpha_t} \bx_0, (1 - \alpha_t) \bI_D),
\end{equation}
which follows a Gaussian distribution with a known mean $\sqrt{\alpha_t} \bx_0$. 
Even with this known mean, we can apply Tweedie's formula~\eqref{equation:tweedie_formula} to estimate the mean parameter of this Gaussian distribution:
\begin{equation}
\sqrt{\alpha_t} \bx_0 = \Exp[\bmu_{\bz_t} \mid \bz_t] = \bz_t + (1 - \alpha_t) \nabla_{\bz_t} \ln p(\bz_t).
\end{equation}
Rearranging the above equation yields a key relation that expresses the original data sample $\bx_0$ in terms of the score function:
\begin{equation}\label{equation:ddpm_x0_score}
\bx_0 = \frac{\bz_t + (1 - \alpha_t) \nabla \ln p(\bz_t)}{\sqrt{\alpha_t}}.
\end{equation}
We now substitute this relation into the mean parameter $\bmu^q_t(\bx_0, \bz_t)$  of the true reverse transition distribution $q(\bz_{t-1} \mid \bz_t, \bx_0)$, whose closed-form expression is provided in~\eqref{eq:mean_reverse}. 
Then we obtain:
\begin{align}
\bmu^q_t(\bx_0, \bz_t) 
&= \frac{1}{\sqrt{1-\beta_t}} \bz_t + \frac{\beta_t}{\sqrt{1-\beta_t}} \nabla \ln p(\bz_t), \label{equation:mux0zt_score}
\end{align}
where we have used the fact that $\alpha_t = (1-\beta_t)\alpha_{t-1}$ in \eqref{equation:ddpm_alpha_t_ddpm}.
Consistent with the Gaussian modeling of the reverse diffusion process  in \eqref{equation:ddpm_reverse_model_gaussian}, where each timestep $t$ adopts a Gaussian distribution parameterized by mean $\bmu_t^\btheta(\bz_t)$, we define the mean of the learned reverse distribution $p_\btheta(\bz_{t-1} \mid \bz_t)$ to match the derived form above:
\begin{equation}
\bmu^\btheta_t(\bz_t) = \frac{1}{\sqrt{1-\beta_t}} \bz_t + \frac{\beta_t}{\sqrt{1-\beta_t}} \bs^\btheta_t(\bz_t).
\end{equation}
Here, the network output $\bs^\btheta_t(\bz_t)$ is trained to approximate the true score function $\nabla_{\bz_t} \ln p(\bz_t)$. 
We further derive the training objective by minimizing the KL divergence term in the ELBO loss~\eqref{equation:ddpm_kl_gaussian}:
\begin{align}
\arg \min_{\btheta} 
&\KL \big[q(\bz_{t-1} \mid \bz_t, \bx_0) \parallel p_{\btheta}(\bz_{t-1} \mid \bz_t)\big]\nonumber \\
&= \arg \min_{\btheta} \KL \big[\normal(\bz_{t-1}\mid \bmu^q_t, \bSigma^q_t) \parallel \normal(\bz_{t-1}\mid \bmu^\btheta_t,\bSigma^q_t)\big] \nonumber\\
&= \arg \min_{\btheta} \frac{1}{2(\sigma^q_t)^2} \normtwo{\frac{1}{\sqrt{\alpha_t}} \bz_t + \frac{\beta_t}{\sqrt{1-\beta_t}} \bs^\btheta_t(\bz_t) - \frac{1}{\sqrt{\alpha_t}} \bz_t - \frac{\beta_t}{\sqrt{1-\beta_t}} \nabla \ln p(\bz_t)}^2 \nonumber\\
&= \arg \min_{\btheta} \frac{1}{2(\sigma^q_t)^2} \frac{\beta_t^2}{1-\beta_t} \left[ \normtwo{\bs^\btheta_t(\bz_t) - \nabla \ln p(\bz_t)}^2 \right].
\label{equation:loss_score_ddpm_final}
\end{align}
All constant terms independent of the network parameters $\btheta$ are omitted, as they contribute no gradient signal and thus do not affect model training.
The final optimized objective takes the form of mean squared error, which is highly consistent with that of original diffusion models (DPMs) and noise-prediction-based diffusion models (DDPMs). 
In this formulation, the true score gradient $\nabla \ln p(\bz_t)$ can be computed directly via automatic differentiation. 
The training procedure for the score-based DDPM is illustrated in Algorithm~\ref{alg:score_ddpm_train}.

\begin{algorithm}
\caption{Score-Based DDPM Training Procedure\index{Denoising diffusion probabilistic models}\index{DDPMs}}
\label{alg:score_ddpm_train}
\begin{algorithmic}[1]
\Require Training data $\mathcalX=\{\bx_n\}$, noise schedule $\{\beta_t\}_t$, the total number of steps $T$ (standard choice: $T = 1,000$, $\beta_1 = 10^{-4}$, $\beta_T = 0.02$);
\State \textbf{initialize:} $\btheta$;
\For{$cnt=0,1,2,\ldots$}
\State $\bx_0\sim \mathcalX$; \Comment{(SBDPM$_1$)} 
\State $t\sim \uniformdist(\{1,2,\ldots,T\})$; \Comment{(SBDPM$_2$)} 
\State $\bepsilon\sim\normal(\bzero, \bI_D)$;  \Comment{(SBDPM$_3$)} 
\State $\bz_t\leftarrow \sqrt{\alpha_t}\bx_0+\sqrt{1-\alpha_t}\bepsilon$; \Comment{(SBDPM$_4$)} 
\State Ground-truth score:
$\bs^{\bthetastar}_t(\bz_t) \leftarrow \nabla_{\bz_t}\ln p(\bz_t) = -\frac{1}{\sqrt{1-\alpha_t}} \bepsilon$; \Comment{(SBDPM$_5$)} 
\State Update network by minimizing loss: $\mathcalJ(\btheta) \leftarrow \normtwo{\bs^\btheta_t(\bz_t) - \bs^{\bthetastar}_t (\bz_t)}^2$; \Comment{(SBDPM$_6$)} 
\EndFor
\State \Return Optimized network parameter $\btheta$;
\end{algorithmic}
\end{algorithm}

\begin{remark}
There are two main approaches to representing the score function $\bs^\btheta(\bz)$ using a deep neural network. Each element $s_i$ of $\bs$ corresponds to an element $z_i$ of $\bz$, so the first method uses a network with an equal number of output and input dimensions. However, the score function is defined as the gradient of a scalar function (the log-probability density), which belongs to a more constrained function class. An alternative approach is to use a network with a single scalar output $\psi(\bz)$ and then compute $\nabla_{\bz} \psi(\bz)$ via automatic differentiation. This second method requires two backpropagation passes and is thus computationally more costly. For this reason, most practical applications adopt the first approach.
\end{remark}

\begin{SCfigure}
\centering
\includegraphics[width=0.6\textwidth]{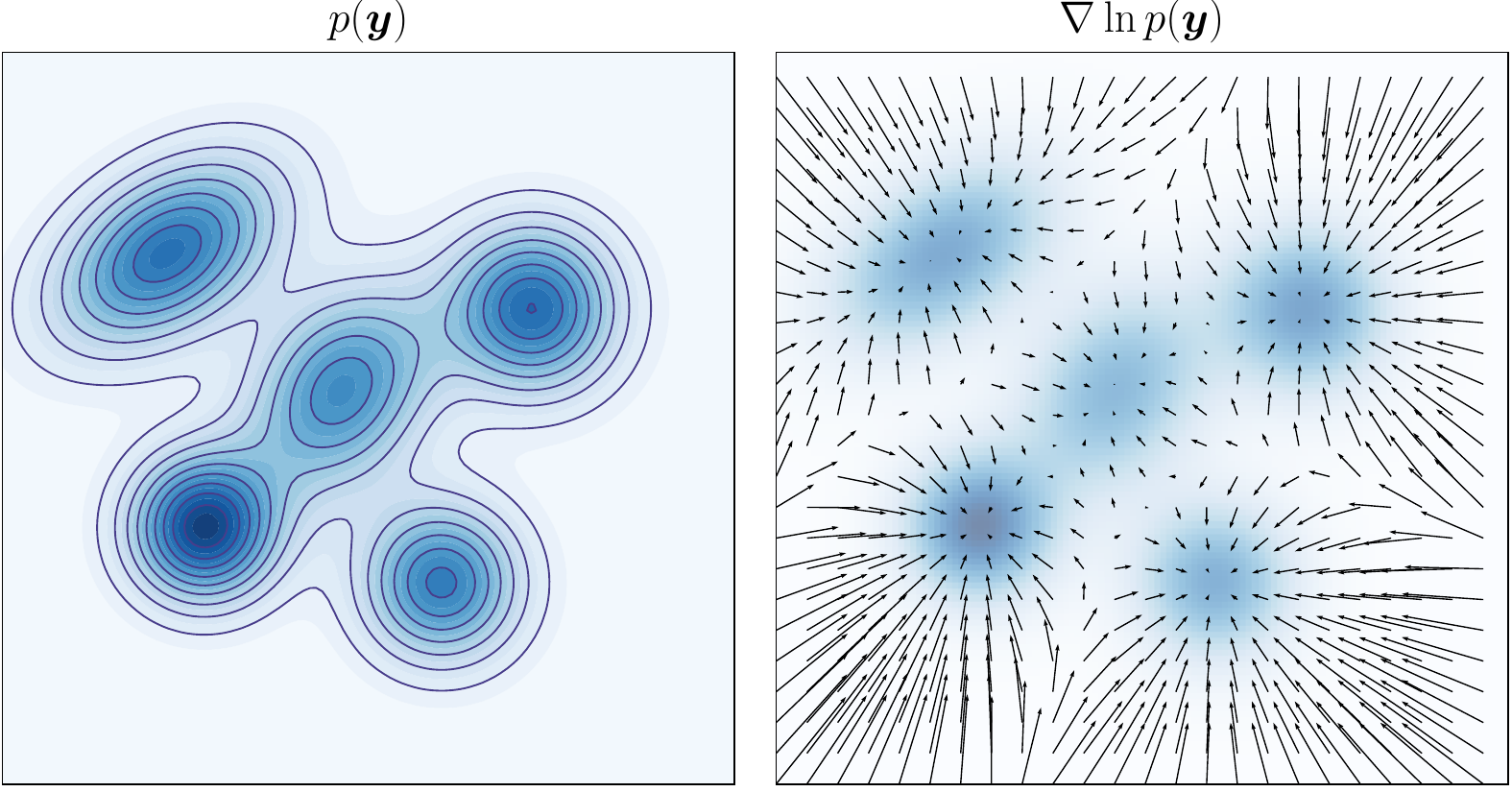}
\caption{Let $p(\by)$ be an arbitrary probability distribution. Then the {score function} of $p$ is defined as $\nabla \ln p(\by)$, i.e., as the gradient of the log-likelihood of $p$ with respect to $\bx$. The score has an intuitive meaning: $\nabla \ln p(\bx)$ points in the direction of steepest ascent of the log-likelihood.}
\label{fig:score_ascent}
\end{SCfigure}

We further establish the intrinsic connection between the score gradient and the diffusion noise. By combining \eqref{equation:ddpm_x0_score} and the diffusion kernel formulation~\eqref{equation:diffusion_kernel_zt_form_x0_ddpm}, we obtain the following equivalence:
\begin{align}
\bx_0 &= \frac{\bz_t + (1-\alpha_t)\nabla \ln p(\bz_t)}{\sqrt{\alpha_t}} = \frac{\bz_t - \sqrt{1-\alpha_t} \bepsilon_t}{\sqrt{\alpha_t}} \\
\implies  \quad \nabla &\ln p(\bz_t) = -\frac{1}{\sqrt{1-\alpha_t}} \bepsilon_t. \label{equation:score_noise}
\end{align}
The gradient direction corresponds to the direction of function maximization: $\nabla \ln p(\bz_t)$ points toward increasing log-likelihood (and thus higher probability density) of $p(\bz_t)$, i.e., an \textit{ascent direction} (see Figure~\ref{fig:score_ascent} illustrative example). 
The above result reveals that the true score gradient is exactly the negative of the diffusion noise, meaning the two vectors point in opposite directions. This opposing relationship between score gradient and injected noise offers clear intuitive interpretability for the diffusion generation process: the ascent direction corresponds to reversing the added noise.

\subsection{Final Remarks: Three Equivalent Views of Diffusion Models}\label{section:ddpm_final_remark}

To this point, we have presented three distinct formulations for diffusion models: (i) directly predicting the initial sample; (ii) predicting noise; and (iii) predicting the score. While these three formulations are theoretically equivalent, they differ considerably in terms of learning difficulty and model flexibility. Below, we summarize and compare these three approaches for clarity and review.

We provide a compact review of the DDPM mechanism. 
In DDPM, real-world data (e.g., images) is modeled as a random variable $\bx_0$, with its underlying data distribution denoted as $p_{\text{data}}(\bx_0)$. Since the true form of $p_{\text{data}}(\bx_0)$ is intractable and unknown, direct sampling from this exact distribution is infeasible.
Fortunately, abundant empirical observations of $\bx_0$ are available. We can therefore approximate $p_{\text{data}}(\bx_0)$ using these observed samples and generate new data from the approximated distribution.

The core principle of DDPM is to construct a Markov chain that gradually injects Gaussian random noise into $\bx_0$, ultimately converting the original structured data into pure standard Gaussian noise. This noise-injection procedure is defined as the forward (or noising) process. The corresponding reverse procedure is a progressive denoising process, which estimates the denoising transition distribution $p(\bz_{t-1} \mid \bz_t)$ at each timestep. New data samples are generated by iteratively denoising pure Gaussian noise $\bz_T$ sampled from the standard normal distribution, where $T$ denotes the total number of diffusion timesteps---namely, the number of noise injection steps in the forward diffusion process and the corresponding number of denoising steps in the reverse process (illustrated in Figure~\ref{fig:diffusion_graphical_model}).

The joint distribution of the entire diffusion process is formulated as $p(\bx_0, \bz_{1:T})$. Following the probability chain rule, the forward diffusion process decomposes the joint distribution via the diffusion kernel in \eqref{equation:ddpm_joint_latent_dist_ddpm}:
\begin{equation}
p(\bx_0, \bz_{1:T}) = p_{\text{data}}(\bx_0) 
q(\bz_1\mid\bx_0)
\prod_{t=2}^{T} q(\bz_t \mid \bz_{t-1}).
\end{equation}
Although the analytical form of $p_{\text{data}}(\bx_0)$ remains unknown, empirical observations of $\bx_0$ are accessible. 
The forward transition  $q(\bz_t \mid \bz_{t-1})$ follows a conditional Gaussian distribution with the probability density function specified in \eqref{equation:forward_step_zt_ddpm_ALL} and we recall below:
\begin{equation}
q(\bz_t \mid \bz_{t-1}) = \normal\left(\bz_t\mid \sqrt{1-\beta_t} \bz_{t-1}, \beta_t \bI_D\right),
\end{equation}
where the sequence $\beta_1<\beta_2<\ldots<\beta_T$ represents the predefined noise schedule. The transition kernel characterizes the Gaussian noise injection operation that transforms $\bz_{t-1}$ to $\bz_t$. Through iterative application of this kernel, the structured original data $\bx_0$ is progressively corrupted into pure Gaussian noise $\bz_T$.

However, iterative step-wise noising computation is computationally inefficient. To address this issue, we leverage a mathematical property of conditional Gaussian distributions to directly compute the latent state $\bz_t$ at any arbitrary timestep from the original data $\bx_0$ in a single step. 
This one-step mapping is formalized by the conditional distribution $q(\bz_t \mid \bx_0)$ from \eqref{equation:ddpm_diffusion_kernel_all}:
\begin{subequations}\label{equation:ddim_ddpm_ztx0_ALL} 
\begin{align}
q(\bz_t \mid \bx_0) &= \int q(\bz_{1:t} \mid \bx_0) \diff\bz_{1:t-1} 
= \normal\left(\bz_t\mid \sqrt{\alpha_t} \bx_0, (1 - \alpha_t) \bI_D\right) \label{equation:ddim_ddpm_ztx0} \\
\iff \quad \bz_t &= \sqrt{\alpha_t} \bx_0 + \sqrt{1 - \alpha_t} \bepsilon_t \quad ,  \, \bepsilon_t \sim \normal(\bzero, \bI_D).  \label{equation:ddim_ddpm_ztx0_prime}
\end{align}
\end{subequations}
The reverse of the noising process is the denoising process, which factorizes the joint distribution $p(\bx_0, \bz_{1:T})$ in reverse chronological order:
\begin{equation}
p(\bx_0, \bz_{1:T}) 
= p(\bz_T) \left\{\prod_{t=T}^{2} q(\bz_{t-1} \mid \bz_t)\right\} 
q(\bx_0\mid\bz_1),
\end{equation}
where $p(\bz_T) \sim \normal(\bzero, \bI_D)$ denotes the standard normal prior distribution for the final latent state, and $q(\bz_{t-1} \mid \bz_t)$ is the reverse transition kernel. This kernel describes the operation of removing partial Gaussian noise from $\bz_t$ to recover the cleaner latent state $\bz_{t-1}$, which defines the denoising process.

The primary optimization objective is to approximate the intractable true reverse transition $q(\bz_{t-1} \mid \bz_t)$ via a parameterized model $p_{\btheta}(\bz_{t-1} \mid \bz_t)$ with learnable parameters $\btheta$.
Since $q(\bz_{t-1} \mid \bz_t)$ lacks a closed-form solution, we approximate the data-dependent reverse process $q(\bz_{t-1} \mid \bz_t, \bx_0)$, which is a conditional Gaussian distribution with mean $\bmu^q_t(\bx_0, \bz_t)$ and diagonal covariance $\bSigma^q_t$, as derived in \eqref{equation:ddpm_reverse_conditional_gaussian}:
\begin{equation}
q(\bz_{t-1} \mid \bz_t, \bx_0) \sim \normal(\bz_{t-1}\mid  \bmu^q_t(\bx_0, \bz_t), \bSigma^q_t),
\quad\text{where $\bSigma^q_t 
= (\sigma^q_t)^2 \bI_D$}.
\end{equation}
The parameterized model learns the distribution $p_\btheta(\bz_{t-1} \mid \bz_t)$ to fit the true conditional distribution $q(\bz_{t-1} \mid \bz_t, \bx_0)$. In practice, model training focuses on fitting the mean term $\bmu^q_t(\bx_0, \bz_t)$ of the true distribution. This mean term can be rearranged into three mathematically equivalent forms, which correspond to the three core learning formulations of diffusion models introduced earlier and we will conclude below.

From the perspective of maximum likelihood estimation, we aim to maximize the log-likelihood of observed real data $\ln p_{\btheta}(\bx_0)$. Notably, only $\bx_0$ is observable among all $T+1$ random variables in the diffusion graphical model. Thus, we marginalize the joint distribution $p_{\btheta}(\bx_0, \bz_{1:T})$ over all latent variables to obtain the marginal data distribution $p_{\btheta}(\bx_0)$:
\begin{equation}\label{equation:ddim_loglike}
\ln p_{\btheta}(\bx_0) = \ln \int p_{\btheta}(\bx_0, \bz_{1:T}) \diff\bz_{1:T} .
\end{equation}
The unobserved variables $\bz_{1:T}$ are treated as latent variables, forming a standard latent variable model estimation problem (refer to Section~\ref{section:lvm}). As shown in \eqref{equation:ddim_loglike}, marginalization over latent variables introduces an intractable integral in the log-likelihood function, preventing direct gradient computation and maximization. To resolve this issue, we apply Jensen's inequality to derive the evidence lower-bound (ELBO) as a surrogate optimization objective parameterized by $\btheta$, see \eqref{equation:ddom_elbo_final}:
\begin{align}
\ln &p_\btheta(\bx_0) 
= \ln \int p_\btheta(\bx_0, \bz_{1:T}) \diff\bz_{1:T}
= \ln \int \frac{p_\btheta(\bx_0, \bz_{1:T}) q(\bz_{1:T}\mid\bx_0)}{q(\bz_{1:T}\mid\bx_0)} \diff\bz_{1:T} \nonumber  \\
&= \ln \Exp_{q(\bz_{1:T}|\bx_0)} \left[ \frac{p_\btheta(\bx_0, \bz_{1:T})}{q(\bz_{1:T}\mid\bx_0)} \right] 
\geq \Exp_{q(\bz_{1:T}|\bx_0)} \left[ \ln \frac{p_\btheta(\bx_0, \bz_{1:T})}{q(\bz_{1:T}\mid\bx_0)} \right] \nonumber\\
&= \Exp_{q(\bz_1|\bx_0)} \left[ \ln p_\btheta(\bx_0\mid\bz_1) \right] - \sum_{t=2}^T \Exp_{q(\bz_t|\bx_0)} \left[ \KL \left( q(\bz_{t-1}\mid\bz_t,\bx_0) \parallel p_\btheta(\bz_{t-1}\mid\bz_t) \right) \right] +\text{const}.
\label{equation:ddim_ddpm_elbo}
\end{align}
Under standard convergence conditions, maximizing this ELBO is equivalent to maximizing the original log-likelihood function.
By substituting and simplifying all ELBO terms, the final training objective reduces to a simple mean squared error (MSE) loss $\mathcalJ_\bzeta$, where $\zeta_t$ denotes constant scaling factors that do not affect optimization convergence and is usually set to 1 (cf. \eqref{equation:ddpm_equaweight_sum_loss2}):
\begin{mybox}
\begin{equation}\label{equation:ddim_ddpmloss}
\mathcalJ_\bzeta(\btheta) \triangleq \sum_{t=1}^T \zeta_t \Exp_{q(\bz_t|\bx_0)} \left[ \normtwo{\bepsilon_t - \bepsilon^\btheta_t(\bz_t)}^2 \right], \quad \bepsilon_t \sim \normal(\bzero, \bI_D). 
\end{equation}
\end{mybox}
For this objective, we provide three different perspectives below.

\paragrapharrow{First form: learning the initial sample.}
The mean $\bmu^q_t(\bx_0, \bz_t)$ of the reverse conditional distribution is given by \eqref{eq:mean_reverse}:
\begin{equation}
\bmu^q_t(\bx_0, \bz_t) = \frac{(1-\alpha_{t-1})\sqrt{1-\beta_t} \bz_t + \sqrt{\alpha_{t-1}}\beta_t \bx_0}{1-\alpha_t}.
\end{equation}
The mean $\bmu^\btheta_t$ of the parameterized distribution $p_\btheta(\bz_{t-1} \mid \bz_t)$  is then defined accordingly:
\begin{equation}
\bmu^\btheta_t(\bz_t) 
= \frac{(1-\alpha_{t-1})\sqrt{1-\beta_t} \bz_t + \sqrt{\alpha_{t-1}}\beta_t \widehatbx^\btheta_t(\bz_t)}{1-\alpha_t}.
\end{equation}
Under this formulation, the model effectively learns to predict the initial sample $\bx_0\approx \widehatbx^\btheta_1(\bz_1)$. This task is notably challenging for the model, resulting in relatively modest performance.

\paragrapharrow{Second form: learning noise.}
A second expression for the mean $\bmu^q_t(\bx_0, \bz_t)$ is provided in \eqref{equation:ddpm_mu_in_terms_of_epsilon}:
\begin{equation}
\bmu^q_t(\bx_0, \bz_t) =  \frac{1}{\sqrt{1 - \beta_t}} \left\{ \bz_t - \frac{\beta_t}{\sqrt{1 - \alpha_t}} \bepsilon_t \right\},
\end{equation}
where $\bepsilon_t$  represents the cumulative noise applied to the sample $\bx_0$ at timestep $t$, rather than just the incremental noise added at timestep $t$; see \eqref{equation:diffusion_kernel_zt_form_ddpm}.
Correspondingly, the mean $\bmu^\btheta_t(\bz_t)$ of the parameterized distribution $p_\btheta(\bz_{t-1} \mid \bz_t)$ is defined as:
\begin{equation}
\bmu^\btheta_t(\bz_t) =  \frac{1}{\sqrt{1 - \beta_t}} \left\{ \bz_t - \frac{\beta_t}{\sqrt{1 - \alpha_t}} \bepsilon^\btheta_t(\bz_t) \right\},
\end{equation}
where $\bepsilon^\btheta_t(\bz_t)$ predicts the total noise corrupting the original sample $\bx_0$.
In this setup, the model learns to predict the noise $\bepsilon_t$. Predicting noise at each step is substantially easier than directly estimating the initial sample $\bx_0$, leading to significantly improved model performance.

\paragrapharrow{Third form: learning the score.}
Using  Tweedie's formula, the mean $\bmu^q$ can be expressed in a score-based form as shown in \eqref{equation:mux0zt_score}:
\begin{equation}
\bmu^q_t(\bx_0, \bz_t) = \frac{1}{\sqrt{1-\beta_t}} \bz_t + \frac{\beta_t}{\sqrt{1-\beta_t}} \nabla \ln p(\bz_t).
\end{equation}
Accordingly, the mean of the parameterized distribution is defined as:
\begin{equation}
\bmu^\btheta_t(\bz_t) = \frac{1}{\sqrt{\alpha_t}} \bz_t + \frac{\beta_t}{\sqrt{1-\beta_t}} \bs^\btheta_t(\bz_t).
\end{equation}
Here, the model learns to predict the score (log-likelihood gradient) $\nabla \ln p(\bz_t)$. Relative to noise prediction, score-based learning offers the key advantage that it enables the use of score-based sampling algorithms during reverse-process data generation. 
A wide variety of score-based sampling methods exist, which greatly enhances overall algorithmic flexibility.

\index{Denoising diffusion implicit models}
\index{DDIMs}
\section{Denoising Diffusion Implicit Models (DDIMs)}\label{section:ddims}

In DDPMs, the data generation process is formulated as the reverse of a Markovian diffusion process. At each step of reverse sampling, the model predicts the injected noise (see Section~\ref{section:ddpm_rev}).
However, \citet{song2020denoising} demonstrated that diffusion processes are not strictly required to follow Markovian assumptions, leading to the proposal of \textit{denoising diffusion implicit models (DDIMs)}. Subsequent studies on score-based diffusion models and stochastic differential equation (SDE)-based diffusion theories have independently validated this conclusion (see Section~\ref{section:score_mat}).

\subsection{Non-Markovian Forward Process}
The DDPM training loss formulated in \eqref{equation:ddpm_equaweight_sum_loss2} or \eqref{equation:ddim_ddpmloss} indicates that model training relies on single-timestep noised sample construction. Specifically, given a clean data $\bx_0$, a noise vector $\bepsilon_t\sim\normal(\bzero, \bI_D)$ is sampled to generate the noised feature $\bz_t = \sqrt{\alpha_t}\bx_0 + \sqrt{1-\alpha_t} \bepsilon_t$ (see \eqref{equation:ddpm_diffusion_kernel_all}). This construction only utilizes the marginal distribution $q(\bz_t\mid \bx_0)$, which characterizes the direct mapping between a clean data and its noisy counterpart at a single diffusion level $t$.
Intermediate latent states along the Markov chain ($\bz_1, \bz_2, \dots, \bz_{t-1}$) are never involved in training. The network is exclusively trained on paired data consisting of original clean data and their corresponding noised versions at arbitrary timesteps $t$, with the sole objective of noise prediction.

Accordingly, the trained diffusion network acts as a standalone function that takes a marginal sample from $q(\bz_t \mid \bx_0)$ and a timestep index $t$ as inputs to estimate the injected noise (or equivalently, to recover the original data $\bx_0$ or compute the score function). Critically, the training objective neither incorporates nor depends on the Markovian nature of the forward diffusion chain. Although the Markovian forward process serves as the theoretical basis for loss derivation, it leaves no inherent constraint or structural fingerprint on the final loss formulation.
This implies the training objective does not uniquely necessitate the Markovian forward process assumed by standard DDPMs.

To address this, DDIM constructs a family of \textit{non-Markovian forward diffusion processes} that exactly preserve the marginal distributions of DDPMs. Due to identical marginal constraints, pre-trained DDPM models can be directly adopted for DDIM sampling without additional fine-tuning or retraining.

\paragrapharrow{Defining DDIMs.}
Based on the above analysis, eliminating the Markovian assumption while preserving full consistency with standard DDPMs (i.e., retaining the original DDPM training objective) only requires invariance of the conditional distributions $q(\bz_t\mid\bx_0)$. We next formalize the construction of the non-Markovian forward diffusion model for DDIMs.

We reparameterize the joint distribution $q(\bz_{1:T}\mid\bx_0)$ by introducing a free hyperparameter $\sigma^2$, which controls the variance of the conditional transition distribution $q_\sigma(\bz_{t-1}\mid\bz_t,\bx_0)$. Detailed analysis of $\sigma^2$ is provided in subsequent discussions.
The joint distribution of the non-Markovian forward process is defined as:
\begin{equation}\label{equation:ddim_qsigma_z1T_x0}
q_\sigma(\bz_{1:T}\mid\bx_0) \triangleq q_\sigma(\bz_T\mid\bx_0) \prod_{t=2}^T q_\sigma(\bz_{t-1}\mid\bz_t,\bx_0), 
\end{equation}
where $T$ denotes the total number of diffusion steps. We enforce the terminal marginal distribution to be identical to that of standard DDPMs, as specified in \eqref{equation:ddpm_diffusion_kernel_all} or  \eqref{equation:ddim_ddpm_ztx0}:
\begin{equation}\label{equation:ddim_ztx0_sig_final}
\textbf{(Terminal marginal)}\qquad \qquad
q_\sigma(\bz_T\mid\bx_0) = \normal\left(\bz_T\mid \sqrt{\alpha_T} \bx_0, (1 - \alpha_T) \bI_D\right). 
\end{equation}
For all timesteps $t>1$, DDIM directly defines the backward transition kernel $q_\sigma(\bz_{t-1}\mid\bz_t,\bx_0)$ without inverting a Markovian diffusion chain, as follows (cf. \eqref{equation:ddpm_reverse_conditional_gaussian} in DDPMs):
\begin{equation}\label{equation:ddim_ztmin_ztx0_sig}
\textbf{(Reverse kernel)}\qquad \qquad
q_\sigma(\bz_{t-1}\mid\bz_t,\bx_0) 
\triangleq 
\normal\left(\bz_{t-1}\mid 
\bmu_t^\sigma(\bx_0, \bz_t),
\bSigma_t^\sigma
\right)
\end{equation}
where the mean and covariance terms are given by:
\begin{align}
\bmu_t^\sigma(\bx_0, \bz_t) 
&= {\sqrt{\alpha_{t-1}} \bx_0 + \sqrt{1 - \alpha_{t-1} - \sigma_t^2} \cdot \frac{\bz_t - \sqrt{\alpha_t} \bx_0}{\sqrt{1 - \alpha_t}}}, \\
\bSigma_t^\sigma 
&= { \sigma_t^2  \bI_D}.
\end{align}
This formulation is explicitly designed such that valid marginal distributions $q(\bz_{t-1} \mid \bx_0)$ are guaranteed for all $\sigma_t$ values, provided that $\bz_t$ adheres to the standard DDPM marginal distribution.
The following theorem verifies the universal marginal consistency across all diffusion steps:
\begin{theoremHigh}\label{theorem:ddim_ztx0_t}
Using the definitions~\eqref{equation:ddim_ztx0_sig_final} and \eqref{equation:ddim_ztmin_ztx0_sig}, the following transition kernel holds for all $1 \leq t \leq T$:
\begin{equation}\label{equation:ddim_ztx0_t}
\textbf{(Marginal)}\qquad \qquad
q_\sigma(\bz_t\mid\bx_0) = \normal\left(\bz_t\mid \sqrt{\alpha_t} \bx_0, (1 - \alpha_t) \bI_D\right).
\end{equation}
\end{theoremHigh}

\begin{proof}[of Theorem~\ref{theorem:ddim_ztx0_t}]
Assume for any $1<t \leq T$, $q_\sigma(\bz_t\mid\bx_0) = \normal(\sqrt{\alpha_t}\bx_0, (1 - \alpha_t)\bI_D)$ holds, if:
$$
q_\sigma(\bz_{t-1}\mid\bx_0) = \normal(\sqrt{\alpha_{t-1}}\bx_0, (1 - \alpha_{t-1})\bI_D),
$$
then we can prove the statement with an induction argument for $t$ from $T$ to $1$, since the base case ($t = T$) already holds.
To see this, we have 
$$
q_\sigma(\bz_{t-1}\mid\bx_0) = \int q_\sigma(\bz_t\mid\bx_0) \cdot q_\sigma(\bz_{t-1}\mid\bz_t, \bx_0) \diff\bz_t,
$$
where we  assume $q_\sigma(\bz_t\mid\bx_0) \sim \normal(\sqrt{\alpha_t}\bx_0, (1 - \alpha_t)\bI_D)$ and $q_\sigma(\bz_{t-1}\mid\bz_t, \bx_0)$  is given by \eqref{equation:ddim_ztmin_ztx0_sig}.
By linear Gaussian models (Problem~\ref{prob:linear_gaus_model}), we have   $q_\sigma(\bz_{t-1}\mid\bx_0) = \normal(\sqrt{\alpha_{t-1}}\bx_0, (1 - \alpha_{t-1})\bI_D)$.
The proof follows by induction.
\end{proof}

Equation \eqref{equation:ddim_ztx0_t} holds for all $1\leq t\leq T$, confirming that the DDIM joint distribution strictly matches the marginal distributions of standard DDPMs. The forward transition kernel can be further derived via Bayes' rule \eqref{equation:bayes_base}:
\begin{equation}
q_{\sigma}(\bz_t \mid \bz_{t-1}, \bx_0) 
= \frac{q_{\sigma}(\bz_{t-1} \mid \bz_t, \bx_0) \, q_{\sigma}(\bz_t \mid \bx_0)}{q_{\sigma}(\bz_{t-1} \mid \bx_0)},
\end{equation}
which also yields a valid Gaussian distribution.

Unlike the strictly Markovian forward process adopted in conventional DPMs and DDPMs \eqref{equation:forward_joint_ddpm}, the DDIM forward process is non-Markovian, as each latent state $\bz_t$ depends on both the preceding state $\bz_{t-1}$ and the original clean data $\bx_0$ (during sampling, $\bx_0$ is replaced by the model's predicted estimate $\widehatbx_0$).  
This breaks the single-step conditional independence constraint inherent to Markovian diffusion chains. 
The hyperparameter $\sigma_t$ governs the stochasticity of the forward diffusion process. In the limiting case where $\sigma_t \rightarrow 0$, the diffusion process becomes fully deterministic: given $\bx_0$ and any intermediate state $\bz_t$, all previous latent states $\bz_{t-1}$ are uniquely determined with no randomness.

\paragrapharrow{Reverse process and objective function.}
We next define a trainable generative process $p_\btheta(\bx_0, \bz_{1:T})$, in which each transition $p_\btheta(\bz_{t-1} \mid \bz_t)$ incorporates the structural formulation of the conditional distribution $q_\sigma(\bz_{t-1} \mid \bz_t, \bx_0)$.
Consistent with standard DPMs and DDPMs, our goal is to approximate the true data-dependent conditional distribution $q(\bz_{t-1} \mid \bz_t, \bx_0)$ (see Sections~\ref{section:dpm_approx} and \ref{section:ddpm_rev}).
Intuitively, given a noisy latent sample $\bz_t$, the model first estimates the corresponding clean data $\bx_0$, and then generates the previous latent state $\bz_{t-1}$ via the predefined reverse conditional distribution $q_\sigma(\bz_{t-1} \mid \bz_t, \bx_0)$.

For a clean data $\bx_0 \sim p_{\text{data}}(\bx_0)$ sampled from the real data distribution and Gaussian noise  $\bepsilon_t \sim \normal(\bzero, \bI_D)$, the noisy latent $\bz_t$ can be constructed following either \eqref{equation:diffusion_kernel_zt_form_ddpm} or \eqref{equation:ddim_ddpm_ztx0_prime}. 
The standard DDPM model $\bepsilon^\btheta_t(\bz_t)$ predicts the accumulated noise $\bepsilon_t$ directly from $\bz_t$ without access to the clean data $\bx_0$. By rearranging the aforementioned diffusion kernel formulations, we obtain an explicit estimate of the denoised clean data from $\bz_t$, denoted as $\widehatbx_0$:
\begin{equation}\label{equation:ddim_wihatbx_0}
\bz_t = \sqrt{\alpha_t} \bx_0 + \sqrt{1 - \alpha_t} \bepsilon_t
\quad\implies\quad 
\widehatbx_0\triangleq f^\btheta_t(\bz_t) = \frac{\bz_t - \sqrt{1-\alpha_t}\bepsilon^\btheta_t(\bz_t)}{\sqrt{\alpha_t}}.
\end{equation}
With this denoising estimator, we formalize the full generative process with a fixed standard Gaussian prior $p(\bx_T)=\normal(\bzero, \bI_D)$. The conditional transition distributions are defined as:
\begin{equation}\label{equation:ddim_ztm_zt_theta}
p_{\btheta}(\cdot)=
\begin{cases}
p_{\btheta}(\bx_0\mid\bz_1) =\normal\left(f^\btheta_1(\bz_1),\textcolor{black}{\sigma_1^2} \bI_D\right), & \text{if } t=1; \\
p_{\btheta}(\bz_{t-1}\mid\bz_t) =q_\sigma(\bz_{t-1}\mid\bz_t, f^\btheta_t(\bz_t)), & \text{if } 1 < t \leq T,
\end{cases} 
\end{equation}
where the conditional term $q_\sigma(\bz_{t-1}\mid\bz_t, f^\btheta_t(\bz_t))$ follows the definition in  \eqref{equation:ddim_ztmin_ztx0_sig}.
For the final generation step $t=1$, we add Gaussian noise with covariance $\sigma_1^2\bI_D$ to ensure full support of the generative distribution over the entire data space.

Following the variational inference framework established in Sections~\ref{section:diff_elbo} and \ref{section:ddpm_rev}, we optimize the model parameters $\btheta$ via a tailored variational objective functional parameterized by the noise prediction network $\bepsilon_t^\btheta$:
\begin{mybox}
\begin{align}
&\mathcalJ_\sigma(\btheta) 
= -\mathcalF_{\sigma}(\btheta) 
=\Exp_{q_{\sigma}}\big[\ln q_\sigma(\bz_{1:T} \mid \bx_0) - \ln p_\btheta(\bx_0, \bz_{1:T})\big]  \nonumber\\
&= \Exp_{q_{\sigma}}\left[\ln q_\sigma(\bz_T \mid \bx_0) 
+ \sum_{t=2}^T \ln \left(
\frac{q_\sigma(\bz_{t-1} \mid \bz_t, \bx_0)}{p_{\btheta, \sigma}(\bz_{t-1} \mid \bz_t) } \right) 
-  \ln p_{\btheta, \sigma}(\bx_0 \mid \bz_1) 
- \ln p(\bz_T)\right]
\end{align}
\end{mybox}
with the expectation operator defined as 
\begin{equation}
\Exp_{q_{\sigma}}[\cdot]=
\Exp_{q_\sigma(\bz_{1:T}\mid\bx_0)}[\cdot],
\end{equation}
where we factorize $q_\sigma(\bz_{1:T} \mid \bx_0)$ according to \eqref{equation:ddim_qsigma_z1T_x0} and $p_\btheta(\bx_0, \bz_{1:T})$ according to \eqref{equation:ddpm_reverse_chain}.

Intuitively, different choices of the hyperparameter sequence $\{\sigma_t\}$ yield distinct variational objectives and corresponding generative processes, which seemingly require independent model training. However, the objective $\mathcalJ_\sigma(\btheta)$ can be exactly mapped to the DDPM loss formulation $\mathcalJ_\bzeta(\btheta)$ defined in \eqref{equation:ddim_ddpmloss} via appropriate reweighting. Specifically, for any valid $\sigma_t > 0$, there exist positive real weights $\bzeta \in \real_{++}^T$ and a constant offset $C \in \real$ such that $\mathcalJ_\sigma(\btheta) = \mathcalJ_\bzeta(\btheta) + C$.
Notably, when $\sigma_t$ is set to the specific form:
\begin{equation}\label{equation:ddim_be_markov}
\sigma_t =  \sqrt{1 - \frac{\alpha_t}{\alpha_{t-1}}} \sqrt{\frac{1 - \alpha_{t-1}}{1 - \alpha_t}}, 
\quad\text{for all $t$},
\end{equation}
$\bmu_t^\sigma(\bx_0, \bz_t) \equiv\bmu_t^q $ in \eqref{eq:mean_reverse}, and $\sigma_t\equiv \sigma^q_t$ in \eqref{eq:variance_reverse}; 
thus, $q_\sigma (\bz_{t-1}\mid \bz_t, \bx_0) \equiv q (\bz_{t-1}\mid \bz_t, \bx_0)$, the true data-conditioned distribution from \eqref{equation:ddpm_reverse_conditional_gaussian}.
The forward process becomes Markovian, and the generative process becomes a DDPM.

\paragrapharrow{Reusing DPMs or DDPMs.}	
Following the reverse conditional definition in  \eqref{equation:ddim_ztm_zt_theta}, the explicit sampling formula for generating $\bz_{t-1}$ is derived as follows:
\begin{equation}\label{equation:ddim_samp}
\begin{aligned}
\bz_{t-1} &= \sqrt{\alpha_{t-1}}\widehatbx_0 + \sqrt{1-\alpha_{t-1}-\sigma_t^2} \cdot \frac{\bz_t - \sqrt{\alpha_t}\widehatbx_0}{\sqrt{1-\alpha_t}} + \sigma_t\bepsilon_t^* \\
&= \sqrt{\alpha_{t-1}} \underbrace{\left( \frac{\bz_t - \sqrt{1-\alpha_t}\bepsilon^\btheta_t(\bz_t)}{\sqrt{\alpha_t}} \right)}_{\text{Predicted } \bx_0}
+ \underbrace{\sqrt{1-\alpha_{t-1}-\sigma_t^2}\bepsilon^\btheta_t(\bz_t)}_{\text{Direction pointing to } \bz_t}
+ \underbrace{\sigma_t\bepsilon_t^*}_{\text{Random noise}},
\end{aligned} 
\end{equation}
where $\bepsilon_t^* \sim \normal(\bzero, \bI_D)$ denotes auxiliary Gaussian noise, and the initial scheduling parameter satisfies $\alpha_0=1$.
Since the marginal distributions in \eqref{equation:ddim_ztx0_t} are strictly consistent with those of standard DDPMs, the noise-prediction network trained for DDPMs is inherently valid for the DDIM generative process. As the DDPM training objective depends solely on timestep-wise marginal distributions rather than chain-specific transitional properties, the learned network function remains fully compatible. This explains the zero-retraining property of DDIM: DDIM does not require model fine-tuning, as it implements a novel sampling algorithm built upon the pre-trained DDPM inference function.
Different configurations of $\sigma_t$ yield distinct generative trajectories while reusing the fixed DDPM noise-prediction model $\bepsilon^\btheta_t$, eliminating the need for parameter retraining.

In summary, the non-Markovian diffusion construction delivers two core advantages.
First, the sequence $\{\sigma_t\}$ constitutes a set of free hyperparameters. Different $\sigma_t$ choices produce distinct reverse diffusion processes while preserving identical marginal distributions to DDPMs. A specific $\sigma_t$ configuration exactly recovers the stochastic sampling scheme of vanilla DDPMs, whereas setting $\sigma_t = 0$ eliminates stochastic noise injection and yields a fully deterministic sampling process.

Second, the DDIM formulation is defined based on independent marginal distributions at individual timesteps, rather than adjacent transitional dependencies in Markov chains. This allows sampling to be performed over arbitrary increasing timestep subsequences (e.g., $0, 50, 100, \dots, 1000$), enabling direct sampling along heavily shortened diffusion schedules. All intermediate updates remain marginally consistent with the original diffusion process, which suppoAmenabilityrts sampling speedups of one to two orders of magnitude without retraining. Unlike rigid Markovian diffusion chains that require consecutive timestep transitions, the non-Markovian DDIM framework removes adjacent-step constraints and only enforces marginal distribution consistency.

\subsection{Accelerated Sampling}

The non-Markovian nature of DDIM diffusion enables flexible acceleration of the sampling procedure. As discussed above, the DDPM training objective depends exclusively on per-timestep marginal distributions and is agnostic to the chain-specific transition order of the forward process. Consequently, pre-trained DDPM models can be generalized to forward processes with fewer diffusion steps, yielding faster generative sampling pipelines without additional model training or fine-tuning.

Instead of constructing the forward diffusion process over the full sequence of latent variables $\bz_{1:T}$, we consider a reduced process defined on a strictly increasing timestep subsequence $\sS = \{\sS_1, \sS_2, \dots, \sS_S\}$ sampled from the full range $[1,2,\dots,T]$, where $S \ll T$ denotes the truncated sequence length. We enforce each subsampled latent state to preserve the exact marginal distribution of the original DDPM process:
$$
q(\bz_{\sS_i} \mid \bx_0) = \normal\big(\sqrt{\alpha_{\sS_i}} \bx_0, (1 - \alpha_{\sS_i}) \bI_D\big).
$$
The generative sampling procedure is then conducted over the reversed subsequence of selected timesteps, defined as the sampling trajectory. By adopting a substantially shortened sampling trajectory, we significantly reduce the total number of iterative update steps, thereby improving the computational efficiency of image generation.

Furthermore, the reverse diffusion update can be reinterpreted from a score-based perspective. Following the theoretical derivation in Section~\ref{section:score_ddpm}, the log-likelihood gradient of the latent variable $\bz_t$ is linearly correlated with the model's predicted noise:
$$
\nabla \ln p(\bz_t) = -\frac{1}{\sqrt{1-\alpha_t}} \bepsilon^\btheta_t(\bz_t) .
$$
This score function allows us to reformulate the predicted clean observation $\widehatbx_0$ by substituting the noise term with the log-probability gradient, as derived in \eqref{equation:ddpm_x0_score}:
\begin{equation}\label{equation:ddim_wihatbx_0_score}
\widehatbx_0 = \frac{\bz_t + (1-\alpha_t)\nabla \ln p(\bz_t)}{\sqrt{\alpha_t}} .
\end{equation}
Substituting this score-based clean image estimate into the DDIM iterative sampling formula \eqref{equation:ddim_samp} yields a gradient-descent-style update rule for the latent state $\bz_{t-1}$:
\begin{equation}\label{equation:ddim_descent}
\small
\begin{aligned}
&\bz_{t-1} 
= \sqrt{\alpha_{t-1}} \widehatbx_0 + \sqrt{1-\alpha_{t-1}-\sigma_t^2} \cdot \frac{\bz_t - \sqrt{\alpha_t}\widehatbx_0}{\sqrt{1-\alpha_t}} + \sigma_t\bepsilon_t^*  \\
&= \frac{\bz_t}{\sqrt{\beta_t}} 
+ \left[ \frac{\sqrt{\alpha_{t-1}}(1-\alpha_t)}{\sqrt{\alpha_t}}  - \sqrt{1-\alpha_{t-1}-\sigma_t^2}\sqrt{(1-\alpha_t)} \right] \nabla \ln p(\bz_t)
+ \sigma_t\bepsilon_t^* +C \\
&\triangleq A\bz_t + B\nabla \ln p(\bz_t) + C + \sigma_t\bepsilon_t^*,
\end{aligned}
\end{equation}
where $C$ denotes a constant bias term aggregated from algebraic rearrangement.
This formulation reveals that the iterative denoising update in reverse diffusion generation essentially performs iterative updates along the log-likelihood ascent direction of the latent samples, which is analogous to parameter optimization via gradient descent. This key insight enables the integration of advanced optimization techniques into diffusion sampling, including adaptive learning rate strategies and mainstream optimization algorithms such as Adam \citep{kingma2014adam}, to further optimize sampling performance and convergence.

\index{Image super-resolution}
\index{Guided diffusion models}
\index{Cascaded diffusion pipelines}
\section{Guidance: Condition on a Prompt}\label{section:guid_ddpm}

Thus far, we have framed diffusion models as tools for approximating the \textit{unguided (unconditional) data density} $p_{\btheta}(\bx)\approx p_{\text{data}}(\bx)$, which is trained on samples $\mathcalX=\{\bx_1, \bx_2, \ldots, \bx_N\}$ independently drawn from the underlying true data distribution $p_{\text{data}}(\bx)$. 
Once training is complete, the fitted model can generate new synthetic data (e.g., synthetic images) by sampling from this learned distribution $p_{\btheta}(\bx)$.
Both DDPM and DDIM adhere to the same generative paradigm for  data synthesis: they take randomly sampled Gaussian noise as input and iteratively refine this noisy input into coherent, realistic outputs such as natural images. 
Since each denoising step constitutes a stochastic transformation, independent sampling runs produce distinct data, endowing generated samples with strong diversity. However, this stochasticity introduces a key limitation: the generation process lacks controllability, meaning the sampling pipeline cannot be systematically directed to produce data content aligned with custom target objectives.

In most practical applications, by contrast, sampling is performed from a \textit{guided (conditional) distribution} $p(\bx \mid \by)$ where the guidance variable $\by$ may correspond to a class label, textual description, or other desired  attribute. 
This conditional formulation underpins numerous downstream applications, including text-guided diffusion generation, image super-resolution, and image inpainting, among others. Representative use cases are detailed below:
\begin{itemize}
\item \textit{Text-guided diffusion models.} 
These models integrate techniques from large language models to accept general text sequences---referred to as \textit{prompts}---as guidance input, rather than restricting guidance to predefined class labels. Text prompts modulate the denoising process through two core mechanisms: first, by concatenating latent representations extracted from transformer-based language models \citep{vaswani2017attention} with the input to the denoising network; second, by enabling cross-attention layers within the denoising network to attend explicitly to text token sequences. See Chapter~\ref{chapter:diffarchitect} for more details.

\item \textit{Image super-resolution.} Guided diffusion models also excel at image super-resolution, which maps low-resolution inputs to corresponding high-resolution counterparts. This task is inherently an ill-posed inverse problem, as multiple valid high-resolution images can correspond to a single low-resolution input. Diffusion-based super-resolution achieves high-fidelity upsampling by denoising high-resolution Gaussian samples conditioned on low-resolution guidance images \citep{saharia2022image}.

\item \textit{Cascaded diffusion pipelines.} Cascaded diffusion pipelines further enhance resolution capabilities, enabling ultra-high-resolution synthesis (e.g., progressive upsampling from $64 \times 64$ to $256 \times 256$, then to $1024 \times 1024$) \citep{ho2022cascaded}. Each stage typically employs a U-Net architecture \citep{ronneberger2015u}, where each U-Net is conditioned on the fully denoised output of the preceding stage.

\item This cascaded strategy is also applicable to standard \textit{image-generation} diffusion models. In this framework, initial denoising occurs at a lower resolution, and a dedicated upsampling network (which may additionally accept text prompts) subsequently upsamples the low-resolution output to produce high-resolution final results \citep{nichol2021glide, saharia2022photorealistic}. This approach serves as an alternative to the latent diffusion models (LDMs) introduced in Section~\ref{section:ldm_intro} and substantially reduces computational overhead. Because denoising requires hundreds of forward passes through the network, operating on low-resolution image spaces avoids the high costs of direct high-dimensional denoising. Notably, these methods still operate directly in the pixel image space rather than latent space.
\end{itemize}
To address the limited controllability of unconditional diffusion models, extensive prior work has developed effective strategies for steering the data generation process. The core intuition is to inject auxiliary guidance information into the diffusion sampling process to regulate model behavior. We denote this guidance information as $\by$, which can take the form of text prompts, reference images, or categorical class labels. With the introduction of $\by$, the model characterizes a conditional diffusion distribution:
$$
p(\bz_{1:T} \mid\by, \bx_0).
$$

\begin{remark}[Terminology for conditional and guided]
To avoid notation and terminology conflicts arising from the term ``conditional"---which conventionally refers to conditioning on the data or latent distribution (e.g., the reverse approximation $ p_{\btheta}(\bz_{t-1}\mid \bz_t)$), we strictly reserve the term ``\textit{guided}" exclusively for conditioning on external guidance variables  $\by$, such as text prompts.
\end{remark}

\paragrapharrow{Forward process with guidance.}
We now analyze how the guidance variable $\by$ affects the forward and reverse diffusion processes.
We define the hat notation $ \widehatq$ to denote probability distributions associated with the \textit{guided diffusion model} conditioned on $\by$, while the unhatted notation $q$ denotes distributions of the \textit{original (unguided) diffusion model}.
We first consider a single diffusion timestep $t$. Diffusion models form Markov chains, meaning the latent state $\bz_t$ depends only on the previous state $\bz_{t-1}$. Consequently, the forward noising transition at each timestep is independent of external guidance $\by$, yielding the key identity:
\begin{equation}
(\text{A5}) \qquad \widehatq(\bz_t \mid \bz_{t-1}, \by) \equiv q(\bz_t \mid \bz_{t-1}) .
\end{equation}
From this property, the marginal forward transition of the guided model is also independent of $\by$, as derived below:
\begin{align}
\widehatq(\bz_t \mid \bz_{t-1}) 
&= \int \widehatq(\bz_t, \by \mid \bz_{t-1}) \diff \by 
= \int \widehatq(\bz_t \mid\by, \bz_{t-1}) \widehatq(\by \mid \bz_{t-1}) \diff \by \nonumber \\
&= \int q(\bz_t \mid \bz_{t-1}) \widehatq(\by \mid \bz_{t-1}) \diff \by 
= q(\bz_t \mid \bz_{t-1}) 
= \widehatq(\bz_t \mid \bz_{t-1}, \by).
\end{align} 
By extension, the full joint forward distribution of the guided diffusion model $\widehatq(\bz_{1:T} \mid \bx_0)$ matches that of the unguided baseline $q(\bz_{1:T} \mid \bx_0)$. 
The joint marginal over all guidance variables recovers the unguided forward chain:
\begin{align}
\widehatq(\bz_{1:T} \mid \bx_0) 
&= \int \widehatq(\bz_{1:T}, \by \mid \bx_0) \diff \by
= \int \widehatq(\by \mid \bx_0) \widehatq(\bz_{1:T} \mid \bx_0, \by) \diff \by \nonumber\\
&= \int \widehatq(\by \mid \bx_0) \prod_{t=2}^{T} \widehatq(\bz_t \mid \bz_{t-1}, \by) \widehatq(\bz_1 \mid \bx_0, \by) \diff \by \nonumber\\
&= \prod_{t=2}^{T} \widehatq(\bz_t \mid \bz_{t-1}) \widehatq(\bz_1 \mid \bx_0) 
= q(\bz_{1:T} \mid \bx_0).
\end{align} 
This derivation verifies that the inclusion of external guidance $\by$ exerts no influence on the forward diffusion process. The noising procedure remains identical for both guided and unguided diffusion models.

\paragrapharrow{Reverse process with guidance.}
We next examine the effect of $\by$ on the reverse generative sampling process, where guidance is intentionally introduced to control synthesis outcomes. For the original unguided diffusion model, the joint distribution of the reverse Markov chain is recalled from \eqref{equation:ddpm_reverse_chain}:
$$
p_\btheta(\bx_0, \bz_{1:T}) = p(\bz_T) \left\{\prod_{t=2}^{T} p_\btheta(\bz_{t-1} \mid \bz_t)\right\}   p_\btheta(\bx_0 \mid \bz_1)  .
$$
After incorporating the guidance variable $\by$, the conditional joint distribution for the guided reverse process becomes:
\begin{equation}
p_\btheta(\bx_0, \bz_{1:T} \mid\by) = p(\bz_T) \left\{\prod_{t=2}^{T} p_\btheta(\bz_{t-1} \mid \bz_t, \by)\right\}  p_\btheta(\bx_0 \mid \bz_1, \by) .
\end{equation}
The entire controllability of guided diffusion generation is exclusively embedded within the reverse denoising process. By conditioning the reverse transition distributions on $\by$, the model reshapes the trajectory of latent state restoration to align with external semantic or structural constraints, while preserving the diversity inherent to stochastic diffusion sampling. This core property explains why naive conditional injection---feeding $\by$ as a network input---can enable conditional generation in principle.

\paragrapharrow{Simple guidance approaches.}
Collectively, these results clarify the essential working mechanism of guided diffusion models: conditional generation does not alter data corruption, but steers data reconstruction. This insight lays the theoretical foundation for advanced guidance strategies, which are designed to amplify guidance signal influence during reverse sampling, balance conditional fidelity and sample diversity, and resolve the under-constrained optimization problem of naive conditional diffusion training.

The most straightforward strategy for implementing conditional guidance is to feed $\by$ as an additional network input and train the model on paired training data $\{\bx_n, \by_n\}_{n=1}^N$.
As summarized in Section~\ref{section:ddpm_final_remark}, diffusion probabilistic models admit three equivalent training interpretations:
\begin{enumerate}[(i)]
\item Directly predict the original data $\bx_0 $: $\widehatbx^\btheta_t(\bz_t) \simeq \bx_0 $.
\item Predict the noise data: $\bepsilon^\btheta_t(\bz_t) \simeq \bepsilon_t $.
\item Predict the score (gradient): $\bs^\btheta_t(\bz_t) \simeq \nabla_{\bz_t} \ln p(\bz_t) $.
\end{enumerate}
In the simplest guided formulation, the guidance variable $\by$ is appended to the network input in the same manner as the timestep embedding $t$. The three learning objectives are modified to accept conditioning:
\begin{enumerate}[(i)]
\item Directly predict the original data $\bx_0 $: $\widehatbx^\btheta_t(\bz_t\mid \textcolor{mylightbluetext}{\by}) \simeq \bx_0 $.
\item Predict the noise data: $\bepsilon^\btheta_t(\bz_t\mid \textcolor{mylightbluetext}{\by}) \simeq \bepsilon_t $.
\item Predict the score (gradient): $\bs^\btheta_t(\bz_t\mid \textcolor{mylightbluetext}{\by} ) \simeq \nabla_{\bz_t} \ln p(\bz_t ) $.
\end{enumerate}
Although this naive conditioning strategy is functional, it suffers from a critical limitation: the network may assign insufficient weight to the guidance signal or overlook it entirely during optimization. This motivates explicit mechanisms to modulate guidance strength, enabling a controllable trade-off between alignment to the guidance condition and sample diversity. Such explicit alignment constraints are formally termed \textit{guidance techniques}. Broadly speaking, guidance methods fall into two primary categories: those relying on an external classifier model and those that do not.

\index{Classifier guidance}
\subsection{Classifier Guidance}

\citet{dhariwal2021diffusion} proposed a framework for guiding image generation using image class labels, referred to as \textit{classifier guidance}. This technique substantially improved the quality of images synthesized by diffusion models, outperforming GANs in FID metrics (see Problem~\ref{prob:imp_arcendec}).
While the paper introduced several enhancements, including modifications to the U-Net architecture \citep{ronneberger2015u}, we focus exclusively on the guidance mechanism in this work.

As noted earlier, the predictive model used in the denoising step of a diffusion model can be interpreted from three perspectives; see Section~\ref{section:ddpm_final_remark}.
One interpretation is that the parameterized neural network predicts the log-gradient of $\bz_t$, also known as the score, which is expressed as:
$$
\bs^\btheta_t(\bz_t) \simeq \nabla_{\bz_t} \ln p(\bz_t) .
$$
When guidance information $\by$ is introduced, the original score $\nabla_{\bz_t} \ln p(\bz_t)$ becomes $\nabla_{\bz_t} \ln p(\bz_t \mid\by)$, defined as the \textit{score of the guided (diffusion) model}, evaluated at arbitrary noise levels $t$. 
Applying Bayes' theorem~\eqref{equation:bayes_base}, the score function of the guided diffusion model can be decomposed as follows:
\begin{align}
\underbrace{\nabla_{\bz_t} \ln p(\bz_t \mid\by)}_{\text{Guided score}} &= \nabla_{\bz_t} \ln \left( \frac{p(\bz_t) p(\by \mid \bz_t)}{p(\by)} \right) \nonumber \\
&= \nabla_{\bz_t} \ln p(\bz_t) + \nabla_{\bz_t} \ln p(\by \mid \bz_t) - \underbrace{\nabla_{\bz_t} \ln p(\by)}_{=0, \text{ independent of } \bz_t} \nonumber\\
&= \underbrace{\nabla_{\bz_t} \ln p(\bz_t)}_{\text{Unguided score}} + \underbrace{\nabla_{\bz_t} \ln p(\by \mid \bz_t)}_{\text{Adversarial gradient}}.  \label{equation:class_guid_decomp}
\end{align}
This expansion yields two distinct terms.
The first term on the right-hand side of~\eqref{equation:class_guid_decomp} is the standard \textit{unguided score function}, while the second term---called the \textit{adversarial gradient}---steers the denoising process toward maximizing the probability of the given label/guidance under the classifier model \citep{song2020score, dhariwal2021diffusion}.
Thus, the guided score $\nabla_{\bz_t} \ln p(\bz_t \mid\by)$ can be interpreted as the unguided score $\nabla_{\bz_t} \ln p(\bz_t)$ combined with an adversarial gradient derived from the classifier $p(\by \mid \bz_t)$.

The input to the classifier $p(\by \mid \bz_t)$ is not the clean data $\bx_0$, but the noisy latent variable $\bz_t$ at the corresponding timestep t.
This classifier must be trained independently prior to training the guided diffusion model. Specifically, during classifier training, $\bz_t$ generated via the forward noising process of the diffusion model (see \eqref{equation:ddpm_diffusion_kernel_all}) serves as the input.
In classifier guidance, the unguided diffusion model score is learned as previously derived, together with a classifier that accepts arbitrary noisy $\bz_t$ and predicts the guidance label $\by$. The overall pipeline consists of three steps:
\begin{enumerate}[(i)]
\item Learn the unguided diffusion model as derived earlier; see \eqref{equation:loss_score_ddpm_final}.
\item Pre-train a noise-conditioned classifier $p(\by \mid \bz_t)$, where $t$ indexes  different timesteps (i.e., different noise levels).
\item During sampling at each timestep $t$ of the original unguided diffusion model:
\begin{enumerate}
\item Compute the unguided score predicted by the diffusion model: $\bs^\btheta_t(\bz_t)$.
\item Obtain the noisy latent $\bz_t$ via forward process \eqref{equation:ddpm_diffusion_kernel_all}, then evaluate the adversarial gradient $p(\by \mid \bz_t)$: $\nabla_{\bz_t} \ln p(\by \mid \bz_t)$.
\item Compute the guided score: $\bs^\btheta_t(\bz_t, \by) = \bs^\btheta_t(\bz_t) + \nabla_{\bz_t} \ln p(\by \mid \bz_t)$.
\item Continue with the standard denoising procedure.
\end{enumerate}
\end{enumerate}
The adversarial gradient $\nabla_{\bz_t} \ln p(\by \mid \bz_t)$ from the noise classifier modulates the sampling process, directing denoising toward the target class $\by$ and producing data that closely align with the specified label. For instance, if $\by = \text{``horse"}$, the model generates horse images with improved realism. However, although classifier guidance enhances visual fidelity, it noticeably \textbf{reduces sample diversity}.

To enable more flexible control over the generation direction, the original work introduces a hyperparameter $\lambda$ as a weighting coefficient. This scalar adjusts the strength of class guidance, yielding the \textit{scaled guided diffusion score}:
\begin{equation}\label{equation:scaled_cond_score}
\widetilde{\nabla_{\bz_t}} \ln p(\bz_t \mid\by) = \nabla_{\bz_t} \ln p(\bz_t) + \lambda \nabla_{\bz_t} \ln p(\by \mid \bz_t),
\end{equation}
where $\lambda\geq 0$ is known as the \textit{guidance scale} (typically set to $\lambda>1$). 
Intuitively, when $\lambda = 0$, the guided model disregards label information entirely. As $\lambda$ increases, the model prioritizes adherence to the guidance signal, which comes at the expense of sample diversity, as it tends to produce only samples that easily reconstruct the given label even under significant noise.
Notably, this formulation is \textbf{heuristic}: for $\lambda \neq 1$, we have $\widetilde{\nabla_{\bz_t}} \ln p(\bz_t \mid\by)  \neq {\nabla_{\bz_t}} \ln p(\bz_t \mid\by) $, meaning it does not represent the ``true" score of the guided diffusion model.

Subsequent work by \citet{liu2023more} generalized the notion of a ``classifier" to support image-to-image and text-to-image generation. While retaining the core mathematical structure, this approach replaces the dedicated classifier with a differentiable semantic scoring function that is not trained on noisy inputs and is not restricted to discrete class labels.

A key limitation of classifier guidance is its dependence on a separately trained classifier $p(\by\mid\bz_t)$. Since most off-the-shelf pre-trained classification models are not optimized for highly noisy inputs, such a classifier must be trained from scratch alongside the diffusion model.
Furthermore, the classifier's label space is inherently finite and cannot cover all possible concepts, making it poorly suited for out-of-distribution or novel categories.
For these reasons, we next consider an alternative framework that eliminates the need for an independent classifier.

\begin{algorithm}
\caption{Classifier-Free Guidance for Score-Based DDPM Training Procedure\index{Denoising diffusion probabilistic models}\index{DDPMs}}
\label{alg:score_ddpm_train_cfg}
\begin{algorithmic}[1]
\Require Paired training data $\mathcalX=\{\bx_n, \by_n\}$, noise schedule $\{\beta_t\}_t$, the total number of steps $T$ (standard choice: $T = 1,000$, $\beta_1 = 10^{-4}$, $\beta_T = 0.02$), neural network $\bs^\btheta_t$, dropout probability $0<\gamma<1$;
\State \textbf{initialize:} $\btheta$;
\For{$cnt=0,1,2,\ldots$}
\State Sample a data example $(\bx_0, \by)$ from the dataset; \Comment{(SCDPM$_1$)} 
\State $t\sim \uniformdist(\{1,2,\ldots,T\})$; \Comment{(SCDPM$_2$)} 
\State $\bepsilon\sim\normal(\bzero, \bI_D)$;  \Comment{(SCDPM$_3$)} 
\State $\bz_t\leftarrow \sqrt{\alpha_t}\bx_0+\sqrt{1-\alpha_t}\bepsilon$; \Comment{(SCDPM$_4$)} 
\State Ground-truth score:
$\bs^{\bthetastar}_t(\bz_t) \leftarrow \nabla_{\bz_t}\ln q_t(\bz_t) = -\frac{1}{\sqrt{1-\alpha_t}} \bepsilon$; \Comment{(SCDPM$_5$)} 
\State With probability $\gamma$ drop label: $\by \leftarrow \varnothing$; \Comment{(SCDPM$_6$)} 
\State Training using the loss term $\mathcalJ(\btheta) \leftarrow \normtwo{\bs^\btheta_t(\bz_t\mid\by) - \bs^{\bthetastar}_t (\bz_t)}^2$; \Comment{(SCDPM$_7$)} 
\EndFor
\State \Return Network parameter $\btheta$;
\end{algorithmic}
\end{algorithm}

\index{Classifier-free guidance (CFG)}
\subsection{Classifier-Free Guidance}\label{section:cfg_ddpm}

Shortly after the classifier guidance work, the Google Brain team introduced  \textit{classifier-free diffusion guidance (CFG)} \citep{ho2022classifier}.
This mechanism was subsequently adopted by several prominent models, including OpenAI's \textit{GLIDE} and\textit{DALL-E 2} \citep{nichol2021glide, ramesh2022hierarchical},  Google's \textit{Imagen} \citep{saharia2022photorealistic}, and StabilityAI's Stable Diffusion series \citep{rombach2022high}.
Classifier-free guidance can be derived through a minor modification of classifier guidance. Rearranging \eqref{equation:class_guid_decomp}, we obtain
$$
\nabla_{\bz_t} \ln p(\by \mid \bz_t) = \nabla_{\bz_t} \ln p(\bz_t \mid\by) - \nabla_{\bz_t} \ln p(\bz_t) .
$$
Substituting this expression into \eqref{equation:scaled_cond_score} yields the scaled score of the guided model in the form
\begin{align}
\widetilde{\nabla_{\bz_t}} \ln p(\bz_t \mid\by) &= \nabla_{\bz_t} \ln p(\bz_t) + \lambda \left( \nabla_{\bz_t} \ln p(\bz_t \mid\by) - \nabla_{\bz_t} \ln p(\bz_t) \right) \nonumber\\
&= \nabla_{\bz_t} \ln p(\bz_t) + \lambda \nabla_{\bz_t} \ln p(\bz_t \mid\by) - \lambda \nabla_{\bz_t} \ln p(\bz_t)\nonumber \\
&= \underbrace{\lambda \nabla_{\bz_t} \ln p(\bz_t \mid\by)}_{\text{Guided score}} + \underbrace{(1 - \lambda) \nabla_{\bz_t} \ln p(\bz_t)}_{\text{Unguided score}}.  \label{equation:cfg_ddpm}
\end{align} 
Once again, $\lambda$ controls the degree to which the trained guided model prioritizes the guidance information.
When $\lambda = 0$, the guided model fully disregards the guidance signal and reduces to an unguided diffusion model. When $\lambda = 1$, the model recovers the standard guided distribution. When $\lambda > 1$, the diffusion model not only emphasizes the guided score function but also actively shifts away from the unguided score. In other words, it lowers the probability of generating unguided samples in favor of those that strictly follow the guidance signal. This again reduces sample diversity, but improves alignment between generated samples and the provided guidance information.

Furthermore, we avoid training \textbf{separate} networks to model $p(\bz_t \mid \by)$ and $p(\bz_t)$ by training a single \textbf{unified guided model}. During training, the guidance variable $\by$ is randomly set to a null value (e.g., $\by = \varnothing$) with a fixed probability $\gamma$, typically in the range of 10–20\%. In this setup, $p(\bz_t)$ is approximated by $p(\bz_t \mid \by = \varnothing)$. This process is analogous to dropout, where guidance inputs are collectively disabled for a random subset of training examples.
Classifier-free guidance is conceptually elegant: it provides finer control over guided generation while requiring only a single diffusion model to be trained. We thus define our \textit{CFG score-based DDPM loss function} as
\begin{mybox}
\begin{align}
\mathcalJ&(\btheta)=\Exp_\diamondsuit\left[\normtwo{\bs^\btheta_t(\bz_t\mid\by) - \bs^{\bthetastar}_t (\bz_t)}^2\right],\\
\diamondsuit &=  t \sim \uniformdist(\{1,\ldots,T\}),\,  (\bx, \by) \sim \textcolor{black}{p_{\text{data}}}(\bx, \by), \,  \bz_t \sim q(\bz_t|\bx), \, \text{replace } \by = \varnothing \text{ with prob. } \gamma. \nonumber
\end{align}
\end{mybox}
The training procedure for the CFG-enabled score-based DDPM is outlined in Algorithm~\ref{alg:score_ddpm_train_cfg}. Alternative formulations can be trained using an analogous approach.

After training, the score function in \eqref{equation:cfg_ddpm} is used to enforce strong weighting of the guidance signal. In practice, classifier-free guidance produces significantly higher-quality outputs than classifier guidance \citep{nichol2021glide, saharia2022photorealistic}.
This improvement arises because a classifier $p(\by \mid \bz_t)$ may ignore most components of the input $\bz_t$ as long as it accurately predicts $\by$. In contrast, classifier-free guidance relies directly on the guided density $p(\bz_t \mid \by)$, which must assign high probability to the full structure of $\bz_t$.

\begin{problemset}

\item \label{prob:ddpm_trans_kernel} Let $q(\bz_t \mid \bz_{t-1}) = \normal\left( \bz_t \mid \sqrt{1 - \beta_t} \bz_{t-1},\, \beta_t \bI_D \right)$. Derive the transition kernel of DPMs or DDPMs given in \eqref{equation:diffusion_kernel_ddpm}: 
$q(\bz_t \mid \bx_0) = \int q(\bz_{1:t} \mid \bx_0) \diff\bz_{1:t-1}
= \normal\left( \bz_t \mid \sqrt{\alpha_t} \bx_0,\, (1 - \alpha_t) \bI_D \right)$. \textit{Hint: Use induction.}

\item \label{prob:ddpm_trans_kernel2}  Let $q(\bz_t \mid \bz_{t-1}) = \normal\left( \bz_t \mid \sqrt{1 - \beta_t} \bz_{t-1},\, \beta_t \bI_D \right)$. 
Similar to the transition kernel of  DPMs or DDPMs in \eqref{equation:diffusion_kernel_ddpm}, consider two timesteps $0<s<t$. Show that 
$$
\begin{aligned}
q(\bz_t\mid\bz_s) 
&= \normal \left(\Big(\prod_{k=s+1}^t \sqrt{1-\beta_k}\Big)\bz_s, 
\Big[1-\Big(\prod_{k=s+1}^{t}(1-\beta_k) \Big)\Big] \bI_D  
\right) \\
&= \normal \left(\sqrt{\frac{\alpha_t}{\alpha_s}}\bz_s, 
\big(1-\frac{\alpha_t}{\alpha_s}\big) \bI_D  
\right),
\end{aligned}
$$
where $\alpha_t \triangleq \prod_{\tau=1}^{t} (1 - \beta_\tau)$.

\item \label{prob:ddpm_trans_kernel3} Following Problem~\ref{prob:ddpm_trans_kernel2}, let $q(\bz_t \mid \bz_{t-1}) = \normal\left( \bz_t \mid \nu_t \bz_{t-1},\, \sigma_t^2 \bI_D \right)$. 
Consider two timesteps $0<s<t$. Show that 
$$
q(\bz_t\mid\bz_s) 
=
\normal \left({\frac{\nu_t}{\nu_s}}\bz_s, 
\Big(\sigma_t^2 - \frac{\nu_t^2}{\nu_s^2}\sigma_s^2\Big)\bI_D  
\right).
$$

\item \label{prob:dpm_zT} Show that when $T\rightarrow \infty$, $\bz_T$ in a DPM or a DDPM converges to the standard Gaussian distribution $\bz_T\sim\normal(\bzero, \bI_D)$. \textit{Hint: Examine the parameters in the transition kernel  \eqref{equation:diffusion_kernel_zt_form_ddpm}.}

\item \textbf{Variance preserving (VP).} Consider the transition kernel in \eqref{equation:ddpm_diffusion_kernel_all}. Show that it satisfies the variance preserving property: for unit-variance data satisfying $\Exp[\rvx_0\rvx_0^\top]=\bI$, the total covariance of $\rvx_t$ remains  $\bI_D$ for all $t$. \textit{Hint: Use the affine transformation of Gaussian distributions; Lemma~\ref{lemma:affine_mult_gauss}.}

\item Derive the relation in \eqref{equation:ddpm_reverse_conditional_gaussian} in details.

\item Derive the relation in \eqref{equation:ddpm_mu_in_terms_of_epsilon}.

\item Derive the relation in \eqref{equation:mux0zt_score}. \textit{Hint: Use the  fact that $\alpha_t = (1-\beta_t)\alpha_{t-1}$ in \eqref{equation:ddpm_alpha_t_ddpm}.}

\item Train your own DDPM model use a neural network structure. Sampling using Algorithm~\ref{alg:diffusion_sampling} and compared it with the results from removing the noise in (SDDPM$_3$) step. What do you observe?

\item \label{prob:kl_bernoulli} \textbf{KL of Bernoullis.} 
Given two Bernoulli distributions $p(x)=\bernoulli(x\mid p)$ and $q(x)=\bernoulli(x\mid q)$, show that 
$$
\KL[p \parallel q] = p\ln\frac{p}{q} + (1-p) \ln \frac{1-p}{1-q}.
$$	

\item \label{prob:bernoulli_marg_kernel} \textbf{Binomial diffusion.}  Derive the marginal kernel of the binomial diffusion model in \eqref{equation:bernoulli_marg_kernel}. \textit{Hint: the Bernoulli distribution is fully determined by its mean. 
Let $m_t\triangleq \Exp[z_t\mid x_0]$ and use the tower property (Problem~\ref{prob:prop_expcond}) to show that $m_t = (1-\beta_t) m_{t_1} + \tfrac{1}{2}\beta_t$. Then prove by recursion. 
}

\item \textbf{Binomial diffusion.} Derive the loss function from the variational bound of the binomial diffusion model in \eqref{equation:bernoulli_elbo}. How about the denoising version of this loss function?

\end{problemset}
\newpage 
\chapter{Flow-Based Models}\label{chapter:flow}
\begingroup
\hypersetup{
linkcolor=structurecolor,
linktoc=page,  
}
\minitoc \newpage
\endgroup

\lettrine{\color{caligraphcolor}I}
In the rapidly advancing field of deep generative modeling, flow-based models have emerged as a flexible, powerful framework capable of modeling complex probability distributions via invertible transformations. Unlike generative adversarial networks (GANs), which depend on adversarial training, and variational autoencoders (VAEs) or denoising diffusion probabilistic models (DDPMs), which are bounded by variational approximations, flow-based models directly parameterize probability densities through a series of invertible mappings termed ``flows." This unique mechanism enables exact unbiased likelihood estimation, efficient sampling, and interpretable latent space representations. Over the past decade, flow-based models have evolved from basic discrete transformations to sophisticated continuous dynamical systems. Major variants include \textit{coupling flows}, \textit{autoregressive flows}, \textit{continuous flows}, and the recently proposed \textit{flow matching} paradigm. Each variant addresses distinct challenges in generative modeling---ranging from high-dimensional data processing to training complexity reduction---collectively advancing the state-of-the-art in density estimation, image synthesis, natural language processing, and broader generative tasks.

\begin{figure}[h!]
\centering                      
\vspace{-0.35cm}                 
\subfigtopskip=2pt               
\subfigbottomskip=2pt            
\subfigcapskip=-5pt              
\subfigure[Base and target distributions.]{\label{fig:flow_goal1}
\includegraphics[width=0.236\linewidth]{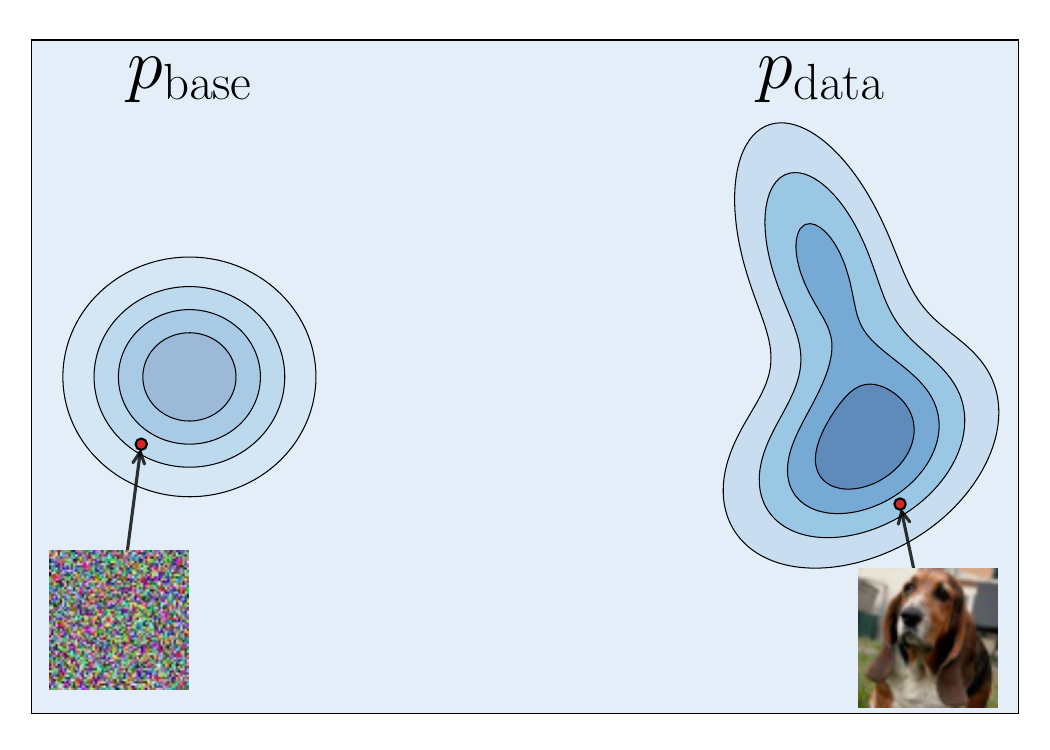}}
\subfigure[Path design in flow-matching.]{\label{fig:flow_goal2}
\includegraphics[width=0.236\linewidth]{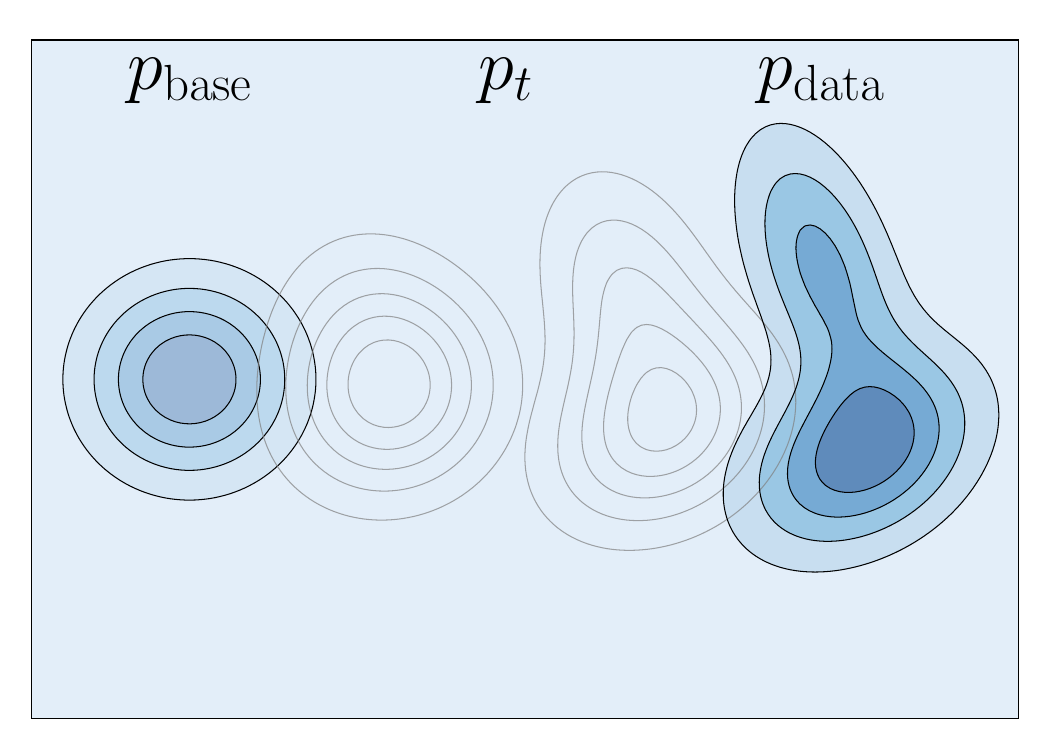}}
\subfigure[Training.]{\label{fig:flow_goal3}
\includegraphics[width=0.236\linewidth]{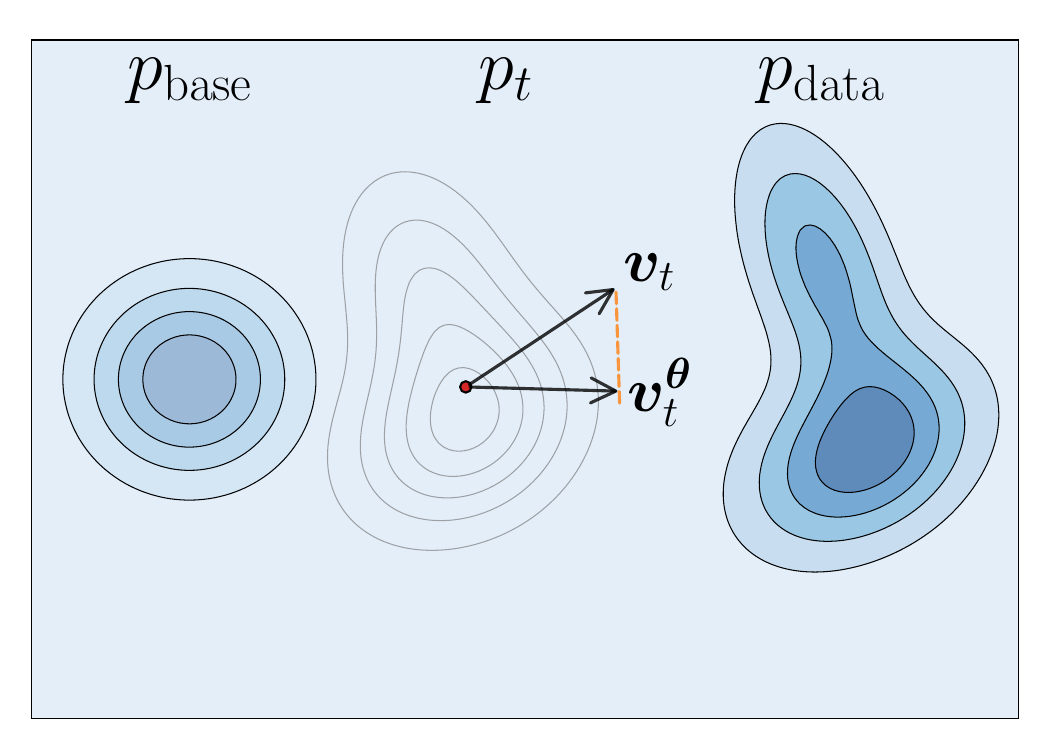}}
\subfigure[Sampling.]{\label{fig:flow_goal4}
\includegraphics[width=0.236\linewidth]{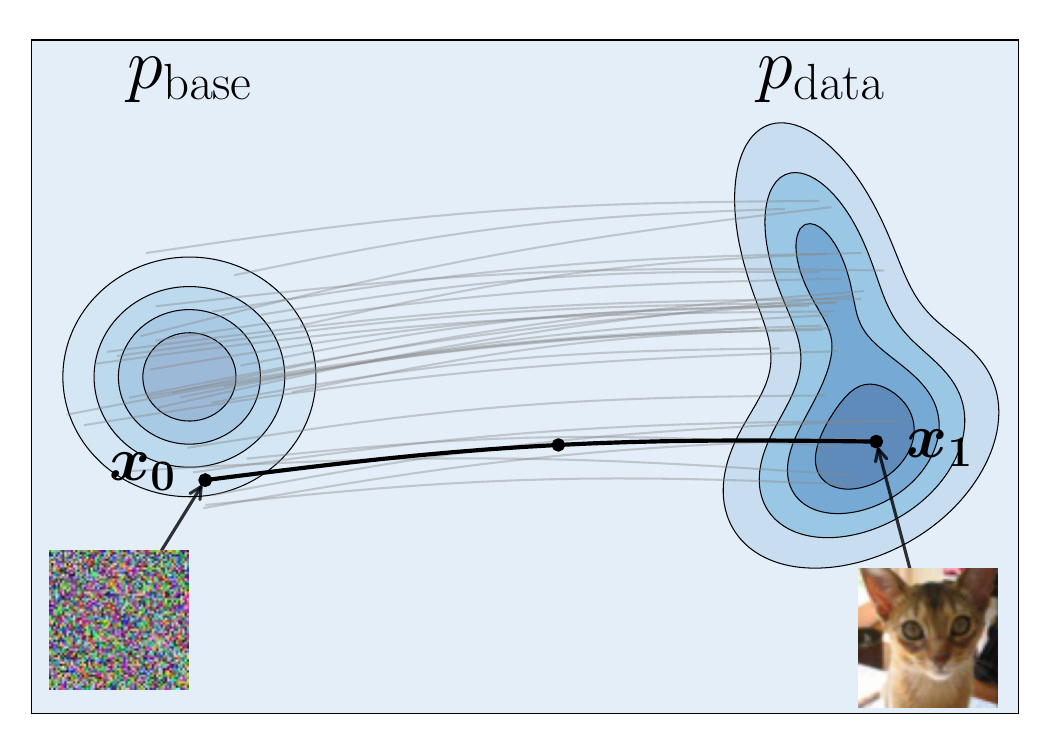}}
\caption{ 
\textit{The flow-based models blueprint.}
Fig~\ref{fig:flow_goal1} The goal of a flow-based model is to find a flow mapping that transforms samples $\rvx_0$ from a known base or noise distribution $p_{\text{base}}$ into samples $\rvx_1$ from an unknown target or data distribution $p_{\text{data}}$. This mapping is typically required to be invertible and differentiable, which is a core characteristic of flow-based models.
Fig~\ref{fig:flow_goal2} To achieve this transformation, two common approaches are adopted:  for traditional flow-based models (e.g., coupling flow, masked autoregressive flow), a sequence of invertible sub-mappings is constructed to directly transform $p_{\text{base}}$ to $p_{\text{data}}$ without a continuous probability path; 
for models involving continuous dynamics (e.g., flow-matching), a time-continuous probability path $\{p_t\}_{0 \leq t \leq 1}$ interpolating between $p_{\text{base}}$ and $p_{\text{data}}$ is designed.
Fig~\ref{fig:flow_goal3} During training, the specific learning objective varies by subcategory:  
for traditional flow-based models (e.g., coupling flow, masked autoregressive flow), the parameters of the invertible sub-mappings are optimized to maximize the log-likelihood of the data, leveraging the change-of-variable theorem to compute the probability density of the target distribution; 
for flow-matching models, regression is used to estimate the vector field $\vf_t$ that generates the probability path $\{p_t\}_{0\leq t\leq 1}$.
Fig~\ref{fig:flow_goal4} To draw a novel target sample $\rvx_1 \sim p_{\text{data}}$, the approach depends on the model type: for traditional flow-based models, apply the learned invertible flow mapping to a sample $\rvx_0 \sim p_{\text{base}}$ to obtain the target sample $\rvx_1$, leveraging the invertibility of the mapping to ensure valid transformation from the base distribution to the target distribution;
for flow-matching models, integrate the estimated vector field $\vf_t^{\widehatbtheta}(\rvx_t)$ from $t=0$ to $t=1$, where $\rvx_0 \sim p_{\text{base}}$ is a novel source sample.
}
\label{fig:flow_goal}
\end{figure}

At their core, all flow-based models share a fundamental principle: by decomposing a complex target distribution into a series of invertible, differentiable transformations applied to a simple base distribution (e.g., a Gaussian), they allow for straightforward computation of the target density using the change of variables theorem  (see Figure~\ref{fig:flow_goal} for a visual illustration).
This principle not only ensures exact \textit{likelihood estimation}---a critical advantage for tasks like density estimation and out-of-distribution detection---but also enables \textit{controllable generation} by manipulating the latent space (\textit{For example, base samples follow a known distribution that is easy to sample, and target samples are given to us in terms of a dataset of finite size. Depending on the application, target samples may constitute images, videos, audio segments, or
other types of high-dimensional, richly structured data. The goal is then to generate random samples of interest}).
This objective aligns closely with the central focus of this book (see Figure~\ref{fig:DDPM_gen_model_idea}).
However, the design of these invertible transformations has undergone significant evolution, leading to the diversity of flow types we see today.

Discrete flow variants, such as coupling flow and autoregressive flow, laid the groundwork for flow-based modeling by introducing structured invertible mappings that balance expressiveness and computational efficiency. \textit{Coupling flow}, popularized by models like \textit{RealNVP}, partitions the input into disjoint subsets and transforms one subset conditional on the other, leveraging the flexibility of conditional neural networks while maintaining invertibility. 
\textit{Autoregressive flow}, by contrast, constructs transformations in a sequential, autoregressive manner, where each dimension of the input is transformed conditional on previous dimensions---offering strong expressiveness at the cost of sequential computation. These discrete flows revolutionized generative modeling by demonstrating that invertible transformations could scale to high-dimensional data, but they also faced limitations: coupling flows often require careful partitioning strategies to avoid bottlenecks, while autoregressive flows suffer from slow sampling due to their sequential nature.

The limitations of discrete flows paved the way for the development of continuous flow, a paradigm that models transformations as continuous dynamical systems evolving over time. Instead of a finite sequence of discrete steps, continuous flow represents the transformation as a smooth, time-dependent flow governed by ordinary differential equations (ODEs), where the input evolves continuously from the base distribution to the target distribution. This continuous formulation eliminates the need for discrete partitioning or sequential processing, enabling more flexible and efficient density estimation. However, training continuous flows historically required computing the Jacobian determinant of the flow, a computationally expensive step that limited their scalability.

\textit{Flow matching}, a recent breakthrough in flow-based modeling, addressed this scalability challenge by reimagining the training of continuous flows. Rather than directly optimizing the likelihood (and thus requiring Jacobian computations), flow matching trains the velocity field of the continuous flow by matching it to a predefined reference vector field, simplifying the training objective to a regression task. This innovation not only reduces computational complexity but also unifies discrete and continuous flow paradigms, enabling the development of hybrid models that combine the strengths of both. Today, flow matching has become a cornerstone of modern flow-based modeling, driving advances in high-quality generation, conditional modeling, and bidirectional transport between distributions---often integrated with coupling mechanisms to form powerful coupled flow matching frameworks.

This chapter aims to provide a comprehensive overview of flow-based models, focusing on the four key variants: coupling flow, autoregressive flow, continuous flow, and flow matching. We will dissect their core principles, mathematical formulations, advantages, and limitations, trace their evolutionary trajectory, and explore how they complement and enhance one another. 
Additionally, the chapter extends flow matching models to score matching models via stochastic differential equations (SDEs).

\index{Change-of-variables formula}
\index{Jacobian matrix}
\index{Normalizing flow}
\index{Diffeomorphism}
\index{Push-forward operator}
\section{Normalizing Flows}
We now discuss normalizing flow models for training nonlinear latent variable models. These models constrain the structure of neural networks such that the data likelihood can be computed exactly without approximation, while retaining straightforward sampling from the trained model. 
Like other generative models we have introduced previously, we first define a \textit{base distribution} $p_{\text{base}}(\bz)$ over the latent variable $\bz$, alongside a nonlinear deep neural network transformation $\bx = \bmathcalD_{\btheta}(\bz)$ that maps the latent space to the data space. When $p_{\text{base}}(\bz)$ is a simple distribution such as a Gaussian, sampling is trivial: one draws a latent sample $\widetildebz \sim p_{\text{base}}(\bz)$ and feeds it through the neural network to produce a corresponding data sample $\widetildebx = \bmathcalD(\widetildebz; \btheta)$.

Evaluating the model's data likelihood requires the data-space probability distribution, which relies on the inverse network transformation $\bz = \bmathcalE_\btheta(\bx)$, satisfying the identity $\bz = \bmathcalE_{\btheta}(\bmathcalD_{\btheta}(\bz))$. For this inverse to be well-defined for all valid parameters $\btheta$, the transformation functions $\bmathcalD_{\btheta}(\bz)$ and $\bmathcalE_\btheta(\bx)$ must be \textbf{invertible} (\textbf{bijective}). 
Bijectivity ensures a one-to-one correspondence between data samples $\bx$ and latent samples $\bz$. We then apply the \textit{change-of-variables formula} to obtain the exact data density:
\begin{equation}\label{equation:jacob_chg_var}
p_{\btheta}(\bx) = p_{\text{base}}(\bmathcalE_\btheta(\bx)) \abs{\det\big(\bJ(\bx)\big)},
\end{equation}
where $\bJ(\bx)\triangleq \partial_{\bx}\bmathcalE(\bx)$ denotes the \textit{Jacobian matrix} of partial derivatives of $\bmathcalE_\btheta(\bx)$, with entries $J_{ij}(\bx) = \frac{\partial \mathcalE^i_{\btheta}(\bx)}{\partial x_j}$. Here, $\mathcalE^i_{\btheta}(\bx)=[\bmathcalE_\btheta(\bx)]_i$ is the $i$-th element of the inverse transformation $\bmathcalE_\btheta(\bx)$, and $\det(\bA)$ denotes the determinant of a square matrix $\bA \in \real^{D\times D}$.
\footnote{	
The \textit{determinant} is a scalar-valued function that maps a square matrix $\bA \in \real^{D \times D}$ to a real number, denoted $\det(\bA)$ or $\abs{\bA}$.
Geometrically, it represents the \textit{signed volume scaling factor} of the linear transformation induced by $\bA$:
it describes how much a unit volume in $\real^D$ is stretched, compressed, or reflected by the transformation.
A nonzero determinant indicates the matrix is invertible (nonsingular), while a zero determinant means the matrix is singular (non-invertible).
} 
Notably, we still refer to $\bz$ as a \textit{latent variable} despite the deterministic, bijective mapping. 
To align with the conventional autoencoder formulation, we adopt the notations $\mathcalE$ and $\mathcalD$ throughout our analysis.
Although each data point $\bx$ maps to a unique, deterministic latent value $\bz$, eliminating posterior uncertainty over latent states, we retain conventional latent-variable terminology to ensure consistency with the preceding context.

To rigorously formalize the change-of-variables likelihood derivation \eqref{equation:jacob_chg_var} in full detail, we next introduce the formal definitions of \textit{diffeomorphisms} and \textit{push-forward operators}, which serve as the mathematical foundation of flow-based generative modeling.
\begin{definition}[Diffeomorphisms and push-forward maps]\label{definition:diffeomorphism}
Let $C^r(\real^m, \real^n)$ denote the space of $r$-times continuously differentiable functions $\bff: \real^m\rightarrow \real^n$, whose higher-order partial derivatives are defined as:
\begin{equation}
\frac{\partial^r f_k}{\partial x_{i_1} \cdots \partial x_{i_r}}, \quad k \in [n], i_j \in [m], f_k=[\bff(\bx)]_k,
\end{equation}
where $[n] \triangleq \{1, 2, \dots, n\}$. 
For notational simplicity, we further define $C^r(\real^n) \triangleq C^r(\real^n, \real)$. 
Under this convention, for instance, $C^1(\real^m)$ represents the set of continuously differentiable scalar-valued functions on $\real^m$.
A fundamental class of functions for our analysis is $C^r$ \textit{diffeomorphism}. 
A function $\bmathcalD \in C^r(\real^n, \real^n)$ is a  $C^r$ {diffeomorphism} if it is bijective, and its inverse satisfies
$\bmathcalD^{-1}  \in C^r(\real^n, \real^n)$.

Given a random variable $\rvz \sim p_\rvz$ with PDF $p_\rvz$, we define a transformed random variable  $\rvx = \bmathcalD(\rvz)$, where $\bmathcalD: \real^D \to \real^D$ is a $C^1$ diffeomorphism. The PDF of $\rvx$, denoted $p_\rvx$, is referred to as the \textit{push-forward of $p_\rvz$} (under $\bmathcalD$). 
We derive $p_\rvx$  via the standard change-of-variables formula for random variables.
By definition of the expectation operator,
$$ 
\Exp[f(\rvx)] 
= \Exp[f(\bmathcalD(\rvz))] 
= \int f(\bmathcalD(\bz)) p_\rvz(\bz) \diff \bz 
= \int f(\bx) p_\rvz(\bmathcalD^{-1}(\bx)) \abs{\det \big(\partial_{\bx} \bmathcalD^{-1}(\bx)\big)} \diff \bx, 
$$
where the third equality is due the change of variables $\bz = \bmathcalD^{-1}(\bx)$.
By matching the integrand to the definition of the expectation with respect to $p_\rvx$, we obtain the explicit form of the push-forward PDF:
\begin{equation}
p_\rvx(\bx) = p_\rvz(\bmathcalD^{-1}(\bx)) \abs{\det \big(\partial_{\bx} \bmathcalD^{-1}(\bx)\big)}.
\end{equation}
We introduce the \textit{push-forward operator} $\sharp$  to compactly denote this transformation. For a diffeomorphism $\bmathcalD$, the push-forward of $p_{\rvz}$ is defined as
\begin{equation}
[\bmathcalD_{\sharp} p_\rvz](\bx) 
=p_\rvz(\bmathcalD^{-1}(\bx)) \abs{\det \big(\partial_{\bx} \bmathcalD^{-1}(\bx)\big)}.
\end{equation}
Conversely, we define the corresponding \textit{pull-back operator} via the inverse variable transformation. Letting $\bx=\bmathcalD(\bz)\triangleq\bmathcalE^{-1}(\bz)$, the pull-back of $p_\rvx$ under $\bmathcalE$ takes the form
\begin{equation}
[\bmathcalE_{\sharp} p_\rvx](\bz) 
=p_\rvx(\bmathcalE^{-1}(\bz)) \abs{\det \big(\partial_{\bz} \bmathcalE^{-1}(\bz)\big)}.
\end{equation}
\end{definition}

In this work, the diffeomorphic mapping $\bmathcalD_{\btheta}(\bz)$ is parameterized by a specialized neural network architecture, which we elaborate on in subsequent sections. A key constraint of diffeomorphic mappings is that the latent space and data space must share identical dimensionality. This dimensional matching often results in large model architectures when processing high-dimensional data such as natural images.
Additionally, the computational cost of evaluating matrix determinants constitutes a major bottleneck in flow-based models. For a $D \times D$ Jacobian matrix, determinant computation incurs a time complexity of $\mathcalO(D^3)$, where $D$ denotes the input dimension or the number of hidden units \citep{lu2021numerical}. 
To alleviate this computational overhead, we introduce structural constraints on the network design to enable efficient Jacobian determinant evaluation.

Consider a training dataset $\mathcalX = \{\bx_1, \bx_2, \ldots, \bx_N\}$ consisting of $N$ independent and identically distributed data samples. Using the change-of-variables result derived above, the log-likelihood function of the dataset given model parameters $\btheta$ is formulated as:
\begin{equation}\label{equation:normflow_cplike}
\ln p_{\btheta}(\mathcalX   ) 
= \sum_{n=1}^N \ln p_{\btheta}(\bx_n )
= \sum_{n=1}^N \left\{ \ln p_{\text{base}}(\bmathcalE_{\btheta}(\bx_n)) + \ln \abs{\det\big(\bJ(\bx_n)\big) } \right\} ,
\end{equation}
where $p_{\text{base}}$ denotes the base latent distribution and $\bJ$ denotes the Jacobian matrix of the inverse transformation; see \eqref{equation:jacob_chg_var}. 
Our training objective is to maximize this log-likelihood function to optimize the network parameters $\btheta$.
This density transformation framework is still consistent with the main focus of this book; see Figure~\ref{fig:DDPM_gen_model_idea}.

To model complex, real-world data distributions with high flexibility, the transformation $\bx = \bmathcalD_{\btheta}(\bz)$ is implemented via deep neural networks. 
To guarantee global invertibility of the overall transformation, we construct the network as a composition of \textit{individually invertible layers}. 
Consider a three-layer composite diffeomorphism as an illustrative example:
$$
\bx = \bmathcalD^A\left( \bmathcalD^B\left( \bmathcalD^C(\bz) \right) \right).
$$
where $\bmathcalD^A, \bmathcalD^B, \bmathcalD^C$ denote the invertible transformations corresponding to three consecutive network layers. The inverse composite transformation is then given by the reverse composition of individual layer inverses:
$$
\bz = \bmathcalE^C\left( \bmathcalE^B\left( \bmathcalE^A(\bx) \right) \right),
$$
where $\bmathcalE^A$, $\bmathcalE^B$, and $\bmathcalE^C$ are the inverse mappings of $\bmathcalD^A$, $\bmathcalD^B$, and $\bmathcalD^C$, respectively. 
The Jacobian determinant of the composite transformation can be efficiently decomposed via the multivariate chain rule. The Jacobian matrix of the inverse composite mapping satisfies:
$$
J_{ij} = \frac{\partial z_i}{\partial x_j} = \sum_k \sum_l \frac{\partial \mathcalE_i^C}{\partial \mathcalE_k^B} \frac{\partial \mathcalE_k^B}{\partial \mathcalE_l^A} \frac{\partial \mathcalE_l^A}{\partial x_j}.
$$
This expression corresponds to the product of the Jacobian matrices of the three individual layers. Leveraging the matrix determinant property that the determinant of a product equals the product of determinants (see Problem~\ref{prob:prod_deter} or \citet{lu2021numerical}), the log-determinant of the full Jacobian matrix decomposes into the sum of log-determinants of per-layer Jacobian matrices:
$$
\det(\bJ_{\text{total}}) = \prod_{k=1}^K \det(\bJ_k)
\quad\implies \quad
\ln\abs{\det(\bJ_{\text{total}})} = \sum_{k=1}^K \ln\abs{\det(\bJ_k)}.
$$ 
This additive structure enables stable and efficient computation of the log-determinant term in the log-likelihood,
avoiding the high cost of directly evaluating large Jacobian determinants, and simplifying gradient-based optimization.

This generative modeling paradigm is termed \textit{normalizing flows}, as the sequential composition of invertible mappings gradually transforms a simple base distribution into a complex data distribution, analogous to the continuous flow of fluid density across a manifold \citep{rezende2015variational, kobyzev2020normalizing, bishop2023deep}. 
Notably, the inverse flow mapping projects the complex empirical data distribution back to a standardized, simple base distribution (most commonly a standard Gaussian distribution).
In the remainder of this section, we introduce the core theoretical and architectural concepts of two dominant families of practical normalizing flows: \textit{coupling flows} and \textit{autoregressive flows}.

\index{RealNVP}
\index{Coupling flow}
\subsection{Coupling Flows}
When constructing coupling-based flow frameworks, we aim to design a standalone invertible functional layer that can be stacked sequentially to yield a highly versatile class of invertible transformations.  
We first consider an affine linear mapping between the observed data vector  $\bx$ and the latent variable $\bz$, formulated as
\begin{equation}
\bx = \gamma\bz + \bbeta.
\end{equation}
This mapping admits a straightforward inverse transformation:
\begin{equation}
\bz = \frac{1}{\gamma}(\bx - \bbeta).
\end{equation}
Nevertheless, linear transformations exhibit \textit{closure under composition},such that cascading multiple linear layers collapses into a single unified linear transformation. 
Furthermore, any affine transformation applied to a Gaussian distribution produces another Gaussian distribution (see Lemma~\ref{lemma:affine_mult_gauss}). Consequently, stacking arbitrarily many linear layers cannot generate non-Gaussian distributions, resulting in fundamentally limited expressive capacity.
This motivates a key research question: can we preserve the exact invertibility of linear mappings while introducing additional structural flexibility to model complex, non-Gaussian data distributions?

\paragrapharrow{RealNVP.}
A canonical solution to this problem is implemented by the \textit{real-valued non-volume-preserving (RealNVP)} normalizing flow architecture proposed in \citet{dinh2014nice}. 
The core mechanism involves splitting the D-dimensional latent vector $\bz$ into two disjoint subvectors: $\bz = [\bz_1; \bz_2]$, where  $\bz_1$ is $d$-dimensional and $\bz_2$ spans the remaining $D - d$ dimensions. 
The observed data vector $\bx$ is partitioned identically as $\bx = [\bx_1; \bx_2]$, with matching dimensionality for $\bx_1$ ($d$-dimensional) and $\bx_2$ ($(D-d)$-dimensional).
The first partition of the output is defined as a direct copy of the corresponding latent partition:
\begin{subequations}\label{equation:coup_zALL}
\begin{equation}\label{equation:coup_z1}
\bx_1 = \bz_1.
\end{equation}
The second partition undergoes an affine transformation whose scaling and shifting coefficients are dynamically modulated by nonlinear neural network functions of $\bz_1$:
\begin{equation}\label{equation:coup_z2}
\bx_2 = \exp\left(\bgamma_{\btheta}(\bz_1)\right) \hadaprod \bz_2 + \bbeta_{\btheta}(\bz_1).
\end{equation}
\end{subequations}
Here, $\bgamma_{\btheta}(\bz_1)$ and $\bbeta_{\btheta}(\bz_1)$ denote real-valued outputs from parameterized neural networks with parameters $\btheta$.
The elementwise exponential operation guarantees positive scaling coefficients for numerical stability. The operator $\hadaprod$ represents the Hadamard (elementwise) vector product, and the exponential function in \eqref{equation:coup_z2} is also applied elementwise across vector entries. While the formulation unifies the network parameters under a single $\btheta$ notation, practical implementations may employ two independent parameterized subnetworks or a single network with dual output branches.

\begin{SCfigure}
\centering
\includegraphics[width=0.45\textwidth]{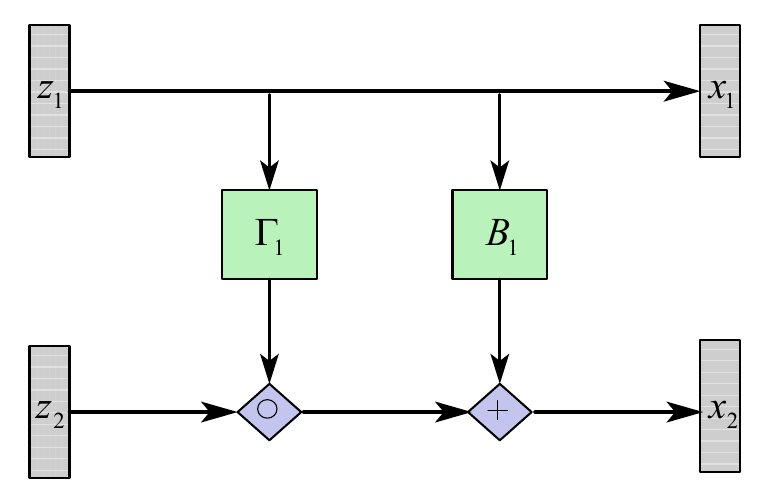} 
\caption{
Conceptual illustration of a  single layer of the RealNVP normalizing flow model. 
Network $\Gamma_1$ evaluates the function $\exp(\bgamma_{\btheta}(\bz_1))$, while network $B_1$ computes  $\bbeta_{\btheta}(\bz_1)$.
The resulting output vector is then specified by  \eqref{equation:coup_zALL}.
}
\label{fig:coupling_flow1}
\end{SCfigure}

By virtue of the nonlinear neural modulation, the resulting mapping from $\bx_1$ to $\bx_2$ possesses substantial expressive flexibility. Critically, the full transformation remains exactly invertible. Given an observed sample $\bx = [\bx_1; \bx_2]$, the latent variables can be recovered via the inverse procedure:
\begin{subequations}
\begin{equation}\label{equation:coup_x1}
\bz_1 = \bx_1.
\end{equation}
Following the recovery of $\bz_1$, we compute the network-modulated coefficients $\bgamma_{\btheta}(\bz_1)$ and $\bbeta_{\btheta}(\bz_1)$, then solve for $\bz_2$ as:
\begin{equation}\label{equation:coup_x2}
\bz_2 = \exp\left(-\bgamma_{\btheta}(\bz_1)\right) \hadaprod \left( \bx_2 - \bbeta_{\btheta}(\bz_1) \right).
\end{equation}
\end{subequations}
The complete coupling transformation pipeline is visualized in Figure~\ref{fig:coupling_flow1}, where network $\Gamma_1$ outputs $\exp(\bgamma_{\btheta}(\bz_1))$ and network $B_1$ outputs the shift term $\bbeta_{\btheta}(\bz_1)$. Notably, the individual neural network mappings $\bgamma_{\btheta}(\bz_1)$ and $\bbeta_{\btheta}(\bz_1)$ are not required to be invertible---only the full composite coupling transformation maintains strict invertibility.

\paragrapharrow{Jacobian and determinant calculation.}
We next examine the computation of the Jacobian matrix and its determinant. Partitioning the Jacobian according to the split of $\bz$ and $\bx$ yields the block-structured matrix:
\begin{equation}\label{equation:coup_jacob_exp}
\bJ = 
\begin{bmatrix}
\bI_d & \bzero \\
\frac{\partial \bz_2}{\partial \bx_1} & \diag\left( \exp(-\bgamma) \right)
\end{bmatrix}
\quad \implies\quad 
\det(\bJ)
=
\prod_{i=d+1}^D\exp(-\gamma_i).
\end{equation}
The top-left block represents the derivatives of  $\bz_1$ with respect to $\bx_1$, which from \eqref{equation:coup_x1} is simply the $d \times d$ identity matrix. 
The top-right block contains derivatives of $\bz_1$ with respect to $\bx_2$,\ which are identically zero, again by \eqref{equation:coup_x1}. 
The bottom-left block corresponds to derivatives of $\bz_2$ with respect to $\bx_1$, which from \eqref{equation:coup_x2} yield complex expressions involving the neural network mappings. 
Finally, the bottom-right block gives derivatives of $\bz_2$ with respect to $\bx_2$, which form a diagonal matrix with entries equal to the elementwise exponentials of $-\bgamma_{\btheta}(\bz_1)$. 
The Jacobian in \eqref{equation:coup_jacob_exp} is thus lower triangular, with all entries above the main diagonal equal to zero. For such matrices, the determinant equals the product of the diagonal entries~\footnote{See, for example, \citet{lu2021numerical}.}, and is therefore independent of the intricate expressions in the bottom-left block. As a result, the Jacobian determinant simplifies directly to the product of the elements of $\exp(-\bgamma_{\btheta}(\bz_1))$.

\begin{figure}[htp]
\centering                      
\vspace{-0.35cm}                 
\subfigtopskip=2pt               
\subfigbottomskip=2pt            
\subfigcapskip=-5pt              
\includegraphics[width=0.9\textwidth]{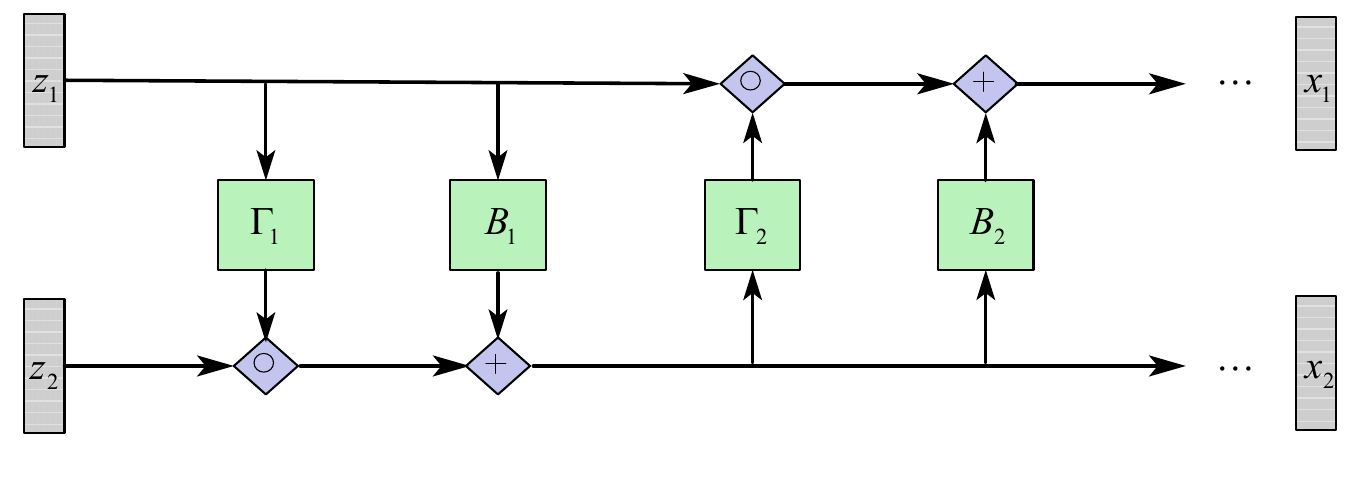} 
\caption{
Conceptual illustration of the full RealNVP framework.
By stacking two layers structured as in Figure~\ref{fig:coupling_flow1}, we construct a more expressive yet fully invertible nonlinear layer. Since each individual sublayer supports exact inversion and admits tractable Jacobian computation, the combined two-layer architecture preserves both desirable properties.
}
\label{fig:coupling_flow2}
\end{figure}

\paragrapharrow{Stacking the structure.}
A clear drawback of this construction is that $\bz_1$ remains unmodified by the transformation. This issue is readily addressed by introducing an additional layer (or multiple layers) where the roles of $\bz_1$ and $\bz_2$ are swapped, as depicted in Figure~\ref{fig:coupling_flow2}. This two-layer pattern can then be iterated repeatedly to construct highly expressive generative models.

\paragrapharrow{Training and density transformation.}
The overall training pipeline operates on mini-batches of data, where each sample contributes to the log-likelihood via \eqref{equation:normflow_cplike},
where the base latent distribution is usually chosen as the standard Gaussian distribution.
The inverse mapping $\bz = \bmathcalE(\bx)$ is computed by chaining inverse transformations of the form \eqref{equation:coup_x2}. And the log Jacobian determinant is obtained by summing the log determinants of each layer, where each term is itself a sum of components of the form $-\gamma^i_{\btheta}(\bx)$, $i=d+1, d+2, \ldots,D$. 
Gradients of the log-likelihood can be efficiently computed via automatic differentiation, and the network parameters updated using stochastic gradient descent; see Algorithm~\ref{alg:realnvp}.

\begin{algorithm}[h]
\caption{RealNVP Normalizing Flow: Forward/Inverse Pass and Training}
\label{alg:realnvp}
\begin{algorithmic}[1]
\Require 
Data point $\bx \in \real^D$, 
neural network parameters $\{\btheta_l\}_{l=1}^L$ for $L$ coupling layers,
base distribution $p_{\text{base}}(\bz) = \normal(\bz\mid \bzero,\bI)$;
\Ensure 
Latent code $\bz$ and log-likelihood contribution $\ln p(\bx)$;

\State \textbf{Forward Pass (Data $\rightarrow$ Latent)}
\State Initialize $\bz^{(0)} \leftarrow \bx$
\For{$l = 1, 2, \ldots, L$}
\State Partition $\bz^{(l-1)}$ into two parts: $\bz_1, \bz_2$;
\Comment{Partition dimension $D$ into two subsets}
\State Compute scale and shift:
\State $\bgamma \leftarrow \text{NN}_\gamma(\bz_1; \btheta_l)$;
\Comment{Scale network, elementwise output}
\State $\bbeta \leftarrow \text{NN}_\beta(\bz_1; \btheta_l)$;
\Comment{Bias network, elementwise output}
\State Transform the second (or first) partition: $\bz_2' \leftarrow \exp(-\bgamma) \hadaprod (\bz_2 - \bbeta)$;
\State Set $\bz^{(l)} \leftarrow \text{concat}(\bz_1, \bz_2')$;
\Comment{Recombine partitions}
\State Accumulate log-Jacobian determinant: $\ln \abs{\det(\bJ_l)} \leftarrow -\sum_i \gamma_i$;
\EndFor
\State Set $\bz \leftarrow \bz^{(L)}$
\State Compute log-likelihood: $\ln p(\bx) \leftarrow \ln p_{\text{base}}(\bz) + \sum_{l=1}^L \ln \abs{\det(\bJ_l)}$;

\vspace{0.5em}
\State \textbf{Inverse Pass (Latent $\rightarrow$ Data)}
\State Initialize $\bx^{(0)} \leftarrow \bz$
\For{$l = L, L-1, \ldots, 1$}
\State Partition $\bx^{(L-l)}$ into $\bx_1, \bx_2$;
\State Compute scale and shift:
\State $\bgamma \leftarrow \text{NN}_\gamma(\bx_1; \btheta_l)$;
\State $\bbeta \leftarrow \text{NN}_\beta(\bx_1; \btheta_l)$;
\State Invert the transformation: $\bx_2' \leftarrow \exp(\bgamma) \hadaprod \bx_2 + \bbeta$;
\State Set $\bx^{(L-l+1)} \leftarrow \text{concat}(\bx_1, \bx_2')$;
\EndFor
\State Set $\bx \leftarrow \bx^{(L)}$;

\vspace{0.5em}
\State \textbf{Training Step (Single Mini-Batch)}
\For{each mini-batch $\{\bx_n\}_{n=1}^N$}
\State Compute forward pass for each $\bx_n$ to get $\bz_n$ and $\ln p(\bx_n)$;
\State Compute average loss: $\mathcalJ \leftarrow -\frac{1}{N}\sum_{n=1}^N \ln p(\bx_n)$;
\State Update parameters $\{\btheta_l\}$ via gradient descent: $\btheta_l \leftarrow \btheta_l - \eta \nabla_{\btheta_l} \mathcalJ$;
\EndFor
\end{algorithmic}
\end{algorithm}

\index{Masked autoregressive flow}
\index{Inverse autoregressive flow}
\subsection{Autoregressive Flows}
A distinct family of normalizing flows can be derived from a fundamental probabilistic property: any joint distribution over a set of random variables can be factorized into a product of univariate conditional distributions. We first define a fixed ordering for the elements of the vector $\bx$ (for example, the raster scan order for image data). With this ordering, the joint distribution can be universally rewritten as follows:
\begin{equation}\label{equation:cond_ar_prob}
p(x_1, x_2, \ldots, x_D) = \prod_{i=1}^D p(x_i \mid  \bx_{1:i-1}),
\end{equation}
where $\bx_{1:i-1}$ denotes the subsequence $x_1, \ldots, x_{i-1}$. 
This modeling paradigm aligns closely with the first class of generative models introduced in Chapter~\ref{chapter_genintroduction}. As illustrated in Figure~\ref{fig:autoregres_comp}, it addresses the fundamental generative question: Given partial observed data, what is the subsequent content? This characteristic property gives rise to the name \textit{autoregressive flows}.

\begin{figure}[h!]
\centering                      
\vspace{-0.35cm}                 
\subfigtopskip=2pt               
\subfigbottomskip=2pt            
\subfigcapskip=-5pt              
\subfigure[Masked autoregressive flow.]{\label{fig:flow_arflow1}
\includegraphics[width=0.35\linewidth]{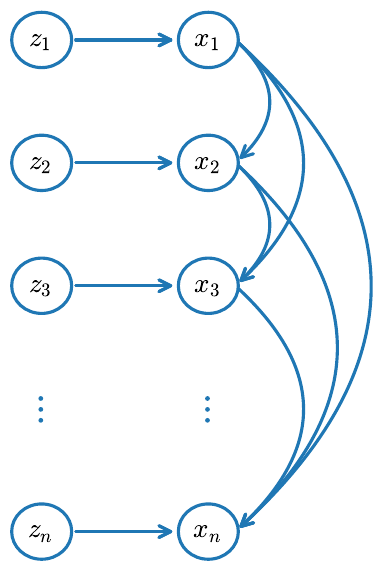}}
\subfigure[Inverse autoregressive flow.]{\label{fig:flow_arflow2}
\includegraphics[width=0.228\linewidth]{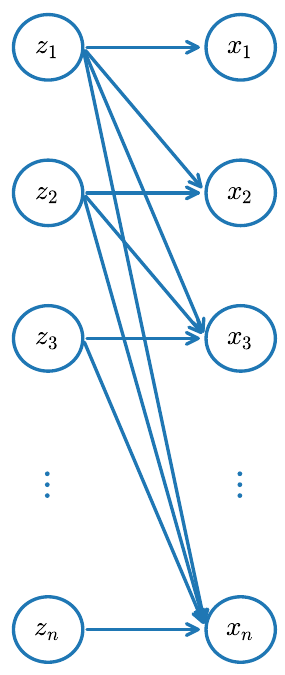}}
\caption{
Conceptual illustration of two alternative architectures for autoregressive normalizing flows.
While the masked autoregressive flow enables efficient likelihood evaluation, its counterpart---the inverse autoregressive flow---supports efficient sampling.
}
\label{fig:flow_arflows}
\end{figure}

\paragrapharrow{Masked autoregressive flow.}
The factorization~\eqref{equation:cond_ar_prob} forms the basis of a normalizing flow variant known as a \textit{masked autoregressive flow (MAF)}, defined as \citep{papamakarios2017masked}:
\begin{equation}\label{equation:maf}
x_i = \phi\left(z_i; \mathcalE_{\btheta_i}^i(\bx_{1:i-1})\right).
\end{equation}
A conceptual illustration is shown in Figure~\ref{fig:flow_arflow1}. 
Here, $\phi(z_i, \cdot)$ represents the coupling function, selected to be easily invertible with respect to $z_i$, 
while $\mathcalE_i$ acts as the conditioner, usually implemented using a deep neural network. 
The term \textit{masked} refers to the use of a single neural network to enforce the set of relations in \eqref{equation:maf}, paired with a binary mask  that zeros out a subset of network weights to enforce the autoregressive constraint in \eqref{equation:cond_ar_prob}.

For likelihood evaluation, the required inverse transformations take the form
\begin{equation}\label{equation:inv_maf}
z_i = \phi^{-1}\left(x_i; \mathcalE_{\btheta_i}^i(\bx_{1:i-1})\right),
\end{equation}
which can be computed efficiently on modern hardware, as the individual evaluations of $z_1,  z_2, \ldots, z_D$ from \eqref{equation:inv_maf} can be performed in \textbf{parallel}. 
The corresponding Jacobian matrix of the transformations \eqref{equation:inv_maf}, with entries $\partial z_i / \partial x_j$, is upper triangular. 
Therefore, its determinant thus equals the product of its diagonal elements, allowing for highly efficient computation. 
By contrast, sampling from this model requires sequential evaluation of \eqref{equation:maf}, which is inherently slow, as each value $x_i$ can only be computed after $x_1, x_2, \ldots, x_{i-1}$ have been determined.

\paragrapharrow{Inverse autoregressive flow.}
To resolve the inefficient sampling inherent to MAF, we adopt an alternative framework known as the \textit{inverse autoregressive flow (IAF)}, defined as \citep{kingma2016improved}:
\begin{equation}\label{equation:iaf}
x_i = \phi\left(z_i; \mathcalD_{\btheta_i}^i(\bz_{1:i-1})\right).
\end{equation} 
A conceptual illustration is shown in Figure~\ref{fig:flow_arflow2}. 
Sampling becomes highly efficient under this formulation: for a fixed latent vector  $\bz$, the components $x_1, x_2, \ldots, x_D$ can be evaluated in \textbf{parallel} using \eqref{equation:iaf}. 
Conversely, computing the inverse transformation---which is required for likelihood evaluation; see \eqref{equation:normflow_cplike}---involves sequential operations of the form
\begin{equation}\label{equation:iaf_inv}
z_i = \phi^{-1}\left(x_i; \mathcalD_{\btheta_i}^i(\bz_{1:i-1})\right),
\end{equation}
making this step inherently sequential and computationally slow. The choice between MAFs and IAFs therefore depends on the requirements of the particular application.

Autoregressive flows and coupling flows share fundamental structural similarities. While autoregressive flows deliver strong modeling expressiveness, their sequential ancestral sampling induces a computational cost that scales linearly with data dimensionality $D$. By contrast, coupling flows can be interpreted as a constrained subclass of autoregressive flows. These models trade full autoregressive generality for better computational efficiency by partitioning variables into only two groups, rather than $D$ individual sequential components.

\index{Ordinary differential equations (ODEs)}
\index{Stochastic differential equations (SDEs)}
\section{Flow Matching}\label{section:flow_match}

In prior discussions, we formalized generative modeling as the process of sampling from the data distribution $p_{\text{data}}$, and established the core objective of generative model development: designing an algorithm capable of producing samples $\bx \sim p_{\text{data}}$. This section elaborates on the construction of generative models via the simulation of carefully designed \textit{differential equations}.
Notably, flow matching and diffusion models  rely on the simulation of \textit{ordinary differential equations (ODEs)} and \textit{stochastic differential equations (SDEs)}, respectively \citep{lipman2022flow, lipman2024flow, holderrieth2025introduction}. 
As introduced in Chapter~\ref{chapter:diff}, diffusion models gradually corrupt input data with Gaussian noise to yield a simple prior distribution, then learn a reverse denoising process to generate high-fidelity realistic samples. In contrast, flow matching builds continuous, invertible flow trajectories that smoothly map a simple noise distribution to the target data distribution.

A key distinction between the two frameworks lies in their mapping properties: flow models enable exact invertible transformations, whereas the forward diffusion process is irreversible. Despite this difference, both paradigms generate data through iterative trajectory evolution and eliminate the need for explicit latent posterior inference.

In this section, we first provide formal definitions of ODEs, alongside a discussion of their numerical simulation. We then introduce deep neural network-based parameterization strategies for ODEs, which lay the foundation for formulating flow matching models as well as their core sampling algorithms. Model training methodologies will be comprehensively explored in subsequent sections.

\index{Trajectory}
\index{Velocity field}
\index{Vector field}
\index{Continuous-time Markov process (CTMP)}
\index{Diffeomorphism}
\index{Markov property}
\subsection{Flow Models}\label{section:flomat_flowmodel}

\begin{figure}[h]
\centering                      
\vspace{-0.35cm}                 
\subfigtopskip=2pt               
\subfigbottomskip=2pt            
\subfigcapskip=-5pt              
\includegraphics[width=0.999\textwidth]{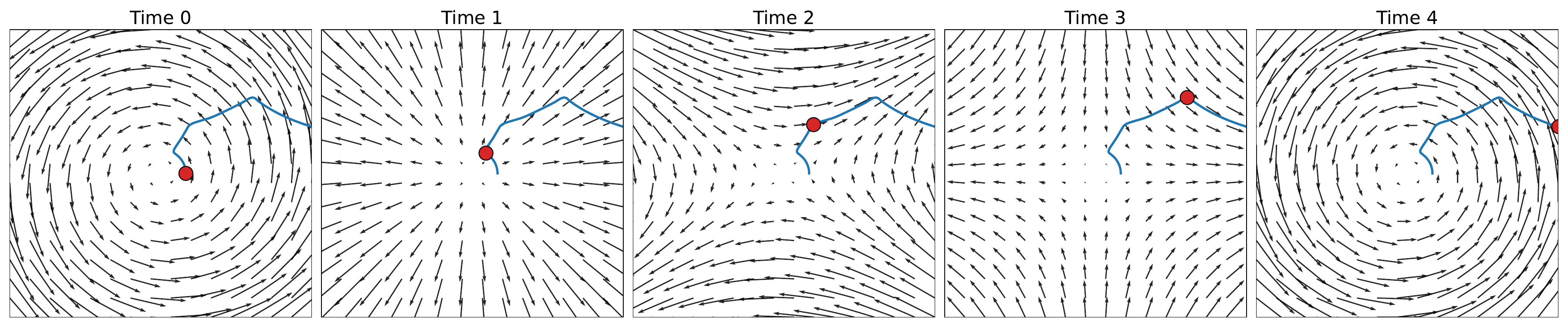}
\caption{
A flow $\bpsi_t : \real^D \to \real^D$ is induced by a time-dependent velocity field $\vf_t : \real^D \to \real^D$, whose instantaneous spatial movements at every coordinate point are visualized via the arrow field in the illustration (with $D=2$ in this setup). 
We demonstrate the flow evolution across five distinct time instances $t$, where the red dot marks the trajectory's spatial position at each corresponding timestamp.
}
\label{figure:flow_field}
\end{figure}

We begin by revisiting the fundamental concept of \textit{ordinary differential equations (ODEs)} \citep{chen2018neural}. 
An ODE solution is characterized by a \textit{trajectory}, visualized in Figure~\ref{figure:flow_field}, which corresponds to a function formulated as
$$ 
\rvx : [0, 1] \to \real^D, \quad t \mapsto \rvx_t.
$$
This function maps each time instance $t\in[0,1]$ to a spatial coordinate in the $D$-dimensional Euclidean space  $\real^D$. 
For notational clarity, we adopt a consistent font convention throughout this work: the regular-font variable $\rvx_t$ represents the collection of all possible spatial positions that a particle can attain at time $t$, whereas the italicized variable $\bx_t$ denotes a specific, fixed position occupied by the particle in a single realization of the dynamical process.

Central to any ODE is a \textit{vector field} (also referred to as a \textit{velocity field}) $\vf$, defined as the function
$$ 
\vf : \real^D \times [0, 1] \to \real^D, \quad (\bz, t) \mapsto \vf_t(\bz), 
$$
which assigns a $D$-dimensional velocity vector $\vf_t(\bz) \in \real^D$ to every spatial location $\bz$ and time $t$, encoding the local spatial velocity at the specified position and timestamp (illustrated in Figure~\ref{figure:flow_field}).

A valid ODE trajectory is constrained to align with the flow dictated by the vector field $\vf_t$ and is initialized at the starting position $\bx_0$. Notably, the initial position $\bx_0$ and vector field $\vf$ uniquely determine the resulting trajectory $\mathcalX$. This dynamical process can be formally defined via the initial-value problem below:
\begin{subequations}\label{equation:ode_def}
\begin{align}
\frac{\diff}{\diff t} \rvx_t &= \vf_t(\rvx_t); & (\text{ODE}) \label{eq:ode1} \\
\rvx_0 &= \bx_0. & (\text{initial conditions}) \label{eq:ode2}
\end{align}
\end{subequations}
Equation \eqref{eq:ode1} enforces that the temporal derivative of the trajectory $\rvx_t$ matches the velocity direction and magnitude prescribed by the vector field $\vf_t$ at the corresponding state. Equation \eqref{eq:ode2} specifies the initial constraint, fixing the particle's position to $\bx_0$ at the initial timestep $t = 0$.

Furthermore, the \textit{flow}, denoted by $\bpsi$, is defined as the ensemble of all trajectories induced by a vector field $\vf$.
Each individual trajectory characterizes the continuous path of a particle propagating through the state space, following the dynamical rules encoded in $\vf$. Given an initial spatial coordinate $\bx_0$ and a time instance $t$, the flow map $\bpsi_t$ yields the particle's subsequent position $\bx_t$ via the relation $\bx_t = \bpsi_t(\bx_0)$. This dynamical evolution is governed by the ODE presented in Equation \eqref{eq:ode1}.
Substituting the flow formulation into the core ODE relation yields the rigorous defining equations for the flow structure, as summarized below:
\begin{subequations}\label{equation:flows_def}
\begin{align}
\bpsi : \real^D \times [0, 1] &\to \real^D, \quad (\bx_0, t) \mapsto \bpsi_t(\bx_0); \label{eq:flow1} \\
\frac{\diff }{\diff t} \bpsi_t(\bx_0) &= \vf_t(\bpsi_t(\bx_0)); &  (\text{flow ODE}) \label{eq:flow2} \\
\bpsi_0(\bx_0) &= \bx_0. & (\text{flow initial conditions})                        \label{eq:flow3}
\end{align}
For a given initial state $\rvx_0 = \bx_0$, the corresponding ODE trajectory can be retrieved as:
\begin{equation}
\rvx_t = \bpsi_t(\rvx_0).
\end{equation}
\end{subequations}
From an alternative perspective, a \textit{flow model} corresponds to a  \textit{continuous-time Markov process (CTMP)} $\{\rvx_t\}_{0\leq t\leq 1}$. 
This process is constructed by applying the flow map $\bpsi_t$ to the initial random state variable $\rvx_0$, formulated as:
\begin{equation}\label{equation:flow_ctmp}
\rvx_t = \bpsi_t(\rvx_0), \quad t\in[0,1], \quad \text{where } \rvx_0\sim p_{\text{base}}. 
\end{equation}
The Markov property of $\{\rvx_t\}$ can be verified through the following derivation. For any time pair satisfying $0 \leq t < s \leq 1$, we obtain:
\begin{equation}\label{equation:flow_ctmp2}
\rvx_s = \bpsi_s(\rvx_0) = \bpsi_s(\bpsi_t^{-1}(\bpsi_t(\rvx_0))) = \bpsi_{s|t}(\rvx_t).
\end{equation}
The above derivation holds for two key reasons. First, $\bpsi_t$ constitutes a \textit{diffeomorphism} (Definition~\ref{definition:diffeomorphism})---a smooth and bijective coordinate transformation with a smooth inverse, as validated in Theorem~\ref{theorem:flow_exist}. Second, substituting the CTMP definition $\rvx_t = \bpsi_t(\rvx_0)$ and defining the conditional flow mapping $\bpsi_{s|t} \triangleq \bpsi_s \circ \bpsi_t^{-1}$ (which also preserves the diffeomorphism property) completes the equality. Equation \eqref{equation:flow_ctmp2} demonstrates that future states $\rvx_s$ depend solely on the present state $\rvx_t$, which confirms the Markov property of the process $\{\rvx_t\}$. Notably, the state dependence inherent to flow-based Markov processes is entirely \textbf{deterministic}.

Intuitively, vector fields, ODEs, and flows offer three equivalent perspectives describing the same underlying dynamical structure: vector fields specify ODEs, and the solution trajectories of these ODEs collectively form flows. As with any differential equation system, two critical questions arise regarding the well-posedness of the defined ODE: whether a valid solution exists and, if existent, whether such a solution is unique. A core mathematical result answers both questions affirmatively under mild regularity conditions on the velocity field $\vf_t$, as formalized in the following theorem.
\begin{theoremHigh}[Flow existence and uniqueness \citep{perko2013differential, lipman2024flow}]\label{theorem:flow_exist}
If $\vf : \real^D \times [0, 1] \to \real^D$ is continuously differentiable with  bounded derivatives, then the ODE system in  \eqref{equation:flows_def} admits a unique solution characterized by the flow map $\bpsi_t$. 
Under these conditions, $\bpsi_t$ is a \textit{diffeomorphism} for all $t\in[0,1]$, meaning $\bpsi_t$ is continuously differentiable and possesses a continuously differentiable inverse  $\bpsi_t^{-1}$.
\end{theoremHigh}

This theorem establishes that vector fields satisfying standard smoothness conditions uniquely induce a valid flow. Importantly, these existence and uniqueness prerequisites are universally satisfied in practical machine learning settings. Since $\vf_t(\bz)$ is parameterized via neural networks in our framework, its derivatives are inherently bounded, fulfilling the theorem's requirements. Consequently, Theorem~\ref{theorem:flow_exist} guarantees robust mathematical well-posedness for our formulation: \textbf{flows always exist and correspond to unique ODE solutions for all scenarios considered in this work}.

Conversely, a continuously differentiable flow $\bpsi_t$ uniquely determines its associated vector field. For any spatial coordinate $\bz\in\real^D$, the corresponding velocity field can be recovered from the flow dynamics via the flow ODE \eqref{eq:flow2}:
$\frac{\diff }{\diff t} \bpsi_t(\widetildebz) = \vf_t(\bpsi_t(\widetildebz))$.
Leveraging the diffeomorphic property of $\bpsi_t$---which ensures invertibility for all $t\in[0,1]$---we substitute $\widetildebz = \bpsi_t^{-1}(\bz)$ to invert the flow mapping. This yields the explicit expression for the unique vector field $\vf_t$ that generates the flow $\bpsi_t$:
\begin{equation}\label{equation:vecfield_from_flow}
\vf_t(\bz) = \dot{\bpsi}_t(\bpsi_t^{-1}(\bz)),
\end{equation}
where $\dot{\bpsi}_t \triangleq  \frac{\diff }{\diff t} \bpsi_t$ denotes the temporal derivative of the flow map. 
Overall, this mutual construction establishes a rigorous equivalence between continuously differentiable, bounded vector fields $\vf_t$ and continuously differentiable diffeomorphic flows $\bpsi_t$.

\begin{example}[Linear vector fields]\label{example:ode_lnvf}
We consider a simple example of a vector field $\vf_t(\bz)$ that is linear in $\bz$, namely $\vf_t(\bz) = \bA \bz$. 
The flow map $\bpsi$ that solves the ODE defined in \eqref{equation:flows_def} is then given by
\begin{equation}\label{equation:linear_vf}
\bpsi_t(\bx_0) = \exp(\bA t) \bx_0,
\end{equation}
where the matrix exponential $\exp(\bA t) $ is defined via the following Taylor series expansion (see Theorem~\ref{theorem:linear_approx}):
$$
\exp(\bA t) = \sum_{k=0}^{\infty} \frac{(\bA t)^k}{k!} = \bI + \bA t  + \frac{\bA^2 t^2}{2!} + \frac{\bA^3 t^3}{3!} + \ldots.
$$
We verify this solution as follows. At $t=0$, we obtain $\bpsi_0(\bx_0) = \bx_0$, which satisfies the initial condition for the flow. 
Differentiating the flow map with respect to time yields the governing flow ODE:
\begin{equation*}
\frac{\diff}{\diff t} \bpsi_t(\bx_0) 
= \frac{\diff}{\diff t} (\exp(\bA t) \bx_0) 
= \bA \exp(\bA t) \bx_0 
= \bA \bpsi_t(\bx_0) = \vf_t(\bpsi_t(\bx_0)).
\end{equation*}
Equation~\eqref{equation:linear_vf} defines an explicit flow: the flow corresponds to a linear transformation acting on the initial system state. This linear vector field flow possesses several fundamental properties, summarized below:
\begin{itemize}
\item \textbf{Linearity}: $\vf(\bz_1 + \bz_2) = \vf(\bz_1) + \vf(\bz_2)$ and $\vf(\alpha\bz) = \alpha \vf(\bz)$ for all $\bz_1, \bz_2, \bz, \alpha$.

\item \textbf{Determined by eigenvalues}: The long-term behavior of the system (e.g., stability, oscillations, exponential growth/decay) is completely determined by the eigenvalues of $\bA$. For example:
\begin{itemize}
\item If all eigenvalues have negative real parts, the system is globally asymptotically stable (trajectories converge to the origin). Specifically, when $\bA=a<0$ is a negative scalar,  all trajectories decay exponentially to the origin $\bz=\bzero$ as time $t$ increases.
\item If any eigenvalue has a positive real part, the system is unstable.
\end{itemize}
\end{itemize}
\end{example}

\index{Euler method}
\index{Second-order midpoint method}
\index{Heun's method}
\index{Predictor-corrector procedure}
\index{Runge--Kutta methods}
\paragrapharrow{Simulating an ODE.} 
In most practical scenarios, the flow map $\bpsi_t$ cannot be derived in closed form when the velocity field $\vf_t$ deviates from the simple cases presented earlier. For such general settings, numerical methods are adopted to perform ODE simulation. Fortunately, numerical ODE solving is a well-established and extensively studied area in numerical analysis, offering a rich collection of robust and efficient algorithms for this purpose \citep{iserles2009first, lipman2024flow, holderrieth2025introduction}.
The \textit{Euler method} represents one of the most fundamental, intuitive numerical approaches for ODE integration,  and it is widely used in numerical softwares \citep{virtanen2020scipy}. When applied to \eqref{equation:diffode_deis}. 
The scheme initializes the particle state as $\rvx_0 = \bx_0$ and iteratively updates the state via
\begin{equation}
\rvx_{t+\tau} = \rvx_t + \tau \cdot \vf_t(\rvx_t), \quad t = 0, \tau, 2\tau, \ldots, 1-\tau ,
\end{equation}
where $\tau = n^{-1} > 0$ denotes  the {step size} and $n \in \naturalset$ corresponds to the total number of discrete simulation steps.
The Euler update is equivalent to a first-order Taylor expansion of  $\rvx_t$, which yields:
\begin{equation}
\rvx_{t+\tau} = \rvx_t + \tau \frac{\diff}{\diff t} \rvx_t  + o(\tau) = \rvx_t + \tau \vf_t(\rvx_t) + o(\tau),
\end{equation}
where the \textit{small-oh function}  $o(\cdot): \real_+\rightarrow \real$ is a one-dimensional function satisfying $\frac{o(\mu)}{\mu}\rightarrow 0$ as $\mu\rightarrow 0^+$.
This formulation implies that the Euler method incurs a local truncation error of $o(\tau)$ per integration step, leading to a global accumulated error of $o(1)$ after $n = 1/\tau$ total steps. As a result, the overall approximation error of the Euler method converges to zero as the step size $\tau$ is reduced toward zero.

The Euler method is only one of many available ODE solvers, with higher-order alternatives offering substantially improved accuracy in practical applications. A widely used second-order alternative is the \textit{second-order midpoint method}. As a second-order numerical scheme, its global error scales quadratically with step size, delivering far more precise approximations than the Euler method for identical step sizes.
Unlike the Euler method, which evaluates the derivative solely at the start of each time interval, the midpoint method approximates the average slope over the interval by computing the derivative at the interval midpoint, yielding a superior representation of the solution curve's local curvature. The method follows a two-stage \textit{predictor-corrector procedure}, summarized as:
\begin{align*}
\rvx_{\text{mid}} &= \rvx_t + \frac{\tau}{2} \vf_t(\rvx_t); && (\text{predictor step})\\
\rvx_{t+\tau} &= \rvx_t + \tau \vf_{t+\tau/2}(\rvx_{\text{mid}}). && (\text{corrector step})
\end{align*}
The \textit{predictor step} performs a ``half-step" Euler update to estimate the system state at the temporal midpoint of the interval. 
The subsequent \textit{corrector step} evaluates the velocity field at this predicted midpoint state and uses this refined slope to compute the full-step state update from the original time-$t$ state $\rvx_t$. By leveraging midpoint gradient information, the method effectively mitigates the linearization error inherent in Euler-style straight-line approximations of curved solution trajectories.

\textit{Heun's method}, also known as the \textit{improved Euler method} or the \textit{explicit trapezoidal rule}, is another popular second-order ODE solver. 
Rather than relying exclusively on either the initial or midpoint interval slope, Heun's method constructs a slope estimate by averaging the derivative evaluated at the start of the interval and the derivative evaluated at an Euler-predicted end state. Its update rules are defined as:
\begin{align*}
\widetilde{\rvx}_{t+\tau} &= \rvx_t + \tau \vf_t(\rvx_t); && (\text{initial guess of new state}) \\
\rvx_{t+\tau} &= \rvx_t + \frac{\tau}{2} (\vf_t(\rvx_t) + \vf_{t+\tau}(\widetilde{\rvx}_{t+\tau})). && (\text{update with average  $\vf$})
\end{align*}
Conceptually, Heun's method first generates a preliminary Euler-based prediction of the next system state, then corrects this tentative update using averaged gradient information across the time interval. Similar to the midpoint method, it achieves second-order accuracy and requires two evaluations of the velocity field per time step.

\textit{Runge--Kutta (RK)} methods generalize the Euler method by using a weighted average of intermediate slopes.
The first-order forward Euler method uses only the slope at the initial point, yielding a per-step local error of $\mathcalO(\tau^2)$ and first-order global accuracy.
An $s$-stage RK method samples $s$ intermediate points within the step size $\tau$, computes the derivative at each point, and combines these derivatives using optimized weights to achieve up to $s$-th order global accuracy.
For example, the \textit{second-order Runge--Kutta method (RK2)} introduces an intermediate slope at the half-step point, requiring two function evaluations per step:
\begin{align*}
\rvv_1 &= \bv_t(\rvx_t); \\
\rvv_2 &= \bv_{t+\tau/2}(\rvx_t + \frac{\tau}{2}\rvv_1); \\
\rvx_{t+\tau} &= \rvx_t + \tau \cdot \rvv_2.
\end{align*}
The widely used \textit{fourth-order Runge--Kutta method (RK4)} is defined as follows:
\begin{align*}
\rvv_1 &= \bv_t(\rvx_t);\\
\rvv_2 &= \bv_{t+\tau/2}(\rvx_t + \frac{\tau}{2} \rvv_1);\\
\rvv_3 &= \bv_{t+\tau/2}(\rvx_t + \frac{\tau}{2} \rvv_2);\\
\rvv_4 &= \bv_{t+\tau}(\rvx_t + \tau \rvv_3);\\
\rvx_{t+\tau} &= \rvx_t + \frac{\tau}{6} (\rvv_1+2\rvv_2+2\rvv_3+\rvv_4).
\end{align*}
The weights $(1,2,2,1)/6$ are derived by matching Taylor expansions, ensuring the first four order terms are exact.
RK4 requires four function evaluations per step---corresponding to four neural network forward passes---and strikes a strong balance between accuracy and computational cost, which explains its widespread use.
In general, an $s$-th order RK method has the property that halving the step size $\tau$ reduces the global numerical error by a factor of $1/2^s$.
For RK4 in particular, halving the step size reduces the error to $1/16$ of its original value, resulting in a substantial improvement in accuracy.

\paragrapharrow{Generative flow models.}

Building on the ODE simulation framework introduced above, we can construct a generative model by parameterizing the vector field via a deep neural network with trainable parameters $\btheta$, denoted as $\vf_t^\btheta$. For the purpose of this discussion, $\vf_t^\btheta$ is defined as a parameterized mapping $\vf_t^\btheta : \real^D \times [0, 1] \to \real^D$ governed by the parameter set $\btheta$; detailed neural network architecture designs will be elaborated in Chapter~\ref{chapter:diffarchitect}.
\footnote{We adopt a consistent notational convention throughout this work: $\rvx_0$ and $\bx_0$ correspond to the random variable and individual sample of the base distribution, while $\rvx_1$ and $\bx_1$ denote the random variable and sample of the target data distribution. For notational flexibility, we occasionally use $\bz$ to represent samples from the base or intermediate distributions, and reserve $\bx$ exclusively for samples drawn from the data distribution.}
The core objective of our generative framework is to produce random samples $\bx \sim p_{\text{data}}$ that follow the target data distribution $p_{\text{data}}$. Notably, ODE systems are inherently deterministic and lack stochasticity, which contradicts the random nature of data sampling. To resolve this discrepancy, stochasticity is introduced by randomizing the ODE's initial condition $\rvx_0$. Specifically, we define a base (initial) distribution $p_{\text{base}}$ for the initial state. In standard implementations, $p_{\text{base}}$ is set to a standard Gaussian distribution $\normal(\bzero, \bI_D)$ for simplicity. 
A critical requirement for the base distribution is tractable sampling at inference time, regardless of the specific distribution chosen.
With these definitions, the generative flow model is fully characterized by the following system of ODE equations:
\begin{subequations}\label{equation:gen_flow_models}
\begin{align}
\rvx_0 &\sim p_{\text{base}}; && (\text{random initialization})  \\
\frac{\diff}{\diff t} \rvx_t &= \vf_t^\btheta(\rvx_t) . && (\text{ODE})
\end{align}
\end{subequations}
The primary learning objective is to train the parameterized vector field such that the terminal state $\rvx_1$ of the ODE trajectory adheres to the target data distribution $p_{\text{data}}$. This objective can be formally expressed as:
\begin{equation*}
\rvx_1 \sim p_{\text{data}} \quad \iff \quad \bpsi_1^\btheta(\rvx_0) \sim p_{\text{data}},
\end{equation*}
where $\bpsi_t^\btheta$ denotes the flow map induced by the neural vector field $\vf_t^\btheta$. It is important to clarify a key conceptual distinction: despite the name ``flow model", the neural network directly parameterizes the underlying vector field rather than the flow map itself. The flow trajectory can only be obtained through numerical ODE simulation based on the parameterized vector field. The complete sampling pipeline for generative flow models is summarized in Algorithm~\ref{alg:samp_nfm_euler}.

\paragrapharrow{Sampling backwards.}
Recall the forward flow ODE, which characterizes the continuous trajectory of the flow map from $t = 0$ to $t = 1$:
$$
\frac{\diff}{\diff t}\bpsi_t(\bx_0) = \vf_t(\bpsi_t(\bx_0)), \quad \bpsi_0(\bx_0) = \bx_0.
$$
Forward simulation integrates the ODE from $t = 0$ to $t = 1$ to obtain the final state at $t=1$. However, likelihood evaluation for observed data samples requires computing the reverse flow path (see Section~\ref{section:like_flow_hutchin}). Given a data point $\bx_1 = \bpsi_1(\bx_0)$ sampled from the target distribution $p_{\text{data}}$, we aim to recover its corresponding initial latent state $\bx_0 = \bpsi_0(\bx_1)$ and calculate the required integral term. This procedure corresponds to backward-time ODE simulation: we initialize the system at $t=1$ with a data sample $\bx_1\sim p_{\text{data}}$ and integrate from $t = 1$ to $t = 0$ to recover the initial latent state $\bx_0$.
The reverse-time ODE can be derived through a straightforward time substitution. We define the reversed time variable $s = 1 - t$, which gives the differential relation $\diff t = -\diff s$. Substituting this transformation into the forward ODE yields the following derivation:
\begin{align*}
\frac{\diff}{\diff t}\bpsi_t(\bx_0) = \vf_t(\bpsi_t(\bx_0)) 
&\implies -\frac{\diff}{\diff s}\bpsi_{1-s}(\bx_1) = \vf_{1-s}(\bpsi_{1-s}(\bx_1)) \\
&\implies \frac{\diff}{\diff s}\bpsi_{1-s}(\bx_1) = -\vf_{1-s}(\bpsi_{1-s}(\bx_1)).
\end{align*}
This derivation indicates that backward flow simulation is equivalent to solving the original ODE system with an inverted time direction and a negated vector field. The complete workflow for reverse-path sampling is summarized in Algorithm~\ref{alg:samp_bkds_nfm_euler}.

\noindent
\begin{minipage}[t]{0.495\linewidth}
\begin{algorithm}[H]
\caption{Sampling Forwards from a Neural  Flow Model with Euler method}
\label{alg:samp_nfm_euler}
\begin{algorithmic}[1]
\Require Neural network vector field $\vf_t^\btheta$, number of steps $n$;
\State Set $t \leftarrow 0$;
\State Set step size $\tau \leftarrow \frac{1}{n}$;
\State Draw a noise sample $\bx_0 \sim p_{\text{base}}$;
\For{$i = 1,2, \ldots, n$}
\State $\bx_{t+\tau} \leftarrow \bx_t + \tau \cdot \vf_t^\btheta(\bx_t)$;
\State Update $t \leftarrow t + \tau$;
\EndFor
\State \Return $\bx_1$;
\end{algorithmic}
\end{algorithm}
\end{minipage}%
\hfil 
\begin{minipage}[t]{0.495\linewidth}
\begin{algorithm}[H]
\caption{Sampling Backwards from a Neural  Flow Model with Euler method}
\label{alg:samp_bkds_nfm_euler}
\begin{algorithmic}[1]
\Require Neural network vector field $\vf_t^\btheta$, number of steps $n$;
\State Set $t \leftarrow 0$;
\State Set step size $\tau \leftarrow \frac{1}{n}$;
\State Draw a data sample $\bx_1 \sim p_{\text{data}}$;
\For{$i = 1,2, \ldots, n$}
\State $\bx_{t-\tau} \leftarrow \bx_t - \tau \cdot \vf_t^\btheta(\bx_t)$;
\State Update $t \leftarrow t - \tau$;
\EndFor
\State \Return $\bx_0$;
\end{algorithmic}
\end{algorithm}
\end{minipage}

\begin{figure}[htp]
\centering  
\subfigtopskip=2pt 
\subfigbottomskip=9pt 
\subfigcapskip=-5pt 
\includegraphics[width=0.99\textwidth]{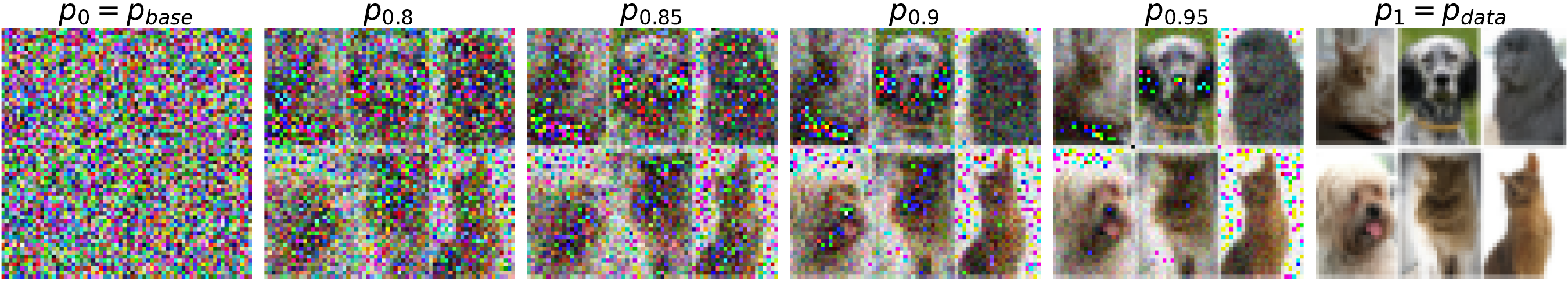}
\caption{We gradually interpolate from random noise to target images along a Gaussian-conditioned probability path.}
\label{fig:flow_interpolation}
\end{figure}

\index{Gaussian probability path}
\index{Conditional probability path}
\index{Marginal probability path}
\index{Dirac delta distribution}
\index{Dirac delta function}
\subsection{Conditional and Marginal Probability Path}\label{section:cond_marg_prob_path}

In the previous subsection, we introduce generative flow models as generative frameworks parameterized by a neural network vector field  $\vf_t^\btheta$, where $\btheta\in\real^P$ denotes the set of learnable model parameters.
The training objective of flow models is to optimize the vector field such that the terminal state $\rvx_1$ follows the target data distribution $p_{\text{data}}$. This core objective is formally defined as:
\begin{equation}\label{equation:flow_model_goal}
\rvx_0 \sim p_{\text{base}}, \quad \diff \rvx_t = \vf_t^\btheta(\rvx_t)\diff t 
\quad\implies\quad 
\rvx_1 \sim p_{\text{data}}.
\end{equation}
Through this training paradigm, the learned generative model is able to produce high-quality, realistic outputs, such as visually coherent images and realistic videos. This learning formulation motivates the development of \textit{flow matching} \citep{lipman2022flow, lipman2024flow}, a simple, scalable, and state-of-the-art algorithm for training neural vector fields $\vf_t^\btheta$.

The foundational step of flow matching is to define a \textit{probability path}, namely a time-dependent sequence of probability distributions $\{p_t\}$. Intuitively, a probability path constructs a smooth, gradual interpolation between the base noise distribution $p_{\text{base}}$ and the target data distribution $p_{\text{data}}$ (see Figure~\ref{fig:flow_interpolation}). This formulation raises a natural question regarding intermediate timesteps: while our ODE trajectory is strictly constrained by the boundary conditions $\rvx_0 \sim p_{\text{base}}$ at $t=0$ and $\rvx_1 \sim p_{\text{data}}$ at $t=1$, the distribution behavior for $0 < t < 1$ is flexible and unconstrained. A probability path mathematically formalizes this intermediate distribution evolution.

Given a data sample $\bx\in\real^D$ drawn from $p_{\text{data}}$, a \textit{conditional (interpolating) probability path} is a family of distributions $p_{t\mid 1}(\cdot\mid \bx)$ over $\real^D$ defined for all $\bx \in \real^D$, satisfying the following boundary conditions:
\begin{equation} \label{equation:flow_conditional_path}
p_{0|1}(\cdot\mid \bx) = p_{\text{base}}, \quad p_{1|1}(\cdot\mid \bx) = \delta_{\bx}, \quad (\text{conditional boundary conditions})
\end{equation}
where $\delta_{\by}\triangleq \delta(\bx-\by)$ denotes the \textit{Dirac delta distribution}  centered at $\by$ (i.e., sampling from $\delta_{\by}$ always returns $\by$).
The \textit{Dirac delta function} $\delta(\by)$ can be informally interpreted as an infinitely tall ``spike" at $\by=\bzero$ with the properties
\begin{equation}\label{equation:dirac_func}
\begin{aligned}
\delta(\by) &= 0, \quad \text{for }  \by \neq \bzero;  \\
\int_{-\infty}^\infty  \delta(\bx) \diff\bx &= 1
\quad\implies\quad 
\int_{-\infty}^\infty f(\bx)\delta(\bx-\by)\diff \bx = f(\by).
\end{aligned}
\end{equation}
In essence, each conditional probability path continuously transforms the initial base noise distribution $p_{\text{base}}$ into a discrete, single data point $\bx$ over the time horizon $t\in[0,1]$. Conceptually, this path can be interpreted as a continuous trajectory traversing the space of probability distributions.

Each conditional probability path $p_{t\mid 1}(\bz\mid \bx)$ induces a corresponding \textit{marginal probability path} $p_t(\bz)$. This marginal distribution is constructed by first sampling a data point $\bx \sim p_{\text{data}}\equiv p_1$ from the target data distribution, then sampling intermediate states from the conditional path $p_{t\mid 1}(\cdot\mid \bx)$. The formal definition and sampling procedure are summarized below:
\begin{subequations}\label{equation:flow_sampling_all}
\begin{align}
\rvx &\sim p_{\text{data}}, \quad \rvz \sim p_{t\mid 1}(\cdot\mid \bx) \quad \implies \rvz \sim p_t, \label{eq:flow_sampling_marginal} \\
\text{where }\; p_t(\bz) &= \int p_{t\mid 1}(\bz\mid \bx) p_{\text{data}}(\bx) \diff \bx;  \label{eq:flow_density_marginal}
\end{align}
\end{subequations}
Notably, while efficient sampling from the marginal distribution $p_t$ is feasible via the two-stage sampling process in \eqref{eq:flow_sampling_marginal}, explicitly computing the density value $p_t(\bz)$ is intractable due to the high-dimensional integral in \eqref{eq:flow_density_marginal}. Owing to the boundary properties of the conditional probability path $p_{t\mid 1}(\cdot\mid \bx)$ in \eqref{equation:flow_conditional_path}, the resulting marginal path $p_t$ achieves valid distribution interpolation. As $t$ progresses from $0$ to $1$, $p_t$ transitions smoothly from the base noise distribution $p_{\text{base}}$ to the target data distribution $p_{\text{data}}$, with all intermediate distributions serving as continuous blends of the two boundary distributions.

\begin{noteb}[Gaussian conditional probability path]\label{note:gaussian_conditional}
Among existing probability path formulations, the \textit{Gaussian probability path} stands out as the most widely adopted option, serving as the standard probability path for modern state-of-the-art models. 
We define two continuously differentiable, monotonic interpolation coefficients $\alpha_t, \beta_t$, referred to as \textit{noise schedulers}, that satisfy the boundary conditions $\alpha_0 = \beta_1 = 0$ and $\alpha_1 = \beta_0 = 1$. A simple and common instantiation is the \textit{linear scheduling setup} $\alpha_t=t$ and $\beta_t=1-t$. Using these coefficients, we construct the Gaussian conditional probability path as:
\begin{equation}\label{equation:gaussian_conditional}
p_{t\mid 1}(\cdot\mid \bx) = \normal(\alpha_t \bx, \beta_t^2 \bI_D).
\end{equation}
By the defined boundary properties of $\alpha_t$ and $\beta_t$, this formulation naturally satisfies the conditional path constraints introduced previously:
\begin{equation*}
p_{0|1}(\cdot\mid \bx) = \normal(\alpha_0 \bx, \beta_0^2 \bI_D) = \normal(\bzero, \bI_D) 
\quad \text{and} \quad 
p_{1|1}(\cdot\mid \bx) = \normal(\alpha_1 \bx, \beta_1^2 \bI_D) = \delta_{\bx}.
\end{equation*}
This holds because a Gaussian distribution with zero variance centered at $\bx$ converges to the Dirac delta distribution  $\delta_{\bx}$. 
Accordingly, this Gaussian-based conditional path fully complies with the boundary conditions in \eqref{equation:flow_conditional_path} under the standard base distribution $p_{\text{base}} = \normal(\bzero, \bI_D)$, confirming its validity as a legitimate conditional probability path.

Visual examples of this Gaussian interpolation process on image data are presented in Figure~\ref{fig:flow_interpolation}. The sampling procedure for the corresponding marginal probability path $p_t$ can be explicitly simplified to a closed-form interpolation operation: given a data sample $\bx \sim p_{\text{data}}$ and a Gaussian noise sample $\bepsilon \sim \normal(\bzero, \bI_D)$, the intermediate state $\bz \sim p_t$ is computed as:
\begin{equation}
\bx \sim p_{\text{data}}, \, \bepsilon \sim p_{\text{base}} = \normal(\bzero, \bI_D) 
\quad\implies\quad
\bz = \alpha_t \bx + \beta_t \bepsilon \sim p_t .
\end{equation}
Intuitively, this interpolation mechanism adjusts the noise magnitude adaptively over time. Lower values of $t$ introduce larger amounts of Gaussian noise into the data sample, gradually corrupting the original data. At $t=0$, the data component vanishes completely, leaving only pure Gaussian noise, which aligns with the definition of the base distribution.
\end{noteb}

\index{Conditional vector field}
\index{Marginal vector field}
\index{Marginalization theorem}
\index{Continuity equation}
\index{Divergence operator}
\subsection{Conditional and Marginal Vector Fields}\label{section:cd_mar_vf}

A probability path $\{p_t\}_{0 \leq t \leq 1}$ defines the distributions  $\rvx_t \sim p_t$ that the points $\rvx_t$ along a trajectory are intended to follow.
The key question is then: how can we construct a vector field under which trajectories $\rvx_t$ adhere to this probability path? Flow matching directly provides such a vector field---known as the \textit{marginal vector field}---which we introduce in this subsection.

\begin{figure}[h]
\centering  
\subfigtopskip=2pt 
\subfigbottomskip=9pt 
\subfigcapskip=-5pt 
\includegraphics[width=0.99\textwidth]{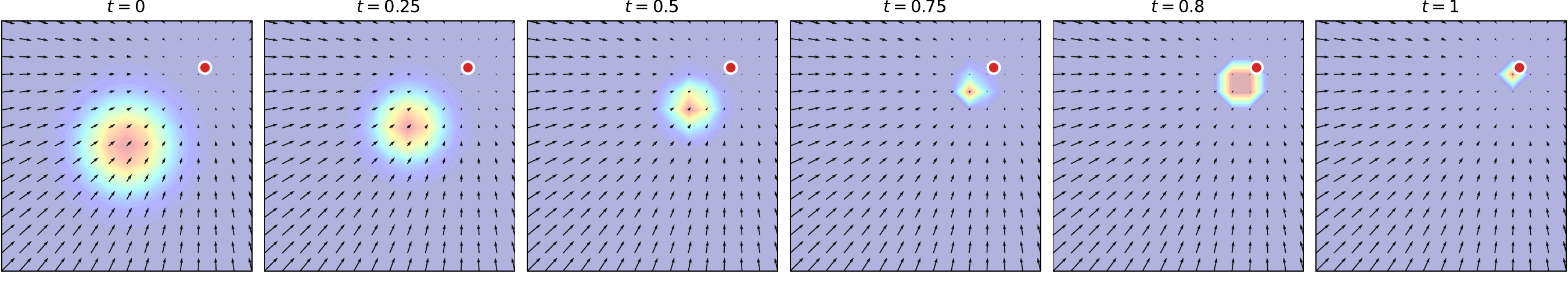}
\caption{ Simulation of ODE dynamics with a conditional vector field for conditional probability path modeling.}
\label{fig:flow_condition_path}
\end{figure}

To derive the marginal vector field, we first define the \textit{conditional vector field}.
Given a data point $\bx \in \real^D$, let $\vf_t^{\bthetastar}(\cdot\mid \bx)$ denote a {conditional vector field} 
such that the corresponding ODE realizes the conditional probability path $p_{t\mid 1}(\cdot\mid \bx)$ from \eqref{equation:flow_conditional_path}. 
Formally, this means
\begin{equation}\label{equation:condition_vf}
\rvx_0 \sim p_{\text{base}}, \quad \frac{\diff }{\diff t}\rvx_t =\vf_t^{\bthetastar}(\rvx_t\mid \bx) 
\quad \implies \quad
\rvx_t \sim p_{t\mid 1}(\cdot\mid \bx) \quad (0 \leq t \leq 1). 
\end{equation}
Meanwhile, by the equivalence between flows and vector fields (Theorem~\ref{theorem:flow_exist}),
there exists a unique flow (given in \eqref{equation:vecfield_from_flow}) satisfying
\begin{align}
\vf_t^{\bthetastar}(\bz\mid \bx) 
&= \dot{\bpsi}_t(\bpsi_t^{-1}(\bz\mid \bx)\mid \bx); 
\label{equ:cond_flow_vf} \\
\bpsi_t(\bz\mid \bx )
&=
\begin{cases}
\bz, & t=0;\\
\bx ,& t=1.
\end{cases} \label{equ:cond_flow_bound}
\end{align}

At first glance, the conditional vector field may appear uninformative, since every ODE trajectory terminates at the fixed endpoint $\rvx_1 = \bx$, meaning we are simply reconstructing known data points $\bx$; an illustration is given in Figure~\ref{fig:flow_condition_path}.
Nevertheless, the conditional vector field acts as a fundamental building block for constructing a vector field that produces genuine samples from the data distribution $p_{\text{data}}\equiv p_1$.
Once again, the overall framework of this core paradigm is visualized in Figure~\ref{fig:DDPM_gen_model_idea} (or Figure~\ref{fig:flow_goal}), which summarizes the key methodology adopted throughout this book.

\begin{theoremHigh}[Marginalization theorem]\label{theorem:flow_mag_tric}
Let $\vf_t^{\bthetastar}(\bz\mid \bx)$ denote the conditional vector field defined in \eqref{equation:condition_vf}. 
We define the \textit{marginal vector field} $\vf_t^{\bthetastar}(\bz)$ via
\begin{equation}\label{equation:marginal_vf_def}
\vf_t^{\bthetastar}(\bz) 
\triangleq  \int \vf_t^{\bthetastar}(\bz\mid \bx) p_{1\mid t}(\bx\mid \bz) \diff \bx
=\int \vf_t^{\bthetastar}(\bz\mid \bx) \frac{p_{t\mid 1}(\bz\mid \bx)p_{\text{data}}(\bx)}{p_t(\bz)} \diff \bx, 
\end{equation}
which exactly follows the \textit{marginal probability path} $p_t$. Formally,
\begin{equation}\label{equation:marginal_vf}
\rvx_0 \sim p_{\text{base}}, \quad \frac{\diff }{\diff t}\rvx_t = \vf_t^{\bthetastar}(\rvx_t) 
\quad \implies \quad
\rvx_t \sim p_t \quad (0 \leq t \leq 1). 
\end{equation}
In particular, the resulting terminal state satisfies $\rvx_1 \sim p_{\text{data}}$. 
In this sense, the vector field $\vf_t^{\bthetastar}$ transforms the base noise distribution $p_{\text{base}}$ into the target data distribution $p_{\text{data}}$.
We emphasize the conceptual distinction between the marginal dynamics \eqref{equation:marginal_vf} and the conditional dynamics \eqref{equation:condition_vf}.
\end{theoremHigh}

As stated by Bayes' rule \eqref{equation:bayes_base}, the quantity
\begin{equation*}
p_{1\mid t}(\bx\mid \bz)=
\frac{p_{t\mid 1}(\bz\mid \bx) p_{\text{data}}(\bx)}{p_t(\bz)} 
= \text{``posterior over data samples } \bx \text{ given noisy data } \bz\text{"},
\end{equation*}
defines the posterior distribution over clean data samples $\bx$, where $p_{\text{data}}(\bx)$ acts as the prior distribution. 
From this perspective, the marginal vector field is essentially a weighted average. For each candidate data point $\bx$, we take the conditional dynamics $\vf_t^{\bthetastar}(\bz\mid \bx)$---the vector field direction that drives trajectories toward $\bx$---and weight it by the posterior probability that the noisy state $\bz$ originates from the clean data point $\bx$. 
Intuitively, Theorem~\ref{theorem:flow_mag_tric} states that averaging these conditional vector fields over all possible data points yields the valid marginal vector field for the full probability path.

We can alternatively interpret this theorem using the conditional expectation formula in \eqref{equation:law_uncons_stat}. Consider the random pair $(\rvz,\rvx)$ with joint density $p_{t,1}(\bz, \bx) = p_{t\mid 1}(\bz\mid \bx)p_{\text{data}}(\bx)$, so that $\rvz \sim p_{t\mid 1}(\cdot\mid \rvx)$. Applying the conditional expectation formula rewrites the marginal vector field definition \eqref{equation:marginal_vf_def} as
$$
\vf_t^{\bthetastar}(\bz) = \Exp\left[\vf_t^{\bthetastar}(\rvz\mid \rvx) \mid \rvz = \bz\right].
$$
This formulation reveals that $\vf_t^{\bthetastar}(\bz)$ can be interpreted as the least-squares optimal approximation of the conditional vector field $\vf_t^{\bthetastar}(\rvz\mid \rvx)$ conditioned on the observation $\rvz = \bz$. 
Further discussion of this probabilistic least-squares interpretation is provided in Section~\ref{section:stat_prob}.

To prove Theorem~\ref{theorem:flow_mag_tric}, we rely on the \textit{continuity equation}, a fundamental relation widely used across mathematics and physics. 
We first define the \textit{divergence operator}  \texttt{div} as
\begin{equation}\label{equation:div_def}
\text{div}\big(\bv (\bz)\big) 
\equiv
\text{div}(\bv)(\bz) 
\triangleq  \sum_{i=1}^D \frac{\partial}{\partial z_i} v_i(\bz),
\quad\text{for all } \bz\in\real^D,
\end{equation}
where $v_i$ denote the $i$-th entry of $\bv(\bz)$.
Notably, the divergence  $\text{div}(\bv)(\bz)$ is equivalent to the trace of the Jacobian matrix $\partial _\bz\bv(\bz)$:
\begin{equation}\label{equation:div_trac}
\text{div}(\bv)(\bz) = \trace(\partial _\bz\bv(\bz)).
\end{equation}

\begin{lemma}[Continuity equation \citep{lipman2022flow, holderrieth2025introduction}]\label{lemma:cont_equa}
Consider a flow model driven by the vector field $\vf_t^{\bthetastar}$, where the initial state follows $\rvx_0 \sim p_{\text{base}} = p_0$. 
Then $\rvx_t \sim p_t$ for all $0 \leq t \leq 1$ if and only if
\begin{equation}\label{equation:cont_equa}
\partial_t p_t(\bz) = -\text{div}(p_t \vf_t^{\bthetastar})(\bz),
\quad \text{for all } \bz \in \real^D,\; 0 \leq t \leq 1,
\end{equation}
where $\partial_t p_t(\bz)\triangleq \frac{\diff }{\diff t} p_t(\bz)$ denotes the time derivative of the time-varying probability density $p_t(\bz)$.
\end{lemma}
The continuity equation can be viewed as a special case of the \textit{Fokker--Planck equation}, as further detailed in Lemma~\ref{lemma:fp_equation}.

To build intuition, we briefly interpret the continuity equation. The left-hand side $\partial_t p_t(\bz)$ quantifies the instantaneous temporal change in probability density at the state $\bz$. Intuitively, this change corresponds to the net influx of probability mass at that location. In flow-based generative models, particle trajectories $\rvx_t$ evolve according to the vector field $\vf_t^{\bthetastar}$. From physical intuition, divergence characterizes the net outward flux of a vector field. Consequently, the negative divergence encodes the net inward flux. When weighted by the local probability density $p_t(\bz)$, the term $-\text{div}(p_t \vf_t^{\bthetastar})$ exactly describes the net inflow of probability mass at $\bz$. Since the total probability mass remains normalized at all times, these two quantities must be equal, which validates the continuity equation.

We now present the formal proof of the marginalization result stated in Theorem~\ref{theorem:flow_mag_tric}.

\begin{proof}[of Theorem~\ref{theorem:flow_mag_tric}]
In view of Lemma~\ref{lemma:cont_equa}, it suffices to verify that the marginal vector field $\vf_t^{\bthetastar}$,  defined  in \eqref{equation:marginal_vf_def}, satisfies the continuity equation:
\begin{align*}
\partial_t p_t(\bz) &= \partial_t \int p_{t\mid 1}(\bz\mid \bx) \cdot p_{\text{data}}(\bx) \diff \bx = \int \partial_t p_{t\mid 1}(\bz\mid \bx) \cdot p_{\text{data}}(\bx) \diff \bx \\
&\stackrel{(\dag)}{=} \int -\text{div}\big(p_{t\mid 1}(\cdot\mid \bx) \vf_t^{\bthetastar}(\cdot\mid \bx)\big)(\bz) \cdot p_{\text{data}}(\bx) \diff \bx \\
&\stackrel{(\ddag)}{=} -\text{div} \left( \int p_{t\mid 1}(\bz\mid \bx) \vf_t^{\bthetastar}(\bz\mid \bx)\cdot p_{\text{data}}(\bx) \diff \bx \right) \\
&= -\text{div} \left( p_t(\bz) \int \vf_t^{\bthetastar}(\bz\mid \bx) \frac{p_{t\mid 1}(\bz\mid \bx)  p_{\text{data}}(\bx)}{p_t(\bz)} \diff \bx \right)(\bz) 
= -\text{div} \left( p_t \vf_t^{\bthetastar} \right)(\bz),
\end{align*}
where  the equality $(\dag)$ follows from the continuity equation for the conditional probability path $p_{t\mid 1}(\cdot\mid \bx)$, 
the equality $(\ddag)$ holds by interchanging the integral and divergence operators using \eqref{equation:div_def}, and the last equality follows from   \eqref{equation:marginal_vf_def}. 
Therefore, the continuity equation is fulfilled for $\vf_t^{\bthetastar}$. 
By Lemma~\ref{lemma:cont_equa},  \eqref{equation:marginal_vf} holds.
This completes the proof.
\end{proof}

We conclude this subsection by introducing the Gaussian conditional vector field and its associated flow models.
\begin{noteb}[Target ODE for Gaussian probability paths\index{Gaussian conditional vector field}\index{Gaussian CondOT probability path}]\label{note:target_ode_gaussian}
Following the preceding setup, let the conditional Gaussian probability path be $p_{t\mid 1}(\cdot\mid \bx) = \normal(\alpha_t \bx, \beta_t^2 \bI_D)$, where the noise scheduling functions $\alpha_t, \beta_t$ are defined in Note~\ref{note:gaussian_conditional}. 
Denote their time derivatives as $\dot{\alpha}_t = \partial_t \alpha_t$ and $\dot{\beta}_t = \partial_t \beta_t$. 
We now prove that the \textit{Gaussian conditional vector field}
\begin{equation}\label{equaiton:gauss_prob_path_weavg}
\vf_t^{\bthetastar}(\bz\mid \bx) 
= \left(\dot{\alpha}_t - \frac{\dot{\beta}_t}{\beta_t}\alpha_t\right)\bx  + \frac{\dot{\beta}_t}{\beta_t} \bz 
\end{equation}
constitutes a valid conditional vector field in the sense of Theorem~\ref{theorem:flow_mag_tric}.
Specifically, if the initial state satisfies $\rvx_0 \sim \normal(\bzero, \bI_D)$,  the ODE trajectories $\rvx_t$ induced by this vector field follow the conditional distribution $\rvx_t \sim p_{t\mid 1}(\cdot\mid \bx) = \normal(\alpha_t \bx, \beta_t^2 \bI_D)$.
As a special case, considering the noise scheduler configuration $\alpha_t=t, \beta_t=1-t$ (the corresponding probability $p_{t\mid 1}(\bz\mid \bx) = \normal(t\bx, (1-t)^2\bI_D)$ is sometimes referred to as the \textit{Gaussian CondOT (conditional  optimal
transport) probability path}), the Gaussian conditional vector field simplifies to
\begin{equation}\label{equaiton:gauss_prob_path_weavg_spec}
\vf_t^{\bthetastar}(\bz\mid \bx)
= \dot{\alpha}_t\bx+\dot{\beta}_t \bepsilon
= \bx-\bepsilon, \quad \bepsilon \sim \normal(\bzero, \bI_D).
\end{equation}

To verify the above results, we first construct a \textit{conditional flow model} $\bpsi_t^{\bthetastar}(\bz\mid \bx)$ via the affine mapping
\begin{equation}\label{equation:gauss_prob_path_flowm}
\bpsi_t^{\bthetastar}(\bz\mid \bx) = \alpha_t \bx + \beta_t \bz. 
\end{equation}
For each $t\in[0,1)$, this conditional flow is an affine transformation with respect to $\bz$ and satisfies the boundary conditions in \eqref{equ:cond_flow_bound}, hence it is referred to as an \textit{affine conditional flow}.
Let   $\rvx_t$ denote the ODE trajectory generated by $\bpsi_t^{\bthetastar}(\cdot\mid \bx)$, with initial distribution $\rvx_0 \sim p_{\text{base}} = \normal(\bzero, \bI_D)$. 
By definition of the flow mapping,
\begin{equation*}
\rvx_t = \bpsi_t^{\bthetastar}(\rvx_0\mid \bx) = \alpha_t \bx + \beta_t \rvx_0 \sim \normal(\alpha_t \bx, \beta_t^2 \bI_D) = p_{t\mid 1}(\cdot\mid \bx).
\end{equation*}
By the definition of the noise schedulers in \eqref{equation:gaussian_conditional}, we have $\alpha_1=1$ and $\beta_1=0$, which confirms that the flow trajectories align with the target conditional probability path and satisfy the condition \eqref{equation:condition_vf}.
We next derive the explicit form of the conditional vector field $\vf_t^{\bthetastar}(\bz\mid \bx)$ from the affine flow model $\bpsi_t^{\bthetastar}(\bz\mid \bx)$. 
Following the flow definition \eqref{eq:flow2}, we have the time evolution relation:
\begin{align*}
&&\gap \frac{\diff }{\diff t} \bpsi_t^{\bthetastar}(\bz\mid \bx) &= \vf_t^{\bthetastar}(\bpsi_t^{\bthetastar}(\bz\mid \bx)\mid \bx) \quad \text{for all } \bz, \bx \in \real^D \\
&&\stackrel{(\dag)}{\iff} \quad \dot{\alpha}_t \bx + \dot{\beta}_t \bz &= \vf_t^{\bthetastar}(\alpha_t \bx + \beta_t \bz\mid \bx) \quad \text{for all } \bz, \bx \in \real^D \\
&&\stackrel{(\ddag)}{\iff} \quad \dot{\alpha}_t \bx + \dot{\beta}_t \left(\frac{\bz - \alpha_t \bx}{\beta_t}\right) &= \vf_t^{\bthetastar}(\bz\mid \bx) \quad \text{for all } \bz, \bx \in \real^D \\
&&\iff \quad \left(\dot{\alpha}_t - \frac{\dot{\beta}_t}{\beta_t}\alpha_t\right) \bx + \frac{\dot{\beta}_t}{\beta_t} \bz &= \vf_t^{\bthetastar}(\bz\mid \bx) \quad \text{for all } \bz, \bx \in \real^D,
\end{align*}
where the implication $(\dag)$ follows from the definition of $\bpsi_t^{\bthetastar}(\bz\mid \bx)$ in \eqref{equation:gauss_prob_path_flowm}, and the implication $(\ddag)$ follows by reparameterizing $\bz \to (\bz - \alpha_t \bx)/\beta_t$. 
This confirms the conditional Gaussian vector field defined in \eqref{equaiton:gauss_prob_path_weavg}. 
\end{noteb}

\subsection{Tractable Likelihood Computation via Hutchinson's Estimator}\label{section:like_flow_hutchin}
A key advantage of flow-based generative models is their ability to compute exact log-likelihoods $\ln p_1(\bx)$ tractably for all $\bx \in \real^D$. This property follows from a continuity relation termed the \textit{instantaneous change of variables} \citep{chen2018neural}:
\begin{equation}\label{equation:inst_cgvar}
\frac{\diff}{\diff t} \ln p_t(\bpsi_t(\bz)) = -\text{div}(\vf_t)(\bpsi_t(\bz))
\stackrel{\eqref{equation:div_trac}}{\equiv} -\trace(\partial \vf_t(\bpsi_t(\bz))).
\end{equation}
This ODE describes the evolution of the log-likelihood  $\ln p_t(\bpsi_t(\bz))$ along the sampling trajectory $\bpsi_t(\bz)$ defined by the flow ODE \eqref{eq:flow2}. 
To derive  \eqref{equation:inst_cgvar}, we differentiate $\ln p_t(\bpsi_t(\bz))$ with respect to time and substitute the continuity equation \eqref{equation:cont_equa} alongside the flow ODE \eqref{eq:flow2}. Integrating both sides of \eqref{equation:inst_cgvar} from $t=0$ to $t=1$ and rearranging terms yields:	
\begin{equation}\label{equation:inst_cgvar2}
\ln p_1(\bpsi_1(\bz)) = \ln p_0(\bpsi_0(\bz)) - \int_0^1 \text{div}(\vf_t)(\bpsi_t(\bz)) \, \diff t,
\end{equation}
where $p_0=p_{\text{base}}$ denotes the base distribution and   $p_1=p_{\text{data}}$ denotes the target data distribution.

\index{Hutchinson's trace estimator}
\paragrapharrow{Hutchinson's trace estimator.}
In practice, direct computation of the divergence term  $\text{div}(\vf_t)(\bz) = \trace[\partial_\bz \vf_t(\bz)] \in \real^{D \times D}$ (see \eqref{equation:div_trac}) becomes increasingly challenging as the dimensionality $D$ increases. 
To resolve this issue, prior studies adopt unbiased estimation via \textcolor{
black}{\textit{Hutchinson's trace estimator}}, defined as \citep{grathwohl2018ffjord} (see Problem~\ref{prob:score_hutchinson}):
\begin{equation}
\text{div}(\vf_t)(\bz) = \trace\left[ \partial_\bz \vf_t(\bz) \right] = \Exp_\bepsilon \trace\left[ \bepsilon^\top \partial_\bz \vf_t(\bz) \bepsilon \right],
\end{equation}
where $\bepsilon\in\real^D$ is a random vector with zero mean and unit covariance. 
A standard choice is the standard Gaussian distribution $\bepsilon \sim \normal(\bzero, \bI)$.
Substituting the above trace estimator into \eqref{equation:inst_cgvar2} and interchanging the order of integration and expectation produces an unbiased estimator for the log-likelihood:
\begin{equation}\label{equation:unbias_hut_flow}
\ln p_1(\bpsi_1(\bz)) = \ln p_0(\bpsi_0(\bz)) - \Exp_\bepsilon \int_0^1 \trace\left[ \bepsilon^\top \partial_\bz \vf_t(\bpsi_t(\bz)) \bepsilon \right] \, \diff t.
\end{equation}
Unlike the original divergence computation in  $\text{div}(\vf_t)(\bpsi_t(\bz))$ in \eqref{equation:inst_cgvar} or \eqref{equation:inst_cgvar2}, for a specific $\bepsilon$, the matrix-vector product $\bA\bepsilon$ can be evaluated efficiently in a \textbf{single backward pass} using \textit{reverse-mode automatic differentiation}. 
The full trace can then be approximated via Monte Carlo sampling as:
\begin{equation}
\trace(\bA) \approx \frac{1}{M} \sum_{m=1}^M \bepsilon_m^\top \bA \bepsilon_m.
\end{equation}
{In practical implementations, $M=1$ is commonly adopted, where a single noise sample is resampled for each data point. Although this introduces stochastic noise into the likelihood estimate, the impact is negligible in practice. This is because the overall training objective is optimized via stochastic gradient descent, which inherently tolerates noisy gradients. Critically, the estimator remains strictly unbiased, meaning its expected value exactly matches the true trace and log-likelihood.}

\index{Reverse-mode automatic differentiation}
\paragrapharrow{Reverse-mode automatic differentiation.}
As discussed above, for a fixed noise sample $\bepsilon$, evaluating the trace term  $\trace\left[ \bepsilon^\top \partial_\bz \vf_t(\bpsi_t(\bz)) \bepsilon \right]$ in \eqref{equation:unbias_hut_flow} can be done with a {single backward pass} using \textit{reverse-mode automatic differentiation (backpropagation)}, reducing the complexity from $\mathcalO(D^2)$ to $\mathcalO(D)$.
Specifically, 
in a flow, we are computing the divergence of a vector field $\bff(\bz)\triangleq \vf_t(\bz): \real^D \to \real^D$. 
To do this, we first need to compute the vector $\bepsilon^\top \bJ\bepsilon$.
Naively, computing the full Jacobian matrix $\bJ$ takes $D$ passes of backpropagation. However, we can compute the product 
$\bepsilon^\top \bJ\bepsilon$ in a single pass.

To formalize the efficient trace estimation, we define the target quadratic form corresponding to Hutchinson's estimator as:
\begin{equation}
\text{Trace Estimator} = \bepsilon^\top \bJ \bepsilon,
\end{equation}
where $\bJ = \frac{\partial \bff}{\partial \bz}$ is the Jacobian matrix of size $D \times D$.
Calculating $\bJ$ explicitly requires $D$ passes of backpropagation (one for each output dimension), which is $\mathcalO(D^2)$. We can avoid forming $\bJ$ explicitly.
Let's define this scalar function as $L(\bz)$:
\begin{equation}
L(\bz) \triangleq  \bepsilon^\top \bff(\bz) = \sum_{i=1}^D \epsilon_i f_i(\bz).
\end{equation}
Now, we compute the gradient of this scalar function $L(\bz)$ with respect to the input $\bz$. Using the chain rule and the fact that  $\bepsilon$ is constant with respect to $\bz$:
\begin{equation}
\frac{\partial L}{\partial \bz} = \frac{\partial}{\partial \bz} \left( \bepsilon^\top \bff(\bz) \right)
=  \frac{\partial \bff}{\partial \bz}\bepsilon  =  \bJ^\top\bepsilon.
~\footnote{
Note that depending on the layout convention (numerator vs. denominator layout), this might appear as  $ \bJ\bepsilon $  or  $ \bJ^\top\bepsilon $. 
In the context of Hutchinson's estimator  $ \bepsilon^\top \bJ \bepsilon $, since the result is a scalar,  $ \bepsilon^\top (\bJ \bepsilon) = (\bJ^\top \bepsilon)^\top \bepsilon $. 
Most deep learning libraries (like PyTorch's `autograd.grad') compute the vector-Jacobian product (VJP). 
To get the Jacobian-vector product (JVP), one often uses specific tricks or forward-mode automatic differentiation; but for the specific case of the trace estimator  $ \bepsilon^\top \bJ \bepsilon $, computing the gradient of the dot product works perfectly.}
\end{equation}
To get the final scalar estimator $\bepsilon^\top \bJ \bepsilon$, we simply take the dot product of this gradient with $\bepsilon$ again:
\begin{equation}
\bepsilon^\top \bJ \bepsilon =  \left( \frac{\partial L}{\partial \bz} \right)^\top \bepsilon .
\end{equation}
In practice, deep learning frameworks (like PyTorch or TensorFlow) implement this efficiently. We do not need to derive the equations manually; we use the built-in \texttt{grad} function.
The trick is known as \textit{reverse-mode automatic differentiation (backpropagation)} and can be concluded in four steps:
\begin{enumerate}[(i)]
\item \textit{Forward pass.} Compute the output of the vector field $\bff(\bz)$ for the input coordinate $\bz$.
\item \textit{Scalar projection.} Calculate the scalar objective $L = \bepsilon^\top \bff(\bz)$. This is a simple dot product (or sum of elementwise multiplication).
\item \textit{Reverse pass (backpropagation).} Evaluate the gradient of the scalar objective $L$ with respect to the input $\bz$.
\begin{equation}
\bg = \nabla_{\bz} L = \nabla_{\bz} \left( \sum_{i=1}^D \epsilon_i f_i(\bz) \right)
\end{equation}
By definition of the gradient, $\bg = \bJ^\top \bepsilon$.
\item \textit{Final dot product.} Compute the scalar trace estimate via $\bg^\top \bepsilon$.
\end{enumerate}
For the complexity, we can conclude that 
\begin{itemize}
\item \textit{Naive approach.} Computing the full Jacobian $\bJ$ requires $D$ backward passes (computing $\partial f_i / \partial \bz$ for each $i$). This results in a quadratic computational cost of $\mathcalO(D) \times \mathcalO(D) = \mathcalO(D^2)$.
\item \textit{Reverse-mode automatic differentiation.} We treat $L=\bepsilon^\top \bff(\bz)$ as a single scalar loss. Computing the gradient of a scalar loss with respect to inputs requires exactly \textbf{one} backward pass, yielding a linear computational cost of $\mathcalO(D)$.
\end{itemize}
This optimization allows the Hutchinson trace estimator to be evaluated in linear time with respect to input dimensionality $D$, rather than the prohibitive quadratic time complexity of naive Jacobian-based computation, making high-dimensional flow model training feasible.

\paragrapharrow{Obtaining the log-likelihood.}
As established, naive explicit Jacobian trace computation incurs prohibitive $\mathcalO(D^2)$ complexity. 
By combining Hutchinson's trace estimator with reverse-mode automatic differentiation, we reduce this computational overhead to $\mathcalO(D)$, enabling efficient log-likelihood estimation for high-dimensional inputs.
In summary, unbiased estimation of the data log-likelihood $\ln p_1(\bx)$ relies on parallel backward simulation of two coupled flow ODEs (implemented via the Euler method in Algorithm~\ref{alg:samp_bkds_nfm_euler}). The two evolving components are defined as follows:
\begin{enumerate}[(i)]
\item \textit{Trajectory flow $\bphi(t)$.} This term corresponds to the standard flow trajectory  $\bpsi_t(\cdot)$,  governed by the original flow ODE:
$$
\frac{\diff}{\diff t}\bphi(t) = \vf_t(\bphi(t)).
$$

\item \textit{Integral term  flow $\omega(t)$:} This auxiliary state tracks the cumulative integral of the estimated trace term from time $t$ to $1$:
$$
\omega(t) = \int_t^1  \left( \bepsilon^\top \partial_\bz \vf_s(\bpsi_s(\bz)) \bepsilon \right) \diff s.
$$
At time $t=1$, $\omega(1)=0$.
Its time derivative, can be obtained using the {fundamental theorem of calculus} (see Problems~\ref{prob:fund_calc_trace1}--\ref{prob:fund_calc_trace2}):
$$
\frac{\diff}{\diff t}\omega(t) = -  \left( \bepsilon^\top \partial_\bz \vf_t(\bphi(t)) \bepsilon \right).
$$

\end{enumerate}
Coupling these two dynamics yields a unified backward-time ODE system evolving from $t=1$ to $t=0$:
\begin{subequations}
\begin{align}
\frac{\diff}{\diff t} 
\begin{bmatrix} 
\bphi(t) \\ 
\omega(t) 
\end{bmatrix} 
&= \begin{bmatrix} \vf_t(\bphi(t)) \\ -\left(  \bepsilon^\top \partial_\bz \vf_t(\bphi(t)) \bepsilon \right) \end{bmatrix};  
\quad &&(\text{flow ODE})\\
\begin{bmatrix} \bphi(1) \\ \omega(1) \end{bmatrix} 
&= \begin{bmatrix} \bx \\ 0 \end{bmatrix}, \quad &&(\text{flow initial conditions})
\end{align}
\end{subequations}
where $\bx$ denotes a sample drawn from the target data distribution  $p_1=p_{\text{data}}$.
The final log-likelihood estimate from \eqref{equation:inst_cgvar2} or \eqref{equation:unbias_hut_flow} for $\bx$ is formulated as:
\begin{equation}
\ln \widehat{p_1(\bx)} = \ln p_0(\bphi(0)) - \omega(0), 
\quad p_0=p_{\text{base}}.
\end{equation}
The full discrete-time implementation using the Euler method is summarized in Algorithm~\ref{alg:backward_flow_loglike}.

\begin{algorithm}[h]
\caption{Backward Simulation of Continuous Normalizing Flow (Euler Method)}
\label{alg:backward_flow_loglike}
\begin{algorithmic}[1]
\Require Time-dependent vector field $\vf_t(\cdot)$, number of steps $n$, data sample $\bx \sim p_1$;
\Ensure Latent sample $\bx_0 = \bpsi_0(\bx)$, log-likelihood estimate $\ln\widehat{ p_1(\bx)}$;

\State $\tau \gets 1 / n$; \Comment{Step size}
\State $t \gets 1$; \Comment{Start from final time}
\State $\bx_t \gets \bx$; \Comment{Initialize flow state}
\State $g_t \gets 0$; \Comment{Initialize trace integral}
\State Sample $\bepsilon \sim \normal(\bzero, \bI)$; \Comment{For Hutchinson's trace estimator}

\For{$i = 1, 2, \ldots,n$}
\State $\vf \gets \vf_t(\bx_t)$; \Comment{Evaluate vector field}
\State $\text{tr} \gets \innerproduct{\bepsilon, \nabla_{\bx_t} \vf_t(\bx_t) \bepsilon}$; \Comment{Trace term estimate}
\State $\bx_{t-\tau} \gets \bx_t - \tau \cdot \vf$; \Comment{Backward Euler step for flow}
\State $g_{t-\tau} \gets g_t + \tau \cdot \text{tr}$; \Comment{Accumulate integral}
\State $\bx_t \gets \bx_{t-\tau}, \quad g_t \gets g_{t-\tau}$;
\State $t \gets t - \tau$;
\EndFor

\State $\bx_0 \gets \bx_t$;
\State $\ln\widehat{ p_1(\bx)} \gets \ln p_0(\bx_0) - g_t$;
\State \Return $\bx_0$, $\ln\widehat{ p_1(\bx)}$;
\end{algorithmic}
\end{algorithm}

\subsection{Learning the Marginal Vector Field}\label{section:train_marg_vf}

We now proceed to formalize the training algorithm. The core objective of flow matching is to train a neural network $\vf_t^\btheta$---parameterized by learnable weights $\btheta\in\real^P$---to approximate the target marginal vector field $\vf_t^{\bthetastar}$. 
Per Theorem~\ref{theorem:flow_mag_tric}, exact alignment between $\vf_t^\btheta$ and $\vf_t^{\bthetastar}$ guarantees that the terminal samples $\rvx_1 \sim p_{\text{data}}$ follow the target data distribution.
A natural and intuitive strategy to achieve $\vf_t^\btheta \approx \vf_t^{\bthetastar}$ is to minimize a mean squared error objective~\footnote{An alternative loss function is discussed in Problems~\ref{prob:breg_dist}--\ref{prob:breg_dist_flow}.}, termed the \textit{flow matching (FM) loss}, which is defined as:
\begin{mybox}
\begin{equation}\label{equation:fmloss_def}
\begin{aligned}
	\mathcalJ_{\text{FM}}(\btheta) 
	&\triangleq  \Exp_{t \sim \uniformdist, \bz \sim p_t} \big[\normtwobig{\vf_t^\btheta(\bz) - \vf_t^{\bthetastar}(\bz)}^2\big] \\
	&= \Exp_{t \sim \uniformdist, \bx \sim p_{\text{data}}, \bz \sim p_{t\mid 1}(\cdot\mid \bx)} \big[\normtwobig{\vf_t^\btheta(\bz) - \vf_t^{\bthetastar}(\bz)}^2\big], 
\end{aligned}
\end{equation}
\end{mybox}
where $p_t(\bz) = \int p_{t\mid 1}(\bz\mid \bx)p_{\text{data}}(\bx) \diff \bx$ denotes the marginal probability path,
and the final equality derives from the sampling procedure outlined in \eqref{equation:flow_sampling_all}. 
Here, ``$\uniformdist$" denotes a uniform distribution over the interval $[0,1]$.
Conceptually, this loss operates via the following sampling pipeline: first, sample a continuous time variable $t \in [0,1]$ uniformly at random. Second, draw a data sample $\bx$ from the empirical data distribution $p_{\text{data}}$, then sample a latent point $\bz$ from the conditional path distribution $p_{t\mid 1}(\cdot\mid \bx)$ (e.g., via noise injection) and evaluate the network output $\vf_t^\btheta(\bz)$. Finally, compute the squared Euclidean discrepancy between the network prediction and the target marginal vector field $\vf_t^{\bthetastar}(\bz)$.

However, this formulation presents a critical limitation. While Theorem~\ref{theorem:flow_mag_tric} provides an exact analytical expression for the marginal vector field $\vf_t^{\bthetastar}$, the associated integral is computationally \textbf{intractable}, precluding direct optimization of the standard flow matching loss.
To resolve this issue, we leverage the tractability of the conditional vector field $\vf_t^{\bthetastar}(\bz\mid \bx)$. We accordingly define the \textit{conditional flow matching (CFM) loss}:
\begin{mybox}
\begin{equation}\label{equation:cfmloss_def}
\mathcalJ_{\text{CFM}}(\btheta) 
\triangleq  \Exp_{t \sim \uniformdist, \bx \sim p_{\text{data}}, \bz \sim p_{t\mid 1}(\cdot\mid \bx)} \big[\normtwobig{\vf_t^\btheta(\bz) - \vf_t^{\bthetastar}(\bz\mid \bx)}^2\big]. 
\end{equation}
\end{mybox}
This objective differs from the original flow matching loss in \eqref{equation:fmloss_def} by replacing the intractable marginal vector field $\vf_t^{\bthetastar}(\bz)$ with its tractable conditional counterpart $\vf_t^{\bthetastar}(\bz\mid \bx)$. 
Since closed-form expressions exist for $\vf_t^{\bthetastar}(\bz\mid \bx)$, the conditional loss is readily optimizable. 

A natural question arises: why optimize against the conditional vector field when the ultimate goal is to match the marginal vector field? Critically, explicit regression on the tractable conditional vector field implicitly induces regression on the intractable marginal vector field. The following theoretical result rigorously validates this key intuition.
\begin{theoremHigh}[FM  and CFM losses \citep{holderrieth2025introduction}]\label{theorem:fmcfm_equiv}
The marginal FM loss and CFM loss differ only by a parameter-independent constant. Their relationship can be formulated as:
\begin{equation*}
\mathcalJ_{\text{FM}}(\btheta) = \mathcalJ_{\text{CFM}}(\btheta) + C,
\end{equation*}
where the constant  $C$ is independent of  the learnable network parameters $\btheta$. 
As a direct consequence, the two loss functions yield identical parameter gradients:
\begin{equation*}
\nabla_\btheta \mathcalJ_{\text{FM}}(\btheta) = \nabla_\btheta \mathcalJ_{\text{CFM}}(\btheta).
\end{equation*}
Accordingly, stochastic gradient descent (SGD) and other standard optimization procedures minimize $\mathcalJ_{\text{CFM}}(\btheta)$ in a manner fully equivalent to optimizing the original intractable $\mathcalJ_{\text{FM}}(\btheta)$. 
\end{theoremHigh}
\begin{proof}[of Theorem~\ref{theorem:fmcfm_equiv}]
The proof proceeds by expanding the mean squared error objective of the FM loss and isolating terms independent of $\btheta$. We first expand the squared norm in the original FM loss definition:
\begin{align*}
\mathcalJ_{\text{FM}}(\btheta) &= \Exp_{t \sim \uniformdist, \bz \sim p_t} \big[\normtwobig{\vf_t^\btheta(\bz) - \vf_t^{\bthetastar}(\bz)}^2\big] \\
&= \Exp_{t \sim \uniformdist, \bz \sim p_t} \big[\normtwobig{\vf_t^\btheta(\bz)}^2 - 2 \vf_t^\btheta(\bz)^\top \vf_t^{\bthetastar}(\bz) + \normtwobig{\vf_t^{\bthetastar}(\bz)}^2\big] \\
&= \Exp_{t \sim \uniformdist, \bx \sim p_{\text{data}}, \bz \sim p_{t\mid 1}(\cdot\mid \bx)} \big[\normtwobig{\vf_t^\btheta(\bz)}^2\big] - 2 \Exp_{t \sim \uniformdist, \bz \sim p_t} [\vf_t^\btheta(\bz)^\top \vf_t^{\bthetastar}(\bz)] + C_1,
\end{align*}
where we define $C_1\triangleq \Exp_{t \sim \uniformdist[0,1], \bz \sim p_t} [\normtwobig{\vf_t^{\bthetastar}(\bz)}^2]$, a fixed constant that does not depend on the network parameters $\btheta$. The final equality holds due to the marginal sampling procedure for $p_t$ specified in \eqref{equation:flow_sampling_all}.
The middle expectation term involves the intractable marginal vector field $\vf_t^{\bthetastar}(\bz)$. We resolve this by substituting the marginalization identity for vector fields in \eqref{equation:marginal_vf_def}, which rewrites the marginal vector field as an integral over the tractable conditional vector field $\vf_t^{\bthetastar}(\bz\mid \bx)$. This yields the following transformation for the middle term:
\begin{equation*}
\small
\begin{aligned}
\Exp_{t \sim \uniformdist, \bz \sim p_t}[\vf_t^\btheta(\bz)^\top \vf_t^{\bthetastar}(\bz)] 
&= \int_0^1 \int p_t(\bz) \vf_t^\btheta(\bz)^\top \vf_t^{\bthetastar}(\bz) \, \diff \bz \, \diff t \\
&= \int_0^1 \int p_t(\bz) \vf_t^\btheta(\bz)^\top \left[ \int \vf_t^{\bthetastar}(\bz\mid \bx) \frac{p_{t\mid 1}(\bz\mid \bx)p_{\text{data}}(\bx)}{p_t(\bz)} \, \diff \bx \right] \, \diff \bz \, \diff t \\
&=\Exp_{t \sim \uniformdist, \bx \sim p_{\text{data}}, \bz \sim p_{t\mid 1}(\cdot\mid \bx)}[\vf_t^\btheta(\bz)^\top \vf_t^{\bthetastar}(\bz\mid \bx)].
\end{aligned}
\end{equation*}
Substituting this result back into the expanded FM loss expression gives:
\begin{equation*}
\small
\begin{aligned}
\mathcalJ_{\text{FM}}(\btheta) 
&= \Exp_{t \sim \uniformdist, \bx \sim p_{\text{data}}, \bz \sim p_{t\mid 1}(\cdot\mid \bx)}
\big[\normtwobig{\vf_t^\btheta(\bz)}^2\big] - 2\Exp_{t \sim \uniformdist, \bx \sim p_{\text{data}}, \bz \sim p_{t\mid 1}(\cdot\mid \bx)}[\vf_t^\btheta(\bz)^\top \vf_t^{\bthetastar}(\bz\mid \bx)] + C_1 \\
&= \mathcalJ_{\text{CFM}}(\btheta) + {C_2 + C_1},
\end{aligned}
\end{equation*}
where $C_2\triangleq \Exp_{t \sim \uniformdist, \bx \sim p_{\text{data}}, \bz \sim p_{t\mid 1}(\cdot\mid \bx)}[-\normtwobig{\vf_t^{\bthetastar}(\bz\mid \bx)}^2]$ denotes a constant in terms of $\btheta$. 
Defining the total constant  $C\triangleq C_1+C_2$ completes the proof of the loss equivalence relation.
\end{proof}

The above equivalence theorem implies that, given a sufficiently expressive neural network parameterization, the minimizer $\widehatbtheta$ of the CFM loss $\mathcalJ_{\text{CFM}}(\btheta)$ exactly recovers the target marginal vector field, such that  $\vf_t^{\widehatbtheta}(\bz) = \vf_t^{\bthetastar}(\bz)$. This validates the core training paradigm of flow matching, which optimizes the tractable conditional flow matching loss as a viable surrogate for the original intractable marginal flow matching objective. The complete flow matching training pipeline is summarized in Algorithm~\ref{alg:flow_matching} (e.g, for Gaussian conditional optimal transport (CondOT) probability paths; see Note~\ref{note:fm_got}).

\begin{algorithm}
\caption{Flow Matching Training Procedure (for Gaussian CondOT Path $p_{t\mid 1}(\bz\mid \bx) = \normal(t\bx, (1-t)^2\bI)$)}
\label{alg:flow_matching}
\begin{algorithmic}[1]
\Require A dataset of samples $\bx \sim p_{\text{data}}$, neural network $\vf_t^\btheta$;
\State \textbf{initialize:} $\btheta$;
\For{$cnt=0,1,2,\ldots$}
\State Sample a data example $\bx$ from the dataset;
\State Sample a random time $t \sim \uniformdist{[0,1]}$;
\State Sample noise $\bepsilon \sim \normal(\bzero, \bI_D)$;
\State Set $\bz \sim p_{t\mid 1}(\cdot \mid \bx)$; \Comment{Gaussian CondOT: $\bz \leftarrow t\bx + (1-t)\bepsilon$} 
\State Loss $\mathcalJ(\btheta) \leftarrow \normtwobig{\vf_t^\btheta(\bz) - \vf_t^{\bthetastar}(\bz\mid \bx)}^2  $;
\Comment{Gaussian CondOT: $=\normtwobig{\vf_t^\btheta(\bz) - (\bx - \bepsilon)}^2$}
\State Update $\btheta \leftarrow \btheta-\eta \nabla \mathcalJ(\btheta)$; \Comment{Gradient descent}
\EndFor
\State \Return $\btheta$;
\end{algorithmic}
\end{algorithm}

This algorithm exhibits three notable and advantageous properties. First, the training process is entirely \textit{simulation-free}: it eliminates the need to solve or roll out ODE trajectories during optimization, which substantially reduces computational overhead compared to traditional flow-based generative models. Second, the training objective reduces to a simple regression task, where the network is trained to fit the analytically known conditional vector field $ \vf_t^{\bthetastar}(\bz\mid \bx)$, making the optimization framework analogous to standard supervised learning. Third, the formulation is structurally concise with minimal computational complexity, rendering it highly practical for large-scale machine learning systems. These merits establish flow matching as a highly competitive approach for high-dimensional generative modeling tasks.
After training the vector field network $ \vf_t^\btheta $, data samples adhering to the target data distribution $p_{\text{data}}$ can be generated by simulating the learned flow ODE:
\begin{equation}
\diff \rvx_t = \vf_t^\btheta(\rvx_t) \diff t, \quad \rvx_0 \sim p_{\text{base}}.
\end{equation}
Numerical integration schemes such as the Euler method (detailed in Algorithm~\ref{alg:samp_nfm_euler}) can be adopted to solve the ODE and produce valid data samples $\rvx_1 \sim p_{\text{data}}$.

\paragrapharrow{Timestep variant.}
A practical extension of the standard CFM loss involves non-uniform time sampling. Instead of sampling the interpolation time $t$ uniformly over $[0,1]$, we can draw $t$ from an arbitrary probability density function $q(t)$ defined on the same interval. This modification yields a \textit{weighted CFM loss} objective, formulated as:
\begin{mybox}
\begin{equation}\label{equation:cfmloss_def_time}
\begin{aligned}
\mathcalJ'_{\text{CFM}}(\btheta) 
&= \Exp_{t \sim q, \bx \sim p_{\text{data}}, \bz \sim p_{t\mid 1}(\cdot\mid \bx)} \big[\normtwobig{\vf_t^\btheta(\bz) - \vf_t^{\bthetastar}(\bz\mid \bx)}^2\big]\\
&=\Exp_{t \sim \uniformdist, \bx \sim p_{\text{data}}, \bz \sim p_{t\mid 1}(\cdot\mid \bx)}\big[ q(t)\cdot \normtwobig{\vf_t^\btheta(\bz) - \vf_t^{\bthetastar}(\bz\mid \bx)}^2\big].
\end{aligned}
\end{equation}
\end{mybox}
While the uniform-sampled and weighted formulations are mathematically equivalent in expectation, empirical studies demonstrate that sampling $t$ from a tailored distribution $q(t)$ delivers superior performance in large-scale image generation tasks compared to applying static weight coefficients to uniform time samples \citep{esser2024scaling}.

To conclude this section, we explicitly instantiate the full flow matching framework for Gaussian conditional probability paths, consistent with the theoretical setup introduced in Notes~\ref{note:gaussian_conditional} and \ref{note:target_ode_gaussian}.
\begin{noteb}[Flow matching for Gaussian conditional probability paths\index{Rectified flow}]\label{note:fm_got}
We revisit the general form of the  Gaussian conditional probability path formulation $p_{t\mid 1}(\cdot\mid \bx) = \normal(\alpha_t \bx; \beta_t^2 \bI_D)$.
Samples along this continuous path can be efficiently generated via a simple noise injection procedure. By sampling standard Gaussian noise $ \bepsilon \sim \normal(\bzero, \bI_D) $, the latent intermediate variable $ \bz_t $ is constructed as:
\begin{equation}\label{equation:fm_got1}
\bepsilon \sim \normal(\bzero, \bI_D) 
\quad \implies \quad
\bz_t = \alpha_t \bx + \beta_t \bepsilon \sim \normal(\alpha_t \bx, \beta_t^2 \bI_D) = p_{t\mid 1}(\cdot\mid \bx). 
\end{equation}
As previously derived in \eqref{equaiton:gauss_prob_path_weavg}, the closed-form expression for the target conditional vector field $\vf_t^{\bthetastar}(\bz\mid \bx)$ is 
\begin{equation}
\vf_t^{\bthetastar}(\bz\mid \bx) = \left(\dot{\alpha}_t - \frac{\dot{\beta}_t}{\beta_t}\alpha_t\right)\bx + \frac{\dot{\beta}_t}{\beta_t}\bz, 
\end{equation}
where $\dot{\alpha}_t = \partial_t \alpha_t$ and $\dot{\beta}_t = \partial_t \beta_t$ denote the time derivatives of the path scaling and diffusion coefficients, respectively. 
Substituting this analytical vector field into the standard CFM loss yields the general objective for Gaussian probability paths:
\begin{align}
\mathcalJ_{\text{CFM}}(\btheta) &= \Exp_{t \sim \uniformdist, \bx \sim p_{\text{data}}, \bz \sim \normal(\alpha_t \bx, \beta_t^2 \bI_D)} 
\Big[\normtwobig{\vf_t^\btheta(\bz) - \left(\dot{\alpha}_t - \frac{\dot{\beta}_t}{\beta_t}\alpha_t\right)\bx - \frac{\dot{\beta}_t}{\beta_t}\bz}^2\Big]  \\
&= \Exp_{t \sim \uniformdist, \bx \sim p_{\text{data}}, \bepsilon \sim \normal(\bzero, \bI_D)} \big[\normtwobig{\vf_t^\btheta(\alpha_t \bx + \beta_t \bepsilon) - (\dot{\alpha}_t \bx + \dot{\beta}_t \bepsilon)}\big],
\end{align}
where the last equality follows by  plugging in \eqref{equation:fm_got1} and replacing $\bz$ with $\alpha_t \bx + \beta_t \bepsilon$. 
This final form highlights the striking simplicity of the CFM training objective: the pipeline only requires sampling a real data point $\bx$, drawing standard Gaussian noise $\bepsilon$, and minimizing the corresponding mean squared regression error.

We further simplify the formulation by considering the widely adopted CondOT path configuration with $\alpha_t = t $ and $\beta_t = 1 - t $ (also known as a special \textit{rectified flow} \citep{liu2022flow}). For this specific setting, the time derivatives simplify to $\dot{\alpha}_t = 1 $ and $\dot{\beta}_t = -1 $, yielding the specialized Gaussian CFM loss:
\begin{equation*}
\mathcalJ_{\text{CFM}}^{\text{G}}(\btheta) = \Exp_{t \sim \uniformdist, \bx \sim p_{\text{data}}, \bepsilon \sim \normal(\bzero, \bI_D)} 
\big[\normtwobig{\vf_t^\btheta(t\bx + (1-t)\bepsilon) - (\bx - \bepsilon)}^2\big].
\end{equation*}
\end{noteb}

\begin{problemset}
\item \label{prob:prod_deter} 	
For any two square matrices $\bA, \bB \in \real^{D \times D}$ of identical dimension, prove that the determinant of a matrix product equals the product of the individual determinants:
$$
\det(\bA\bB) = \det(\bA) \det(\bB).
$$
By induction, this property extends to any finite sequence of square matrices:
$$
\det(\bA_1 \bA_2 \ldots \bA_K)
= \det(\bA_1) \det(\bA_2) \ldots \det(\bA_K).
$$

\item  \label{prob:fund_calc_trace1} For any continuous function $h(s)$, prove the following differentiation rule for integrals with variable lower bounds:
$$
g(t) = \int_t^b h(s) \, ds 
\quad \implies \quad 
\frac{\diff}{\diff t} g(t) = -h(t).
$$
\textit{Hint: Use the fact that $
\int_t^b h(s) \, ds = \int_0^b h(s) \, ds - \int_0^t h(s) \, ds
$, and differentiate both sides w.r.t. $t$.}

\item \label{prob:fund_calc_trace2}
Let 
$ \omega(t) = \int_t^1  \left( \bepsilon^\top \partial_\bz \vf_s(\bz) \bepsilon \right) \diff s.$ At time $t=1$, $\omega(1)=0$.
Show that 
$$
\frac{\diff}{\diff t}\omega(t) = -  \left( \bepsilon^\top \partial_\bz \vf_t(\bz) \bepsilon \right).
$$
\textit{Hint: Use Problem~\ref{prob:fund_calc_trace1}.}

\item \textbf{Bregman divergence.}\label{prob:breg_dist} 
The \textit{Bregman distance (or Bregman divergence)} generalizes the concept of squared Euclidean distance (mean squared loss) and measures the difference between two points in a space.
It is defined for a  convex and differentiable function $ \phi: \sS \rightarrow \real $, where $ \sS \subseteq \real^D $ is a convex set. The Bregman divergence from point $ \by $ to point $ \bx $ with respect to $ \phi $ is given by:
$$
\mathcalD_{\phi}(\bx,\by) = \phi(\bx) - \phi(\by) - \innerproduct{\nabla \phi(\by), \bx - \by }.
$$
Show that $\mathcalD_{\phi}(\bx,\by)\geq 0$ for any $\bx,\by\in\real^D$.
Show that the Bregman divergence gradient is \textit{affine invariant} with respect to the first argument:
$$
\nabla_\by \mathcalD_{\phi}(a\bx_1+b\bx_2,\by)
= a\cdot \nabla_\by \mathcalD_{\phi}(\bx_1,\by)+ b\cdot \nabla_\by\mathcalD_{\phi}(\bx_2,\by), 
\quad\text{for any } a+b=1.
$$

\item \label{prob:breg_dist_flow}  \textbf{Flow matching with Bregman.}
Define the FM and CFM losses in \eqref{equation:fmloss_def} and \eqref{equation:cfmloss_def} using the Bregman divergence rather than the mean squared loss. Show that Theorem~\ref{theorem:fmcfm_equiv} still holds.
\textit{Hint: Use Problem~\ref{prob:breg_dist}.}


\item \label{prob:score_hutchinson} \textbf{Hutchinson's trace estimator.} 
Let $\bA\in\real^{D\times D}$ be an arbitrary square matrix, and let $\bzeta$ follows a distribution with zero mean and unit variance, i.e., $\Exp[\bzeta\bzeta^\top]=\bI_D$. 
Show that $\trace(\bA) = \Exp[\bzeta^\top\bA\bzeta]$. \textit{Hint: Use the trace trick $\trace(\bB\bC)=\trace(\bC\bB)$.}

\item Let the vector field $v_t(x) = 2\sqrt{x}$. Does the unique ODE exist? \textit{Hint: Consider $x_t=0$ or $x_t=t^2$ and the condition in  Theorem~\ref{theorem:flow_exist}.}


\item \label{prob:ode_sol_ftut} Consider the following  probability flow ODE
$$
\frac{\diff\rvx_t}{\diff t} = \bF_t \rvx_t +\bu_t.
$$
Show that the exact solution to this ODE is
$$
\bx_t = \bPsi(t, s) \bx_s + \int_s^t \bPsi(t, \tau)\bu_\tau \diff \tau,
$$
where the transition matrix $\bPsi(t, s)$ satisfies
$\frac{\partial}{\partial t} \bPsi(t, s) = \bF_t \bPsi(t, s)$, $ \bPsi(s, s) = \bI$.
\textit{Hint: 
Prove that the transition matrix satisfies (1). $\frac{\partial \bPsi}{\partial t}(s, t) = -\bPsi(s, t) \bF_t$ (derived from the inverse property of $\bPsi$), (2). $\bPsi(t, s) \bPsi(s, t) = \bI$, and (3). $\bPsi(t, s) \bPsi(s, \tau) = \bPsi(t, \tau)$.
Consider the Leibniz rule for integrals: 
$$
\frac{\diff}{\diff t} \int_{a(t)}^{b(t)} f(x, t)  \diff x = \int_{a(t)}^{b(t)} \frac{\partial f}{\partial t}(x, t) \diff x + f(b(t), t) \cdot b'(t) - f(a(t), t) \cdot a'(t).
$$
}


\end{problemset}

\newpage 
\chapter{Score Matching and Stochastic Differential Equations}\label{chapter:scorematch}
\begingroup
\hypersetup{
	linkcolor=structurecolor,
	linktoc=page,  
}
\minitoc \newpage
\endgroup

\index{Energy-based models}
\lettrine{\color{caligraphcolor}W}
We introduce the score-based interpretation of DDPMs in Section~\ref{section:score_ddpm}. This relationship can also be characterized from a reverse process perspective \citep{song2019generative, bishop2023deep}. To elaborate on why optimizing a score function yields valid generative modeling, we revisit \textit{energy-based models} for foundational context \cite{lecun2006tutorial}. Arbitrarily flexible probability distributions can be formulated in the standard energy-based form:
\begin{equation}\label{equation:score_general_func}
p(\bx) = \frac{1}{C_\bphi}\exp(-f_\bphi(\bx)),
\end{equation}
where $f_\bphi(\bx)$ denotes a flexible, parameterizable energy function, typically instantiated as a neural network, and $C_\bphi$ is the normalizing constant that enforces the valid probability density condition $\int p(\bx)\diff \bx = 1$. Maximum likelihood estimation is a conventional approach to learn such distributions. However, this method requires explicit computation of the intractable normalizing constant $C_\bphi = \int \exp(-f_\bphi(\bx))\diff \bx$, which is infeasible for highly complex energy functions $f_\bphi(\bx)$.

A canonical strategy to bypass explicit normalization constant computation is to train a neural network $\bs^\btheta(\bx)$ to approximate the score function $\nabla \ln p(\bx) \equiv \nabla_{\bx} \ln p(\bx)$ of the target distribution $p(\bx)$. This approach is motivated by differentiating the log-transformed form of Equation \eqref{equation:score_general_func}, which yields the following exact equivalence and neural network approximation:
\begin{align}
\nabla_\bx \ln p(\bx)
&= \nabla_\bx \ln\left(\frac{1}{C_\bphi}\exp(-f_\bphi(\bx))\right)
= -\nabla_\bx f_\bphi(\bx) 
\simeq  \bs^\btheta(\bx).
\end{align}
Notably, the resulting score approximation is fully normalization-free and can be flexibly parameterized by neural networks.

Score matching frameworks are formalized via a loss function that trains the model score $\bs^\btheta(\bx)$ to align with the true data score $\nabla_{\bx} \ln p(\bx)$ of the underlying data-generating distribution $p(\bx)\triangleq p_{\text{data}}(\bx)$. The standard training objective is the expected squared error between the predicted and ground-truth score functions, known as the \textit{Fisher divergence (FD)} between the model and data distributions:
\begin{mybox}
\begin{equation}\label{equation:ddpm_score_loss}
\mathcalJ(\btheta)=
\Exp_{p(\bx)}\left[\normtwo{\bs^\btheta(\bx) - \nabla_{\bx} \ln p(\bx)}^2\right]
= \int p(\bx) \normtwo{\bs^\btheta(\bx) - \nabla_{\bx} \ln p(\bx)}^2 \diff \bx.
\end{equation}
\end{mybox}
Fisher divergence directly measures discrepancies in the score vectors that constitute Fisher information. Minimizing FD aligns the local Fisher curvature of the learned model distribution with the intrinsic information geometry of the true data distribution. Unlike KL divergence, which operates on raw probability density values, FD relies exclusively on log-density gradients (scores). This key property eliminates the need to compute intractable normalization constants $C_\bphi$, making FD particularly suitable for training diffusion generative models.

\begin{figure}[h]
\centering
\includegraphics[width=0.99\textwidth]{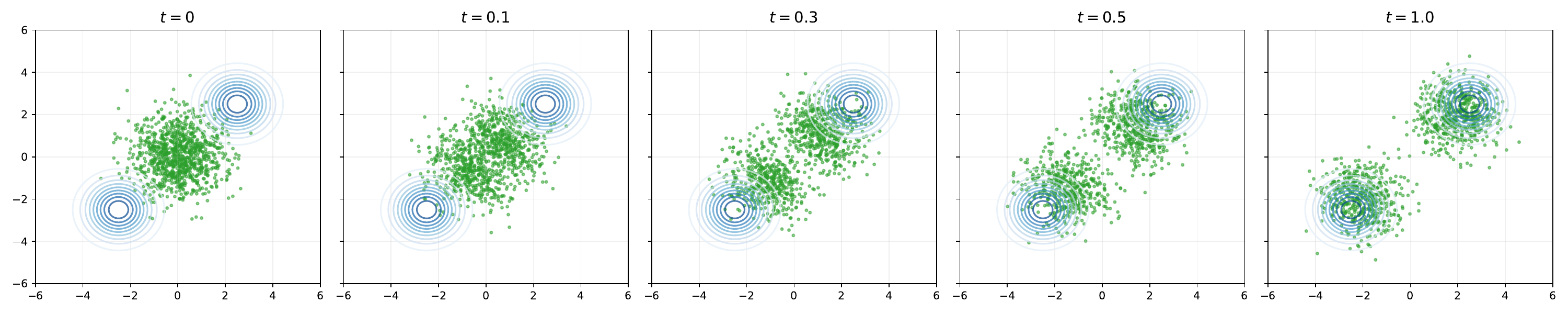}
\caption{Particles evolving under the Langevin dynamics given by \eqref{equation:lagevin_dyn_def_ddpm}, with $p(\bx)$ taken to
be a Gaussian mixture with two modes.}
\label{fig:langevin_move}
\end{figure}

\begin{figure}[h]
\centering
\includegraphics[width=0.95\textwidth]{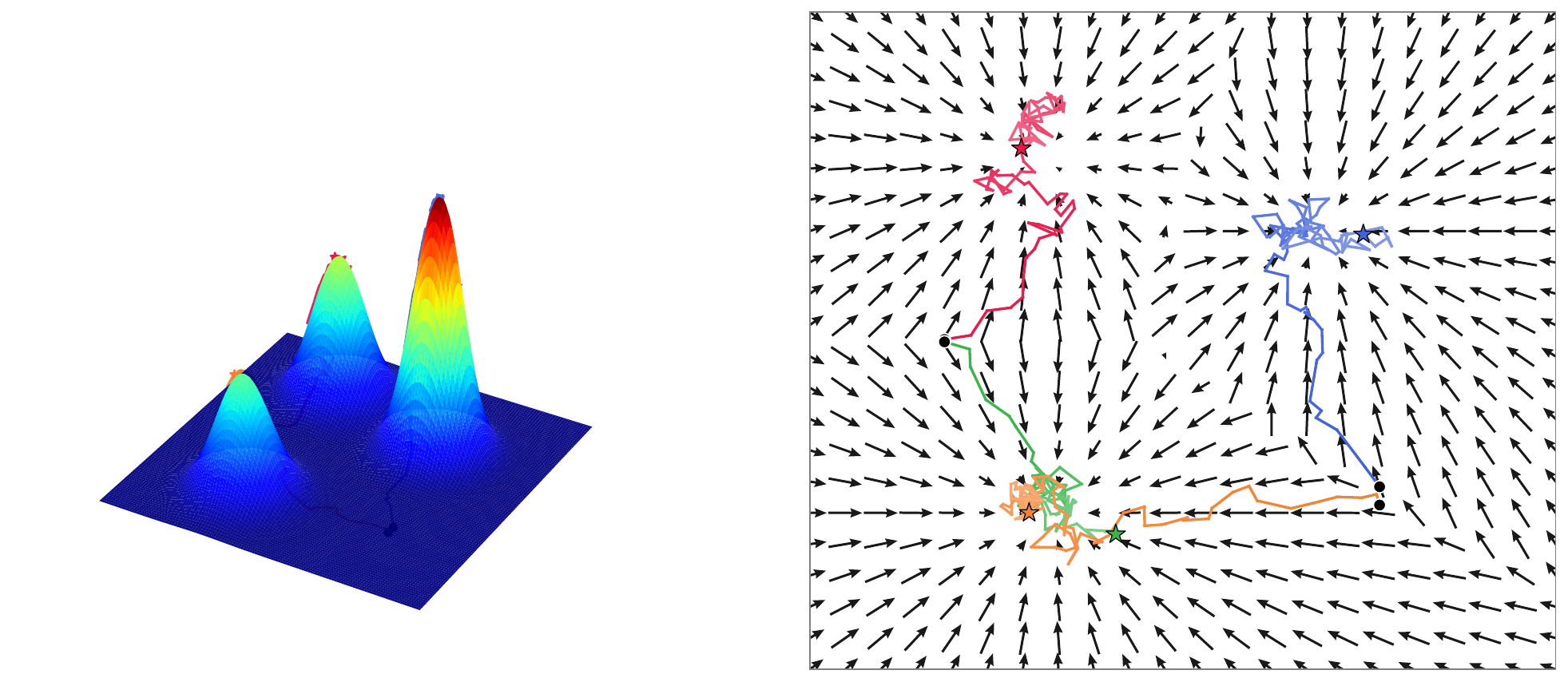}
\caption{
Presents four distinct sampling trajectories generated by Langevin dynamics for a Gaussian mixture model, all initialized from a starting point between two modes. 
The left subfigure visualizes these trajectories over a three-dimensional contour landscape of the target distribution, whereas the right subfigure overlays the trajectories against the ground-truth score function field. Owing to the stochastic noise term embedded within the Langevin sampling update rule, samples originating from a single fixed initialization can converge to different distribution modes across independent trials. In the absence of this noise component, sampling initiated at a fixed point would follow a fully deterministic trajectory and invariably converge to the identical mode in every run.}
\label{fig:langevin_mog}
\end{figure}

The score function admits an intuitive geometric interpretation: for each data point $\bx$, the gradient of the log-likelihood with respect to $\bx$ indicates the direction in data space along which the likelihood of the sample increases~\footnote{See, for example, \citet{lu2025practical}.}. Collectively, the score function defines a continuous vector field over the entire data space, pointing toward the high-density modes of the data distribution. This geometric structure is visualized in the right panel of Figure~\ref{fig:langevin_mog}.
By accurately learning the score field of the true data distribution, we can generate novel samples via iterative optimization: starting from arbitrary initial points in the ambient data space, we iteratively follow the score vector field to converge toward data modes. This generative sampling paradigm is termed \textit{Langevin dynamics}, defined mathematically as:
\begin{equation}\label{equation:lagevin_dyn_def_ddpm}
\bx_{t+1} \leftarrow \bx_t + \frac{\zeta}{2}\nabla_{\bx}\ln p(\bx_t) + \sqrt{\zeta}\,\bepsilon,\quad t=0,1,\ldots,T,
\end{equation}
where the initial sample $\bx_0$ is randomly sampled from a pre-specified prior distribution (e.g., a uniform distribution), and $\bepsilon\sim \normal(\bzero, \bI_D)$ denotes Gaussian noise. The injected stochastic noise prevents generated samples from collapsing deterministically to distribution modes, thereby preserving sample diversity, as discussed in Section~\ref{section:gen_ddpm}. The evolution of sample particles under Langevin dynamics is illustrated in Figure~\ref{fig:langevin_move}.
Furthermore, since the learned score function is deterministic, the additive noise term introduces essential stochasticity to the generative process and avoids rigid deterministic sampling trajectories. This stochasticity is especially beneficial when sampling initial points located between multiple data modes. A comprehensive visual illustration of Langevin dynamics sampling and the regularization benefits of noise injection is presented in Figure~\ref{fig:langevin_mog}.

\index{Langevin dynamics}
\section{Score Matching}\label{section:score_mat_main}
However, the loss function defined in \eqref{equation:ddpm_score_loss} suffers from a critical practical limitation: neither the original loss nor its direct transformations can be optimized efficiently, as the formulation is computationally intractable. To address this issue, we introduce two practical variants of the score-matching loss: \textit{denoising score matching loss} and \textit{sliced score matching loss}.

\index{Dirac delta distribution}
\index{Dirac delta function}
\index{Parzen estimator}
\index{Fisher divergence}
\index{Kernel density estimator}
\index{Denoising score matching}
\index{Gaussian Parzen kernel}
\index{Snergy score matching}
\subsection{Denoising Score Matching}

A fundamental limitation of the loss function in \eqref{equation:ddpm_score_loss} is that it cannot be minimized directly, since the ground-truth data score $\nabla_{\bx} \ln p_{\text{data}}(\bx)$ of the true data distribution is analytically unknown. In practice, we only have access to a finite empirical dataset $\mathcalX = \{\bx_1, \bx_2, \ldots, \bx_N\} \sim p_{\text{data}}(\bx)$, from which we construct the empirical data distribution:
\begin{equation}\label{equation:score_empirical_density}
p_\mathcalX(\bx) = \frac{1}{N} \sum_{n=1}^{N} \delta_{\bx_n}(\bx),
\end{equation}
where $\delta_{\bx_n}=\delta(\bx-\bx_n)$ is the \textit{Dirac delta distribution} (see \eqref{equation:dirac_func}) centered at $\bx_n$ (i.e., sampling from
$\delta_{\bx_n}$ always returns $\bx_n$).

The empirical Dirac-based distribution in \eqref{equation:score_empirical_density} is non-differentiable with respect to $\bx$, making its corresponding score function unavailable for direct computation. To overcome this obstacle, we smooth the discrete data points via a noise injection strategy, yielding a continuous, differentiable density estimate. This kernel-based smoothing technique corresponds to a \textit{Parzen estimator}, or \textit{kernel density estimator}, formulated as:
\begin{equation}\label{equation:parzen_estimator}
q_\sigma(\bz) = \int q_\sigma(\bz \mid \bx) p_{\text{data}}(\bx) \diff\bx,
\end{equation}
where $q_\sigma(\bz \mid \bx)$ denotes  the \textit{noise kernel}. 
The most commonly adopted choice is the \textit{Gaussian Parzen kernel}, defined as:
\begin{equation}\label{equation:gaussian_parzen_kernel}
q_\sigma(\bz \mid \bx) = \normal(\bz \mid \bx, \sigma^2 \bI).
\end{equation}

Instead of minimizing the original intractable score-matching loss in \eqref{equation:ddpm_score_loss}, we optimize the loss formulated over the smoothed Parzen density, termed the \textit{energy score matching (ESM) loss}:
\begin{subequations}
\begin{equation}\label{equation:ddpm_modified_score_loss}
\small
\begin{aligned}
\mathcalJ_{\text{ESM}}(\btheta) = 
\Exp_{q_\sigma(\bz)}\left[\normtwo{\bs^\btheta(\bz) - \nabla_{\bz} \ln q_\sigma(\bz)}^2\right]
= \int \normtwo{\bs^\btheta(\bz) - \nabla_{\bz} \ln q_\sigma(\bz)}^2 q_\sigma(\bz) \diff\bz.
\end{aligned}
\end{equation}
A core theoretical result established in \citet{vincent2011connection} proves that substituting the Parzen smoothing framework \eqref{equation:parzen_estimator} into the ESM loss yields an equivalent, computationally feasible reformulation known as the \textit{denoising score matching (DSM) loss} (see Problem~\ref{prob:ddpmscore_vincent_form}):
\begin{equation}\label{equation:ddpmscore_vincent_form}
\mathcalJ_{\text{DSM}}(\btheta) =  \int \normtwo{\bs^\btheta(\bz) - \nabla_{\bz} \ln q_\sigma(\bz \mid \bx)}^2 q_\sigma(\bz \mid \bx) p_{\text{data}}(\bx) \diff\bz \diff\bx + \text{const}.
\end{equation}
Replacing the true data distribution $p_{\text{data}}(\bx)$ with the empirical distribution $p_{\mathcal{X}}(\bx)$ from \eqref{equation:score_empirical_density} produces the final empirical DSM loss:
\begin{mybox}
\begin{equation}\label{equation:ddpmscore_empirical_modified_loss}
\mathcalJ'(\btheta) = \frac{1}{N} \sum_{n=1}^{N} \int \normtwo{\bs^\btheta(\bz) - \nabla_{\bz} \ln q_\sigma(\bz \mid \bx_n)}^2 q_\sigma(\bz \mid \bx_n) \diff\bz + \text{const}.
\end{equation}
\end{mybox}
\end{subequations}
For the Gaussian Parzen kernel defined in \eqref{equation:gaussian_parzen_kernel}, the corresponding conditional score function admits a closed-form expression:
\begin{equation}\label{equation:gaussian_parzen_score}
\nabla_{\bz} \ln q_\sigma(\bz \mid \bx) = -\frac{1}{\sigma^2} \bepsilon,
\end{equation}
where $\bepsilon = \bz - \bx$. 
In practice, this formulation simplifies training significantly: we only sample standard Gaussian noise $\bepsilon^*$ and construct noisy data samples via the standard reparameterization trick $\bz =\bx + \sigma\bepsilon^*$, which obeys the noisy data distribution $q_\sigma(\bz\mid \bx) \sim \normal(\bz\mid \bx, \sigma^2 \bI)$.

\begin{remark}[Equivalence to DDPMs]
By adopting the diffusion transition kernel from \eqref{equation:ddpm_diffusion_kernel_all}, we derive the diffusion-specific score function:
\begin{equation}\label{equation:diffusion_parzen_score}
\nabla_{\bz_t} \ln q(\bz_t \mid \bx_0) 
= \textcolor{black}{\frac{\sqrt{\alpha_t}\bx_0 - \bz_t}{{1 - \alpha_t}}}
= -\frac{1}{\sqrt{1 - \alpha_t}} \bepsilon_t, 
\quad t=1,2,\ldots,T.
\end{equation}
This result demonstrates that the empirical score-matching loss in \eqref{equation:ddpmscore_empirical_modified_loss} essentially minimizes the prediction error between the network output and the Gaussian noise injected during the diffusion forward process. As a consequence, this loss shares an identical optimal solution with the standard noise-prediction loss used in DDPMs, as shown in \eqref{equation:ddpm_kl_noise_prediction} and the score-based DDPM formulation \eqref{equation:loss_score_ddpm_final}. The score network $\bs^\btheta(\bz)$ is functionally equivalent to the DDPM noise-prediction network $\bepsilon^\btheta(\bz)$, differing only by a constant scaling factor $-1/\sqrt{1 - \alpha_t}$ \citep{song2019generative}. This optimization objective defines denoising score matching, which establishes a rigorous theoretical bridge between score-based generative models and denoising diffusion probabilistic models.
\end{remark}

\index{Sliced score matching}
\index{Hutchinson's trace estimator}
\subsection{Sliced Score Matching}
As discussed previously, direct optimization of Fisher divergence is infeasible, as it requires evaluating the intractable score function of the unknown true data distribution. Rather than performing explicit denoising on observed data, we circumvent the need to explicitly compute the underlying data distribution $p_{\text{data}}(\bx)$ by directly training a score network $\bs^\btheta(\bx)$ to approximate the data score $\nabla_{\bx} \ln p_{\text{data}}(\bx)$. This paradigm is grounded in the following theoretical result:
\begin{theoremHigh}[Score matching loss]\label{theorem:sm_loss}
The intractable Fisher divergence loss in \eqref{equation:ddpm_score_loss} can be equivalently rewritten as a fully computable objective:
\begin{equation}\label{equation:sm_loss}
\mathcalJ''(\btheta)
=\Exp_{p_{\text{data}}(\bx)}\left[ \trace\left( \nabla_{\bx} \bs^\btheta(\bx) \right) + \frac{1}{2} \normtwo{\bs^\btheta(\bx)}^2 \right],
\end{equation}
where $\nabla_{\bx} \bs^\btheta(\bx)$ denotes the Jacobian matrix of the score network with respect to input $\bx$, and $\trace(\cdot)$ denotes the matrix trace, corresponding to the sum of diagonal entries of the Jacobian.
\end{theoremHigh}
\begin{proof}[of Theorem~\ref{theorem:sm_loss}]
For simplicity, we abbreviate the data distribution as $p(\bx)\triangleq p_{\text{data}}(\bx)$ and expand the original score matching loss in \eqref{equation:ddpm_score_loss} as follows:
$$
\small
\begin{aligned}
\Exp_{p} \left[ \normtwo{\bs^{\btheta}(\bx) - \nabla_{\bx} \ln p(\bx) }^2 \right]
= \int p(\bx) \left[ \normtwo{\bs^{\btheta}(\bx)}^2 - 2 \left( \nabla_{\bx} \ln p(\bx) \right)^\top \bs^{\btheta}(\bx) + \normtwo{ \nabla_{\bx} \ln p(\bx)}^2 \right] \diff \bx.
\end{aligned}
$$
The final term inside the integral is constant with respect to the model parameter $\btheta$ and can thus be discarded during optimization. The first term is directly computable from network outputs, leaving only the cross term requiring further derivation:
$$
\begin{aligned}
&\int p(\bx) \left[ -2 \left( \nabla_{\bx} \ln p(\bx) \right)^\top \bs^{\btheta}(\bx) \right] \diff \bx
= -2 \int p(\bx) \left[ \sum_{i=1}^D \frac{\partial \ln(p(\bx))}{\partial x_i} \bs^\btheta_{i}(\bx) \right] \diff \bx \\
&= -2 \sum_{i=1}^D \int p(\bx) \left[ \left( \frac{1}{p(\bx)} \frac{\partial p(\bx)}{\partial x_i} \right) \bs^\btheta_{i}(\bx) \right] \diff \bx
= -2 \sum_{i=1}^D \int \left[ \frac{\partial p(\bx)}{\partial x_i} \bs^\btheta_{i}(\bx) \right] \diff \bx \\
&\stackrel{\dag}{=} -2 \sum_{i=1}^D \int \left( \frac{\partial(p(\bx) \bs^\btheta_{i}(\bx))}{\partial x_i} - p(\bx) \frac{\partial \bs^\btheta_{i}(\bx)}{\partial x_i} \right) \diff \bx
\stackrel{\ddag}{=} 2 \int p(\bx) \left[ \sum_{i=1}^D \frac{\partial \bs^\btheta_{i}(\bx)}{\partial x_i} \right] \diff \bx \\
&= 2 \int p(\bx) \left( \trace\left( \frac{\partial \bs^{\btheta}(\bx)}{\partial \bx} \right) \right) \diff \bx,
\end{aligned}
$$
where $\bs^\btheta_{i}(\bx)$ denotes the $i$-th element of $\bs^\btheta(\bx)$, the equality ($\dag$) follows from integration by parts, 
and the equality ($\ddag$) follows since 
$
\int \left( \frac{\partial(p(\bx) \bs^\btheta_{i}(\bx))}{\partial x_i} \right) \diff \bx 
= \left. p(\bx) \bs^\btheta_{i}(\bx) \right|_{-\infty}^{+\infty} = 0
$ (we assume $p(\infty) \to 0$, so this term is omitted).
Substituting this simplified cross term back into the expanded loss yields the final result, which completes the proof.
\end{proof}

Despite eliminating the intractable ground-truth score term, the loss function $\mathcalJ''(\btheta)$ in \eqref{equation:sm_loss} still suffers from severe computational overhead for high-dimensional data (e.g., images with hundreds or thousands of pixel and channel dimensions). Exact computation of the Jacobian trace $\trace\left( \nabla_{\bx} \bs^\btheta(\bx) \right)$ requires dimension-wise backpropagation, which scales prohibitively with data dimensionality. To resolve this bottleneck, \citet{song2020sliced} proposed \textit{sliced score matching} as an efficient approximation.

As outlined above, exact Jacobian trace computation is computationally prohibitive for high-dimensional inputs. To address this issue, we adopt a stochastic estimation strategy via \textit{Hutchinson's trace estimator} \citep{grathwohl2018ffjord}: we sample a random vector $\bzeta$ from a zero-mean, unit-variance distribution (e.g., standard Gaussian distribution) to approximate the exact trace, yielding the following identity (see Problem~\ref{prob:score_hutchinson}):
\begin{equation}\label{equation:score_hutchinson}
\trace\left( \nabla_{\bx} \bs^\btheta(\bx) \right)
=
\Exp_{\bzeta} \trace \left[\bzeta^\top \left( \nabla_{\bx} \bs^\btheta(\bx) \right) \bzeta\right].
\end{equation}
Based on matrix trace properties, we project the Jacobian matrix onto the random vector $\bzeta$ to form a Jacobian-vector product, which simplifies computation significantly (see Section~\ref{section:like_flow_hutchin}):
$$
\bzeta^\top \frac{\partial \bs^\btheta(\bx)}{\partial \bx} \bzeta
= \bzeta^\top \frac{\partial \left[ \bzeta^\top \bs^\btheta(\bx) \right]}{\partial \bx}.
$$
Substituting this estimator into the original score matching loss \eqref{equation:sm_loss} yields the final optimized training objective:
\begin{mybox}
\begin{equation}\label{equation:sm_hutchinson}
\mathcalJ''(\btheta)
= \Exp_{p_{\bzeta}(\bzeta)p_{\text{data}}(\bx)}
\left[
\bzeta^\top \frac{\partial \left[ \bzeta^\top \bs^\btheta(\bx) \right]}{\partial \bx}
+ \frac{1}{2} \normtwo{\bs^\btheta(\bx)}^2
\right].
\end{equation}
\end{mybox}
Unlike standard network training pipelines, this optimized loss requires one additional backpropagation pass. The first backpropagation computes the core training objective, while a second subsequent backpropagation step calculates the model gradients for parameter update. Further technical details regarding this two-pass training mechanism are provided in Section~\ref{section:like_flow_hutchin}.

\subsection{Annealed Langevin Dynamics}

We have introduced the procedure for learning the score function from training datasets and generating new samples from the learned distribution via Langevin sampling. Nevertheless, three key limitations exist for this standard framework \citep{song2019generative}.

\paragrapharrow{Manifold hypothesis.}
First, we elaborate on the \textit{manifold hypothesis}, which states that real-world data typically resides on a \textit{low-dimensional manifold} embedded within a high-dimensional ambient space. This indicates that raw or preprocessed data contains substantial redundant information across most dimensions, such that only a small subset of dimensions is required for complete data representation. Consequently, data occupies only a constrained subspace (a low-dimensional manifold) rather than uniformly filling the entire high-dimensional encoding space.
This inherent property leads to non-convergent loss optimization in the training of score-based generative models. Specifically, the score function $\nabla_{\bx} \ln p(\bx)$ is mathematically defined over the entire encoding space. However, for manifold-restricted data distributions, the score function is ill-defined in all regions outside the data-supporting manifold.

\paragrapharrow{Inaccurate score estimation.}
The score matching objectives defined in \eqref{equation:ddpm_score_loss} and \eqref{equation:sm_loss} yield consistent score estimators only if the data distribution has full support over the entire ambient space. Consistency breaks down when data lies on a low-dimensional manifold.
Score estimation accuracy further degrades in low-data-density regions, due to density-based loss weighting and insufficient training samples covering these distribution regions. Such inaccurate score estimates corrupt the sampling trajectories of Langevin dynamics \eqref{equation:lagevin_dyn_def_ddpm}. Since Langevin sampling initializes from random high-dimensional noise and evolves strictly following the learned score function, noisy or biased score estimates produce suboptimal generated samples and require substantially more iteration steps to achieve convergence.

\paragrapharrow{Large deviation between generated distribution and reality.}
Inaccurate score estimation inevitably degrades generation quality, resulting in generated samples that fail to align with the true underlying data distribution. Even with perfectly accurate score function estimation, standard Langevin sampling may still produce biased samples for data distributions composed of disjoint mixture components.
Consider a mixture data distribution formulated as $p_{\text{data}}(\bx) = \pi p_1(\bx) + (1-\pi) p_2(\bx)$, where $p_1(\bx)$ and $p_2(\bx)$ denote normalized distributions with disjoint supports, and $\pi \in (0,1)$ denotes the mixing coefficient. Within the support of $p_1(\bx)$, the score satisfies $\nabla_{\bx} \ln p_{\text{data}}(\bx) = \nabla_{\bx} \left( \ln \pi + \ln p_1(\bx) \right) = \nabla_{\bx} \ln p_1(\bx)$. Similarly, within the support of $p_2(\bx)$, $\nabla_{\bx} \ln p_{\text{data}}(\bx) = \nabla_{\bx} \left( \ln(1-\pi) + \ln p_2(\bx) \right) = \nabla_{\bx} \ln p_2(\bx)$.

Notably, the score function of the mixture distribution is independent of the mixing coefficient $\pi$. As Langevin dynamics generate samples solely guided by $\nabla_{\bx} \ln p_{\text{data}}(\bx)$, the resulting samples fail to reflect the true mixing weights. This conclusion also holds for practical scenarios with approximately disjoint modes, where distinct modes are separated by low-density transitional regions. Although Langevin sampling is theoretically valid in such cases, it requires extremely small step sizes and vast iteration counts to achieve mode mixing.

This phenomenon is visualized in Figure~\ref{fig:langevin_mog}, where the ground-truth score function is invariant to the varying weights of the three Gaussian mixture components. From the illustrated initialization point, standard Langevin dynamics yield nearly equal sampling probability for all modes, even when the rightmost mode possesses a significantly higher weight in the true Gaussian mixture model.

\index{Annealed Langevin dynamics}
\paragrapharrow{Annealed Langevin dynamics.}
All aforementioned issues can be mitigated by introducing Gaussian noise with sufficiently large variance $\sigma^2$ via the kernel function in \eqref{equation:gaussian_parzen_kernel}, which effectively smooths the original data distribution. First, Gaussian noise spans full support over the entire ambient space, lifting manifold-constrained data to the complete high-dimensional space and eliminating undefined score regions. Second, large-magnitude Gaussian noise expands the effective support of each data mode, enriching training signals in low-density regions. Third, hierarchical noise injection with progressively increasing variance yields intermediate smoothed distributions that faithfully preserve the ground-truth mixture mixing coefficients.

Nevertheless, excessively large noise variance introduces severe distortion to the original data distribution, inducing new biases in score function modeling. To resolve this trade-off, we adopt a decreasing sequence of noise variances $\sigma_1^2 > \sigma_2^2 > \ldots > \sigma_L^2$. The smallest variance $\sigma_L^2$ is set sufficiently small to retain accurate original distribution characteristics, while the largest variance $\sigma_1^2$ is large enough to eliminate the aforementioned manifold and estimation artifacts.

The score network is accordingly modified to take the noise variance as an auxiliary input, denoted $\bs^\btheta(\bx, \sigma)$. The network is trained via a weighted combination of standard score matching losses \eqref{equation:ddpm_score_loss}, where each loss term corresponds to a specific noise level and its perturbed data distribution. For a given data sample $\bx_n$, the training objective is formulated as:
\begin{mybox}
\begin{equation}\label{equation:annealed_score_loss}
\frac{1}{2} \sum_{i=1}^{L} \lambda(i) \int \normtwo{\bs^\btheta(\bz, \sigma_i) - \nabla_{\bz} \ln q_{\sigma_i}(\bz \mid \bx_n)}^2 q_{\sigma_i}(\bz \mid \bx_n) \diff\bz,
\end{equation}
\end{mybox}
where $\lambda(i)$ denotes positive weight coefficients conditioned on the noise level $i$. This training paradigm is highly analogous to the optimization strategy for hierarchical denoising networks, and the objective function closely matches the final loss formulation for variational diffusion models in \eqref{equation:loss_score_ddpm_final}.

After training, sample generation proceeds by sequentially executing Langevin sampling steps across all noise levels $i = 1, 2, \ldots, L$. This paradigm is defined as \textit{annealed Langevin dynamics}, which shares core design principles with the sampling algorithm for denoising diffusion models (Algorithm~\ref{alg:diffusion_sampling}). The full pseudocode of annealed Langevin dynamics is provided in Algorithm~\ref{alg:anneal_langevin}.
Sampling initialization is sampled from a fixed prior distribution (e.g., uniform distribution), and the sampling trajectory of each noise level inherits the final samples from the previous level. By progressively reducing noise levels and sampling step sizes, the generated samples gradually refine and converge to valid data modes. This sampling mechanism is consistent with the iterative refinement process in the Markovian HVAE interpretation of variational diffusion models, where randomly initialized data vectors are optimized iteratively under decreasing noise perturbations.

\begin{algorithm}
\caption{Annealed Langevin Dynamics}
\label{alg:anneal_langevin}
\begin{algorithmic}[1]
\Require $\{\sigma_i\}_{i=1}^L$, $\epsilon$, $T$, score network output $\bs^\btheta_t(\cdot)$
\State \textbf{initialize:} $\bx_0$;
\For{$i \leftarrow 1$ to $L$}
\State $\alpha_i \leftarrow \epsilon \cdot \sigma_i^2 / \sigma_L^2$; \Comment{$\alpha_i$ is the step size.}
\For{$t \leftarrow 1$ to $T$}
\State Draw noise $\bepsilon \sim \normal(\bzero, \bI)$;
\State $\bx_t \leftarrow \bx_{t-1} + \frac{\alpha_i}{2} \bs^\btheta_{t-1}(\bx_{t-1}, \sigma_i) + \sqrt{\alpha_i} \bepsilon$;
\EndFor
\State $\bx_0 \leftarrow \bx_T$;
\EndFor
\State \Return $\bx_T$;
\end{algorithmic}
\end{algorithm}

\section{Stochastic Differential Equations and Diffusion Models}\label{section:score_mat}
In the preceding Section~\ref{section:score_mat_main}, we establish the foundational framework of score matching---a powerful technique for estimating the gradient of the log-density (score function) of a data distribution---and explore key variants like denoising score matching, sliced score matching, and their connection to annealed Langevin dynamics for sampling. While these methods provide a direct route to modeling and generating data, they often frame the problem in terms of static density estimation and iterative sampling steps, without fully unpacking the continuous-time stochastic processes that underpin modern generative modeling.

This section shifts this perspective to a more fundamental and unifying lens: stochastic differential equations (SDEs) and the continuous probability paths they describe. This SDE perspective reinterprets diffusion models not as a discrete sequence of noise injection and denoising steps, but as a continuous-time stochastic process that gradually transforms a complex data distribution into a simple Gaussian distribution (and vice versa, for sampling). By modeling the evolution of probability distributions along continuous-time paths, we gain deeper mathematical insights into the relationship between score matching, diffusion processes, and generative sampling.

\index{Stochastic differential equations (SDEs)}
\subsection{Diffusion Models: Revisited}\label{section:diff_model}

We begin by revisiting diffusion models  through the lens of \textit{stochastic differential equations (SDEs)}, showing how the discrete forward/backward processes of standard diffusion models emerge as discretizations of continuous-time stochastic processes. 
SDEs generalize the deterministic trajectory solutions of ODEs~\eqref{equation:ode_def} by introducing stochastic trajectory dynamics \citep{mao2007stochastic, lipman2024flow, holderrieth2025introduction}. 
Such stochastic trajectories are formally defined as a stochastic process $\{\rvx_t\}_{0 \leq t \leq 1}$, which satisfies the following properties: 
\begin{align*}
	\rvx_t &\text{ is a random variable for every } 0 \leq t \leq 1;\\
	\rvx: [0, 1] \to \real^D, \quad t \mapsto \rvx_t &\text{ is a random trajectory for every draw of } \rvx.
\end{align*}
Notably, independent simulations of the same stochastic process yield distinct trajectory realizations, as the governing dynamics are inherently randomized.

\index{Brownian motion}
\index{Wiener process}
\paragrapharrow{Brownian Motion.} 
The construction of SDEs relies fundamentally on \textit{Brownian motion}, also referred to as the \textit{Wiener process}. 
Serving as the canonical mathematical model for continuous-time random motion, Brownian motion can be intuitively interpreted as a continuous analog of discrete random walks. 
A physical intuition for Brownian motion is the erratic movement of tiny particles---such as dust motes in air or pollen grains suspended in water. These particles undergo continuous, random collisions with surrounding microscopic molecules from all directions, resulting in unstructured, unpredictable jittering and drifting motion, which is precisely characterized by Brownian motion.
Formally, a Brownian motion $\rvb = \{\rvb_t\}_{0 \leq t \leq 1}$ is a stochastic process with continuous sample trajectories $t \mapsto \rvb_t$ that adheres to the following three axiomatic properties:
\begin{enumerate}[(i)]
	\item \textit{Zero initialization.} The process starts at the origin, i.e., $\rvb_0 = \bzero$.
	\item \textit{Normal increments.} For all $0 \leq s < t$, the temporal increment obeys $\rvb_t - \rvb_s \sim \normal(\bzero, (t-s)\bI_D)$, meaning incremental displacements follow a Gaussian distribution with variance scaling linearly with time ($\bI_D$ denotes the $D$-dimensional identity matrix). 
	\item \textit{Independent increments.} For any arbitrary time sequence $0 \leq t_0 < t_1 < \ldots < t_n = 1$, the incremental terms $\rvb_{t_1} - \rvb_{t_0}, \ldots, \rvb_{t_n} - \rvb_{t_{n-1}}$ form mutually independent random variables.
\end{enumerate}
Brownian motion can be numerically approximated via discrete-time simulation with a small step size $\tau > 0$. Starting from the zero initial condition $\rvb_0 = \bzero$, the iterative update rule is formulated as (see Problem~\ref{prob:brownian_motion}):
\begin{equation}\label{equation:brownian_motion}
	\rvb_{t+\tau} = \rvb_t + \sqrt{\tau} \bepsilon_t, \quad \bepsilon_t \sim \normal(\bzero, \bI_D) \quad (t = 0, \tau, 2\tau, \ldots, 1 - \tau).
\end{equation}

Brownian motion occupies a foundational role in stochastic process theory, analogous to the central position of Gaussian distributions in probability theory. Beyond machine learning, it finds extensive applications across diverse disciplines, including quantitative finance, statistical physics, and epidemiology. For instance, Brownian motion is widely adopted in financial modeling to characterize the price dynamics of sophisticated financial derivatives. 
Mathematically, Brownian motion exhibits striking and counterintuitive properties: its sample paths are globally continuous, enabling unbroken graphical rendering, yet possess infinite arc length over any finite time interval, rendering perpetual trajectory tracing impossible.

\paragrapharrow{From ODEs to SDEs.} 
SDEs extend the deterministic evolutionary dynamics of ODEs~\eqref{equation:ode_def} by incorporating stochastic perturbations driven by Brownian motion. Due to the inherent randomness of stochastic systems, the standard time derivative definition applicable to deterministic ODEs in \eqref{eq:ode1} is no longer valid. This necessitates an alternative ODE formulation that avoids explicit derivative operations. To this end, we reformulate the continuous trajectories $\{\rvx_t\}_{0 \leq t \leq 1}$ of standard ODEs via discrete infinitesimal updates, as illustrated in the following derivation:
\begin{align*}
\frac{\diff}{\diff t} \rvx_t &= \vf_t(\rvx_t) && (\text{from ODE})  \\
\stackrel{\dag}{\implies} \quad \frac{1}{\tau} (\rvx_{t+\tau} - \rvx_t) &= \vf_t(\rvx_t) + \rve_t(\tau) && (\text{infinitesimal updates})\\
\iff \quad \rvx_{t+\tau} &= \rvx_t + \tau \vf_t(\rvx_t) + \tau \rve_t(\tau).
\end{align*}
Here, $\rve_t(\tau)$ denotes a negligible residual term for sufficiently small step sizes $\tau$, satisfying $\lim_{\tau \to 0} \rve_t(\tau) = \bzero$, and the implication $(\dag)$ holds by the fundamental definition of temporal derivatives. This derivation restates the core property of deterministic ODE trajectories: at each infinitesimal time step, the system evolves a small increment along the directional vector field $\vf_t(\rvx_t)$. Building on this discrete update framework, we introduce stochasticity to construct the general SDE formulation. Specifically, SDE trajectories inherit the deterministic drift of ODE systems while incorporating random fluctuations sourced from Brownian motion, yielding the discrete update rule:
\begin{equation}\label{equation:sde_def}
\rvx_{t+\tau} = \rvx_t + \underbrace{\tau \vf_t(\rvx_t)}_{\text{Deterministic}} + \underbrace{\sigma_t (\rvb_{t+\tau} - \rvb_t)}_{\text{Stochastic}} + \underbrace{\tau \rve_t(\tau)}_{\text{Error term}}.
\end{equation}
In this expression, the nonnegative parameter $\sigma_t \geq 0$ is defined as the \textit{diffusion coefficient} that governs the magnitude of stochastic perturbations. The term $\rve_t(\tau)$ represents a stochastic residual error, whose root mean squared norm vanishes asymptotically with step size reduction, i.e., $\Exp[\normtwo{\rve_t(\tau)}^2]^{1/2} \to \bzero$ as $\tau \to 0$. The above discrete formulation fully characterizes SDE dynamics, which is conventionally simplified into a compact, informal symbolic form for theoretical analysis \citep{holderrieth2025introduction} (cf. \eqref{equation:ode_def}):
\begin{subequations}\label{equation:sde_def2}
\begin{align}
\diff \rvx_t &= \vf_t(\rvx_t)\diff t + \sigma_t \diff \rvb_t; && (\text{SDE}) \\
\rvx_0 &= \bx_0. && (\text{initial conditions})
\end{align}
\end{subequations}

Unlike deterministic ODEs, SDEs do not admit well-defined deterministic flow maps. This is because the system state $\rvx_t$ cannot be uniquely determined by the initial state $\rvx_0 \sim p_{\text{base}}$, as the temporal evolution is governed by probabilistic stochastic dynamics. Nevertheless, the well-posedness of SDE solutions can be guaranteed under standard regularity conditions, analogous to the existence and uniqueness theory for ODEs:
\begin{theoremHigh}[SDE solution existence and uniqueness \citep{mao2007stochastic}]
If $\vf : \real^D \times [0, 1] \to \real^D$  is continuously differentiable with bounded gradients and the diffusion coefficient $\sigma_t$ is continuous over time, the SDE system defined in \eqref{equation:sde_def2} admits a unique stochastic process solution $\{\rvx_t\}_{0 \leq t \leq 1}$ that satisfies the discrete update formulation in \eqref{equation:sde_def}.
\end{theoremHigh}

Notably, ODEs can be regarded as a special subclass of SDEs corresponding to the zero-diffusion case where $\sigma_t = 0$ for all $t$  \citep{albergo2025stochastic}. 
For all subsequent discussions in this book, we adopt a unified perspective that treats conventional ODE systems as degenerate SDE systems with vanishing stochastic components.

\begin{figure}[h]
\centering
\includegraphics[width=0.99\textwidth]{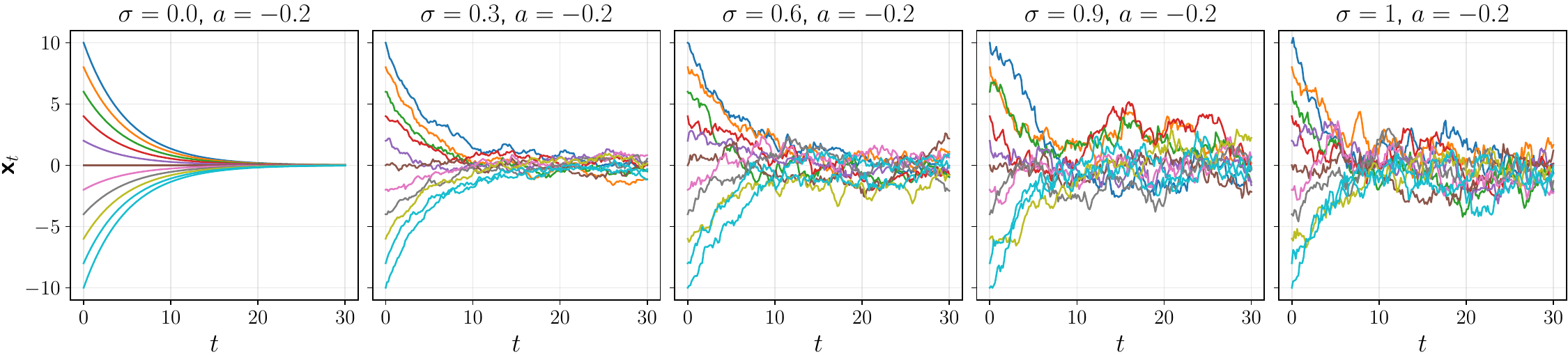}
\caption{Shown are one-dimensional OU processes~\eqref{equation:ouprocess} with $a = 0.2$, where $\sigma$ increases from left to right. The $\sigma = 0$ case yields a deterministic flow: smooth trajectories that decay monotonically to the origin as $t \to \infty$. In contrast, $\sigma > 0$ introduces randomness, with sample paths converging to the stationary Gaussian distribution $\normal(0, {\sigma^2}/{(-2a)})$ in the limit $t \to \infty$.}
\label{figure:ou_motion}
\end{figure}
\begin{example}[Ornstein--Uhlenbeck Process\index{Ornstein--Uhlenbeck process}]\label{example:ou_process}
We consider a representative SDE setup with constant diffusion coefficient $\sigma_t = \sigma \geq 0$ and time-invariant linear vector field $\vf_t(\bz) = a \bz$ with $a < 0$, which yields the canonical SDE formulation:
\begin{equation}\label{equation:ouprocess}
\diff \rvx_t = a\rvx_t \diff t + \sigma \diff \rvb_t.
\end{equation}
In the noise-free scenario with $\sigma = 0$, the system reduces to the deterministic linear vector field analyzed in Example~\ref{example:ode_lnvf}.
The unique stochastic process solving this equation is termed the \textit{Ornstein--Uhlenbeck (OU) process}, whose sample trajectories are visualized in Figure~\ref{figure:ou_motion}. 
The linear vector field $a \rvx_t$ acts as a restoring force, consistently pulling the system state toward the origin $\bzero$ by counteracting positional deviations. Meanwhile, the diffusion term injects persistent Gaussian noise to introduce stochastic fluctuations. In the long-time limit $t \to \infty$, this process converges to a stationary Gaussian distribution $\normal(\bzero, \sigma^2/(-2a)\bI )$. 
\end{example}

\index{Euler--Maruyama method}
\paragrapharrow{Simulating an SDE.} 
The most fundamental numerical scheme for discretizing and simulating SDE dynamics is the \textit{Euler--Maruyama method}, which serves as the stochastic counterpart of the standard Euler method for deterministic ODEs \citep{mao2007stochastic}. 
Given a predefined number of discretization steps $n \in \naturalset$, we set the discrete step size to $\tau = n^{-1} > 0$. The simulation procedure initializes the system state as $\rvx_0 = \bx_0$ and iteratively updates the trajectory via the following rule:
\begin{equation}
\rvx_{t+\tau} = \rvx_t + \tau \vf_t(\rvx_t) + \sqrt{\tau} \sigma_t \bepsilon_t, 
\qquad \bepsilon_t \sim \normal(\bzero, \bI_D).
\end{equation}
Intuitively, each Euler--Maruyama iteration consists of two components: a deterministic update driven by the vector field $\vf_t(\rvx_t)$, and a stochastic Gaussian noise term scaled by both the diffusion coefficient $\sigma_t$ and the square root of the step size $\sqrt{\tau}$.

\begin{algorithm}
\caption{Sampling from a Diffusion Model (Euler--Maruyama Method)}
\label{algorithm:gen_diff_sample}
\begin{algorithmic}[1]
\Require Neural network $\vf_t^\btheta$, number of steps $n$, diffusion coefficient $\sigma_t$;
\State Set $t = 0$;
\State Set step size $\tau = \frac{1}{n}$;
\State Draw a sample $\rvx_0 \sim p_{\text{base}}$;
\For{$i = 1, 2, \ldots, n$}
\State Draw a sample $\bepsilon \sim \normal(\bzero, \bI_D)$;
\State $\rvx_{t+\tau} = \rvx_t + \tau \vf_t^\btheta(\rvx_t) + \sigma_t \sqrt{\tau} \bepsilon$;
\State Update $t \leftarrow t + \tau$;
\EndFor
\State \Return $\rvx_1$;
\end{algorithmic}
\end{algorithm}

\paragrapharrow{Diffusion models.} 
Equipped with the SDE formulation and its numerical discretization, we can construct generative diffusion models following the same core paradigm used for ODE-based flow models. Recall that the central objective of such generative frameworks is to transform a simple, tractable base distribution $p_{\text{base}}$ into the complex target data distribution $p_{\text{data}}$ (see Figure~\ref{fig:DDPM_gen_model_idea}). 
Similar to ODE-based generation, simulating a stochastic process initialized from $\rvx_0 \sim p_{\text{base}}$ provides a principled approach to this distribution transformation.
To endow the SDE with learnable dynamics, we parameterize its governing vector field $\vf_t$ via a parameterized neural network $\vf_t^\btheta$.
This construction yields the formal definition of a diffusion model, summarized via the following stochastic system
\begin{subequations}\label{equation:gen_diff_models}
\begin{align}
\diff \rvx_t &= \vf_t^\btheta(\rvx_t)\diff t + \sigma_t \diff \rvb_t; && (\text{SDE})\\
\rvx_0 &\sim p_{\text{base}}; && (\text{random initialization}) \\
\rvx_1 &\sim p_{\text{data}}. && (\text{goal}) 
\end{align}
\end{subequations}
We note a key terminology discrepancy with standard DDPM sampling (presented in Algorithm~\ref{alg:diffusion_sampling}): DDPM literature conventionally defines the latent-to-data mapping as the forward process, which is opposite to the convention commonly used for flow-based models.
The full sampling pipeline for diffusion models using the Euler--Maruyama discretization is explicitly outlined in Algorithm~\ref{algorithm:gen_diff_sample}.

\index{Marginal score function}
\index{Conditional score function}
\index{Probability path}
\index{Gaussian probability path}
\index{Conditional probability path}
\index{Marginal probability path}
\subsection{Conditional and Marginal Score Functions}
While diffusion models (reviewed in Sections~\ref{section:dpms} and~\ref{section:diff_model}) are conventionally formulated in terms of noise injection, they can also be interpreted from an alternative perspective rooted in score functions. Accordingly, this section reformulates the preceding developments using the unified language of score functions, vector fields, and probability paths.

We revisit the conditional probability paths $p_{t\mid 1}(\bz\mid\bx)$ and marginal probability paths $p_t(\bz)$ introduced in Section~\ref{section:cond_marg_prob_path}. We formally define the \textit{conditional score function} as $\nabla \ln p_{t\mid 1}(\bz\mid\bx)$ and the \textit{marginal score function} as $\nabla \ln p_t(\bz)$. Analogous to the marginal vector field characterization in Theorem~\ref{theorem:flow_mag_tric}, the marginal score function can be decomposed as a weighted average of the conditional score function $\nabla \ln p_{t\mid 1}(\bz\mid\bx)$, as derived below:
\begin{align}
\nabla \ln p_t(\bz) 
&= \frac{\nabla p_t(\bz)}{p_t(\bz)} = \frac{\nabla \int p_{t\mid 1}(\bz\mid\bx)p_{\text{data}}(\bx)  \diff \bx}{p_t(\bz)}
= \frac{\int \nabla p_{t\mid 1}(\bz\mid\bx)p_{\text{data}}(\bx) \diff \bx}{p_t(\bz)} \nonumber\\
&= \int \nabla \ln p_{t\mid 1}(\bz\mid\bx) \frac{p_{t\mid 1}(\bz\mid\bx)p_{\text{data}}(\bx)}{p_t(\bz)} \diff \bx.
\label{equation:marg_score_func_avg}
\end{align}

Notably, both the conditional vector field $\vf_t(\bz\mid\bx)$ and the score function associated with Gaussian probability paths (see Notes~\ref{note:target_ode_gaussian} and~\ref{note:scfunc_gauss}) take linear forms with respect to $\bz$ and $\bx$. This linearity enables bidirectional conversion between the two quantities, as formalized in the following note.

\begin{noteb}[Score function and conversion formula for Gaussian probability paths]\label{note:scfunc_gauss}
Consider the Gaussian conditional probability path $p_{t\mid 1}(\bz\mid\bx) = \normal(\bz\mid \alpha_t \bx, \beta_t^2 \bI_D)$, where the noise scheduling parameters $\alpha_t$ and $\beta_t$ are defined in Note~\ref{note:gaussian_conditional}. By the definition of the multivariate Gaussian probability density (Definition~\ref{definition:multivariate_gaussian}), the conditional score function admits the closed-form expression (see Problem~\ref{prob:score_multigauss}):
\begin{equation}\label{equation:score_multigauss}
\nabla \ln p_{t\mid 1}(\bz\mid\bx) = \nabla \ln \normal(\bz\mid \alpha_t \bx, \beta_t^2 \bI_D) = -\frac{\bz - \alpha_t \bx}{\beta_t^2}.
\end{equation}
The conditional and marginal vector fields are rigorously linked to their corresponding score functions via the following bidirectional identities:
\begin{subequations}\label{equation:sco_cv_all}
\begin{align}
\vf_t^{\bthetastar}(\bz\mid\bx) 
&= a_t \nabla \ln p_{t\mid 1}(\bz\mid\bx) + b_t \bz;   \label{equation:sco_cv_cd} \\
\vf_t^{\bthetastar}(\bz)
&= a_t \nabla \ln p_t(\bz) + b_t \bz, \label{equation:sco_cv_mar}
\end{align}
\end{subequations}
where the time-dependent coefficients are defined as 
$a_t \triangleq \big( \beta_t^2 \frac{\dot{\alpha}_t}{\alpha_t} - \dot{\beta}_t \beta_t \big)$ 
and 
$ b_t \triangleq \frac{\dot{\alpha}_t}{\alpha_t}$.
This formulation establishes that the conditional (resp., marginal) vector field can be uniquely recovered from the conditional (resp., marginal) score function, and vice versa, enabling full mutual conversion between the two representations.
\end{noteb}
\begin{proof}[of Note~\ref{note:scfunc_gauss}]
We start from the closed-form expression of the Gaussian conditional vector field provided in Note~\ref{note:target_ode_gaussian} and perform algebraic rearrangement:
\begin{align*}
\vf_t^{\bthetastar}(\bz\mid\bx) 
&= \left( \dot{\alpha}_t - \frac{\dot{\beta}_t}{\beta_t} \alpha_t \right) \bx + \frac{\dot{\beta}_t}{\beta_t} \bz 
= \left( \beta_t^2 \frac{\dot{\alpha}_t}{\alpha_t} - \dot{\beta}_t \beta_t \right) \left( \frac{\alpha_t \bx - \bz}{\beta_t^2} \right) + \frac{\dot{\alpha}_t}{\alpha_t} \bz \\
&= \left( \beta_t^2 \frac{\dot{\alpha}_t}{\alpha_t} - \dot{\beta}_t \beta_t \right) \nabla \ln p_{t\mid 1}(\bz\mid\bx) + \frac{\dot{\alpha}_t}{\alpha_t} \bz.
\end{align*}
We then extend this identity to the marginal vector field via the marginalization principle in Theorem~\ref{theorem:flow_mag_tric}. Substituting the decomposed conditional vector field into the marginalization integral yields:
\begin{align*}
\vf_t^{\bthetastar}(\bz) 
&= \int \vf_t^{\bthetastar}(\bz\mid\bx) \frac{p_{t\mid 1}(\bz\mid\bx) p_{\text{data}}(\bx)}{p_t(\bz)} \diff \bx \\
&= \int [a_t \nabla \ln p_{t\mid 1}(\bz\mid\bx) + b_t \bz] \frac{p_{t\mid 1}(\bz\mid\bx) p_{\text{data}}(\bx)}{p_t(\bz)} \diff \bx
= a_t \nabla \ln p_t(\bz) + b_t \bz,
\end{align*}
where the final equality follows directly from the marginal score decomposition in \eqref{equation:marg_score_func_avg} and the normalization property of the posterior density.
\end{proof}

The identities in \eqref{equation:sco_cv_all} from Note~\ref{note:scfunc_gauss} reveal a fundamental equivalence: the optimal vector field $\vf_t^{\bthetastar}$ and the marginal score function $\nabla \ln p_t(\bz)$ contain identical structural information, with each fully determining the other. This equivalence motivates the standard practice of parameterizing and learning the score function $\nabla \ln p_t(\bz)$ via neural networks in modern diffusion models; see Section~\ref{section:sm_flow} for further details.

\index{Laplacian operator}
\index{Fokker--Planck equation}
\subsection{Sampling with SDEs}

Thus far, we have presented the construction of the ODE trajectory $\{\rvx_t\}_{0\leq t\leq 1}$ (see Equations~\eqref{equation:ode_def} and \eqref{equation:flows_def}) that adheres to the target marginal probability path $p_t$ (see Equations~\eqref{equation:flow_conditional_path} and \eqref{equation:flow_sampling_all}) via the marginal vector field $\vf_t^{\bthetastar}$, as formalized in Theorem~\ref{theorem:flow_mag_tric}. Nevertheless, this construction is exclusively limited to flow-based models. A natural extension is to generalize the above results to diffusion models. In the following, we leverage score functions to extend our ODE-based framework to SDEs, as defined in \eqref{equation:sde_def2}.

\begin{theoremHigh}[SDE extension theorem]\label{theorem:sde_ext}
We adopt the prior definitions of the conditional and marginal vector fields $\vf_t^{\bthetastar}(\bz\mid\bx)$ and $\vf_t^{\bthetastar}(\bz)$ (see \eqref{equation:condition_vf} and Theorem~\ref{theorem:flow_mag_tric}). 
For arbitrary diffusion models formulated in \eqref{equation:sde_def2} with a nonnegative diffusion coefficient $\sigma_t \geq 0$ ($0 \leq t \leq 1$), we can construct a valid SDE by augmenting the deterministic dynamics of the original ODE system with stochastic dynamical components. The explicit SDE formulation is presented below:
\begin{align}
\rvx_0 &\sim p_{\text{base}}, \quad \diff \rvx_t = \vf_t^{\bthetastar}(\rvx_t)\diff t + \frac{\sigma_t^2}{2} \nabla \ln p_t(\rvx_t)\diff t + \sigma_t \diff \rvb_t \label{eq:sde_ext1} \\
\implies   \rvx_t &\sim p_t \quad (0 \leq t \leq 1). \label{eq:sde_ext2}
\end{align}
As a direct consequence, the terminal sample satisfies $\rvx_1 \sim p_{\text{data}}$ under this constructed SDE. 
\end{theoremHigh}

Unlike the smooth trajectories of deterministic ODEs, the SDE trajectories exhibit zig-zag fluctuations, which intuitively demonstrate the stochastic nature of SDE evolution (see Figure~\ref{fig:langevin_mog} or Figure~\ref{figure:ou_motion}).  
Critically, despite the stochastic perturbations, the marginal distribution $p_t$ at each time step $t$ remains strictly consistent with the target path, as guaranteed by Theorem~\ref{theorem:sde_ext}. A remarkable property of this result is that the diffusion coefficient $\sigma_t \geq 0$ can be \textbf{arbitrarily} chosen even after the model training stage. Theoretically, the validity of Theorem~\ref{theorem:sde_ext} holds for all nonnegative $\sigma_t$ values.

To prove Theorem~\ref{theorem:sde_ext}, we introduce the \textit{Fokker--Planck equation}, which generalizes the continuity equation for ODEs (established in Lemma~\ref{lemma:cont_equa}) to stochastic SDE systems. To facilitate the proof, we first define the \textit{Laplacian operator} $\Delta$ for a scalar field $u_t : \real^D \to \real$ as follows:
\begin{equation}\label{equation:laplaci_oper}
\Delta u_t(\bz) = \sum_{i=1}^D \frac{\partial^2}{\partial z_i^2} u_t(\bz) = \text{div}(\nabla u_t)(\bz), 
\quad\text{for all }\bz\in\real^D.
\end{equation}

\begin{lemma}[Fokker--Planck equation \citep{holderrieth2025introduction}]\label{lemma:fp_equation}
Let $p_t$ denote a probability path, and consider the SDE
$$ 
\rvx_0 \sim p_{\text{base}}, \quad \diff \rvx_t = \vf_t(\rvx_t)\diff t + \sigma_t \diff \rvb_t. 
$$
The random variable $\rvx_t$ follows the distribution $p_t$ for all $0 \leq t \leq 1$ if and only if the \textit{Fokker--Planck equation} is satisfied:
\begin{equation}\label{equation:fp_equation}
\partial_t p_t(\bz) = -\text{div}(p_t \vf_t)(\bz) + \frac{\sigma_t^2}{2} \Delta p_t(\bz), \quad \text{for all } \bz \in \real^D,\; 0 \leq t \leq 1,
\end{equation}
where $\partial_t p_t(\bz) \triangleq \frac{\diff }{\diff t} p_t(\bz)$ denotes the time derivative of the time-varying probability density $p_t(\bz)$.
\end{lemma}
\begin{proof}[of Lemma~\ref{lemma:fp_equation}]
\textbf{Necessity.}
We first establish the necessity of the Fokker--Planck equation: if $\rvx_t \sim p_t$, then $p_t$ obeys the Fokker--Planck equation. 
The proof relies on test functions---smooth, infinitely differentiable functions $f: \real^D \to \real$ with compact support (i.e., vanishing outside a bounded domain). A key integral characterization of pointwise function equality underpins the derivation: for any integrable functions $g_1, g_2: \real^D \to \real$,
\begin{equation}\label{equation:test_int}
g_1(\bx) = g_2(\bx) \, \forall\, \bx \in \real^D \;\; \Leftrightarrow \;\; \int f(\bx) g_1(\bx) \diff \bx = \int f(\bx) g_2(\bx) \diff \bx \, \forall\text{ test functions } f.
\end{equation}
This equivalence converts pointwise  equality into integral equality over test functions, enabling rigorous manipulation via integration by parts. For compactly supported smooth functions $f_1, f_2$ (such that $f_1(\bx)f_2(\bx)|_{-\infty}^{+\infty}=0$), the integration-by-parts formula for partial derivatives states
\begin{equation}
\int f_1(\bx) \frac{\partial}{\partial x_i} f_2(\bx) \diff \bx = - \int f_2(\bx) \frac{\partial}{\partial x_i} f_1(\bx) \diff \bx.
\end{equation}
Combining this identity with the definitions of the divergence and Laplacian operators (see \eqref{equation:div_def} and \eqref{equation:laplaci_oper}) yields two core integral identities:
\begin{align}
\int \nabla f_1^\top(\bx) \bff_2(\bx) \diff \bx &= - \int f_1(\bx) \text{div}(\bff_2)(\bx) \diff \bx && (f_1: \real^D \to \real, \bff_2: \real^D \to \real^D) \label{equation:int1} \\
\int f_1(\bx) \Delta f_2(\bx) \diff \bx &= \int f_2(\bx) \Delta f_1(\bx) \diff \bx && (f_1: \real^D \to \real, f_2: \real^D \to \real) \label{equation:int2}
\end{align}

We now proceed with the main proof, starting from the SDE trajectory update rule in \eqref{equation:sde_def}. For small time increments $\tau > 0$, the SDE admits the approximation
\begin{align}
\rvx_{t+\tau} &= \rvx_t + \tau \vf_t(\rvx_t) + \sigma_t(\rvb_{t+\tau} - \rvb_t) + \tau \rve_t(\tau)\nonumber \\
&\approx \rvx_t + \tau \vf_t(\rvx_t) + \sigma_t(\rvb_{t+\tau} - \rvb_t) \triangleq \rvx_t +  \tau \vf_t(\rvx_t)+ \sigma_t\bdelta_t, \label{equation:sde_approx}
\end{align}
where $\bdelta_t \triangleq  \rvb_{t+\tau} - \rvb_t$, and the residual error term $\rve_t(\tau)$ is neglected for simplicity, as it vanishes in the limit $\tau \to 0$.
We expand $f(\rvx_{t+\tau}) - f(\rvx_t)$ via second-order Taylor expansion around $\rvx_t$ (Theorem~\ref{theorem:quad_app_theo}), denoted by equality $(\dag)$:
$$
\small
\begin{aligned}
&f(\rvx_{t+\tau}) - f(\rvx_t)  = f(\rvx_t + \tau \vf_t(\rvx_t)+ \sigma_t\bdelta_t) - f(\rvx_t)\\
&\stackrel{(\dag)}{=} \nabla f(\rvx_t)^\top \big(\tau \vf_t(\rvx_t)+ \sigma_t\bdelta_t\big)  + \frac{1}{2} \big(\tau \vf_t(\rvx_t)+ \sigma_t\bdelta_t\big)^\top \nabla^2 f(\rvx_t) \big(\tau \vf_t(\rvx_t)+ \sigma_t\bdelta_t\big) +o(\normtwo{\rvx_{t+\tau}-\rvx_{t}}^2) \\
& \approx\big[ \tau \nabla f(\rvx_t)^\top \vf_t(\rvx_t) + \sigma_t \nabla f(\rvx_t)^\top \bdelta_t\big] 
+ \frac{1}{2} \tau^2 \vf_t(\rvx_t)^\top \nabla^2 f(\rvx_t) \vf_t(\rvx_t) \\
& \quad  + \tau \sigma_t \vf_t(\rvx_t)^\top \nabla^2 f(\rvx_t) \bdelta_t + \frac{1}{2} \sigma_t^2 \bdelta_t^\top \nabla^2 f(\rvx_t) \bdelta_t.
\end{aligned}
$$
The approximation holds since  $\Exp[\rvb_{t+\tau} - \rvb_t \mid \rvx_t] = \bzero$, and we will take $\tau\rightarrow 0$. 
By the properties of Brownian motion, the increment satisfies  $\bdelta_t \mid \rvx_t \sim \normal(\bzero, \tau \bI_D)$. Taking the conditional expectation of the function difference given $\rvx_t$ eliminates all linear and cross terms in $\bdelta_t$, yielding:
\begin{align*}
& \Exp[f(\rvx_{t+\tau}) - f(\rvx_t) \mid  \rvx_t] \\
= & \tau \nabla f(\rvx_t)^\top \vf_t(\rvx_t) + \frac{1}{2} \tau^2 \vf_t(\rvx_t)^\top \nabla^2 f(\rvx_t) \vf_t(\rvx_t) + \frac{\tau}{2} \sigma_t^2 \Exp_{\bepsilon_t \sim \normal(\bzero, \bI_D)}[\bepsilon_t^\top \nabla^2 f(\rvx_t) \bepsilon_t] \\
\stackrel{(\dag)}{=} & \tau \nabla f(\rvx_t)^\top \vf_t(\rvx_t) + \frac{1}{2} \tau^2 \vf_t(\rvx_t)^\top \nabla^2 f(\rvx_t) \vf_t(\rvx_t) + \frac{\tau}{2} \sigma_t^2 \trace( \nabla^2 f(\rvx_t)) \\
= & \tau \nabla f(\rvx_t)^\top \vf_t(\rvx_t) + \frac{1}{2} \tau^2 \vf_t(\rvx_t)^\top \nabla^2 f(\rvx_t) \vf_t(\rvx_t) + \frac{\tau}{2} \sigma_t^2 \Delta f(\rvx_t),
\end{align*}
where the equality  $(\dag)$ follows from the fact that $\Exp_{\bepsilon_t \sim \normal(\bzero, \bI_D)}[\bepsilon_t^\top \bA \bepsilon_t] = \trace(\bA)$ (see Problem~\ref{prob:score_hutchinson}), and the last equality follows from the definition of the Laplacian operator \eqref{equation:laplaci_oper} and the Hessian matrix. 
Therefore, we have 
$$
\small
\begin{aligned}
& \partial_t \Exp[f(\rvx_t)] 
= \lim_{\tau \to 0} \frac{1}{\tau} \Exp[f(\rvx_{t+\tau}) - f(\rvx_t)]
\stackrel{\eqref{equation:tower_property}}{=} \lim_{\tau \to 0} \frac{1}{\tau} \Exp\big[\Exp[f(\rvx_{t+\tau}) - f(\rvx_t) \mid \rvx_t]\big] \\
& = \Exp\left[\lim_{\tau \to 0} \frac{1}{\tau} \left( \tau \nabla f(\rvx_t)^\top \vf_t(\rvx_t) + \frac{1}{2} \tau^2 \vf_t(\rvx_t)^\top \nabla^2 f(\rvx_t) \vf_t(\rvx_t) + \frac{\tau}{2} \sigma_t^2 \Delta f(\rvx_t) \right)\right] \\
& = \Exp[\nabla f(\rvx_t)^\top \vf_t(\rvx_t) + \frac{1}{2} \sigma_t^2 \Delta f(\rvx_t)]
\stackrel{(\dag)}{=} \int \nabla f(\bz)^\top \vf_t(\bz) p_t(\bz) \diff \bz + \int \frac{1}{2} \sigma_t^2 \Delta f(\bz) p_t(\bz) \diff \bz \\
& \stackrel{(\ddag)}{=} - \int f(\bz) \text{div}(\vf_t p_t)(\bz) \diff \bz + \int \frac{1}{2} \sigma_t^2 f(\bz) \Delta p_t(\bz) \diff \bz 
= \int f(\bz) \left( -\text{div}(\vf_t p_t)(\bz) + \frac{\sigma_t^2}{2}  \Delta p_t(\bz) \right) \diff \bz,
\end{aligned}
$$
where the equality  ($\dag$) follows from  the assumption that $p_t$ is the distribution of $\rvx_t$,
and the equality ($\ddag$) follows from the observation in \eqref{equation:int1}. 
This manipulation requires $p_t \vf_t$ to be integrable, i.e., $\int p_t(\bz) \normtwo{\vf_t(\bz)} \diff \bz < \infty$, a condition universally satisfied in machine learning due to bounded data domains and numerically constrained function magnitudes.
Therefore, it holds that
$$
\small
\begin{aligned}
\int f(\bz) \left( -\text{div}(p_t \vf_t)(\bz) + \frac{\sigma_t^2}{2} \Delta p_t(\bz) \right) \diff \bz &=\partial_t \Exp[f(\rvx_t)]  
\stackrel{(\dag)}{=}  \int f(\bz) \partial_t p_t(\bz) \diff \bz  \quad (\forall\, f \text{, } 0 \leq t \leq 1)\\
\stackrel{(\ddag)}{\iff} \quad \partial_t p_t(\bz) &= -\text{div}(p_t \vf_t)(\bz) + \frac{\sigma_t^2}{2} \Delta p_t(\bz) \quad (\forall\, \bz \in \real^D, 0 \leq t \leq 1) ,
\end{aligned}
$$
where the equality $(\dag)$ follows from  the assumption that $\rvx_t \sim p_t$ and swapping the derivative with the integral, 
and the implication $(\ddag)$ follows from \eqref{equation:test_int}. This shows that the Fokker--Planck equation is a necessary condition.

\paragraph{Sufficiency.}
We finally establish sufficiency: if $p_t$ satisfies the Fokker--Planck equation \eqref{equation:fp_equation}, then $p_t$ is exactly the marginal distribution of $\rvx_t$ for all $t \in [0,1]$.
The Fokker--Planck equation is a linear parabolic partial differential equation  (PDE). As established in well-posedness results for parabolic PDEs (e.g., \citet{evans2022partial}) and analogous to the existence and uniqueness result in Theorem~\ref{theorem:flow_exist}, such equations admit a unique solution for fixed initial conditions. Let $q_t$ denote the true marginal distribution of the SDE solution $\rvx_t$. From the necessity argument above, $q_t$ must also satisfy the same Fokker--Planck equation \eqref{equation:fp_equation}. By construction of the probability interpolation path, the initial conditions coincide: $p_0 = q_0 = p_{\text{base}}$.
Uniqueness of parabolic PDE solutions forces $p_t = q_t$ for all $0 \leq t \leq 1$, meaning $\rvx_t \sim p_t$. This completes the proof of sufficiency.
\end{proof}

Setting the diffusion coefficient $\sigma_t = 0$ recovers the continuity equation derived in Lemma~\ref{lemma:cont_equa} as a special case of the Fokker--Planck equation. The additional Laplacian diffusion term $\Delta p_t$ in the Fokker--Planck equation  is introduced here as a mathematical diffusion term that characterizes the stochastic spreading of the probability density $p_t$ induced by the SDE's Brownian motion component. Equipped with the Fokker--Planck equation, we now proceed to prove Theorem~\ref{theorem:sde_ext}.

\begin{proof}[Theorem~\ref{theorem:sde_ext}]
By the equivalence result in Lemma~\ref{lemma:fp_equation}, verifying that the SDE defined in \eqref{eq:sde_ext1} generates the probability path $p_t$ reduces to confirming that $p_t$ satisfies the corresponding Fokker--Planck equation.
We start from the continuity equation in Lemma~\ref{lemma:cont_equa}, which gives the base temporal evolution of the density:
\begin{align*}
\partial_t p_t(\bz) 
&= -\text{div}(p_t \vf_t^{\bthetastar})(\bz) 
= -\text{div}(p_t \vf_t^{\bthetastar})(\bz) - \frac{\sigma_t^2}{2} \Delta p_t(\bz) + \frac{\sigma_t^2}{2} \Delta p_t(\bz) \\
&\stackrel{(\dag)}{=} -\text{div}(p_t \vf_t^{\bthetastar})(\bz) - \text{div}\left(\frac{\sigma_t^2}{2} \nabla p_t\right)(\bz) + \frac{\sigma_t^2}{2} \Delta p_t(\bz) \\
&\stackrel{(\ddag)}{=} -\text{div}(p_t \vf_t^{\bthetastar})(\bz) - \text{div}\left(p_t \left[\frac{\sigma_t^2}{2} \nabla \ln p_t\right]\right)(\bz) + \frac{\sigma_t^2}{2} \Delta p_t(\bz) \\
&= -\text{div}\left(p_t \left[\vf_t^{\bthetastar} + \frac{\sigma_t^2}{2} \nabla \ln p_t\right]\right)(\bz) + \frac{\sigma_t^2}{2} \Delta p_t(\bz),
\end{align*}
where the equality ($\dag$) follows from the definition of the Laplacian (see \eqref{equation:laplaci_oper}), the equality ($\ddag$) follows form the fact  that $\nabla \ln p_t = \frac{\nabla p_t}{p_t}$, and the last equality follows from the linearity of the divergence operator. 
This derivation confirms that the density $p_t$ associated with the SDE in \eqref{eq:sde_ext1} satisfies the Fokker--Planck equation. Invoking Lemma~\ref{lemma:fp_equation}, we conclude that the SDE solution obeys $\rvx_t \sim p_t$ for all $0 \leq t \leq 1$, which completes the proof.
\end{proof}

\paragrapharrow{Connection to Langevin dynamics.}
The stochastic dynamical system formulated in Theorem~\ref{theorem:sde_ext}  \eqref{eq:sde_ext1} exhibits core properties consistent with classical Langevin dynamics~\eqref{equation:lagevin_dyn_def_ddpm}. 
Specifically, it introduces structured Gaussian noise to the dynamical system while rigorously preserving the time-evolving marginal distribution $p_t$. 
The construction underlying Theorem~\ref{theorem:sde_ext} and \eqref{eq:sde_ext1} admits a well-known special case corresponding to a stationary probability path, where $p_t = p^*$ for some fixed target distribution $p^*$. Under this stationary setting, we set $\vf_t^{\bthetastar} = \bzero$, which reduces the general SDE to the simplified form:
\begin{equation}
\diff \rvx_t = \frac{\sigma_t^2}{2} \nabla \ln p^*(\rvx_t) \diff t + \sigma_t \diff \rvb_t,
\end{equation}
This equation defines standard Langevin dynamics, in which all time-dependent directional flow components vanish, leaving only gradient-induced drift and Brownian motion noise (cf.~\eqref{equation:lagevin_dyn_def_ddpm}).
The stationarity condition $p_t \equiv p^*$ implies $\partial_t p_t(\bz) = 0$. Direct substitution via Theorem~\ref{theorem:sde_ext} verifies that the resulting dynamics satisfy the Fokker--Planck equation associated with the constant distribution path $p_t = p^*$. This confirms that $p^*$ constitutes the stationary distribution of Langevin dynamics, namely:
$$ 
\rvx_0 \sim p^* 
\quad \implies \quad 
\rvx_t \sim p^* \quad (t \geq 0). 
$$
Consistent with the general behavior of Markov processes, Langevin dynamics converge to their stationary distribution $p^*$ under standard regularity conditions. For an initial distribution $\rvx_0 \sim p' \neq p^*$ with corresponding time-evolving marginal $p'_t$, the marginal distribution converges to $p^*$ asymptotically under mild technical assumptions. 
Notably, the Ornstein--Uhlenbeck process (presented in Example~\ref{example:ou_process}) emerges as a Gaussian-specialized instance of Langevin dynamics, and it forms the original mathematical foundation for early diffusion model formulations.

To conclude this subsection, we show that for Gaussian probability paths, learning the marginal vector field yields the corresponding score function directly, with no additional optimization required.
\begin{noteb}[SDE extension trick for Gaussian probability paths]
Consider the Gaussian conditional probability path $p_{t\mid 1}(\bz\mid\bx) = \normal(\bz\mid \alpha_t \bx, \beta_t^2 \bI_D)$ where the noise scheduling coefficients $\alpha_t, \beta_t$  are defined in  Note~\ref{note:gaussian_conditional}. 
Leveraging the Gaussian score function characterization established in Note~\ref{note:scfunc_gauss}, we can rewrite the SDE derived in Theorem~\ref{theorem:sde_ext} entirely in terms of score functions:
\begin{align}
\rvx_0 &\sim p_{\text{base}}, \quad \diff \rvx_t = \left[ \left(a_t + \frac{\sigma_t^2}{2}\right) \nabla \ln p_t(\rvx_t) + b_t \rvx_t \right] \diff t + \sigma_t \diff \rvb_t \\
\implies   \rvx_t &\sim p_t \quad (0 \leq t \leq 1), 
\end{align}
where  
$a_t = \big( \beta_t^2 \frac{\dot{\alpha}_t}{\alpha_t} - \dot{\beta}_t \beta_t \big)$ 
and 
$ b_t = \frac{\dot{\alpha}_t}{\alpha_t}$.
In particular, $\rvx_1\sim p_{\text{data}}$ for this SDE.
\end{noteb}

\index{Score network}
\index{Score matching loss}
\index{Denoising score matching loss}
\index{Conditional score matching loss}
\subsection{Score Matching: Revisited}\label{section:sm_flow}
We next introduce a general framework for learning the marginal score function $\nabla \ln p_t(\bz)$ via probability path modeling. As demonstrated above, the marginal vector field $\vf_t^{\bthetastar}(\bz)$ can be readily converted into the score function for Gaussian probability paths using the results in Note~\ref{note:scfunc_gauss}. Nevertheless, a more general learning formulation is required for arbitrary probability paths. We show that marginal score functions can be learned directly via supervised optimization. To approximate the marginal score $\nabla \ln p_t$, we again adopt a parameterized neural network $\bs_t^\btheta: \real^D \times [0,1] \to \real^D$, termed the \textit{score network}. 
Following the previously established optimization paradigm, we define two canonical training objectives: the \textit{score matching (SM) loss} and the  \textit{conditional score matching (CSM) loss}, both of which fall under the family of denoising score matching objectives (cf. \eqref{equation:fmloss_def} and \eqref{equation:cfmloss_def}) \citep{hyvarinen2005estimation, vincent2011connection, song2020sliced}:
\begin{mybox}
\begin{align*}
\mathcalJ_{\text{SM}}(\btheta) &= \Exp_{t \sim \uniformdist, \bx \sim p_{\text{data}}, \bz \sim p_{t\mid 1}(\cdot|\bx)} \left[ \normtwo{\bs_t^\btheta(\bz) - \nabla \ln p_t(\bz)}^2 \right]; &&  \text{(SM loss)} \\
\mathcalJ_{\text{CSM}}(\btheta) &= \Exp_{t \sim \uniformdist, \bx \sim p_{\text{data}}, \bz \sim p_{t\mid 1}(\cdot|\bx)} \left[ \normtwo{\bs_t^\btheta(\bz) - \nabla \ln p_{t\mid 1}(\bz\mid\bx)}^2 \right]. &&  \text{(conditional SM loss)}
\end{align*}
\end{mybox}
The core distinction between the two objectives lies in the regression target: the SM loss optimizes toward the marginal score $\nabla \ln p_t(\bz)$, whereas the CSM loss targets the conditional score $\nabla \ln p_{t\mid 1}(\bz\mid\bx)$.
Ideally, training is performed by minimizing the SM loss; however, direct optimization is infeasible since the marginal score $\nabla \ln p_t(\bz)$ is analytically intractable in practice. Consistent with our prior analysis, the denoising score matching framework provides a tractable surrogate objective via the CSM loss, with theoretical equivalence guaranteed by the following theorem.

\begin{theoremHigh}[SM  and CSM losses]\label{theorem:smcsm_equiv}
The SM loss and CSM loss differ only by a constant term independent of network parameters. Specifically,
$$ \mathcalJ_{\text{SM}}(\btheta) = \mathcalJ_{\text{CSM}}(\btheta) + C, $$
where $C$ denotes a constant independent of  $\btheta$. 
As a direct consequence, the two objectives yield identical parameter gradients:
$$ 
\nabla_\btheta \mathcalJ_{\text{SM}}(\btheta) = \nabla_\btheta \mathcalJ_{\text{CSM}}(\btheta). 
$$
\end{theoremHigh}
\begin{proof}[of Theorem~\ref{theorem:smcsm_equiv}]
The marginal score $\nabla \ln p_t$ (characterized in \eqref{equation:marg_score_func_avg}) shares an identical structural form with the marginal conditional vector field $\vf_t^{\bthetastar}$ defined in \eqref{equation:marginal_vf_def}. Accordingly, the proof follows exactly the same logic as that of Theorem~\ref{theorem:fmcfm_equiv}, with the vector field $\vf_t^{\bthetastar}$ replaced by the marginal score function $\nabla \ln p_t$.
\end{proof}

This equivalence theorem implies that the optimal parameter minimizer $\widehatbtheta$ of the CSM loss also minimizes the original SM loss. Therefore, the trained score network satisfies $\bs_t^{\widehatbtheta} = \nabla \ln p_t$, yielding an exact approximation of the target marginal score function.

\begin{noteb}[Denoising diffusion models: Score matching for Gaussian probability paths]
We now instantiate the CSM loss for the Gaussian conditional probability path $p_{t\mid 1}(\bz\mid\bx) = \normal(\alpha_t \bx, \beta_t^2 \bI_D)$. As derived in Note~\ref{note:scfunc_gauss}, the closed-form expression for the corresponding conditional score function is given by:
\begin{equation}
\nabla \ln p_{t\mid 1}(\bz\mid\bx) = -\frac{\bz - \alpha_t \bx}{\beta_t^2}. 
\end{equation}
Substituting this analytical form into the CSM loss definition yields the expanded objective:
\begin{align*}
\mathcalJ_{\text{CSM}}(\btheta) 
&= \Exp_{t \sim \uniformdist, \bx \sim p_{\text{data}}, \bz \sim p_{t\mid 1}(\cdot|\bx)} \left[ \normtwo{\bs_t^\btheta(\bz) + \frac{\bz - \alpha_t \bx}{\beta_t^2}}^2 \right] \\
&\stackrel{(\dag)}{=} \Exp_{t \sim \uniformdist, \bx \sim p_{\text{data}}, \bepsilon \sim \normal(\bzero, \bI_D)} \left[ \normtwo{\bs_t^\btheta(\alpha_t \bx + \beta_t \bepsilon) + \frac{\bepsilon}{\beta_t}}^2 \right] \\
&= \Exp_{t \sim \uniformdist, \bx \sim p_{\text{data}}, \bepsilon \sim \normal(\bzero, \bI_D)} \left[ \frac{1}{\beta_t^2} \normtwo{\beta_t \bs_t^\btheta(\alpha_t \bx + \beta_t \bepsilon) + \bepsilon}^2 \right],
\end{align*}
where the equality ($\dag$) follows by plugging in Note~\ref{note:fm_got}~\eqref{equation:fm_got1} and replacing $\bz$ with $\alpha_t \bx + \beta_t \bepsilon$. 

This formulation reveals that the score network $\bs_t^\btheta$ essentially learns to recover the Gaussian noise injected during the data corruption process, which motivates the naming of denoising score matching. Despite its effectiveness, this original loss formulation suffers from numerical instability when $\beta_t \to 0$, meaning the objective is only well-behaved with sufficiently large injected noise. To resolve this issue, early denoising diffusion model works (refer to Algorithm~\ref{alg:diffusion_training}) propose discarding the scaling constant $\frac{1}{\beta_t^2}$ and reparameterizing the score network into a dedicated \textit{noise predictor network} $\bepsilon_t^\btheta: \real^D \times [0,1] \to \real^D$ via the transformation:
$$ 
-\beta_t \bs_t^\btheta(\bz) = \bepsilon_t^\btheta(\bz) \quad \implies \quad \mathcalJ_{\text{DDPM}}(\btheta) = \Exp_{t \sim \uniformdist, \bx \sim p_{\text{data}}, \bepsilon \sim \normal(\bzero, \bI_D)} \left[ \normtwo{ \bepsilon_t^\btheta(\alpha_t \bx + \beta_t \bepsilon) - \bepsilon}^2 \right].
$$
That is, this reparameterization simplifies the training objective to the standard DDPM loss (cf. \eqref{equation:ddpm_equaweight_sum_loss2}).
The resulting noise predictor network retains the core learning objective of estimating the noise corrupting clean data samples. We summarize the complete training pipeline for Gaussian probability paths in Algorithm~\ref{alg:score_gauss_path}.
\end{noteb}

\begin{algorithm}[H]
\caption{Score Matching Training Procedure (for Gaussian Probability Path $p_{t\mid 1}(\bz\mid \bx) = \normal(\alpha_t\bx, \beta_t^2\bI)$)}
\label{alg:score_gauss_path}
\begin{algorithmic}[1]
\Require  Training data $\mathcalX=\{\bx_n\}$, score network $\bs_t^\btheta$ (or noise predictor $\bepsilon_t^\btheta$);
\State \textbf{initialize:} $\btheta$;
\For{$cnt=0,1,2,\ldots$}
\State  Sample a data example $\bx$ from the dataset;
\State  Sample a random time $t \sim \uniformdist{[0,1]}$;
\State  Sample noise $\bepsilon \sim \normal(\bzero, \bI_D)$;
\State  Set  $\bz \sim p_{t\mid 1}(\cdot\mid \bx)$;\Comment{Gaussian path: $\bz \leftarrow  \alpha_t \bx + \beta_t \bepsilon$}
\State  Loss $\mathcalJ(\btheta) \leftarrow \normtwo{ \bs_t^\btheta(\bz) - \nabla \ln p_{t\mid 1}(\bz|\bx)}^2$; \Comment{Gaussian path: 
$=\begin{cases}
\normtwo{\bs_t^\btheta(\bz) + \frac{\bepsilon}{\beta_t}}^2\\
\normtwo{\bepsilon_t^\btheta(\bz) - \bepsilon}^2
\end{cases}	$
}
\State  Update $\btheta \leftarrow \btheta-\eta \nabla \mathcalJ(\btheta)$; \Comment{Gradient descent}
\EndFor
\State \Return Optimized network parameter $\btheta$;
\end{algorithmic}
\end{algorithm}

After completing network training, we can generate novel data samples by simulating the parameterized SDE with arbitrary nonnegative diffusion coefficients $\sigma_t \geq 0$. The full SDE formulation is as follows (via Theorem~\ref{theorem:sde_ext}):
\begin{align}
\rvx_0 \sim p_{\text{base}}, \quad 
\diff \rvx_t &= \left[ \vf_t^\widehatbtheta(\rvx_t) + \frac{\sigma_t^2}{2}\bs_t^\widehatbtheta(\rvx_t) \right] \diff t + \sigma_t \diff \rvb_t  \\
&\stackrel{\text{Gauss path}}{\longeq} \left[ \left(a_t + \frac{\sigma_t^2}{2}\right) \bs_t^\widehatbtheta(\rvx_t) + b_t \rvx_t \right] \diff t + \sigma_t \diff \rvb_t .
\end{align}
Proper simulation of this SDE yields approximate samples $\rvx_1$ aligned with the target data distribution $p_{\text{data}}$, where the optimal diffusion schedule $\sigma_t \geq 0$ can be determined via empirical validation.

\index{Guided diffusion model}
\index{Guided flow model}
\section{Guidance: Revisited}\label{section:guidance_flow}

We previously introduced guidance for DDPMs in Section~\ref{section:guid_ddpm} from the score function perspective, as formulated in \eqref{equation:class_guid_decomp} and \eqref{equation:cfg_ddpm}; these frameworks enable sample generation conditioned on supplementary contextual information. In this section, we extend this guidance paradigm to stochastic differential equation (SDE) and ordinary differential equation (ODE) generative frameworks, which are governed by a core vector field component (see \eqref{equation:gen_flow_models} and \eqref{equation:gen_diff_models}). For Gaussian probability paths, the SDE/ODE and score-based perspectives are mathematically equivalent, as validated by the conversion formula presented in Note~\ref{note:scfunc_gauss}.

\subsection{Simple Guidance}

We first elaborate on the simple construction strategy for guided generative models. The core methodology is straightforward: the input condition prompt $\by$ is fed into the network during both training and inference, following the standard pipeline of unconditional generative modeling. We formalize this conditional modeling framework as follows. We define the conditioning variable (prompt) $\by$ residing in a general space $\sY$. Specifically, $\sY$ corresponds to the full text embedding space for text prompts, or a discrete set for categorical class labels, with no restrictive assumptions imposed on the structure of $\sY$.

We define a \textit{guided diffusion model} via a neural network-parameterized \textit{guided vector field} $\vf_t^\btheta(\bz\mid\by)$ paired with a time-varying diffusion coefficient $\sigma_t$ (see \eqref{equation:sde_def2}). 
The formal definitions are specified below:
\begin{align*}
	\textbf{Neural network: } & \vf^\btheta : \real^D \times \sY \times [0, 1] \to \real^D, \quad (\bz, \by, t) \mapsto \vf_t^\btheta(\bz\mid\by); \\
	\textbf{Fixed: } & \sigma_t : [0, 1] \to [0, \infty), \quad t \mapsto \sigma_t.
\end{align*}
Here, the vector field $\vf_t^\btheta$ is explicitly conditioned on the input variable $\by \in \sY$. 
Given any valid guidance signal $\by \in \sY$, model sampling proceeds via the following SDE dynamics:
\begin{subequations}\label{equation:all_sgd_guided}
	\begin{align}
		\rvx_0 &\sim p_{\text{base}}; && (\text{initial condition}) \\
		\diff \rvx_t &= \vf_t^\btheta(\rvx_t\mid\by)\diff t + \sigma_t \diff \rvb_t; && (\text{SDE}) \\
		\rvx_1 &\sim p_{\text{data}}(\cdot\mid\by). && (\text{Goal}) 
	\end{align}
\end{subequations}
The special case $\sigma_t = 0$ recovers a \textit{guided flow model}.
For concise exposition, the subsequent analysis focuses primarily on flow matching and pure flow models, though all theoretical conclusions generalize directly to the full guided diffusion framework in \eqref{equation:all_sgd_guided}.

We next address the training objective for the guided flow model $\vf_t^\btheta(\bz\mid\by)$. 
For a fixed guidance prompt  $\by$, the data distribution reduces to the conditional distribution $p_{\text{data}}(\bx\mid\by)$, which degenerates to the unconditional generative modeling setup.
This allows us to adopt a conditional flow matching (CFM) training objective of the form:
\begin{equation}
	\Exp_{\bx \sim p_{\text{data}}(\cdot\mid\by), \bz \sim p_{t\mid 1}(\cdot\mid\bx)} 
	\big[\normtwobig{\vf^\btheta_t(\bz\mid\by) - \vf_t^{\bthetastar}(\bz\mid\bx)}^2\big],
\end{equation}
where $\vf_t^{\bthetastar}(\bz\mid\bx)$ denotes the optimal target conditional vector field.
Notably, the conditional prompt $\by$ does not alter the underlying conditional probability path $p_t(\cdot\mid\bx)$ or the target conditional vector field $\vf_t^{\bthetastar}(\bz\mid\bx)$. 
By marginalizing the objective over all valid prompts $\by$ sampled from the joint data distribution, we derive the final \textit{guided conditional flow matching loss} (cf. \eqref{equation:cfmloss_def}):
\begin{mybox}
	\begin{equation}\label{equation:cfmloss_def_guided}
		\mathcalJ_{\text{CFM}}^{\text{guided}}(\btheta) = \Exp_{t \sim \uniformdist[0,1], (\bx,\by) \sim p_{\text{data}}(\bx,\by), \bz \sim p_{t\mid 1}(\cdot\mid\bx)} 
		\big[\normtwobig{\vf^\btheta_t(\bz\mid\by) - \vf_t^{\bthetastar}(\bz\mid\bx)}^2\big].
	\end{equation}
\end{mybox}
The key distinction between this guided loss and the unconditional baseline in \eqref{equation:cfmloss_def} lies in the data sampling procedure. Instead of sampling only data samples $\bx$, we sample paired data-condition tuples $(\bx, \by)$ from the joint distribution $p_{\text{data}}(\bx,\by)$, which characterizes the paired dataset (e.g., paired images and text prompts) used for guidance generative modeling.

\subsection{Classifier and Classifier-Free Guidance}\label{section:cfg_flow}
In principle, vanilla guidance yields generation that faithfully follows the conditional data distribution $p_{\text{data}}(\cdot\mid\by)$. 
Nevertheless, as outlined in Section~\ref{section:guid_ddpm}, empirical observations reveal that samples produced via this standard guidance often fail to align sufficiently well with the target guidance $\by$ \citep{dhariwal2021diffusion, ho2022classifier}. This mismatch stems from multiple factors, including model underfitting---where the network fails to accurately recover the true underlying vector field---and inherent noise in real-world training data, such as noisy or inaccurate web-crawled text-image pairs. To enhance sample alignment with the given prompt, it is therefore necessary to artificially strengthen the guidance signal $\by$. The dominant technique for achieving this improvement is classifier-free guidance, a standard component in modern state-of-the-art diffusion and flow matching models. 
In this subsection, we revisit and formalize this approach from a vector-field perspective.

\index{Classifier guidance}
\paragrapharrow{Classifier guidance.} 
We begin by introducing classifier guidance under the vector field framework \citep{dhariwal2021diffusion, holderrieth2025introduction}. 
We restrict our discussion to Gaussian probability paths for simplicity. As established in Note~\ref{note:gaussian_conditional}, the Gaussian conditional probability path is defined as $p_{t\mid 1}(\cdot\mid \bx) = \normal(\alpha_t \bx, \beta_t^2 \bI_D)$. 
Here, the noise schedule coefficients $\alpha_t$ and $\beta_t$ are continuously differentiable and strictly monotonic, adhering to the boundary conditions $\alpha_0 = \beta_1 = 0$ and $\alpha_1 = \beta_0 = 1$. 
Building on Note~\ref{note:scfunc_gauss}, we reformulate the {\textit{target guided vector field} $\vf_t^{\bthetastar}(\bz\mid\by)$} via the \textit{guided score function} $\nabla \ln p_t(\bz\mid\by)$
\begin{equation}
	\vf_t^{\bthetastar}(\bz\mid\by) = a_t \nabla \ln p_t(\bz\mid\by) + b_t \bz,
	\footnote{Note that $p_{t\mid 1}(\bz\mid\bx)$ denotes the conditional distribution where the data $\bx $ follows  $p_1= p_{\text{data}}(\bx)$, while $p_t(\bz\mid\by)$ represents the conditional distribution conditioned on the guidance variable $\by$.}
\end{equation}
where the time-dependent coefficients are defined as 
$a_t \triangleq \big( \beta_t^2 \frac{\dot{\alpha}_t}{\alpha_t} - \dot{\beta}_t \beta_t \big)$ 
and 
$ b_t \triangleq \frac{\dot{\alpha}_t}{\alpha_t}$.
Given that $p_t(\bz\mid\by)$ is a conditional probability density, we apply Bayes' rule~\eqref{equation:bayes_base} to decompose the guided score function:
\begin{align}
	p_t(\bz\mid\by) &= \frac{p_t(\bz) p_t(\by\mid\bz)}{p_t(\by)}\\
	\implies \quad \nabla_\bz \ln p_t(\bz\mid\by) &= \nabla_\bz \ln \left( \frac{p_t(\bz) p_t(\by\mid\bz)}{p_t(\by)} \right) = \nabla_\bz \ln p_t(\bz) + \nabla_\bz \ln p_t(\by\mid\bz). 
	\label{equation:score_equiv_cond}
\end{align}
The simplification above holds because the gradient operator $\nabla$ acts exclusively on $\bz$, such that $\nabla \ln p_t(\by) = \bzero$. 
Substituting this decomposed score function and using Note~\ref{note:scfunc_gauss} again yield an expanded form of the guided vector field:
\begin{equation}
	\vf_t^{\bthetastar}(\bz\mid\by) = b_t \bz + a_t (\nabla \ln p_t(\bz) + \nabla \ln p_t(\by\mid\bz)) = \vf_t^{\bthetastar}(\bz) + a_t \nabla \ln p_t(\by\mid\bz).
\end{equation}
This decomposition reveals that the conditional guided vector field $\vf_t^{\bthetastar}(\bz\mid\by)$ consists of two components: the unguided vector field $\vf_t^{\bthetastar}(\bz)$, and an additional gradient term corresponding to the log-likelihood of the guidance variable $\by$ given the noisy sample $\bz$. 

In practical diffusion generation, generated samples often fail to sufficiently align with the input guidance prompt $\by$. To mitigate this issue, a common heuristic is to amplify the contribution of the conditional log-likelihood gradient term, leading to the definition of the \textit{scaled guided vector field} for classifier guidance:
\begin{equation}\label{equation:scaled_class_guidance}
	\widetildebv_t(\bz\mid\by) 
	\triangleq 
	\vf_t^{\bthetastar}(\bz) + \lambda a_t \nabla \ln p_t(\by\mid\bz),
\end{equation}
where $\lambda \geq 0$ denotes the \textit{guidance scale} (typically set to $\lambda>1$ in practice). 
The log-likelihood term $\ln p_t(\by\mid\bz)$ can be interpreted as a classification function for noisy data, which quantifies the log-probability of the guidance condition $\by$ given the noisy latent $\bz$. This function can be directly optimized via supervised learning, forming the core of standard classifier guidance \citep{dhariwal2021diffusion}. 
Note again that the scaled guidance formulation is a heuristic modification: for all $\lambda \neq 1$, the scaled vector field satisfies $\widetildebv_t(\bz\mid\by) \neq \vf_t^{\bthetastar}(\bz\mid\by)$, meaning it no longer corresponds to the mathematically exact guided vector field.

\index{Classifier-free guidance (CFG)}
\paragrapharrow{Classifier-free guidance.}
Although classifier-based guidance is theoretically feasible, it suffers from notable practical limitations. First, this paradigm requires training a dedicated classifier network in parallel with a flow or diffusion model, doubling the total number of learnable networks from one to two. Second, for high-dimensional conditioning variables $\by$---such as textual prompts, rather than discrete class labels---the conditional distribution $p_t(\by\mid\bz)$ becomes extremely challenging to model, making the corresponding gradient term $\nabla \ln p_t(\by\mid\bz)$ difficult to compute reliably. To address these limitations, \textit{classifier-free guidance  (CFG)} was proposed in \citep{ho2022classifier} as a substitute approach that achieves theoretically identical guidance performance while eliminating the need for an auxiliary classifier network.

Similar to the CFG for diffusion models (Section~\ref{section:cfg_ddpm}), the derivation of CFG in the context of vector fields starts by substituting the equivalence relation in \eqref{equation:score_equiv_cond} into the classifier guidance formulation, yielding the following transformation:
\begin{align}
	\widetildebv_t(\bz\mid\by) &= \vf_t^{\bthetastar}(\bz) + \lambda a_t \nabla \ln p_t(\by\mid\bz) 
	= \vf_t^{\bthetastar}(\bz) + \lambda a_t [\nabla \ln p_t(\bz\mid\by) - \nabla \ln p_t(\bz)] \nonumber\\
	&= \big(\vf_t^{\bthetastar}(\bz) - [\lambda b_t \bz + \lambda a_t \nabla \ln p_t(\bz)]\big) + [\lambda b_t \bz + \lambda a_t \nabla \ln p_t(\bz\mid\by)] \nonumber\\
	&= (1 - \lambda) \vf_t^{\bthetastar}(\bz) + \lambda \vf_t^{\bthetastar}(\bz\mid\by). \label{equation:scaled_class_guidance2}
\end{align}
This result demonstrates that the scaled guided vector field $\widetildebv_t(\bz\mid\by)$ can be decomposed as a linear interpolation between the unguided vector field $\vf_t^{\bthetastar}(\bz)$ and the conditionally guided vector field $\vf_t^{\bthetastar}(\bz\mid\by)$. A naive implementation of this formulation would require training two distinct model branches: an unguided branch optimized via the unconditional flow matching loss in \eqref{equation:cfmloss_def}, and a conditional branch trained with the guided flow matching loss in \eqref{equation:cfmloss_def_guided}. The two branches are then linearly combined during inference to produce the final guided vector field $\widetildebv_t(\bz\mid\by)$.

Instead of training separate networks, we can unify both learning objectives within a single model by introducing an additional \textbf{null label} $\varnothing$ to represent the absence of conditional guidance. With this augmentation, the unguided vector field can be formally redefined as the conditionally guided vector field conditioned on the null label, i.e., $\vf_t^{\bthetastar}(\bz) = \vf_t^{\bthetastar}(\bz\mid\varnothing)$.
Substituting this null-label formulation into \eqref{equation:scaled_class_guidance2} yields the standard \textit{classifier-free guided vector field}:
\begin{align}
	\widetildebv_t(\bz\mid\by) 
	&= (1 - \lambda)\vf_t^{\bthetastar}(\bz\mid \varnothing) + \lambda \vf_t^{\bthetastar}(\bz\mid\by) \nonumber\\
	&\simeq\vf_t^{\bthetastar}(\bz\mid\by)  \qquad \text{replace } \by = \varnothing \text{ with prob. } \gamma. \label{equation:scaled_class_guidance_free}
\end{align}
This design completely removes the need for an auxiliary classifier network to enable conditional guidance. The unified training strategy that jointly learns conditional and unconditional generation capabilities within a single model, followed by linear interpolation at inference time to strengthen conditional alignment.

\paragrapharrow{Training classifier-free guidance.}
To implement CFG in practice, we revise the guided CFM training objective defined in \eqref{equation:cfmloss_def_guided} to accommodate the null conditioning label $\varnothing$. A key obstacle arises from the standard data sampling pipeline: paired training samples $(\bx, \by)$ drawn from the data distribution $p_{\text{data}}$ never naturally contain the null label $\varnothing$. Consequently, we artificially introduce null conditioning during training via a stochastic masking strategy. Specifically, we define a hyperparameter $\gamma$ that controls the probability of discarding the original conditional label $\by$ and replacing it with $\varnothing$ for each training sample. This augmentation yields the final \textit{CFG conditional flow matching (CFG-CFM) loss function}, formulated as follows:
\begin{mybox}
\begin{align}
\mathcalJ_{\text{CFM}}^{\text{CFG}}&(\btheta) 
= \Exp_{\diamondsuit} \big[\normtwobig{\vf_t^\btheta(\bz\mid\by) - \vf_t^{\bthetastar}(\bz\mid\bx)}\big]^2  , \label{equation:cfmloss_def_cfg}\\
\diamondsuit &=  t \sim \uniformdist[0, 1],\,  (\bx, \by) \sim p_{\text{data}}(\bx, \by), \,  \bz \sim p_{t\mid 1}(\cdot\mid\bx), \, \text{replace } \by = \varnothing \text{ with prob. } \gamma. \nonumber
\end{align}
\end{mybox}
The procedure is described in Algorithm~\ref{alg:flow_matching_cfg}.
\begin{algorithm}[h]
\caption{Classifier-free guidance training for Gaussian probability path $p_{t\mid 1}(\bz\mid\bx) = \normal(\bz\mid  \alpha_t \bx, \beta_t^2 \bI_D)$, e.g., $p_{t\mid 1}(\bz\mid\bx) = \normal(\bz\mid  t \bx, (1-t)^2 \bI_D)$ in the Gaussian CondOT path case (cf. Algorithm~\ref{alg:flow_matching}).}
\label{alg:flow_matching_cfg}
\begin{algorithmic}[1]
\Require Paired dataset $(\bx, \by) \sim p_{\text{data}}$, neural network $\vf_t^\btheta$, dropout probability $0<\gamma<1$;
\State \textbf{initialize:} $\btheta$;
\For{$cnt=0,1,2,\ldots$}
\State Sample a data example $(\bx, \by)$ from the dataset;
\State Sample a random time $t \sim \uniformdist{[0,1]}$;
\State Sample noise $\bepsilon \sim \normal(\bzero, \bI_D)$;
\State Set $\bz \sim p_{t\mid 1}(\cdot \mid \bx) = \alpha_t \bx + \beta_t \bepsilon$;  \Comment{see \eqref{equation:gaussian_conditional}}
\State With probability $\gamma$ drop label: $\by \leftarrow \varnothing$;
\State Compute loss $\small \mathcalJ(\btheta) \leftarrow \normtwobig{\vf_t^\btheta(\bz\mid\by) - \vf_t^{\bthetastar}(\bz\mid\bx)}= \normtwobig{\vf_t^\btheta(\bz\mid\by) - (\dot{\alpha}_t \bx + \dot{\beta}_t \bepsilon)}^2 $; \Comment{\eqref{equaiton:gauss_prob_path_weavg_spec}}
\State Update  $\btheta\leftarrow \btheta-\eta \nabla \mathcalJ(\btheta)$; \Comment{Gradient descent}
\EndFor
\State \Return $\btheta$;
\end{algorithmic}
\end{algorithm}

At inference time, for a fixed choice of $\by$, we may sample via
\begin{align*}
\rvx_0 & \sim p_{\text{base}}(\bz); \quad && \text{Initialize with simple distribution (such as a Gaussian)} \\
\diff \rvx_t & = \widetildebv_t^{\btheta}(\rvx_t\mid\by) \diff t; & & (\text{Simulate ODE from } t=0 \text{ to } t=1) \\
\rvx_1&\sim p_{\text{data}}(\cdot\mid\by).   && \text{Goal is for } \rvx_1 \text{ to adhere to the guiding variable } \by
\end{align*}
Note that the distribution of $\rvx_1$ is not necessarily aligned with $\rvx_1 \sim p_{\text{data}}(\cdot\mid\by)$ anymore if we use a weight $\lambda > 1$. However, empirically, this shows better alignment with conditioning. Classifier-free guidance is therefore a \textbf{heuristic} that is predominantly justified by its excellent empirical results. In fact, almost any image or video that you see that is AI-generated relied heavily on classifier-free guidance $\lambda \geq 4$ \citep{ho2022classifier}.

\begin{problemset}
\item \label{prob:ddpmscore_vincent_form} Prove that the ESM and DSM lose functions defined in \eqref{equation:ddpm_modified_score_loss} and   \eqref{equation:ddpmscore_vincent_form} are mathematically equivalent. \textit{Hint: Expand the squared norm in both formulations and compare the individual terms.}

\item \label{prob:brownian_motion} Let $\Delta t$ denote a small timestep (infinitesimal in the limit $\Delta t\rightarrow 0$), and let $\diff \rvb_t$ denote the increment of a Wiener process (Brownian motion; see Section~\ref{section:diff_model}). And let $\bepsilon\sim\normal(\bzero, \bI)$. Show that $\bepsilon (\Delta t)^{1/2} = \diff \rvb_t$. \textit{Hint: Consider the covariance of $\bepsilon (\Delta t)^{1/2}$ and $\Delta \rvb_t =  \rvb_{t+\Delta t}-\rvb_t$.}

\item \textbf{OU process stationary limit.} Prove that in the long-time limit $t \to \infty$, this OU process~\eqref{equation:ouprocess} 
converges asymptotically to a stationary Gaussian distribution as $t \to \infty$, with stationary distribution $\normal(\bzero, \sigma^2/(-2a)\bI)$.

\item \label{prob:score_multigauss} Derive the score function of a multivariate Gaussian probability density $\normal(\bx\mid \bmu, \bSigma)$ (Definition~\ref{definition:multivariate_gaussian}): $\nabla_\bx\normal(\bx\mid \bmu, \bSigma) = -\bSigma^{-1}(\bx-\bmu)$. 
\textit{Hint: For a symmetric matrix $\bA$ and vector $\bv$,  the gradient identity $\nabla_\bv(\bv^\top\bA\bv)=2\bA\bv$ holds.}

\item \label{prob:int12} Provide rigorous proofs for the integral identities  \eqref{equation:int1} and \eqref{equation:int2}. \textit{Hint: For the first equality, establish the integration-by-parts relation $$\int f_{1}(\bx) \frac{\partial}{\partial x_i}  \bff_2(\bx)_i \diff \bx = -\int \bff_2(\bx)_i \frac{\partial}{\partial x_i}  f_1(\bx)\diff \bx,$$ then complete the general proof.}

\item Following the framework established in Section~\ref{section:cfg_flow}, discuss how classifier guidance can be extended to probability paths that are non-Gaussian.

\item \label{prob:sde_meanvar} \textbf{Solution of linear SDEs.} Consider the SDE: $\diff \rvx_t =\bF(t)\rvx_t \diff t + \bG(t) \diff \rvb_t$, where $\rvb_t\in\real^{P}$ denotes a Brownian motion (Section~\ref{section:diff_model}), $\bF(t)\in\real^{D\times D}$, $\bG(t)\in\real^{D\times P}$. Show that 
\begin{equation}
\rvx_t=\bPhi(t)\rvx_0 + \int_0^t {\bPhi(t)} \bPhi(s)^{-1} \bG(s)\diff \rvb_s,
\end{equation}
where $\bPhi(t)= \exp\left(\int_0^t \bF(s)\diff s\right)$.
Using this result to show that 
\begin{equation}
\begin{aligned}
\Exp[\rvx_t\mid \rvx_0=\bx_0] &= \bPhi(t)\bx_0;\\
\qquad 
\Cov[\rvx_t\mid \rvx_0=\bx_0] 
&= \Cov[\int_0^t {\bPhi(t)} \bPhi(s)^{-1} \bG(s)\diff \rvb_s] \\
&= \bPhi(t) \left(\int_0^t \big(\bPhi(s)^{-1} \bG(s)\big) \big(\bPhi(s)^{-1} \bG(s)\big)^\top \diff s \right)  \bPhi(t)^\top.
\end{aligned}
\end{equation}
\textit{Hint: 
Let $\bM_t \triangleq \bPhi(t)^{-1}$ and multiply $\bM_t$ to both sides of the SDE.
Consider the \text{vector It\^o isometry}, which states that for an adapted matrix-valued process $\rmH_s \in \real^{D \times P}$ (measurable with respect to the filtration generated by a $P$-dimensional Wiener process $\rvb_s$ up to time $s$) satisfying
$
\Exp\left[ \int_0^t \normf{\rmH_s}^2 \, \diff s \right] < \infty,
$
the following identity holds:
$$
\Exp\left[ \left( \int_0^t \rmH_s \, \diff \rvb_s \right) \left( \int_0^t \rmH_s \, \diff \rvb_s \right)^\top \right]
= \Exp\left[ \int_0^t \rmH_s \rmH_s^\top \, \diff s \right].
$$
}

\item \label{prob:sde_meanvar2}  \textbf{SDE for a DPM or DDPM \citep{anderson1982reverse, song2020score, kingma2021variational}.}
Consider the forward process in a DPM or DDPM given in \eqref{equation:ddpm_diffusion_kernel_all}: $\bx_t=\nu_t\bx_0 + \sigma_t\bepsilon_t$, $\bepsilon_t\sim\normal(\bzero, \bI_D)$. Show that this can be modeled by the SDE
\begin{equation}\label{equation:sde_meanvar2}
\diff \rvx_t = f(t)\rvx_t \diff t + g(t)\diff \rvb_t, \quad  \rvx_0\sim p_{\text{data}}(\bx),
\end{equation}
where 
\begin{equation}
f(t) = \frac{\diff \ln \nu_t}{\diff t}, \qquad 
g^2(t) = \frac{\diff \sigma_t^2}{\diff t}  - 2\frac{\diff \ln \nu_t}{\diff t}\sigma_t^2\equiv -2\sigma_t^2 \frac{\diff  \xi_t}{\diff  t},
\end{equation}
and  $\xi_t \triangleq \ln\left(\nu_t / \sigma_t\right)$ denotes the log-SNR (signal-to-noise ratio).
Now let $\rvx_t\sim p_t(\bx)$. We consider the reverse SDE from $t=1$ to $0$ (in the normalized timestep case). Show that 
\begin{equation}
\diff \rvx_t = [f(t)\rvx_t -g^2(t) \nabla \ln p_t(\rvx_t)]\diff t + g(t)\diff \overline{\rvb}_t, \quad \rvx_1\sim \normal(\bzero, \bI_D),
\end{equation}
where $\overline{\rvb}_t$ denotes the Wiener process in the reverse time.
\textit{Hint: Use Problem~\ref{prob:sde_meanvar}.}
\end{problemset}

\newpage 
\chapter{Advanced Sampling for Diffusion Models}\label{chapter:dpmsampler}
\begingroup
\hypersetup{
	linkcolor=structurecolor,
	linktoc=page,  
}
\minitoc \newpage
\endgroup
\lettrine{\color{caligraphcolor}D}
Diffusion models have emerged as a powerful framework for generative modeling, achieving state-of-the-art results across image, audio, and text generation tasks (Chapters~\ref{chapter:diff}--\ref{chapter:scorematch}). 
At their core, these models operate by iteratively corrupting data with Gaussian noise (the forward noising process) and learning to reverse this process to recover clean samples (the reverse denoising process). This bidirectional dynamics can be rigorously formalized using stochastic differential equations (SDEs) for the forward process and their corresponding ordinary differential equations (ODEs) for the deterministic reverse process. However, a critical practical challenge remains: efficiently sampling high-quality outputs from these models, as naive numerical integration of the underlying ODE/SDE often requires hundreds of steps to produce coherent results.

This chapter is dedicated to advancing the state of the art in diffusion sampling by exploring two influential families of fast solvers: \textit{DEIS} (\textit{diffusion exponential integrator sampler}) and \textit{DPM-Solver} (\textit{diffusion probabilistic model solver}). 
These methods are designed to drastically reduce the number of function evaluations needed for high-fidelity sampling, making diffusion models more practical for real-world applications.

We begin by revisiting the foundational SDE and ODE formulations of diffusion models (Section~\ref{section:dm_sampler}), establishing the mathematical groundwork for the sampling methods that follow. We clarify the relationship between the forward noising SDE, the reverse denoising ODE, and their analytical solutions---key insights that enable the design of efficient numerical integrators.

Building on this foundation, we introduce DEIS (Section~\ref{section:deis}), a family of samplers that leverages exponential integrators to precisely handle the stiff linear components of the diffusion ODE. We demonstrate how DEIS outperforms naive Euler methods, improves accuracy through noise prediction, and extends to higher-order schemes via polynomial extrapolation, enabling high-quality sampling with as few as 10--20 steps.

Next, we turn to DPM-Solver (Section~\ref{section:dpmsolver}), another class of fast ODE solvers that exploits analytically computable coefficients to achieve high-order accuracy with minimal computational overhead. We cover its single-step and multistep variants, showing how it estimates derivative terms to further accelerate convergence. We then extend this discussion to DPM-Solver++ (Section~\ref{section:dpmsolver++}), an enhanced framework that unifies and improves upon earlier DPM-Solver variants, offering a more flexible and robust approach to diffusion sampling. We also compare these methods and explore their variants, highlighting their trade-offs in speed, quality, and numerical stability.

\index{Variance preserving}
\section{Diffusion Model with SDE: Revisited}\label{section:dm_sampler}

Note that in the diffusion model Chapter~\ref{chapter:diff}, the forward process refers to the noise-adding process, and the reverse process refers to the denoising process; intermediate states are denoted by $\bz_t$ to reflect the latent variable structure.
In the flow- and score-matching Chapters~\ref{chapter:flow}--\ref{chapter:scorematch}, the forward process denotes the transformation from a noise distribution to the target data distribution; intermediate states are denoted by $\bx_t$ to reflect the probabilistic transformation structure.
To avoid inconsistency, in this chapter we use the terms \textbf{noising process}  and \textbf{denoising process} for the two procedures, and denote intermediate states by $\bx_t$.

When discussing diffusion models, we aim to approximate the target probability distribution $p_0 \triangleq p_{\text{data}}(\rvx_0)$.
This is accomplished by defining a noising process $\{\bx_t\}_{t \in [0,T]}$, in which $\bx_t$ accumulates increasing noise as time $t$ increases.
For a given data point $\bx_0$, the state at time $t$ satisfies:
\begin{equation}\label{equation:dpm_diff_kernel}
q(\bx_t \mid \bx_0) = \normal\left(\bx_t\mid \nu_t \bx_0, \sigma_t^2 \bI_D\right)
\quad \implies \quad 
\bx_t = \nu_t \bx_0 + \sigma_t \bepsilon_t, \quad \bepsilon_t \sim \normal(\bzero, \bI_D),
\end{equation}
where $ \nu_t $ and $ \sigma_t $ are predefined, differentiable functions of $ t $ ($\nu_t=\sqrt{\alpha_t}$ and $\sigma_t=\sqrt{1-\alpha_t}$ in \eqref{equation:ddpm_diffusion_kernel_all} such that $\nu_t^2+\sigma_t^2=1$, i.e., \textit{variance preserving (VP)}~\footnote{A \textit{variance exploding (VE)} extension is discussed in Problem~\ref{prob:ve} \citep{song2020score}.}).
Intuitively, these parameters control the signal-to-noise ratio of the diffusion process: $ \nu_t $ decreases and $ \sigma_t $ increases with $ t $.
This process can be described by a stochastic differential equation (SDE):
\begin{equation}\label{equation:nois_diff_sde}
\diff \rvx_t = f(t) \rvx_t \, \diff t + g(t) \, \diff \rvb_t, \quad \rvx_0 \sim p_{\text{data}}(\rvx_0) ,
\end{equation}
where $ \rvb_t $ is a standard Wiener process (see Section~\ref{section:diff_model}), and
\begin{equation}
f(t) = \frac{\diff \ln \nu_t}{\diff t}, \quad g^2(t) = \frac{\diff \sigma_t^2}{\diff t} - 2 \frac{\diff \ln \nu_t}{\diff t} \sigma_t^2; 
\end{equation}
see Problems~\ref{prob:sde_meanvar}--\ref{prob:sde_meanvar2} or Theorem~\ref{theorem:dpmforward}.

All quantities above are known.
Equation \eqref{equation:dpm_diff_kernel} allows us to gradually corrupt the target distribution into a Gaussian distribution.
DPMs or DDPMs learn to recover data $\bx_0\sim p_{\text{data}}$ from the noisy input $\bx_T\sim\normal(\bzero, \bI_D)$ using a sequential denoising procedure. There are three alternative formulations of the model (Section~\ref{section:ddpm_final_remark}).
The \textit{noise prediction model} $\bepsilon^\btheta_t(\bx_t)$ seeks to predict the noise $\bepsilon_t$ from $\bx_t$, optimizing the parameter $\btheta$ via the following objective (see \eqref{equation:ddim_ddpmloss}):
\begin{equation}\label{equation:sampler_ddpmloss}
\min_\btheta \left\{\mathcalJ(\btheta)=\Exp_{\bx_0,\bepsilon,t}\left[\zeta_t\normtwo{\bepsilon^\btheta_t(\bx_t) - \bepsilon_t}^2\right]\right\}, 
\end{equation}
where $\bx_0 \sim p_{\text{data}}(\bx_0)$, $\bepsilon_t \sim \normal(\bzero, \bI_D)$, $t \sim \uniformdist([0, T])$, and $\zeta_t > 0$ is a weighting function. Alternatively, the \textit{data prediction model} $\bx^\btheta_t(\bx_t)$ predicts the original data $\bx_0$ from the noisy state $\bx_t$. Its relationship to $\bepsilon^\btheta_t(\bx_t)$ is given by   (see \eqref{equation:dpm_diff_kernel})
\begin{equation}\label{equation:dpmsolver_data_pred}
\bx^\btheta_t(\bx_t) \triangleq \frac{1}{\nu_t}(\bx_t - \sigma_t \bepsilon^\btheta_t(\bx_t)).
\end{equation}
The \textit{score prediction model} $\bs^\btheta_t(\bx_t)$ predicts the score function (see \eqref{equation:score_noise}):
\begin{equation}\label{equation:score_noise_sampler}
\bs^\btheta_t(\bx_t) \triangleq -\frac{1}{\sigma_t}\bepsilon^\btheta_t(\bx_t).
\end{equation}

For generative sampling, we require the reverse process of \eqref{equation:nois_diff_sde}, which maps a Gaussian distribution back to the target data distribution.
Under suitable conditions, it can be shown that this denoising process is also an SDE (see Problems~\ref{prob:sde_meanvar}--\ref{prob:sde_meanvar2} \citep{anderson1982reverse, song2020score}):
\begin{equation}\label{equation:denois_diff_sde}
\diff \rvx_t = \left[ f(t) \rvx_t - g^2(t) \nabla \ln p_t(\rvx_t) \right] \diff t + g(t) \, \diff \overline{\rvb}_t, \quad \rvx_T \sim p_T(\rvx_T) \equiv\normal(\bzero, \bI_D),
\end{equation}
where $\overline{\rvb}_t$ denotes the Wiener process in the reverse time.
The only unknown term is the score function $ \nabla \ln p_t(\rvx_t) $.
Once this term is estimated, we can solve \eqref{equation:denois_diff_sde}.

\index{log-signal-to-noise ratio}
\index{log-SNR}
\subsection{The Noising SDE}

To obtain the parameters $f(t), g(t)$ in the denoising SDE \eqref{equation:denois_diff_sde}, we may instead derive the noising SDE \eqref{equation:nois_diff_sde}.
We previously showed that the forward process of a DPM or DDPM corresponds to an SDE in Problems~\ref{prob:sde_meanvar}--\ref{prob:sde_meanvar2} using solutions to linear SDEs. Here we provide an alternative, direct derivation.
In diffusion models, the interpretation of the coefficients $f(t), g(t)$ is less intuitive than that of the noise schedule parameters $\nu_t, \sigma_t$. We therefore seek to express $f(t), g(t)$ in terms of $\nu_t, \sigma_t$. Recall that $\nu_t, \sigma_t$ are defined via the perturbation kernel in \eqref{equation:dpm_diff_kernel}.
Furthermore, let $\xi_t$ denote the \textit{log-signal-to-noise ratio (log-SNR)}:
\begin{equation}\label{equation:logsnr}
\xi_t \triangleq \ln\left(\frac{\nu_t  }{\sigma_t}\right).
\end{equation}
\begin{theoremHigh}[DPM, DDPM forward SDE]\label{theorem:dpmforward}
The forward process of a DPM or a DDPM corresponds to an SDE of the form
$$
\diff \rvx_t = f(t)\rvx_t \diff t  + g(t) \diff \rvb_t,
$$
where the relationships between $f(t), g(t)$ and $\nu_t, \sigma_t, \xi_t$ are given by
\begin{equation}\label{equation:dpmforward}
f(t) = \frac{\diff  \ln \nu_t}{\diff  t}, 
\quad 
g^2(t) = -2\sigma_t^2 \frac{\diff  \xi_t}{\diff  t} =\frac{\diff \sigma_t^2}{\diff t}  - 2\frac{\diff \ln \nu_t}{\diff t}\sigma_t^2.
\end{equation}
Using the standard DPM/DDPM parameterization from  \eqref{equation:ddpm_diffusion_kernel_all} ($\nu_t=\sqrt{\alpha_t}$ and $\sigma_t=\sqrt{1-\alpha_t}$), these expressions simplify to
\begin{equation}\label{equation:dpmforward2}
f(t) = \frac{1}{2}\frac{\diff \ln \alpha_t}{\diff t}
, \quad 
g^2(t) = -\frac{\diff \ln \alpha_t}{\diff t}.
\end{equation}
That is, $f(t) = -g^2(t)$.
\end{theoremHigh}
\begin{proof}[of Theorem~\ref{theorem:dpmforward}]
For this Markov process, consider the transition probability from time $s$ to time $t$ ($s < t$). Solving via the perturbation kernel yields (see Problems~\ref{prob:ddpm_trans_kernel2}--\ref{prob:ddpm_trans_kernel3}):
\begin{align*}
q(\bx_t \mid \bx_s) &= \normal\left( \bx_t\mid \frac{\nu_t}{\nu_s} \bx_s, \left( \sigma_t^2 - \sigma_s^2 \frac{\nu_t^2}{\nu_s^2} \right) \bI_D \right);\\
\bx_t &= \frac{\nu_t}{\nu_s} \bx_s + \sqrt{\sigma_t^2 - \sigma_s^2 \frac{\nu_t^2}{\nu_s^2}} \, \bepsilon, \quad \bepsilon \sim \normal(\bzero, \bI_D).
\end{align*}
It follows that
\begin{equation}\label{equation:dpmode1}
\bx_t - \bx_s = \frac{\nu_t - \nu_s}{\nu_s} \bx_s + \sigma_t \sqrt{1 - e^{2(\xi_t - \xi_s)}} \, \bepsilon, \quad \bepsilon \sim \normal(\bzero, \bI_D).
\end{equation}
Taking the limit $s \to t$ (see Problem~\ref{prob:dpmode2}) gives :
\begin{equation}\label{equation:dpmode2}
\diff \bx_t = \frac{\dot{\nu}_t}{\nu_t} \bx_t \diff t 
+ \sigma_t \sqrt{-2\dot{\xi}_t} \diff \rvb_t.
\end{equation}
Note that $\xi_t$ is monotonically decreasing, so the expression under the square root is positive (see Section~\ref{section:forward_rev_ddpm}). Comparing this with the general diffusion SDE
$\diff \bx_t = f(t)\bx_t \diff t  + g(t) \diff \rvb_t$
establishes the relationships between $f(t), g(t)$ and $\nu_t, \sigma_t, \xi_t$.
\end{proof}

\subsection{The Denoising ODE}

Equation \eqref{equation:denois_diff_sde} is a continuous-time SDE.
To solve it numerically, we must discretize it by replacing $ \diff t $ with a timestep $ \Delta t $.
Due to the stochastic nature of the Wiener process, large timesteps introduce substantial errors and may hinder convergence.
In practice, DDPMs often use 1000 or 4000 sampling steps, which is computationally costly.
We therefore seek an equivalent ODE formulation.

\begin{theoremHigh}[Equivalent ODE for diffusion SDE]\label{theorem:ode_diffsde}
The denoising process SDE \eqref{equation:denois_diff_sde} admits a corresponding ODE that preserves the same marginal distribution $p_t$ at each timestep.
Identical results can be obtained by solving the ODE:
\begin{subequations}\label{equation:diff_denoid_ode}
\begin{align}
\frac{\diff \rvx_t}{\diff t} &= f(t) \rvx_t - \frac{1}{2} g^2(t) \nabla \ln p_t(\rvx_t), \quad \rvx_T \sim p_T(\rvx_T) \equiv\normal(\bzero, \bI_D); \label{equation:diff_denoid_ode1}\\
\frac{\diff \rvx_t}{\diff t} &= f(t) \rvx_t - \frac{1}{2} g^2(t) \bs^\btheta_t(\rvx_t), \quad \rvx_T \sim p_T(\rvx_T)\equiv\normal(\bzero, \bI_D); \label{equation:diff_denoid_ode3}\\
\frac{\diff \rvx_t}{\diff t} &= f(t) \rvx_t + \frac{g^2(t)}{2\sigma_t} \bepsilon^\btheta_t(\rvx_t), \quad \rvx_T \sim p_T(\rvx_T)\equiv\normal(\bzero, \bI_D). \label{equation:diff_denoid_ode2}
\end{align}
\end{subequations}
where $\bs^\btheta_t(\rvx_t)$ denotes the score network, $\bepsilon^\btheta_t(\rvx_t)$ denotes the noise network, and the third equality follows from the relationship between score and diffusion noise  (see \eqref{equation:score_noise_sampler}).
\end{theoremHigh}
\begin{proof}[of Theorem~\ref{theorem:ode_diffsde}]
Consider any SDE of the form:
$$
\diff  \rvx_t = \bu_t(\rvx_t) \diff t + \zeta_t \diff \rvb_t.
$$
The time evolution of its probability density $p_t(\rvx_t)$ is governed by the {Fokker--Planck equation} (Lemma~\ref{lemma:fp_equation}):
$$
\frac{\partial p_t}{\partial t} = -\text{div} \left( p_t \bu_t \right) + \frac{\zeta^2_t}{2} \Delta p_t.
$$
where $\texttt{div}$ denotes the divergence operator (see \eqref{equation:div_def}) and $\Delta$ denotes the Laplacian (see \eqref{equation:laplaci_oper}).
For a deterministic ODE:
$$
{\diff \rvx_t} = \bv_t(\rvx_t) \diff t,
$$
the density evolution follows the {continuity equation} (Lemma~\ref{lemma:cont_equa}):
$$
\frac{\partial p_t}{\partial t} = -\text{div} \left(  p_t \bv_t \right).
$$
Requiring identical marginal distributions implies that the two density evolution equations must be equivalent.
The vector field of the reverse SDE is:
$$
\bu_t(\rvx_t) = f(t) \rvx_t - g^2(t) \nabla \ln p_t(\rvx_t),
$$
with diffusion coefficient $\zeta_t = g(t)$. Substituting into the Fokker--Planck equation yields:
$$
\frac{\partial p_t}{\partial t} = -\text{div} \left[ \left( f(t)\rvx_t - g^2(t) \nabla \ln p_t \right) p_t \right] + \frac{g^2(t)}{2} \Delta p_t .
$$
Since Tthe continuity equation for the ODE is:
$\frac{\partial p_t}{\partial t} = -\text{div} \left( p_t\bv_t  \right)$, we equate the right-hand sides:
$$
-\text{div} \left( p_t\bv_t \right) = -\text{div} \left[ \left( f(t)\rvx_t - g(t)^2 \nabla \ln p_t \right) p_t \right] + \frac{g(t)^2}{2} \Delta p_t.
$$
Using the identity $\nabla \ln p_t = \frac{\nabla p_t}{p_t}$, we have $g(t)^2 \nabla \ln p_t \cdot p_t = g(t)^2 \nabla p_t$. Expanding the first divergence term on the right-hand side:
$$
\text{div} \left[ \left( f(t)\rvx_t - g(t)^2 \nabla \ln p_t \right) p_t \right] = \text{div} \left( f(t)\rvx_t p_t \right) - \text{div} \left( g(t)^2 \nabla p_t \right).
$$
Substitute back and rearrange:
$$
\text{div} \left( p_t\bv_t \right) 
= \text{div} \left( f(t)\rvx_t p_t \right) - \text{div} \left( g(t)^2 \nabla p_t \right) + \frac{g(t)^2}{2} \text{div} (\nabla p_t)
= \text{div} \left[ f(t)\rvx_t p_t - \frac{g(t)^2}{2} \nabla p_t \right].
$$
For this to hold for all $\rvx_t$, the vector fields inside the divergence must be equal (divergence-free fields do not affect density evolution and can be ignored):
$$
p_t\bv_t = f(t)\rvx_t p_t - \frac{g^2(t)}{2} \nabla p_t.
$$
Dividing  both sides by $p_t$ and substituting  $\frac{\nabla p_t}{p_t} = \nabla \ln p_t$ yields the vector field of the probability flow ODE:
$$
\bv_t(\rvx_t) = f(t) \rvx_t - \frac{g^2(t)}{2} \nabla \ln p_t(\rvx_t),
$$
which matches the first expression in \eqref{equation:diff_denoid_ode} and completes the proof.
\end{proof}

Equation \eqref{equation:diff_denoid_ode1} is a semi-linear stiff ODE {\citep{hochbruck2010exponential}} that consists of a linear term $f(t) \rvx_t$ and a nonlinear term $\nabla \ln p_t(\rvx_t)\approx \bs^\btheta_t(\rvx_t)$.
The distribution of the ODE
trajectories of \eqref{equation:diff_denoid_ode1} with terminal distribution $\bx_T\sim p_{\text{data}}$ coincides with that of the SDE \eqref{equation:denois_diff_sde} with initial distribution $\bx_T \sim p_T$. 
In other words, the ODE \eqref{equation:diff_denoid_ode1} and the SDE \eqref{equation:denois_diff_sde} share the same probability law.
Consequently, in principle, we can generate new samples from the data distribution $p_{\text{data}}$ by simulating the backward diffusion ODE \eqref{equation:diff_denoid_ode1}.
ODE solvers are well-established in numerical analysis.
By solving the ODE instead of the SDE, the number of sampling steps in diffusion models can be drastically reduced---for example, from 1000 steps to just 10.

However, solving \eqref{equation:diff_denoid_ode1} requires evaluating the score function $\nabla \ln p_t(\bx)$, which is not directly available.
The core idea of diffusion models is to use a time-dependent network $\bs^\btheta_t(\bx)$, referred to as a score network, to approximate the score $\nabla \ln p_t(\bx)$. 
This is accomplished using score-matching techniques (Chapter~\ref{chapter:scorematch}), where the score network $\bs^\btheta_t$ is trained by minimizing the conditional score matching loss
\begin{equation}\label{equation:score_lss_sampler}
\mathcalJ(\btheta)
= \Exp_{t \sim \uniformdist[0,T], p(\bx_0), q(\bx_t\mid \bx_0)} \left[ \normtwo{\nabla \ln q(\bx_t\mid \bx_0) - \bs^\btheta_t(\bx_t)}^2 \right]. 
\end{equation}
Here, $\nabla \ln q(\bx_t\mid \bx_0)$ admits  a closed-form expression because $q(\bx_t\mid \bx_0)$ is a simple Gaussian distribution (see \eqref{equation:dpm_diff_kernel}). 
This loss can be estimated from empirical samples via Monte Carlo methods, allowing standard stochastic optimization algorithms to be used for training.

\subsection{Diffusion Denoising ODE and SDE Solutions}

In this subsection, we present a detailed derivation of the solutions to the diffusion denoising SDE and ODE.

\paragrapharrow{Diffusion SDE solution.}
Sampling using diffusion models can be implemented by solving the \textit{diffusion SDE} given in \eqref{equation:denois_diff_sde}:
\begin{equation}\label{equation:denois_diff_sde_sol}
\diff \rvx_t = \left[f(t)\rvx_t + \frac{g^2(t)}{\sigma_t}\bepsilon^\btheta_t(\rvx_t)\right]\diff t + g(t)\diff\overline{\rvb}_t, \quad \rvx_T \sim p_T(\rvx_T) \equiv\normal(\bzero, \bI_D),
\end{equation}
where $\overline{\rvb}_t$ denotes the reverse-time Wiener process over the interval from $T$ to $0$, i.e., from $t=1$ to $t=0$ in the normalized timestep setting. 

\begin{theoremHigh}[Exact solution formula for diffusion SDE]\label{theorem:sde_semilineare}
The exact solution formula for the SDE~\eqref{equation:denois_diff_sde_sol} is given by:
\begin{equation}\label{equation:sde_semilineare}
\rvx_t = \underbrace{e^{\int_s^t f(\tau)\diff\tau}\rvx_s}_{\text{Deterministic linear flow}} 
+ \underbrace{\int_s^t e^{\int_\tau^t f(m)\diff m} \frac{g^2(\tau)}{\sigma_\tau}\bepsilon_\tau^\btheta(\rvx_\tau)\diff\tau}_{\text{Nonlinear score-matching term}} 
+ \underbrace{\int_s^t e^{\int_\tau^t f(m)\diff m} g(\tau)\diff\overline{\rvb}_\tau}_{\text{Stochastic Brownian motion term}}.
\end{equation}
The first term in the expression can be evaluated analytically in closed form.
\end{theoremHigh}
\begin{proof}[of Theorem~\ref{theorem:sde_semilineare}]
We isolate the linear term of the diffusion SDE~\eqref{equation:denois_diff_sde_sol} on the left-hand side:
$$
\diff\rvx_t - f(t)\rvx_t\diff t = \frac{g^2(t)}{\sigma_t}\bepsilon_t^\btheta(\rvx_t)\diff t + g(t)\diff\overline{\rvb}_t.
$$
Define a function $\psi(t)$ such that:
$$
\diff\psi(t) = -\psi(t)f(t)\diff t.
$$
Integrating this ODE gives:
$$
\psi(t) = \psi(0)e^{-\int_0^t f(\tau)\diff\tau}
\qquad \implies\qquad 
\frac{\psi(s)}{\psi(t)} = e^{\int_s^t f(\tau)\diff\tau}.
$$
Multiply both sides of the rearranged SDE by $\psi(t)$:
$$
\psi(t)\diff\rvx_t - \psi(t)f(t)\rvx_t\diff t = \psi(t)\frac{g^2(t)}{\sigma_t}\bepsilon_t^\btheta(\rvx_t)\diff t + \psi(t)g(t)\diff\overline{\rvb}_t.
$$
By It\^o's lemma (since $\psi(t)$ is deterministic), the left-hand side corresponds to the stochastic differential of $\psi(t)\rvx_t$:
$$
\diff\left(\psi(t)\rvx_t\right) = \psi(t)\frac{g^2(t)}{\sigma_t}\bepsilon_t^\btheta(\rvx_t)\diff t + \psi(t)g(t)\diff\overline{\rvb}_t.
$$
Integrate both sides over the interval $[s, t]$:
$$
\int_s^t \diff\left(\psi(\tau)\rvx_\tau\right) = \int_s^t \psi(\tau)\frac{g^2(\tau)}{\sigma_\tau}\bepsilon_\tau^\btheta(\rvx_\tau)\diff\tau + \int_s^t \psi(\tau)g(\tau)\diff\overline{\rvb}_\tau.
$$
Evaluating the left-hand side and rearranging to isolate $\rvx_t$:
$$
\psi(t)\rvx_t = \psi(s)\rvx_s + \int_s^t \psi(\tau)\frac{g^2(\tau)}{\sigma_\tau}\bepsilon_\tau^\btheta(\rvx_\tau)\diff\tau + \int_s^t \psi(\tau)g(\tau)\diff\overline{\rvb}_\tau.
$$
Dividing by $\psi(t)$ and substituting $\frac{\psi(s)}{\psi(t)} = e^{\int_s^t f(\tau)\diff\tau}$ (and $\frac{\psi(\tau)}{\psi(t)} = e^{\int_\tau^t f(m)\diff m}$) yields the desired result.
\end{proof}

Substituting the relationship from Theorem~\ref{theorem:dpmforward}~\eqref{equation:dpmforward} into \eqref{equation:sde_semilineare} produces:
\begin{mybox}
\begin{equation}\label{equation:sde_int_chg}
\bx_t = \frac{\nu_t}{\nu_s} \bx_s 
- 2\nu_t \int_s^t \left( \frac{\diff  \xi_\tau}{\diff  \tau} \right) \frac{\sigma_\tau}{\nu_\tau} \bepsilon^\btheta_\tau(\bx_\tau)  \diff\tau 
+ \sqrt{2}\nu_t\int_s^t \frac{\sigma_\tau}{\nu_\tau}    \sqrt{- \frac{\diff \xi_\tau}{\diff \tau}} \diff\overline{\rvb}_\tau.
\end{equation}
\end{mybox}
Since the signal-to-noise ratio defined in~\eqref{equation:logsnr} is monotonic, we can establish a bijection between the timestep $\tau$ and the signal-to-noise ratio $\xi_\tau$. 
We also analyze the diffusion SDEs with respect to the log-SNR $\xi$.
Let $\diff \rvb_\xi \triangleq \sqrt{-\frac{\diff t}{\diff \xi}}\diff \overline{\rvb}_{t_\xi}$ denote the corresponding Wiener process with respect to the log-SNR $\xi$. 
Additionally, let $\widehatbepsilon_\btheta(\widehatbx_\xi, \xi) \triangleq  \bepsilon^\btheta_{t_\xi}(\bx_{t_\xi})$ represent the change-of-variable form of $\bepsilon^\btheta$ with respect to $\xi$---that is, the model parameterized by the log-SNR (where  $t_\xi$ denotes the timestep corresponding to $\xi$).
For simplicity, we define $\widehatbx_\xi \triangleq \bx_{t_\xi}$.
When $\tau=s$, $\xi=\xi_s$; when $\tau=t$, $\xi=\xi_t$.
Changing the integration variable from timestep to log-SNR then gives:
\begin{mybox}
\begin{subequations}\label{equation:sde_int_eps_ALL}
\begin{align}
\bx_t 
&= {\frac{\nu_t}{\nu_s} \bx_s}
- 2\nu_t \int_{\xi_s}^{\xi_t} e^{-\xi} \widehatbepsilon_\btheta(\widehatbx_\xi, \xi) \diff \xi
+ \sqrt{2}\nu_t \int_{\xi_s}^{\xi_t} e^{-\xi} \diff \rvb_\xi
\label{equation:sde_int_eps}\\
&= \frac{\sigma_t}{\sigma_s}e^{-(\xi_t - \xi_s)}\bx_s + 2\nu_t \int_{\xi_s}^{\xi_t} e^{-2(\xi_t - \xi)} \widehatbx_\btheta(\widehatbx_\xi, \xi)\diff \xi + \sqrt{2}\sigma_t \int_{\xi_s}^{\xi_t} e^{-(\xi_t - \xi)} \diff \rvb_\xi,
\label{equation:sde_int_eps2}
\end{align}
\end{subequations}
\end{mybox}
where the second equality follows from the relation between data and noise prediction models given in \eqref{equation:dpmsolver_data_pred} (see Problem~\ref{prob:logsnrdiff_sol}).
The It\^o-integrals can be evaluated as
\begin{subequations}
\begin{align}
\int_{\xi_s}^{\xi_t} e^{-\xi} \diff \rvb_\xi &= \left(\sqrt{\int_{\xi_s}^{\xi_t} e^{-2\xi}\diff \xi}\right) \bepsilon_s = \frac{e^{-\xi_t}}{\sqrt{2}} \sqrt{e^{2(\xi_t - \xi_s)} - 1} \bepsilon_s,
\\
\int_{\xi_s}^{\xi_t} e^{\xi} \diff \rvb_\xi &= \left(\sqrt{\int_{\xi_s}^{\xi_t} e^{2\xi}\diff \xi}\right) \bepsilon_s = \frac{e^{\xi_t}}{\sqrt{2}} \sqrt{1 - e^{-2(\xi_t - \xi_s)}} \bepsilon_s,
\end{align}
\end{subequations}
where $\bepsilon_s \sim \normal(\bzero, \bI)$.
We only need to compute the integral term $\int_{\xi_s}^{\xi_t} e^{-\xi} \widehatbepsilon_\btheta(\widehatbx_\xi, \xi) \diff \xi$ or $\int_{\xi_s}^{\xi_t} e^{-2(\xi_t - \xi)} \widehatbx_\btheta(\widehatbx_\xi, \xi)\diff \xi$ (referred to as the \textit{exponentially weighted integral}) using numerical methods.
Further details are provided in the following sections.

\index{Semi-linear structure}
\paragrapharrow{Diffusion ODE solution.}
Compared with the diffusion SDE~\eqref{equation:denois_diff_sde}, the diffusion ODE~\eqref{equation:diff_denoid_ode} is better suited for accelerating sampling using large step sizes, as it contains no stochastic components. We therefore focus on the denoising diffusion ODE~\eqref{equation:diff_denoid_ode}.
Solving this ODE from time T to 0 corresponds to the generative (denoising) process. Specifically, given $\bx_s$, integration over the interval $[s, t]$ gives:
\begin{equation}
\bx_t = \bx_s + \int_s^t \left( f(\tau)\bx_\tau + \frac{g^2(\tau)}{2\sigma_\tau} \bepsilon^\btheta_\tau(\bx_\tau) \right) \diff\tau.
\end{equation}
Researchers have applied various black-box numerical ODE solvers to this differential equation, which amounts to approximating the above integral using different schemes. For instance, the Euler method approximates the integral using rectangular rules, while the improved Euler method uses trapezoidal approximations (see Section~\ref{section:flomat_flowmodel}).
However, such methods yield unsatisfactory performance when the number of discretization steps is small.
We will exploit the special structure of the diffusion ODE and investigate improved numerical integration strategies.

Note that \eqref{equation:diff_denoid_ode} can be decomposed into two parts: $f(t)\rvx_t$ is the linear term in $\rvx_t$, and $\frac{g^2(t)}{2\sigma_t}\bepsilon^\btheta_t(\rvx_t)$ is the nonlinear term, whose nonlinearity arises from the neural network. This gives the diffusion ODE a \textit{semi-linear structure}.
The benefit of this semi-linear form is that part of the solution can be derived analytically and thus does not require numerical approximation.

The exact solution to ODE~\eqref{equation:diff_denoid_ode} is derived as follows (see Problem~\ref{prob:ode_sol_ftut}):
\begin{mybox}
\begin{equation}\label{equation:dpm_semilineare_gen}
\bx_t = \bPsi(t, s)\bx_s + \int_{s}^{t} \bPsi(t, \tau)\left[ \frac{g^2(\tau)}{2\sigma_\tau}  \bepsilon^\btheta_\tau(\bx_\tau) \right] \diff \tau,
\end{equation}
\end{mybox}
where $\bPsi(t, s)$---satisfying $\frac{\partial}{\partial t}\bPsi(t, s) = \bF_t \bPsi(t, s)$, $\bPsi(s, s) = \bI$---is known as the \textit{transition matrix} from time $s$ to $t$. 
We derive the explicit form of this transition matrix in the following theorem.

\begin{theoremHigh}[Semi-linear solution formula for diffusion ODE]\label{theorem:dpm_semilineare}
The exact solution formula for the ODE~\eqref{equation:diff_denoid_ode} is given by:
\begin{equation}\label{equation:dpm_int}
\bx_t 
= \underbrace{e^{\int_s^t f(\tau)\diff\tau} \bx_s}_{\text{Deterministic linear flow}} 
+ \underbrace{\int_s^t \left( e^{\int_\tau^t f(m)\diff m} \cdot \frac{g^2(\tau)}{2\sigma_\tau} \bepsilon^\btheta_\tau(\bx_\tau) \right) \diff\tau}_{\text{Nonlinear score-matching term}} .
\end{equation}
The first term in the expression can be evaluated analytically in closed form.
\end{theoremHigh}
\begin{proof}[of Theorem~\ref{theorem:dpm_semilineare}]
We rearrange equation~\eqref{equation:diff_denoid_ode} by moving the linear term to the left-hand side and multiplying both sides by an integrating factor $\psi(t)$:
$$
\psi(t)\diff \rvx_t - \psi(t)f(t)\rvx_t \diff t  = \frac{\psi(t)g^2(t)}{2\sigma_t} \bepsilon^\btheta_t(\rvx_t) \diff t .
$$
We choose the integrating factor such that the left-hand side becomes an exact differential:
$$
\diff \left( \psi(t) \cdot \rvx_t \right) = \frac{\psi(t)g^2(t)}{2\sigma_t} \bepsilon^\btheta_t(\rvx_t) \diff t .
$$
Integrate both sides from $s$ to $t$ yields
$\psi(t)\rvx_t - \psi(s)\rvx_s = \int_s^t \left( \frac{\psi(\tau)g^2(\tau)}{2\sigma_\tau} \bepsilon^\btheta_{\tau}(\rvx_\tau) \right) \diff\tau$.
Solving for $\rvx_t$:
\begin{equation}\label{equation:dpm_semilineare1}
\rvx_t = \frac{\psi(s)}{\psi(t)} \rvx_s + \int_s^t \left( \frac{\psi(\tau)}{\psi(t)} \frac{g^2(\tau)}{2\sigma_\tau} \bepsilon^\btheta_{\tau}(\rvx_\tau) \right) \diff\tau .
\end{equation}

It remains to determine the integrating factor $\psi(t)$. By construction, it must satisfy:
$$
\diff\left( \psi(t) \cdot \rvx_t \right) = \rvx_t \diff \psi(t) + \psi(t)\diff \rvx_t = \psi(t)\diff \rvx_t - \psi(t)f(t)\rvx_t \diff t ,
$$
which implies:
$$
\diff \psi(t) = -\psi(t)f(t)\diff t 
\qquad \implies\qquad 
\frac{\diff \psi(t)}{\psi(t)} = -f(t)\diff t .
$$
This is a first-order linear ODE.
Integrating both sides from $0$ to $t$:
$$
\ln \psi(t) - \ln \psi(0) = -\int_0^t f(\tau)\diff\tau.
$$
Thus:
$$
\psi(t) = \psi(0) e^{-\int_0^t f(\tau)\diff\tau}
\qquad \implies\qquad 
\frac{\psi(s)}{\psi(t)} = \frac{\psi(0) e^{-\int_0^s f(\tau)\diff\tau}}{\psi(0) e^{-\int_0^t f(\tau)\diff\tau}} = e^{\int_s^t f(\tau)\diff\tau}.
$$
Substituting this into \eqref{equation:dpm_semilineare1} yields the desired result.
\end{proof}

Substituting the relation from Theorem~\ref{theorem:dpmforward}~\eqref{equation:dpmforward} into \eqref{equation:dpm_int} yields:
\begin{mybox}
\begin{equation}\label{equation:dpm_int_chg}
\bx_t = \frac{\nu_t}{\nu_s} \bx_s - \nu_t \int_s^t \left( \frac{\diff  \xi_\tau}{\diff  \tau} \right) \frac{\sigma_\tau}{\nu_\tau} \bepsilon^\btheta_\tau(\bx_\tau)  \diff\tau .
\end{equation}
\end{mybox}
As expected, the resulting expression is significantly more compact.
Once again, since the signal-to-noise ratio~\eqref{equation:logsnr} is monotonic, we can establish a bijection between the timestep $\tau$ and the signal-to-noise ratio $\xi_\tau$. 
Denote $\widehatbepsilon_\btheta(\widehatbx_\xi, \xi) \triangleq  \bepsilon^\btheta_{t_\xi}(\bx_{t_\xi})$ as the change-of-variable form of $\bepsilon^\btheta$ for $\xi$, i.e., the model parameterized by the log-SNR (where timestep $t_\xi$ corresponds to $\xi$).
Changing the integration variable from timestep to log-SNR then produces:
\begin{mybox}
\begin{subequations}\label{equation:dpm_int_eps_ALL}
\begin{align}
\bx_t 
&= {\frac{\nu_t}{\nu_s} \bx_s} - \nu_t {\int_{\xi_s}^{\xi_t} e^{-\xi} \widehatbepsilon_\btheta(\widehatbx_\xi, \xi) \diff \xi}
\label{equation:dpm_int_eps}\\
&= \frac{\sigma_t}{\sigma_s}e^{-(\xi_t - \xi_s)}\bx_s + \nu_t \int_{\xi_s}^{\xi_t} e^{-2(\xi_t - \xi)} \widehatbx_\btheta(\widehatbx_\xi, \xi)\diff \xi .
\label{equation:dpm_int_eps2}
\end{align}
\end{subequations}
\end{mybox}
Once again, only the integral terms $\int_{\xi_s}^{\xi_t} e^{-\xi} \widehatbepsilon_\btheta(\widehatbx_\xi, \xi) \diff \xi$ or $\int_{\xi_s}^{\xi_t} e^{-2(\xi_t - \xi)} \widehatbx_\btheta(\widehatbx_\xi, \xi)\diff \xi$ require numerical evaluation.

\index{DEIS}
\index{Diffusion exponential integrator sampler}
\index{EI}
\index{Exponential integrator}
\section{DEIS}\label{section:deis}

We now analyze the discretization error arising from solving the probability flow ODE \eqref{equation:diff_denoid_ode}, restated here for convenience:
\begin{equation}\label{equation:diffode_deis}
\frac{\diff \rvx_t}{\diff t} = f(t) \rvx_t - \frac{1}{2} g^2(t) \bs^\btheta_t(\rvx_t).
\end{equation}
The exact solution to this ODE is established in Theorem~\ref{theorem:dpm_semilineare}, and can also be derived in the form \eqref{equation:dpm_semilineare_gen} (see Problem~\ref{prob:ode_sol_ftut}):
\begin{equation}\label{equation:diffode_deis_sol}
\bx_t = \bPsi(t, s)\bx_s + \int_{s}^{t} \bPsi(t, \tau)\left(-\frac{1}{2} g^2(\tau) \bs^\btheta_\tau(\bx_\tau)\right) \diff \tau,
\end{equation}
where $\bPsi(t, s)$ is the transition matrix from time $s$ to $t$ corresponding to $f(t)$, satisfying $\frac{\partial}{\partial t}\bPsi(t, s) = f(t) \bPsi(t, s)$, $\bPsi(s, s) = \bI$.
A variety of numerical ODE solvers exist for \eqref{equation:diffode_deis}, each employing a distinct discretization strategy to approximate the integral solution \eqref{equation:diffode_deis_sol} \citep{griffiths2010numerical, butcher2016numerical}. As the discretization step size tends to zero, all such methods converge to the true solution of \eqref{equation:diffode_deis}. Nevertheless, their performance can differ drastically when using finite, large step sizes. Meanwhile, rapid sampling using \eqref{equation:diffode_deis} requires approximating its solution with only a small number of discretization steps, which inherently implies large step sizes. This provides the motivation for designing an efficient discretization scheme tailored specifically to the structure of ODE \eqref{equation:diffode_deis}.

\subsection{Exponential Integrator over Euler Method}
The Euler method represents the most basic explicit numerical scheme for ordinary differential equations (see Section~\ref{section:flomat_flowmodel}).
When applied to ODE \eqref{equation:diffode_deis}, it takes the form
\begin{equation}\label{equation:ftgt_euler}
\bx_{t}= \bx_s - \left[ f(s) \bx_s - \frac{1}{2} g^2(s) \bs^\btheta_s(\bx_s) \right] \Delta t, 
\quad 
\Delta t \triangleq s-t>0.
\end{equation}
This method, however, suffers from low accuracy and can even become unstable unless the step size is chosen sufficiently small.
From the exact solution \eqref{equation:diffode_deis_sol}, given $\bx_s$, the true value of $\bx_{t}$ satisfies 
\begin{equation}
\bx_{t} 
= \bPsi(t, s)\bx_s +\int_{s}^{t} \bPsi(t, \tau)\left(-\frac{1}{2} g^2(\tau) \bs^\btheta_\tau(\bx_\tau)\right) \diff \tau.
\end{equation} 
Yet the second term in this expression cannot be evaluated analytically.
To approximate this integral term, \citet{zhang2022fast} introduced the \textit{exponential integrator (EI)}, which exploits the semi-linear structure of ODE \eqref{equation:diffode_deis} by approximating the nonlinear term as constant:  $\bs^\btheta_\tau(\bx_\tau) \approx \bs^\btheta_s(\bx_s)$:
\begin{mybox}
\begin{equation}\label{equation:ei_sol}
\text{(EI)}
\qquad 
\bx_{t}= \bPsi(t, s)\bx_s + \left[ \int_{s}^{t} -\frac{1}{2}\bPsi(t, \tau)g^2(\tau) \diff \tau \right] \bs^\btheta_s(\bx_s).
\end{equation}
\end{mybox}
This scheme performs well when the nonlinear contribution $\bs^\btheta_s(\bx_s)$ varies slowly along the trajectory.
Indeed, for any fixed time step $\Delta t=s-t$, the exponential integrator \eqref{equation:ei_sol} yields the exact solution to ODE \eqref{equation:diffode_deis} whenever $\bs^\btheta_s(\bx_s)$ remains constant over the interval $[t, s]$.

\subsection{Improvement using Noise Prediction}
Nevertheless, \citet{zhang2022fast} demonstrate that the EI scheme~\eqref{equation:ei_sol} performs inferiorly to the Euler method~\eqref{equation:ftgt_euler}.
This behavior stems from the fact that the score function of the real data distribution $\nabla \ln p_t(\bx_t)$ evolves rapidly as $t \to 0$ \citep{dockhorn2021score}.
The difficulties induced by this rapidly varying score $\nabla \ln p_t(\bx_t)$ are not limited to sampling alone, but also arise during the training of diffusion models. 
To mitigate these issues, an alternative parameterization of the score network is commonly employed. Specifically, the parameterization $\nabla \ln p_t(\bx_t) \approx -\frac{1}{\sigma_t} \bepsilon^\btheta_t(\bx_t)$
(see \eqref{equation:score_noise_sampler}) has been shown to yield substantial accuracy gains.
The motivation for this parameterization comes from reformulating the conditional score matching loss \eqref{equation:score_lss_sampler} into the DDPM loss \eqref{equation:sampler_ddpmloss}:
\begin{equation}
\mathcalJ(\btheta) = \Exp_{t \sim \uniformdist[0,T], p_{\text{data}}(\bx_0), \bepsilon_t \sim \normal(\bzero, \bI_D)} 
\left[ \zeta_t\normtwo{\bepsilon^\btheta_t(\bx_t) - \bepsilon_t}^2 \right].
\end{equation}
The network $\bepsilon^\btheta_t$ is trained to match the noise variable $\bepsilon_t$, which is drawn from a standard Gaussian distribution and therefore has bounded magnitude. 
In contrast, the direct score parameterization $\bs^\btheta_t = -\frac{1}{\sigma_t} \bepsilon^\btheta_t$ can take extremely large values as $\sigma_t \to 0$ near $t=0$. It is thus more favorable to approximate $\bepsilon^\btheta_t$ rather than $\bs^\btheta_t$ using a neural network.

We adopt this parameterization and rewrite the ODE \eqref{equation:diffode_deis} as
\begin{equation}\label{equation:diffode_deis_epsilon}
\frac{\diff \rvx_t}{\diff t} = f(t) \rvx_t + \frac{1}{2\sigma_t} g^2(t)  \bepsilon^\btheta_t(\rvx_t).
\end{equation}
Applying the exponential integrator to this reformulated ODE yields the following update rule:
\begin{mybox}
\begin{equation}\label{equation:ei_sol_epsilon}
\text{(EI$'$)}
\qquad 
\bx_{t} 
= \bPsi(t, s)\bx_s + \left[ \int_{s}^{t} \frac{1}{2\sigma_\tau}\bPsi(t, \tau)g^2(\tau)  \diff \tau \right] \bepsilon^\btheta_s(\bx_s).
\end{equation}
\end{mybox}
In contrast to the original EI scheme~\eqref{equation:ei_sol}, the modified update~\eqref{equation:ei_sol_epsilon} uses $-\frac{1}{\sigma_\tau} \bepsilon^\btheta_s(\bx_s)$ rather than $\bs^\btheta_s(\bx_s) = -\frac{1}{\sigma_s} \bepsilon^\btheta_s(\bx_s)$ to approximate the score $\bs^\btheta_\tau(\bx_\tau)$ over the time interval $\tau \in [t, s]$. 
This adjustment  from $\frac{1}{\sigma_s}$ to $\frac{1}{\sigma_\tau}$ proves critical, as the coefficient $\frac{1}{\sigma_\tau}$ \textbf{varies sharply} over the interval $\tau \in [t, s]$. 
With this refinement, the exponential integrator substantially outperforms the Euler method \citep{zhang2022fast}.

Notably, the EI-based discretization~\eqref{equation:ei_sol_epsilon} reduces to the widely used deterministic DDIM sampler when the forward diffusion SDE~\eqref{equation:nois_diff_sde} follows the variance preserving  formulation.
\begin{proposition}[Relationship between DEIS and DDIM]\label{proposition:deis_ddim}
When the forward diffusion SDE {\eqref{equation:nois_diff_sde}} is variance-preserving (with $f(t), g(t)$ defined as in \eqref{equation:dpmforward} and satisfying $\nu_t^2+\sigma_t^2=1$), the EI discretization \eqref{equation:ei_sol_epsilon} simplifies to
\begin{equation}
\bx_{t} 
= {\frac{\nu_{t}}{\nu_s}} \bx_s + \left[ \sigma_{t} - {\frac{\nu_{t}}{\nu_s}} \sigma_s \right] \bepsilon^\btheta_s(\bx_s).
\end{equation}
This update exactly matches the noise-free DDIM update rule~\eqref{equation:ddim_samp}  (with no noise).
\end{proposition}
\begin{proof}[of Proposition~\ref{proposition:deis_ddim}]
From \eqref{equation:dpm_int_chg}, the transition matrix for the variance-preserving case ($\nu_1^2+\sigma_t^2=1$) is given by
$\bPsi(t, s) = {\frac{\nu_t}{\nu_s}}\bI$.
Substituting this into the integral term gives
\begin{align*}
\int_s^t \bPsi(t, \tau) \frac{1}{2} g^2(\tau) \bL_\tau^{-1} \diff \tau
&= \int_s^t -\frac{1}{2} {\frac{\nu_t}{\nu_\tau}} \frac{\diff \ln \nu_\tau^2}{\diff \tau} \frac{1}{\sigma_\tau} \diff \tau 
=-{\nu_t} \int_{\nu_s}^{\nu_t} \left( \frac{1}{\nu_\tau^2 \sqrt{1-\nu_\tau^2}} \right) {\diff\nu_\tau} \\
&= -\nu_t \left.(-\frac{\sqrt{1-\nu^2}}{\nu})\right|_{\nu_s}^{\nu_t}
= \sigma_t- {\frac{\nu_t}{\nu_s}} \sigma_s,
\end{align*}
which completes the proof.
\end{proof}

This result offers an alternative explanation for the effectiveness of DDIM in the context of variance-preserving SDEs, grounded in numerical ODE discretization theory.

\begin{SCfigure}
\centering
\includegraphics[width=0.45\textwidth]{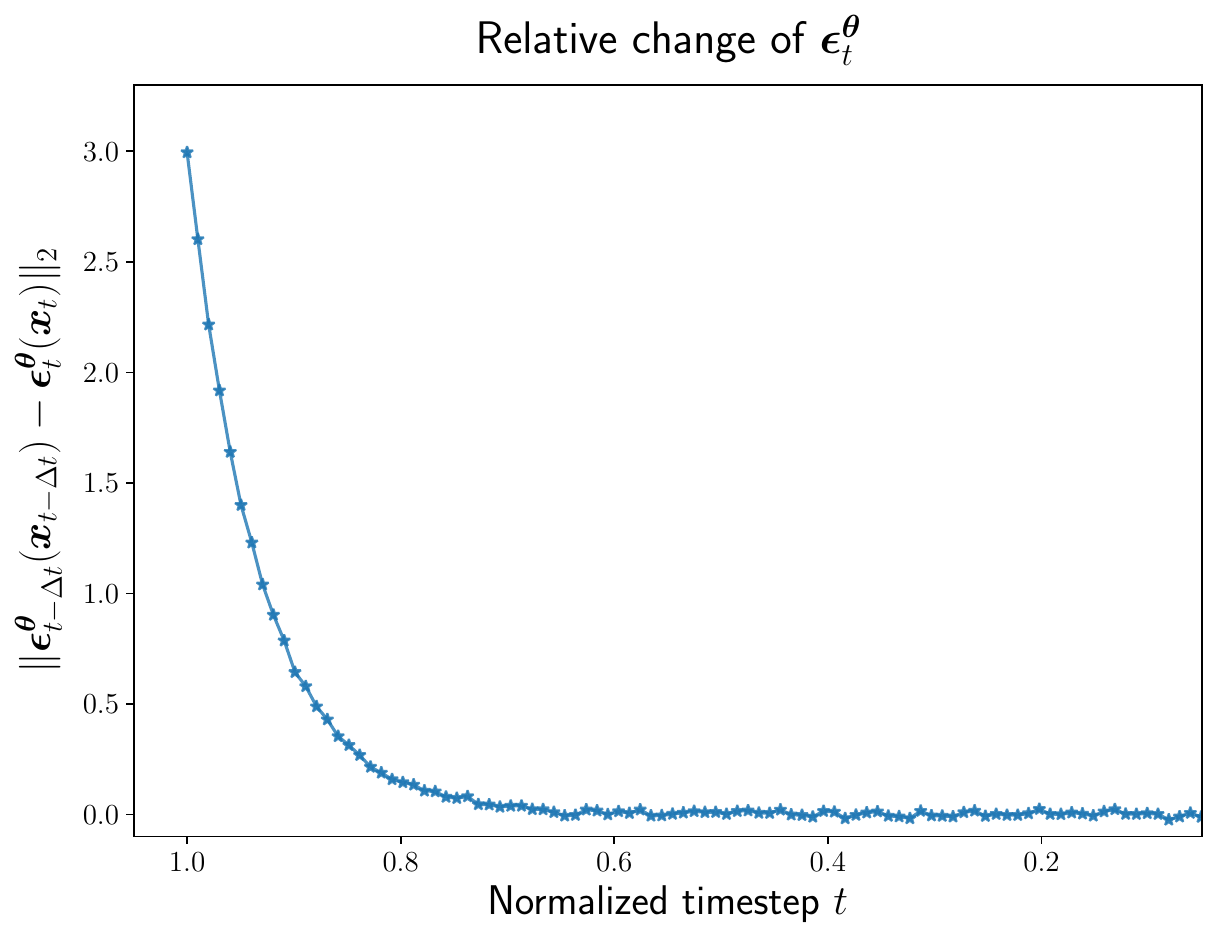}
\caption{Relative changes of $\bepsilon^\btheta_t(\bx_t)$ with respect to normalized timestep $t$ are relatively small, especially when $t <0.7$.}
\label{fig:epsilon_relchg}
\end{SCfigure}
\subsection{Polynomial Extrapolation}
In the discretization~\eqref{equation:ei_sol_epsilon}, we approximate $\bepsilon^\btheta_\tau(\bx_\tau)$ by $\bepsilon^\btheta_s(\bx_s)$ for all $\tau \in [t, s]$, which constitutes a zeroth-order approximation.
Comparing~\eqref{equation:ei_sol_epsilon} with the exact integral solution~\eqref{equation:diffode_deis_sol}, it is clear that this approximation error dominates the overall discretization accuracy.

To address this question, we analyze how $\bepsilon^\btheta_t(\bx_t)$ volves along a ground-truth trajectory $\{\bx_t\}$ from $t = T$ to $t = 0$. We plot the relative change in $\ell_2$-norm change of $\bepsilon^\btheta_t(\bx_t)$ in Figure~\ref{fig:epsilon_relchg}, which shows that the relative variation remains small across most timesteps. 
This observation motivates us to use backward history values of $\bepsilon^\btheta$ up to timestep $s$ to extrapolate $\bepsilon^\btheta_\tau(\bx_\tau)$ for $\tau \in [t, s]$.

We incorporate high-order polynomial extrapolation of $\bepsilon^\btheta$ into the exponential integrator framework. 
To this end, we introduce a time discretization $\{t_i\}_{i=0}^M$ with $t_0 = T$ and $t_M = 0$. 
At each  step $i$ (i.e., timestep $t_i$), we fit a degree-$r$ polynomial $P_r(t)$  using the interpolation points $(t_{i-j}, \bepsilon^\btheta_{t_{i-j}}(\bx_{t_{i-j}})), 0 \leq j \leq r$. 
This polynomial admits the explicit form \citep{zhang2022fast}:
\begin{equation}\label{equation:deis_extrapo}
P_r(\tau) 
= \sum_{j=0}^r \left[ \left(\prod_{\substack{k=0 \\ k \neq j}}^r \frac{\tau - t_{i-k}}{t_{i-j} - t_{i-k}}\right) \bepsilon^\btheta_{t_{i-j}}(\bx_{t_{i-j}})\right], 
\quad \tau \in [t_{i+1}, t_i].
\end{equation}
The core idea is to use the recent timesteps $t_i, t_{i-1}, \ldots, t_{i-r}$  to approximate $\bepsilon^\btheta_\tau(\bx_\tau)$ over the interval $\tau\in[t_{i+1}, t_i]$. 
For $i > M - r$, we naturally switch to lower-degree polynomials for the approximation.

Analogous to the EI method~\eqref{equation:ei_sol_epsilon}, which uses a zeroth-order score approximation in~\eqref{equation:diffode_deis_sol}, the $r$-th-order update rule is derived by substituting the polynomial approximation~\eqref{equation:deis_extrapo} into the exact solution~\eqref{equation:diffode_deis_sol}. The resulting scheme reads
\begin{mybox}
\begin{align}
\text{(DEIS)}
	\qquad 
\bx_{t_{i+1}} 
&= \bPsi(t_{i+1}, t_i)\bx_{t_i} + \sum_{j=0}^r \left[ C_{ij} \bepsilon^\btheta_{t_{i-j}}(\bx_{t_{i-j}}) \right]; 
\label{equation:tabdeis1}\\
C_{ij} 
&= \int_{t_i}^{t_{i+1}} \frac{1}{2\sigma_\tau} \bPsi(t_{i+1}, \tau) g^2(\tau)  \prod_{\substack{k=0 \\ k \neq j}}^r \left[ \frac{\tau - t_{i-k}}{t_{i-j} - t_{i-k}} \right] \diff \tau.
\label{equation:tabdeis2}
\end{align}
\end{mybox}
We note that the update~\eqref{equation:tabdeis1} is a linear combination of  $\bx_{t_i}$ and and past network outputs $\bepsilon^\btheta_{t_{i-j}}(\bx_{t_{i-j}})$. 
The weights $\bPsi(t_{i+1}, t_i)$ and $C_{ij}$ depend only on the forward noising SDE~\eqref{equation:nois_diff_sde} and the chosen time discretization; they can be precomputed once and reused across all batches.
When analytical formulas are unavailable, these coefficients can be accurately computed via numerical integration. In typical DDPM settings, the integral~\eqref{equation:tabdeis2} is at most two-dimensional and thus straightforward to evaluate numerically. 
This approach closely mirrors the classical \textit{Adams--Bashforth (AB) method} \citep{hochbruck2010exponential}, and the resulting sampler is referred to as \textit{$t$AB-DEIS} (diffusion exponential integrator sampler) \citep{zhang2022fast}.

\index{{DPM-Solver}}
\section{DPM-Solver}\label{section:dpmsolver}

Diffusion models generate high-fidelity samples by solving the reverse-time denoising ODE (see \eqref{equation:dpm_int_chg}--\eqref{equation:dpm_int_eps_ALL}). However, naive numerical integration schemes such as Euler's method typically require hundreds of steps to produce coherent samples, which restricts their practical use in latency-sensitive applications. To overcome this limitation, DPM-Solver (diffusion probabilistic model solver) was introduced as a family of advanced ODE solvers specifically designed for the structure of diffusion dynamics. It enables high-quality sampling with significantly fewer steps (often 10--20 steps or fewer).
This section systematically presents the core design principles and technical innovations of DPM-Solver.

\subsection{Analytically Computable Coefficients}\label{section:dpmsolverana}
We start by exploiting the distinctive analytical structure of the diffusion ODE to derive analytically computable coefficients. These remove the need for costly numerical quadrature operations and form the basis of efficient single-step solvers.

While the exponentially weighted integral in the ODE solution \eqref{equation:dpm_int_eps} can be directly approximated numerically, following the principle of maximizing analytical computation wherever possible, \citet{lu2022dpm} further refine the integral term.
Once again, we consider a decreasing sequence $\{t_i\}_{i=0}^M$  from $t_0 = T$ to
$t_M = 0$.
Given the previous state ${\bx}_{t_{i-1}}$ at time $t_{i-1}$, the goal of our solver is to approximate the exact solution at time $t_i$.
At timestep $t_{i-1}$, we take the $(k-1)$-th order Taylor expansion of $\widehatbepsilon_\btheta(\widehatbx_\xi, \xi)$ around  $\xi_{t_{i-1}}$ 
for  $\xi \in [\xi_{t_{i-1}}, \xi_{t_i}]$:
$$
\widehatbepsilon_\btheta(\widehatbx_\xi, \xi) = \sum_{n=0}^{k-1} \frac{(\xi - \xi_{t_{i-1}})^n}{n!} \widehatbepsilon_\btheta^{(n)}(\widehatbx_{\xi_{t_{i-1}}}, \xi_{t_{i-1}}) + \mathcalO\left((\xi - \xi_{t_{i-1}})^k\right),
$$
where $\widehatbepsilon_\btheta^{(n)}(\widehatbx_\xi, \xi) \triangleq   \frac{\diff ^n \widehatbepsilon_\btheta(\widehatbx_\xi, \xi)}{\diff \xi^n}$ denotes the $n$-th order derivative of $\widehatbepsilon_\btheta$ w.r.t. the log-SNR $\xi$ (see definition in \eqref{equation:logsnr}).
Substituting this Taylor expansion into \eqref{equation:dpm_int_eps} with $s = t_{i-1}$ and $t = t_i$ yields:
\begin{mybox}
\begin{equation}\label{equation:dpm_int_tay}
\bx_{t_i} = \frac{\nu_{t_i}}{\nu_{t_{i-1}}} \bx_{t_{i-1}} - \nu_{t_i} \sum_{n=0}^{k-1} \underbrace{\widehatbepsilon_\btheta^{(n)}(\widehatbx_{\xi_{t_{i-1}}}, \xi_{t_{i-1}})}_{\text{Derivatives}} \underbrace{\int_{\xi_{t_{i-1}}}^{\xi_{t_i}} e^{-\xi} \frac{(\xi - \xi_{t_{i-1}})^n}{n!} \diff \xi}_{\triangleq C_n} + \mathcalO\left(h_i^{k+1}\right) ,
\end{equation}
\end{mybox}
where $h_i \triangleq \xi_{t_i} - \xi_{t_{i-1}}$. 
The coefficients $C_n$  in the above expression can be computed analytically. Specifically, using integration by parts:
\begin{align}
C_n &= \int_{\xi_{t_{i-1}}}^{\xi_{t_i}} e^{-\xi} \frac{(\xi - \xi_{t_{i-1}})^n}{n!} \diff \xi 
= -\int_{\xi_{t_{i-1}}}^{\xi_{t_i}} \frac{(\xi - \xi_{t_{i-1}})^n}{n!} \diff e^{-\xi} \nonumber\\
&= \left.\left( -\frac{(\xi - \xi_{t_{i-1}})^n}{n!} e^{-\xi} \right) \right|_{\xi_{t_{i-1}}}^{\xi_{t_i}} + \int_{\xi_{t_{i-1}}}^{\xi_{t_i}} e^{-\xi} \frac{(\xi - \xi_{t_{i-1}})^{n-1}}{(n-1)!} \diff \xi 
= -\frac{h_i^n}{n!} e^{-\xi_{t_i}} + C_{n-1}. \label{equation:dpmsolverc_n}
\end{align}
The coefficients follow a simple recurrence relation, and the base term $C_0$ is given by:
$$
C_0 = \int_{\xi_{t_{i-1}}}^{\xi_{t_i}} e^{-\xi} \diff \xi = e^{-\xi_{t_{i-1}}} - e^{-\xi_{t_i}} = \frac{\sigma_{t_i}}{\nu_{t_i}} \left( e^{h_i} - 1 \right).
$$
From this, we can compute $C_1$ and $C_2$:
\begin{align*}
C_1 &= e^{-\xi_{t_{i-1}}} - (1 + h_i)e^{-\xi_{t_i}} = \frac{\sigma_{t_i}}{\nu_{t_i}} \left( e^{h_i} - 1 - h_i \right);\\
C_2 &= e^{-\xi_{t_{i-1}}} - \left( 1 + h_i + \frac{h_i^2}{2} \right) e^{-\xi_{t_i}} = \frac{\sigma_{t_i}}{\nu_{t_i}} \left( e^{h_i} - 1 - h_i - \frac{h_i^2}{2} \right).
\end{align*}

In practical applications, we only consider $k = 1, 2, 3$, so higher-order coefficients  $C_n$ are unnecessary. 
Setting  $k = 1, 2, 3$ yields the corresponding solvers, referred to as \textit{DPM-Solver-1}, \textit{DPM-Solver-2}, and \textit{DPM-Solver-3}, which are 1st-, 2nd-, and 3rd-order ODE solvers, respectively.

For example, when $k = 1$, substituting $C_0$ into \eqref{equation:dpm_int_tay} produces the update rule for DPM-Solver-1:
\begin{mybox}
\begin{align}
\bx_{t_i} &= \frac{\nu_{t_i}}{\nu_{t_{i-1}}} \bx_{t_{i-1}} - \nu_{t_i} \widehatbepsilon_\btheta\left(\widehatbx_{\xi_{t_{i-1}}}, \xi_{t_{i-1}}\right) C_0 + \mathcalO\left(h_i^2\right) \nonumber\\
&= \frac{\nu_{t_i}}{\nu_{t_{i-1}}} \bx_{t_{i-1}} - \sigma_{t_i} \left( e^{h_i} - 1 \right) \widehatbepsilon_\btheta\left(\widehatbx_{\xi_{t_{i-1}}}, \xi_{t_{i-1}}\right) + \mathcalO\left(h_i^2\right) \nonumber\\
&= \frac{\nu_{t_i}}{\nu_{t_{i-1}}} \bx_{t_{i-1}} - \sigma_{t_i} \left( e^{h_i} - 1 \right) \bepsilon^\btheta_{t_{i-1}}\left(\bx_{t_{i-1}}\right) + \mathcalO\left(h_i^2\right)\nonumber\\
&= \frac{\nu_{t_i}}{\nu_{t_{i-1}}} \bx_{t_{i-1}} - \nu_{t_i} \left( \frac{\sigma_{t_{i-1}}}{\nu_{t_{i-1}}} - \frac{\sigma_{t_{i}}}{\nu_{t_{i}}} \right) + \mathcalO\left(h_i^2\right).
\label{equation:dpmsolverk1}
\end{align}
\end{mybox}
One can observe that this update formula matches exactly that of DDIM~\eqref{equation:ddim_samp} under the parameterization $\nu_t=\sqrt{\alpha_t}$ and $\sigma_t=\sqrt{1-\alpha_t}$ (without noise), implying that DDIM is equivalent to a first-order DPM-Solver. 
This result provides an alternative explanation for the effectiveness of DDIM in the context of variance-preserving SDEs, (cf. Proposition~\ref{proposition:deis_ddim}).

\subsection{Higher-Order Approximation:  Estimating Derivative Terms}\label{section:dpmsolverhigh}
Building on the analytically computable coefficients derived earlier, we extend the framework to higher-order approximations, demonstrating how to estimate derivative terms from past function evaluations to achieve faster convergence and improved sample quality.
When computing higher-order DPM-Solvers, we must  calculate the higher-order derivative terms $\widehatbepsilon_\btheta^{(n)}(\widehatbx_\xi, \xi)$ in \eqref{equation:dpm_int_tay}. 
These terms can be estimated via numerical differentiation. For example, to approximate the first derivative, define an intermediate log-SNR value $\xi_{s_i} = \rho \xi_{t_i} + (1 - \rho)\xi_{t_{i-1}}$ with $\rho \in (0,1)$. 
The first derivative is then approximated as:
$$
\begin{aligned}
\widehatbepsilon_\btheta^{(1)}(\widehatbx_{\xi_{t_{i-1}}}, \xi_{t_{i-1}}) 
&\approx \frac{\widehatbepsilon_\btheta(\widehatbx_{\xi_{s_i}}, \xi_{s_i}) - \widehatbepsilon_\btheta(\widehatbx_{\xi_{t_{i-1}}}, \xi_{t_{i-1}})}{\xi_{s_i} - \xi_{t_{i-1}}} 
= \frac{\widehatbepsilon_\btheta(\widehatbx_{\xi_{s_i}}, \xi_{s_i}) - \widehatbepsilon_\btheta(\widehatbx_{\xi_{t_{i-1}}}, \xi_{t_{i-1}})}{\rho h_i}.
\end{aligned}
$$
Using this approximation, we derive the update formula for DPM-Solver-2 by setting $k=2$ in \eqref{equation:dpm_int_tay}:
\begin{mybox}
{
\small
\begin{align}
\small
&\bx_{t_i} = \frac{\nu_{t_i}}{\nu_{t_{i-1}}} \bx_{t_{i-1}} - \nu_{t_i} \left( C_0 \cdot \widehatbepsilon_\btheta(\widehatbx_{\xi_{t_{i-1}}}, \xi_{t_{i-1}}) + C_1 \cdot \widehatbepsilon_\btheta^{(1)}(\widehatbx_{\xi_{t_{i-1}}}, \xi_{t_{i-1}}) \right) + \mathcalO(h_i^3) \nonumber\\
&\approx \frac{\nu_{t_i}}{\nu_{t_{i-1}}} \bx_{t_{i-1}} - \sigma_{t_i}(e^{h_i} - 1)\widehatbepsilon_\btheta(\widehatbx_{\xi_{t_{i-1}}}, \xi_{t_{i-1}}) - \sigma_{t_i}(e^{h_i} - 1)\frac{h_i}{2}\widehatbepsilon_\btheta^{(1)}(\widehatbx_{\xi_{t_{i-1}}}, \xi_{t_{i-1}})  \nonumber\\
&= \frac{\nu_{t_i}}{\nu_{t_{i-1}}} \bx_{t_{i-1}} - \sigma_{t_i}(e^{h_i} - 1)\widehatbepsilon_\btheta(\widehatbx_{\xi_{t_{i-1}}}, \xi_{t_{i-1}}) - \frac{\sigma_{t_i}}{2\rho}(e^{h_i} - 1)\left( \widehatbepsilon_\btheta(\widehatbx_{\xi_{s_i}}, \xi_{s_i}) - \widehatbepsilon_\btheta(\widehatbx_{\xi_{t_{i-1}}}, \xi_{t_{i-1}}) \right) \nonumber\\
&= \frac{\nu_{t_i}}{\nu_{t_{i-1}}} \bx_{t_{i-1}} - \sigma_{t_i}(e^{h_i} - 1)\bepsilon^\btheta_{t_{i-1}}(\bx_{t_{i-1}}) - \frac{\sigma_{t_i}}{2\rho}(e^{h_i} - 1)\left( \bepsilon^\btheta_{s_i}(\bx_{s_i}) - \bepsilon^\btheta_{t_{i-1}}(\bx_{t_{i-1}}) \right) ,
\label{equation:dpmsolver2}
\end{align}}
\end{mybox}
\noindent
where $s_i =t_{\xi} (\xi_{s_i})$ denotes the timestep corresponding to log-SNR $\xi_{s_i}$ ($t_{\xi}(\cdot)$ is the inverse function of $\xi(t)$, obtaining the log-SNR from timestep).
The second approximation holds because:
$$
(e^{h_i} - 1 - h_i) - (e^{h_i} - 1)\frac{h_i}{2} 
= \frac{2e^{h_i} - 2 - h_i - h_i e^{h_i}}{2} 
\stackrel{\text{Problem }\ref{prob:tay_order4}}{\longeq} \frac{-h_i^3/6 + \mathcalO(h_i^4)}{2} = \mathcalO(h_i^3).
$$
That is, we combine the previous state $\bx_{t_{i-1}}$ at time $t_{i-1}$ with an intermediate state $\bx_{s_i}$ at time $s_i$ (satisfying $t_{i-1}>s_i>t_i$) to compute the state ${\bx}_{t_i}$ at time $t_i$.
At each step, given $\bx_{t_{i-1}}$, the second-order DPM-Solver (DPM-Solver-2) performs the following updates:
\begin{align}
\bx_{s_i} &\leftarrow \frac{\nu_{s_i}}{\nu_{t_{i-1}}}\bx_{t_{i-1}} - \sigma_{s_i}\big(e^{\rho h_i} - 1\big)\bepsilon^\btheta_{t_{i-1}}(\bx_{t_{i-1}}); \qquad (\text{DPM-Solver-1}~\eqref{equation:dpmsolverk1}) \nonumber \\
\bx_{t_i} &\leftarrow \frac{\nu_{t_i}}{\nu_{t_{i-1}}}\bx_{t_{i-1}} 
- \sigma_{t_i}\big(e^{h_i} - 1\big)
\underbrace{\Big(\big(1- \frac{1}{2 \rho}\big)\bepsilon^\btheta_{t_{i-1}}(\bx_{t_{i-1}}) +\frac{1}{2 \rho}\bepsilon^\btheta_{s_i}(\bx_{s_i}) \Big)}_{\text{Convex combination}}.
\label{equation:dpmsolver2_update}
\end{align}
The complete pseudocode for DPM-Solver-2 is presented in Algorithm~\ref{alg:dpmsolver2}.
Note that while this first-order expansion improves accuracy, it comes at the cost of two forward passes through the neural network per step. We achieve higher accuracy at the expense of increased computation time per step. Although greater accuracy enables high-quality sampling with fewer total steps, each step requires roughly twice as much computation. This naturally raises a question: which approach is faster---sampling with $M$ steps using a zeroth-order expansion, or sampling with $M/2$ steps using a first-order expansion?
According to results reported in \citet{lu2022dpm}, DPM-Solver-2 outperforms traditional explicit Runge--Kutta (RK) ODE solvers (see Section~\ref{section:flomat_flowmodel}) under an equal \textit{number of function evaluations (NFE)}, defined as the total number of neural network forward passes.

As an alternative, we may explicitly introduce an additional intermediate timestep $s_i$ between $t_{i-1}$ and $t_i$, then approximate the derivative using function values at $s_i$ and $t_{i-1}$. This follows the standard formulation for single-step ODE solvers \cite{atkinson2009numerical}. In total, this requires $2M + 1$ timesteps, consisting of $\{t_i\}_{i=0}^M$ and $\{s_i\}_{i=1}^M$, such that $t_0 > s_1 > t_1 > \ldots > t_{M-1} > s_M > t_M$.
The modified procedure is described in Algorithm~\ref{alg:dpmsolver2prime}.

\begin{algorithm}[htbp]
\caption{DPM-Solver-2 \citep{lu2022dpm}}
\label{alg:dpmsolver2}
\begin{algorithmic}[1]
\Require initial value $\bx_T$, $M$ timesteps $\{t_i\}_{i=0}^M$, noise prediction model $\bepsilon^\btheta_t$, $0<\rho<1$ (e.g., $\rho=0.5$);
\State $\bx_{t_0} \leftarrow \bx_T\sim \normal(\bzero, \bI_D)$;
\For{$i \leftarrow 1$ to $M$}
\State $h_i \leftarrow \xi_{t_i} - \xi_{t_{i-1}}$;
\State $s_i \leftarrow t_\xi\left( \rho \xi_{t_i} + (1 - \rho)\xi_{t_{i-1}} \right)$; \Comment{Obtain the timestep $s_i$}
\State $\bx_{s_i} \leftarrow \frac{\nu_{s_i}}{\nu_{t_{i-1}}} \bx_{t_{i-1}} - \sigma_{s_i} \left( e^{\rho{h_i}} - 1 \right) \bepsilon^\btheta_{t_{i-1}}\left( \bx_{t_{i-1}}\right)$; \Comment{DPM-Solver-1 \eqref{equation:dpmsolverk1}}
\State $\small \bx_{t_i} \leftarrow \frac{\nu_{t_i}}{\nu_{t_{i-1}}}\bx_{t_{i-1}} 
- \sigma_{t_i}\big(e^{h_i} - 1\big)\Big(\big(1- \frac{1}{2 \rho}\big)\bepsilon^\btheta_{t_{i-1}}(\bx_{t_{i-1}}) +\frac{1}{2 \rho}\bepsilon^\btheta_{s_i}(\bx_{s_i}) \Big)$;
\EndFor
\State \Return $\bx_{t_M}$;
\end{algorithmic}
\end{algorithm}

\begin{algorithm}[H]
\caption{DPM-Solver-2$'$ \citep{lu2022dpm}}
\label{alg:dpmsolver2prime}
\begin{algorithmic}[1]
\Require initial value $\bx_T$, timesteps $\{t_i\}_{i=0}^M$ and $\{s_i\}_{i=1}^M$, noise prediction model $\bepsilon^\btheta_t$;
\State $\bx_{t_0} \leftarrow \bx_T\sim \normal(\bzero, \bI_D)$;
\For{$i \leftarrow 1$ to $M$}
\State $h_i \leftarrow \xi_{t_i} - \xi_{t_{i-1}}$;
\State $\rho_i \leftarrow \frac{\xi_{s_i} - \xi_{t_{i-1}}}{h_i}$;
\State $\bx_{s_i} \leftarrow \frac{\nu_{s_i}}{\nu_{t_{i-1}}} \bx_{t_{i-1}} - \sigma_{s_i} \left( e^{\rho_i{h_i}} - 1 \right) \bepsilon^\btheta_{t_{i-1}}\left( \bx_{t_{i-1}}\right)$; \Comment{DPM-Solver-1 \eqref{equation:dpmsolverk1}}
\State $\small \bx_{t_i} \leftarrow \frac{\nu_{t_i}}{\nu_{t_{i-1}}} \bx_{t_{i-1}} - \sigma_{t_i}(e^{h_i} - 1)\bepsilon^\btheta_{t_{i-1}}(\bx_{t_{i-1}}) - \frac{\sigma_{t_i}}{2\rho_i}(e^{h_i} - 1)\left( \bepsilon^\btheta_{s_i}(\bx_{s_i}) - \bepsilon^\btheta_{t_{i-1}}(\bx_{t_{i-1}}) \right)$;
\EndFor
\State \Return $\bx_{t_M}$;
\end{algorithmic}
\end{algorithm}

One important consideration for the algorithms presented above is the choice of timestep discretization, which is typically performed using uniform spacing in time (linear time) or in log-SNR space.
For normalized timesteps with $t_0 = 1$ and $t_M = 10^{-3}$, uniform discretization along the time axis is defined as:
\begin{equation}
t_i = t_0 - \frac{i}{M}(t_0 - t_M).
\end{equation}
\citet{karras2022elucidating} proposed a noise-level discretization scheme optimized for visual quality, while leaving the solver mathematics unchanged.
Specifically, noise levels $\sigma_i$ are distributed according to a power-law rule:
\begin{equation}
\sigma_i = \left( \sigma_{\text{max}}^{1/\kappa} + \frac{i}{M-1} (\sigma_{\text{min}}^{1/\kappa} - \sigma_{\text{max}}^{1/\kappa}) \right)^\kappa,
\end{equation}
where $\kappa$ (usually set to 7) controls the curvature of the distribution. This scheme yields two key properties:
\begin{itemize}
\item \textit{Denser steps at low $\sigma$:} More computational effort is spent in the late stages of denoising where fine details and high-frequency textures are resolved.
\item \textit{Sparser steps at high $\sigma$:} Larger jumps are taken in the early, high-noise stages where the signal-to-noise ratio is low and coarse structures are formed.
\end{itemize}

\begin{algorithm}[ht]
\caption{DPM-Solver-2M }
\label{alg:dpmsolver2m}
\begin{algorithmic}[1]
\Require initial value $\bx_T$, timesteps $\{t_i\}_{i=0}^M$, noise prediction model $\bepsilon^\btheta_t$.
\State ${\bx}_{t_0} \leftarrow \bx_T\sim\normal(\bzero, \bI_D)$. Initialize an empty buffer $\sB$;
\State $\sB \xleftarrow{\text{buffer}} \bepsilon^\btheta_{t_0}({\bx}_{t_0})$;
\State $h_1 \leftarrow \xi_{t_1} - \xi_{t_{0}}$;
\State ${\bx}_{t_1} \leftarrow \frac{\nu_{t_1}}{\nu_{t_0}}{\bx}_{t_0} - \sigma_{t_1}\big(e^{h_1} - 1\big)\bepsilon^\btheta_{t_0}({\bx}_{t_0})$;
\State $\sB \xleftarrow{\text{buffer}} \bepsilon^\btheta_{t_1}({\bx}_{t_1})$;
\For{$i \leftarrow 2$ to $M$}
\State $h_i \leftarrow \xi_{t_i} - \xi_{t_{i-1}}$;
\State $\rho_i \leftarrow \frac{h_{i-1}}{h_i}$;
\State $\bd_i \leftarrow \big(1 + \frac{1}{2 \rho_i}\big)\bepsilon^\btheta_{t_{i-1}}({\bx}_{t_{i-1}} ) - \frac{1}{2 \rho_i}\bepsilon^\btheta_{t_{i-2}}({\bx}_{t_{i-2}})$; \Comment{Linear multistep  approximation}
\State ${\bx}_{t_i} \leftarrow \frac{\nu_{t_i}}{\nu_{t_{i-1}}}{\bx}_{t_{i-1}} - \sigma_{t_i}\big(e^{h_i} - 1\big)\bd_i$;
\If{$i < M$}
\State $\sB \xleftarrow{\text{buffer}} \bepsilon^\btheta_{t_i}({\bx}_{t_i})$;
\EndIf
\EndFor
\State \Return ${\bx}_{t_M}$;
\end{algorithmic}
\end{algorithm}

\index{Linear multistep approximation}
\subsection{From Singlestep to Multistep}\label{section:dpmsolvermulti}
At each step (from $t_{i-1}$ to $t_i$), the single-step solver described earlier requires two sequential evaluations of the neural network $\bepsilon^\btheta_t$. Furthermore, the intermediate values $\bx_{s_i}$ are used only once and then discarded. This approach discards prior information and can be computationally inefficient. In this section, we introduce an alternative second-order diffusion ODE solver that leverages historical information at each step.

In general, a widely used alternative for approximating the derivatives $\widehatbepsilon_\btheta^{(n)}$ in \eqref{equation:dpm_int_tay} for $n \geq 1$ is the family of \textit{multistep methods}  \citep{atkinson2009numerical}. Given the previous states $\{{\bx}_{t_j}\}_{j=0}^{i-1}$ up to time $t_{i-1}$, multistep methods reuse these historical values to approximate higher-order derivatives. Empirically, multistep methods are more efficient than single-step methods, especially under a limited number of function evaluations.
We integrate multistep design principles with the Taylor expansions in \eqref{equation:dpm_int_tay} to obtain a multistep second-order solver for diffusion ODEs based on $\bepsilon^\btheta_t$.
The full algorithm is given in Algorithm~\ref{alg:dpmsolver2m}, referred to as \textit{DPM-Solver-2M}. This method computes the state $\bx_{t_i}$ by combining the two previous states $\bx_{t_{i-1}}$ and $\bx_{t_{i-2}}$ using a \textit{linear multistep (LMS)} approximation, with no need for extra intermediate points $\bx_{s_i}$.
A convergence analysis is provided in \citet{lu2025dpm}.

For a fixed total function evaluation budget $N$, multistep methods can perform $M = N$ steps, whereas $k$-th order single-step methods are limited to at most $M = N/k$ steps. As a result, the step size $h_i$ for multistep methods is roughly $1/k$ that of single-step methods, meaning the high-order error terms $\mathcalO(h_i^{k+1})$ in \eqref{equation:dpm_int_tay} can be significantly smaller for multistep methods.

\index{Flow-DPM-solver}
\index{DPM-Solver++}
\section{DPM-Solver++}\label{section:dpmsolver++}

The primary objective of \textit{DPM-Solver++} compared to the original DPM-Solver is to address critical failures of noise-prediction-based high-order solvers under large guidance scales in conditional diffusion sampling \citep{lu2025dpm}. 
While DPM-Solver achieves fast, stable unconditional generation by discretizing ODEs derived from the noise network $\bepsilon^\btheta_t$, it suffers from severe numerical instability and train-test distribution mismatch when guidance strength is high (see Section~\ref{section:guid_ddpm}), often yielding degraded samples that perform worse than the low-order DDIM method. DPM-Solver++ reparameterizes the diffusion ODE around the data prediction model $\bx^\btheta_t$ instead of $\bepsilon^\btheta_t$, which reduces the magnitude of error amplification in high-order discretization terms and natively supports dynamic thresholding to constrain outputs within the valid range of training data. It further introduces both single-step (2S) and multistep (2M) variants; as mentioned previously,  the multistep design reduces effective step sizes to broaden the solver's convergence radius and eliminate instability. Ultimately, DPM-Solver++ retains DPM-Solver's high-order exponential integrator framework and fast sampling speed of 15–20 NFEs, while specializing the solver for practical large-guidance text-to-image generation pipelines where the original DPM-Solver breaks down.

\subsection{The DPM-Solver++ Methodology}
Following the transition kernel \eqref{equation:dpm_diff_kernel} in diffusion models, we define the prediction of the original data $\bx_0$ at timestep $t$ as $\bx_t^\btheta(\bx_t)$,given in \eqref{equation:dpmsolver_data_pred}:
\begin{equation}\label{equation:dpmsolver_data_pred2}
\bx^\btheta_t(\bx_t) \triangleq \frac{\bx_t - \sigma_t \bepsilon^\btheta_t(\bx_t)}{\nu_t}
\qquad\implies\qquad 
\bepsilon^\btheta_t(\bx_t) = \frac{\bx_t -\nu_t \bx^\btheta_t(\bx_t) }{\sigma_t}.
\end{equation}
The equivalent ODE  (Theorem~\ref{theorem:ode_diffsde}) expressed in terms of the data prediction model $\bx^\btheta_t$ is obtained by substituting \eqref{equation:dpmsolver_data_pred2} into \eqref{equation:diff_denoid_ode2}:
\begin{equation}\label{equation:dpmode_data}
\frac{\diff \rvx_t}{\diff t} 
= \left(f(t) + \frac{g^2(t)}{2\sigma_t^2}\right)\rvx_t - \frac{\nu_t g^2(t)}{2\sigma_t^2}\bx^\btheta_t(\rvx_t), \quad \rvx_T \sim \normal(\bzero, \bI_D).
\end{equation}
To solve the diffusion ODE in \eqref{equation:dpmode_data} formulated with $\bx^\btheta_t$ in \eqref{equation:dpmode_data}, we present a simplified expression for its exact solution. 
Once again, this formulation computes the linear term in \eqref{equation:dpmode_data} exactly and leaves only an exponentially weighted integral of $\bx^\btheta$ to be approximated. 
Similarly, let $\widehatbx_\btheta(\widehatbx_\xi, \xi) \triangleq  \bx^\btheta_{t_\xi}(\bx_{t_\xi})$ denote the change-of-variable form of $\bx^\btheta$ with respect to $\xi$. 
We then obtain the following result (cf. \eqref{equation:dpm_int_eps}):
\begin{corollary}[Exact solution of diffusion ODEs using $\bx^\btheta_t$\index{Leibniz integral rule}]\label{corollary:dpm_int_eps_data}
Given an initial value $\bx_s$ at time $s > 0$, the solution $\bx_t$ at time $t \in [0, s]$ of diffusion ODE in \eqref{equation:dpmode_data} is:
\begin{equation}\label{equation:dpm_int_eps_data}
\bx_t = \underbrace{\frac{\sigma_t}{\sigma_s}\bx_s}_{\text{Exactly computed}} 
+ \underbrace{\sigma_t \int_{\xi_s}^{\xi_t} e^\xi \widehatbx_\btheta(\widehatbx_\xi, \xi) \diff \xi}_{\text{Exponentially weighted integral}}.
\end{equation}
\end{corollary}
\begin{proof}[of Corollary~\ref{corollary:dpm_int_eps_data}]
Differentiating \eqref{equation:dpm_int_eps_data} with respect to $t$ gives
\begin{align*}
\frac{\diff \rvx_t}{\diff t} 
&= \frac{\diff \sigma_t}{\diff t} \frac{\rvx_s}{\sigma_s} + \frac{\diff \sigma_t}{\diff t} \int_{\xi_s}^{\xi_t} e^\xi \widehatbx_\btheta(\widehatbx_\xi, \xi) \diff \xi + \frac{\diff \xi_t}{\diff t} \sigma_t e^{\xi_t} \widehatbx_\btheta(\widehatbx_{\xi_t}, \xi_t) \\
&= \frac{\diff \sigma_t}{\diff t}\left( \frac{\rvx_s}{\sigma_s} +  \int_{\xi_s}^{\xi_t} e^\xi \widehatbx_\btheta(\widehatbx_\xi, \xi) \diff \xi\right) 
+ \frac{\diff \xi_t}{\diff t} \sigma_t e^{\xi_t} \widehatbx_\btheta(\widehatbx_{\xi_t}, \xi_t) \\
&= \frac{\diff \sigma_t}{\diff t} \frac{\rvx_t}{\sigma_t} + \frac{\diff \xi_t}{\diff t} \sigma_t e^{\xi_t} \widehatbx_\btheta(\widehatbx_{\xi_t}, \xi_t) 
= \left(f(t) + \frac{g^2(t)}{2\sigma_t^2}\right) \rvx_t - \frac{\nu_t g^2(t)}{2\sigma_t^2} \bx^\btheta_t(\rvx_t),
\end{align*}
where the first equality follows from the product rule and the Leibniz integral rule (see Problem~\ref{prob:ode_sol_ftut}),
and the last equality follows from the definitions of $f(t)$ and $g^2(t)$ in \eqref{equation:dpmforward}.
\end{proof}

As the diffusion ODEs in \eqref{equation:diff_denoid_ode2} (based on $\bepsilon^\btheta_t$) and \eqref{equation:dpmode_data} (based on $\bx^\btheta_t$) are equivalent, their exact solution formulas \eqref{equation:dpm_int_eps} and \eqref{equation:dpm_int_eps_data} are also equivalent. 
However, from the perspective of ODE solver design, these two formulations differ significantly:
\begin{itemize}
\item First, \eqref{equation:dpm_int_eps}  computes the linear term $\frac{\nu_t}{\nu_s}\bx_s$ exactly, while \eqref{equation:dpm_int_eps_data} computes the distinct linear term $\frac{\sigma_t}{\sigma_s}\bx_s$ exactly.
\item Second, for solver construction, the first solution \eqref{equation:dpm_int_eps} requires approximating the integral $\int_{\xi_s}^{\xi_t} e^{-\xi} \widehatbepsilon_\btheta (\widehatbx_\xi, \xi)\diff \xi$, while  the second solution \eqref{equation:dpm_int_eps_data} requires approximating $\int e^\xi \widehatbx^\btheta(\widehatbx_\xi, \xi) \diff \xi$. 
These integrals are fundamentally different, recalling that $\bx^\btheta_t = (\bx_t - \sigma_t \bepsilon^\btheta_t)/\nu_t$). 
\end{itemize}
Consequently, high-order solvers derived from \eqref{equation:dpm_int_eps} and \eqref{equation:dpm_int_eps_data} are inherently distinct. We now describe the general framework for designing high-order ODE solvers based on \eqref{equation:dpm_int_eps_data}.

Given the previous state ${\bx}_{t_{i-1}}$ at timestep $t_{i-1}$, our solver aims to approximate the exact solution at timestep $t_i$.  
For any $k \geq 1$, we take the $(k-1)$-th Taylor expansion  of $\widehatbx_\btheta$ around $\xi_{t_{i-1}}$  for $\xi \in [\xi_{t_{i-1}}, \xi_{t_i}]$, and substitute it into \eqref{equation:dpm_int_eps_data} with $s = t_{i-1}$ and $t = t_i$, yielding  (cf. \eqref{equation:dpm_int_tay}):
\begin{mybox}
\begin{equation}\label{equation:dpm_int_tay++}
{\bx}_{t_i} = \frac{\sigma_{t_i}}{\sigma_{t_{i-1}}}{\bx}_{t_{i-1}} 
+ \sigma_{t_i} \sum_{n=0}^{k-1} 
\underbrace{\widehatbx_\btheta^{(n)}(\widehatbx_{\xi_{t_{i-1}}}, \xi_{t_{i-1}})}_{\text{Derivatives}} 
\underbrace{\int_{\xi_{t_{i-1}}}^{\xi_{t_i}} e^\xi \frac{(\xi - \xi_{t_{i-1}})^n}{n!} \diff \xi}_{\triangleq C_n} + \mathcalO(h_i^{k+1}),
\end{equation}
\end{mybox}
where again $h_i \triangleq \xi_{t_i} - \xi_{t_{i-1}}$, and $\widehatbx_\btheta^{(n)}(\widehatbx_\xi,\xi) \triangleq  \frac{\diff ^n \widehatbx_\btheta(\widehatbx_\xi, \xi)}{\diff \xi^n}$ denotes the $n$-th order  derivative of $\widehatbx_\btheta(\widehatbx_\xi, \xi)$ w.r.t. log-SNR $\xi$ (see definition in \eqref{equation:logsnr}). 
Analogous to \eqref{equation:dpmsolverc_n}, the integral coefficients $C_n$ can be computed analytically via integration by part. 
Thus, to construct  a $k$-th order ODE solver, we only need to estimate the $n$-th order derivatives $\widehatbx_\btheta^{(n)}(\widehatbx_{\xi_{t_{i-1}}}, \xi_{t_{i-1}})$ for $n \leq k-1$ after discarding  the $\mathcalO(h_i^{k+1})$ high-order error terms. 
These techniques are well-established and were discussed in detail in the DPM-Solver section. As with the original DPM-Solver, the special case $k = 1$ recovers a solver equivalent to DDIM (see \eqref{equation:dpmsolverk1}).

For $k = 2$, we apply a similar strategy to DPM-Solver-2 (Algorithm~\ref{alg:dpmsolver2}) to estimate the derivative $\widehatbx_\btheta^{(1)}(\widehatbx_{\xi_{t_{i-1}}}, \xi_{t_{i-1}})$.
At each step, given ${\bx}_{t_{i-1}}$,  \textit{DPM-Solver++(2S)} performs the following updates (cf. \eqref{equation:dpmsolver2_update}):
\begin{align}
\bx_{s_i} &\leftarrow \frac{\sigma_{s_i}}{\sigma_{t_{i-1}}}{\bx}_{t_{i-1}} - \nu_{s_i}\big(e^{-\rho h_i} - 1\big)\bx^\btheta_{t_{i-1}}({\bx}_{t_{i-1}});  \qquad (\text{DPM-Solver++(1)})\nonumber \\
{\bx}_{t_i} &\leftarrow \frac{\sigma_{t_i}}{\sigma_{t_{i-1}}}{\bx}_{t_{i-1}} - \nu_{t_i}\big(e^{-h_i} - 1\big)
\underbrace{\Big(\big(1 - \frac{1}{2 \rho}\big)\bx^\btheta_{t_{i-1}}({\bx}_{t_{i-1}}) + \frac{1}{2 \rho}\bx^\btheta_{s_i}(\bx_{s_i})\Big)}_{\text{Convex combination}},
\label{equation:dpmsolver++2_update}
\end{align}
where $0<\rho<1$ (typically $\rho=0.5$).
The full procedure is given in Algorithm~\ref{alg:dpmsolver++2s} ({DPM-Solver++(2S)}).
Similarly, we may explicitly use $2M+1$ timesteps, consisting of ($\{t_i\}_{i=0}^M$ and intermediate points $\{s_i\}_{i=1}^M$), such that $t_0 > s_1 > t_1 > \ldots > t_{M-1} > s_M > t_M$, to estimate the solution trajectory. We omit the detailed derivation here for brevity (cf. Algorithm~\ref{alg:dpmsolver2prime}).

\begin{algorithm}[ht]
\caption{DPM-Solver++(2S) \citep{lu2025dpm}}
\label{alg:dpmsolver++2s}
\begin{algorithmic}[1]
\Require initial value $\bx_T$, timesteps $\{t_i\}_{i=0}^M$, data prediction model $\bx^\btheta_t$, $0<\rho<1$ (e.g., $\rho=0.5$);
\State ${\bx}_{t_0} \leftarrow \bx_T\sim \normal(\bzero, \bI)$;
\For{$i \leftarrow 1$ to $M$};
\State $h_i \leftarrow \xi_{t_i} - \xi_{t_{i-1}}$;
\State $s_i \leftarrow t_\xi\left( \rho \xi_{t_i} + (1 - \rho)\xi_{t_{i-1}} \right)$; \Comment{Obtain the timestep $s_i$}
\State $\bx_{s_i} \leftarrow \frac{\sigma_{s_i}}{\sigma_{t_{i-1}}}{\bx}_{t_{i-1}} - \nu_{s_i}\big(e^{-\rho h_i} - 1\big)\bx^\btheta_{t_{i-1}}({\bx}_{t_{i-1}} )$ ; \Comment{DPM-Solver++(1)}
\State ${\bx}_{t_i} \leftarrow \frac{\sigma_{t_i}}{\sigma_{t_{i-1}}}{\bx}_{t_{i-1}} - \nu_{t_i}\big(e^{-h_i} - 1\big)\Big(\big(1 - \frac{1}{2 \rho}\big)\bx^\btheta_{t_{i-1}}({\bx}_{t_{i-1}}) + \frac{1}{2 \rho}\bx^\btheta_{s_i}(\bx_{s_i})\Big)$;
\EndFor
\State \Return ${\bx}_{t_M}$;
\end{algorithmic}
\end{algorithm}


Similar to the multistep DPM-Solver in Algorithm~\ref{alg:dpmsolver2m}, the DPM-Solver++(2S) method can also be extended to a multistep formulation, as presented in Algorithm~\ref{alg:dpmsolver++2m}. We omit the repetitive technical details here.

Furthermore, while both the DPM-Solver and DPM-Solver++ frameworks are primarily designed for diffusion model sampling, extended variants support sampling for flow-based models. One representative example is \textit{Flow-DPM-Solver} \citep{xie2024sana}.

\begin{algorithm}[ht]
\caption{DPM-Solver++(2M) \citep{lu2025dpm}}
\label{alg:dpmsolver++2m}
\begin{algorithmic}[1]
\Require initial value $\bx_T$, timesteps $\{t_i\}_{i=0}^M$, data prediction model $\bx^\btheta_t$;
\State ${\bx}_{t_0} \leftarrow \bx_T\sim\normal(\bzero, \bI_D)$. Initialize an empty buffer $\sB$;
\State $\sB \xleftarrow{\text{buffer}} \bx^\btheta_{t_0}({\bx}_{t_0})$;
\State $h_1 \leftarrow \xi_{t_1} - \xi_{t_{0}}$;
\State ${\bx}_{t_1} \leftarrow \frac{\sigma_{t_1}}{\sigma_{t_0}}{\bx}_{t_0} - \nu_{t_1}\big(e^{-h_1} - 1\big)\bx^\btheta_{t_0}({\bx}_{t_0})$;
\State $\sB \xleftarrow{\text{buffer}} \bx^\btheta_{t_1}({\bx}_{t_1})$;
\For{$i \leftarrow 2$ to $M$}
\State $h_i \leftarrow \xi_{t_i} - \xi_{t_{i-1}}$;
\State $\rho_i \leftarrow \frac{h_{i-1}}{h_i}$;
\State $\bd_i \leftarrow \big(1 + \frac{1}{2 \rho_i}\big)\bx^\btheta_{t_{i-1}}({\bx}_{t_{i-1}} ) - \frac{1}{2 \rho_i}\bx^\btheta_{t_{i-2}}({\bx}_{t_{i-2}})$; \Comment{Linear multistep  approximation}
\State ${\bx}_{t_i} \leftarrow \frac{\sigma_{t_i}}{\sigma_{t_{i-1}}}{\bx}_{t_{i-1}} - \nu_{t_i}\big(e^{-h_i} - 1\big)\bd_i$;
\If{$i < M$}
\State $\sB \xleftarrow{\text{buffer}} \bx^\btheta_{t_i}({\bx}_{t_i})$;
\EndIf
\EndFor
\State \Return ${\bx}_{t_M}$;
\end{algorithmic}
\end{algorithm}

\subsection{Comparison between DPM-Solvers}
We now compare DPM-Solver-2~\eqref{equation:dpmsolver2_update} and DPM-Solver++(2S)~\eqref{equation:dpmsolver++2_update}.
Recall that, given $\bx_{t_{i-1}}$ at each step,   DPM-Solver-2 performs the following updates:
\begin{subequations}\label{equation:dpmsolverrel_ALL}
{\small
\begin{align}
\bx_{s_i} &\leftarrow \frac{\nu_{s_i}}{\nu_{t_{i-1}}}\bx_{t_{i-1}} - \sigma_{s_i}\big(e^{\rho h_i} - 1\big)\bepsilon^\btheta_{t_{i-1}}(\bx_{t_{i-1}});   \\
\bx_{t_i} &\leftarrow \frac{\nu_{t_i}}{\nu_{t_{i-1}}}\bx_{t_{i-1}} - \sigma_{t_i}\big(e^{h_i} - 1\big)\bepsilon^\btheta_{t_{i-1}}(\bx_{t_{i-1}}) 
- \frac{\sigma_{t_i}}{2 \rho}\big(e^{h_i} - 1\big)\big(\bepsilon^\btheta_{s_i}(\bx_{s_i}) - \bepsilon^\btheta_{t_{i-1}}(\bx_{t_{i-1}})\big).
\end{align}}
\end{subequations}
Using the relationship between the data prediction and noise prediction models from \eqref{equation:dpmsolver_data_pred2},
$$
\bx^\btheta_t(\bx_t) = \frac{\bx_t - \sigma_t \bepsilon^\btheta_t(\bx_t)}{\nu_t} = \frac{1}{\nu_t}\bx_t - e^{-\xi_t}\bepsilon^\btheta_t(\bx_t),
$$
we can rewrite DPM-Solver++(2S)~\eqref{equation:dpmsolver++2_update} in terms of the noise prediction model $\bepsilon^\btheta_t$ (see Problem~\ref{prob:dpmsolverrel}):
\begin{subequations}\label{equation:dpmsolverrel_ALL+}
{\small\begin{align}
\bx_{s_i} &= \frac{\nu_{s_i}}{\nu_{t_{i-1}}}{\bx}_{t_{i-1}} - \sigma_{s_i}\big(e^{\rho h_i} - 1\big)\bepsilon^\btheta_{t_{i-1}}({\bx}_{t_{i-1}});
\label{equation:dpmsolverrel1}\\
\bx_{t_i} &= \frac{\nu_{t_i}}{\nu_{t_{i-1}}}{\bx}_{t_{i-1}} - \sigma_{t_i}\big(e^{h_i} - 1\big)\bepsilon^\btheta_{t_{i-1}}({\bx}_{t_{i-1}}) - \frac{\sigma_{t_i}}{2 \rho}\big(e^{h_i} - 1\big)
\textcolor{mylightbluetext}{\underbrace{e^{-\rho h_i}}_{\triangleq D}}
\big(\bepsilon^\btheta_{s_i}(\bx_{s_i}) - \bepsilon^\btheta_{t_{i-1}}({\bx}_{t_{i-1}})\big).
\label{equation:dpmsolverrel2}
\end{align}}
\end{subequations}
\noindent
The only difference between DPM-Solver-2 \eqref{equation:dpmsolverrel_ALL} and DPM-Solver++(2S) \eqref{equation:dpmsolverrel_ALL+} lies in the constant $D\triangleq e^{-\rho h_i}<1$ applied to the second term, which corresponds to the approximation of the first-order total derivative $\widehatbepsilon_\btheta^{(1)}$.
Specifically, we have
$$
\bepsilon^\btheta_{s_i}(\bx_{s_i}) - \bepsilon^\btheta_{t_{i-1}}({\bx}_{t_{i-1}}) = (\widehatbepsilon^\btheta_{t_{i-1}})^{(1)}({\bx}_{t_{i-1}}) + \mathcalO(h_i).
$$
Since DPM-Solver++(2S) multiplies a smaller coefficient into the $\mathcalO(h_i)$ error term, the leading constant associated with its high-order error is smaller than that of DPM-Solver-2.
While both methods correspond to second-order discretizations of the diffusion ODE, a smaller error constant leads to lower discretization error and reduced numerical instability, particularly under large guidance scales (see Section~\ref{section:cfg_ddpm}).
Thus, employing the data prediction model is key to stabilizing the sampling process, and DPM-Solver++(2S) is consequently more stable than DPM-Solver-2.

\subsection{Diffusion SDEs Solvers Variants}

Sampling with diffusion models can also be performed by solving the diffusion SDE \eqref{equation:denois_diff_sde}, which we restate below:
\begin{equation}\label{equation:denois_diff_sde_sol22}
\diff \rvx_t = \left[f(t)\rvx_t + \frac{g^2(t)}{\sigma_t}\bepsilon^\btheta_t(\rvx_t)\right]\diff t + g(t)\diff\overline{\rvb}_t, \quad \rvx_T \sim p_T(\rvx_T) \equiv\normal(\bzero, \bI_D).
\end{equation}
The solution to this diffusion SDE is established in Theorem~\ref{theorem:sde_semilineare}; and solutions expressed in terms of the log-SNR $\xi$ for noise and data predictions are given in \eqref{equation:sde_int_eps} and \eqref{equation:sde_int_eps2}, respectively.

For brevity, given a state $\bx_s$ at timestep $s$, we aim to compute the predicted state $\bx_t$ at timestep $t<s$.
Let $h\triangleq \xi_t-\xi_s$. 
The DPM-Solver framework admits the following \textit{SDE-DPM-Solver} variants:
\begin{enumerate}[(i)]
\item \textit{SDE-DPM-Solver-1.}
Under the approximation  $\widehatbepsilon_\btheta(\widehatbx_\xi, \xi) \approx \bepsilon^\btheta_s(\bx_s)$ (analogous to the exponential integrator approximation in \eqref{equation:ei_sol}), we obtain
\begin{equation}\label{equation:sdedpmsolver1}
\bx_t = \frac{\nu_t}{\nu_s}\bx_s - 2\sigma_t(e^h - 1)\bepsilon^\btheta_s(\bx_s) + \sigma_t \sqrt{e^{2h} - 1}\bepsilon_s.
\end{equation}

\item \textit{SDE-DPM-Solver++(1).}
Under the approximation $\widehatbx_\btheta(\widehatbx_\xi, \xi) \approx \bx^\btheta_s(\bx_s)$, we derive
\begin{equation}\label{equation:sdedpmsolver1++}
\bx_t = \frac{\sigma_t}{\sigma_s}e^{-h}\bx_s + \nu_t(1 - e^{-2h})\bx^\btheta_s(\bx_s) + \nu_t \sqrt{1 - e^{-2h}}\bepsilon_s.
\end{equation}

\item \textit{SDE-DPM-Solver-2M.}
Suppose we have a prior state $\bx_r$ with corresponding model output $\bepsilon^\btheta_r(\bx_r)$ at timestep $r > t$, and let $\rho = \frac{\xi_r - \xi_s}{h}$. 
Using the linear approximation $\widehatbepsilon_\btheta(\widehatbx_\xi, \xi) \approx \bepsilon^\btheta_s(\bx_s) + \frac{\xi - \xi_s}{\rho h}\big(\bepsilon^\btheta_r(\bx_r) - \bepsilon^\btheta_s(\bx_s)\big)$, we arrive at
\begin{equation}\label{equation:sdedpmsolver2m}
\bx_t = \frac{\nu_t}{\nu_s}\bx_s - 2\sigma_t(e^h - 1)\bepsilon^\btheta_s(\bx_s) - \sigma_t(e^h - 1)\frac{\bepsilon^\btheta_r(\bx_r) - \bepsilon^\btheta_s(\bx_s)}{\rho} + \sigma_t \sqrt{e^{2h} - 1}\bepsilon_s.
\end{equation}

\item \textit{SDE-DPM-Solver++(2M).}
Similarly, suppose we have a prior state $\bx_r$ with model output $\bx^\btheta_r(\bx_r)$ at timestep $r > t$, and let $\rho = \frac{\xi_r - \xi_s}{h}$. 
Using the linear approximation $\widehatbx_\btheta(\widehatbx_\xi, \xi) \approx \bx^\btheta_s(\bx_s) + \frac{\xi - \xi_s}{\rho h}\big(\bx^\btheta_r(\bx_r) - \bx^\btheta_s(\bx_s)\big)$, we obtain
\begin{equation}\label{equation:sdedpmsolver2m++}
\bx_t = \frac{\sigma_t}{\sigma_s}\frac{1}{e^{h}}\bx_s + \nu_t(1 - \frac{1}{e^{2h}})\bx^\btheta_s(\bx_s) + \frac{\nu_t(1 - \frac{1}{e^{2h}})}{2}\left(\frac{\bx^\btheta_r(\bx_r) - \bx^\btheta_s(\bx_s)}{\rho}\right) + \nu_t \sqrt{1 - \frac{1}{e^{2h}}}\bepsilon_s.
\end{equation}
\end{enumerate}
To verify these results, for \eqref{equation:sdedpmsolver1}, substituting the approximation $\widehatbepsilon_\btheta(\widehatbx_\xi, \xi) \approx \bepsilon^\btheta_s(\bx_s)$ into \eqref{equation:sde_int_eps}, we have 
\begin{align*}
2\nu_t \int_{\xi_s}^{\xi_t} e^{-\xi} \widehatbepsilon_\btheta(\widehatbx_\xi, \xi) \diff \xi
\approx 2\nu_t \left( \int_{\xi_s}^{\xi_t} e^{-\xi} \diff \xi \right) \bepsilon^\btheta_s(\bx_s)
= 2\sigma_t (e^h - 1) \bepsilon^\btheta_s(\bx_s).
\end{align*}
For \eqref{equation:sdedpmsolver1++}, substituting $\widehatbx_\btheta(\widehatbx_\xi, \xi) \approx \bx^\btheta_s(\bx_s)$ into \eqref{equation:sde_int_eps2} gives
\begin{align*}
2\nu_t \int_{\xi_s}^{\xi_t} e^{-2(\xi_t - \xi)} \widehatbx_\btheta(\widehatbx_\xi, \xi) \diff \xi
\approx 2\nu_t e^{-2\xi_t} \left( \int_{\xi_s}^{\xi_t} e^{2\xi} \diff \xi \right) \bx^\btheta_s(\bx_s)  
= \nu_t (1 - e^{-2h}) \bx^\btheta_s(\bx_s). 
\end{align*}
For \eqref{equation:sdedpmsolver2m}, substituting the linear approximation of $\widehatbepsilon_\btheta(\widehatbx_\xi, \xi) \approx \bepsilon^\btheta_s(\bx_s) + \frac{\xi - \xi_s}{\rho h}\big(\bepsilon^\btheta_r(\bx_r) - \bepsilon^\btheta_s(\bx_s)\big)$ into \eqref{equation:sde_int_eps} leads to
{\small
\begin{align*}
2\nu_t \int_{\xi_s}^{\xi_t} e^{-\xi} \widehatbepsilon_\btheta(\widehatbx_\xi, \xi) \diff \xi  
&\approx 2\nu_t \left( \int_{\xi_s}^{\xi_t} e^{-\xi} \diff \xi \right) \bepsilon^\btheta_s(\bx_s) + \frac{2\nu_t}{\rho h} \left( \int_{\xi_s}^{\xi_t} e^{-\xi} (\xi - \xi_s) \diff \xi \right) \big(\bepsilon^\btheta_r(\bx_r) - \bepsilon^\btheta_s(\bx_s)\big)  \\
&= 2\sigma_t (e^h - 1) \bepsilon^\btheta_s(\bx_s) + 2\sigma_t \left( \frac{e^h - 1 - h}{h} \right) \left( \frac{\bepsilon^\btheta_r(\bx_r) - \bepsilon^\btheta_s(\bx_s)}{\rho} \right)\\
&\approx 2\sigma_t (e^h - 1) \bepsilon^\btheta_s(\bx_s) + \frac{\sigma_t (e^h - 1)}{\rho} \big(\bepsilon^\btheta_r(\bx_r) - \bepsilon^\btheta_s(\bx_s)\big). 
\end{align*}
}
\noindent
where the final approximation uses the relation$\frac{e^h - 1 - h}{h} \approx \frac{h}{2} + \mathcalO(h^2) \approx \frac{e^h - 1}{2}$ (Problem~\ref{prob:tay_order4}).
For \eqref{equation:sdedpmsolver2m++}, substituting the linear approximation of $\widehatbx_\btheta(\widehatbx_\xi, \xi) \approx \bx^\btheta_s(\bx_s) + \frac{\xi - \xi_s}{\rho h}\big(\bx^\btheta_r(\bx_r) - \bx^\btheta_s(\bx_s)\big)$ into \eqref{equation:sde_int_eps2} results in
{\small
\begin{align*}
\int_{\xi_s}^{\xi_t} e^{-2(\xi_t - \xi)} \widehatbx_\btheta(\widehatbx_\xi, \xi) \diff \xi 
&\approx  e^{-2\xi_t} \left[\left( \int_{\xi_s}^{\xi_t} e^{2\xi} \diff \xi \right) \bx^\btheta_s(\bx_s) +  \left( \int_{\xi_s}^{\xi_t} e^{2\xi} (\xi - \xi_s) \diff \xi \right) \frac{\bx^\btheta_r(\bx_r) - \bx^\btheta_s(\bx_s)}{\rho h}\right]  \\
&= \frac{1}{2} (1 - e^{-2h}) \bx^\btheta_s(\bx_s) + \frac{1}{2} \left( \frac{e^{-2h} - 1 + 2h}{2h} \right) \left( \frac{\bx^\btheta_r(\bx_r) - \bx^\btheta_s(\bx_s)}{\rho} \right)\\
&\approx \frac{1}{2} (1 - e^{-2h}) \bx^\btheta_s(\bx_s) + \frac{ (1 - e^{-2h})}{4} \left( \frac{\bx^\btheta_r(\bx_r) - \bx^\btheta_s(\bx_s)}{\rho} \right),
\end{align*}
}
where the last approximation follows from the approximation $\frac{e^{-2h} - 1 + 2h}{2h} \approx h + \mathcalO(h^2) \approx \frac{1 - e^{-2h}}{2}$ (Problem~\ref{prob:tay_order4}).

\begin{problemset}
\item \label{prob:ve} \textbf{Variance exploding.} 
Consider the distribution of $\bx_t$ defined by the following Markov chain:
\begin{equation}
\bx_t = \bx_{t-1} + \sqrt{\sigma_t^2 - \sigma_{t-1}^2} \bepsilon_{t-1}, \quad t = 1, \ldots, T,
\end{equation}
where $\bepsilon_{t-1} \sim \normal(\bzero, \bI_D)$,  $\sigma_0 = 0$,  and $\sigma_t$ is increasing. Derive the SDE corresponding to this chain.
	
\item \label{prob:dpmode2} Starting from \eqref{equation:dpmode1}, derive \eqref{equation:dpmode2}.
\textit{Hint: Use use the first-order Taylor expansion of $\nu_t$ around $m$: $\nu_t-\nu_m\approx\dot{\nu}(t-m) =\dot{\nu}\Delta t$.
Similarly, we have $\Delta \xi = \xi_t-\xi_m\approx\dot{\xi}(t-m) =\dot{\xi}\Delta t$.
Then apply the Taylor series $1-e^{2\Delta \xi}\approx-2\Delta \xi -\frac{(2\Delta\xi)^2}{2!}-\ldots \approx -2\dot{\xi}\Delta t$.
Since $\bepsilon \sim \normal(\bzero, \bI_D)$, we can write $\bepsilon \Delta t^{1/2} = \diff \rvb_t$ (Wiener increment; see Problem~\ref{prob:brownian_motion} and \eqref{equation:brownian_motion}).
}

\item \label{prob:tay_order4} Show that ${2e^{h} - 2 - h - h e^{h}} = {-h^3/6 + \mathcalO(h^4)}$. 
Additionally, show that:
\begin{itemize}
\item $\frac{e^h - 1 - h}{h} = \frac{h}{2} + \mathcalO(h^2) $ and $  \frac{e^h - 1}{2} =\frac{h}{2} + \mathcalO(h^2)$.
\item $\frac{e^{-2h} - 1 + 2h}{2h} = h + \mathcalO(h^2) $ and $ \frac{1 - e^{-2h}}{2} =h + \mathcalO(h^2)$.
\end{itemize}
\textit{Hint: Use the Taylor series expansion $e^{h} = 1 + h + \frac{h^2}{2} + \frac{h^3}{3} + \frac{h^4}{4} + \mathcalO(h^5)$.}

\item Using a derivation analogous to that of DPM-Solver-2 in \eqref{equation:dpmsolver2}, derive the DPM-Solver-3 with order  $k=3$ based on the approximation \eqref{equation:dpm_int_tay}. What is the total number of forward passes in the DPM-Solver-3 algorithm?

\item Using a derivation analogous to that of DPM-Solver-2 in \eqref{equation:dpmsolver2}, derive the DPM-Solver++(1) used in Algorithm~\ref{alg:dpmsolver++2s} with order $k=1$ from the approximation \eqref{equation:dpm_int_tay++}. 
Similarly, derive the DPM-Solver++(2S) with order $k=2$ from the approximation \eqref{equation:dpm_int_tay++}.

\item  \label{prob:dpmsolverrel} Prove the forms of DPM-Solver++(2S) in \eqref{equation:dpmsolverrel1} and \eqref{equation:dpmsolverrel2}. 
\textit{Hint: Use the fact that $\xi_{s_i} = \rho\xi_{t_i} + (1-\rho)\xi_{t_{i-1}}$ such that $\rho h_i=\rho(\xi_{t_i} - \xi_{t_{i-1}}) = \xi_{s_i} - \xi_{t_{i-1}}$.
For the second identity, you may also want to show that $e^{\xi_{t_{i-1}}}\big(\bx^\btheta_{s_i}(\bx_{s_i}) - \bx^\btheta_{t_{i-1}}({\bx}_{t_{i-1}})\big) = e^{-\rho h_i}\big(\bepsilon^\btheta_{t_{i-1}}(\bx_{t_{i-1}}) - \bepsilon^\btheta_{s_i}(\bx_{s_i})\big)$.
}

\item \label{prob:logsnrdiff} Let $\nu^2+\sigma^2=1$, and let $\xi\triangleq \ln(\nu/\sigma)$ denote the log-SNR. Show that $\frac{\diff \ln \nu}{\diff \xi}=\sigma^2$ and $\frac{\diff \ln \sigma}{\diff \xi} = -\nu^2$.
\textit{Hint: Differentiate $\nu^2+\sigma^2=1$ w.r.t. $\xi$ to show that $\nu\frac{\diff \nu}{\diff \xi} + \sigma \frac{\diff \sigma}{\diff \xi}=0$, with which we can show that $\nu^2\frac{\diff \ln\nu}{\diff \xi}+ \sigma^2\frac{\diff \ln \sigma}{\diff \xi} = 0$.
Also show that $1=\frac{\diff \ln \nu}{\diff \xi} - \frac{\diff \ln \sigma}{\diff \xi}$.
Using these results to establish the final identities.}

\item \label{prob:logsnrdiff_sol}
Consider the same setup as in \eqref{equation:sde_int_eps}, and revisit the diffusion SDEs parameterized by the log-SNR $\xi$ given in \eqref{equation:logsnr}.
Define $\diff \rvb_\xi \triangleq \sqrt{-\frac{\diff t}{\diff \xi}}\diff \overline{\rvb}_{t_\xi}$ as the corresponding Wiener process w.r.t. log-SNR $\xi$. For simplicity, let $\widehatbx_\xi \triangleq \bx_{t_\xi}$, $\sigma_\xi \triangleq \sigma_{t_\xi}$, and $\widehatbepsilon_\btheta(\widehatbx_\xi, \xi) \triangleq \bepsilon^\btheta_{t_\xi}(\bx_{t_\xi})$. For {variance preserving  diffusion models}  (satisfying $\nu_t^2 + \sigma_t^2 = 1$), we have $\frac{\diff \ln \nu_\xi}{\diff \xi} = \sigma_\xi^2$ and $\frac{\diff \ln \sigma_\xi}{\diff \xi} = -\nu_\xi^2$ (see Problem~\ref{prob:logsnrdiff}).  
Use these identities to show that the diffusion SDE~\eqref{equation:denois_diff_sde_sol} w.r.t. $\xi$ is
\begin{align}
\diff \widehat{\rvx}_\xi 
&= \left[\sigma_\xi^2 \widehat{\rvx}_\xi - 2\sigma_\xi \widehatbepsilon_\btheta(\widehat{\rvx}_\xi, \xi)\right]\diff \xi + \sqrt{2}\sigma_\xi \diff \rvb_\xi
\nonumber\\
&= \left[-(1 + \nu_\xi^2)\widehat{\rvx}_\xi + 2\nu_\xi \widehatbx_\btheta(\widehat{\rvx}_\xi, \xi)\right]\diff \xi + \sqrt{2}\sigma_\xi \diff \rvb_\xi, \nonumber
\end{align}
where the second equality follows from the relationship between data and noise prediction models in \eqref{equation:dpmsolver_data_pred}.
Following the proof of Theorem~\ref{theorem:sde_semilineare}, use the first expression to derive \eqref{equation:sde_int_eps} via an alternative route, and use the second expression to derive \eqref{equation:sde_int_eps2}.

\end{problemset}
\newpage 
\chapter{Generative Architectures for Image and Video Synthesis}\label{chapter:diffarchitect}
\begingroup
\hypersetup{
linkcolor=structurecolor,
linktoc=page,  
}
\minitoc \newpage
\endgroup

\lettrine{\color{caligraphcolor}O}
Over the past decade, generative models have revolutionized visual content creation and manipulation, enabling the synthesis of photorealistic images, coherent videos, and immersive multimedia at unprecedented scales and quality. Prominent among these advances are open-sourced large-scale generative models including \textit{Stable Diffusion (SD)}, \textit{Flux}, and \textit{W{\"u}rstchen} \citep{rombach2022high, bfl2024flux, pernias2024wurstchen}. These models represent paradigm-shifting breakthroughs in generative AI, with their key scientific and industrial contributions summarized as follows:
\begin{itemize}
\item \textit{Implementation of practical latent diffusion models.} The latent diffusion model (LDM) paradigm relocates diffusion denoising from high-dimensional pixel space to a perceptually equivalent low-dimensional latent space, reducing computational complexity by approximately 100$\times$ \citep{rombach2022high}. This innovation renders diffusion-based generative modeling viable on consumer-grade hardware and establishes latent-space processing as the standard design principle for efficient diffusion/flow models.
	
\item \textit{Standardization of flexible guided diffusion/flow architectures.} By integrating cross-attention layers into the U-Net/DiT backbone \citep{ronneberger2015u, peebles2023scalable}  and coupling this structure with CLIP/OpenCLIP cross-modal encoding \citep{radford2021learning, ilharco_gabriel_2021_5143773}, these models establish a unified framework for multimodal conditional generation. This framework supports diverse input guidance---including text, image, layout, and semantic cues---and eliminates the need for task-specific architectural redesign.

\item \textit{Acceleration of open-source generative AI research.} As the first fully open-sourced state-of-the-art text-to-image diffusion model (releasing both code and pre-trained weights), Stable Diffusion dismantled the proprietary monopoly on high-performance generative models. It fostered a global, decentralized research ecosystem that has driven rapid progress in diffusion model controllability (e.g., ControlNet \citep{zhang2023adding}), lightweight fine-tuning (e.g., LoRA \citep{hu2022lora}), and large-scale model scaling, becoming the fundamental benchmark platform for worldwide LDM variant research.

\item \textit{Translation of academic diffusion/flow research to industrial deployment.} These models bridge the gap between academic diffusion/flow research and real-world industrial applications. They support edge and on-premise deployment to address data privacy constraints and enable widespread scalable adoption across creative design, digital media, and broader AIGC industries. Iterative updates including SDXL and SD3 further advance high-resolution (1024$\times$1024$+$) image synthesis and real-time generation via 1–4 step turbo sampling, setting state-of-the-art benchmarks for generation fidelity and prompt alignment.

\item \textit{Establishment of open generative AI ecosystem norms.} The mainstream success of open latent diffusion models validates open-source development as a sustainable paradigm for cutting-edge generative AI research. It has spawned a comprehensive lineage of open latent diffusion and flow-matching models and promoted standardized generative AI toolchains. Additionally, it highlights critical research directions regarding diffusion model safety, bias mitigation, and generative content governance for both scientific and societal impact.
\end{itemize}

\noindent 
In essence, alongside closed-source proprietary systems such as DALL-E \citep{ramesh2021zero, ramesh2022hierarchical} and Imagen \citep{saharia2022photorealistic}, these open-source generative models advance the practical implementation of diffusion and flow models. By democratizing access to high-performance generative frameworks, they reshape the landscape of generative AI ecosystems and industrial applications, laying solid technical and community foundations for modern latent-based generative modeling. Built upon the pioneering frameworks of variational autoencoders (VAEs), generative adversarial networks (GANs), diffusion models, and flow-based models---each of which establishes core theories for latent space learning, adversarial training, iterative denoising, and invertible transformations---this chapter systematically elaborates on the core components and large-scale architectures underlying state-of-the-art image and video generative systems, starting with the fundamental building blocks of mainstream models such as Stable Diffusion.

Early generative models primarily focused on reconstructing or generating simple, constrained datasets. In contrast, modern generative systems require robust multimodal understanding, precise controllability over generation attributes, and strong scalability for high-resolution, long-form visual content. This evolutionary shift has driven the emergence of novel architectural modules that unify language understanding, visual perception, and sequential modeling, while scaling to billions of parameters.

This chapter first dissects the core building blocks of modern generative architectures (Section~\ref{section:build_block_archi}), starting with \textit{contrastive language-image pre-training (CLIP)} \citep{radford2021learning}---a foundational component for aligning generative outputs with textual prompts. It then elaborates on the embedding strategies for conditional guidance in generation, followed by an in-depth analysis of advanced \textit{attention mechanisms} \citep{vaswani2017attention} that enable long-range context modeling, a prerequisite for coherent and high-fidelity visual synthesis.

Based on these foundational modules, this chapter further investigates large-scale image and video generative models (Section~\ref{section:large_gen}). It begins with the \textit{U-Net} architecture, the dominant backbone of contemporary diffusion-based generators, before introducing \textit{diffusion transformers (DiTs)}, which replace conventional convolutional U-Nets with transformer-based structures to enhance global dependency modeling. Finally, it covers \textit{multimodal diffusion transformers (MM-DiTs)}, which extend DiT frameworks to natively process diverse input modalities (text, image, audio), enabling fully unified multimodal generative pipelines.

By concluding this chapter, we will acquire a systematic understanding of how individual functional components and integrated architectures constitute modern state-of-the-art generative systems. It also clarifies how these advanced designs inherit and innovate upon core principles of VAEs, diffusion models, and flow-based models to continuously expand the boundaries of high-quality visual synthesis.

\section{Building Blocks for Generative Architectures}\label{section:build_block_archi}

Before delving into the design of scalable neural network architectures for diffusion and flow models tailored to visual modalities such as natural images and videos, we first introduce the core building blocks ubiquitously adopted in modern diffusion and flow-based generative models. These modular components constitute the fundamental infrastructure for encoding conditional guidance, modeling long-range spatial dependencies, and aligning generated outputs with semantic and structural constraints---core capabilities that empower these models to produce high-fidelity, precisely controllable visual content.

In the subsequent sections, we elaborate on \textcolor{black}{three} essential foundational components: the CLIP framework for cross-modal semantic conditioning, embedding strategies for guidance variables (including timesteps, text prompts, and control signals), and attention mechanisms for modeling global spatial relationships. Collectively, these modules underpin the representational expressiveness and functional flexibility of state-of-the-art diffusion and flow architectures, serving as the fundamental building blocks for constructing scalable, task-adaptive generative systems.

\begin{figure}[htbp]
	\centering
\subfigure[CLIP training.]{\label{fig:clip_train}
\includegraphics[width=0.455\linewidth]{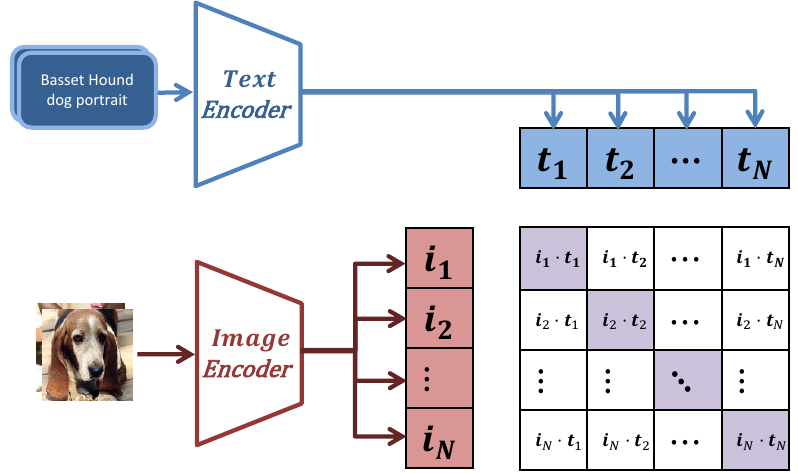}}
\subfigure[Zero-shot prediction.]{\label{fig:clip_zeroshot}
\includegraphics[width=0.455\linewidth]{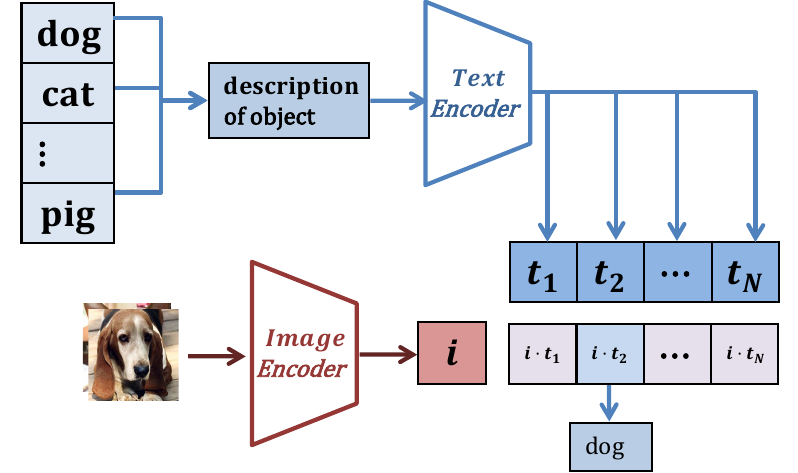}}
\caption{
Conceptual illustration of CLIP. 
Whereas standard image models jointly train an image feature extractor and a linear classifier to predict predefined class labels, CLIP jointly trains an image encoder and a text encoder to predict the correct pairings within a batch of (image, text) training examples. At test time, the learned text encoder synthesizes a zero-shot linear classifier by embedding the names or textual descriptions of the target dataset's classes. Adapted from \citet{radford2021learning}.
}
\label{fig:clip_all}
\end{figure}

\subsection{Contrastive Language-Image Pre-training (CLIP)}\label{section:clip}
\textit{Contrastive language-image pre-training (CLIP)} learns a joint embedding space for images and text via a dual-encoder contrastive training paradigm. Specifically, it trains paired image and text encoders to maximize the semantic alignment between matched image-text pairs while minimizing the similarity between unmatched pairs \citep{radford2021learning}. In this shared embedding space, text and images with consistent semantic meanings are mapped to adjacent vector points---for instance, the text prompt ``a photo of a cat" and a real-world cat image yield highly similar embedding representations.
CLIP employs two specialized encoders to project discrete image and text inputs into a unified $D$-dimensional embedding space, as detailed below:
\begin{itemize}
\item Image encoder $\bff_{\text{img}}$: Implemented via a ResNet or vision transformer (ViT) backbone \citep{he2016deep, dosovitskiy2020image}, which encodes an input image $\bx_i^{\text{img}}$ into a compact feature representation.  
\item Text encoder $\bff_{\text{txt}}$: Built upon a standard Transformer architecture, which converts a text prompt $\bx_j^{\text{txt}}$ into a corresponding textual feature vector.
\end{itemize}
Each encoded feature is subsequently projected into the joint multimodal embedding space via a dedicated linear projection layer ($\bW_I$ for images, $\bW_T$ for text) and normalized using $\ell_2$ normalization, formulated as:
$$
\bi_i = \frac{\bW_I \, \bff_{\text{img}}(\bx_i^{\text{img}})}{\normtwo{\bW_I \, \bff_{\text{img}}(\bx_i^{\text{img}})}}, \qquad
\bt_j = \frac{\bW_T \, \bff_{\text{txt}}(\bx_j^{\text{txt}})}{\normtwo{\bW_T \, \bff_{\text{txt}}(\bx_j^{\text{txt}})}}.
$$
This normalization enforces unit vector magnitude for all embeddings, such that $\normtwo{\bi_i } = \normtwo{\bt_j } = 1$. During training, given a batch of $N$ distinct paired image-text samples $\{(\bi_i, \bt_i)\}_{i=1}^N$, CLIP constructs an $N \times N$ pairwise similarity matrix. Thanks to $\ell_2$ normalization, the pairwise dot product directly corresponds to cosine similarity between embeddings (see Figure~\ref{fig:clip_train} for visualization).
The similarity scores are scaled by a learnable temperature parameter $\tau$ (stabilized via exponential reparameterization $\tau = \exp(t)$) to generate classification logits:
$$
\text{logits}_{ij} = \frac{\bi_i \cdot \bt_j}{\tau} = \exp(t)\,(\bi_i \cdot \bt_j).
$$
Within the similarity matrix, diagonal entries $\bi_i \cdot \bt_i$ represent semantically matched positive pairs, while all off-diagonal entries correspond to negative mismatched pairs that are explicitly dissociated during optimization.

The training objective formulates a bidirectional cross-modal classification task. Each row of the similarity matrix serves as an $N$-way classification task for matching images to corresponding text prompts, while each column performs the inverse task of matching text prompts to images. The bidirectional cross-entropy loss functions are defined as follows:
\begin{align}
(\text{Image-to-text direction})\qquad 
\mathcalJ_{\text{i}\to\text{t}} &= -\frac{1}{N}\sum_{i=1}^{N} \ln \frac{\exp(\bi_i \cdot \bt_i / \tau)}{\sum_{j=1}^{N} \exp(\bi_i \cdot \bt_j / \tau)}; \\
(\text{Text-to-image direction})\qquad 
\mathcalJ_{\text{t}\to\text{i}} &= -\frac{1}{N}\sum_{j=1}^{N} \ln \frac{\exp(\bi_j \cdot \bt_j / \tau)}{\sum_{i=1}^{N} \exp(\bi_i \cdot \bt_j / \tau)}.
\end{align}
The overall training loss is computed as the average of the two directional losses:
\begin{equation}
\mathcalJ = \tfrac{1}{2}\left(\mathcalJ_{\text{i}\to\text{t}} + \mathcalJ_{\text{t}\to\text{i}}\right).
\end{equation}
Notably, the supervision signal relies solely on pairwise identity matching, applying symmetric cross-entropy optimization across both image-text and text-image matching axes.

For zero-shot downstream classification tasks with predefined classes $\{c_1, c_2, \ldots, c_K\}$, CLIP generates class-specific text embeddings by encapsulating each class label within a universal prompt template (e.g., ``A photo of a \{object\}"). The normalized text embedding for class $k$ is formulated as:
$$
\bt_k = \frac{\bW_T \, \bff_{\text{txt}}(\text{``a photo of a } c_k\text{''})}{\normtwo{\cdot}}, \quad k = 1,2, \ldots, K.
$$
These $K$ text embeddings function as static classification weights, enabling zero-shot inference without task-specific training data. Performance can be further improved via prompt ensembling, which averages embeddings from multiple diverse prompt templates per class. For an input query image $\bx$, the model first extracts its image embedding $\bi$, then predicts the most semantically consistent class to yield zero-shot classification outputs (illustrated in Figure~\ref{fig:clip_zeroshot}). The classification probability and final prediction are defined as:
\begin{equation}
p(y = k \mid \bx) = \frac{\exp(\bi \cdot \bt_k / \tau)}{\sum_{k'=1}^{K} \exp(\bi \cdot \bt_{k'} / \tau)}, \qquad
\widehaty = \arg\max_{k} \; \bi \cdot \bt_k.
\end{equation}

Benefiting from this robust contrastive pre-training paradigm, CLIP has become the dominant text encoder for state-of-the-art generative models, including Stable Diffusion (SD), DALL-E, and Flux. Its strong open-vocabulary semantic comprehension enables it to interpret complex, fine-grained textual prompts (e.g., ``a watercolor painting of a cyberpunk cat riding a skateboard at sunset") and abstract stylistic concepts (e.g., ``serene", ``retro-futuristic"). Unlike conventional text encoders such as BERT, CLIP is pre-trained on approximately 400 million web-scale image-text pairs, optimized for open-domain, free-form natural language descriptions---making it perfectly compatible with the flexible prompt inputs required for generative visual modeling.
As the foundational backbones of modern image generation, diffusion and flow models require fixed-dimensional conditional signals to guide controlled generation. CLIP outputs fixed-length text embeddings (typically 768 or 1024 dimensions) that can be seamlessly integrated into the cross-attention blocks or U-Net attention layers of generative models (see Section~\ref{section:sd_architecture_unet}). These semantic embeddings serve as precise conditional control signals, guiding diffusion and flow models to prioritize target visual features throughout the iterative denoising/flow process.

\subsection{Embedding the Guidance Variables}\label{section:emb_guidance_var}

To enable precise, controllable generative modeling, we next elaborate on the embedding strategies for core guidance signals---including timesteps, class labels, and text prompts---to produce high-dimensional representations compatible with mainstream neural network architectures.

\paragrapharrow{Embedding time.}
For simplified baseline models, directly concatenating raw scalar timestep values $t$ to network inputs suffices to yield decent training performance. For practical high-performance systems, however, scalar timesteps are typically projected into high-dimensional feature spaces via \textit{Fourier feature embedding}. This design enables the model to capture fine-grained, high-frequency temporal dependencies more accurately throughout the generative process \citep{vaswani2017attention, tancik2020fourier, su2024roformer}. The explicit timestep featurization is formulated as:
\begin{equation}
\texttt{TimeEmb}(t) = \sqrt{\frac{2}{D}} 
\begin{bmatrix} 
\cos( \sigma_1 t), \, \ldots , \,\cos( \sigma_{D/2} t) ,\, \sin( \sigma_1 t) ,\, \ldots ,\, \sin( \sigma_{D/2} t) 
\end{bmatrix}^\top,
\end{equation}
where the frequency parameters $\sigma_i$ are spaced logarithmically and geometrically across low-to-high frequency ranges, defined as:
\begin{equation}
\sigma_i = 2\pi \sigma_{\min} \left( \frac{\sigma_{\max}}{\sigma_{\min}} \right)^{\frac{i-1}{D/2 - 1}}, \quad i = 1, \ldots, D/2.
\end{equation}
This Fourier-based timestep embedding is a standard design in modern generative models, though alternative formulations are also valid. A key advantage of this specific implementation is that it produces strictly unit-normalized $D$-dimensional embeddings ($\normtwo{\texttt{TimeEmb}(t)} = 1$), guaranteed by the trigonometric identity $\sin^2(\cdot) + \cos^2(\cdot) = 1$.

\paragrapharrow{Embedding class labels.} 
For discrete classification-guided generation, where the raw conditional input $\by_{\text{raw}} \in \sY \triangleq \{0,1, \ldots, N\}$ denotes a categorical class label, the standard practice is to learn a dedicated embedding vector for each of the $N+1$ possible label values. The resulting embedding vector serves as the guidance variable $\by$ for model conditioning.
These class embedding parameters are optimized end-to-end alongside the backbone neural network parameters $\btheta$. They provide conditional guidance for the learned noise predictor, vector field, or score function---corresponding to $\bepsilon_t^\btheta$, $\vf^\btheta_t$, and $\bs^\btheta_t$ for diffusion, flow matching, and score-based generative frameworks, respectively (see Algorithms~\ref{alg:diffusion_training}, \ref{alg:flow_matching}, and \ref{alg:score_gauss_path}).

\paragrapharrow{Embedding textual input.}
Unlike discrete class labels, raw text prompts require sophisticated embedding pipelines, which predominantly rely on frozen pre-trained multimodal models. These models encode discrete natural language inputs into continuous semantic embeddings that preserve textual contextual and descriptive information.
As introduced in Section~\ref{section:clip}, CLIP \citep{radford2021learning} is the most widely adopted text encoder for this task. Pre-trained via contrastive learning, CLIP constructs a unified image-text embedding space that aggregates semantically aligned image-text pairs and separates mismatched cross-modal samples. We adopt the frozen pre-trained CLIP model to generate text conditional embeddings:
\begin{equation}
\by = \texttt{TextEmb}_{\text{CLIP}}(\by_{\text{raw}}) \in \real^{D_{\text{CLIP}}}.
\end{equation}
Global average pooling of sequence-level text features yields a single holistic embedding vector, which may discard fine-grained sequential textual details. To retain full prompt sequence information, researchers additionally employ pre-trained Transformer encoders to produce sequence-aware text embeddings. Combining multiple complementary pre-trained embedding modules is also a common strategy to integrate diverse semantic priors and boost generation quality \citep{polyak2024movie}.
In this work, we uniformly define the output of sequence-level text embedding models as:
\begin{equation}
\texttt{TextEmb}_{\text{SEQ}}(\by_{\text{raw}}) \in \textcolor{black}{\real^{M \times D_\tau}},
\end{equation}
where $M$ denotes the maximum text sequence length and $D_\tau$ denotes the dimension of per-token textual features. Representative practical implementations include \textit{CLIP ViT-L/14} for the Stable Diffusion v1.x series \citep{radford2021learning} and \textit{CLIP ViT-H/14} for the Stable Diffusion v2.x series \citep{ilharco_gabriel_2021_5143773}.~\footnote{Comprehensive ablation studies on text encoder selection for generative models are provided in the Imagen paper \citet{saharia2022photorealistic}.}

\index{Cross-attention}
\index{Self-attention}
\index{Attention}
\subsection{Attention Mechanism}

\textit{Cross-attention} serves as a fundamental attention primitive that enables dynamic cross-sequence information fusion and conditional guidance, constituting a core component of multimodal architectures including the Stable Diffusion v1.x series \citep{rombach2022high} and \textit{vision transformers (ViTs)} \citep{dosovitskiy2020image}, alongside mainstream text-to-image generative frameworks. Unlike \textit{self-attention} \citep{vaswani2017attention}, which exclusively models intra-sequence dependencies within a single feature space, cross-attention builds semantic alignment across heterogeneous modalities. It enables one feature representation to actively query, retrieve, and aggregate contextually relevant information from a distinct auxiliary feature sequence, laying the foundation for multimodal conditional generation.
Formally,  cross-attention computes adaptive feature fusion based on three linearly projected feature subspaces: \textit{Queries} ($\bQ$), \textit{Keys} ($\bK$), and \textit{Values} ($\bV$), with distinct functional roles for source and target inputs:
\begin{itemize}
\item \textit{Query projection (target modality)}: The primary target input, which requires external conditional guidance, is projected to query features. These queries encode the contextual demands and semantic requirements of individual elements within the target representation.

\item \textit{Key-Value projection (guidance modality)}: The secondary conditional guidance input is projected into key and value feature spaces. Keys characterize the semantic attributes of each guidance element for similarity matching, while Values store the intrinsic feature information to be transferred to the target representation.

\item \textit{Adaptive attention weighting}: The pairwise similarity between Queries and Keys is computed and scaled by the square root of the feature dimension to stabilize numerical gradients. A row-wise softmax function normalizes raw similarity scores into a valid probability distribution, forcing the model to prioritize the most semantically consistent guidance components for each target element.

\item \textit{Contextual feature aggregation}: The final conditioned output is obtained via a weighted summation of Value features using the normalized attention weights. This mechanism delivers spatially adaptive, semantically precise conditioning, wherein each position in the target representation selectively absorbs information from the most relevant segments of the guidance signal.

\item \textit{Training stability design}: Integrated with residual connections and layer normalization (illustrated in Figure~\ref{fig:unet_sd_simplified}), cross-attention maintains stable gradient propagation and intact feature representation in deep neural networks. This structured information routing empowers text-to-image models to faithfully interpret textual prompts and generate semantically coherent visual content, rendering cross-attention indispensable for controllable, high-fidelity generative modeling.
\end{itemize}
We next elaborate on the mathematical formulation of scaled dot-product attention---covering both self-attention and cross-attention variants---which endows generative networks with the capability to dynamically weight spatial and modal contextual information for precise conditional image synthesis.

\paragrapharrow{Scaled dot-product attention mechanism.} iven query, key, and value matrices $\bQ \in \real^{N \times D_H}$, $\bK \in \real^{M \times D_H}$, and $\bV \in \real^{M \times D_H}$, the standard scaled dot-product attention operation is defined as:
\begin{equation}
\texttt{Attention}(\bQ, \bK, \bV) = \underbrace{\texttt{softmax}\left( \frac{\bQ\bK^\top}{\sqrt{D_H}} \right)}_{\text{Atten. matrix }\bA} \bV \in \real^{N \times D_H}, 
\end{equation}
where the softmax function is applied independently to each row of the similarity matrix. In cross-attention pipelines, $N$ denotes the spatial dimension of the target visual representation (e.g., latent image pixels), while $M$ denotes the sequence length of the conditional guidance input (e.g., text prompt tokens).

The full computation pipeline proceeds as follows. First, pairwise compatibility scores between all query-key pairs are calculated to quantify cross-modal relevance:
$$
\bS = \frac{\bQ \bK^\top}{\sqrt{D_H}}\in \real^{N \times M}.
$$
In DiT and diffusion-based generative models, each entry $s_{ij}$ measures the semantic relevance of the $j$-th text token to the $i$-th spatial pixel position in the visual latent space. The scaling factor $\sqrt{D_H}$ mitigates gradient vanishing and explosion by constraining the magnitude of dot-product similarity scores during training.
Subsequently, row-wise softmax normalization is applied along the guidance sequence dimension $M$ for every spatial position $i$:
$$
a_{ij} = \frac{\exp(s_{ij})}{\sum_{k=1}^{M} \exp(s_{ik})}.
$$
This produces the normalized attention matrix $\bA = [a_{ij}] \in \real^{N \times M}$, with each row satisfying $\sum_{j=1}^{M} a_{ij} = 1$. Notably, the attention distribution $\bA_{i,:}$ is \textbf{spatially variant across visual positions}: pixels corresponding to foreground objects assign dominant weights to semantically matching text tokens, whereas background pixels prioritize background-related textual descriptors.
Finally, context-aware feature aggregation is performed by weighting value vectors with the adaptive attention coefficients:
$$
\bZ_{attn} = \bA \bV\in {\real^{ N \times D_H}}.
$$
The \textbf{contextualized feature vector} for each individual spatial location $i$ is explicitly formulated as:
$$
\bz_i = \sum_{j=1}^{M} a_{ij} \bv_j.
$$
This operation yields a uniquely contextualized feature embedding for every pixel position, where each visual feature is a weighted fusion of textual semantics dynamically adapted to local visual content.

In diffusion generative architectures (detailed in Section~\ref{section:sd_architecture_unet}), the cross-attention map $\bA$ is derived from interactions between latent visual queries ($\bQ$) and text guidance keys ($\bK$) sampled from pre-trained textual embedding spaces. This mechanism functions as a soft, dynamic selective gating system driven by query-key matching. It adaptively modulates the contribution of conditional value features ($\bV$) throughout the generative inference trajectory. Rather than accepting static global conditioning, the generative model actively queries and retrieves step-specific semantic context from the textual guidance space, enabling precise, prompt-aligned visual synthesis.

\index{Multi-head attention}
\paragrapharrow{Multi-head attention.} 
Let $H$ denote the total number of attention heads and $D_H = \frac{D_q}{H}$ denote the feature dimension allocated to each individual head. For each head $h \in \{1,2, \ldots, H\}$, we learn dedicated projection matrices $\bW_Q^{(h)}\in\real^{D_q\times D_H}$ for queries, as well as $\bW_K^{(h)}, \bW_V^{(h)} \in \real^{D_{kv} \times D_H}$ for keys and values, respectively. The computation for a single attention head is defined as:
\begin{equation}
\texttt{Head}_h(\bx, \bz) = \texttt{Attention}(\bx\bW_Q^{(h)}, \bz\bW_K^{(h)}, \bz\bW_V^{(h)}) 
\in\real^{N\times D_H}, 
\end{equation}
where the \textit{source sequence} $\bz$ supports two distinct conditioning settings corresponding to self-attention and cross-attention:
\begin{align*}
\bz &= \bx\in\real^{N\times D_q}, \qquad (\text{self-attention on patches with } D_q=D_{kv}) \\
\text{or } \bz &= \by\in\real^{M\times D_{kv}}. \qquad (\text{cross-attention to the prompt or guidance information})
\end{align*}
Consistent with the convention adopted throughout this book, all input and latent variables are formulated in vector-based form, even when the underlying representations are structured as matrices or tensors. 
The outputs of all individual heads are concatenated and projected via a unified output projection matrix $\bW_O \in \real^{D_q \times D_q}$ to produce the final multi-head attention output:
\begin{subequations}\label{equation:muthead_attn}
\begin{equation}
\texttt{MultiHeadAttn}(\bx, \bz) 
= \texttt{Concat}\big(\texttt{Head}_1(\bx, \bz), \ldots, \texttt{Head}_H(\bx, \bz)\big) \cdot\bW_O 
\in \real^{N \times D_q}. 
\end{equation}
An equivalent alternative formulation folds the final projection operation into each individual head branch. Specifically, each head is equipped with a dedicated output projection $\bW_O^{(h)}\in\real^{D_H\times D_q}$, and the projected head outputs are summed to aggregate multi-scale contextual information:
\begin{equation}
\texttt{MultiHeadAttn}(\bx, \bz) 
= \sum_{h=1}^H \texttt{Head}_h(\bx, \bz) \cdot \bW_O^{(h)}
\in \real^{N \times D_q}.
\end{equation}
\end{subequations}

\index{QK-normalization}
\paragrapharrow{QK-normalization.}
Standard scaled dot-product attention calculates attention logits as $\texttt{softmax}(\bQ\bK^\top / \sqrt{D_H})$. However, this formulation suffers from inherent training instability, particularly in large-scale Transformer-based architectures and under mixed-precision or low-precision training regimes. As training progresses, the magnitudes of query and key feature entries tend to grow excessively due to weight drift. This causes unbounded growth of query-key dot products, leading to softmax saturation where a single logit dominates the entire probability distribution and gradients of all other entries vanish. In severe cases, attention logit overflow occurs, triggering abrupt loss spikes and training divergence. Such instability is widely observed in large ViT and DiT-style generative models (see Section~\ref{section:dits}).
\textit{QK-normalization} is an effective stabilization technique that mitigates these issues by normalizing per-head query and key features prior to dot-product computation \citep{henry2020query}. The modified attention formulation is defined as:
\begin{equation}\label{equation:qknormalization}
\texttt{Attention}'(\bQ, \bK, \bV) 
= \texttt{softmax}\left( \frac{\texttt{Norm}(\bQ)\texttt{Norm}(\bK)^\top \cdot\gamma}{\sqrt{D_H}} \right) \bV \in \real^{N \times D_H}.
\end{equation}
In practical implementations, \texttt{RMSNorm} is typically applied along the per-head feature dimension, paired with a learnable scaling factor $\gamma$ for adaptive magnitude adjustment. By constraining the norm of query and key features, QK-normalization decouples attention logit magnitudes from drifting projection weight scales. The resulting logit values depend solely on the cosine similarity between query and key feature directions, modulated by the learnable scale parameter. This design tightly confines logits within a stable numerical range, effectively eliminating softmax saturation and numerical overflow. Consequently, it enables stable training with larger learning rates and lower numerical precision, avoiding catastrophic loss explosion in large-scale generative  models.

\section{Large-Scale Image or Video Generative Models}\label{section:large_gen}
We now focus on designing scalable neural network architectures for flow and diffusion models tailored to image-based modalities, including static images and dynamic videos. Specifically, we elaborate on the practical implementation of parameterized modeling functions for guided generative tasks, namely the noise vector, vector field, or score function $\bp^\btheta_t(\bz\mid\by)$. Here, $\bp^\btheta_t$ corresponds to $\bepsilon_t^\btheta$, $\vf^\btheta_t$, and $\bs^\btheta_t$ for noise prediction, vector field regression, and score matching formulations, respectively (see Algorithms~\ref{alg:diffusion_training}, \ref{alg:flow_matching}, and \ref{alg:score_gauss_path}), with $\btheta$ denoting the trainable model parameters.
The neural network in these frameworks requires three mandatory inputs: a latent vector $\bz \in \real^D$, a conditioning guidance variable $\by \in \sY$, and a time parameter $t \in [0, 1]$ (or a discrete timestep $t\in\{1,2,\ldots,T\}$), alongside a single output: the predicted vector $\bp^\btheta_t(\bz\mid\by) \in \real^D$. 

For low-dimensional data distributions, such as the MNIST example \citep{lecun2010mnist}, a standard \textit{multi-layer perceptron (MLP)}, also referred to as a \textit{fully connected neural network}, is sufficient to parameterize $\bp^\btheta_t(\bz\mid\by)$ \citep{lecun2015deep, goodfellow2016deep}.
In this simplified low-dimensional setting, the forward pass of $\bp^\btheta_t(\bz\mid\by)$ operates by concatenating the three input components ($\bz$, $\by$, and $t$) and feeding the combined vector into an MLP for prediction. However, MLPs lack the inductive biases required to model complex high-dimensional distributions inherent to images, videos, and biological protein data. For such real-world data, domain-specific specialized architectures are universally adopted instead. The remainder of this section centers on image generation, with all discussed principles extendable to video generation tasks.
We introduce and analyze three dominant architectural paradigms for conditional generative modeling: the \textit{U-Net} along with its variant using \textit{ControlNet} \citep{ronneberger2015u, zhang2023adding}, the \textit{diffusion transformer (DiT)} \citep{peebles2023scalable, ma2024sit, chen2024pixartt}, and the \textit{multimodal DiT (MM-DiT)} \citep{esser2024scaling}.

\subsection{U-Net}\label{section:sd_architecture_unet}

\begin{figure}[h]
\centering
\includegraphics[width=0.99\textwidth]{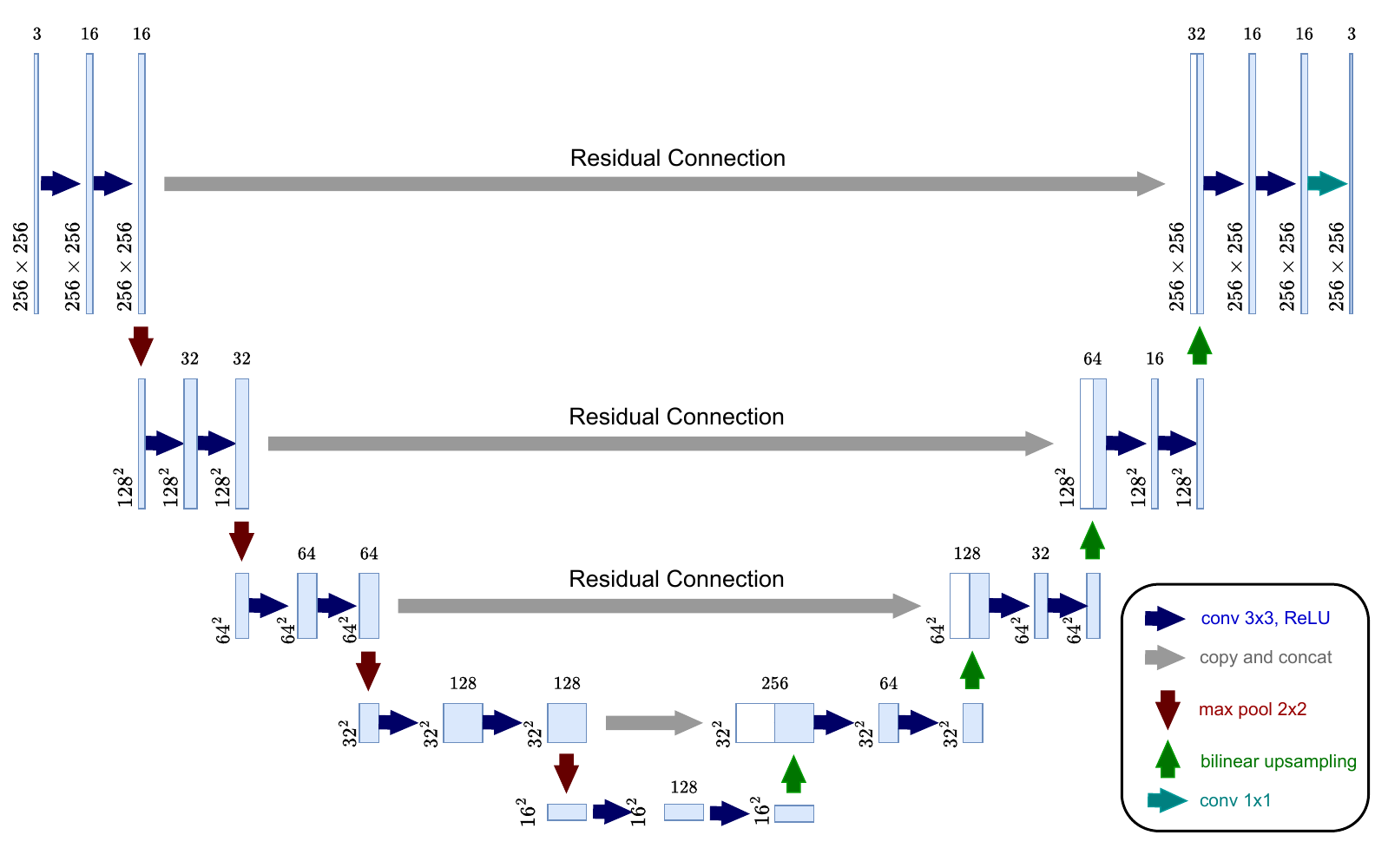} 
\caption{The original U-Net architecture {(example for $16\times 16$ pixels in the lowest resolution)}. Each blue
box corresponds to a multi-channel feature map. The number of channels is denoted
on top of the box. The h-w-size is provided at the lower-left edge of the box. White
boxes represent copied feature maps. The arrows denote the different operations.
Adapted from \citet{ronneberger2015u}.
}
\label{fig:unet}
\end{figure}

The U-Net architecture is a specialized convolutional neural network (CNN) composed of a \textit{contracting (encoding) path} and a symmetric \textit{expansive (decoding) path} built from stacked neural network blocks, originally proposed for semantic segmentation \citep{ronneberger2015u}. During encoding, the contracting path progressively reduces spatial resolution while hierarchically extracting multi-scale discriminative features. In the expansive decoding path, low-resolution latent features are upsampled and fused with corresponding high-resolution features from the contracting path via skip connections (residual connections).
These skip connections concatenate fine-grained spatial information from shallow encoder layers with high-level semantic information from deep decoder layers, compensating for spatial detail lost during downsampling. The architecture derives its name from the symmetric U-shaped structure formed by its encoder and decoder modules (see Figure \ref{fig:unet}).

A defining advantage of U-Net is its image-to-image mapping property: both input and output are spatial feature maps with arbitrary channel dimensions. This property makes U-Net uniquely suitable for parameterizing core generative model mappings. Specifically, it can implement the noise prediction function $\bz \mapsto \bepsilon^\btheta_t(\bz\mid\by)$ in DDPMs and the vector field regression $\bz \mapsto \vf^\btheta_t(\bz\mid\by)$ in flow-based models. For fixed condition $\by$ and timestep $t$, the network maintains consistent spatial dimensions between input and output. Owing to this flexibility and robust feature-learning capability, U-Nets have become the backbone architecture in early diffusion model research, including mainstream frameworks such as \textit{SD v1.x}, \textit{DALL-E 2}, \textit{Imagen}, and \textit{Kandinsky 2} \citep{ho2020denoising, rombach2022high, ramesh2022hierarchical, saharia2022photorealistic, razzhigaev2023kandinsky}.

The detailed U-Net structure is visualized in Figure \ref{fig:unet}. Note that practical implementation details may vary across literature and engineering deployments, while the core architectural paradigm remains consistent. The full framework comprises three components: a left-side contracting path consisting of sequential \textit{encoder blocks} $\mathcalE_i$, a symmetric right-side expansive path composed of corresponding \textit{decoder blocks} $\mathcalD_i$, and an intermediate latent processing module termed the \textit{midcoder} $\mathcalM$ in this work.
The contracting path follows standard convolutional downsampling design. Each encoder block iteratively applies two successive $3\times 3$ unpadded convolutions, each followed by a rectified linear unit (ReLU) activation. A $2\times 2$ max-pooling operation with a \texttt{stride} of 2 is then performed for spatial downsampling. Critically, the number of feature channels is doubled at each downsampling stage to enrich semantic representation while compressing spatial dimensions.

Each decoder block in the expansive path executes a multi-step feature reconstruction process. First, feature maps are upsampled via a $2\times 2$ transposed convolution (up-convolution). Next, the upsampled latent features are concatenated with spatially aligned feature maps from the corresponding encoder layer via skip connections,  followed by quartering the number of feature channels. 
When spatial mismatches occur between paired feature maps, center cropping is applied to ensure dimension consistency before concatenation. The fused features are further refined by two consecutive $3\times 3$ convolutions with ReLU activation. Finally, a $1\times 1$ convolution layer projects the refined feature maps to the target channel dimension, adapting the network for segmentation classification, diffusion noise prediction, or flow vector field estimation tasks.

We illustrate the complete U-Net forward propagation pipeline using the $256\times256$ input example in Figure \ref{fig:unet}, with input dimensions defined as $(C_{\text{input}}, H, W) = (3, 256, 256)$ for the input latent representation $\bz_t \in \real^{3 \times 256 \times 256}$:
\begin{align*}
\bz_t^{\text{input}} &\in \real^{3 \times 256 \times 256} ;&& \quad  \text{(Input to the U-Net)} \\
\bz_t^{\text{latent}} = \mathcalE(\bz_t^{\text{input}}) &\in \real^{128 (\uparrow) \times 16(\downarrow) \times 16(\downarrow)}; &&\quad  \text{(Pass through encoders to obtain latent)} \\
\bz_t^{\text{latent}} = \mathcalM(\bz_t^{\text{latent}}) &\in \real^{128(=) \times 16(=) \times 16(=)} ;&&\quad  \text{(Pass latent through midcoder)} \\
\bz_t^{\text{output}} = \mathcalD(\bz_t^{\text{latent}}) &\in \real^{3(\downarrow) \times 256(\uparrow) \times 256(\uparrow)}. &&\quad  \text{(Pass through decoders to obtain output)}
\end{align*}
As demonstrated in the forward pass, encoder processing progressively increases feature channel counts while reducing spatial height and width, trading spatial granularity for high-dimensional semantic context. In practical implementations, the raw input is typically processed by a pre-encoding convolutional block to expand channel dimensions before entering the first encoder layer. In addition, dense skip connections are universally deployed between encoder and decoder blocks to preserve low-level spatial details throughout hierarchical feature propagation.

\begin{figure}[h]
\centering
\includegraphics[width=0.9\textwidth]{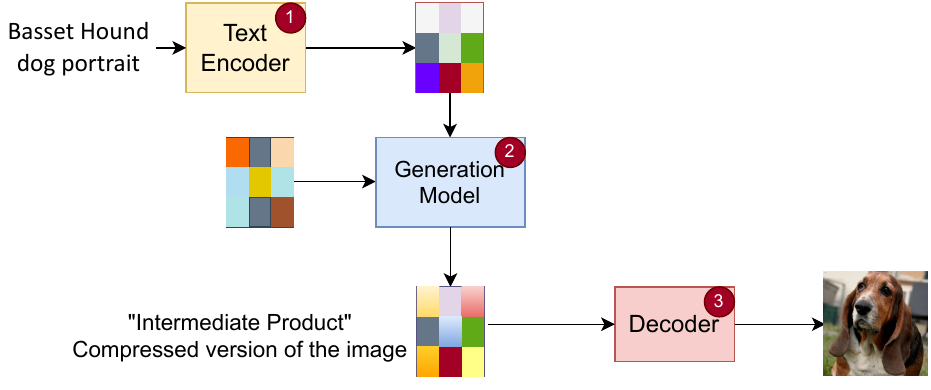} 
\caption{Three components in generative model architectures. The middle component can generate both a low-resolution image that can be recognized by human (e.g., in Imagen \citep{saharia2022photorealistic}) or a noise image that cannot be recognized (e.g., in SD v1.x series \citep{rombach2022high}).
The ``decoder" can be trained without lablelled data: i). it decodes smaller (that can be recognized by a human) to larger image (for example in Imagen); ii). it decodes latent representation to image (e.g., VAE;  for example in SD). 
The ``generation model"  requires (text-image) paired data; e.g., LAION5B \citep{schuhmann2022laion}.
}
\label{fig:SD_three_comps}
\end{figure}

As mentioned previously, the {SD v1.x series}, {DALL-E 2}, {Imagen}, and {Kandinsky 2} all adopt the U-Net architecture as their core backbone for generative modeling. The mathematical inference pipelines of these models are compared and summarized below:
\begin{itemize}[leftmargin=2pt, labelsep=5pt]
\item  \textit{Stable Diffusion (SD) v1.x \citep{rombach2022high}}:
$$
\text{Text } \by \xrightarrow{\text{CLIP ViT-L/14 Text}} \widetildeby
\xrightarrow{\text{U-Net (Latent Diffusion, cross-attn)}} \bz_{t}
\xrightarrow{\text{VAE Decoder}} \text{Image } \bx.
$$
This model operates as a pure latent diffusion model without a prior module, as introduced in Section~\ref{section:ldm_intro}. The generation process initiates with random latent noise $\bz_t$, as illustrated in Figure~\ref{fig:SD_three_comps}. The U-Net progressively denoises the latent features, guided by text embeddings injected through cross-attention conditioning.

\item \textit{DALL-E 2 (unCLIP) \citep{ramesh2022hierarchical}}:
$$
\small
\text{Text } \by \xrightarrow{\text{CLIP Text}} \bz_t
\xrightarrow{\text{Prior (Transformer)}} \bz_i
\xrightarrow{\text{U-Net Decoder (Diffusion)}} \bx_{64}
\xrightarrow{\text{U-Net Upsamplers}} \text{Image } \bx_{1024}.
$$
DALL-E 2 performs pixel-space diffusion with an additional transformer-based prior module. In this pipeline, $\bz_t$ denotes the CLIP text embedding and $\bz_i$ denotes the CLIP image embedding. High-resolution image generation is implemented via two cascaded U-Net super-resolution stages, which upscale images from $64\times64$ to $256\times256$ and finally to $1024\times1024$ resolution.

\item \textit{Imagen \citep{saharia2022photorealistic}}:
$$
\text{Text } \by \xrightarrow{\text{T5-XXL}} \bc
\xrightarrow{\text{U-Net (Diffusion)}} \bx_{64}
\xrightarrow{\text{Efficient U-Net SR}} \bx_{256}
\xrightarrow{\text{Efficient U-Net SR}} \text{Image } \bx_{1024}.
$$
Imagen conducts the entire diffusion process directly in pixel space, requiring no prior module or latent space compression. Its base diffusion U-Net is conditioned on T5-XXL text embeddings $\bc$ \citep{raffel2020exploring}. Two successive Efficient U-Net super-resolution modules are then applied to iteratively upscale the low-resolution generated image to full $1024\times1024$ resolution.

\item \textit{Kandinsky (v2.x) \citep{razzhigaev2023kandinsky}}:
$$
\text{Text } \by \xrightarrow{\text{CLIP Text {+ XLMR}}} \bz_t
\xrightarrow{\text{Prior (Trans.)}} \bz_i
\xrightarrow{\text{U-Net (Lat. Diff.)}} \bz_{\text{lat}}
\xrightarrow{\text{MoVQ Decoder}} \text{Image } \bx.
$$
Kandinsky 2.x integrates the core designs of DALL-E 2 and Stable Diffusion. It inherits DALL-E 2's transformer prior to map CLIP text embeddings $\bz_t$ to CLIP image embeddings $\bz_i$, while implementing latent-space diffusion similar to Stable Diffusion. The final high-fidelity image is reconstructed from latent features via a MoVQ vector-quantized decoder.

\end{itemize}

\begin{figure}[h]
\centering
\includegraphics[width=\textwidth]{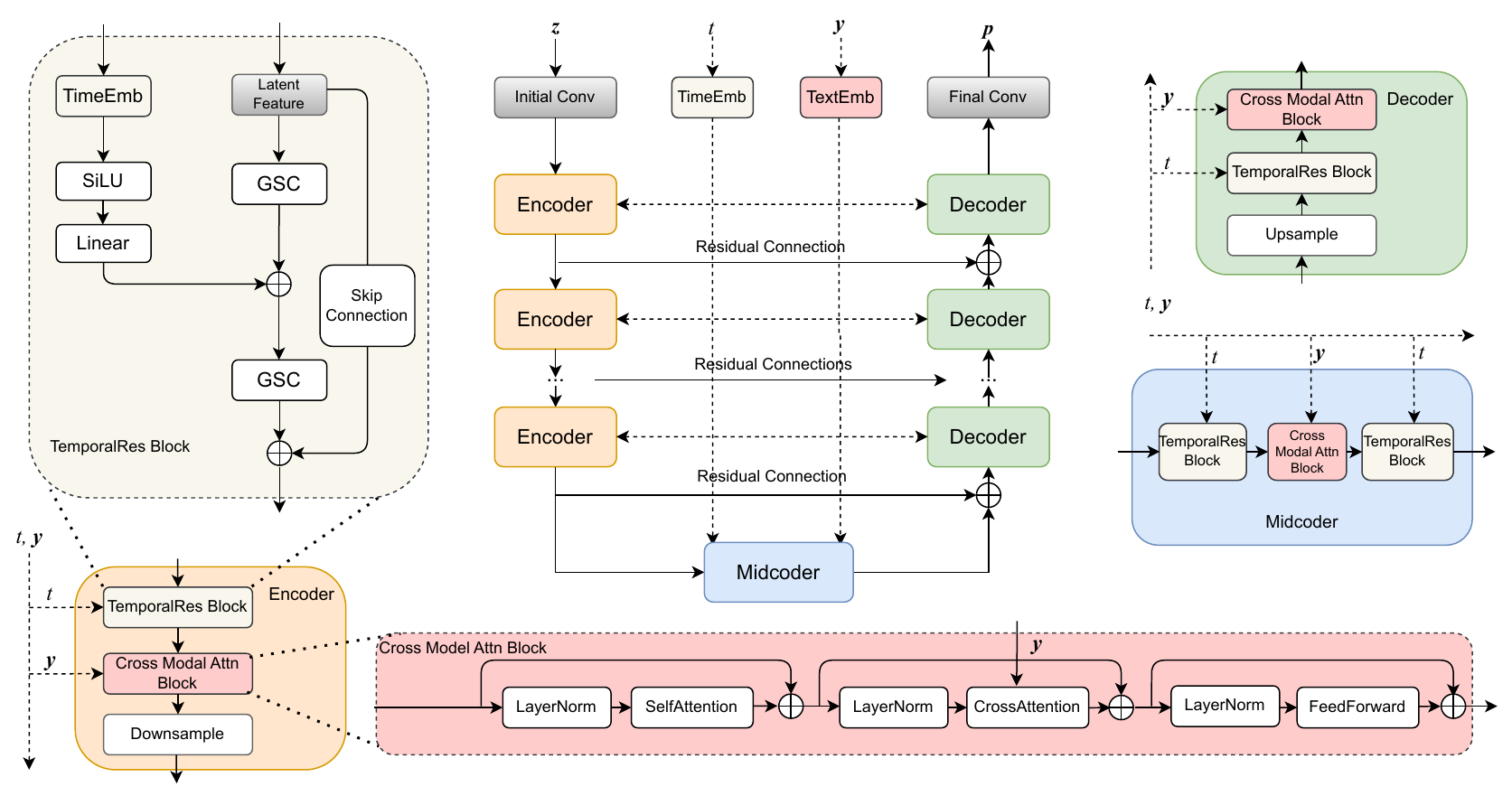}
\caption{
Conceptual illustration of a simplified U-Net architecture used in the SD v1.x series, adapted from the original U-Net framework in Figure~\ref{fig:unet}. The notation ``\texttt{GSC}" refers to the sequential composition of \texttt{GroupNorm}, \texttt{SiLU}, and \texttt{Conv} layers.
White blocks represent individual basic building blocks, while colored blocks represent composite structures formed from multiple simpler blocks (except the input latent or embedding vectors). 
Note that SD v1.x interleaves the residual and transformer
(self- and cross-attention) blocks only at certain resolutions---the lower-resolution
stages (and the Midcoder) carry the attention blocks, while the highest-resolution
stages are often residual-only; the figure  describe a stage that contains both.
}
\label{fig:unet_sd_simplified}
\end{figure}
\subsubsection{Case Study: Stable Diffusion 1}

Stable Diffusion (SD), Imagen, DALL-E, and Kandinsky are series of state-of-the-art image generation models. 
These frameworks were among the earliest to adopt large-scale latent diffusion paradigms for visual synthesis.
Specifically, SD v1.x, v2.x, and XL series, alongside Imagen, DALL-E 2, and Kandinsky 2, are fundamentally built on the U-Net architecture. This widespread adoption stems from four core inherent strengths of U-Net, outlined as follows:
\begin{itemize}
\item \textit{Encoder compression capability}: As a foundational function of the encoder module, downsampling is applied to input images to extract compact high-dimensional feature representations. The resulting latent features occupy a substantially smaller spatial dimension than the original input, achieving effective feature compression. This mechanism perfectly matches the latent space design adopted by Stable Diffusion, LDMs, and DDPMs.

\item \textit{Decoder denoising capability}: The decoder module preserves its native upsampling and reconstruction functionality, which serves as the core structural support for the iterative denoising process in diffusion-based image generation.

\item \textit{Architectural simplicity, stability and efficiency}: U-Net features a streamlined, lightweight structural design that robustly accommodates the iterative denoising workflow of DDPMs. Its reliable operational stability and computational efficiency effectively sustain the end-to-end image generation pipeline.

\item \textit{High encoder-decoder compatibility}: The symmetric encoder-decoder paradigm of U-Net delivers excellent extensibility and compatibility. It can be seamlessly integrated with advanced modules such as Transformer-based attention mechanisms, and is adaptable to both image segmentation and generative modeling tasks, making it suitable for generation diffusion model optimization.

\end{itemize}
Nearly all modern U-Net variants inherit the above core architectural characteristics. 
Notably, the baseline architecture discussed previously adopts a fully convolutional design, while mainstream U-Net variants commonly integrate attention layers within encoder and decoder blocks to enhance global feature modeling. This subsection briefly elaborates on the architectural design of the SD v1.x series.
In the SD v1.x architecture, all conditional guidance signals (i.e., timestep $t$ and text prompt $\by$) are processed through two independent conditioning pathways. The timestep value is mapped to a high-dimensional embedding vector to modulate the feature learning of each residual block. Meanwhile, the input text prompt is encoded into a semantic sequence, which is subsequently processed via cross-attention layers.
The corresponding embedding formulations are defined as:
$$
\widetildebt = \texttt{MLP}\big(\texttt{TimeEmb}(t)\big) \in \real^{D},
\qquad
\widetildeby \in \textcolor{black}{\real^{M \times D_\tau}}.
$$
For the SD v1.x series, the fixed text token length is set to $M=77$.
Timestep conditioning is implemented via additive embedding within residual blocks (yellow components in Figure~\ref{fig:unet_sd_simplified}), whereas text semantic conditioning is realized through cross-attention mechanisms (red components in Figure~\ref{fig:unet_sd_simplified}), forming two distinct conditioning paradigms in the model.

\paragrapharrow{TemporalRes block.}
For a residual block taking a input feature map $\bz \in \real^{N\times D}$ (assuming the tensor is flattened to a matrix),
timestep information is injected via linear projection and elementwise addition, following the formulations below:
\begin{subequations}
\begin{align}
\bh &= \texttt{GSC}(\bz) \triangleq \texttt{Conv}\big(\texttt{SiLU}(\texttt{GroupNorm}(\bz))\big); \\
\bh &\leftarrow \bh + \texttt{Linear}\big(\texttt{SiLU}(\widetildebt)\big); \\
\texttt{TemporalResBlock}_t(\bz) &= \texttt{Skip}(\bz) + \texttt{GSC}(\bh),
\end{align}
\end{subequations}
where $\texttt{Skip}(\cdot)$ denotes an identity mapping when the input and output channel dimensions are consistent; otherwise, a $1\times1$ convolution layer is applied for channel dimension alignment.

\paragrapharrow{Cross model attention  block.}
At each spatial resolution, the temporal residual block is followed by a cross-model attention block, which injects text semantic guidance into latent features through cross-attention operations. The detailed computation pipeline is formulated as:
\begin{subequations}\label{equation:unet_crossattn}
\begin{align}
\bz  &\leftarrow \texttt{TemporalResBlock}_t(\bz); \\
\bz   &\leftarrow \bz +  \texttt{MultiHeadAttn}(\texttt{LayerNorm}(\bz), \texttt{LayerNorm}(\bz)); \\
\bz &\leftarrow \bz + \texttt{MultiHeadAttn}\big(\texttt{LayerNorm}(\bz),\,
\widetildeby\big); \\
\texttt{CrossModelAttnBlock}_{\by} &\leftarrow \bz + \texttt{PointwiseFeedForward}\big(\texttt{LayerNorm}(\bz)\big).
\end{align}
\end{subequations}
In this configuration, cross-attention constructs queries from normalized latent feature representations and retrieves corresponding keys and values from precomputed text embeddings, following the multi-head attention mechanism illustrated in \eqref{equation:muthead_attn}.
The core advantage of cross-attention mechanisms for conditional generation lies in their ability to enable dynamic, content-aware feature fusion between the generative diffusion process and external text guidance signals. Unlike static feature concatenation or fixed additive modulation strategies, cross-attention implements a differentiable, query-driven interaction paradigm. This allows the model to adaptively extract, weight, and integrate task-relevant textual contextual information for each spatial position and feature token, achieving fine-grained conditional control over image generation.

\paragrapharrow{Composite stage blocks.}
The contracting (encoder) path is constructed by stacking sequential modules: one or more pairs of temporal residual blocks and cross-model attention blocks, followed by a strided convolutional downsampling operation. The overall encoder transformation is defined as:
\begin{subequations}
\begin{equation}
\texttt{Encoder}(\bz) =
\texttt{DownSample}\Big(\big[\texttt{CrossModelAttnBlock}_{\by}\circ\texttt{TemporalResBlock}\big](\bz)\Big),
\end{equation}
where $\texttt{DownSample}(\cdot)$ represents the spatial downsampling operation. 
The network middle layer adopts a symmetric sandwich structure, with one cross-model attention block inserted between two temporal residual blocks, formulated as:
\begin{equation}
\texttt{Midcoder}(\bz) =
\texttt{TemporalResBlock}\circ\texttt{CrossModelAttnBlock}_{\by}\circ\texttt{TemporalResBlock}\,(\bz). 
\end{equation}
The expansive (decoder) path mirrors the structural design of the encoder path. Before executing residual and attention computations at each stage, the decoder first concatenates skip-connected features from the corresponding encoder layer at the same resolution, followed by upsampling via interpolation and convolution refinement. The decoder operation is defined as:
\begin{equation}
\texttt{Ddecoder}(\bz, \bz') =
\texttt{Up}\Big(\big[\texttt{CrossModelAttnBlock}_{\by}\circ\texttt{TemporalResBlock}\big]
\big(\,[\bz \,\|\, \bz'\,]\,\big)\Big), 
\end{equation}
\end{subequations}
where $\texttt{Up}(\cdot)$ denotes the spatial upsampling operation, and $[\cdot\,\|\,\cdot]$ denotes channel-wise feature concatenation.

\paragrapharrow{Full forward pass.} 
The complete U-Net forward propagation follows a standard U-shaped hierarchical paradigm, consisting of encoder downsampling, midcoder feature transformation, and decoder upsampling stages. Cross-resolution skip connections fuse high-fidelity encoder features into the decoder to compensate for spatial detail loss during downsampling. The overall forward process is summarized as:
$$
\bz_t \;\xrightarrow{\text{Encoder}\,\downarrow}\;
\xrightarrow{\text{Midcoder}}\;
\xrightarrow{\text{Decoder}\,\uparrow\,(\oplus\,\text{skips})}\;
\bp^\btheta_t(\bz_t \mid \by).
$$
See Figure~\ref{fig:unet_sd_simplified} for an illustration.
The final prediction is used for the prediction of the noise added to the original representation (or for the prediction of the vector field in a flow-based model).

\index{ControlNet}
\subsection{Guidance with ControlNet}
\begin{figure}[htbp]
\centering
\includegraphics[width=0.99\textwidth]{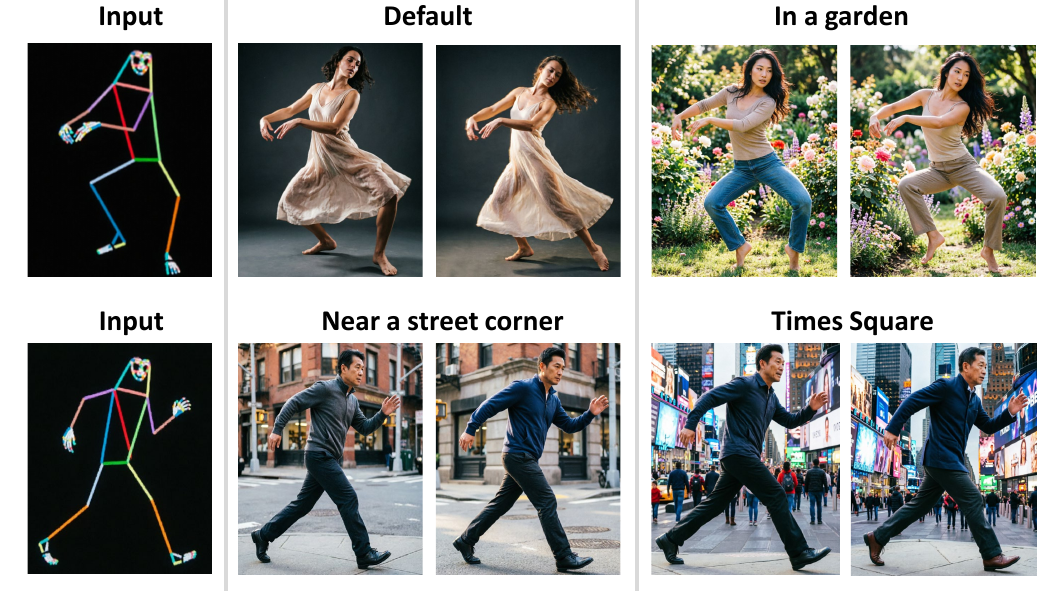}
\caption{Using human pose maps to control image generation, producing character images that match the pose diagrams.}
\label{fig:controlnet_pose}
\end{figure}

A central challenge in conditional image generation lies in achieving precise, controllable generation outputs. 
Previous chapters introduced guidance strategies for diffusion and flow-based generative models (see Sections~\ref{section:guid_ddpm} and \ref{section:guidance_flow}), and pre-trained Stable Diffusion natively supports text-based conditional guidance. Nevertheless, natural language exhibits inherent limitations in descriptive granularity, as numerous fine-grained visual details cannot be accurately and comprehensively conveyed through text alone. This issue is further exacerbated by the keyword-centric design of current text prompt paradigms, which prioritize short phrases and lack robust support for syntactically complete, detailed sentences. For example, accurately specifying precise human pose configurations remains challenging via textual descriptions. Accordingly, developing effective guidance mechanisms that incorporate fine-grained conditional information has become a prominent research direction in image generation. A intuitive and effective solution involves leveraging auxiliary visual conditions to regulate the generation process. As demonstrated in Figure~\ref{fig:controlnet_pose}, human pose maps can serve as explicit structural guidance to synthesize images that strictly align with the target pose layout.

\textit{ControlNet} \citep{zhang2023adding}, introduced in this subsection, addresses this limitation by enabling visual conditional control. Built upon the pre-trained Stable Diffusion U-Net backbone, ControlNet introduces an additional visual input modality, including pose maps, line art, hand-drawn sketches, and other structural guidance images. This auxiliary visual signal acts as an explicit conditioning constraint, steering the diffusion generation process to produce outputs that faithfully preserve the structural and semantic features of the input control image.

Stable Diffusion inherently supports conditional guidance, with its core network forward formulation expressed as:
\begin{equation}
\bepsilon^\btheta_t(\bz_t, \bomega^\bphi(\by)),
\end{equation}
where $\bepsilon^\btheta$ denotes the core U-Net-based network of Stable Diffusion at diffusion timestep $t$ (see Algorithm~\ref{alg:diffusion_training} or Section~\ref{section:sd_architecture_unet}). 
The network takes the noisy latent feature $\bz_t$, timestep variable $t$, and conditioning embedding $\bomega^\bphi(\by)$ as inputs. Standard Stable Diffusion realizes text conditioning via a pre-trained CLIP text encoder, such that the conditioning term corresponds to text semantic embeddings. The resulting text-guided formulation is written as:
\begin{equation}
\bepsilon^\btheta_t(\bz_t,  \bomega_{\text{CLIP}}(\text{text prompt})).
\end{equation}

\begin{figure}[h]
\centering
\includegraphics[width=0.88\textwidth]{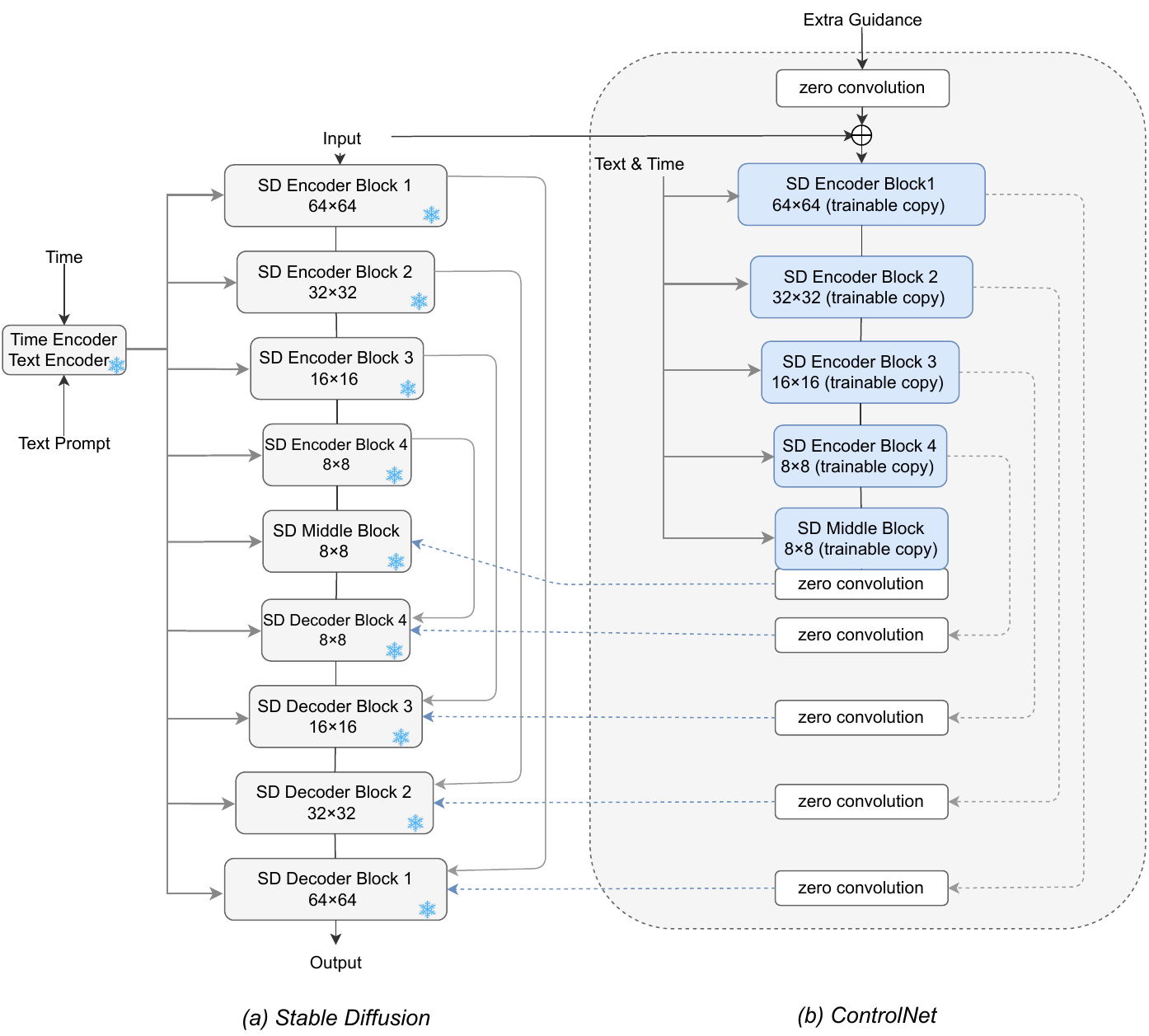}
\caption{Left (a): A standard Stable Diffusion architecture. Right (b): Additional architecture in ControlNet. 
The ``snowflake" symbol indicates that the corresponding component is frozen during training.
Adapted from \citet{zhang2023adding}.}
\label{fig:controlnet_unet}
\end{figure}

\subsection*{Principle of ControlNet}
We further extend the guidance representation $\bomega^\bphi(\by)$ by incorporating additional image control information. Since the introduced guidance control information originates from visual images, image encoding (i.e., feature transformation and extraction) is required for processing. The most intuitive solution is to adopt a dedicated image encoder to handle guidance image data---for instance, directly leveraging the image encoder module from CLIP (see Section~\ref{section:clip}). This paradigm yields the following formulation:
\begin{equation}
\bepsilon^\btheta_t(\bz_t,  \bomega_{\text{CLIP}}(\text{text prompt}), \bomega_{\text{CLIP}}(\text{image prompt}))
\end{equation}
ControlNet deviates from this conventional design. Instead of employing an external standalone image encoder, it repurposes the native U-Net architecture of Stable Diffusion for image encoding.
Figure~\ref{fig:controlnet_unet} (a) visualizes the standard U-Net architecture of Stable Diffusion, where the canonical U-shaped structure is restructured into a linear layout (adapted from Figure~\ref{fig:unet_sd_simplified}). 
Figure~\ref{fig:controlnet_unet} (b) illustrates the parallel ControlNet network appended to this baseline U-Net structure. The ControlNet architecture replicates the encoder and middle blocks on the left side of the original U-Net while fully discarding the decoder module. Each block in the ControlNet branch is augmented with an additional $1 \times 1$  convolutional layer, denoted as a ``zero convolution" in the illustrative diagram.

\begin{figure}[h]
\centering
\includegraphics[width=0.63\textwidth]{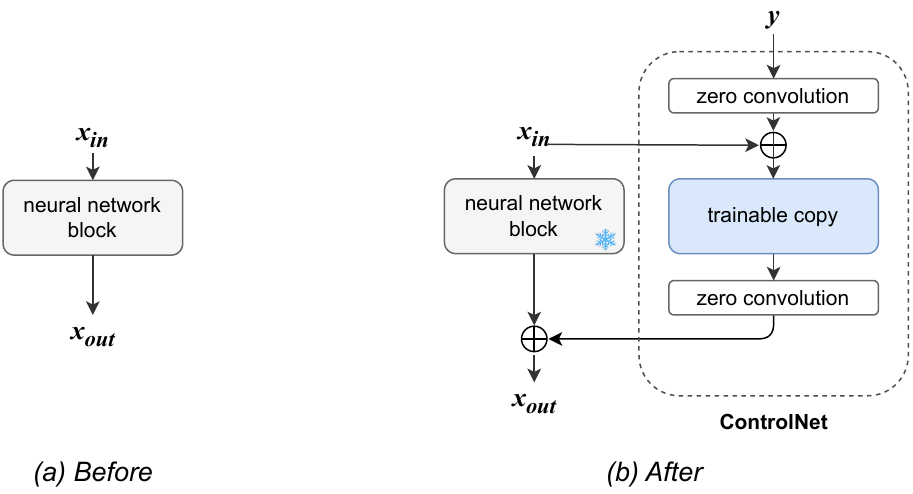}
\caption{Detailed view of a ControlNet block. Left (a): a standard Stable Diffusion block. Right (b): its mirrored copy in ControlNet with an appended $1 \times 1$ convolution. The ``snowflake" symbol indicates that the corresponding component is frozen during training.
Adapted from \citet{zhang2023adding}.}
\label{fig:controlnet_block}
\end{figure}

Figure~\ref{fig:controlnet_block} presents a fine-grained view of a single network block. The left subfigure (a) demonstrates a standard block from the original Stable Diffusion architecture, whereas the right subfigure (b) shows its mirrored counterpart in ControlNet, with $1 \times 1$ zero convolutions applied to both the block's input and output terminals. All ControlNet blocks maintain a one-to-one correspondence with the encoder and middle blocks of the original U-Net.

The output of each ControlNet block is elementwise summed with the output of its corresponding U-Net encoder or middle block. The fused feature is subsequently delivered to the matched U-Net decoder block through skip connections. Notably, this aggregated feature is not propagated to subsequent layers of the original U-Net encoder or middle blocks; it is exclusively transmitted to the corresponding decoder block via skip pathways. This core design ensures that the original structural layout and pre-trained weights of the Stable Diffusion U-Net remain completely intact after ControlNet integration.

In summary, ControlNet eliminates the need for an external image encoder by repurposing the built-in U-Net of Stable Diffusion to encode visual guidance signals. Specifically, it duplicates the pre-trained encoder and middle submodules of the U-Net (excluding the decoder) to construct a dedicated control signal encoding branch. Each block within this duplicated branch is equipped with an additional $1 \times 1$ zero convolution layer, and the output of each ControlNet block is connected to the corresponding decoder layer of the original U-Net via skip connections.

\subsection*{Training Process}
During ControlNet training, the original Stable Diffusion U-Net is fully frozen, with its pre-trained parameters fixed and excluded from optimization. Only the parameters belonging to the ControlNet branch are updated throughout the training process. This training paradigm relies on two critical initialization strategies for stable optimization:
\begin{enumerate}[(i)]
\item The replicated ControlNet blocks derived from the U-Net are initialized with the corresponding pre-trained parameter values of the original Stable Diffusion U-Net. 
\item All $1 \times 1$ convolutional layers within the ControlNet branch are initialized with zero weights and zero biases. 
\end{enumerate}
Owing to the zero initialization of these $1 \times 1$ convolutional layers, all ControlNet blocks produce zero outputs at the initial training stage. In this early state, ControlNet exerts no influence on the network's behavior, and the model performs identically to the vanilla Stable Diffusion model. As training iterations proceed, ControlNet progressively learns to inject precise visual guidance signals into the generation process, while fully preserving the original generative performance of the pre-trained Stable Diffusion model. This carefully designed initialization scheme prevents significant performance degradation of the pre-trained backbone during fine-tuning.

A prevalent concern regarding this design is whether zero-initialized convolutional layers can generate valid gradients and undergo normal parameter updates during backpropagation. This problem can be intuitively explained via a basic linear layer formulation:
\begin{equation}
	x_{\text{out}} = \theta \cdot x_{\text{in}} + b
\end{equation}
The partial derivatives of the output with respect to the trainable parameter $\theta$, input feature $x_{\text{in}}$, and bias term $b$ are derived as follows:
\begin{align*}
	\frac{\partial x_{\text{out}}}{\partial \theta} &= x_{\text{in}}; \\
	\frac{\partial x_{\text{out}}}{\partial x_{\text{in}}} &= \theta; \\
	\frac{\partial x_{\text{out}}}{\partial b} &= 1.
\end{align*}
Notably, initializing the weight parameter $\theta$ to zero does not nullify its corresponding gradient, as the gradient of $\theta$ depends solely on the input feature $x_{\text{in}}$ rather than the weight value itself. Therefore, zero-initialized convolutional layers are capable of computing valid, meaningful gradients and completing standard parameter updates during backpropagation.

With the ControlNet module integrated into the framework, the overall noise prediction network can be formalized as:
\begin{equation}
	\bepsilon^\btheta_t\big(\bz_t, \bomega_{\text{CLIP}}(\text{text prompt}), \bomega_{\text{ControlNet}}(\text{image prompt})\big).
\end{equation}
Based on this augmented noise prediction network, the sampling and image generation pipeline remains consistent with the diffusion sampling procedure illustrated in Algorithm~\ref{alg:diffusion_sampling}.

This core ControlNet design paradigm has been successfully extended to modern flow-matching and rectified-flow-based generative models, including Stable Diffusion v3.x and FLUX. These advanced architectures adopt the identical fundamental principle: attaching a trainable duplicate of network blocks to inject spatial conditional cues such as edges, depth, and human pose. 
Such adaptations differ from the original ControlNet framework, as recent flow-matching models employ transformer-based DiT backbones (see Section~\ref{section:dits}) and flow-matching optimization objectives, replacing the classic U-Net and DDPM diffusion setup for which the original ControlNet was designed. Although the conditioning mechanism is reimplemented to adapt to the structural characteristics of transformer blocks, the core concept of spatial controllability is fully inherited.
Furthermore, alternative conditioning strategies have been proposed to better fit flow-matching and DiT-based models, including in-context conditioning via control token concatenation and lightweight adapter modules \citep{mou2024t2i}. These lightweight approaches are often preferred over full ControlNet block duplication due to their superior computational efficiency.

\subsection{Diffusion Transformers (DiTs)}\label{section:dits}

\begin{figure}[h]
\centering
\includegraphics[width=1\textwidth]{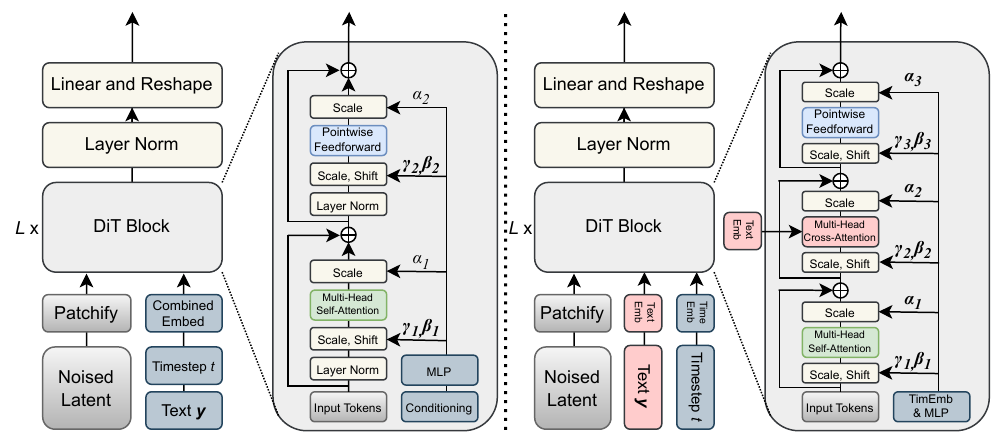}
\caption{An overview of the diffusion transformer architecture. Elementwise addition is denoted by $\oplus$.
The input latent representation  is decomposed into patches and processed by several (here $L$) DiTBlocks. 
Adapted from \citet{peebles2023scalable, chen2024pixartt}. }
\label{fig:dit_all}
\end{figure}

The \textit{diffusion transformer (DiT)} architecture \citep{peebles2023scalable, ma2024sit} 
serves as a powerful alternative to the conventional U-Net backbone and represents a specialized class of neural network designs for generative modeling. While the original DiT framework leveraged class-label conditional guidance, text-conditioned adaptations have become the dominant variant in practical applications. The most prominent advantage of DiT lies in its strong scalability. Extensive experiments demonstrate that the generated image quality improves in a predictable manner with increasing model size and computational resources. This robust scaling capability lays the foundation for state-of-the-art generative models, including OpenAI's Sora and the Stable Diffusion v3.x series \citep{openaicreating, esser2024scaling}.

Diffusion transformers (DiTs) and their variants constitute a dominant family of modern generative architectures, which adopt the self-attention mechanism \citep{vaswani2017attention} as the core network building block \citep{peebles2023scalable, ma2024sit}. Multiple DiT variants have been proposed for diverse tasks; this subsection introduces a generic unified DiT design, while acknowledging that specific model instantiations may vary across different implementations and application scenarios.

Before elaborating on the detailed architectural design, we revisit the basic image representation defined in the introduction section: an image can be formalized as a tensor $\bx \in \real^{C \times H \times W}$, where $C$ denotes the number of image channels (typically $C = 3$ for standard RGB color images), $H$ represents the image height in pixels, and $W$ represents the image width in pixels.
For the remainder of this subsection, we define consistent notation: $D$ denotes the network hidden dimension, $L$ denotes the total number of transformer layers, and $H$ denotes the number of attention heads per layer. DiT architectures are built upon the paradigm of \textit{vision transformers (ViTs)} \citep{dosovitskiy2020image}. The core ViT principle partitions a full image into a set of local patches, embeds these patches into discrete token sequences, and processes the token sequence via standard self-attention operations. A final depatchification module recovers the complete image tensor with the original spatial dimensions.
The initial \textit{patchification} step reorganizes the raw image tensor $\bx \in \real^{C \times H \times W}$ into a patch sequence:
$$ 
\texttt{Patchify}(\bx) \in \real^{N \times C'} ,
$$
where $C' = CP^2, N = (H/P) \cdot (W/P)$, with $P$ specifying the patch size. 
The patchified features are subsequently projected via a linear transformation and integrated with learnable positional embeddings $``\text{PosEmb}"\in\real^{N\times D}$. Notably, the original DiT implementation employs fixed, frozen 2D sinusoidal positional embeddings instead of learnable ones. The final patch embedding is formulated as:
\begin{equation}\label{equation:patch_emb_ND}
\texttt{PatchEmb}(\bx) 
= \underbrace{\texttt{Patchify}(\bx) \cdot \bW}_{\texttt{Linear}(\bx)} 
+ \text{PosEmb}
\in \real^{N \times D} ,
\end{equation}
where $\bW \in \real^{C' \times D}$ is a trainable weight matrix. 
The output $\texttt{PatchEmb}(\bx)$ yields a \textit{continuous tokenization} of the original input, consisting of a sequence of continuous latent feature tokens.
Nearly all modern large-scale image and video generative models---including latent diffusion models (LDMs, introduced in Section~\ref{section:ldm_intro}) and flow-based generative variants---operate entirely in the latent space. For such frameworks, patch embedding is performed on compressed latent representations rather than raw pixel images. Specifically, a variational autoencoder (VAE) encoder first compresses the original image $\bx\in\real^{3\times H\times W}$ into a low-dimensional latent tensor; for instance, the Stable Diffusion v1.x series maps raw images to latent features $\bz\in\real^{4\times H/8\times W/8}$. Patch embedding is then applied to this compressed latent tensor. 
In this book, across all DiT variants, the output of the patch embedding module maintains a unified shape of $N\times D$, where $N$ denotes the total number of spatial tokens and $D$ denotes the (transformer) hidden dimension size.

DiT takes three core conditional inputs: time embeddings, text prompt embeddings  (detailed in Section~\ref{section:emb_guidance_var}), and patchified latent image features, defined as follows:
\begin{align*}
\widetildebt &= \texttt{TimeEmb}(t) \in \real^D;\\
\widetildeby &= \texttt{TextEmb}_{\text{SEQ}}(\by) \in \textcolor{black}{\real^{M \times D}};\\
\widetildebz_0 &= \texttt{PatchEmb}(\bx) \in \real^{N \times D}.
\end{align*}
All input modalities are projected to the consistent hidden dimension $D$, matching the transformer network dimension requirement. The DiT network iteratively refines the latent token features $\widetildebz_l$ through $L$ stacked transformer-based DiTBlocks for $l = 0, 1, \ldots, L-1$ (see Equations \eqref{equation:ditblock_form1} and \eqref{equation:ditblock_form2} for detailed DiTBlock formulations):
\begin{equation}
\widetildebz_{l+1} = \texttt{DiTBlock}(\widetildebz_l, \widetildebt, \widetildeby) \in \real^{N \times D}, \quad l = 0,1, \ldots, L-1.
\end{equation}
Following the hierarchical feature refinement via all $L$ DiT layers, a final \textit{depatchification} operation converts the refined latent token sequence back to the original image/latent tensor dimension:
$$ 
\bp = \texttt{Depatchify}(\widetildebz_N \widetildebW) \in \real^{C \times H \times W}, 
$$
where $\widetildebW \in \real^{D \times C'}$ denotes the final projection weight matrix. 
The output tensor $\bp$ corresponds to the model's final prediction, which serves as either the predicted velocity field $\vf_t^\btheta(\bz\mid\by)$ or the predicted noise residual $\bepsilon_t^\btheta(\bz\mid\by)$ for the generative diffusion/flow process conditioned on the input prompt $\by$.

\subsubsection{One DiT Block}
For completeness, this subsection elaborates on the mathematical formulation of a single DiT block. While we provide sufficient technical details to facilitate a comprehensive understanding of the DiT model family, we prioritize core algorithmic designs over trivial architectural intricacies. Let $\bz \in \real^{N \times D}$ denote the latent patch token sequence at the current layer (i.e., $\bz = \widetildebz_l$) and $\by \in \textcolor{black}{\real^{M \times D}}$ denote the embedded conditional guidance feature (i.e., $\by = \widetildeby$). A standard DiT block updates the latent token sequence $\bz$ via three core operations: (i) patch self-attention, (ii) prompt/time-guided cross-attention, and (iii) time/text conditioning implemented via adaptive layer normalization (AdaLN).

\paragrapharrow{DiTBlock Form 1: Guidance embedding adaptive normalization.}
All conditional guidance signals, including the diffusion timestep $t$ and text/label embeddings $\by$, are fused and projected into a unified conditioning vector $\widetildebc \in \real^{D}$ via a dedicated embedding function:
\begin{equation}
\widetildebc = \texttt{CombinedEmbed}(t, \by) \in \real^{D}.
\end{equation}
This design creates a key distinction from the U-Net backbone adopted in Stable Diffusion v1.x models: DiT unifies timestep and textual/label conditions into a single compact conditioning vector $\widetildebc$, rather than processing them separately. This unified conditioning vector is further fed into a multilayer perceptron (MLP) to generate \textit{AdaLN-Zero modulation} parameters, which govern the adaptive normalization process:
$$
(\bgamma_1, \bbeta_1, \alpha_1, \bgamma_2, \bbeta_2, \alpha_2) = \texttt{MLP}(\widetildebc),
$$
where $ \bgamma_i, \bbeta_i \in \real^D $ (scale/shift) and $ \alpha_i \in \real $ (residual scaling).
Given the DiTBlock input latent tokens $\bz \in \real^{N \times D}$, the block executes sequential self-attention and pointwise feed-forward submodules with adaptive normalization modulation, as formalized below:
\begin{subequations}\label{equation:ditblock_form1}
\begin{align}
\texttt{AdaNorm1}_{\widetildebc}(\bz) &= \bgamma_1 \hadaprod \texttt{LayerNorm}(\bz) + \bbeta_1 ;
\quad\quad\quad\quad\quad (\text{Scale \& Shift}) \\
\widehatbx &= \bz + \alpha_1 \hadaprod \texttt{MultiHeadAttn}(\texttt{AdaNorm1}_{\widetildebc}(\bz), \texttt{AdaNorm1}_{\widetildebc}(\bz)) ;\\
\texttt{AdaNorm2}_{\widetildebc}(\bz) &= \bgamma_2 \hadaprod \texttt{LayerNorm}(\widehatbx) + \bbeta_2 ;
\quad\quad\quad\quad\quad (\text{Scale \& Shift}) \\
\bz_{\text{out}} &= \widehatbx + \alpha_2 \hadaprod\texttt{PointwiseFeedForward}(\texttt{AdaNorm2}_{\widetildebc}(\bz)),
\end{align}
\end{subequations}
where $\hadaprod$ denotes elementwise multiplication with automatic broadcasting along the token dimension $N$. The structural workflow corresponding to this formulation is visualized in the left portion of Figure~\ref{fig:dit_all}. In this paradigm, positional embeddings preserve the spatial order of patch tokens, while self-attention mechanisms model long-range global spatial dependencies across the entire latent feature map.

\paragrapharrow{Contrast with SD U-Net.}
A fundamental architectural distinction lies in how conditional signals are integrated and how spatial features are processed. DiT embeds all conditional information into adaptive normalization applied to a single flat token sequence. In comparison, the U-Net backbone adopted by standard Stable Diffusion (SD) models employs two distinct conditioning pathways and operates on a convolutional pyramid structure instead of a uniform token grid. Specifically, SD U-Net applies additive timestep modulation within residual blocks and leverages cross-attention layers to inject textual prompt conditions, separating temporal and linguistic guidance signals.
The core structural and functional differences between DiT (Form 1) and the conventional SD v1.x U-Net are systematically summarized in Table~\ref{tab:diff_sdunet_dit}.
\begin{center}
\setlength{\tabcolsep}{1.3pt}    
\begin{tabular}{lll}
\toprule
\textbf{Aspect} & \textbf{DiT (Form 1)} & \textbf{SD v1.x U-Net} \\
\midrule
Cond. fusion & $t$ and $\by$ merged into one $\widetildebc$
& kept separate ($\widetildebt$ vs $\widetildeby$) \\
Time mechanism  & adaLN (scale/shift/gate from $\widetildebc$)
& additive embedding inside res blocks \\
Text mechanism      & also folded into adaLN $\widetildebc$
& cross-attention (keys/values from $\widetildeby$) \\
Spatial structure   & flat token sequence+PosEmb
& multi-resolution conv hierarchy+skips \\
Normalization       & adaptive LayerNorm (modulated)
& GroupNorm ($t$ added separately) \\
\bottomrule
\end{tabular}
\label{tab:diff_sdunet_dit}
\end{center}

\index{CrossDiT}
\paragrapharrow{DiTBlock Form 2: Time-guided  adaptive normalization.} 
The original DiT architecture implements all conditional injections (timestep, class label, and text conditions) exclusively through the AdaLN-Zero modulation mechanism, without incorporating cross-attention modules. In contrast, advanced DiT variants decouple conditional modulation functionalities: cross-attention replaces partial AdaLN-based text conditioning, enabling fine-grained, token-wise feature modulation guided by textual semantic information. This refined block design---which arranges self-attention, text-driven cross-attention, and MLP layers in sequence within each block and leverages cross-attention for textual conditioning---is formally defined as \textit{CrossDiT} in existing literature, first employed in the PixArt-$\alpha$ model \citep{chen2024pixartt}.

Let $\widetildebt \in \real^D$ denote the  timestep embedding. A standard design paradigm across DiT architectures leverages $\widetildebt$ to generate channel-wise scale and shift parameters for modulating normalized feature activations \citep{perez2018film}. Concretely, we define an MLP function $\bg: \real^D \to \real^{2D}$ that outputs modulation parameters conditioned solely on the timestep embedding:
$$ 
(\bgamma, \bbeta) = \bg(\widetildebt), 
$$
where $\bgamma, \bbeta \in \real^D$. 
In practical implementations, independent $(\bgamma, \bbeta)$ parameter pairs may be allocated for different sublayers (i.e., attention and MLP modules) to support layer-specific modulation; see the right portion of Figure~\ref{fig:dit_all}. 
Given a patch token matrix $\bz \in \real^{N \times D}$ and a normalization operator $\texttt{Norm}(\cdot)$ (e.g., {\texttt{LayerNorm}}),  the time-aware adaptive normalization is formulated as:
\begin{equation}
\texttt{AdaNorm}_{\widetildebt}(\bz) = (1 + \bgamma) \hadaprod \texttt{LayerNorm}(\bz) + \bbeta,
\end{equation}
where $\hadaprod$ denotes elementwise multiplication with automatic  broadcasting along  the token dimension ($N$).
This time-dependent adaptive normalization dynamically rescales and shifts normalized features according to timestep information, allowing the network to adapt its behavior to varying noise levels throughout the generative process.

The complete computational pipeline of the improved DiTBlock (CrossDiT) is defined as follows:
\begin{subequations}\label{equation:ditblock_form2}
\begin{align}
\bz &\leftarrow \bz + \bg_{\text{self}}(\widetildebt) \hadaprod \texttt{MultiHeadAttn}\big(\texttt{AdaNorm}_{\widetildebt}(\bz), \texttt{AdaNorm}_{\widetildebt}(\bz)\big); \\
\bz &\leftarrow \bz + \bg_{\text{Cross}}(\widetildebt) \hadaprod \texttt{MultiHeadAttn}\big(\texttt{AdaNorm}_{\widetildebt}(\bz), \by\big) ;\\
\bz_{\text{out}} &= \bz + \bg_{\text{MLP}}(\widetildebt) \hadaprod \texttt{PointwiseFeedForward}\big(\texttt{AdaNorm}_{\widetildebt}(\bz)\big),
\end{align}
\end{subequations}
where $\bg_{\text{self}}(\widetildebt)$, $\bg_{\text{Cross}}(\widetildebt)$, and $\bg_{\text{MLP}}(\widetildebt)\in\real^D$ represent learnable timestep-dependent modulation parameters that control the contribution strength of each sublayer. 
The self-attention module models intrinsic pairwise interactions among spatial patch tokens, while the cross-attention module injects external textual conditional information $\by$ into the token sequence. The overall block structure is illustrated in the right portion of Figure~\ref{fig:dit_all}.
All gating parameters $\bg_{\ldots}(\widetildebt)$ are initialized to zero, a core design of AdaLN-Zero that stabilizes early-stage model training. Meanwhile, stacked residual connections preserve smooth gradient propagation and mitigate gradient vanishing issues, which is essential for training deep transformer architectures \citep{he2016deep}. 

Further lightweight DiT variants have also been developed to reduce computational overhead. Instead of deploying an independent full MLP for each block to generate modulation parameters, these designs employ a single shared global MLP that computes base $\bgamma$ and $\bbeta$ parameters from the timestep embedding. Each DiT block is then equipped with a small learnable offset vector to fine-tune the shared modulation parameters block-wisely. This strategy retains the full expressive capability of block-specific modulation while drastically reducing the total number of trainable parameters. The detailed implementation of these lightweight variants is omitted here for brevity.

The output feature $\bz_{\text{out}}$ of the $l$-th DiTBlock serves as the input token sequence for the subsequent layer, denoted as $\widetildebz_{l+1}\in \real^{N \times D}$ in our notation. After processing through $L$ sequentially stacked DiTBlocks, the final refined latent token sequence is obtained as:
$$
\widetildebz_{L} = \texttt{DiTBlocks}(\widetildebz_0, \widetildebt, \widetildeby) \in \real^{N \times D}.
$$

\begin{figure}[h]
\centering
\includegraphics[width=0.8\textwidth]{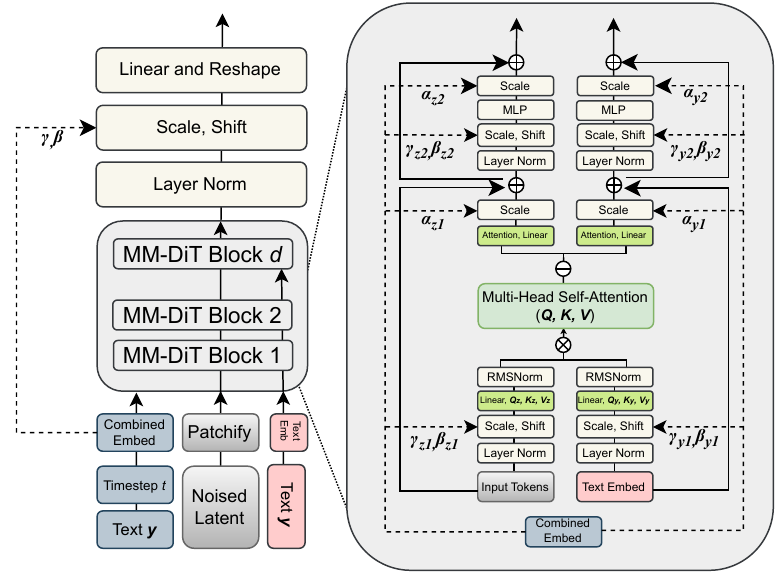}
\caption{An overview of the multimodal diffusion transformer (MM-DiT) architecture. 
Concatenation is denoted by $\otimes$; split is indicated by $\ominus$; and elementwise addition is denoted by $\oplus$.
The input latent is decomposed into patches and processed by several (here $d$) MM-DiT Blocks. 
Adapted from \citet{esser2024scaling}. }
\label{fig:dit_sd3}
\end{figure}

\index{Multimodal diffusion transformers (MM-DiTs)}
\subsection{Multimodal Diffusion Transformers (MM-DiTs)}

Similar to the vanilla DiT design, the \textit{Stable Diffusion v3.x series (SD v3.x series)} adopts a \textit{multimodal DiT (MM-DiT)} backbone, which features disjoint image and text feature streams, \textit{joint self-attention mechanisms}, and AdaLN-Zero modules for unified timestep and textual conditioning \citep{esser2024scaling}. 
The overall MM-DiT architectural design is visualized in Figure~\ref{fig:dit_sd3}.

\subsection*{Overview and Conditioning Inputs}

\paragrapharrow{Pooled (global) conditioning vector $\widetildeby$.}
Global text representations are derived from pooled CLIP outputs. Unlike per-token sequence embeddings that preserve individual word-level features, pooled CLIP embeddings compress the entire text caption into a single fixed-length global semantic vector. The final global conditioning representation is constructed via concatenation of two CLIP pooled features in SD v3.x series:
\begin{equation}
\widetildeby = \texttt{Concat}(\by_{\text{CLIP-G}},\, \by_{\text{CLIP-L}}) .
\end{equation}
This global vector encodes only coarse-grained textual semantic information for high-level conditional guidance.


\paragrapharrow{Combined modulation vector $\widetildebc$.}
To generate the block-wise adaptive modulation signal, both  timestep information and global text conditions are embedded and fused. The timestep $t$ is first encoded via sinusoidal embedding and processed by a dedicated MLP, while the pooled global text vector $\widetildeby$ is independently projected through another MLP. The two feature streams are summed to form the final conditioning vector:
\begin{equation}\label{equation:mmdit_pooled_y}
\widetildebc = \texttt{MLP}(\textcolor{black}{\texttt{TimeEmb}(t)}) + \texttt{MLP}(\widetildeby).
\end{equation}
This fused conditioning vector $\widetildebc$ serves as the core modulation signal that drives the AdaLN-based adaptive normalization mechanism within every MM-DiT block.

\paragrapharrow{Data (image) stream $\widehatbz$.}
Similar to \eqref{equation:patch_emb_ND}, 
the noised latent feature $\bz_t \in \real^{H\times W\times C}$ is first partitioned into $2\times 2$ patches and flattened into a token sequence with a length of $\frac{1}{2} H \cdot \frac{1}{2} W$. The resulting patch sequence undergoes linear projection and is augmented with positional embeddings:
\begin{equation}
\widehatbz = \operatorname{Patchify}(\bz_t)\,\bW_{\text{patch}} + \text{PosEmb}\in \real^{N \times D}.
\end{equation}

\paragrapharrow{Text stream $\widehatby$.}
The fine-grained textual sequence embeddings are linearly projected to match the transformer hidden dimension:
\begin{equation}
\widehatby = \texttt{TextEmb}_{\text{SEQ}}(\by)\bW_{\text{ctxt}} \in \real^{M \times D}.
\end{equation}
Here, the network hidden dimension is defined as $D = 64\cdot d$, where $d$ denotes the model depth (i.e., the number of MM-DiT blocks). The number of self-attention heads is consistently set equal to $d$. It is worth noting that minor implementation variations may exist across different practical MM-DiT deployments, while the core architectural and conditioning paradigms remain identical.

\subsection*{One MM-DiT Block}

Each MM-DiT block consists of \textbf{two parallel, parameter-independent transformer streams} dedicated to the text token sequence $\widehatby$ and image latent token sequence $\widehatbz$, respectively. Although the two streams maintain separate learnable parameters, their feature representations are fused via a unified shared attention operation.

\paragrapharrow{Modulation parameters.}
From the fused conditioning vector $\widetildebc$, a sequential SiLU activation and linear layer generate a complete set of adaptive modulation parameters for each individual stream. We use the subscript $y$ to denote text-stream parameters and $z$ to denote image-stream parameters:
\begin{align}
(\bgamma_{y1}, \bbeta_{y1}, \balpha_{y1}, \bgamma_{y2}, \bbeta_{y2}, \balpha_{y2})
&= \texttt{Linear}(\textcolor{black}{\texttt{SiLU}}(\widetildebc)), \\
(\bgamma_{z1}, \bbeta_{z1}, \balpha_{z1}, \bgamma_{z2}, \bbeta_{z2}, \balpha_{z2})
&= \texttt{Linear}(\texttt{SiLU}(\widetildebc)).
\end{align}
These two groups of parameters independently regulate the adaptive normalization and residual scaling behaviors of the text and image streams.

\index{QK-normalization}
\paragrapharrow{Pre-attention: modulated LayerNorm + QKV projection.}
For each feature stream $\bmm \in \{\widehatby, \widehatbz\}$, the input feature is first normalized and adaptively affine-modulated using the corresponding scale parameter $\bgamma$ and shift parameter $\bbeta$:
\begin{subequations}\label{equation:mmdit_preatten}
\begin{equation}
	\widetildebm = \bgamma_{m1} \hadaprod \texttt{LayerNorm}(\bmm) + \bbeta_{m1}.
\end{equation}
The modulated features are then linearly projected to produce stream-specific query, key, and value matrices:
\begin{equation}
	(\bQ_{m}, \bK_{m}, \bV_{m}) = \widetildebm\,\bW_{\text{qkv}}^{(m)}.
\end{equation}
Optionally, RMS normalization is applied to the query and key matrices to stabilize training dynamics (i.e., QK-normalization~\eqref{equation:qknormalization}):
\begin{equation}
	\bQ_{m} \leftarrow \operatorname{RMSNorm}(\bQ_{m}), \qquad
	\bK_{m} \leftarrow \operatorname{RMSNorm}(\bK_{m}).
\end{equation}

\end{subequations}

\index{Joint-attention}
\paragrapharrow{Joint-attention (the key fusion step).}
A single self-attention operation (detailed in \eqref{equation:muthead_attn}) is performed on the concatenated sequence, defining the \textit{joint-attention mechanism}. 
This design enables every text token to interact with every image token and vice versa, establishing fully \textit{bidirectional cross-modal information flow}:
\begin{equation}
\bQ = [\bQ_y\,;\,\bQ_z], \qquad \bK = [\bK_y\,;\,\bK_z], \qquad \bV = [\bV_y\,;\,\bV_z].
\end{equation}
A single self-attention (see \eqref{equation:muthead_attn} for details) runs over the joined sequence (hence the name \textit{joint-attention} mechanism), so every text token attends to
every image token and vice versa---a genuinely \textit{bidirectional} flow of information:
\begin{equation}
\bZ = \texttt{softmax}\left(\frac{\bQ \bK^\top}{\sqrt{D_H}}\right)\bV,
\end{equation}
where $D_H$ denotes the per-head dimension. The equation is simplified with single-head attention for illustration; all practical implementations adopt standard multi-head attention, as referenced in \eqref{equation:muthead_attn}. Following global attention computation, the unified output sequence is split back into disjoint text and image feature branches:
\begin{equation}
\bZ = [\bZ_y\,;\,\bZ_z].
\end{equation}
This joint-attention paradigm essentially integrates the functionalities of text-image cross-attention and intra-modal self-attention within one unified module.

\paragrapharrow{Post-attention residual (per stream).}
Each individual stream processes its split attention output independently. The attention features are projected through a stream-specific output layer, scaled by the corresponding residual gating factor $\balpha_{m1}$, and fused with the original input via a residual connection:
\begin{equation}
\bmm' = \bmm + \balpha_{m1} \hadaprod \big(\bZ_{m}\,\bW_{\text{out}}^{(m)}\big).
\end{equation}

\paragrapharrow{Feed-forward sub-block (per stream).}
A second round of adaptive layer normalization is applied to the post-attention features, governed by the second group of stream-specific modulation parameters. The normalized features are fed into a pointwise MLP, and the output is scaled by $\balpha_{m2}$ before residual addition:
\begin{equation}
\bmm'' = \bmm' + \balpha_{m2} \hadaprod
\texttt{MLP}\!\big(\bgamma_{m2} \hadaprod \texttt{LayerNorm}(\bmm') + \bbeta_{m2}\big).
\end{equation}
The final updated feature $\bmm''$ serves as the input representation for the subsequent MM-DiT block. Notably, both text and image streams are iteratively updated throughout the network depth. Different from conventional fixed text conditioning paradigms, the textual feature stream evolves dynamically alongside the image latent stream, enabling progressive cross-modal alignment and refinement.

\subsection*{Stacking and Output}
The aforementioned MM-DiT block is stacked sequentially $d$ times to form the complete backbone network. After the final block iteration, only the refined image data stream is retained for generative prediction. A final modulation operation, linear projection, and unpatching process are applied to map the latent token sequence back to the structured latent feature grid:
\begin{equation}
\bp_t^\btheta(\bz_t\mid \by) =
\texttt{Unpatchify}\big(\texttt{Linear}(\texttt{Modulation}(\widehatbz_d))\big).
\end{equation}
This network output parameterizes either the velocity field $\vf_t^\btheta$ of the conditional flow-matching ODE or the predicted noise residual $\bepsilon_t^\btheta$ for standard DDPM-based diffusion modeling.

The core innovation of the MM-DiT architecture lies in its dual-stream design paradigm. Conceptually, MM-DiT instantiates two modality-independent transformer branches for image and text features, respectively. The two modalities maintain isolated parameter spaces and feature representations throughout feed-forward propagation, while their token sequences are concatenated and fused exclusively within the joint-attention module. This design allows each modality to preserve its intrinsic feature characteristics while enabling comprehensive bidirectional cross-modal interaction.

This mechanism fundamentally differs from conventional text-to-image generative models, which inject static, fixed text embeddings via one-way cross-attention. In contrast, MM-DiT jointly optimizes and updates text and image representations across all stacked layers. This dynamic co-evolution of textual and visual features is credited for the model’s superior text alignment capability and enhanced typographic generation quality \citep{esser2024scaling}.

Two key practical trade-offs accompany this design. First, the dual independent parameter streams nearly double the per-block parameter count compared to conventional single-stream transformers. Second, joint attention computation exhibits quadratic complexity with respect to the total number of concatenated text and image tokens $(M + N)$, meaning longer text sequences directly increase computational overhead during attention inference and training.

\subsubsection{Case Study: Stable Diffusion 3}

Stable Diffusion v3.x (SD v3.x) adopts the conditional flow-matching objective formalized in {Algorithms~\ref{alg:flow_matching} and \ref{alg:score_gauss_path}} \citep{esser2024scaling}.
Extensive ablation studies in the original work demonstrate that flow matching outperforms alternative diffusion  generative objectives. During training, the model leverages classifier-free guidance, as elaborated in Sections~\ref{section:cfg_ddpm} and \ref{section:cfg_flow}. Consistent with the SD v1.x and v2.x series, SD v3.x follows the latent generative paradigm introduced in Section~\ref{section:ldm_intro} and illustrated in Figure~\ref{fig:SD_three_comps}, performing all training and generation within the compressed latent space of a pretrained autoencoder. The development of high-quality perceptual autoencoders constituted a core technical contribution of the original Stable Diffusion framework.

To strengthen textual conditional alignment, SD v3.x integrates multiple complementary text embedding modalities, including CLIP textual features \citep{radford2021learning} and sequential embeddings extracted from the pretrained T5-XXL text encoder developed by Google \citep{raffel2020exploring}. This multi-encoder embedding strategy aligns with prior designs proposed in existing text-to-image generative models \citep{balaji2022ediff, saharia2022photorealistic}.
Specifically, two CLIP variants, CLIP-ViT-G/14 and CLIP-ViT-L/14 \citep{cherti2023reproducible, radford2021learning}, provide both global pooled textual vectors for the modulation formulation in \eqref{equation:mmdit_pooled_y} and fine-grained per-token sequence features for the joint-attention computation in \eqref{equation:mmdit_preatten}. The T5-XXL encoder further supplements additional high-resolution sequential text tokens to enrich the conditional context in the attention module.

This design represents a clear paradigm shift from single text encoder pipelines toward a multi-model ``committee of experts" embedding strategy. Early Stable Diffusion variants (SD v1.x) relied solely on CLIP embeddings, which excel at capturing high-level visual concepts and stylistic attributes but struggle with complex spatial composition and detailed textual rendering. SDXL, which incorporates dual CLIP encoders, improves textual nuance understanding yet retains limitations in structural logical reasoning \citep{podell2024sdxl}.
The hybrid CLIP+T5 embedding scheme adopted by SD v3.x (and FLUX.1) marks a substantial advancement in conditional modeling \citep{radford2021learning, labs2025flux}. In this collaborative framework, CLIP encoders define the core visual concepts and stylistic characteristics of the generated content, answering what to generate. In contrast, the T5 encoder captures linguistic syntax, spatial constraints, and textual spelling rules, governing how visual content is structurally organized. This functional division compensates for the inherent deficiencies of individual encoders, collectively achieving superior prompt alignment accuracy and high-fidelity typographic generation.

CLIP embeddings deliver coarse, global semantic summaries of input prompts, while T5-XXL sequential embeddings provide dense, fine-grained contextual information that enables the model to attend to precise substructures within textual instructions. To fully exploit sequential textual conditioning, SD v3.x extends the standard DiT architecture to support bidirectional cross-modal attention between image patch tokens and text sequence tokens. This upgrade evolves the original class-conditional DiT design into the sequential context-aware MM-DiT architecture detailed above.
The largest SD v3.x model comprises 8 billion parameters. For inference, generation adopts 50 sampling steps with an Euler ODE solver, and the classifier-free guidance weight is configured within the range of 2.0 to 5.0 to balance generation fidelity and diversity.

Beyond the aforementioned architectural designs, SD v3.x introduces two critical upgrades over prior Stable Diffusion series:
First, the pretrained VAE employed in SD v3.x outputs a 16-dimensional latent space, a substantial upgrade from the 4-channel latent representation used in SD v1.x and SD v2.x. The 16-channel VAE adopted by SD v3.x reduces compression intensity, effectively preserving fine visual details including facial expressions, intricate textures, and small textual patterns within latent features, thereby significantly improving fine-detail generation quality.
Second, native resolution capabilities are progressively enhanced across generations: SD v1.x natively generates 512$\times$512 pixel images, and SD v2.x supports 768$\times$768 pixel outputs. By comparison, SD v3.x features a native resolution of 1024$\times$1024 pixels and can stably generate images up to 2048$\times$2048 pixels without prominent upsampling artifacts.

\begin{problemset}
\item \label{prob:global_clss_cross}\textbf{Degenerate attention mechanism.} 
Analyze the behavior of Equation \eqref{equation:unet_crossattn} when the guidance tensor $\widetildeby\in\real^{M\times D_\tau}$ has $M=1$. In this setting, the guidance signal is non-spatial and uniform across all spatial locations of the feature map.
\textit{Hint: Derive the resulting attention scores and discuss their structural shape and implications.}

\item \label{prob:imp_arcendec} \textbf{Impact of encoder and network size \citep{ramesh2022hierarchical}.} 
Train a standard DDPM model on a publicly available dataset such as MNIST or ImageNet. Systematically increase the size of the text encoder (CLIP) and the overall network architecture (e.g., U-Net capacity). Evaluate and compare model performance using the \textit{Fréchet Inception Distance (FID)} (\textit{FID is a standard evaluation metric for generative models such as GANs and diffusion models. It quantifies the distributional discrepancy between real and generated images by comparing their latent feature statistics extracted from a pretrained network; lower FID values indicate higher generation quality \citep{heusel2017gans}})
and \textit{CLIP score} (\textit{the CLIP score evaluates text-image alignment by embedding generated images and their corresponding text prompts into a shared pretrained CLIP embedding space and computing their cosine similarity; higher CLIP scores indicate better text-image consistency \citep{radford2021learning}}).
Report and interpret your observations across the two model scaling configurations.


\end{problemset}

\newpage
\vskip 0.2in
\addcontentsline{toc}{chapter}{Bibliography}
\bibliography{bib}

\clearpage

\printindex
\addcontentsline{toc}{chapter}{Index}

\end{document}